\documentclass[runningheads]{llncs}

\usepackage{accv}

\usepackage{accvabbrv}
\usepackage[percent]{overpic}

\usepackage{graphicx}
\usepackage{booktabs}

\usepackage[accsupp]{axessibility}  

\usepackage{wasysym}    
\newcommand\BigSquare{\raisebox{-0.3mm}{\scalebox{1.5}{$\Box$}}}
\newcommand\BigPentagon{\raisebox{-0.3mm}{\scalebox{1.5}{\pentagon}}}
\newcommand\BigCircle{\raisebox{-0.3mm}{\scalebox{1.4}{\Circle}}}
\newcommand\BigRedCircle{\raisebox{-0.3mm}{\scalebox{1.4}{\textcolor{red}{\Circle}}}}

\newcommand{\greentick}{\textcolor{ForestGreen}{\cmark}\xspace}
\newcommand{\redxmark}{\textcolor{red}{\xmark}\xspace}

\newcommand{\na}{\textcolor{gray}{n/a}\xspace}
\definecolor{MyLightGray}{rgb}{0.95, 0.95, 0.95}
\definecolor{MyLightYellow}{rgb}{0.98, 0.97, 0.91}

\definecolor{green83}{RGB}{77,177,93} 
\definecolor{green88}{RGB}{36,157,83} 
\definecolor{green91}{RGB}{22,147,76} 

\newcommand{\model}[1]{{\texttt{#1}\xspace}}
\definecolor{gpt_green}{RGB}{22,163,127} 
\definecolor{gemini_blue}{RGB}{81,134,209} 
\definecolor{sonnet3_brown}{RGB}{204,154,123} 
\definecolor{sonnet35_brown}{RGB}{216, 119, 87}

\renewcommand{\gemini}{\model{Gemini-\textcolor{gemini_blue}{1.5}}\xspace}
\newcommand{\geminifull}{\model{Gemini-\textcolor{gemini_blue}{1.5} Pro}\xspace}
\newcommand{\geminilogo}{{\includegraphics[scale=0.7]{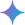}}\xspace}

\newcommand{\sonnet}{\model{Sonnet-\textcolor{sonnet3_brown}{3}}\xspace}
\newcommand{\sonnetfull}{\model{Claude-\textcolor{sonnet3_brown}{3} Sonnet}\xspace}
\newcommand{\sonnetlogo}{\includegraphics[scale=0.02]{{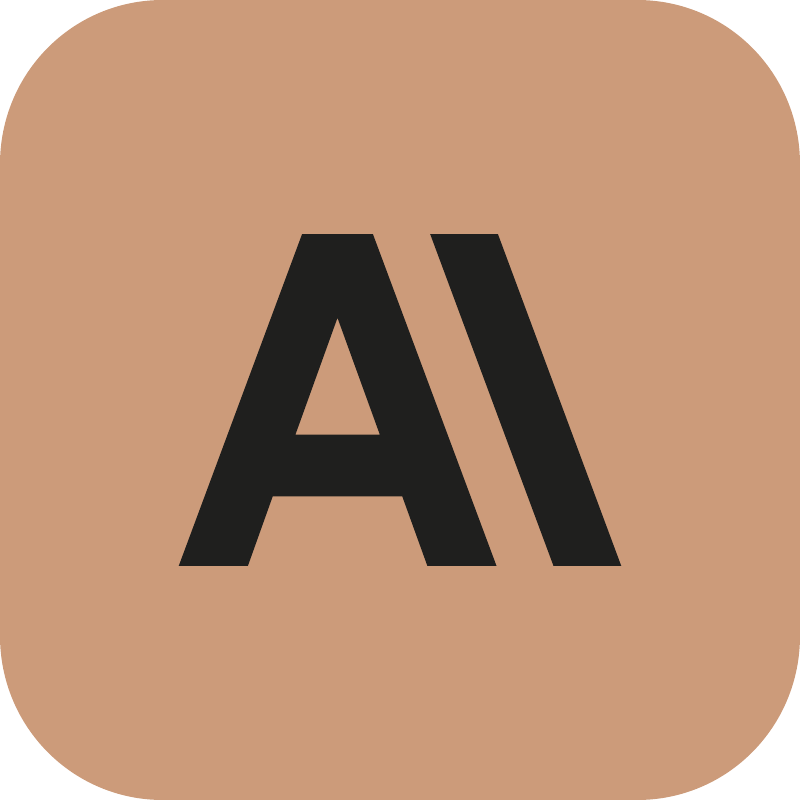}}\xspace}

\newcommand{\geminithinkfull}{\model{Gemini \textcolor{gemini_blue}{2.0} Flash-Thinking}\xspace}

\newcommand{\geminiflashfull}{\model{Gemini \textcolor{gemini_blue}{2.0} Flash}\xspace}

\newcommand{\qvq}{\model{QVQ}\xspace}
\newcommand{\qvqfull}{\model{QVQ-Preview}\xspace}
\newcommand{\qwen}{\model{Qwen2-VL}\xspace}

\newcommand{\gpt}{\model{GPT-\textcolor{gpt_green}{4o}}\xspace}
\newcommand{\gptlogo}{{\includegraphics[scale=0.02]{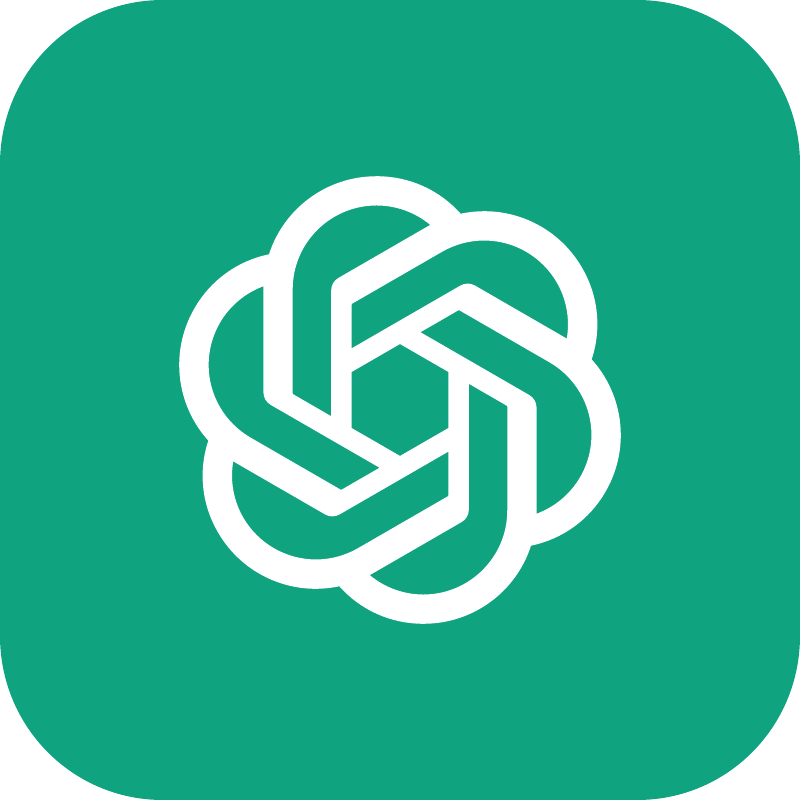}}\xspace}

\usepackage{xspace}    
\newcommand{\llavaone}{\model{LLaVA-OneV}\xspace}
\newcommand{\llavaonefull}{\model{LLaVA OneVision-qwen2}\xspace}
\newcommand{\llavaSI}{\model{LLaVA-OneV-S}\xspace}

\newcommand{\llavalogolarge}{{\includegraphics[scale=0.01]{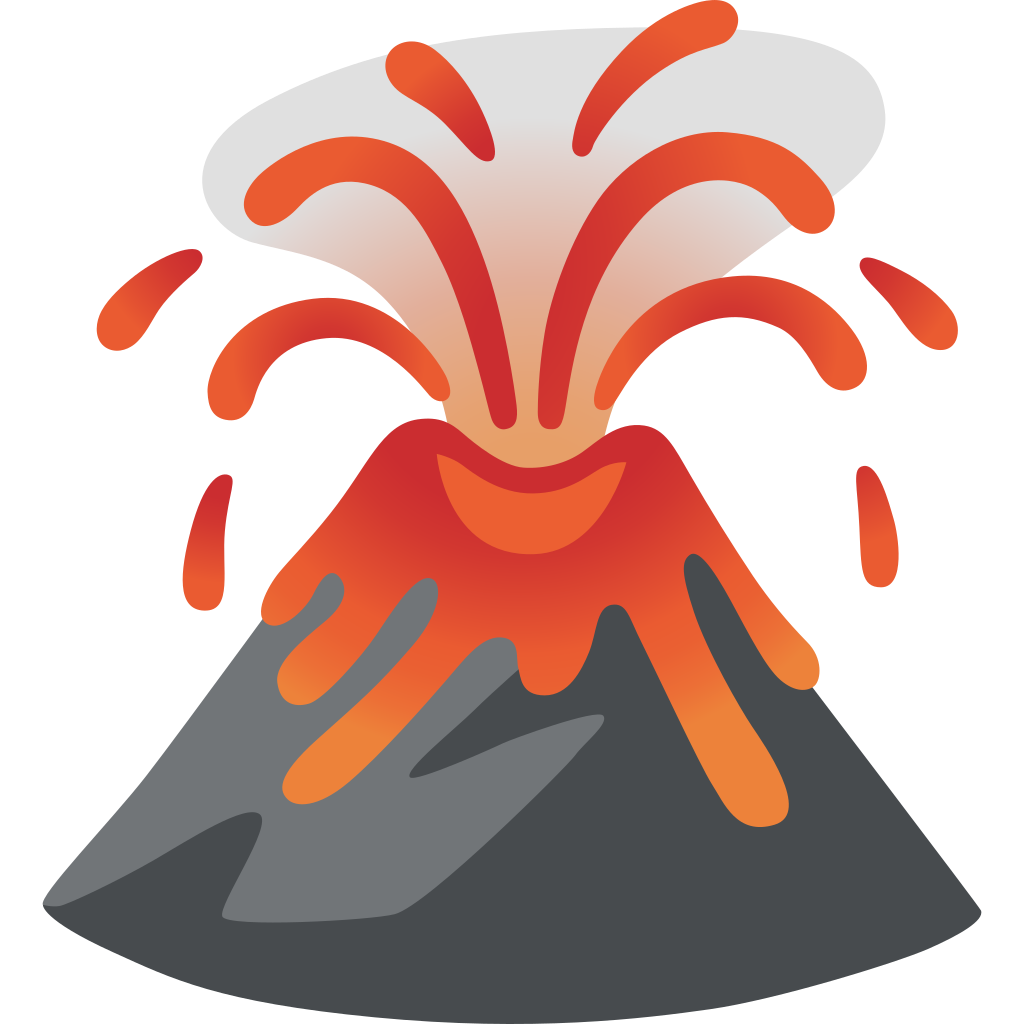}}\xspace}
\newcommand{\llavalogosmall}{{\includegraphics[scale=0.01]{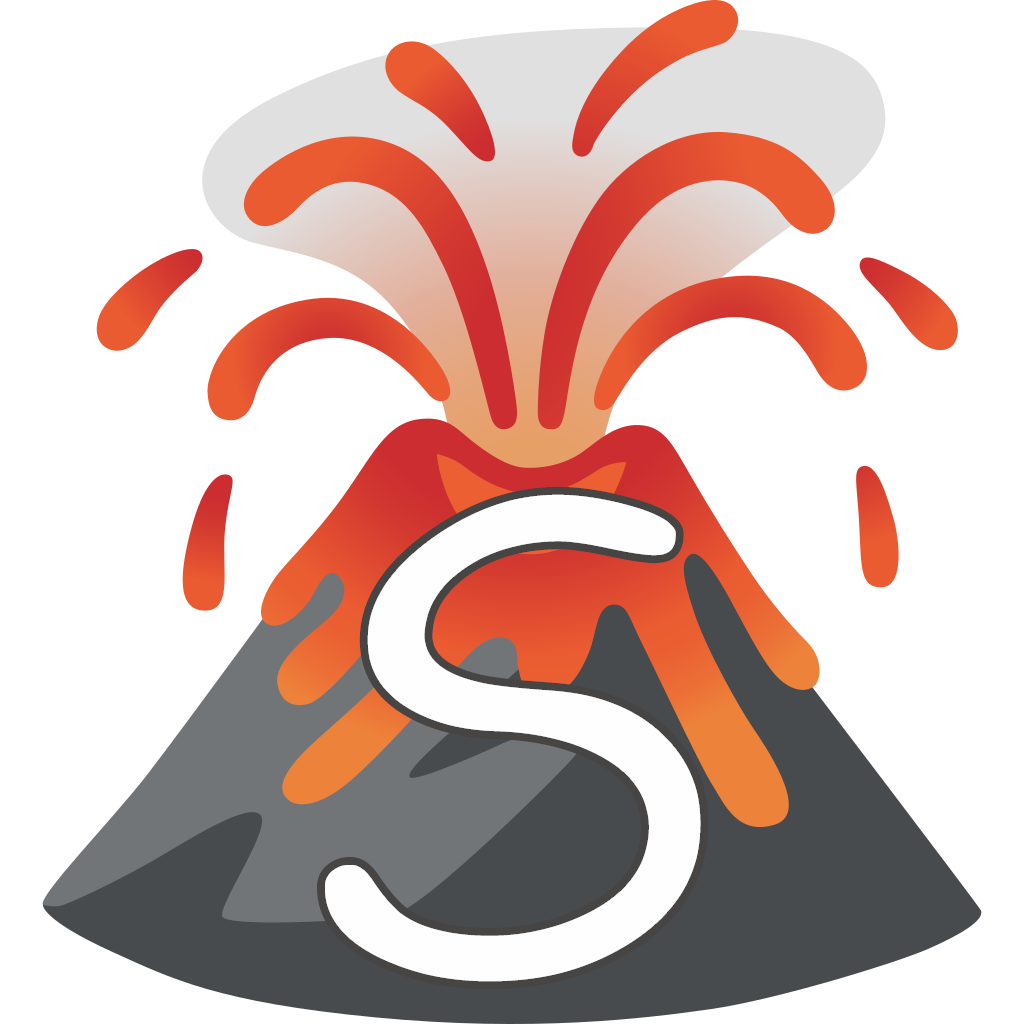}}\xspace}
\newcommand{\phivision}{\model{Phi-3.5}\xspace}
\newcommand{\phivisionfull}{\model{Phi-3.5-vision-instruct}\xspace}
\newcommand{\philogo}{{\includegraphics[scale=0.018]{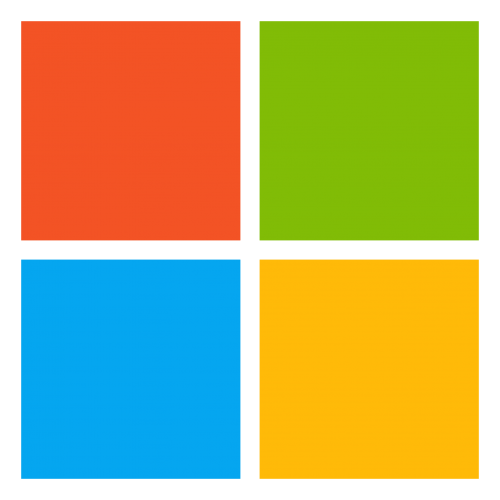}}\xspace}
\newcommand{\internvl}{\model{InternVL-2}\xspace}

\newcommand{\newsonnet}{\model{Sonnet-\textcolor{sonnet35_brown}{3.5}}\xspace}
\newcommand{\newsonnetfull}{\model{Claude-\textcolor{sonnet35_brown}{3.5} Sonnet}\xspace}
\newcommand{\newsonnetlogo}{{\includegraphics[scale=0.013]{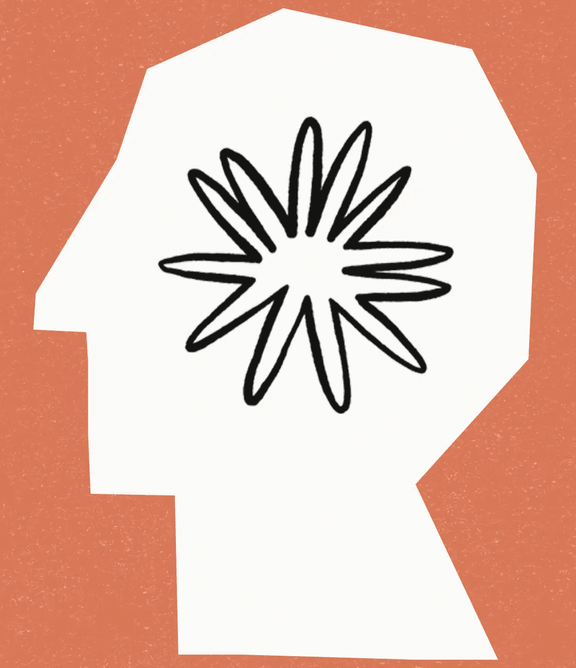}}\xspace}

\newcommand{\gptnumber}[1]{\textcolor{gpt_green}{\textbf{#1}}}
\newcommand{\gemininumber}[1]{\textcolor{gemini_blue}{\textbf{#1}}}

\newcommand{\newsonnetnumber}[1]{\textcolor{sonnet35_brown}{\textbf{#1}}}

\newcommand{\blindtest}{\texttt{BlindTest}\xspace}

\newcommand{\subsec}[1]{\noindent\textbf{#1}~~}
\newcommand{\subsecspace}[1]{\vspace*{0.5cm}\noindent\textbf{#1}~~}

\usepackage{pifont}
\newcommand{\cmark}{\ding{51}}%
\newcommand{\xmark}{\ding{55}}%
\usepackage{colortbl}
\usepackage{soul}
\usepackage{float}

\definecolor{yellow_sizes}{RGB}{255, 252, 130}
\definecolor{amber}{RGB}{252, 213, 123}
\definecolor{aureolin}{rgb}{0.99, 0.93, 0.0}
\definecolor{forestgreen(web)}{RGB}{125, 192, 125}
\definecolor{mediumslateblue}{rgb}{0.48, 0.41, 0.93}

\newcommand{\lineintersectlogo}{\includegraphics[scale=0.020]{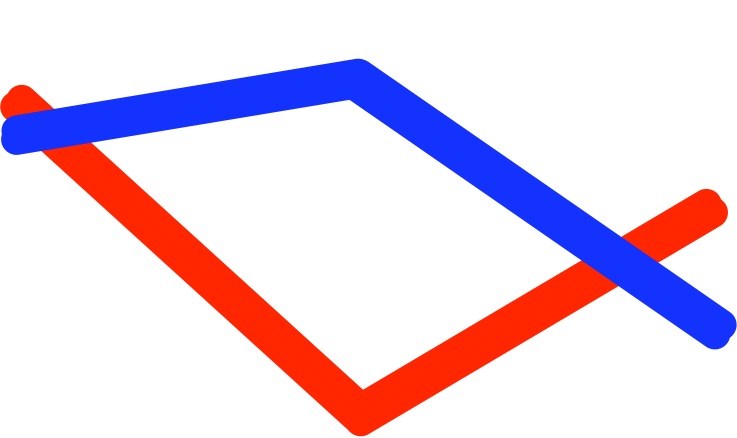}\xspace}
\newcommand{\twocirclelogo}{\includegraphics[scale=0.15]{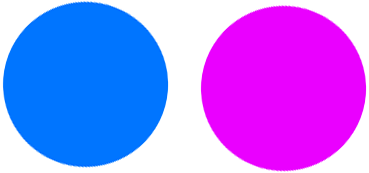}\xspace}
\newcommand{\nestedsquareslogo}{\includegraphics[scale=0.04]{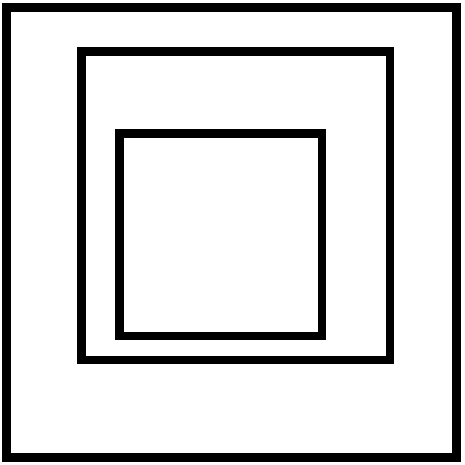}\xspace}
\newcommand{\gridlogo}{\includegraphics[scale=0.02]{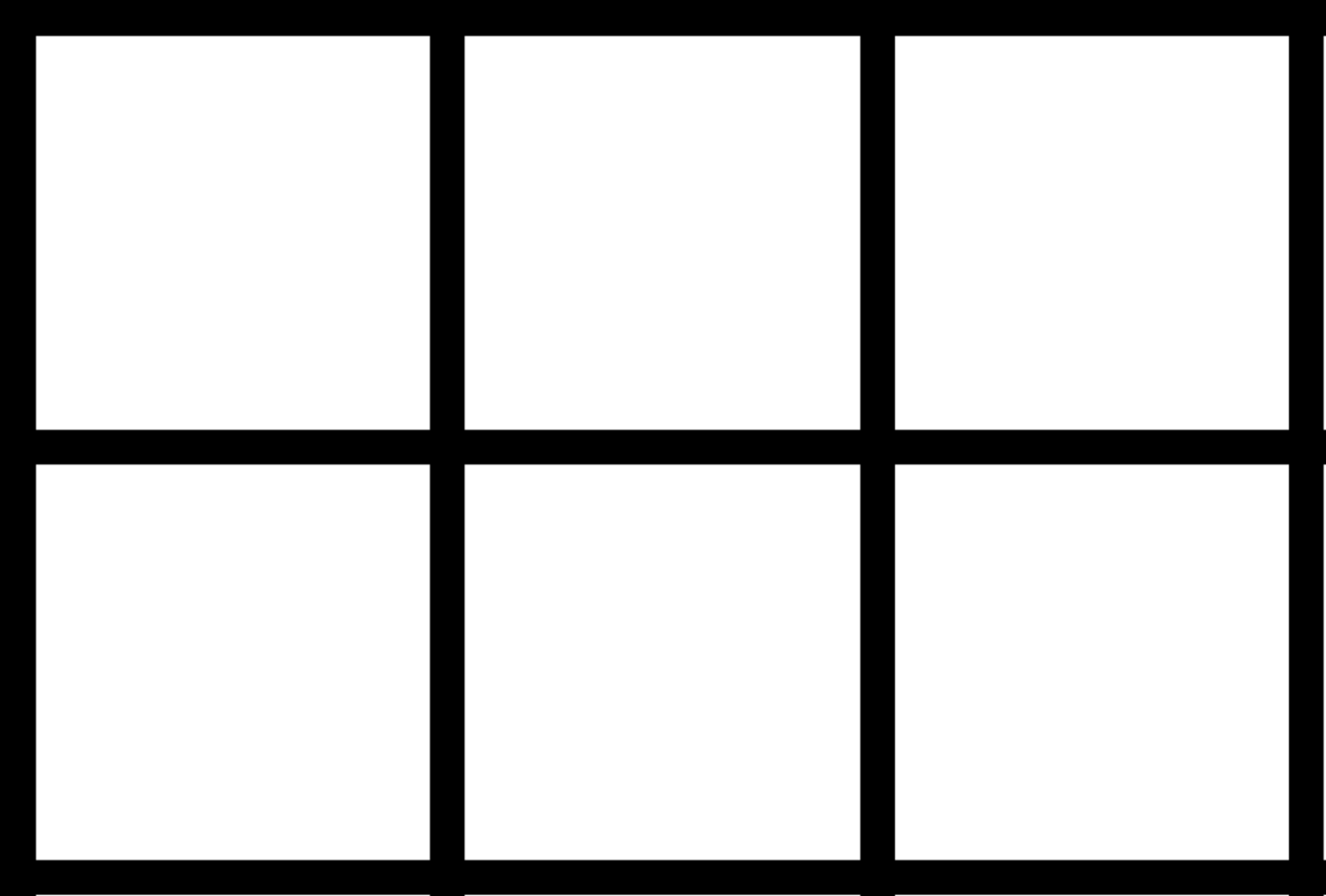}\xspace}
\newcommand{\olympiclogo}{\BigCircle~\hspace{-0.3cm}\BigCircle~\hspace{-0.3cm}\BigCircle\xspace}
\newcommand{\pentagonlogo}{\BigPentagon~\hspace{-0.3cm}\BigPentagon~\hspace{-0.3cm}\BigPentagon\xspace}
\newcommand{\pathfollowinglogo}{\includegraphics[scale=0.07]{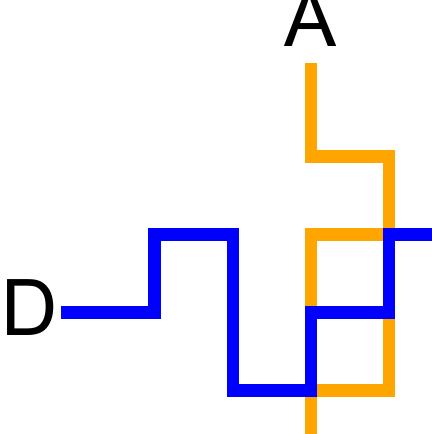}\xspace}

\newcommand{\increase}[1]{(\textcolor{ForestGreen}{+#1})}
\newcommand{\increasenoparent}[1]{\textcolor{ForestGreen}{+#1}}

\newcommand{\decreasenoparent}[1]{\textcolor{red}{-#1}}

\usepackage{tikz}

\definecolor{pyblue}{RGB}{50,115,247} 
\definecolor{pypurple}{RGB}{215,46,246} 
\newcommand{\bluecircle}{\tikz\draw[pyblue, fill=pyblue] (0,0) circle (.7ex); }
\newcommand{\purplecircle}{\tikz\draw[pypurple, fill=pypurple] (0,0) circle (.7ex); }
\usepackage{hyperref}
\definecolor{mydarkblue}{rgb}{0,0.53,0.96}
\hypersetup{
    colorlinks=true,
    linkcolor=magenta,
    citecolor=mydarkblue,
    filecolor=magenta,      
    urlcolor=magenta,
    pdftitle={Overleaf Example},
    pdfpagemode=FullScreen,
}

\usepackage{tcolorbox}
\global
\tcbset{
  aibox/.style={
    width=\textwidth,
    top=10pt,
    colback=white,
    colframe=black,
    colbacktitle=black,
    center,
  }
}
\newtcolorbox{AIbox}[2][]{aibox,title=#2,#1}
\tcbset{
  aiboxsmall/.style={
    width=0.62\textwidth,
    top=10pt,
    colback=white,
    colframe=black,
    colbacktitle=black,
    enhanced,
    top,
    attach boxed title to top left={yshift=-0.1in,xshift=0.15in},
    boxed title style={boxrule=0pt,colframe=white,},
  }
}
\newtcolorbox{AIboxSmall}[2][]{aiboxsmall,title=#2,#1}
\definecolor{aigold}{RGB}{244,210, 1} 
\definecolor{aired}{RGB}{255,180,181}
\newlength\savewidth

\definecolor{defaultcolor}{gray}{0.9}

\usepackage{multirow}

\newcommand{\smallredcircle}{%
    \tikz[baseline=(char.base)]{
        \node[inner sep=0pt] (char) {};
        \draw[red, line width=0.5mm] (char) ++(0,0.1) circle [radius=0.12cm];
    }%
}

\newcommand{\redcircle}[1]{%
    \tikz[baseline=(char.base)]{
        \node[inner sep=0pt] (char) {#1};
        \draw[red, line width=0.7mm] (char) ++(0,0.1) circle [radius=0.15cm];
    }%
}

\usepackage[T1]{fontenc}
\usepackage{helvet} 
\usepackage{opensans} 
\newcommand{\helvetica}[1]{{\fontfamily{phv}\selectfont\small #1}}
\newcommand{\opensansfont}[1]{{\fontfamily{fos}\selectfont #1}}

\newcommand{\overbar}[1]{\mkern 1.5mu\overline{\mkern-1.5mu#1\mkern-1.5mu}\mkern 1.5mu}

\newcommand{\overlapping}{${O}$\xspace}
\newcommand{\notoverlapping}{$\overbar{O}$\xspace}

\newcommand{\touching}{${T}$\xspace}
\newcommand{\nottouching}{$\overbar{T}$\xspace}

\newcommand{\circledword}[1]{{\textsf{\footnotesize#1}}\xspace}
\newcommand{\Subdermatoglyphic}{\scalebox{0.9}{\circledword{Subdermatoglyphic }}}
\newcommand{\SubdermatoglyphicCaption}{\scalebox{0.9}{\circledword{Subdermatoglyphic}}}
\newcommand{\subdermatoglyphicCircled}{\raisebox{-0.08cm}{\includegraphics[scale=0.15]{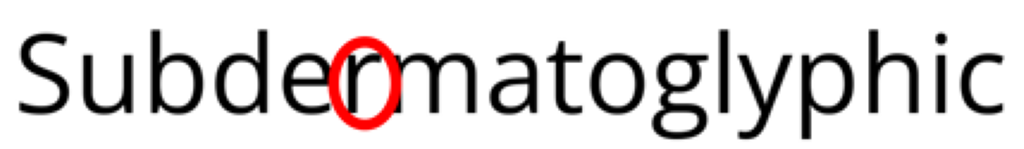}}}

\newcommand{\Acknowledgement}{\scalebox{0.9}{\circledword{Acknowledgement }}}
\newcommand{\tHyUiKaRbNqWeOpXcZvM}{\scalebox{0.9}{\circledword{tHyUiKaRbNqWeOpXcZvM }}}

\newcommand{\questionWord}{\scalebox{0.9}{\circledword{question }}}
\newcommand{\syndrome}{\scalebox{0.9}{\circledword{syndrome }}}
\newcommand{\environmental}{\scalebox{0.9}{\circledword{environmental }}}
\newcommand{\tournament}{\scalebox{0.9}{\circledword{tournament }}}
\newcommand{\civilization}{\scalebox{0.9}{\circledword{civilization }}}

\newcommand{\orangedistance}[1]{\sethlcolor{amber}\hl{#1}}

\newcommand{\distances}{\sethlcolor{amber}\hl{distances}}

\newcommand{\yellowsize}[1]{\sethlcolor{aureolin} \hl{#1}}

\definecolor{hallucinateyellow}{RGB}{255, 245, 234} 
\newcommand{\hallucinatepanel}[1]{\sethlcolor{hallucinateyellow}\hl{#1}}

\usepackage{array}
\usepackage{tabularx}
\usepackage{ragged2e}
\usepackage{makecell}
\usepackage{booktabs}

\usepackage{graphicx}
\usepackage{subcaption}

\begin{document}


\newcommand{\papertitle}{Vision language models are blind: Failing to translate detailed visual features into words}
\title{\papertitle}

\titlerunning{VLMs fail to translate detailed visual features into words}

\author{Pooyan Rahmanzadehgervi \inst{\textcolor{orange}{1}}\thanks{All first three authors contributed to running experiments, analyzing results, and writing the paper.}
~~~~~~~~Logan Bolton\inst{\textcolor{orange}{1}}$^{\star}$\\
~~~~~~~~\email{\footnotesize pooyan.rmz@gmail.com}\hspace{1.2cm}\email{\footnotesize logan.bolton@auburn.edu}\\
\vspace*{0.3cm}
Mohammad Reza Taesiri\inst{\textcolor{ForestGreen}{2}}$^{\star}$~~~
Anh Totti Nguyen\inst{\textcolor{orange}{1}}\\
\email{\footnotesize mtaesiri@gmail.com}\hspace{1.5cm}\email{\footnotesize anh.ng8@gmail.com}
}

\authorrunning{Rahmanzadehgervi, Bolton, Taesiri, Nguyen.}

\institute{Auburn University, USA \and
University of Alberta, Canada
}

\maketitle

\begin{abstract}
While large language models with vision capabilities (VLMs), \eg, \gpt and \geminifull, score high on many vision-understanding benchmarks, they are still struggling with low-level vision tasks that are easy to humans.
Specifically, on \blindtest, our suite of 7 very simple tasks, including identifying (a) whether two circles overlap; (b) how many times two lines intersect; (c) which letter is being circled in a word; and (d) the number of circles in an Olympic-like logo, four state-of-the-art VLMs are only 58.07\% accurate on average.
\newsonnet performs the best at 77.84\% accuracy, far from the human expected accuracy of 100\%.
Across different image resolutions and line widths, VLMs including slow-thinking models consistently struggle with those tasks that require precise spatial information when geometric primitives overlap or are close.
Yet, VLMs perform at near-100\% accuracy when much more space is added to separate shapes and letters.
Linear probing experiments show that vision encoders contain sufficient visual information to solve \blindtest and that language models fail to decode this information into correct answers.
Code and data are at: \href{https://vlmsareblind.github.io}{VLMsAreBlind.github.io}

\end{abstract}

\section{Introduction}

The advent of VLMs, starting with GPT-4V(ision) \cite{openai2023gpt4}, has enabled numerous, unprecedented image-text processing applications \cite{yang2023som}.
VLMs can accurately identify objects in a scene \cite{rasheed2023glamm,gemini2024groundingdemo,yang2023som} and perform complex tasks based on these detected objects, \eg, calculating the cost of beers on a table from an image of the scene and an image of the menu \cite{yang2023dawn}.
Interestingly, VLMs advance so quickly that describing \emph{unusual} activities in an image \cite{taesiri2024glitchbench} (\eg, a man ironing on a moving taxi) has become a standard sanity check \cite{dai2023instructblip}.

Existing VLM benchmarks cover a wide range of tasks \cite{yue2023mmmu,kembhavi2016diagram,masry2022chartqa}.
However, they often assess a high-level human-vs-machine performance gap conflating both visual and non-visual abilities.
Interestingly, the input images in so many questions, \eg, 42.9\% of MMMU \cite{yue2023mmmu}, are \emph{not} even necessary \cite{chen2024we} for determining the correct answer.
Many answers (1) can be inferred from the textual question and choices alone \cite{chen2024we,hegde2023analyzing}; and (2) are memorized by VLMs from their Internet-scale training \cite{chen2024we}.
In contrast, it is important to exclusively \textbf{measure the visual capabilities} of VLMs, independent of their strong language abilities.

In this paper, we test VLMs' ability to \emph{see} (not reasoning) on low-level vision tasks inspired by the ``visual acuity'' tests \cite{bailey2013visual} given to humans by optometrists.
We test four state-of-the-art (SOTA) VLMs: \gpt \cite{gpt4o}, \geminifull \cite{reid2024gemini}, \sonnetfull \cite{anthropic2024claude}, and \newsonnetfull \cite{sonnet35}
on our suite of 7 extremely simple visual tasks that involve only 2D geometric primitives (\eg, lines and circles) \cite{hughes2014computer} and require minimal world knowledge.
Our key findings are:

\begin{enumerate} 

    \item Despite excellent performance on chart and diagram benchmarks \cite{gpt4o,anthropic2024claude},
    VLMs cannot reliably tell whether two lines (or two circles) are intersecting, especially when close together.
    Accuracy in detecting 0, 1 or 2 intersections in a line chart \lineintersectlogo of two 2-segment piecewise-linear functions ranges from $\sim$41\% to 76\% (\cref{sec:res-2line}).
    For the two-circle \bluecircle\hspace{-0.1cm}\purplecircle task, VLMs perform better ($\sim$75--93\% accuracy), but still far from the expected 100\% (\cref{sec:res-2circles}).

    \item VLMs can perfectly recognize a circle (\smallredcircle) and a word (\Subdermatoglyphic) separately. 
    Yet, when the circle is superimposed on the word (\raisebox{-0.18\height}{\includegraphics[scale=0.15]{images/CircledLetter/InlineImages/Subde_r_matoglyphic.png}}), models tend to struggle to identify which letter is being circled (\cref{sec:res-redoval}).

    \item VLMs can accurately count shapes, \eg, circles (\BigCircle), that are disjoint and far apart.
    However, all VLMs struggle to count intersecting circles \raisebox{-0.05\height}{\includegraphics[scale=0.15]{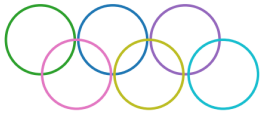}} (like the Olympic logo), and, generally, primitive shapes (\BigCircle, \BigSquare, \BigPentagon) that are overlapping or nested (\cref{sec:res-count-shapes}).

    \item Tiling up squares into a grid \raisebox{-0.1\height}{\gridlogo}, we find VLMs to fail to count the number of rows or columns in the grid, whether empty or containing text (\cref{sec:res-gridcount}).
    This is in stark contrast to VLM high performance ($\geq$ 90\% accuracy) \cite{gpt4o,reid2024gemini} on DocVQA \cite{mathew2021docvqa}, which includes many questions with tables.

    \item When tasked with tracing colored paths in a simplified subway map of only 2 to 8 paths and a total of 4 stations, VLMs often fail to count the paths between two stations, \ie, with an accuracy of $\sim$31\% to 58\% (\cref{sec:res-paths}).

    \item \gpt is better than \geminifull on 7 existing complex VLM benchmarks \cite{gpt4o,reid2024gemini} but worse on \blindtest.
    On average across all 7 tasks, VLMs perform at 58.07\% accuracy with \newsonnet being the best (77.84\% accuracy), which is still far lower than the expected 100\% accuracy of humans (see \cref{tab:all_model_acc}).
    In sum, \blindtest reveals some remarkable VLM limitations that are not measured in prior benchmarks.

    \item The SOTA ``slow-thinking'' VLMs are at best on par with the regular VLMs on \blindtest (71.59\% vs. 72.75\%; \cref{sec:eval_reasoning}), \ie, longer inference is not immediately addressing the low-level visual challenge posed by \blindtest.
    
    \item Most VLMs perform better (with some reaching $\sim$100\% accuracy) on \emph{simplified} versions of \blindtest (\cref{sec:spacing}) where we increasingly add more space between letters (\cref{sec:spaced-words}) or between shapes (\cref{sec:spaced-shapes-count}), and gradually reduce the number of turns in subway maps (\cref{sec:spaced-subway-count}). 
    For instance, \llavaone-72B-ov and \sonnet's accuracy is increased by \increasenoparent{20} and \increasenoparent{63} on the circled-letter task and counting overlapping circles, respectively, when the images are simplified.

    \item Linear-probing the features of the vision encoders before and after the projection layer in the \emph{smallest} open-source VLMs (\llavaSI and \phivision) shows that these vision encoders already contain sufficient information to solve the
    two-circle \twocirclelogo and counting the intersections in the line chart \lineintersectlogo tasks with $\geq$99.47\% accuracy (\cref{sec:linear_feature_probe}).
    That is, the challenge lies in decoding the visual information into the correct language outputs.

\end{enumerate}

\begin{figure}[h]
\centering
\begin{AIbox}{Examples from \blindtest benchmark with VLMs' responses}
    \hspace{-0.045\textwidth}
    \begin{tabular}{lp{0.8cm}c|p{0.8cm}c|p{0.8cm}c|p{0.8cm}c|p{0.8cm}c|p{0.8cm}c|p{0.8cm}c}
    &\multicolumn{2}{c}{\textbf{P1}}&\multicolumn{2}{c}{\textbf{P2}}&\multicolumn{2}{c}{\textbf{P3}}&\multicolumn{2}{c}{\textbf{P4}}&\multicolumn{2}{c}{\textbf{P5}}&\multicolumn{2}{c}{\textbf{P6}}&\multicolumn{2}{c}{\textbf{P7}}\\
     &\multicolumn{2}{c}{\includegraphics[width=0.14\textwidth]{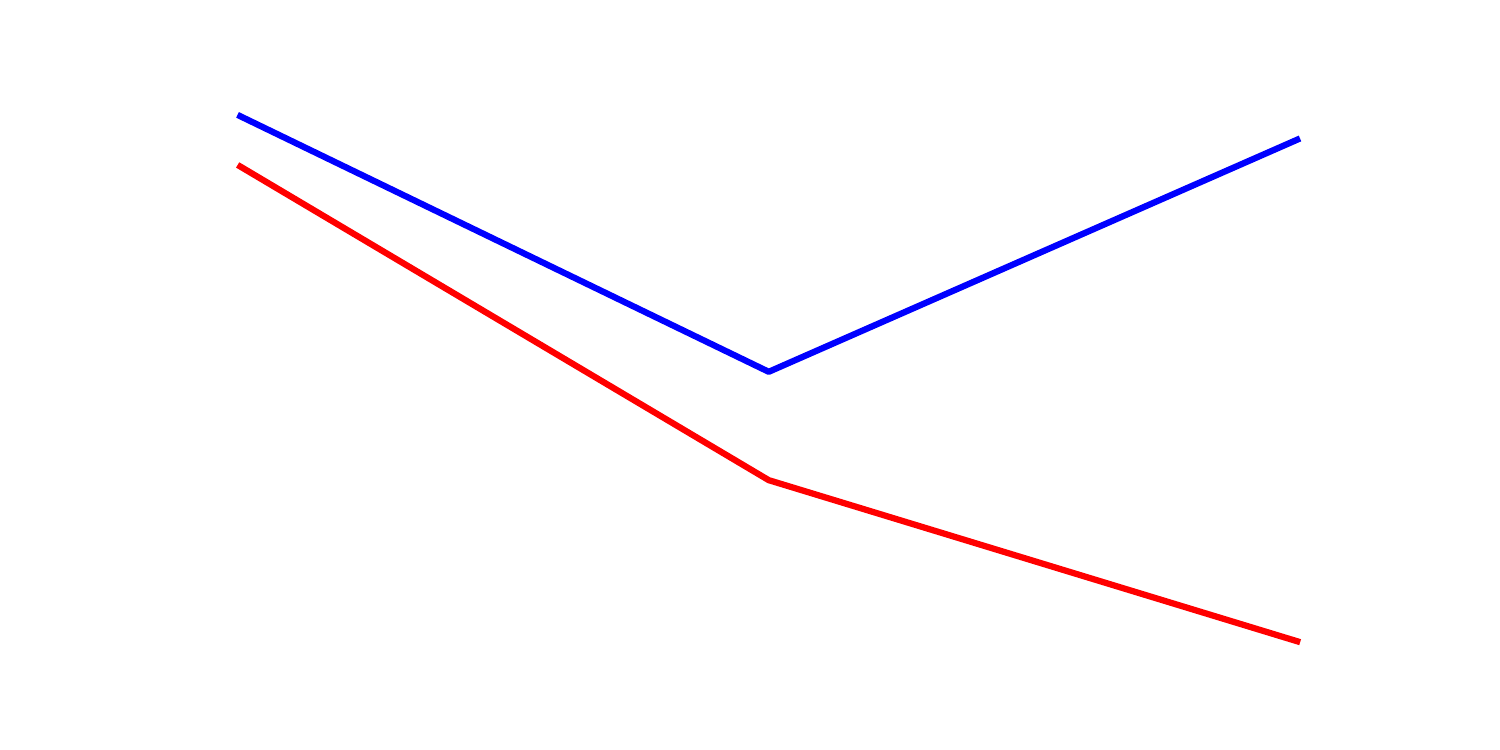}} & \multicolumn{2}{c}{\includegraphics[width=0.10\textwidth]{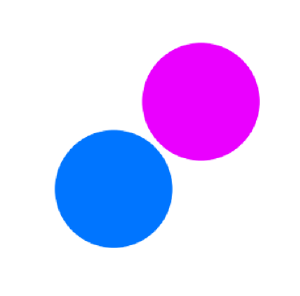}} &\multicolumn{2}{c}{\includegraphics[width=0.14\textwidth]{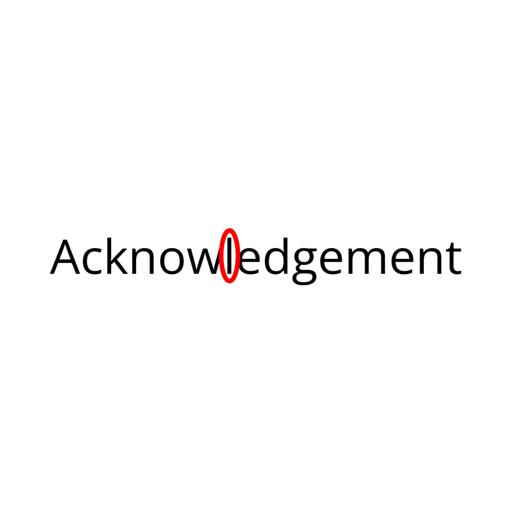}} &\multicolumn{2}{c}{\includegraphics[width=0.14\textwidth]{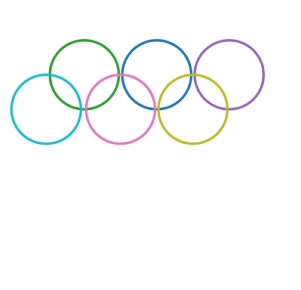}} &\multicolumn{2}{c}{\includegraphics[trim={130 150 200 130},clip,width=0.14\textwidth]{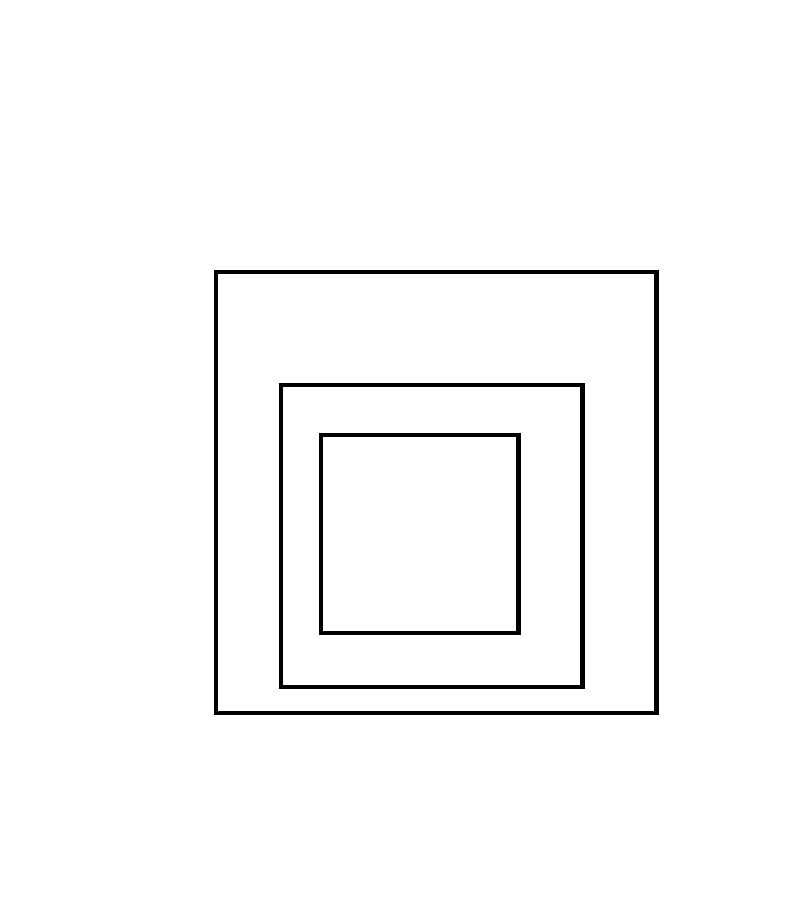}}&\multicolumn{2}{c}{\includegraphics[width=0.14\textwidth]{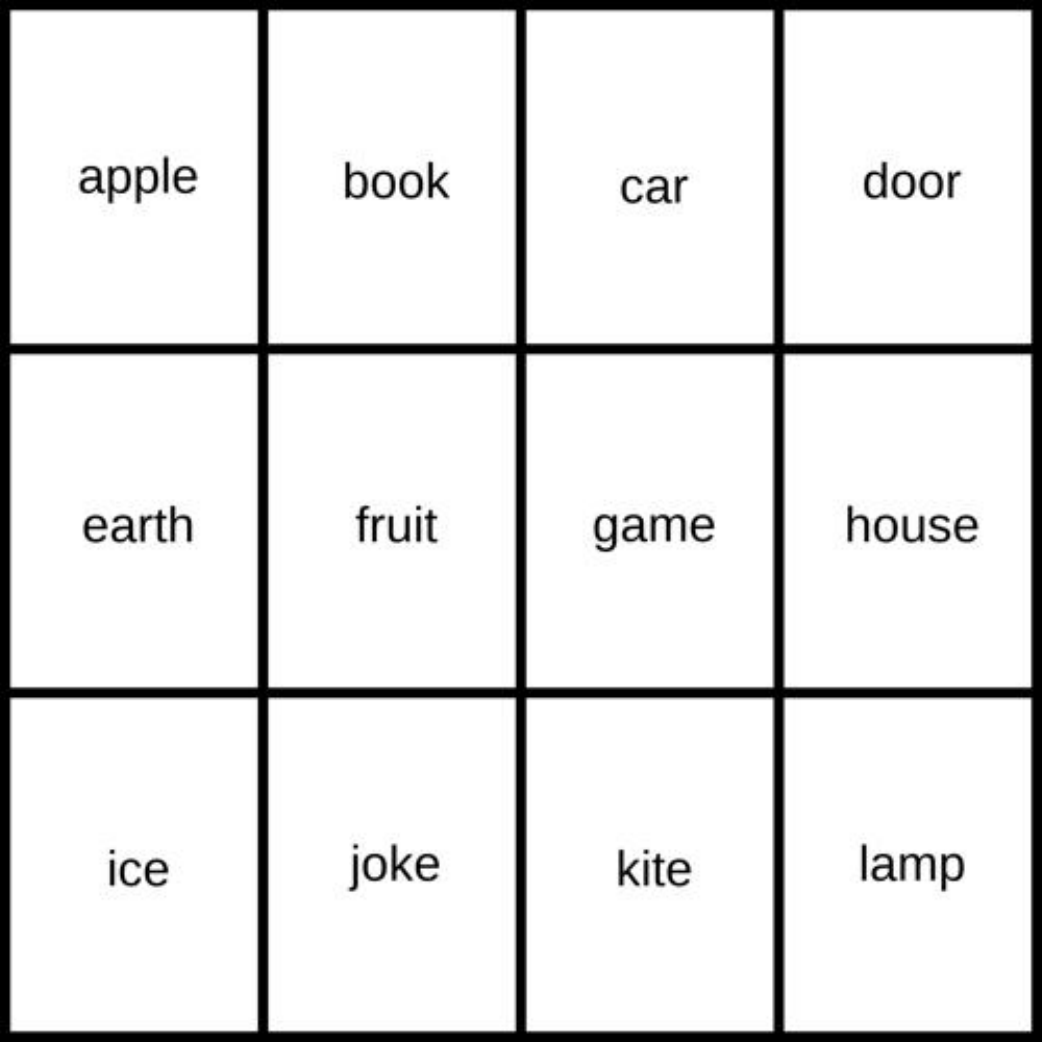}}&\multicolumn{2}{c}{\includegraphics[width=0.14\textwidth]{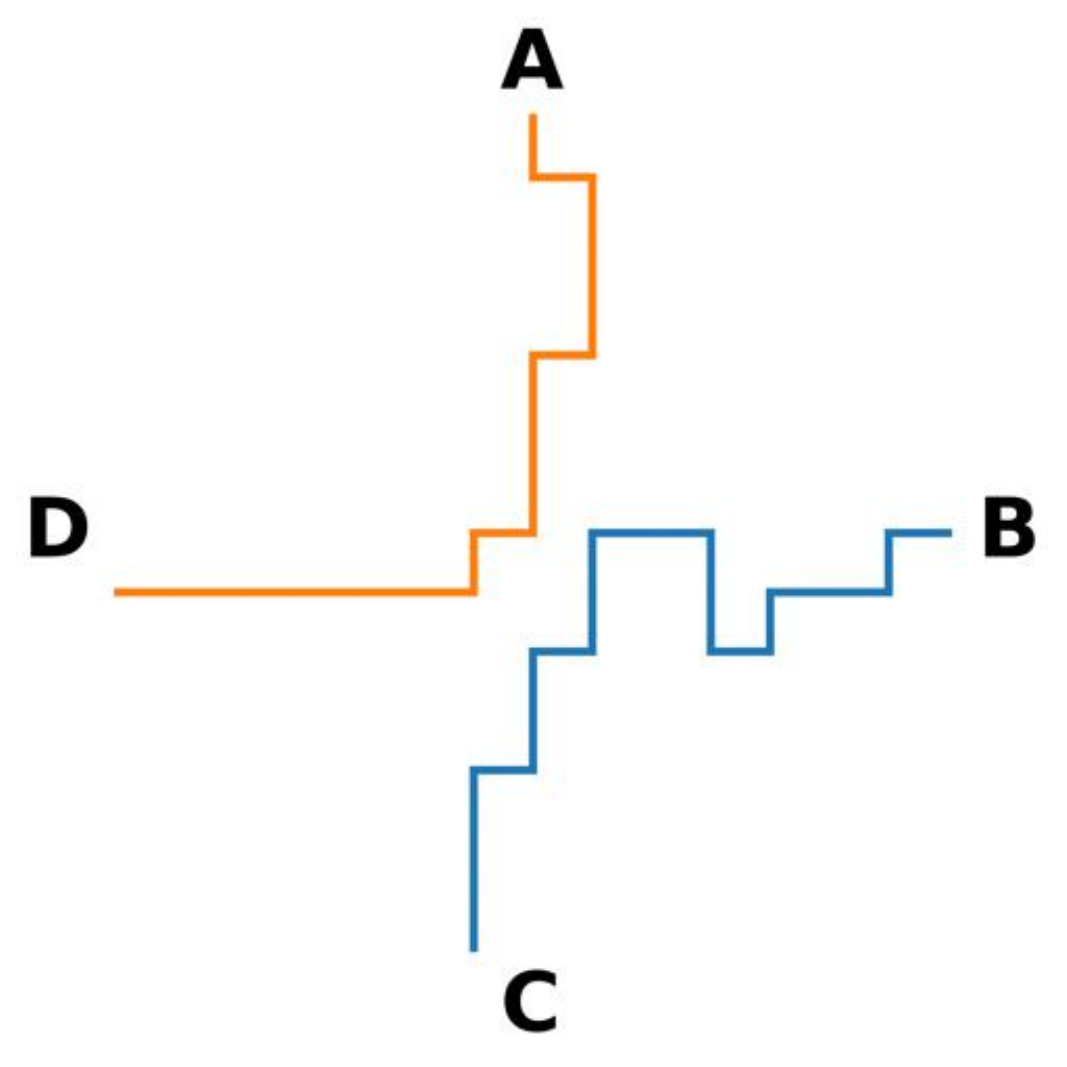}}\\
     \rowcolor{lightgray}
     \raisebox{-0.2\height}\gptlogo & \centering 1 & \textcolor{red}{\xmark} &  \centering Yes & \textcolor{red}{\xmark}&  \centering o & \textcolor{red}{\xmark} &  \centering 6 & \textcolor{ForestGreen}{\cmark} &  \centering 5 & \textcolor{red}{\xmark}&  \centering 3$\times$4 & \textcolor{ForestGreen}{\cmark}&  \centering 1 & \textcolor{ForestGreen}{\cmark}\\
     \raisebox{-0.2\height}\geminilogo & \centering 1 & \textcolor{red}{\xmark} & \centering No & \textcolor{ForestGreen}{\cmark}&  \centering w & \textcolor{red}{\xmark} &  \centering 5 & \textcolor{red}{\xmark} &  \centering 3 & \textcolor{ForestGreen}{\cmark}&  \centering 3$\times$4 & \textcolor{ForestGreen}{\cmark}&  \centering 2 & \textcolor{red}{\xmark}\\
     \rowcolor{lightgray}
     \raisebox{-0.2\height}\sonnetlogo & \centering 1 & \textcolor{red}{\xmark} & \centering Yes & \textcolor{red}{\xmark}&  \centering o & \textcolor{red}{\xmark} &  \centering 5 & \textcolor{red}{\xmark} &  \centering 4 & \textcolor{red}{\xmark}&  \centering 4$\times$4 & \textcolor{red}{\xmark}&  \centering 2 & \textcolor{red}{\xmark}\\
     \raisebox{-0.2\height}\newsonnetlogo & \centering 0 & \textcolor{ForestGreen}{\cmark} & \centering No & \textcolor{ForestGreen}{\cmark}&  \centering l & \textcolor{ForestGreen}{\cmark} &  \centering 6 & \textcolor{ForestGreen}{\cmark} &  \centering 3 & \textcolor{ForestGreen}{\cmark}&  \centering 3$\times$4 & \textcolor{ForestGreen}{\cmark}&  \centering 1 & \textcolor{ForestGreen}{\cmark}\\
     \end{tabular}

    \vspace{4pt}
    \hspace{0.13\textwidth}
    \centering
    \begin{tabular}{cccccccccccccc}
    \raisebox{-0.1\height}\gptlogo & \gpt &&&
      \raisebox{-0.1\height}\geminilogo & \gemini & &&\raisebox{-0.1\height}\sonnetlogo & \sonnet & && \raisebox{-0.12\height}\newsonnetlogo & \newsonnet \\
      \end{tabular}
    \hfill \break
    \hfill \break
    \raggedright
    \footnotesize
      \textbf{P1:} How many times do the blue and red lines touch each other? Answer with a number in curly brackets, e.g., \{5\}. \\
      \textbf{P2:} Are the two circles overlapping? Answer with Yes/No. \\
      \textbf{P3:} Which character is being highlighted with a red oval?  Please provide your answer in curly brackets{,} e.g. \{a\} \\
      \textbf{P4:} How many circles are in the image? Answer with only the number in numerical format. \\
      \textbf{P5:} How many squares are in the image? Please answer with a number in curly brackets e.g., \{10\}.\\
      \textbf{P6:} Count the number of rows and columns and answer with numbers in curly
brackets. For example, rows=\{5\} columns=\{6\}. \\
      \textbf{P7:} How many single-color paths go from A to D? Answer with a number in curly
brackets e.g. \{3\}. \\

\end{AIbox}
\caption{VLMs fail on the simple tasks of \blindtest.}
\label{fig:teaser}
\end{figure}

\section{Vision language models}

Our goal is to evaluate how SOTA VLMs perceive simple images composed of \emph{interacting} geometric primitives. 

\subsec{4 commercial}
We test four SOTA models: \gpt (\raisebox{-0.16\height}\gptlogo), \geminifull (\raisebox{-0.16\height}\geminilogo~\gemini), \sonnetfull (\raisebox{-0.16\height}\sonnetlogo~\sonnet), and \newsonnetfull (\raisebox{-0.2\height}\newsonnetlogo~\newsonnet) that are ranking highest on 7 recent vision-language benchmarks (see \cite{gpt4o} and Table 10 in \cite{reid2024gemini}), which cover multi-discipline, college-level subjects in MMMU \cite{yue2023mmmu}, science diagrams in AI2D \cite{kembhavi2016diagram}, mathematics in MathVista \cite{lu2023mathvista}, charts in ChartQA \cite{masry2022chartqa}, documents in DocVQA \cite{mathew2021docvqa}, and videos in ActivityNet-QA \cite{yu2019activitynet} \& EgoSchema \cite{mangalam2024egoschema}.
We initially run experiments with Claude 3 \texttt{Opus} \cite{opus} but swap it with \newsonnet, which performs more accurately on \blindtest and costs 5$\times$ less.
All models tested are described in \cref{sec:models}.

\subsecspace{8 open-source} For completeness, we also test 8 \emph{open-source} models of varying sizes (from 0.5B to 72B parameters) across three different families: \llavaonefull (\raisebox{-0.16\height}\llavalogolarge \llavaone) \cite{li2024llava}, \phivisionfull (\raisebox{-0.16\height}\philogo \phivision) \cite{abdin2024phi}, and \internvl \cite{chen2023internvl}.
Yet, they underperform the four closed-source models described above (see results in \cref{app:open-source}).

\subsecspace{2 slow-thinking}
Given that \blindtest intuitively does not require slow, iterative thinking, our main goal is to evaluate standard VLMs (\eg, \gpt instead of \texttt{GPT-o1}).
However, for completeness, we also test two slow-thinking models \cite{kahneman2011thinking}: A closed-source \geminithinkfull and an open-source \qvqfull (\qvq) \cite{Qwen2VL,qvq-72b-preview} (\cref{sec:eval_reasoning}). 
However, these models generally do not outperform standard ones on \blindtest, supporting our hypothesis that \blindtest is a low-level visual task that does not require high-level reasoning.


\section{\blindtest benchmark of 7 tasks}

\subsec{Eye exams} Like humans' visual acuity tests \cite{bailey2013visual}, we design a set of 7 very simple, yet novel tasks that are composed of common geometric primitives.
We do not use the existing tests designed for human-eye exams for two reasons.
First, we avoid using the questions that exist on the Internet, which may provide an inflated measure of vision capabilities \cite{yu2023mm,chen2024we,hegde2023analyzing}.
Second, our preliminary experiments show that \gpt \emph{already performs very well} on humans' eye exams, which typically contain single, separate symbols---\eg, the Snellen chart \cite{bailey2013visual}, tumbling E \cite{bailey2013visual}, and contrast sensitivity charts \cite{mantyjarvi2001normal,zeiss_eye_exams}.
\\
\subsec{Motivation} Our \blindtest benchmark tests VLMs on identifying known geometric primitives when they are close together, overlapping, or intersecting.
We hypothesize that VLMs will struggle because they mostly rely on ``late fusion'' \cite{liu2024visual,team2024chameleon}, \ie, first extracting visual representations \emph{without} considering the textual question, and then feeding them to a large language model (LLM) for processing.
Therefore, while geometric primitives in \blindtest are well known, their exact spatial information on a white canvas (\eg, the size and position of a \BigCircle) is typically not describable in natural language, even for humans, and may not be captured by the vision encoders trained mostly on natural images.
\\
\subsec{Controls}
For each test image, we prompt VLMs using \textbf{two} different, yet semantically equivalent questions.
Furthermore, we test VLMs on multiple versions of each task across \textbf{three} different image sizes (\cref{sec:count_intersections,sec:two_circles,sec:olympic_logo,sec:count_grid,sec:subway}) and \textbf{two} to \textbf{three} line thickness values (\cref{sec:count_intersections,sec:count_nested_squares,sec:olympic_logo,sec:count_grid,sec:subway}).

\subsection{Task 1: Counting line intersections \lineintersectlogo}
\label{sec:count_intersections}

Given the impressive accuracy of VLMs on answering questions on diagrams and charts (\eg, \newsonnet scoring 94.7\% on AI2D and 90.8\% on ChartQA) \cite{sonnet35}, a reasonable hypothesis is that VLMs must be able to see if two graphs intersect in a chart.
Here, we test this hypothesis by asking VLMs to count the number of intersections (0, 1 or 2) between two 2-segment piece-wise linear functions.

\subsec{Images}  
We create 1,800 images (\cref{fig:2dline}) of 2D line plots drawn on an image of size of $C\times C$, where $C\in\{384, 768, 1152\}$. Each line plot consists of two line segments, defined by three points whose x-coordinates are fixed at $\{0,\frac{C}{2}, C\}$px (see \cref{fig:2dline}).
The y-coordinates are randomly sampled from a pre-defined, invisible 12$\times$12 grid to ensure there is sufficient spacing between two plots and that there are exactly 0, 1 or 2 intersections.
See \cref{app:2dline-bench} for more details.

\subsection{Task 2: Two circles \twocirclelogo}
\label{sec:two_circles}

In the task of counting line intersections (\cref{sec:count_intersections}), each image contains two long, thin colored lines on a large white canvas.
Here, we test models in a complementary setting where the two interacting objects (here, two same-sized filled circles \twocirclelogo) are larger while their gap is smaller.
This task evaluates VLM ability in detecting (1) a small gap between two circles; and (2) that two circles are overlapping, \ie, no gaps.
We vary circle and gap sizes and ask VLMs if two circles are (a) overlapping or (b) touching each other.

\subsec{Images}
Given a blank image of size $C\times C$, we draw two same-sized circles of diameter $\phi \in \{ \frac{C}{4}, \frac{C}{5}, \frac{C}{6}, \frac{C}{7} \}$ with a boundary-to-boundary distance $= \phi \times d$ where $d \in$
\{-0.25, -0.2, -0.15, -0.1, -0.05, 0.0, 0.05, 0.1, 0.15, 0.2, 0.25, 0.3, 0.35, 0.4, 0.45, 0.5\} to cover all three cases: overlapping, tangent, and disjoint (see \cref{fig:2circles_examples_main}a).
The two circles are arranged in four different orientations, making a $90^\circ$, $0^\circ$, $-45^\circ$, and $45^\circ$ angle with the x-axis (\cref{fig:2circles_examples_main}b).
The whole grid sampling generates 224 images per image size.
We replicate the procedure for 3 image sizes, \ie, $C = 384, 769, 1155$ px to create a total of 3$\times$224 = 768 images.
See \cref{app:2circles-bench} for more details.

\subsection{Task 3: The circled letter \subdermatoglyphicCircled}
\label{sec:circled_letter}

Consistent with prior reports \cite{yang2023dawn,yang2023som,shtedritski2023does}, we find that VLMs can 100\% accurately identify a primitive shape (\eg, a red circle \smallredcircle) \cite{shtedritski2023does} and can perfectly read an English word (\eg, \Subdermatoglyphic) alone.
Here, we superimpose the red circle on every letter, one at a time, in the word, and ask VLMs to identify which letter is being circled.
While the task is easy to humans, our hypothesis is that if a VLM's vision is ``blurry'', it might not be able to identify the exact letter being circled since there is tiny spacing between the adjacent letters.

\subsec{Images}
We choose three strings \Acknowledgement, \Subdermatoglyphic, and\\ \tHyUiKaRbNqWeOpXcZvM because they contain characters of variable widths and heights.
Furthermore, all four tested VLMs can read out all characters in these strings when they are input to the models as an image.
While \Acknowledgement is a common English word, \Subdermatoglyphic is the longest word without repetitive letters.
We also test VLMs on the random string \tHyUiKaRbNqWeOpXcZvM to estimate how much model accuracy is due to its familiarity with the word.

For each (string, circled-letter) pair, we render a 512$\times$512 image by choosing among 3 red oval line-thickness levels, 2 font families, and 4 different values of image padding for a total of 24 images.
That is, we generate 360, 408, and 480 images for \Acknowledgement (15 letters), \Subdermatoglyphic (17 letters), and \tHyUiKaRbNqWeOpXcZvM (20 letters), respectively.
We ensure each letter to be circled fits completely the oval \smallredcircle ~(see \cref{fig:red_oval}).
See \cref{app:redoval-bench} for more details.

    


\subsection{Task 4: Counting overlapping shapes \olympiclogo}
\label{sec:olympic_logo}

Aligned with prior research \cite{yang2023dawn}, we also find VLMs to be able to count disjoint circles (\BigCircle\BigCircle\BigCircle).
Yet, here, we test VLMs on counting circles that are \emph{intersecting} (\olympiclogo) like in the Olympic logo---a common cognitive development exercise for preschoolers \cite{preSchoolerCountShapes,preSchoolerCountShapes2}.
Our hypothesis is that a ``blurry'' vision may not see the intersection between two circles clearly and therefore unable to trace circles and count them.
For generalization of our findings, we repeat the experiment with pentagons (\BigPentagon) as well (instead of circles).

\subsec{Images}
In an image of size $C\times C$, where $C\in\{384, 769, 1155\}$px. 
We draw $N\in\{5, 6, 7, 8, 9\}$ overlapping, same-sized circles arranged in two rows like the Olympic logo (see \cref{fig:olympic_examples_main}).
A circle diameter $\phi \in\{\frac{C}{7}, \frac{C}{10}\}$.
We repeat the images with two different line thicknesses for rendering circles.
This procedure renders 3 resolutions $\times$ 5 values of $N$ $\times$ 2 diameters $\times$ 2 line widths $\times$ 2 color options = 120 images. 
We also vertically flip all 120 images, resulting in a total of 240 images. 
We repeat for pentagons (\BigPentagon) in addition to circles (\BigCircle), resulting in 240$\times$2 shapes = 480 images in total.
For pentagons, their side length $d\in\{\frac{C}{7}, \frac{C}{10}\}$.
See \cref{app:olympic-bench} for more details.

\subsection{Task 5: Counting the nested squares \nestedsquareslogo}
\label{sec:count_nested_squares}

In addition to testing VLMs on counting the \textbf{intersecting} circles (\cref{sec:olympic_logo}), here, we test a complementary setting by arranging the shapes so that their edges do \emph{not} intersect.
That is, each shape is \textbf{nested} entirely inside another (see \cref{fig:nested_sq}).
For completeness, we test squares (\BigSquare) in this task.

\subsec{Images} 
In an image of size 1000$\times$1000px,
we render $N\in\{2, 3, 4, 5\}$ nested squares one at a time from the largest to the smallest.
First, the outermost square is rendered using a random edge length $d$.
And each subsequent smaller square is placed randomly inside the previous one and has an edge length of 75\% of that of the outer square.
We render squares using a line width of $\{3, 4, 6\}$px and ensure no squares touch by edges.
For each line width, we generate 10 images (where squares have different, random locations) to create 3 $\times$ 10 = 30 images.
Repeating the process for all $N$ values results in 4 $\times$ 30 = 120 images. 
See \cref{app:squares-bench} for more details.


    



\subsection{Task 6: Counting the rows and columns of a grid \gridlogo}
\label{sec:count_grid}

The results from prior tasks show VLMs cannot always count shapes that are overlapping (\cref{sec:olympic_logo}) or nested \nestedsquareslogo (\cref{sec:count_nested_squares}). 
What about adjacent shapes \BigSquare\hspace{-0.135cm}\BigSquare?
Here, we tile up shapes (specifically, \BigSquare) into a grid and challenge VLMs to count---a task that is supposedly simple to VLMs given their remarkable performance ($\geq$ 90\% accuracy) \cite{gpt4o,reid2024gemini} on DocVQA \cite{mathew2021docvqa}, which includes many questions with tables.
To simplify the task, we ask models to count the number of rows and columns in a given table (either empty or text-containing).

\subsec{Images}
A grid may have $N$$\times$$N$, $N$$\times$$N'$, or $N'$$\times$$N$ cells, where $N\in\{3, 4, 5, 6, 7, 8,\\ 9\}$ and $N' = N + 1$. 
We also include grids of size 10$\times$10 to balance the benchmark with the row and column sizes.
Each grid is rendered with two different line widths on a canvas of size $C\times C$ where $C\in\{500, 1250, 2000\}$px.
Besides empty grids, we also replicate the procedure to make grids contain text (which is more common in real-world tables) where each cell contains a single random English word (see \cref{fig:grid-sub-examples}). 
Both versions (empty and text-containing) combined have 2$\times$132 = 264 images. 
See \cref{app:grid-bench} for more details.

\subsection{Task 7: Following single-colored paths \pathfollowinglogo}
\label{sec:subway}

It is important for VLMs to be able to follow paths in order to read maps or charts \cite{masry2022chartqa}, interpret graphs \cite{li2024visiongraph}, and user annotations (\eg, arrows) in input images \cite{yang2023dawn}.
To assess path-following capability, this task asks models to count the unique-color paths between two given stations in a simplified subway map.


\subsec{Images}
We create each subway map on an image of size $C\times C$, where $C\in\{512, 1024\}$px (see \cref{fig:sub-examples-f}). 
We write 4 station names ({\textsf{\footnotesize A}}, {\textsf{\footnotesize B}}, {\textsf{\footnotesize C}}, {\textsf{\footnotesize D}}) at 4 fixed coordinates $\in\{(\frac{C}{2}, C), (C, \frac{C}{2}), (\frac{C}{2}, 0), (0, \frac{C}{2})\}$, respectively. 
We divide the canvas into an invisible grid of 18$\times$18 cells and initialize 3 path-starting points $\frac{C}{18}$px away from each station. 
We draw a path, using the depth-first search algorithm starting from a random station and a random starting point, where a valid move is one cell in any direction: North, south, east or west.
We repeat the process so that each station has exactly $N\in\{1, 2, 3\}$ outgoing paths, for a total of 180 maps. 
See \cref{app:subway-bench} for details.





\section{Results}
\label{sec:results}

\begin{table}[h]
    \centering
    \caption{
    The \hallucinatepanel{mean accuracy} (\%) of all four VLMs over 7 \blindtest tasks is \hallucinatepanel{58.07}\%, substantially higher than random chance (24\%), which is computed considering each task as a single-label, $N$-way classification problem.
    \newsonnet is the best (77.84\%) but still far from the expected 100\% accuracy.
    Note that the best open-source VLM (\cref{tab:open_model_acc}) is only on par with \sonnetlogo and the best slow-thinking VLM is slightly worse than \newsonnetlogo (\cref{sec:eval_reasoning}).
    }
    \label{tab:all_model_acc}
    \begin{tabular}{lccccccccc}
\toprule
 & a. & b. & c. & d. & e.  & f. & g. & h. & i.\\
Model & \lineintersectlogo & \twocirclelogo & \smallredcircle & \olympiclogo & \pentagonlogo  & \nestedsquareslogo & \gridlogo & \pathfollowinglogo & Task mean\\
\midrule
Random           & 33.33           & 50.00          & 5.77 &  20.00 & 20.00 & 25.00  & 4.55 & 33.33 & 24.00 \\
\midrule
\raisebox{-.1\height}\gptlogo~\gpt           & 41.61           & 75.91          & 74.23 & 41.25 & 20.21 & 55.83 & 39.58 & 53.19 & 50.23  \\
\raisebox{-.1\height}\geminilogo~\gemini~~~   & 66.94  &  \textbf{93.62} & 83.29 & 20.25 & 24.17  & 87.08 & 39.39 & 53.13 & 58.48 \\
\raisebox{-.1\height}\sonnetlogo~\sonnet        & 43.41 & 86.46 &  72.06 & 29.79 & 1.87 & 65.00 & 36.17 & 31.11 & 45.73 \\
\raisebox{-.2\height}\newsonnetlogo~\newsonnet & \textbf{75.36} &	90.82 &	\textbf{87.88} &	\textbf{66.46} &	\textbf{77.71} &	\textbf{92.08} &	\textbf{74.26} &	\textbf{58.19} &	\textbf{77.84} \\
\midrule
Mean & 56.84 & 86.70 & 79.36 & 39.44 & 30.99 & 74.99 & 47.35 & 48.90& \hallucinatepanel{58.07}\\
\bottomrule
\end{tabular}
\end{table}


\subsection{VLMs cannot reliably count line intersections \lineintersectlogo}
\label{sec:res-2line}

\subsec{Experiment}
We parse every model's response to extract the final answer and then compare it to the groundtruth.  
We report the mean accuracy of every model on two prompts and analyze how accuracy changes as we vary hyperparameters (\eg, line widths and image sizes).

\begin{figure}[ht]
    \centering
    \begin{minipage}[t]{0.43\textwidth}
        \centering
        \captionof{table}{The accuracy breakdown by line width in pixels (where $C$ = image width), averaged over two prompts, shows that VLMs cannot reliably count the intersections between two simple 2D line plots.}
        \label{tab:consolidated_model_performance_thickness_average}
        \resizebox{\textwidth}{!}{%
            \begin{tabular}{lcccc}
                \toprule
                \multicolumn{1}{l}{{Line width}} & \multicolumn{1}{c}{~~\gptlogo} & \multicolumn{1}{c}{~~\geminilogo} & \multicolumn{1}{c}{~~\sonnetlogo} & \multicolumn{1}{c}{~~\newsonnetlogo} \\                \midrule
                $0.005\times C$ & ~~45.00 & ~~67.55 & ~~45.22 & ~~75.83\\
                $0.010\times C$ & ~~38.22 & ~~66.33 & ~~41.61 & ~~74.88\\
                \midrule
                Mean & ~~41.61 & ~~66.94 & ~~43.41 & ~~\textbf{75.36}\\
                \bottomrule
            \end{tabular}
        }
    \end{minipage}
    \hfill
    \begin{minipage}[t]{0.49\textwidth}
         \vspace*{\fill}
        \centering
        \begin{tabular}{cc}
            \raisebox{-0.8\height}\gptlogo & \multirow{4}{*}{\includegraphics[width=\textwidth]{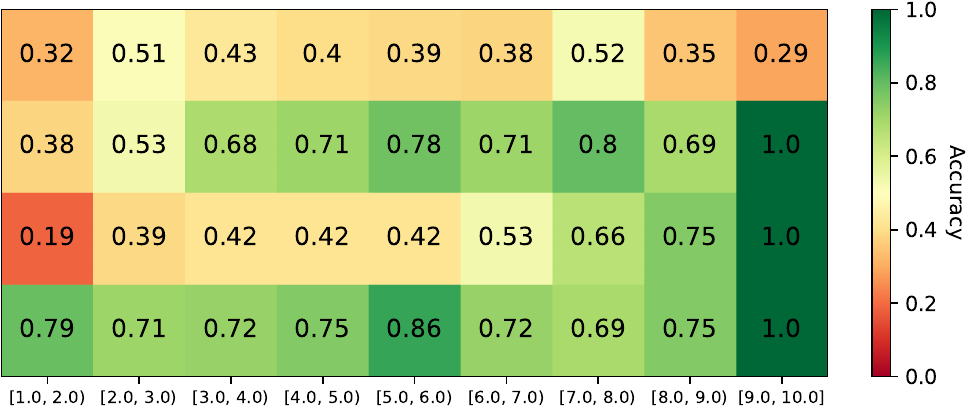}}\\
            \raisebox{-1.2\height}\geminilogo &\\
            \raisebox{-.9\height}\sonnetlogo &\\
            \raisebox{-.9\height}\newsonnetlogo &\\
        \end{tabular}
        \vspace{1em}  
        \captionof{figure}{
        As a line-plot image is divided into a 12$\times$12 grid, the x-axis shows the mean \textbf{distance} (in grid cells) over 3 pairs of points of two 2-segment plots.
        VLMs are often more confused when two plots are closer together (left) than when they are further apart (right).
        }
        \label{fig:line_intersection_distance_heatmap} 
    \end{minipage}
\end{figure}

\subsecspace{Results} 
First, across two prompts and two line widths, all VLMs are 56.84\% accurate (\cref{tab:all_model_acc}a), far from the expected 100\% accuracy on this easy task (\cref{appfig:lineintersection_prompt2}).
The best accuracy is only 75.36\% (\newsonnet)
(\cref{tab:consolidated_model_performance_thickness_average}).
Specifically, \textbf{VLMs tend to perform worse when the distance between two plots narrows} (\cref{fig:line_intersection_distance_heatmap}).
As each line plot is composed of three key points, the distance between two plots is computed as the mean distance over three corresponding point pairs.
See \cref{fig:teaser,app:2dline-qual} for more samples of VLM predictions.
VLMs perform similarly across three image sizes (\cref{app:2dline-thickness}).

Our findings are in stark contrast to the high accuracy of VLMs on ChartQA \cite{reid2024gemini,gpt4o}, suggesting that VLMs can recognize the overall trend of a line plot but unable to ``zoom in'' to extract details, \eg, seeing which lines are intersecting.

\subsection{VLMs cannot clearly see if two circles overlap or not \twocirclelogo}
\label{sec:res-2circles}

Motivated by VLM poor performance in counting line intersections (\cref{sec:res-2line}), here, we replace lines by large, filled circles and ask VLMs explicitly if the two circles are touching (or overlapping).

\subsec{Experiment}
Since we instruct VLMs to output a binary answer (Yes/No), we use Python to extract VLMs' formatted answer from their responses for comparing with groundtruth.

\begin{figure}[h]
    \centering
    \begin{minipage}[t]{0.45\textwidth}
        \centering
        \captionof{table}{\gptlogo and \geminilogo perform more consistently over the two different prompts (``\emph{overlapping}'' and ``\emph{touching}'') than \sonnetlogo and \newsonnetlogo.}
        \label{tab:overall_acc}
        \resizebox{\textwidth}{!}{%
            \begin{tabular}{lccc}
\toprule
Model & ~~Overlapping &~~Touching &~~Mean \\
\midrule
\raisebox{-.1\height}\gptlogo \gpt           & ~~74.74           & ~~77.08          & ~~75.91          \\
\raisebox{-.1\height}\geminilogo \gemini        & ~~\textbf{94.01}  & ~~93.23          & ~~\textbf{93.62} \\
\raisebox{-.1\height}\sonnetlogo \sonnet       & ~~89.58           & ~~83.33          & ~~86.46          \\
\raisebox{-.2\height}\newsonnetlogo \newsonnet    & ~~86.46           & ~~\textbf{95.18}  &~~90.82           \\
\bottomrule
\end{tabular}
        }
    \end{minipage}
    \hfill
    \begin{minipage}[t]{0.53\textwidth}
         \vspace*{\fill}
        \centering
            \begin{tabular}{cc}
            &\hspace{-1.1cm}\begin{tabular}{p{0.1cm}p{0.2cm}p{0.4cm}p{0.5cm}p{0.2cm}p{0.1cm}p{0.2cm}p{0.5cm}p{0.4cm}p{0.2cm}p{0.1cm}p{0.2cm}p{0.4cm}p{0.35cm}p{0.4cm}p{0.1cm}}
                 \includegraphics[trim={50 130 50 50},clip,width=0.1\textwidth]{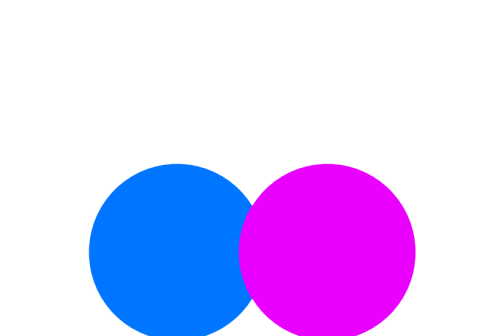}&&&&&\includegraphics[trim={50 130 50 50},clip,width=0.1\textwidth]{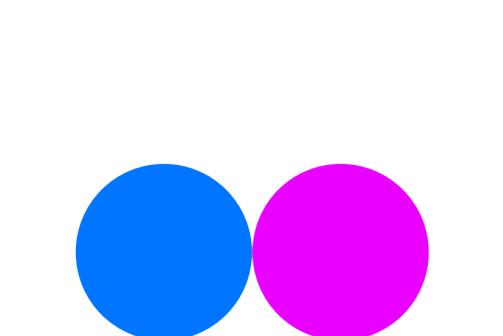}&&&&&\includegraphics[trim={50 130 50 50},clip,width=0.1\textwidth]{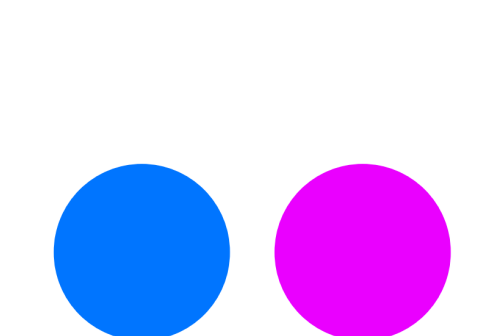}&&&&&\includegraphics[trim={50 130 50 50},clip,width=0.1\textwidth]{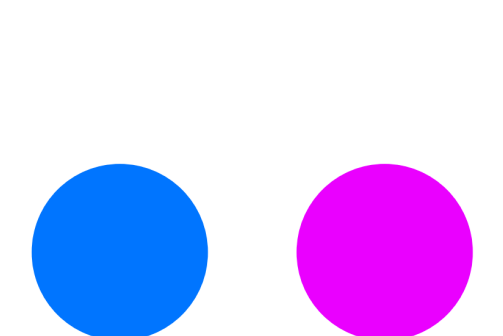}\\
            \end{tabular}\\
    \raisebox{-.35\height}\gptlogo & \multirow{4}{*}{\includegraphics[width=\textwidth]{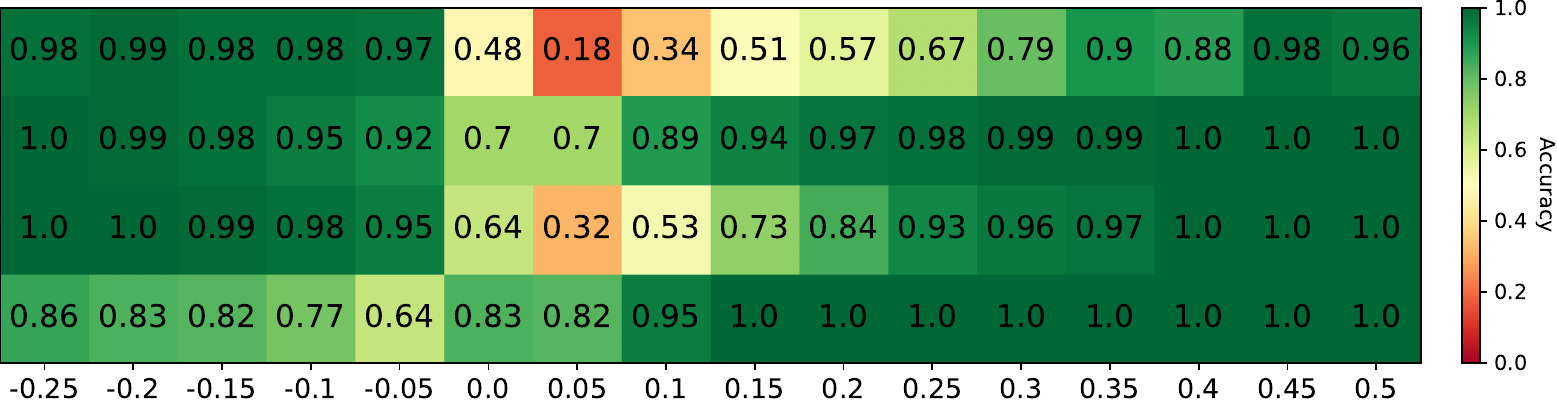}}\\
    \raisebox{-.28\height}\geminilogo  &\\
     \raisebox{-.26\height}\sonnetlogo &\\
      \raisebox{-.28\height}\newsonnetlogo &\\
    \end{tabular}
        \vspace{1em}  
        \captionof{figure}{VLMs perform poorly when two circles are tangent ($d = 0.0$) or close together ($d = 0.05, 0.1$).     
        Yet, \newsonnetlogo is better at $d \geq 0.0$. 
        (perhaps due to its tendency to answer ``{No}'').
        }
        \label{fig:heat-2cricles}
    \end{minipage}
\end{figure}

\subsec{Results}
Surprisingly, even on this task where objects (\purplecircle\bluecircle\hspace{-0.10cm}) are large and clearly visible to humans, no VLMs are able to solve it perfectly---their mean accuracy is 86.70\% (\cref{tab:all_model_acc}b).
The best accuracy is 93.62\% (\gemini) over all images and two prompts (\cref{tab:overall_acc}).
A common trend is \textbf{when two circles are closer together, VLMs tend to perform more poorly}, making educated guesses, \eg, \newsonnet often answers ``{No}'' conservatively (\cref{fig:qual_overlap}).
\gpt performs the worst and shockingly is \emph{not} 100\% accurate even when the distance between two circles is as large as one radius (\cref{fig:heat-2cricles}; $d = 0.5$).
That is, consistent with the results from \cref{sec:res-2line}, VLMs seem to be unable to always detect the gap or intersection between two filled circles (\cref{fig:teaser,fig:qual_overlap}).

An explanation is that due to the late-fusion mechanism \cite{team2024chameleon}, VLMs extract visual features from the image \emph{before} even looking at the question, causing this ``blindness''.
In contrast, if a model first knows that the question asks it to focus on the area between the two circles, it then might be able to extract accurate visual information to answer such simple questions.

While VLMs perform similarly across three image resolutions (\cref{app:2circle-resolution}), every model performs the best at a specific circle orientation (\cref{app:2circle-rotation}). Moreover, VLMs' performance does not change substantially ($\pm 5.79$ points for \newsonnet and $\pm$10.81 for \gpt) when tested against different colors (\cref{app:color-circles}), ruling out the impact of color on their performance on the task.
More examples of VLMs' answers are in \cref{app:2circle-qual}.

\subsection{VLMs do not always see the letter inside the red circle \protect\smallredcircle}
\label{sec:res-redoval}

\subsec{Experiment}
To evaluate the models' ability to recognize individual characters in an image, we place a red circle over one character in a word. 
We prompt VLMs to put their prediction in \{curly braces\} and then we compare the lowercase version of this character to the lowercase version of the groundtruth character.

\begin{figure}[ht]
\centering
\begin{AIbox}{Which character is being highlighted with a red oval? Please provide your answer in curly brackets{,} e.g. \{a\}}
\centering

    \begin{tabular}{lp{0.8cm}cp{0.8cm}cp{0.8cm}cp{0.8cm}cp{0.8cm}cp{0.8cm}c} 
     &\multicolumn{2}{c}{\includegraphics[width=0.15\textwidth]{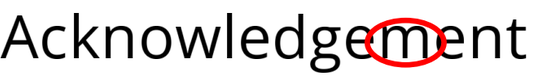}} & 
     \multicolumn{2}{c}{\includegraphics[width=0.15\textwidth]{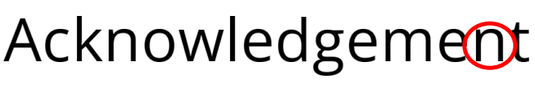}} &
     \multicolumn{2}{c}{\includegraphics[width=0.15\textwidth]{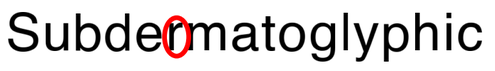}} &
     \multicolumn{2}{c}{\includegraphics[width=0.15\textwidth]{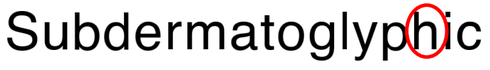}} &
     \multicolumn{2}{c}{\includegraphics[width=0.15\textwidth]{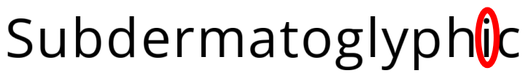}}&
     \multicolumn{2}{c}{\includegraphics[width=0.15\textwidth]{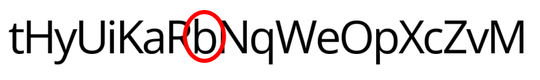}}\\ 
     \rowcolor{lightgray}
     \raisebox{-0.2\height}\gptlogo & \centering e & \textcolor{red}{\xmark} &  \centering t& \textcolor{red}{\xmark}&  \centering o& \textcolor{red}{\xmark} &  \centering h & \textcolor{ForestGreen}{\cmark} &  \centering o & \textcolor{red}{\xmark}&  \centering  b & \textcolor{ForestGreen}{\cmark}\\ 
     
     \raisebox{-0.2\height}\geminilogo & \centering m& \textcolor{ForestGreen}{\cmark} & \centering e& \textcolor{red}{\xmark}&  \centering e& \textcolor{red}{\xmark} &  \centering h& \textcolor{ForestGreen}{\cmark} &  \centering o & \textcolor{red}{\xmark}&  \centering b & \textcolor{ForestGreen}{\cmark}\\ 
     
     \rowcolor{lightgray}
     \raisebox{-0.2\height}\sonnetlogo & \centering e& \textcolor{red}{\xmark} & \centering t& \textcolor{red}{\xmark}&  \centering o& \textcolor{red}{\xmark} &  \centering i& \textcolor{red}{\xmark} &  \centering c& \textcolor{red}{\xmark}&  \centering n & \textcolor{red}{\xmark}\\ 
     
       \raisebox{-0.2\height}\newsonnetlogo & \centering e& \textcolor{red}{\xmark} & \centering e & \textcolor{red}{\xmark}&  \centering r& \textcolor{ForestGreen}{\cmark} &  \centering i& \textcolor{red}{\xmark} &  \centering t & \textcolor{red}{\xmark}&  \centering D & \textcolor{red}{\xmark}\\ 

     \end{tabular}

    \vspace{4pt}
    \centering
    \begin{tabular}{cccccccccccccc}
    \raisebox{-0.1\height}\gptlogo & \gpt &&&
      \raisebox{-0.1\height}\geminilogo & \gemini & &&\raisebox{-0.1\height}\sonnetlogo & \sonnet & && \raisebox{-0.12\height}\newsonnetlogo & \newsonnet \\
      \end{tabular}
     
\end{AIbox}
\caption{
Identifying the letter being circled is non-trivial for VLMs across both English words ({\Acknowledgement} \& \SubdermatoglyphicCaption) and a random string ({\tHyUiKaRbNqWeOpXcZvM}).
When making mistakes, VLMs tend to predict letters adjacent to the one being circled.
}
\label{appfig:circled_word_prompt1}
\end{figure}

\subsec{Results}
All VLMs can accurately spell out the string when there is a red oval \smallredcircle ~superimposed on the image.
Yet, interestingly, reading out which letter is being circled turns out to be a challenge (mean model accuracy: 79.7\%; \cref{tab:all_model_acc}c).

\textbf{When the letters are close together, VLMs often predict letters adjacent to the one being circled} (see the confusion matrix in \cref{fig:confusion-main} and more results in \cref{fig:teaser,app:redoval_adjacent}).
Sometimes models hallucinate, \eg, coming up with characters non-existent in \Subdermatoglyphic (\eg, ``{9}'', ``n'', ``\copyright'') despite having the ability to accurately spell out the word (see \cref{fig:confusion-main}).
We also observe that VLMs, on average, fail to see the circled letter across various common English words (mean accuracy is 86.43\% in \cref{app:english_words}). However, as the words get shorter in length and there is no repetitive letters in them, VLMs tend to perform better.
More failure cases are reported in \cref{app:redoval-confuses_oval,app:redoval_qual}.

On average, models perform better (\increasenoparent{0.46} to \increasenoparent{13} points) on the two English words compared to the random string (\cref{tab:circledwords_combined_average}), suggesting that \textbf{familiarity with the word help VLMs make better educated guesses}, slightly improving accuracy.

\newsonnet and \gemini are the top-2 models (87.88\% and 83.29\%) and are better than \gpt and \sonnet by a large margin of nearly \increasenoparent{15} points (\cref{tab:circledwords_combined_average}).
VLMs perform similarly across two prompts (\cref{app:redoval_prompt}) and two font families (\cref{app:redoval_font}). See also \cref{fig:hallucinate} for an example of \gpt and \gemini making educated guesses on the color of the overlapping area between two overlapping circles (Task 2).

\subsection{VLMs struggle to count overlapped and nested shapes}
\label{sec:res-count-shapes}

\subsec{Experiment}
We run all VLMs on images of overlapping shapes (\BigCircle and \BigPentagon) (\cref{sec:olympic_logo}) and nested squares (\cref{sec:count_nested_squares}).
We prompt VLMs to output the predicted number of shapes in a formatted answer (which is extracted and compared with groundtruth).
For each shape (\BigCircle, \BigPentagon or \BigSquare), we run two different prompts.

\subsec{Results}
On counting overlapping circles, pentagons, and nested squares, VLM mean accuracy is 39.44\%, 30.99\%, and 74.99\%, respectively (\cref{tab:all_model_acc}d--f).
That is, counting shapes is not easy to models regardless of whether the shapes are overlapped or nested (\cref{fig:teaser,appfig:circles_p1}).
On nested squares, model accuracies vary widely---\gpt (55.83\%) and \sonnet (65.00\%) are at least \decreasenoparent{25} points behind \gemini (87.08\%) and \newsonnet (92.08\%).
This gap is even larger on overlapped circles and pentagons---\newsonnet outperforms other models by multiple times (\eg, 77.71\% vs. 1.87\% of \sonnet; \cref{tab:all_model_acc}).

\begin{figure}[H]
\centering
\resizebox{0.75\columnwidth}{!}{
\begin{tabular}{ccccc}
    & (a) Circles \olympiclogo & (b) Pentagons \pentagonlogo & (c) Squares \nestedsquareslogo\\    
     \raisebox{-1\height}\gptlogo & \multirow{4}{*}{\includegraphics[width=0.3\textwidth]{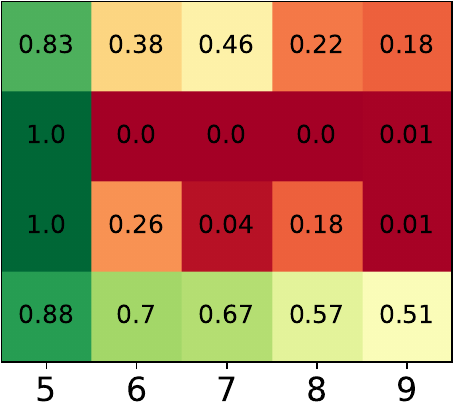}} & \multirow{4}{*}{\includegraphics[width=0.3\textwidth]{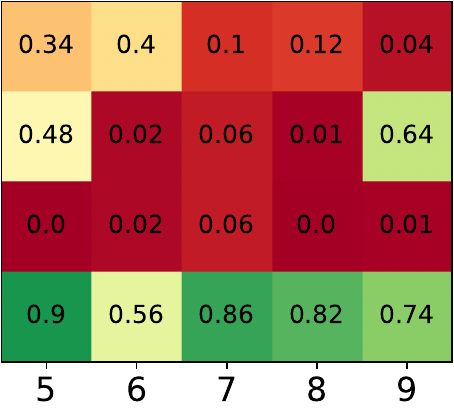}} & \multirow{4}{*}{\includegraphics[width=0.241\textwidth]{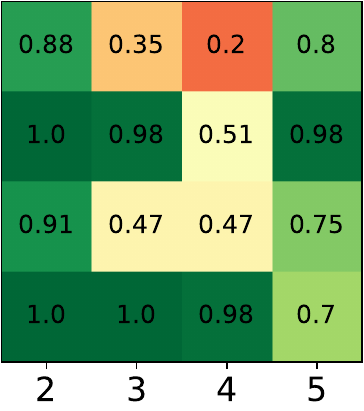}} & \multirow{4}{*}{\raisebox{-0.1\height}{\includegraphics[width=0.045\textwidth]{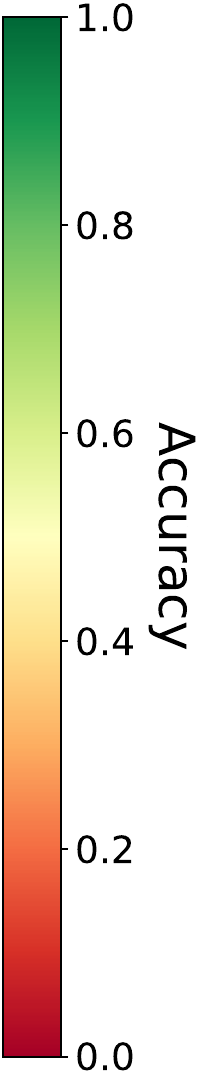}}}\\
     \raisebox{-1.6\height}\geminilogo  &&&&\\
     \raisebox{-1.6\height}\sonnetlogo &&&&\\
      \raisebox{-1.45\height}\newsonnetlogo &&&&\\
      &&&&\\&&&&\\
      \multicolumn{5}{c}{Number of shapes}
\end{tabular}}
\caption{All four VLMs can count 5 circles well (leftmost; {\sethlcolor{green83}\hl{0.83}}), but only \newsonnetlogo can count 5 overlapping pentagons well at 0.9 accuracy (b). 
Counting from 6--9 shapes (either \BigCircle~or \BigPentagon) is {\sethlcolor{red!60}\hl{challenging}} to VLMs.
}
\label{fig:olymp-heatmap}
\end{figure}

All four models are at least 83\% accurate in counting 5 circles.
Yet, surprisingly, \textbf{increasing the number of circles by only one causes accuracy to dip substantially to near zero} for all models, except \newsonnet (\cref{fig:olymp-heatmap}; column 6--9).
In counting pentagons, all VLMs (except \newsonnet) perform poorly even at 5 pentagons.
Overall, \textbf{counting from 6 to 9 shapes (both circles and pentagons) is hard for all models}.

Why are VLMs nearly perfect at counting 5 circles (\cref{fig:olymp-heatmap}a), but struggle to count 5 pentagons or more than 5 shapes?
When there are more than 5 circles (\BigCircle) and VLMs predict an incorrect count, \gemini predicts ``5'' 99.74\% of the time regardless of the actual number of circles (\cref{tab:biased-pred}).
For other models, this frequency is also much higher than that in the case of pentagons.
Our results show strong evidence for an explanation that \textbf{VLMs are biased towards the well-known 5-circle Olympic logo} (more results on this bias in \cref{app:bias-gemini-fig}).

Note that there are only 2 to 5 squares in each image in the task of counting nested squares, and these squares do not intersect (\cref{fig:nested_squares}).
Surprisingly, \gpt and \sonnet are still unable to perfectly count two and three nested squares (\cref{fig:olymp-heatmap}c).
When the count increases to four and five, most models are far from 100\% accurate (\cref{fig:olymp-heatmap}c).

\gpt performs better on colored shapes than on black shapes, and \newsonnet is increasingly better as the image size increases.
However, the accuracy of the three other models only changes marginally as colors (\cref{app:color-olympic}) and image resolutions (\cref{app:resolu-olympic}) change.

\subsection{VLMs cannot reliably count rows and columns in a grid \gridlogo}
\label{sec:res-gridcount}

Since VLMs struggle with counting simple shapes when the shape edges intersect (\cref{sec:olympic_logo}) or separate (\cref{sec:count_nested_squares}), here, we test the remaining case where these shapes are adjacent and share edges. 
That is, multiple squares tile up into a single grid.
Given the impressive accuracy of VLMs \cite{gpt4o,reid2024gemini,sonnet35} on questions involving tables and spreadsheets in DocVQA \cite{mathew2021docvqa}, we hypothesize that VLMs must be able to count rows and columns of a grid.

\subsec{Experiment}
We run all four VLMs on all images of empty grids and text-containing grids (\cref{sec:count_grid}) and analyze their formatted answers.

\subsec{Results}
VLMs surprisingly perform poorly (34.37\% accuracy) in counting rows and columns in an \textbf{empty} grid (see \cref{tab:overall_acc_grid}).
Specifically, they are often off by one or two (\eg, \gpt predicts 4$\times$4 and \gemini predicts 5$\times$5 for a 4$\times$5 grid; \cref{fig:qual_grid,fig:teaser}).
This finding suggests that VLMs can extract important content from a table to answer table-related questions in DocVQA \cite{mathew2021docvqa} but do not clearly ``see'' a table cell-by-cell as a human does.

\begin{figure}[t]
\centering
\begin{tabular}{cc}
\centering
\begin{tabular}{cc}

\begin{tabular}{cccc}

&(a) Empty& (b) Text-containing \\
\raisebox{-0.4\height}\gptlogo & \multirow{4}{*}{\includegraphics[width=0.3\textwidth]{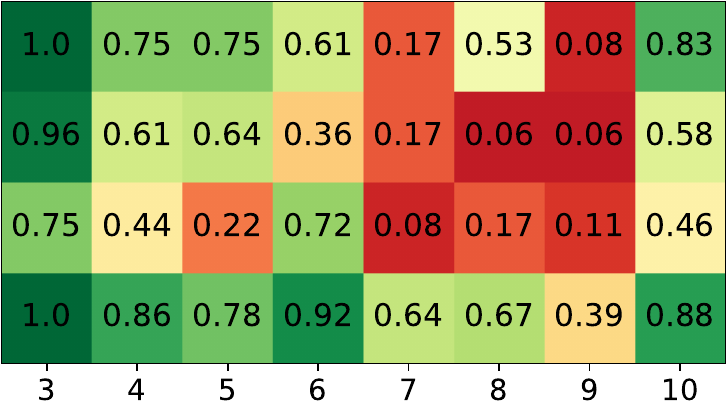}} & \multirow{4}{*}{\includegraphics[width=0.3\textwidth]{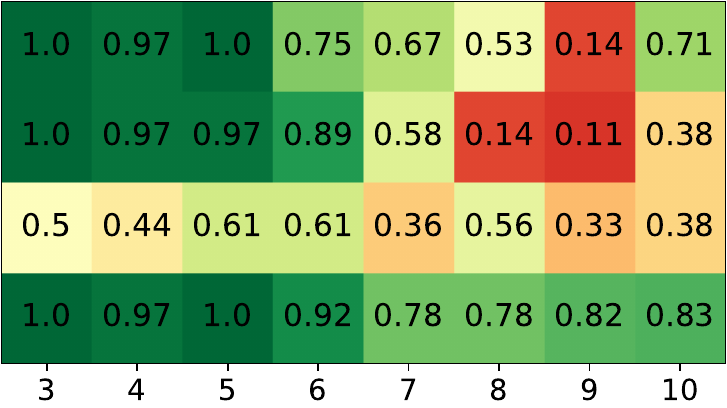}}\\
 \raisebox{-0.6\height}\geminilogo  &&\\
 \raisebox{-0.65\height}\sonnetlogo &&\\
  \raisebox{-0.65\height}\newsonnetlogo &&\\
 
\end{tabular}  & \rotatebox{-90}{\hspace{-0.6cm} (c) Rows}\\

\\

\begin{tabular}{cccc}

\raisebox{-0.4\height}\gptlogo & \multirow{4}{*}{\includegraphics[width=0.3\textwidth]{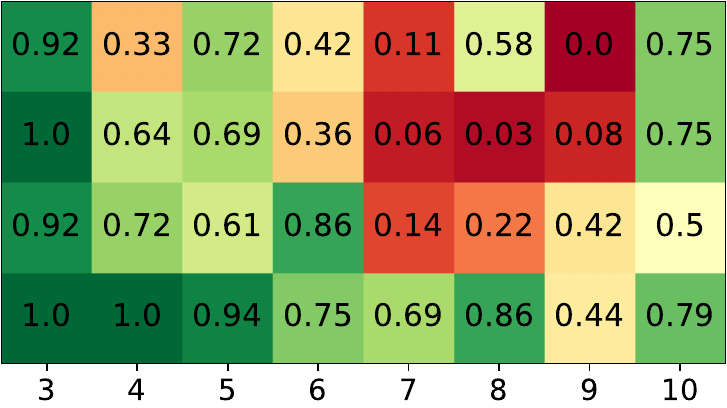}} & \multirow{4}{*}{\includegraphics[width=0.3\textwidth]{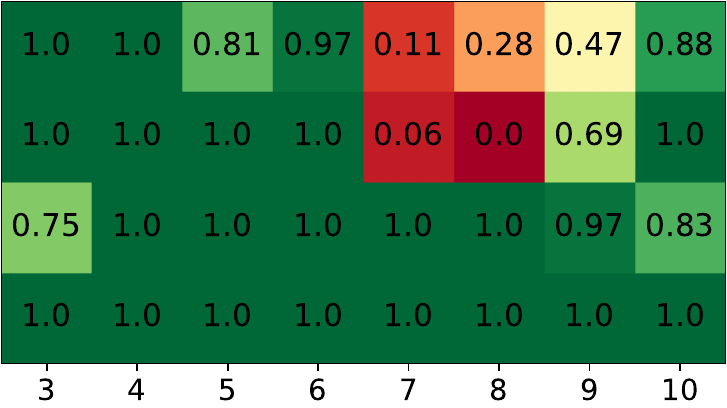}}\\
 \raisebox{-0.6\height}\geminilogo  &&\\
 \raisebox{-0.65\height}\sonnetlogo &&\\
  \raisebox{-0.65\height}\newsonnetlogo &&\\
 
\end{tabular} & \rotatebox{-90}{\hspace{-1cm} (d) Columns} \\

\end{tabular} & \raisebox{-0.55\height}{\includegraphics[width=0.05\textwidth]{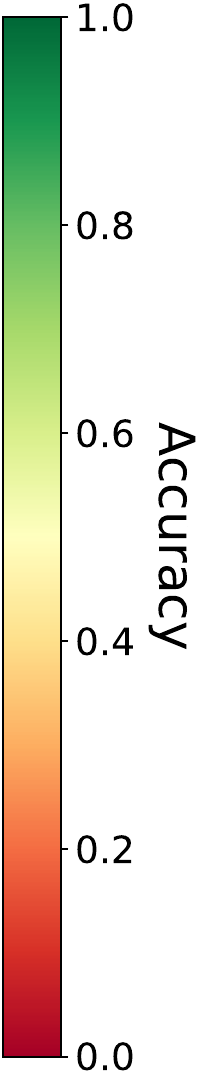}}\\
\end{tabular}
\caption{
Accuracy of counting rows (c) vs. columns (d) (here, analyzed separately) when the grids are empty (a) vs. contain text (b).
VLMs (especially, \geminilogo~and \sonnetlogo) generally count much more accurately when grids contain text vs. empty grids (b vs. a).
Interestingly, columns are also easier for VLMs to count than rows (d vs. c).
}
\label{fig:grid_rc}
\end{figure}

This might be because tables in documents are often non-empty, which are more familiar to VLMs.
Aligned with that hypothesis, after \textbf{adding a single word to each cell, we observe the accuracy of all VLMs to almost double} (\eg, from 26.13\% to 53.03\% for \gpt) (\cref{tab:overall_acc_grid}).
Yet, no models can solve this task with the best model (\newsonnet) performing at 88.68\% on text-containing grids and 59.84\% on empty grids (\cref{fig:grid_rc}a vs. b).

Interestingly, \textbf{VLMs are better at counting columns than rows}---70.53\% vs. 60.83\% accuracy (\cref{fig:grid_rc}c vs. d).
However, these numbers are still far from 100\% showing that VLMs currently cannot count neither rows or columns in a table reliably.
See \cref{app:grids-row,app:grids-qual} for more results.

\subsection{VLMs struggle to count single-colored paths \scalebox{0.7}{\pathfollowinglogo}}
\label{sec:res-paths}

This path-counting task tests a VLM's ability in recognizing a path of a unique color and \emph{trace} it from a given starting station to the destination, an important task in reading maps and graphs in general \cite{masry2022chartqa}. 

\subsec{Experiment}
From a subway map (\cref{sec:subway}), we randomly sample 2 connected stations and prompt every model to count the single-colored paths that connect them. 
We extract numbers from VLM templated responses and compare them with the groundtruth.

\subsec{Results}
Overall, VLMs perform poorly at a mean accuracy of 48.90\% (\cref{tab:all_model_acc}h).
Even when there is only \emph{one path} between two stations, no models can reach 100\% accuracy (the best is \newsonnet at 93.33\% and the worst is 20\%; \cref{fig:overall_acc_subway}).
VLM predicted counts are often off by 1 to 3 paths (\cref{fig:qual_subway3}).
VLM accuracy reduces substantially, \eg, \newsonnet from 93.33\% to 58.33\% and 22.91\% as the complexity of the maps increases from 1, 2 to 3 paths, respectively (\cref{fig:overall_acc_subway}).
More samples of VLM responses are in \cref{fig:teaser,app:subway-qual}.

\subsection{Long-inference, slow-thinking VLMs perform similarly to regular VLMs on \blindtest}
\label{sec:eval_reasoning}

From math to coding, spending more time thinking before responding enables LLMs to perform substantially better in many tasks \cite{jaech2024openai}.
Here, we aim to test whether such slow thinking also enables VLMs to perform better on \blindtest where we argue that reasoning in the text space might not help.
We perform this test for two SOTA models by comparing them with their corresponding slow-thinking versions.


\subsec{Experiment} We run 2 SOTA slow-thinking VLMs: \geminithinkfull and \qvq on \blindtest, and compare them with their non-thinking, regular versions, \ie, \geminiflashfull and \qwen.

\begin{table}[h]
    \centering
    \caption{SOTA \sethlcolor{MyLightYellow}\hl{slow-thinking} models (bottom) perform even worse than their regular counterpart (top) \blindtest, showing that the longer inference has no positive impact on \blindtest tasks.
    \qvq is the slow-thinking counterpart of \qwen.
    }
    \label{tab:reasoning_models}
    \resizebox{\columnwidth}{!}{
    \begin{tabular}{lrccccccccc}
    Model & Size & ~~\lineintersectlogo & \twocirclelogo & \smallredcircle & \olympiclogo & \pentagonlogo  & \nestedsquareslogo & \gridlogo & \pathfollowinglogo & Task mean\\
    \midrule    
    \geminiflashfull & \na & ~~\textbf{85.44} & 80.95 & \textbf{81.73} & \textbf{55.62} & 44.16 & \textbf{96.25} & 66.85 & \textbf{70.97} & \textbf{72.75}\\
    \qwen & 72B & ~~64.97 & 76.41 & 73.56 & 28.12 & 35.42 & 74.58 & 20.64 & 59.31 & 54.13\\
    \midrule
         \rowcolor{MyLightYellow}
            \geminithinkfull & \na &  ~~77.50 & \textbf{88.24} & 74.03 & 47.92 & \textbf{57.08} & 93.75 & \textbf{70.45} & 63.75 & 71.59\\
            \rowcolor{MyLightYellow}
            \qvq & 72B & ~~37.05 & 67.26 & 51.60 & 29.58 & 26.66 & 53.75 & 36.74 & 37.22 & 42.48\\
            \bottomrule
    \end{tabular}}
\end{table}

\begin{figure}[h]
\centering
\resizebox{0.95\columnwidth}{!}{
\begin{AIbox}{Reasoning on counting the overlapping circles}
\centering
\includegraphics[trim={40 90 30 90},clip,width=0.7\textwidth]{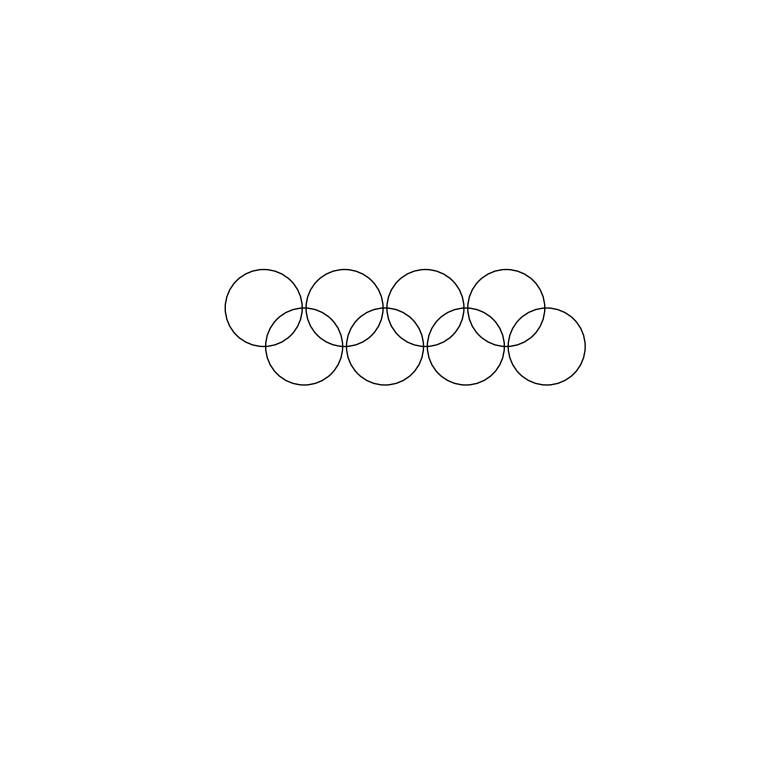}
\vspace{-2cm}
\begin{flushleft}
\textcolor{Orange}{\textbf{Prompt:}} Count the circles in the image. Answer with a number in curly brackets e.g. \{3\}.
\newline
\textcolor{ForestGreen}{\textbf{GT:}} 8
\end{flushleft}

\begin{tabular}{p{11cm}}
    \midrule
     \geminiflashfull
     \newline
     \newline
     \textcolor{Blue}{\textbf{Response:}} \{7\} \redxmark\\
    \midrule
    \geminithinkfull 
    \newline
    \newline
    \textcolor{Purple}{\textbf{Thoughts:}}
    The user wants me to count the circles in the image.
    I can see 7 circles in the image.
    I will count them again to be sure.
    Yes, there are 7 circles.
    I will answer with {7}.
    \newline
    \textcolor{Blue}{\textbf{Response:}} \{7\} \redxmark
     \\     
\end{tabular}
\end{AIbox}}
\caption{\geminithinkfull cannot count the circles in the image even during its thought process and responds similarly to its non-thinking version (\geminiflashfull).}
\label{fig:main-think}
\end{figure}

\begin{figure}[h]
\centering
\resizebox{0.9\columnwidth}{!}{
\begin{AIbox}{Reasoning on counting the single-color paths}
\centering
\includegraphics[width=0.3\textwidth]{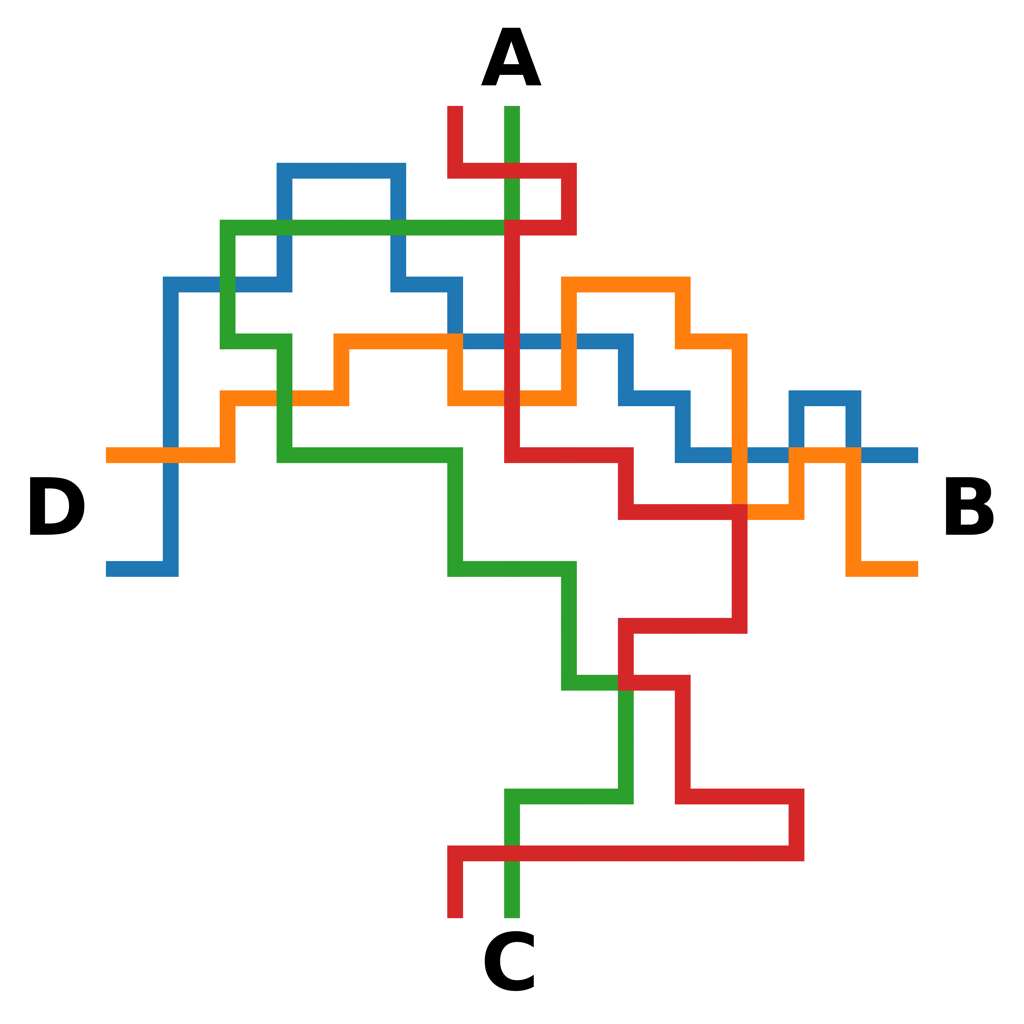}
\begin{flushleft}
\textcolor{Orange}{\textbf{Prompt:}} How many single-color paths go from C to A? Answer with a number in curly brackets e.g. \{3\}.
\newline
\textcolor{ForestGreen}{\textbf{GT:}} 2
\end{flushleft}

\begin{tabular}{p{11cm}}
    \midrule
     \geminiflashfull
     \\
     \\
     \textcolor{Blue}{\textbf{Response:}} \{1\} \redxmark \\
    \midrule
    \geminithinkfull 
    \\
    \\
    \textcolor{Purple}{\textbf{Thoughts:}}
    The user is asking to count the number of single-color paths from point C to point A in the given image.
I need to examine the image to identify the paths and count them.
The image contains four lines of different colors: red, green, blue and orange.
I need to find paths that start at C and end at A and are of a single color.
Let's trace each color starting from C.
Point C is at the bottom.
The line starting from C is red. Let's trace the red line.
The red line goes up, then right, then up again, then left, then up and finally reaches A.
So there is one red path from C to A.
Now let's check for other colored paths.
Point C is only connected to the red line.
Therefore, there are no other single-color paths from C to A other than the red path.
So the number of single-color paths from C to A is 1.
    \newline
    \textcolor{Blue}{\textbf{Response:}} \{1\} \redxmark
\end{tabular}
     
\end{AIbox}}
\caption{While \geminithinkfull makes an effort to follow the lines in the subway-like map, it fails to see all the single-color lines connected to point C, providing the same incorrect final answer as \geminiflashfull.}
\label{fig:main-think-subway}
\end{figure}

\subsec{Results} On average over 7 tasks, \geminithinkfull, is on par with its non-thinking counterpart, \geminiflashfull (\cref{tab:reasoning_models}; 71.59 vs 72.75\%). 
This shows that the ``slow-thinking'' capability (\ie, long, scaled-up inference) does not address the main challenge that \blindtest poses to VLMs.
Qualitatively examining the thinking tokens of \geminithinkfull shows that the hidden thoughts are in text space and have no benefits on \blindtest (see \cref{fig:main-think,fig:main-think-subway}). 
Moreover, \qvq, the SOTA open-source slow-thinking model is also \decreasenoparent{11.65} points behind its non-thinking counterpart, \qwen (\cref{tab:reasoning_models}). 

\section{VLMs can solve most tasks when there is more space between shapes and letters}
\label{sec:spacing}

Since none of the tested SOTA VLMs, including the slow-thinking ones, is capable of solving any task at near-100\% accuracy (\cref{sec:results}), there are two hypotheses to explain this result: (1) prompts may be suboptimal (no prompt-engineering performed);
(2) the questions are out-of-distribution for VLMs and they are not capable of solving them, \eg, counting to 10;
(3) the shapes being so close together that they are overlapping or nesting or adjacent causes challenges to VLMs in extracting accurate features (and therefore answering questions).

To tease apart hypotheses (1) \& (2) from (3), we increasingly add more space between letters and shapes to hopefully make the tasks easier (\eg, space between characters in \subdermatoglyphicCircled ~or between circles \olympiclogo) and report model-accuracy changes (\ie, how a VLM would reach near 100\% accuracy).



\subsection{VLMs perform better as the spacing between letters in the circled letter task increases}
\label{sec:spaced-words}

\subsec{Experiment}
On the circled-letter task \BigRedCircle, VLMs often cannot reliably tell which letter is being circled among adjacent ones (see \cref{fig:confusion-main}).
Here, we add 1 to 3 ASCII space characters between adjacent letters of a string and evaluate VLMs using the exact same prompts as done in \cref{sec:res-redoval}.

\begin{figure}[h]
    \centering
    \begin{tabular}{cc}
         \begin{tabular}{lc}
     \raisebox{1\height}{0 (baseline)} & \includegraphics[trim={0 300 0 300},clip,width=0.45\linewidth]{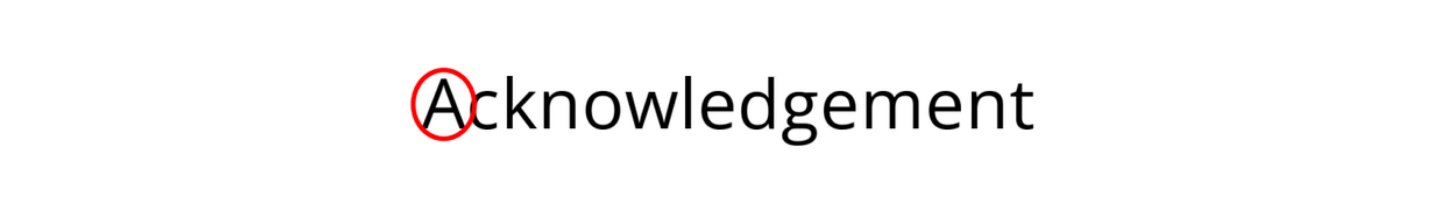} \\ 
     \raisebox{1.5\height}{1} & \includegraphics[trim={0 300 0 300},clip,width=0.45\linewidth]{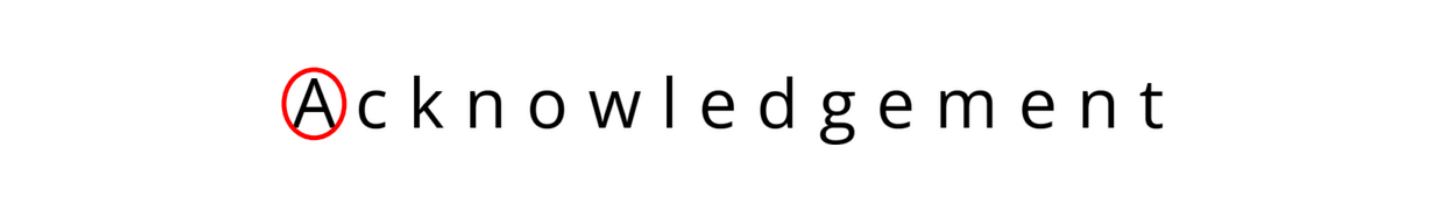}\\
     \raisebox{1.3\height}{2} &\includegraphics[trim={0 300 0 300},clip,width=0.45\linewidth]{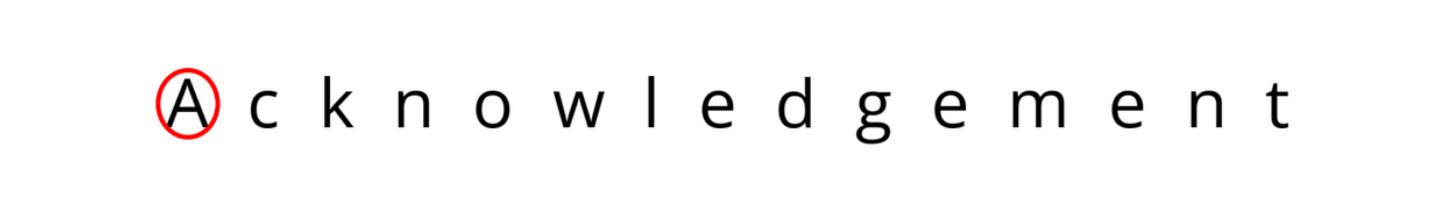}\\
     \raisebox{1.3\height}{3} &\includegraphics[trim={0 300 0 300},clip,width=0.45\linewidth]{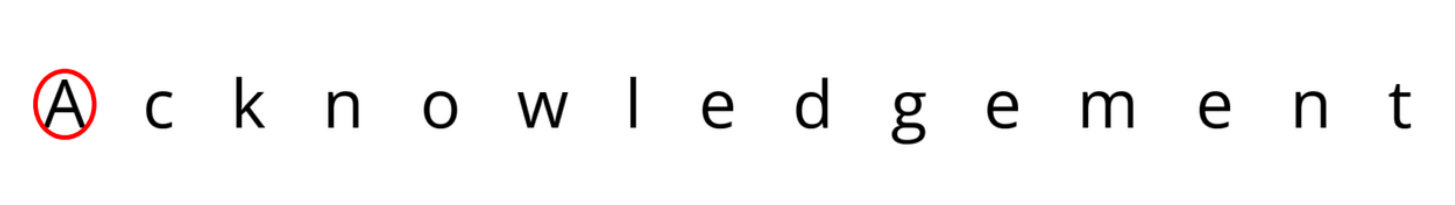}\\
     \# of spaces & example image\\
\end{tabular}
     & 
     \begin{tabular}{rc}
        \raisebox{-1\height}\gptlogo & \multirow{7}{*}{\includegraphics[width=0.25\textwidth]{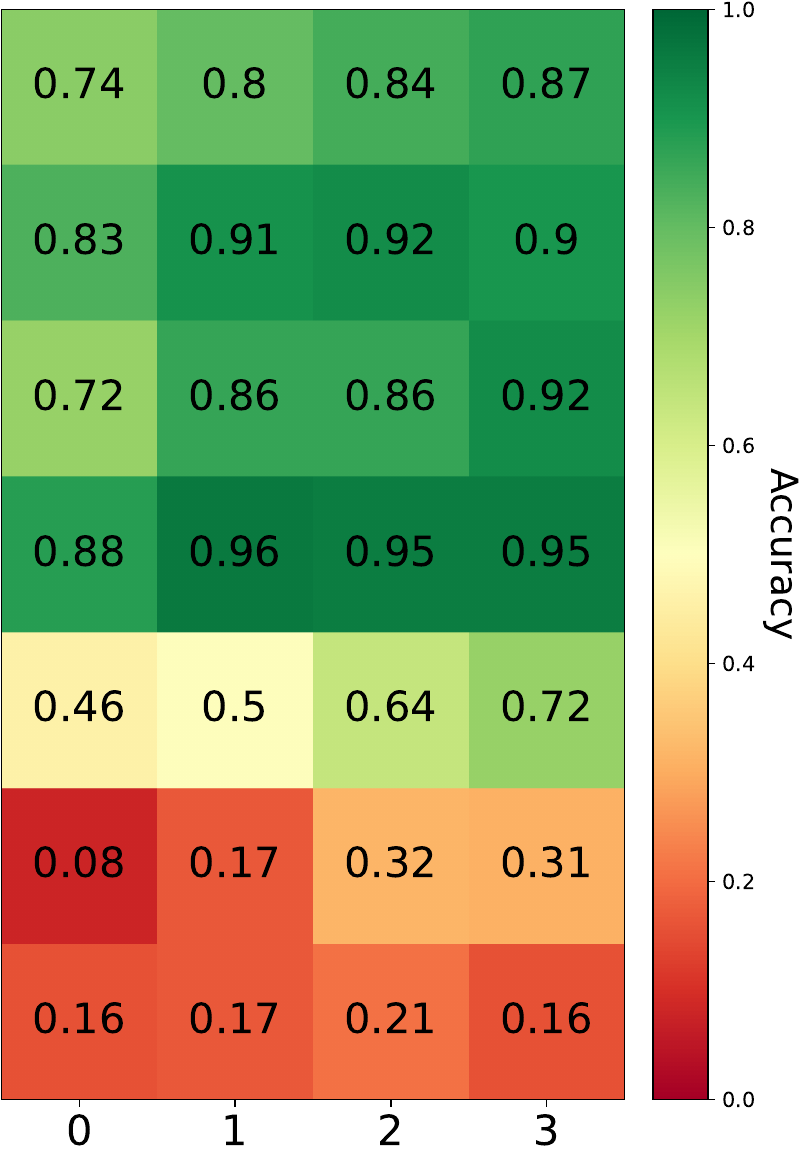}}\\
        \raisebox{-1\height}\geminilogo &\\
        \raisebox{-1.1\height}\sonnetlogo &\\
        \raisebox{-1.1\height}\newsonnetlogo &\\
        \raisebox{-.85\height}{\llavalogolarge}&\\
        \raisebox{-.85\height}{\llavalogosmall}&\\
        \raisebox{-1.\height}{\philogo} &\\
        \\
        & \small{Number of spaces}\\
    \end{tabular}
     \\
     \\
     (a)& (b) \\
    \end{tabular}
    \caption{(a) By adding more space $\in\{1, 2, 3\}$ to baseline images in the circled letter task, we create a simpler version of them. (b) VLMs generally perform better when there is $\geq$1 space between letters of the words ($\triangle$ for \sonnet and \gpt is \increasenoparent{21} and \increasenoparent{13} from 0 to 3 spaces, respectively).}
        \label{fig:spaced-words}
\end{figure}

\subsec{Results} 
\textbf{All VLMs} perform better on this task when there is $\geq$1 extra space between characters compared to when there is no extra space (\cref{fig:spaced-words}b).
However, the increase ($\triangle$) in accuracy varies across models.
For instance, \sonnetlogo and \llavalogolarge accuracies increase by over \increasenoparent{20} points to 92\% and 72\% from 72\% and 46\%, respectively (\cref{fig:spaced-words}b).

Out of 7 tested VLMs, \newsonnet reaches the highest accuracy of 95\% when there are 3 extra spaces between letters (\cref{fig:spaced-words}a).
Qualitatively, the remaining 5\% error (40 samples) includes: (1) 12 mispredictions of adjacent letters and (2) 13 instances of confusing the red circle as part of the letter, \eg, `@' for `a', and (3) 15 cases of predicting `g' instead of `q'.


Note that while the vision encoders in \llavaone with 72B (\llavalogolarge) and 0.5B (\llavalogosmall) are identical (400M SigLIP), the VLM with a larger \texttt{Qwen2} language decoder (72B) substantially outperforms the counterpart with smaller decoder (\cref{fig:spaced-words}b).
This interestingly suggests that, at least in \llavaone family, the language decoder plays a major role in ``reading out'' which letter is being circled.
On this task there is no slow-thinking \cite{kahneman2011thinking} required and therefore, the language model (\texttt{Qwen2}) may act as an extended ``vision'' encoder on top of the \texttt{SigLIP} vision encoder, bottlenecking the accuracy of the \llavaonefull-si-0.5B (\llavalogosmall~\llavaSI).

\subsection{VLMs can more accurately count disjoint shapes}
\label{sec:spaced-shapes-count}

Inspired by the results of adding spaces to the circled letter task (\cref{sec:spaced-words}), where VLMs see the letters better when there is more space between them, we also study the effects of the overlap area, \ie, the space between shapes, in counting overlapping shapes (for both \olympiclogo and \pentagonlogo). 
We aim to evaluate whether reducing the overlap area between shapes would improve VLM accuracy in counting them.

\begin{figure}[h]
\centering
\resizebox{\columnwidth}{!}{
\begin{tabular}{cccccccccc}
     \includegraphics[trim={140 190 165 165},clip,width=0.26\linewidth]{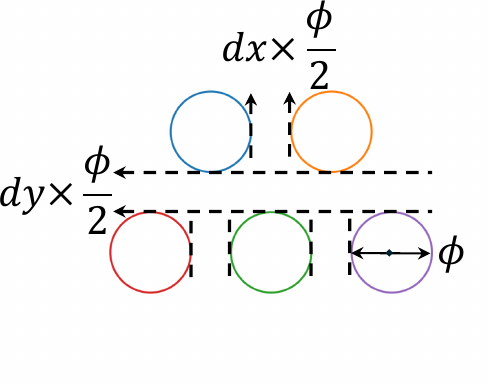}
     & 
     \includegraphics[trim={245 260 245 260},clip,width=0.18\linewidth]{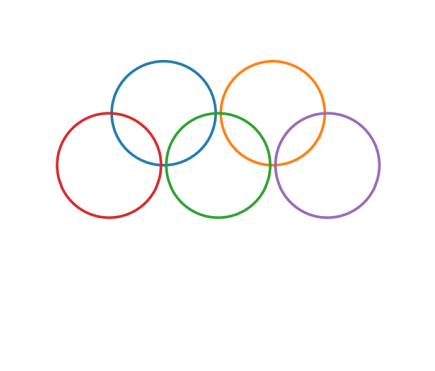}
    &
    \includegraphics[trim={245 260 245 260},clip,width=0.18\linewidth]{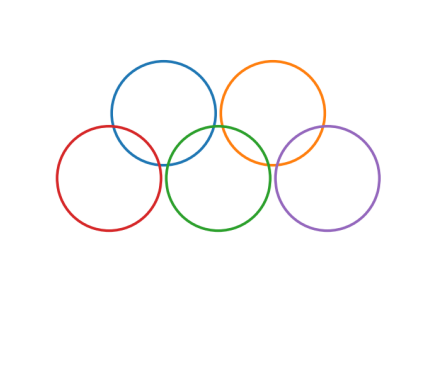}
    &
    \includegraphics[trim={245 260 245 260},clip,width=0.18\linewidth]{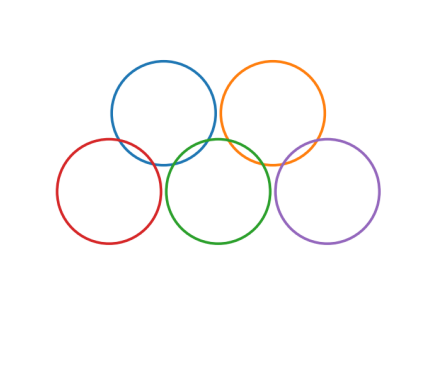}
    &
    \includegraphics[trim={245 260 245 260},clip,width=0.18\linewidth]{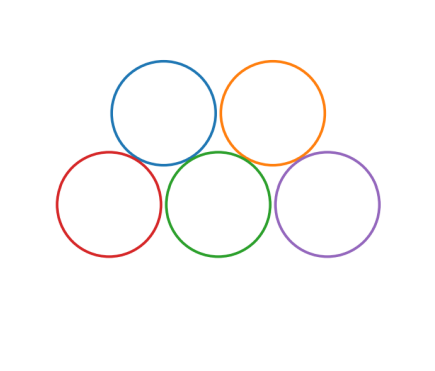}
    &
    \includegraphics[trim={245 260 245 260},clip,width=0.18\linewidth]{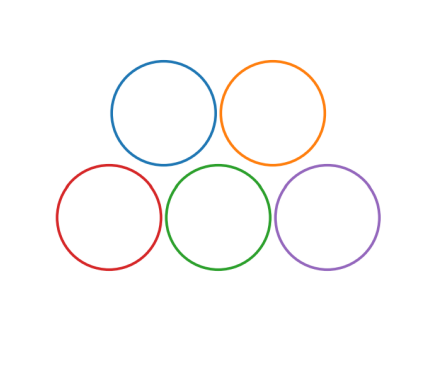}
    &
    \includegraphics[trim={245 260 245 260},clip,width=0.18\linewidth]{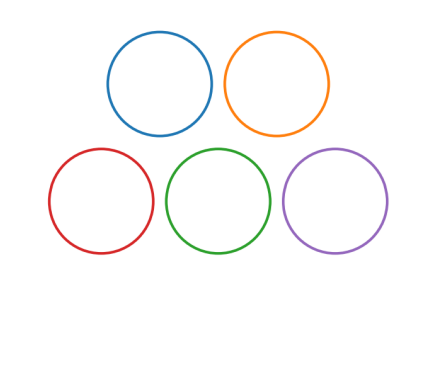}
    &
    \includegraphics[trim={245 260 245 260},clip,width=0.18\linewidth]{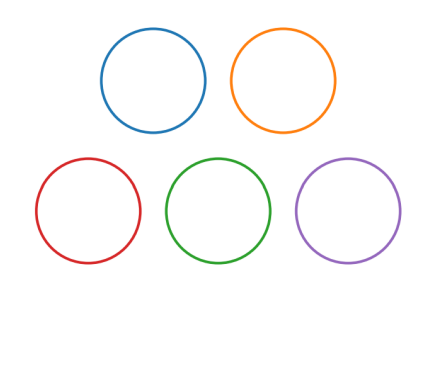}
    &
    \includegraphics[trim={245 260 245 260},clip,width=0.18\linewidth]{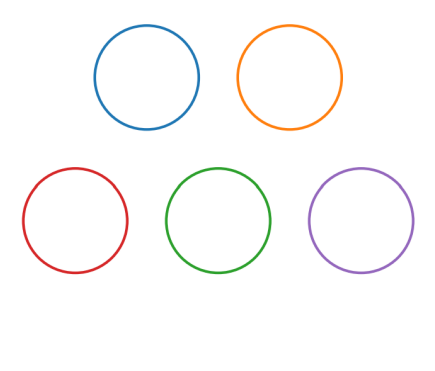}
    &
    \includegraphics[trim={245 260 245 260},clip,width=0.18\linewidth]{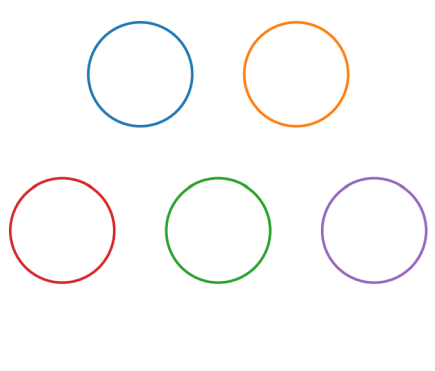}
     \\
     \small($dy$, $dx$) & baseline: \small(-1.0, 0.1)& \small(-0.75, 0.1)& \small(-0.5, 0.1)& \small(-0.25, 0.1)& \small(0.0, 0.1)& \small(0.25, 0.25)& \small(0.5, 0.5)& \small(0.75, 0.75)& \small(1.0, 1.0)\\
\end{tabular}
}
\caption{We reduce the overlap area between circles by increasing the boundary-to-boundary distances along the X and Y axes, \ie, $dx\times \frac{\phi}{2}$ and $ dy\times \frac{\phi}{2}$, respectively.}
\label{fig:spaced_circles_ex}
\end{figure}

\begin{figure}[h]
\centering
\resizebox{\columnwidth}{!}{
\begin{tabular}{ccccccccc}
     \includegraphics[trim={150 190 165 190},clip,width=0.24\linewidth]{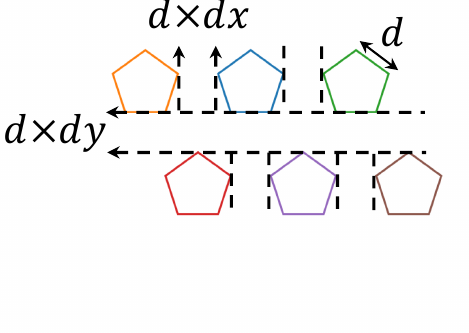}
     & 
     \includegraphics[trim={245 260 230 260},clip,width=0.18\linewidth]{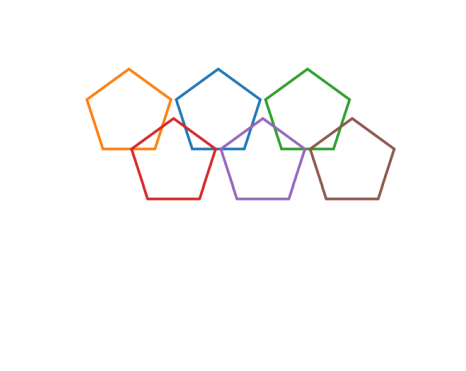}
    &
    \includegraphics[trim={245 260 230 260},clip,width=0.18\linewidth]{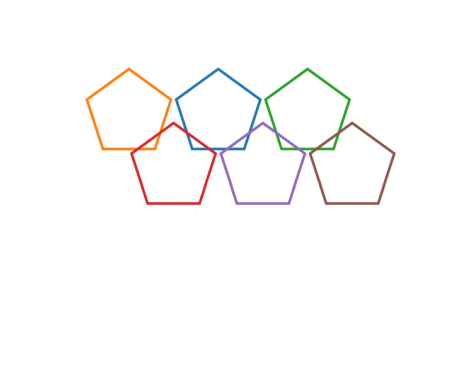}
    &
    \includegraphics[trim={245 260 230 260},clip,width=0.18\linewidth]{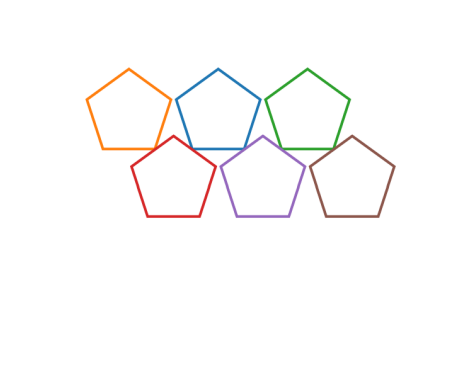}
    &
    \includegraphics[trim={245 260 230 260},clip,width=0.18\linewidth]{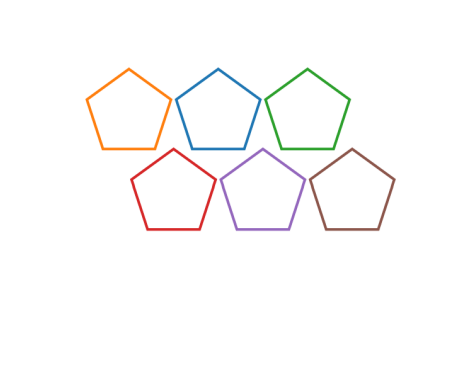}
    &
    \includegraphics[trim={245 260 230 260},clip,width=0.18\linewidth]{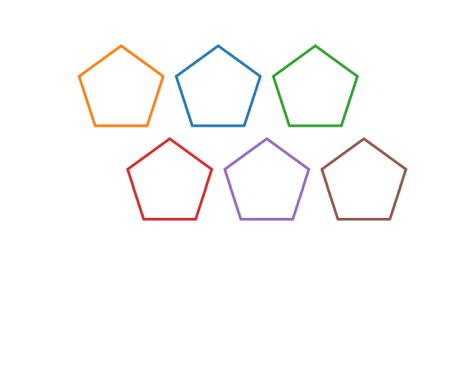}
    &
    \includegraphics[trim={245 260 230 260},clip,width=0.18\linewidth]{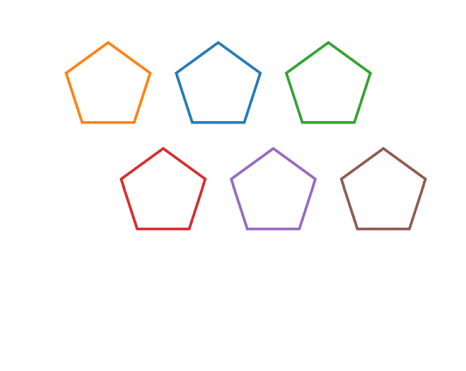}
    &
    \includegraphics[trim={245 260 230 260},clip,width=0.18\linewidth]{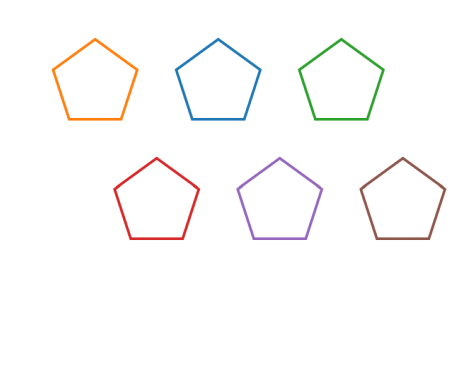}
    &
    \includegraphics[trim={245 260 230 260},clip,width=0.18\linewidth]{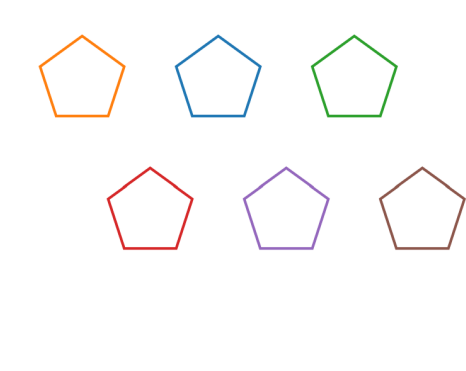}
     \\
     \small($dy$, $dx$) & baseline: \small(-0.58, 0.1)& \small(-0.5, 0.1)& \small(-0.25, 0.1)& \small(0.0, 0.1)& \small(0.25, 0.25)& \small(0.5, 0.5)& \small(0.75, 0.75)& \small(1.0, 1.0)\\
\end{tabular}
}
\caption{We increase the boundary-to-boundary distances between pentagons along the X and Y axes, \ie, $d\times dx$ and $ d\times dy$, respectively.
}
    \label{fig:spaced_pentagons_ex}
\end{figure}

\subsec{Experiment}
The overlapping shapes in the baseline images (\cref{sec:olympic_logo}) are distanced in X and Y directions. 
To generate simplified versions of the images, we gradually increase these distances such that it reduces the overlapping area between the shapes. Specifically, the boundary-to-boundary distance between circles in the original images along the X and Y directions is $\in\{dx\times \frac{\phi}{2}, dy\times \frac{\phi}{2}\}$, where $\phi$ is the diameter of the circles and $dx$ and $dy$ are multipliers (see \cref{fig:spaced_circles_ex}). 
We push the shapes away by increasing $dx$ and $dy$ for circles. 
We repeat the same procedure for pentagons with the boundary-to-boundary distance of $\{d\times dx, d\times dy\}$, where $d$ is the side length of the pentagons (see \cref{fig:spaced_pentagons_ex}). 
We use the same prompts as in \cref{sec:olympic_logo} to evaluate VLMs on the new pushed-away sets and compare their accuracy to the baseline results (in \cref{sec:results}).

\begin{figure}[t]
\centering
\resizebox{\textwidth}{!}{
\begin{tabular}{cc}
     \begin{tabular}{rcc}
           & (a) Circles \olympiclogo & (b) Pentagons \pentagonlogo \\
          \raisebox{-0.8\height}\gptlogo & \multirow{7}{*}{\includegraphics[width=0.491\textwidth]{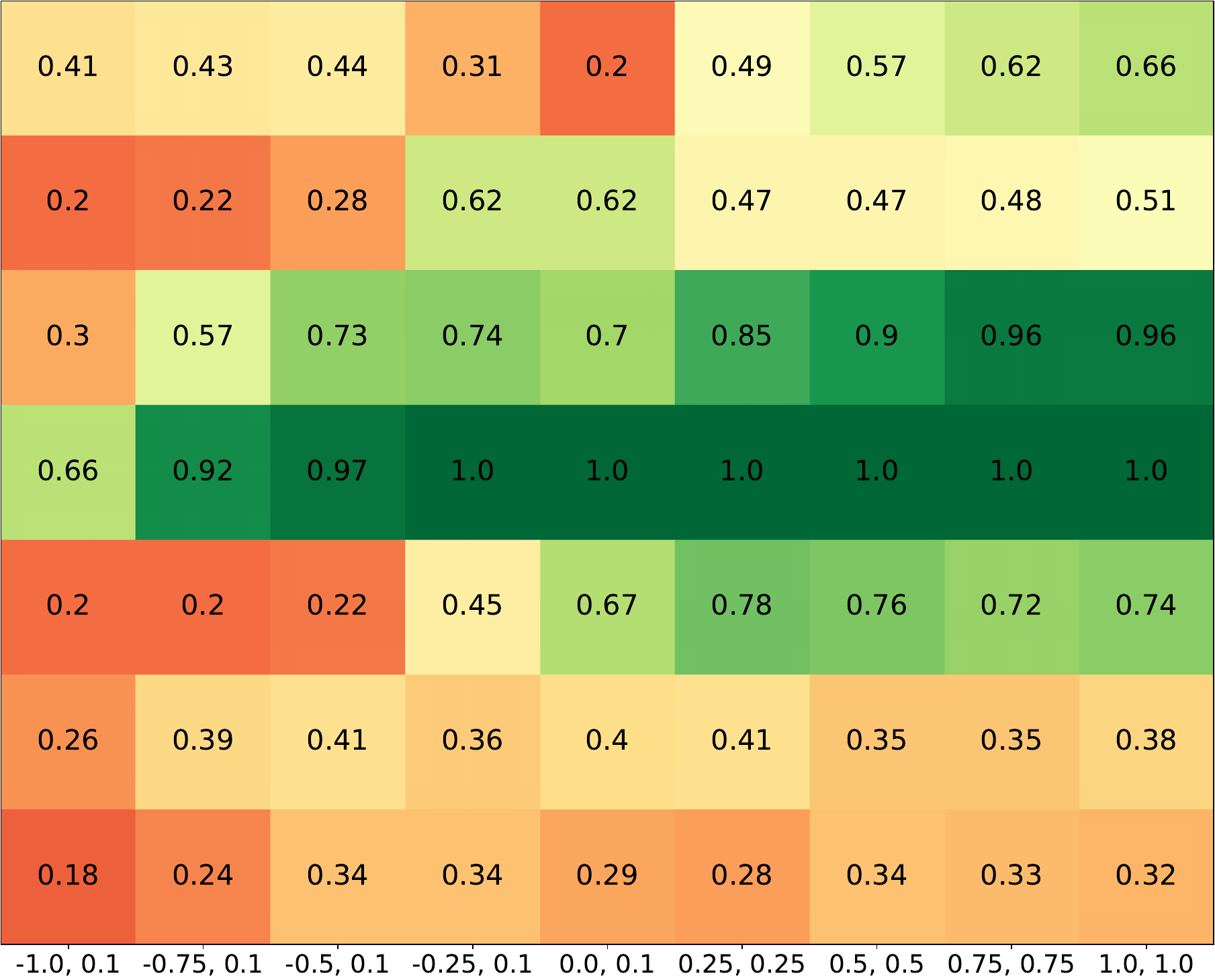}}& \multirow{7}{*}{\includegraphics[width=0.437\textwidth]{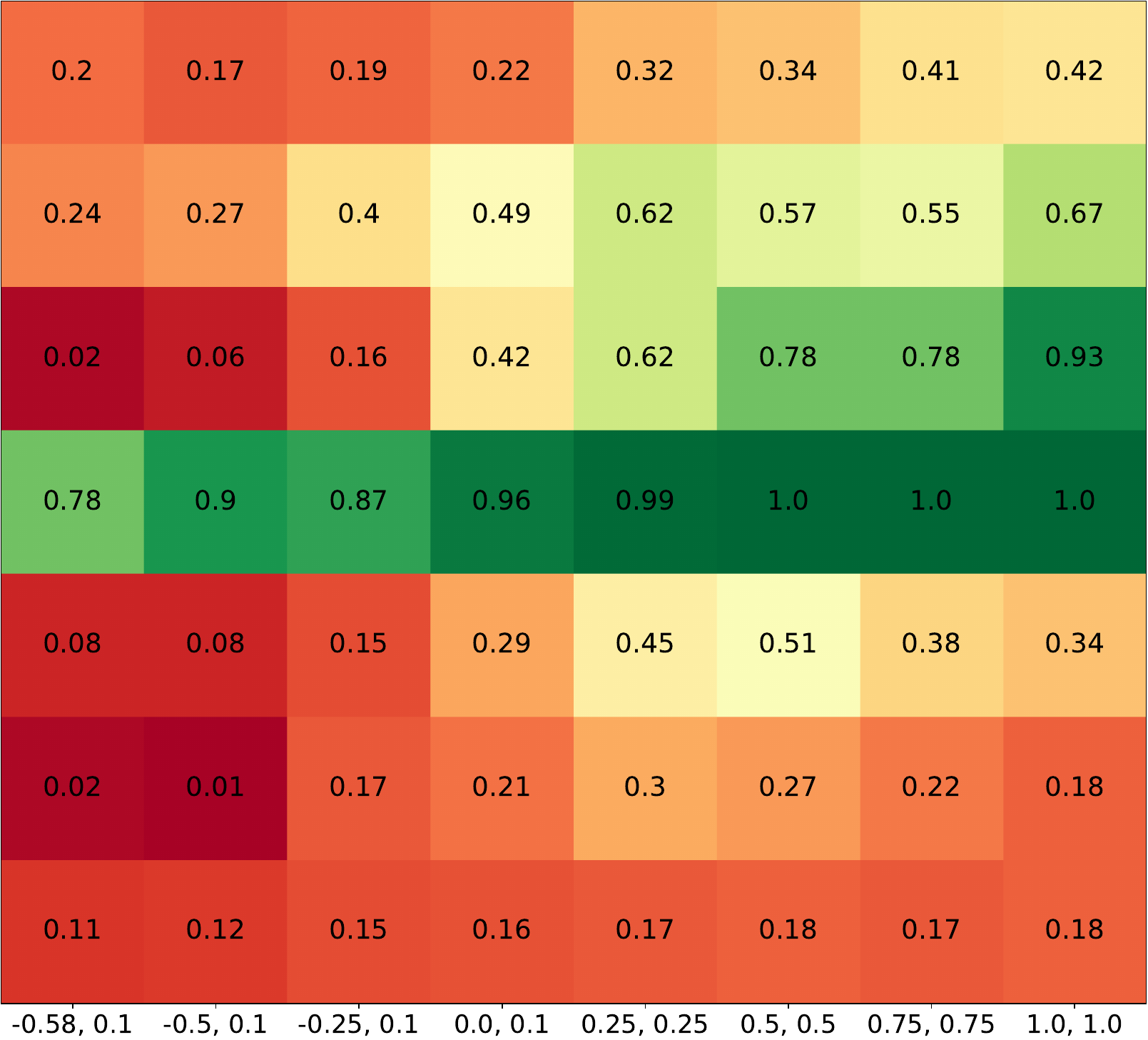}}\\
          \raisebox{-1.4\height}\geminilogo  &&\\
         \raisebox{-1.4\height}\sonnetlogo &&\\
          \raisebox{-1.3\height}\newsonnetlogo &&\\
          \raisebox{-1.\height}{\llavalogolarge}&&\\
          \raisebox{-1.\height}{\llavalogosmall}&&\\
          \raisebox{-1.45\height}{\philogo} &&\\
          \\
          \raisebox{1.3\height}{\tiny$dy$ \rotatebox[origin=c]{90}{$\leftrightarrow$},$dx\leftrightarrow$}&&\\
     \end{tabular}
     & \raisebox{-0.46\height}{\includegraphics[width=0.07\textwidth]{images/CountGridRowColumns/colorbar_only.pdf}} \\
\end{tabular}}
\caption{As we increase the distance between the shapes along the X and Y axis for both circles and pentagons, VLMs' accuracy improves.
For example, \newsonnet (\newsonnetlogo) accuracy increases from 0.66 to 1.0 on \olympiclogo (a) and from 0.78 to 1.0 on \pentagonlogo (b). 
}
\label{fig:heat-cir-dist}
\end{figure}

\subsec{Results}
VLMs, in general, can count shapes more accurately when there is no overlapping area between shapes (\cref{fig:heat-cir-dist}). 
Yet, accuracy increases vary between models. 
For example, both \sonnet and \newsonnet reach $\geq$96\% (\cref{fig:heat-cir-dist}; $dx = 0.75$).
Similarly, 72B-\llavaone (\llavalogolarge) achieves 72\% accuracy on counting disjoint circles (\cref{fig:heat-cir-dist}; $dx = 0.75$).

This shows that most VLMs struggle to count the shapes in the baseline images (\cref{fig:heat-cir-dist}; $dy = -1$ and $dx = 0.1$) because the shapes overlap, which poses a challenge to VLMs in counting.

All \sethlcolor{MyLightGray}\hl{closed-source} VLMs, except for \gemini, consistently benefit from increasing the distance along both directions in counting overlapping shapes. The most significant improvement is for \sonnet with $\triangle$=91\% and the least is for \gpt with  $\triangle$=22\% (\cref{fig:heat-cir-dist}b). 
Qualitatively, this is mainly because counting more than 6 shapes is ``hard'' for \gemini regardless of the amount of space between them due to its strong bias towards the Olympic logo (\cref{sec:res-count-shapes}). 

Similarly to the circled letter task (\cref{sec:spaced-words}), 72B \llavaone performs much better than 0.5B \llavaSI on the simplified versions (\cref{fig:heat-cir-dist}).
This further suggests that the language decoder in \llavaone significantly affects the VLM's ability to see, count, and speak the number of shapes in the image.

\subsection{VLMs can count simplified, more straight paths}
\label{sec:spaced-subway-count}

\begin{figure}
    \centering
    \resizebox{\textwidth}{!}{
    \begin{tabular}{ccc}
        \multicolumn{3}{c}{(a) $P$ values}\\
         \includegraphics[width=0.14\linewidth]{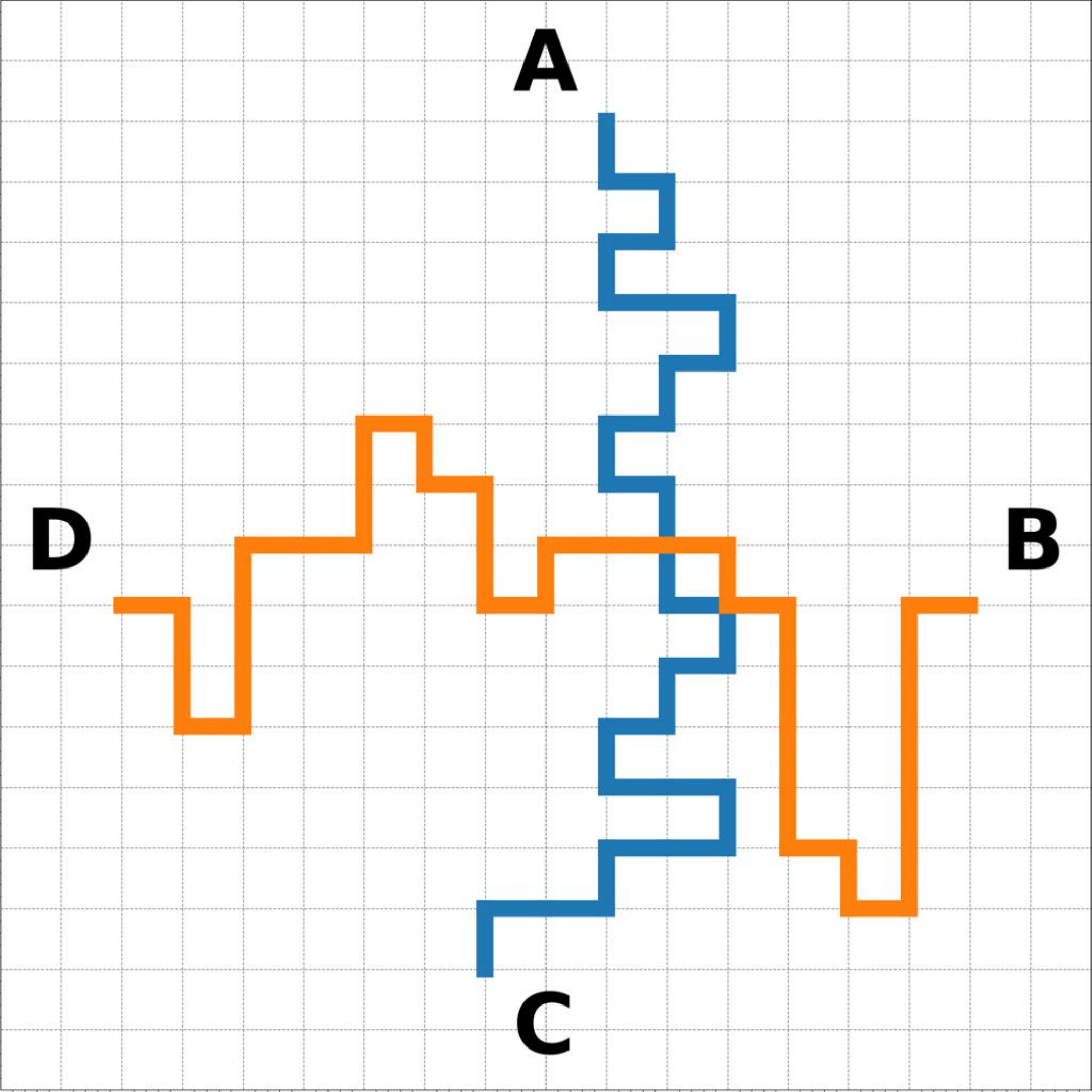}
         & 
         \includegraphics[width=0.14\linewidth]{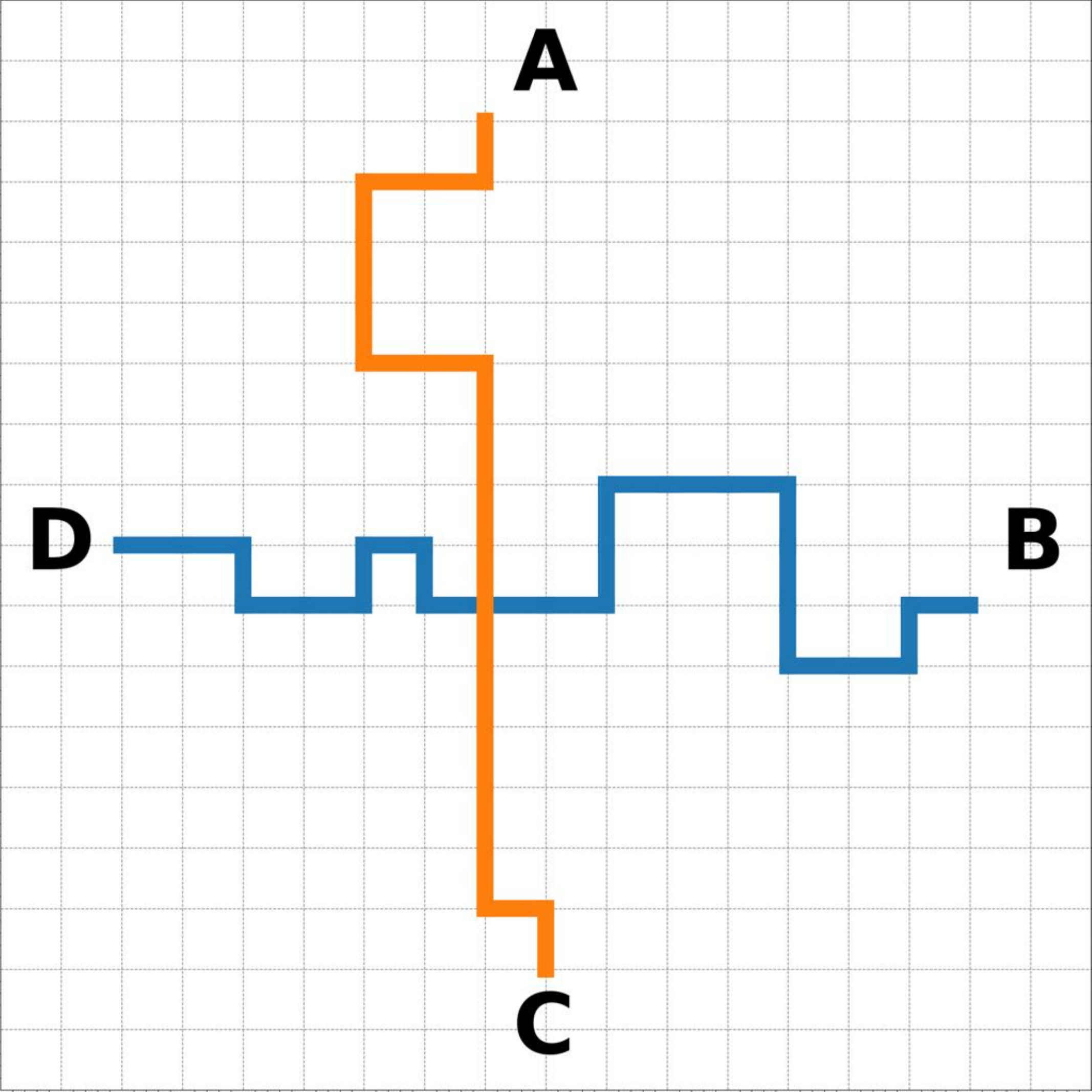}
         &
         \includegraphics[width=0.14\linewidth]{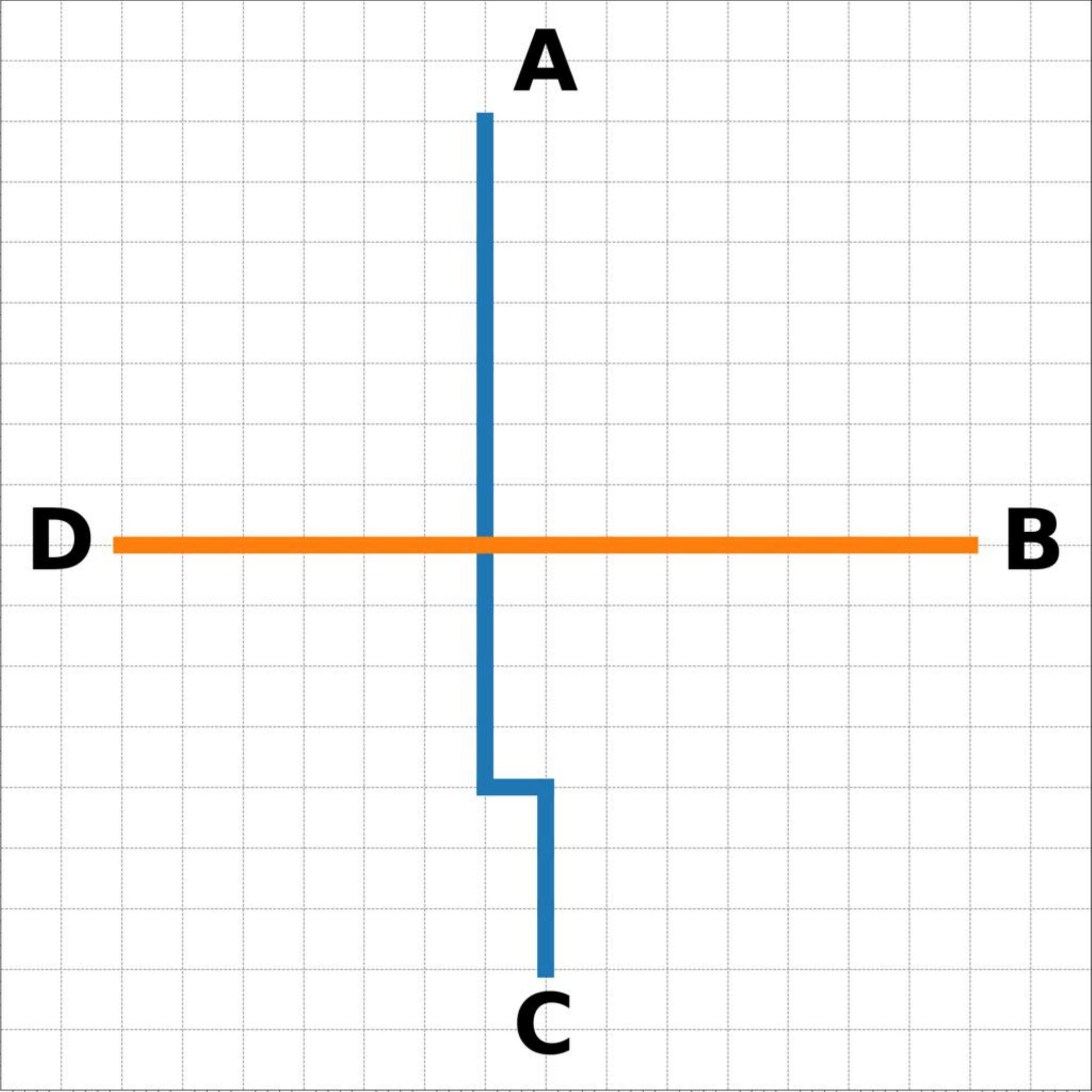} \\
         \makecell{\centering 0.33 (baseline)} & \makecell{\centering 0.6} & \makecell{\centering 0.9}\\
    \end{tabular}
    \begin{tabular}{c}
    (b) \\
    \begin{tabular}{c@{}c}
    \raisebox{-0.7\height}\gptlogo &
    \multirow{7}{*}{
    \begin{tabular}{ccc}
         \includegraphics[width=0.14\linewidth]{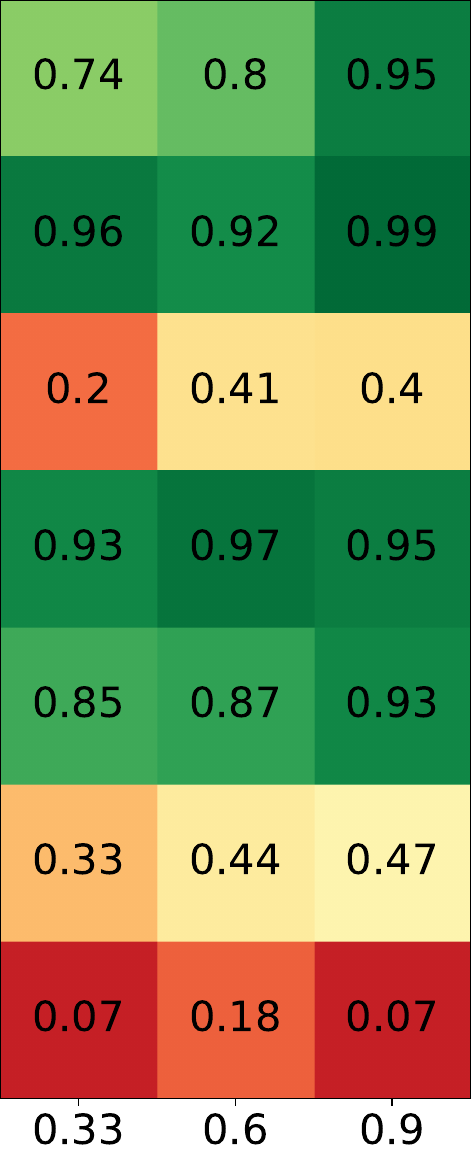}
         & 
         \includegraphics[width=0.14\linewidth]{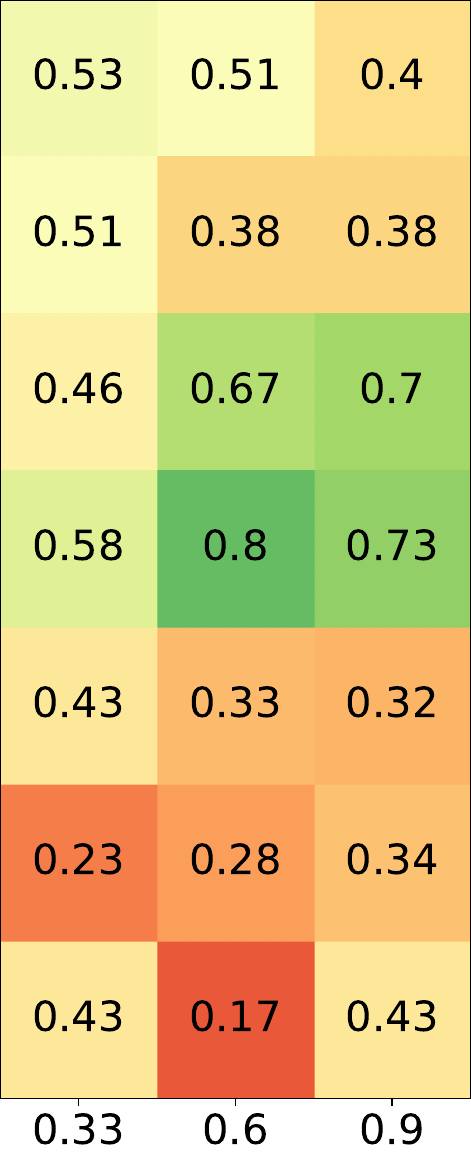}
         & 
         \includegraphics[width=0.192\linewidth]{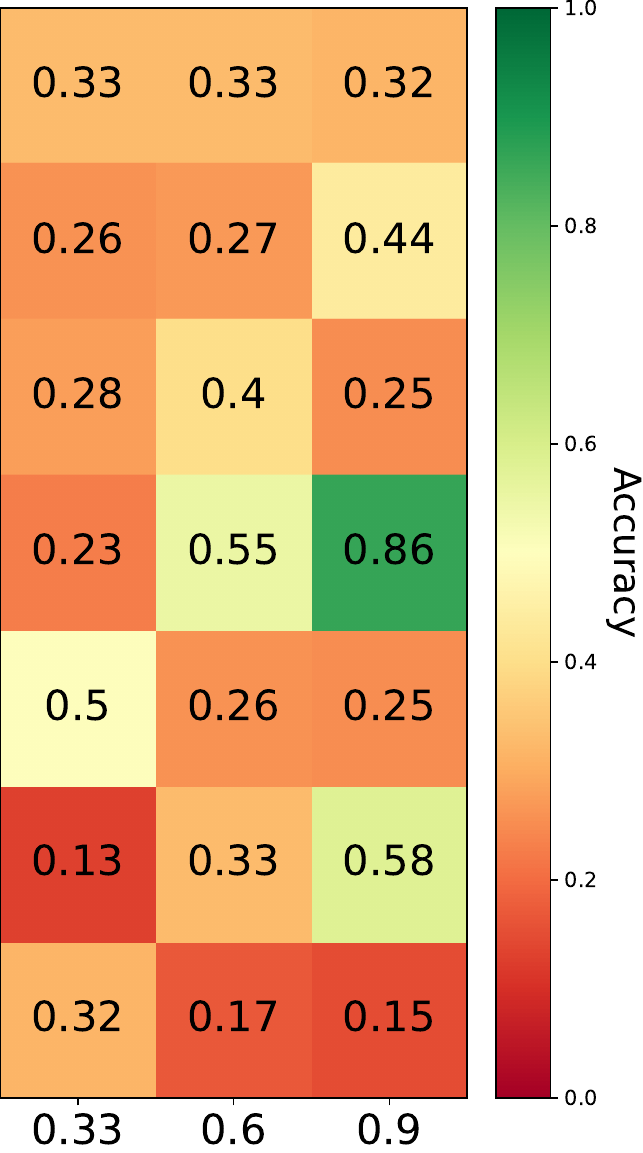} \\
         1 path & 2 paths & 3 paths \\
    \end{tabular}}  \\
    \raisebox{-1\height}\geminilogo &\\
    \raisebox{-1\height}\sonnetlogo &\\
    \raisebox{-1\height}\newsonnetlogo &\\
    \raisebox{-0.8\height}\llavalogolarge &\\
    \raisebox{-0.8\height}\llavalogosmall &\\
    \raisebox{-1.12\height}\philogo & \\
    \end{tabular} \\
    \end{tabular}
    }
    \vspace{2em}
    \caption{We increase the probability $P$ of choosing a straight next move (as opposed to making a turn) and generate two simplified versions of subway-like maps (a).  
    As we increase the probability $P$ from 0.33 to 0.6 and 0.9 (b), some VLMs can reach a near-perfect accuracy (0.99 for \geminilogo~\gemini on 1 path).}
    \label{fig:simp_subway_perpath}
\end{figure}

VLMs' ability to count can also affect the overall accuracy of SOTA models on counting single-colored paths (\raisebox{-0.3\height}{\pathfollowinglogo}) in subway-like maps. 
To investigate whether VLMs are not able to count in general or whether the zigzag patterns of paths (\cref{fig:simp_subway_perpath}a) poses the main challenge to VLMs, we render simplified versions of the original maps, where each path in the map contains fewer intersections with other paths. 
We evaluate the VLMs on the new set and compare their performance with the baseline images.

\subsec{Experiment}
We re-render the images by forcing each path to have fewer $90^\circ$ turns than the baseline. The baseline images (\cref{sec:subway}) are implemented by choosing a direction on a grid using a random depth-first algorithm, where the probability ($P$) of choosing a straight direction is 0.33. Therefore, we gradually increase the $P$ from the baseline ($P$=0.33) to 0.6 and 0.9, such that it yields images with fewer intersections and turns (\cref{fig:simp_subway_perpath}a). We use the prompts in \cref{sec:subway} and compare VLMs' performance with the baseline accuracy.

    

\subsec{Results}
On average, all VLMs more accurately count the single-colored paths when there are fewer turns, \ie, as $P$ increases (see \cref{fig:simp_subway_perpath}b).
The accuracy gain significantly varies between models, \eg, \increasenoparent{2} points for \gpt and \increasenoparent{30} for \newsonnet (\cref{app:simp_sub}). 
This shows that SOTA VLMs mostly struggle to count the paths in original images (\cref{sec:results}) due to the visual complexity of zigzag patterns of paths and their intersection.

Analyzing the accuracy by the number of paths connected to each station, we find that some VLMs even score near-100 accuracy (\eg, 0.95, 0.99, and 0.95 for \gptlogo, \geminilogo, and \newsonnetlogo, respectively at $P=0.9$; \cref{fig:simp_subway_perpath}).
This substantially better accuracy on simplified images is in stark contrast to the poor accuracy reported for the original subway maps ($P=0.33$), confirming that the visual complexity of the paths poses challenges to VLMs.


\section{The vision encoder in open-source VLMs can extract sufficient information to solve \blindtest}
\label{sec:linear_feature_probe}

We find that VLMs surprisingly fail at simple tasks, \eg, determining whether two circles overlap or not (\cref{sec:results}).
Furthermore, our results in \cref{sec:spacing} show supporting evidence that these failures are not due to the uncommon nature of the questions in \blindtest or insufficient prompt engineering.
Specifically, VLMs achieve near-100\% accuracy when there is much more space between shapes and letters (\cref{sec:spacing}).
Therefore, here, we question whether VLMs can ``see'' the key visual information in \blindtest images, \eg, the gap between two circles \twocirclelogo in order to decide whether they overlap.
Specifically, we run linear probing experiments to test whether the visual encoders of open-source VLMs extract sufficient information for solving \blindtest.




\subsec{Models}
From the SOTA open-source VLM families that we test in \cref{tab:open_model_acc}, we select two models, 0.5B \llavaSI (\llavalogosmall) and 4.2B \phivision (\philogo) for \textbf{two reasons}.
First, these two VLMs use the two most common vision encoders (VEs)---\llavalogosmall uses SigLIP \cite{zhai2023sigmoid} while \philogo uses CLIP \cite{radford2021learning}.
That is, our findings on these two VEs would generalize to most VLMs.
Second, \llavalogosmall and \philogo are among the smallest VLMs, and therefore, if their VEs contain sufficient information, the same is likely true with larger and commercial VLMs.


\subsec{Tasks} We choose (1) the two circles (\twocirclelogo) and (2) the counting line-intersections (\lineintersectlogo) for this experiment
because they represent arguably the simplest images and questions in \blindtest (the two circles and the line intersections tasks are 2-way and 3-way classification problems).


We generate 11,100 two-circle \twocirclelogo images with a uniform distribution of images over circle distances ($dx$, $dy$). 
We divide these images into (8880, 1110, 1110) images for train, val, and test sets, respectively. 
Repeating this generation process for the line-intersection task (\lineintersectlogo) with an equal distribution of mean distances between the y coordinates of the red and blue lines (similar to \cref{fig:line_intersection_distance_heatmap}) for a total of 6,300 images, \ie, divided into (4410, 945, 945) for train, val, and test sets.

\begin{figure}[ht]
    \centering
    \includegraphics[width=0.50\textwidth]{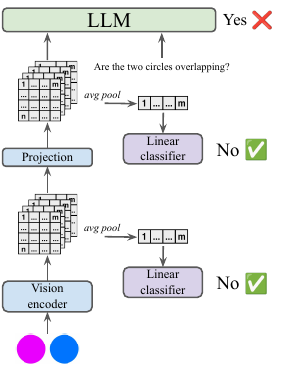}
    \caption{We train a linear-probing classifier on the frozen features extracted from the (1) vision encoder and the (2) projection layer for the two circles \twocirclelogo and the line chart \lineintersectlogo tasks separately. 
    Evaluating the linear classifiers shows that the information necessary to solve these two tasks exists before and after the project layer but is lost in the LLM, resulting in poor VLM accuracy in \cref{sec:results}.
     }
     \label{fig:linprobe-frame}
\end{figure}

\subsec{Method}
We average-pool the image-patch features at the layer right \emph{before} the projection layer (\cref{fig:linprobe-frame}).
Then, we train a logistic-regression linear classifier on top of the frozen features on each task.
We train each classifier for 1,000 epochs with $L_2$ regularization weight of 1.0 (\cref{apptab:ex_lin_prob_details}) and choose the best model based on the validation set.
The dimensions of the features are described in \cref{appsec:implementation_linear}. 
For completeness, we repeat the experiment for the layer right \emph{after} the projection layer to understand the impact of the projection layer.



\subsec{Results} 
The linear-probing accuracy of the CLIP features, \textbf{before projection} layer, in \llavalogosmall is $\geq$ 99.47\% on both tasks (\cref{tab:linprpobecomparison}). 
This suggests that the necessary low-level information to solve these tasks is preserved in CLIP. 
Similarly, the same conclusion holds for the VE in \philogo, a variant of SigLIP, that performs 100\% on both tasks (\cref{tab:linprpobecomparison}). Moreover, using the frozen features \textbf{after the projection} layer in both \llavalogosmall and \philogo yields a linear classification accuracy of $\geq$ 99.58\% on both tasks (\cref{tab:linprpobecomparison}). 
This result shows that most visual information from VEs is preserved before and after the projection layer. 
Contrasting these high linear-classification VE accuracy scores with fairly lower accuracy of both VLMs on \blindtest (see \cref{tab:linprpobecomparison}; rightmost column), we conclude that the \textbf{language models in these VLMs have access to the necessary visual information to solve \blindtest tasks but fail to decode it into correct language outputs}.

Interestingly, in the two-circles \twocirclelogo task, training on frozen features of images, where the two circles are close ($distance = -0.05, 0.05$) produces a linear classifier that performs perfectly at 100\% accuracy not only on that training distribution but also on splits containing circles that are farther apart (\eg, $distance = -0.25, 0.25$) regardless of the VE (\llavalogosmall~and \philogo).
However, the opposite does not hold---training on the images where circles are farther apart ($distance = -0.25, 0.25$) results in a classifier that performs poorly on circles that are closer.
More details in \cref{app:probe-results_splits}.
This result shows evidence that (1) training on the harder cases ($distance = -0.05, 0.05$) results in a more robust classifier and (2) our classifiers do not overfit to training data.

\begin{table}[h]
    \centering
    \caption{
    The output features from the vision encoders right before (a) the projection layer in \llavaSI (\llavalogosmall) and \phivision (\philogo), \ie, CLIP and SigLIP, respectively, contain sufficient information to solve the \twocirclelogo and \lineintersectlogo (linear-probing accuracy is $\geq$ 99.47\%).
    The same conclusion holds for after (b) the projection layer.
    However, the language model in these VLMs fails to decode this information into correct answers, resulting in poor accuracy on the tasks (c).
    }
    \label{tab:linprpobecomparison}
    \begin{tabular}{lcccccccc}
    \toprule
    & \multicolumn{2}{c}{(a) Before} && \multicolumn{2}{c}{(b) After} && \multicolumn{2}{c}{(c) VLM} \\
    \cmidrule(lr){2-3}
    \cmidrule(lr){5-6}
    \cmidrule(lr){8-9}
        Model & ~~\lineintersectlogo & \twocirclelogo && ~~\lineintersectlogo & \twocirclelogo && ~~\lineintersectlogo & \twocirclelogo \\
        \midrule
        \llavalogosmall & ~~99.47 & 99.82&& ~~99.58 & 99.73 && ~~33.14 & 73.21\\
        \philogo & ~~100.0 & 100.0 && ~~100.0 & 100.0 && ~~37.78 & 83.63\\ 
        \bottomrule
    \end{tabular}
\end{table}

\section{Related Work}
\subsec{Benchmarking VLM vision understanding}
College-level topics \cite{yue2023mmmu}, charts \cite{masry2022chartqa}, documents \cite{mathew2021docvqa} or videos \cite{yu2019activitynet} are among the common benchmarks for assessing VLM vision understanding \cite{gpt4o,anthropic2024claude,reid2024gemini,sonnet35} and are witnessing VLMs' recent rapid progress---\eg, \newsonnet is reaching 95.2\% on DocVQA, 90.8\% on ChartQA, and 94.7\% on AI2D \cite{sonnet35}.
However, most of the vision benchmarks attempt to evaluate VLMs on real-world, topic-specific data that require extensive prior knowledge \cite{liang2024scemqa,wang2024measuring,chen2024we}, which has a ``data leakage'' problem, \ie, VLMs many times can answer accurately without even the input image \cite{chen2024we}.
Furthermore, most benchmarks test VLMs on the data that humans have to deal with to provide a high-level sense of the human-machine intelligence gap \cite{yu2023mm,liu2023mmbench}.
In contrast, our \texttt{BlindTest} benchmark differs significantly from prior benchmarks because it is (1) \textbf{extremely easy to humans and can be solved by a 5-year-old} (unlike \cite{yue2023mmmu,masry2022chartqa,mathew2021docvqa}); (2) the first low-level, visual sanity check for VLMs; (3) requiring minimal to zero prior knowledge; (4) requiring minimal commonsense or complex reasoning (unlike \cite{chollet2019measure,zellers2019recognition})---\ie, \textbf{a strong language model is of little use here when it is non-natural for humans to describe \blindtest images in language}. 

The ARC benchmark \cite{mitchell2023comparing,chollet2019measure} also contains abstract images made up of simple shapes; however, it challenges VLMs to understand and reason based on those patterns.
That is, ARC assumes VLMs can identify the abstract shapes in order to reason.
In contrast, our \blindtest directly evaluates VLM capabilities in recognizing these primitive shapes.

\subsec{Improving VLM vision capabilities}
Most recent recipes for improving SOTA VLMs involve finetuning a pretrained LLM coupled with vision encoders to solve high-level vision tasks \cite{liu2023visual}. 
Such late-fusion approaches fuse visual representations learned from the tokenized image with a powerful thinking brain \cite{McKinzie2024MM1MA,liu2024llavanext,liu2023improvedllava}.
However, current vision approaches for VLMs are facing challenges as models sometimes are ``blind''---unable to see natural objects exist in a real photo \cite{Tong_2024_CVPR}.
In contrast, we are showing VLMs are visually impaired at low-level abstract images, \eg, inability to count 6 overlapping circles or 3 nested squares.

Our circled-letter task (\cref{sec:circled_letter}) is inspired by VLM abilities in recognizing content inside a red circle over real objects in natural images \cite{yang2023som,shtedritski2023does,yang2023dawn}.
In contrast, we show that VLMs can fail at a low-level, optical character recognition as opposed to recognizing real objects.
To the best of our knowledge, no prior attempts have been made to address the exact limitations raised in our paper: (1) identifying and counting simple lines, shapes and geometric primitives when they interact (\cref{sec:res-2line} to \cref{sec:res-gridcount}); (2) following colored paths (\cref{sec:res-paths}).
Solving these limitations may be the foundation for VLMs to progress on some existing vision benchmarks on graphs, \eg, \cite{li2024visiongraph}, visual math \cite{lu2023mathvista} and some existing blind-spots in natural images (\eg, understanding the direction an object is facing \cite{Tong_2024_CVPR}).

\section{Discussion and Conclusion} 

We propose \blindtest, a benchmark of seven novel low-level visual tasks for testing VLM ability to ``see'' simple geometric primitives (such as line, circles, squares, intersections) that are the basic building blocks for many image tasks.
The tasks are designed from scratch and require minimal to zero knowledge.
As the tasks did not exist on the Internet before and require minimal world knowledge, there is minimal chance that VLMs can solve \blindtest by memorization or by not using the input image---an issue in some prior benchmarks \cite{chen2024we,hegde2023analyzing}. 

Furthermore, we also test common prompting techniques (\cref{app:prompting-tech}) including 2-shot, chain-of-thought \cite{wei2022chain}, and meta-prompting \cite{mirza2024meta} but do not obtain better accuracy, which (1) suggests that VLMs understand \blindtest questions and (2) confirms that these visual tasks do not benefit from thinking aloud \cite{wei2022chain}.



The poor performance of VLMs on \blindtest suggests that models will perform poorly on the real-world visual tasks that require them to follow arrow directions or paths, (\eg, reading subway maps in \cref{fig:real_task_failures_subway}, street maps or directed graphs in \cref{fig:real_task_failures}), perceive lines and intersections (\eg, reading music sheets; \cref{fig:real_task_failures}), identify and counts objects in a crowded scene.



\subsection*{Acknowledgement} 

We thank Hung H. Nguyen, Thang Pham, Ali Yildirim, Giang Nguyen, and Tin Nguyen at Auburn University for feedback and discussions of the earlier results.
We are also thankful for the API research credits from Anthropic and \texttt{together.ai} to MRT.
AN was supported by the NSF Grant No. 1850117 \& 2145767, and donations from NaphCare Foundation \& Adobe Research.

\clearpage
%
%
\bibliographystyle{splncs04}
\bibliography{main.bbl}

\renewcommand{\thesection}{\arabic{section}}
\renewcommand{\thesubsection}{\thesection.\arabic{subsection}}
\newcommand{\beginsupplementary}{%
	\setcounter{table}{0}
	\renewcommand{\thetable}{T\arabic{table}}%
	\setcounter{figure}{0}
	\renewcommand{\thefigure}{F\arabic{figure}}%
	\setcounter{section}{0}
}
\beginsupplementary

\appendix

\newcommand{\toptitlebar}{
    \hrule height 4pt
    \vskip 0.25in
    \vskip -\parskip%
}
\newcommand{\bottomtitlebar}{
    \vskip 0.29in
    \vskip -\parskip%
    \hrule height 1pt
    \vskip 0.09in%
}

\newcommand{\suptitle}{Appendix for:\\\papertitle}

\newcommand{\maketitlesupp}{
    \newpage
    \onecolumn
        \null
        \vskip .375in
        \begin{center}
            \toptitlebar
            {\Large \bf \suptitle\par}
            \bottomtitlebar
            \vspace*{24pt}
            {
                \large
                \lineskip=.5em
                \par
            }
            \vskip .5em
            \vspace*{12pt}
        \end{center}
}

\maketitlesupp%

\section{Description of models tested}
\label{sec:models}

On our benchmark, we find that some chat interfaces perform \emph{worse} than their API counterparts (\eg, the system on \texttt{gemini.google.com} is worse than \geminifull on \texttt{aistudio.google.com}) perhaps due to their extra finetuning \cite{ouyang2022training} or specific \texttt{system} prompts \cite{system_prompt_GPT4} that attempt to align VLMs with a company's policies.
Similarly, we find \gpt and Claude 3 models in \url{perplexity.ai} to perform worse than the original API models.
To make sure we test the best VLMs available, we access all {four} models via their available APIs on \href{https://platform.openai.com/}{OpenAI}, \href{https://aistudio.google.com}{Google}, and \href{https://claude.ai/}{Anthropic}.

We describe below the exact API versions and settings for each model.

\subsection{\gpt}

We access the API for \gpt (gpt-4o-2024-05-13) via \url{platform.openai.com} and use all \emph{default} settings including: 
\begin{itemize}
    \item \texttt{temperature}: 1.0 
    \item \texttt{detail}: auto (see \href{https://platform.openai.com/docs/guides/vision}{API doc}), \ie, the system will automatically decide whether to use the ``low-res'' (85 tokens) or ``high-res'' mode (85 tokens and a set of 170 tokens for every 512$\times$512 tile).
\end{itemize}

\subsection{\geminifull}
\gemini (gemini-1.5-pro-latest) API is accessible via \url{aistudio.google.com}, and we use all \emph{default} settings.

\subsection{\sonnetfull}
We access the anthropic API via \url{console.anthropic.com} to use \sonnet (claude-3-sonnet-20240229) with \emph{default} settings.

\subsection{\newsonnetfull}
We follow the same process as for \sonnet and use \url{console.anthropic.com} to access \newsonnet (claude-3-5-sonnet-20240620) API with \emph{default} settings.

\subsection{\llavaonefull}
We run the publicly available \href{https://github.com/LLaVA-VL/LLaVA-NeXT}{code} of \llavaonefull locally with \texttt{temperature}=0.2.

\subsection{\phivisionfull}
We host \phivisionfull on a local machine via their \href{https://huggingface.co/microsoft/Phi-3.5-vision-instruct}{Hugging Face page} with \texttt{temperature}=1.0.

\subsection{\internvl}
We access \internvl via their \href{https://huggingface.co/collections/OpenGVLab/internvl-20-667d3961ab5eb12c7ed1463e}{Hugging Face repository}, and use the \texttt{temperature}=0.2.

\subsection{\qwen}
We use the publicly available codebase on \href{https://huggingface.co/Qwen/Qwen2.5-VL-72B-Instruct}{Huggingface}, and run the evaluation with the \texttt{temperature}=0.7.

\subsection{\geminiflashfull}
We use the API via \url{aistudio.google.com} with \texttt{temperature}=0.7.

\subsection{\geminithinkfull}
We use the model available via \url{aistudio.google.com} with \texttt{temperature}=0.7.

\subsection{\qvqfull}
We run the publicly available model on \href{https://huggingface.co/Qwen/QVQ-72B-Preview}{Huggingface}, with the \texttt{temperature}=0.7.

\clearpage
\section{Hallucinations and educated guesses are among VLMs' common failures}
\label{app:haluci_ed_guess}

\begin{figure}[ht]
\centering
    \begin{AIbox}{What is the color of the overlapping area between two circles?}
    {\centering
    \hspace{0.4\textwidth}\includegraphics[width=0.15\textwidth]{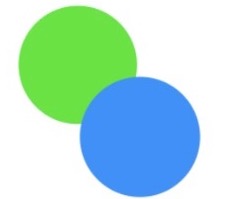}
    }
    \\
    \footnotesize

    \gptlogo~~The overlapping area between the two circles is turquoise (a shade of \textcolor{red}{cyan}).

    \geminilogo~~The overlapping area of a green circle and a blue circle would be \textcolor{red}{**cyan**}. 

    \sonnetlogo~~There is \textcolor{red}{no overlapping area} between the two circles in this image. The image shows two separate circles - one green circle and one blue circle - that are not intersecting or overlapping with each other. \redxmark

    \newsonnetlogo~~I apologize, but there is \textcolor{red}{no overlapping area} between the two circles in this image. The image shows two distinct circles - one green and one blue \redxmark \textcolor{gray}{[...]}

    \end{AIbox}   

\caption{Sometimes VLMs (\gptlogo~and \geminilogo) fail to see clearly and make educated guesses.}
\label{fig:hallucinate}
\end{figure}

\clearpage
\section{Evaluating open-source models on \blindtest}
\label{app:open-source}


On average, 8 well-known open-source models perform \emph{worse} than the commercial, \sethlcolor{MyLightGray}\hl{closed-source} VLMs on \blindtest. 
However, the best-performing open-source model (\llavaone-72B-ov), performs on-par with \sonnet (45.92 vs. 45.73; \cref{tab:open_model_acc}).
As model size increases, VLMs tend to perform better.
For instance, the accuracy increase from \llavaone-ov 0.5B to 72B is \increasenoparent{27.1} (18.82 $\to$ 45.92; \cref{tab:open_model_acc}). 
In sum, \textbf{open-source VLMs exhibit similar limitations to closed-source models on \blindtest}.


\begin{table}[ht]
    \centering
    \caption{Open-source VLMs underperform the \sethlcolor{MyLightGray}\hl{closed-source} ones on \blindtest.
    }
    \label{tab:open_model_acc}
    \resizebox{\columnwidth}{!}{
    \begin{tabular}{lrccccccccc}
Model & Size & ~~\lineintersectlogo & \twocirclelogo & \smallredcircle & \olympiclogo & \pentagonlogo  & \nestedsquareslogo & \gridlogo & \pathfollowinglogo & Task mean\\
\midrule
Random  & & ~~33.33           & 50.00          & 5.77 &  20.00 & 20.00 & 25.00  & 4.55 & 33.33 & 24.00 \\
\midrule
\rowcolor{MyLightGray}
\gpt& \na & ~~41.61           & 75.91          & 74.23 & 41.25 & 20.21 & 55.83 & 39.58 & 53.19 & 50.23  \\
\rowcolor{MyLightGray}
\gemini & \na & ~~66.94  &  \textbf{93.62} & 83.29 & 20.25 & 24.17  & 87.08 & 39.39 & 53.13 & 58.48 \\
\rowcolor{MyLightGray}
\sonnet& \na & ~~43.41 & 86.46 &  72.06 & 29.79 & 1.87 & 65.00 & 36.17 & 31.11 & 45.73 \\
\rowcolor{MyLightGray}
\newsonnet & \na & ~~\textbf{75.36} &	90.82 &	\textbf{87.88} &	\textbf{66.46} &	\textbf{77.71} &	\textbf{92.08} &	\textbf{74.26} &	\textbf{58.19} &	\textbf{77.84}\\
\midrule
\llavaone-ov & 72B & ~~45.83& 90.92 & 44.71 & 20.00 & 11.74 & 87.07 & 8.95 & 58.06 & 45.92 \\
\llavaone-si & 72B & ~~45.33 & 83.48 & 38.14 & 20.00 & 11.46  & 57.50 & 10.23 & 48.06 & 38.41 \\
\llavaone-ov & 7B & ~~48.17 & 83.93 & 42.79 & 20.00 & 7.29 & 42.92 & 21.02 & 47.22 & 39.17 \\
\llavaone-si & 7B & ~~44.50 & 84.67 & 40.22 & 20.00 & 7.29 & 58.75 & 14.01 & 55.00 & 40.00 \\
\llavaone-ov & 0.5B & ~~17.28 & 75.07 & 9.78 & 12.50 & 9.58 & 20.42 & 0.38 & 5.56 & 18.82 \\
\llavaone-si & 0.5B & ~~33.14 & 73.21 & 6.25 & 27.29 & 2.50 & 14.58 & 1.13 & 26.11 & 23.03 \\
\internvl & 8B & ~~47.28 & 91.00 & 57.69 & 20.00 & 13.96 & 28.33 & 7.57 & 60.28 & 40.76 \\
\phivision & 4.2B & ~~37.78 & 83.63 & 16.51 & 18.75 & 11.46 &  32.50 & 11.74 & 19.72 & 29.01\\
\midrule
Mean  & & ~~45.55 & 84.39 & 47.79 & 26.35 & 16.60 & 53.50 & 22.03 & 42.97 & 42.28 \\
\bottomrule
\end{tabular}
}
\end{table}


\clearpage
\section{Advanced prompting techniques}
\subsection{Finding: Meta-prompting and 2-shot examples do not improve the VLMs' performance on the two circles task (\twocirclelogo)}
\label{app:prompting-tech}
We run \gpt and \newsonnet on the two circle task (\twocirclelogo) with \textbf{2-shot} (providing 2 example images with answers) and \textbf{meta-prompting}\footnote{\emph{Describe the image in detail first, and then answer: Are the two circles overlapping? Y/N.}} \cite{mirza2024metaApp}.
We find them to perform worse than our baseline prompts (\cref{tab:adv_prom}).
An explanation is that the VLMs already understand the questions but are limited by the ability to ``see''. These techniques are not helpful perhaps because \blindtest tasks intuitively do not benefit from thinking aloud.


\begin{table}[ht]
    \centering
    \caption{In-context examples and meta-prompting do not improve the overall accuracy of the \newsonnet and \gpt on the \twocirclelogo task.}
    \begin{tabular}{lcc}
    \toprule
    Prompt & ~~\newsonnetlogo & ~~\gptlogo \\
    \midrule
    Baseline & ~~\textbf{91.66} & ~~\textbf{72.69} \\
    Meta-prompting \cite{mirza2024metaApp} & ~~90.53 & ~~65.62 \\
    2-shot & ~~77.93 & ~~68.00 \\
    \bottomrule
    \end{tabular}
    \label{tab:adv_prom}
\end{table}

\clearpage

\section{Image reconstruction using \newsonnet}
Inspired by Pixel Value Prediction \cite{gou2024well} task in VLMs, we formulate the tasks in \blindtest as a reconstruction problem. However, we rely on VLMs' existing capabilities, such as coding, to reform our tasks. Specifically, instead of framing the prompts into question answering, given the input image, we ask the best performing VLM (\newsonnet) to explicitly use its coding ability and reconstruct the input images.

\subsection{Sonnet-3.5 cannot use its coding ability to regenerate the images from \blindtest}

\subsec{Experiment} We ask \newsonnet to generate various formats of the 6 samples from each of 7 tasks in \blindtest, specifically SVG, HTML/CSS, Javascript (using the canvas HTML element to draw the image), and the Python PIL package. We aim to see if the model can maintain the key features of the input image in its generated image, \eg, if \newsonnet can reconstruct images of the \lineintersectlogo task with the same number of intersections as in the input image.

\begin{figure}
    \centering
    \begin{tabular}{ccccc}
         \includegraphics[width=0.19\linewidth]{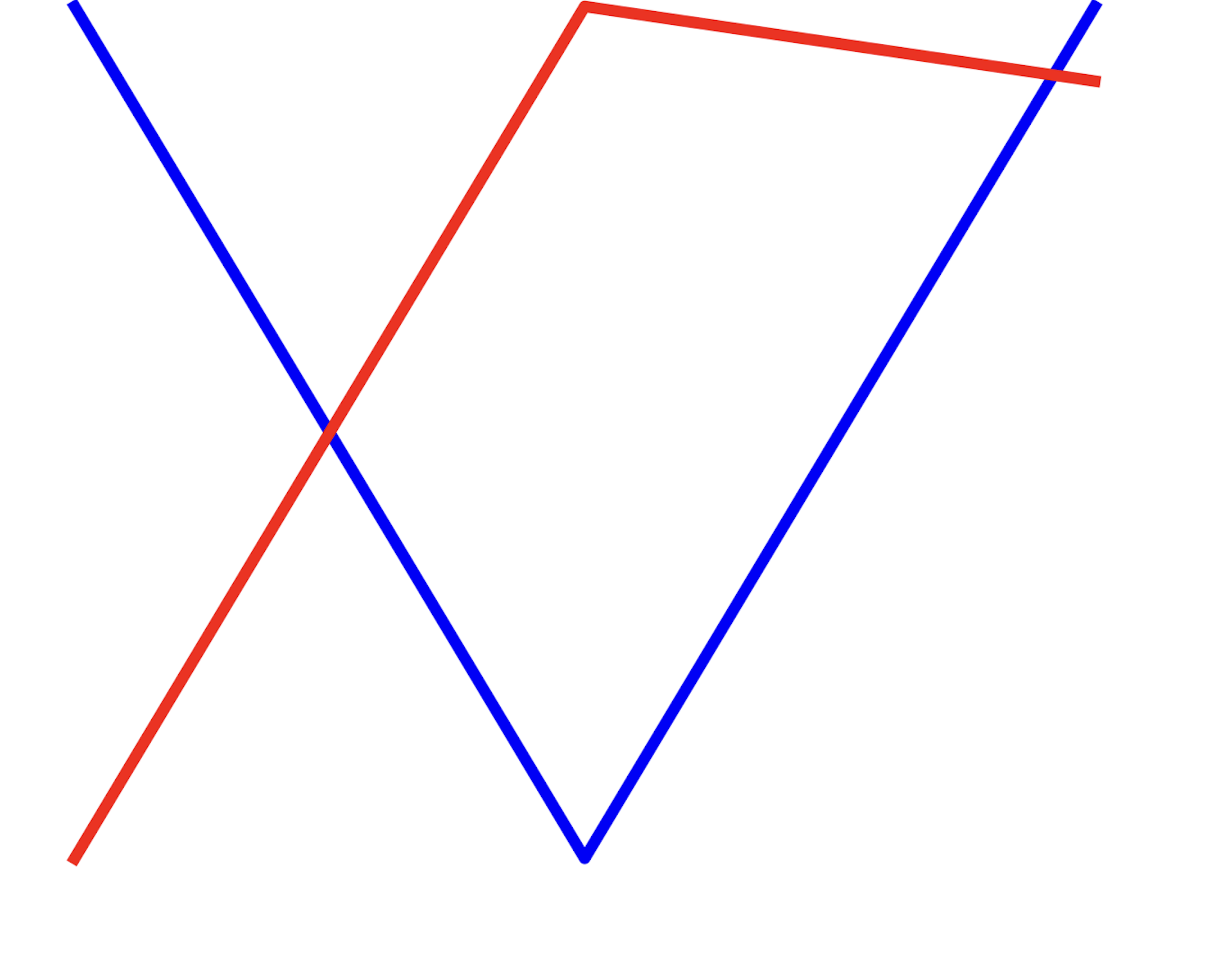}
         &
         \includegraphics[width=0.19\linewidth]{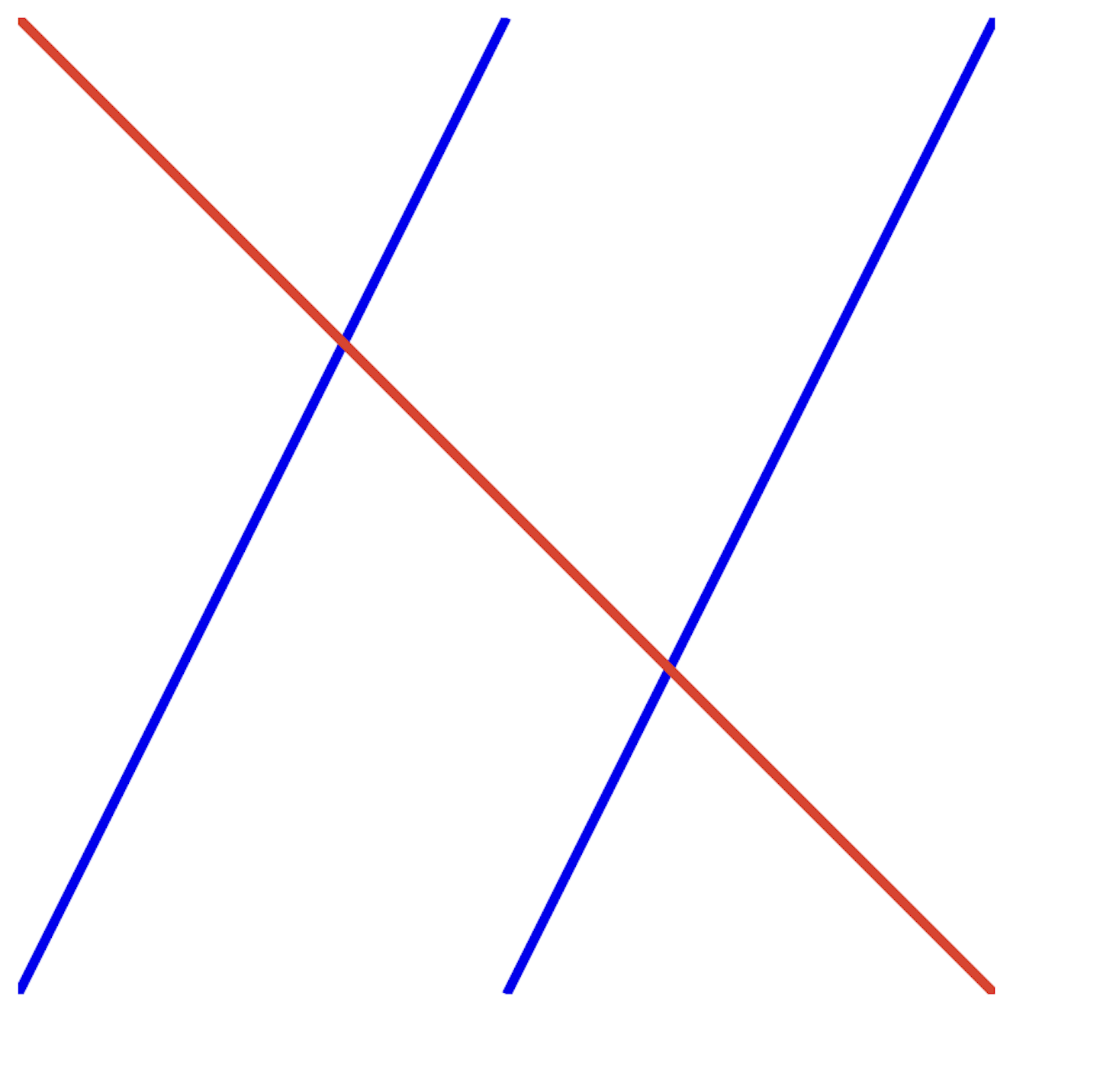}
         &
         \includegraphics[width=0.19\linewidth]{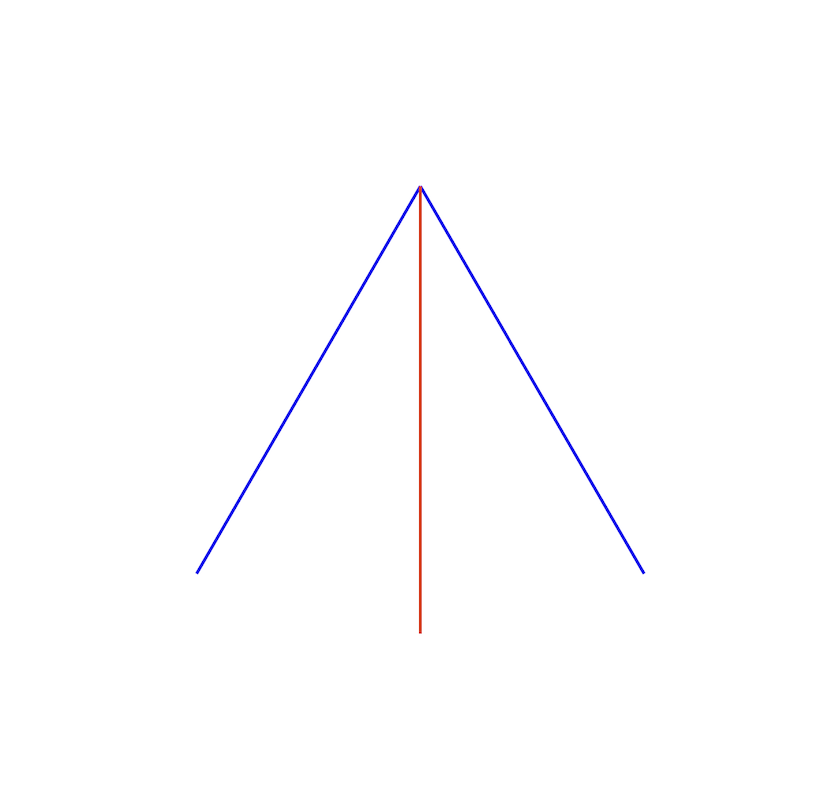}
         &
         \includegraphics[width=0.19\linewidth]{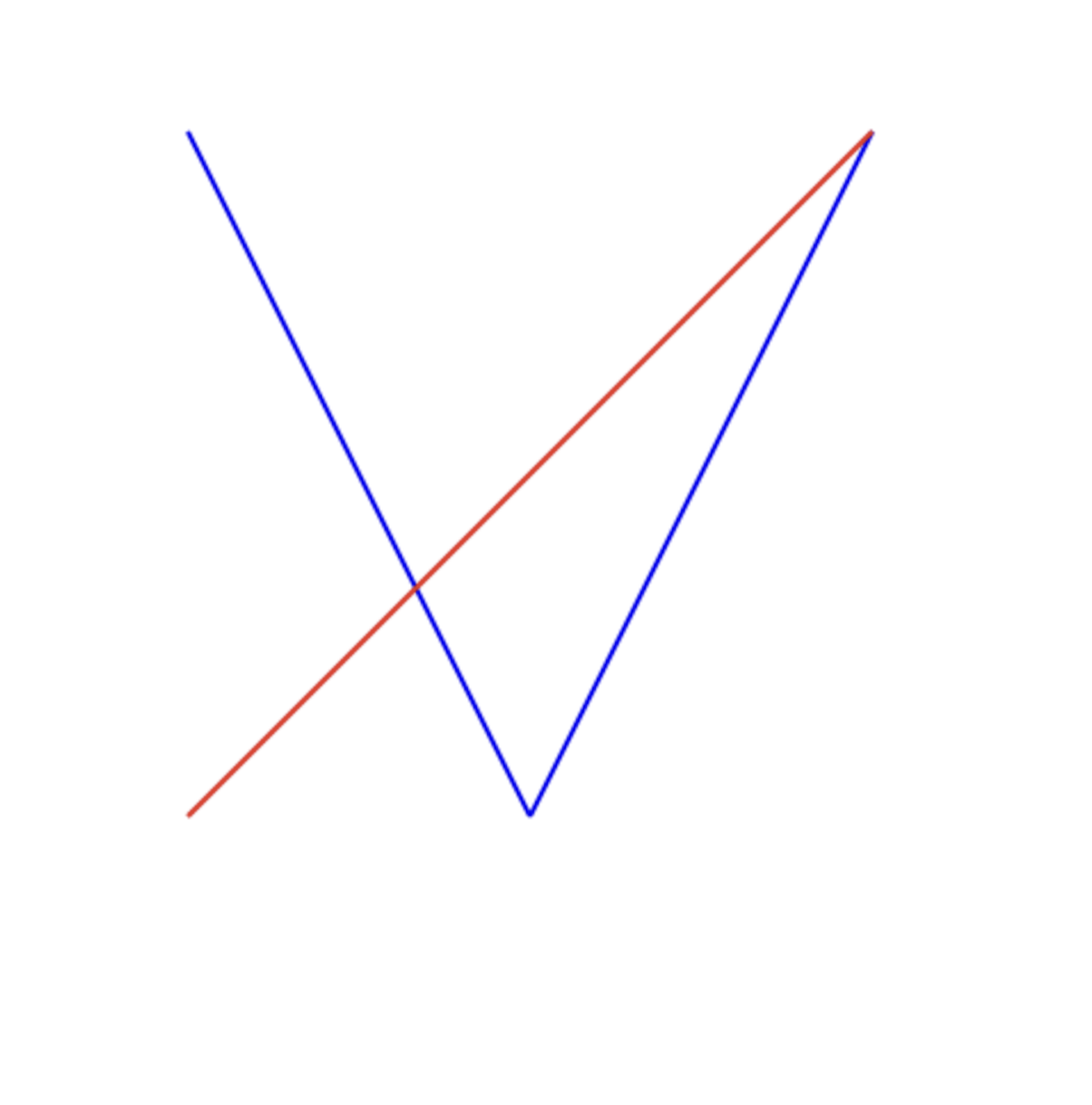}
         &
         \includegraphics[width=0.19\linewidth]{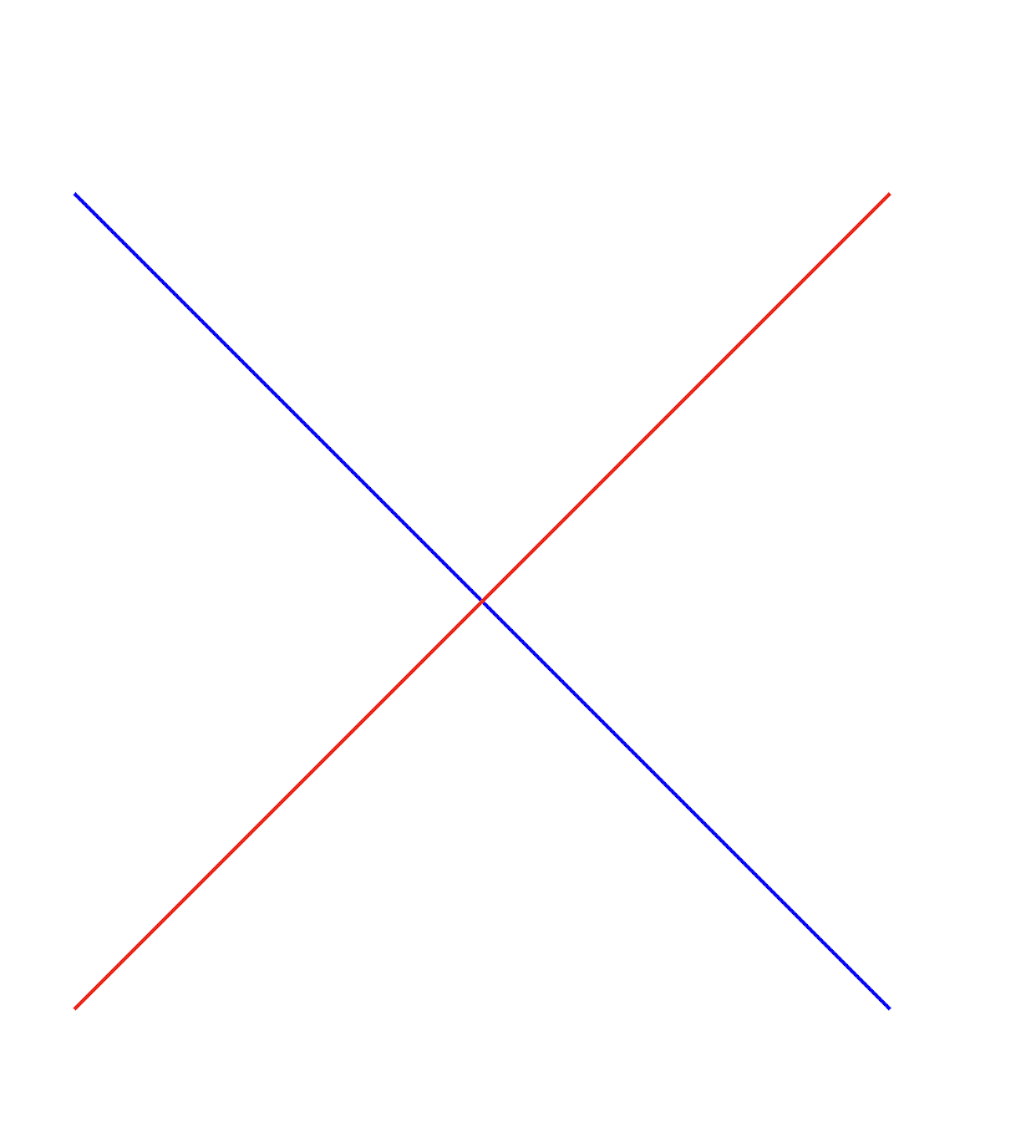}
         \\
         \includegraphics[width=0.19\linewidth]{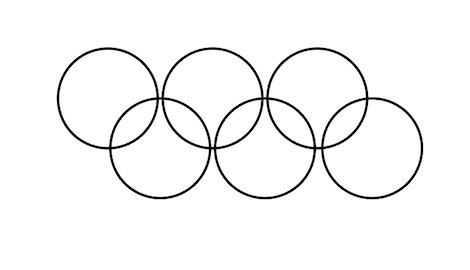}
         &
         \includegraphics[width=0.19\linewidth]{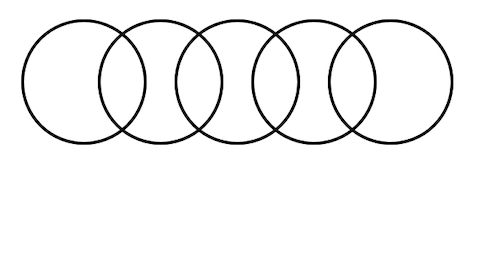}
         &
         \includegraphics[width=0.19\linewidth]{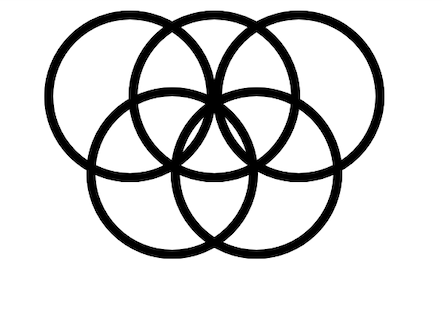}
         &
         \includegraphics[width=0.19\linewidth]{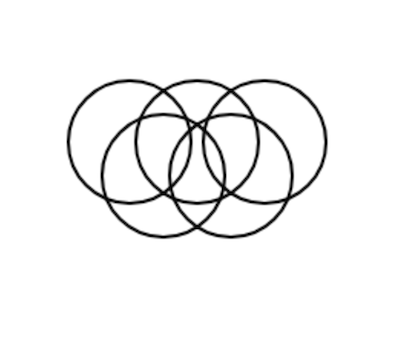}
         &
         \includegraphics[width=0.19\linewidth]{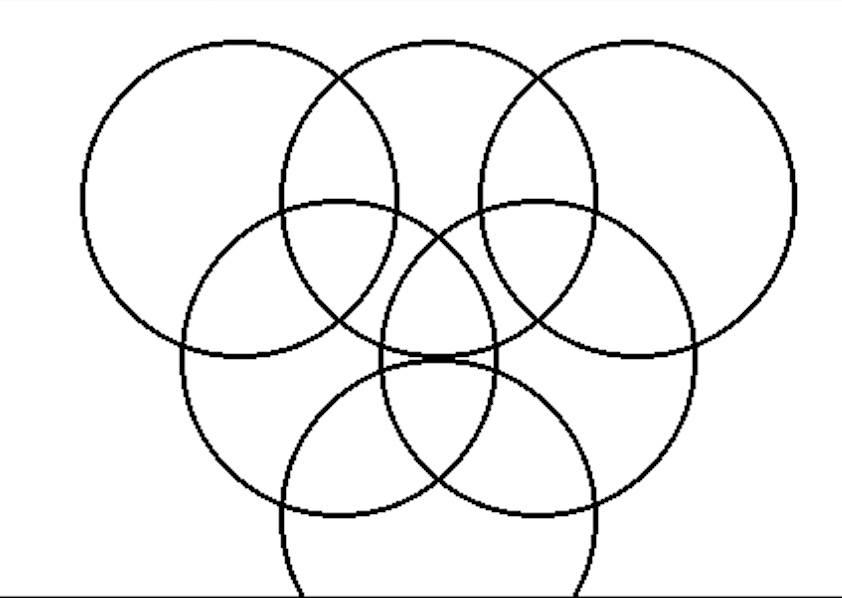}
         \\
         \includegraphics[width=0.19\linewidth]{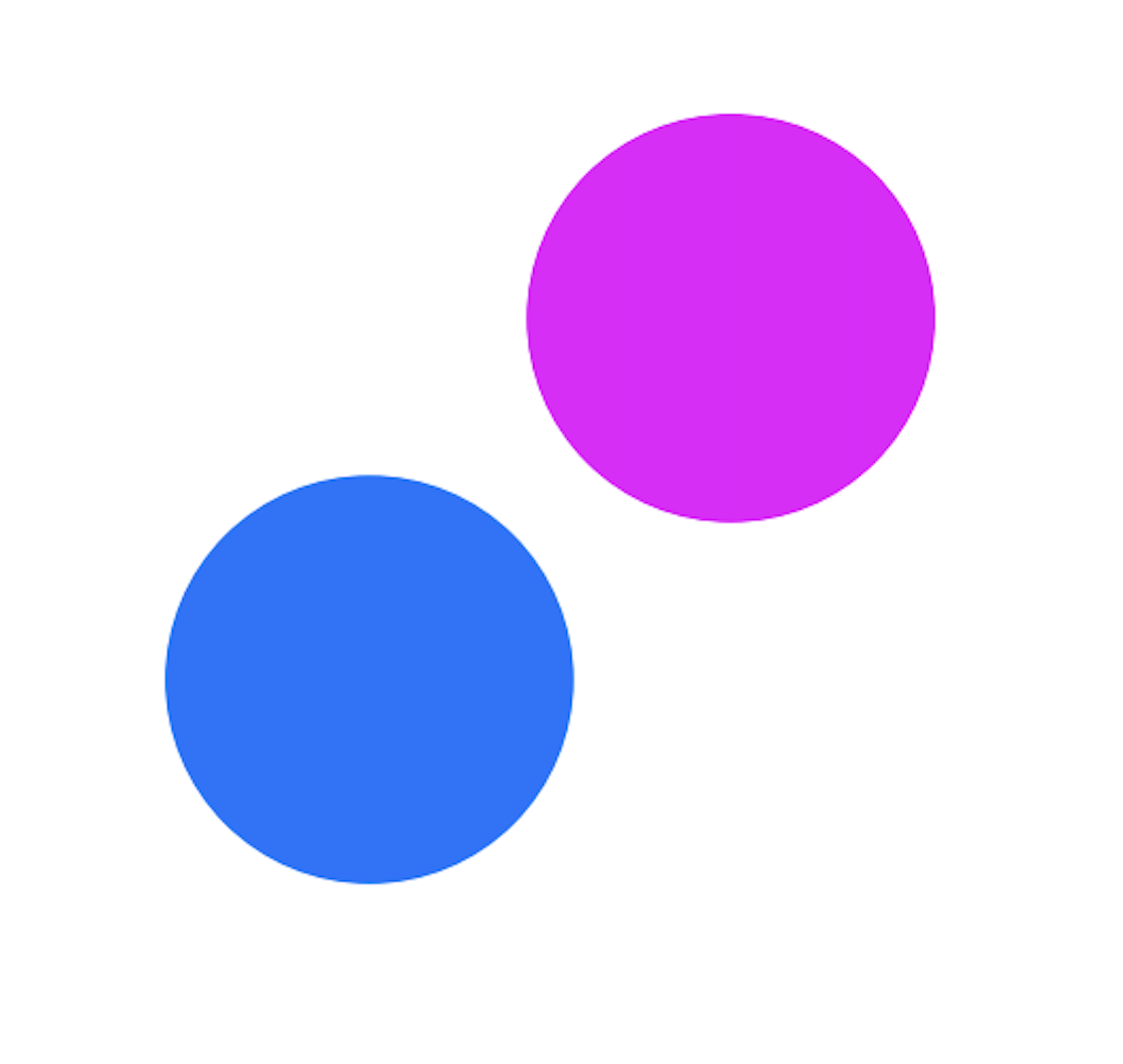}
         &
         \includegraphics[width=0.19\linewidth]{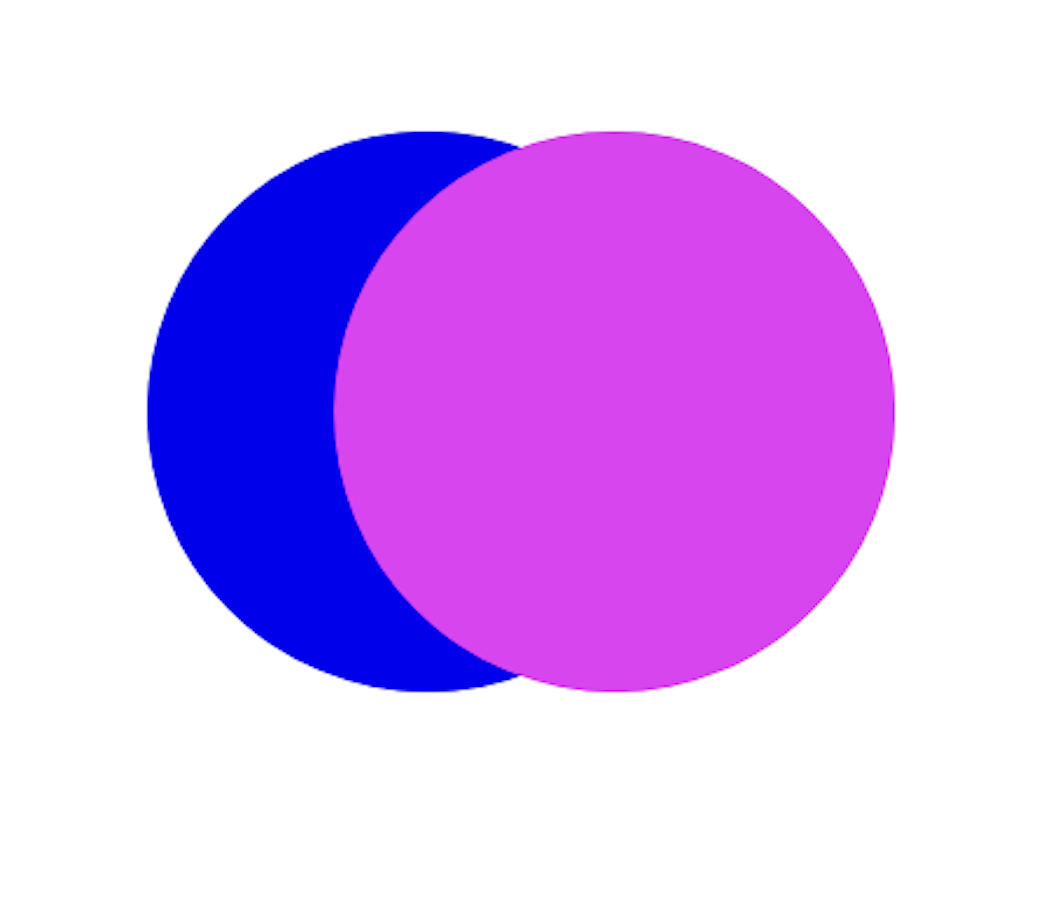}
         &
         \includegraphics[width=0.19\linewidth]{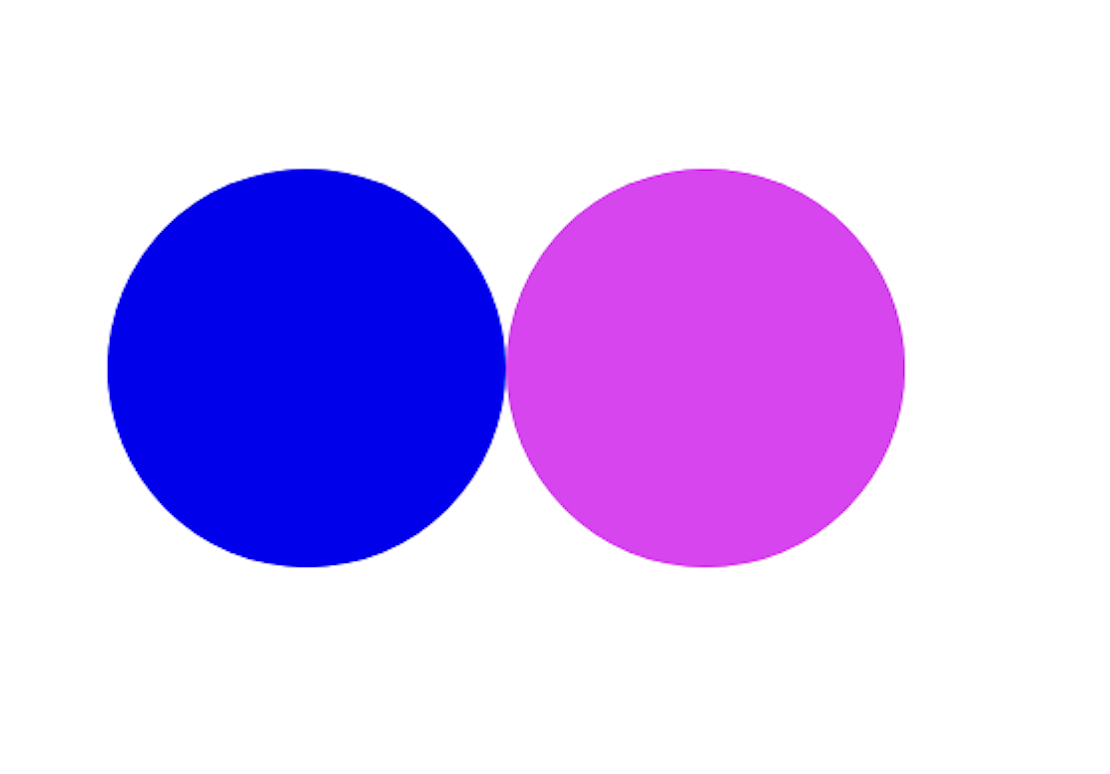}
         &
         \includegraphics[width=0.19\linewidth]{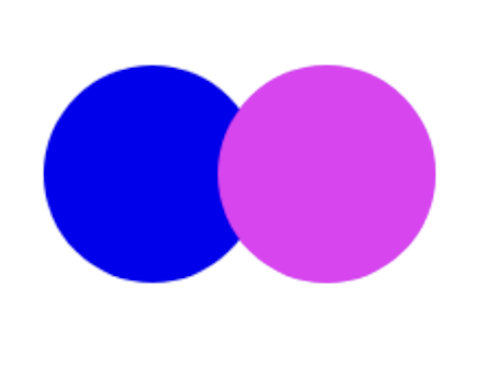}
         &
         \includegraphics[width=0.19\linewidth]{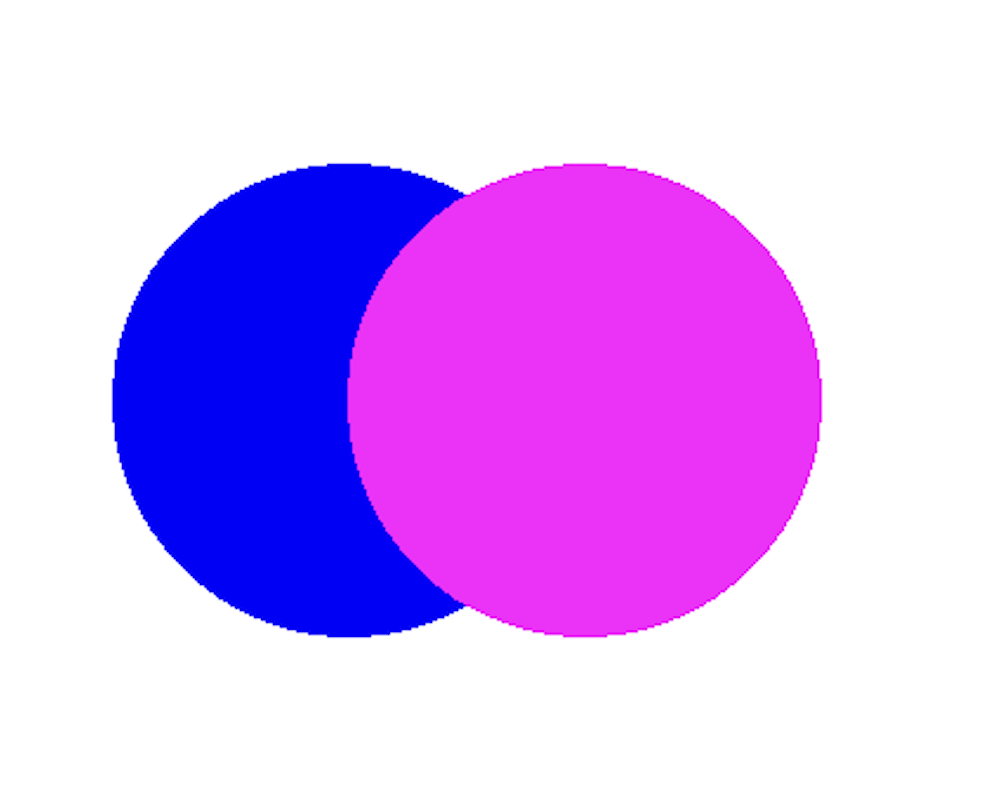}
         \\
         \includegraphics[width=0.19\linewidth]{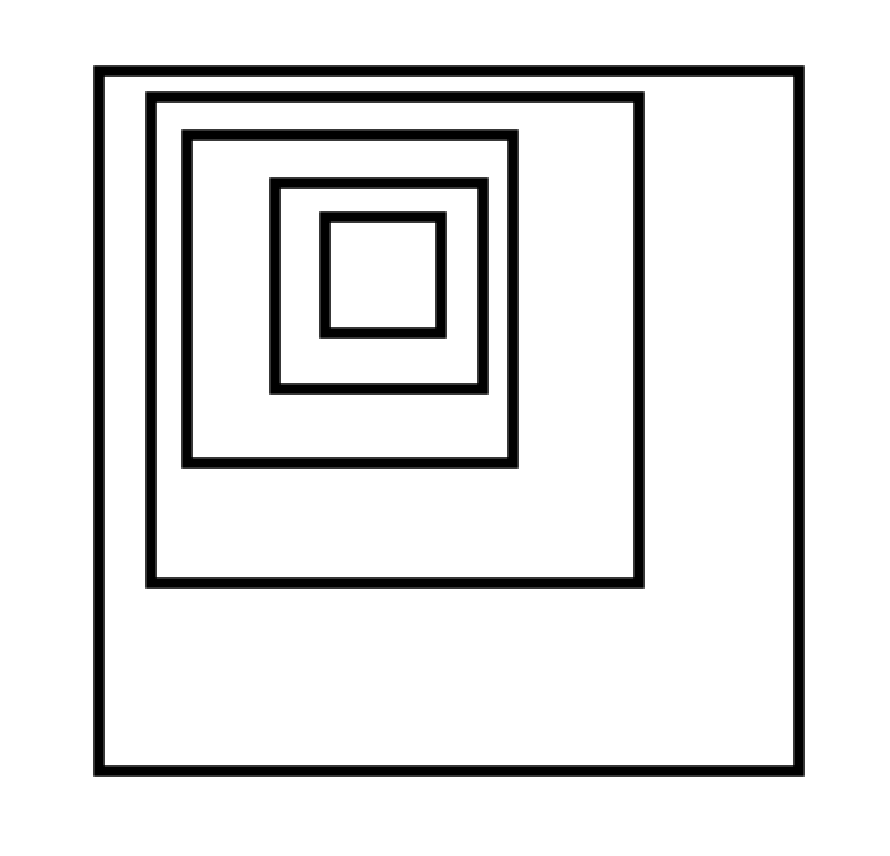}
         &
         \includegraphics[width=0.19\linewidth]{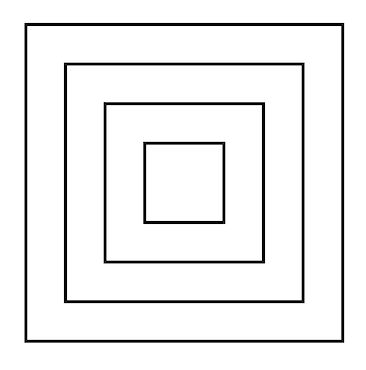}
         &
         \includegraphics[width=0.19\linewidth]{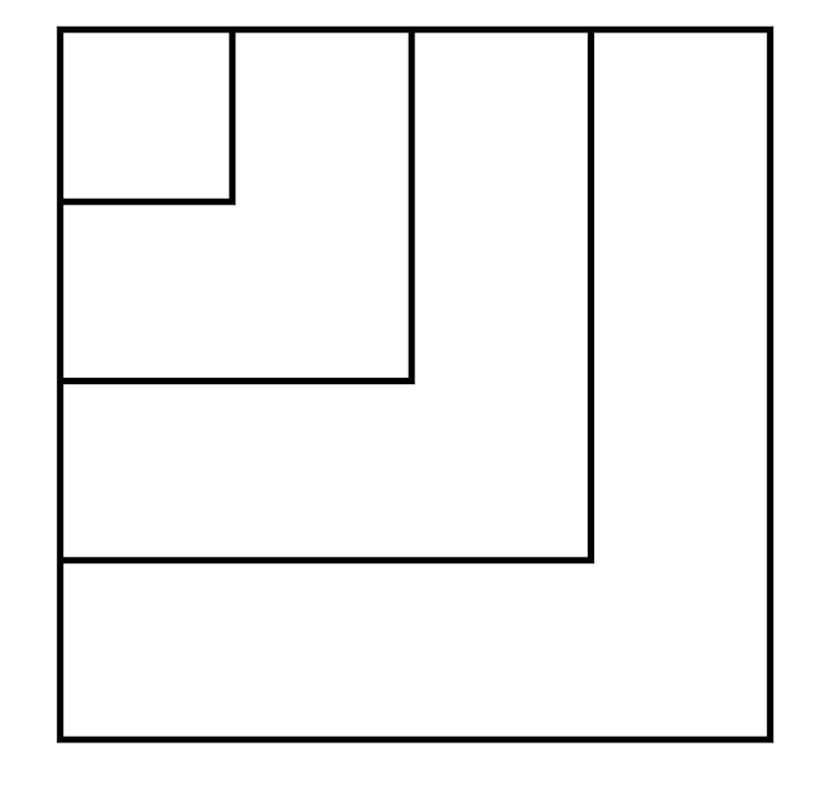}
         &
         \includegraphics[width=0.19\linewidth]{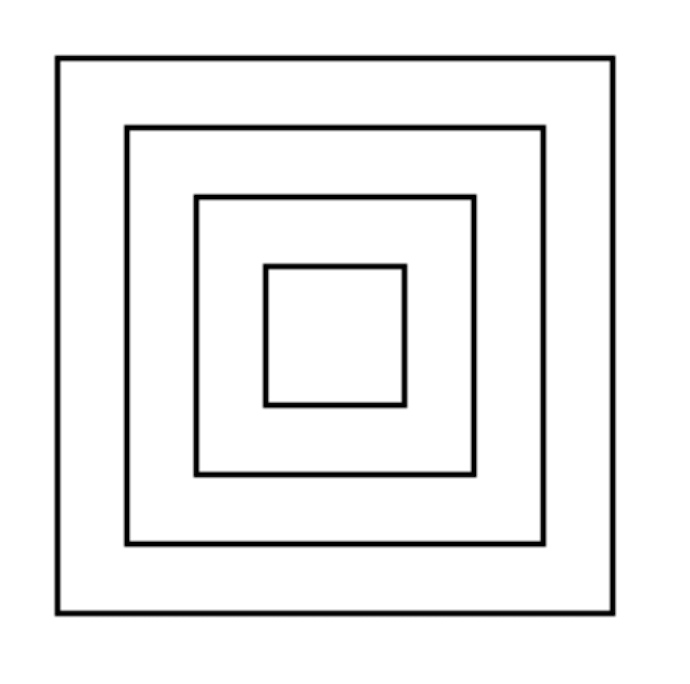}
         &
         \includegraphics[width=0.19\linewidth]{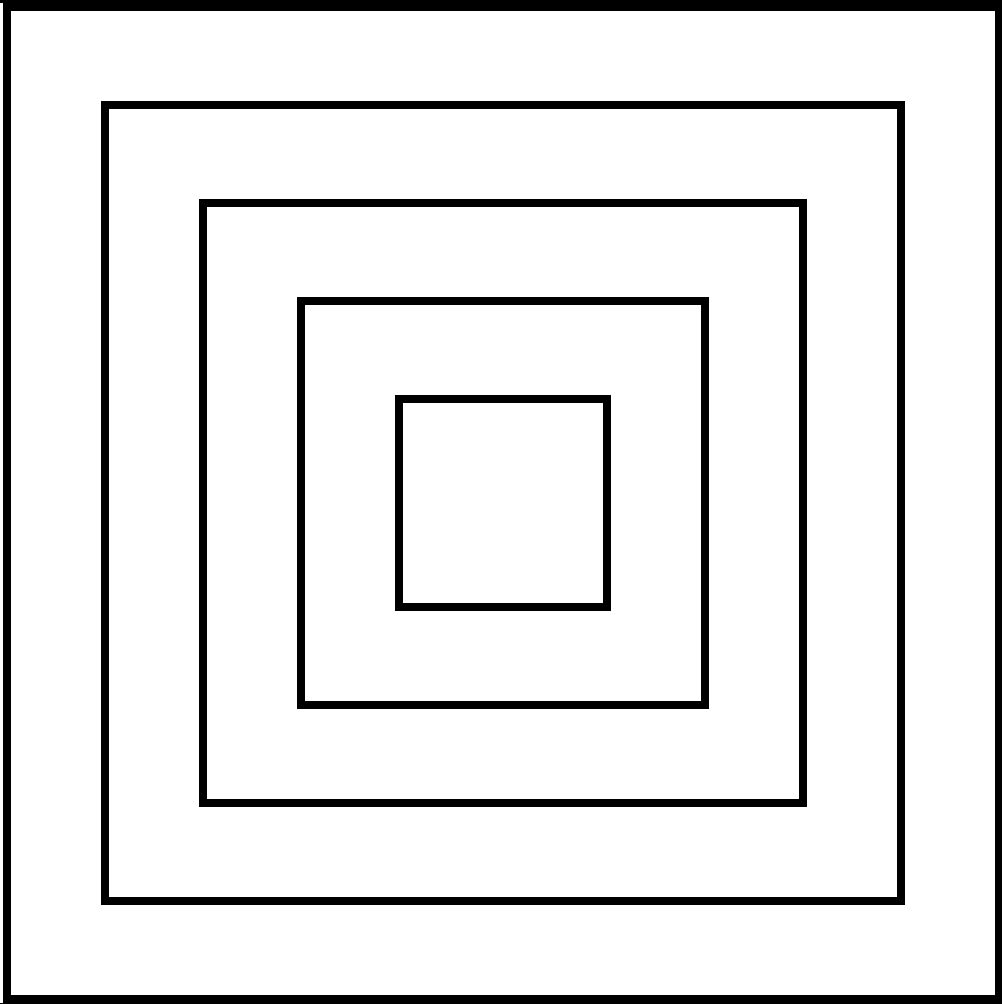}
         \\
         (a) Input image
         &
         (b) SVG
         &
         (c) HTML/CSS
         &
         (d) JavaScript
         &
         (e) Python \\
    \end{tabular}
    
    \caption{Sonnet-3.5 cannot recreate images from \blindtest (a) using its coding ability and different programming languages.}
    \label{fig:line-reconstruct}
\end{figure}

\subsec{Results} Overall, the coding ability does not help \newsonnet to reconstruct the images from \blindtest (see \cref{fig:line-reconstruct}). Furthermore, across various tasks, \newsonnet often cannot faithfully recreate the important features such as the number of intersections, number of shapes, and touching vs. non-touching circles. This shows that \newsonnet can see the overall shapes in the image, but it is not sufficient for code generation.


\clearpage

\section{Two touching circles task}
\label{app:touching-circles}

\subsection{Benchmark Construction Details}
\label{app:2circles-bench}

To create our benchmark, we use 5 parameters to control the diversity of the samples.

\begin{itemize}
    \item \textbf{Color}: We fix the colors for each circle to use \{\emph{magenta}, \emph{dodgerblue}\}.
    \item \textbf{Image size}: We use the physical size, and the DPI arguments in \emph{matplotlib} to initialize the image size. The physical size is fixed to $5\times 5$, and the DPI $\in\{100, 200, 300\}$. The output image sizes are \{384, 769, 1155\}px.
    \item \textbf{Diameter}: We use uniform diameters for both circles and choose the value proportional to the image size, where the diameter is \(\{\frac{1}{4}, \frac{1}{5}, \frac{1}{6}, \frac{1}{7}\}\) of the image size.
    \item \textbf{Distance}: The boundary-to-boundary distance between circles is a fraction of the diameter chosen from \{-0.25, -0.2, -0.15, -0.1, -0.05, 0.0, 0.05, 0.1, 0.15, 0.2, 0.25, 0.3, 0.35, 0.4, 0.45, 0.5\}. Based on our definition, center-to-center distance is (2$+$distance)$\times$diameter.
    \item \textbf{Rotation}: We include 2 main rotations (vertical and horizontal), and 2 diagonal rotations.
\end{itemize}

\begin{figure}
\centering
\begin{tabular}{c@{}cc@{}cc@{}c}
\multicolumn{2}{c}{(a) Rotation} & \multicolumn{2}{c}{(b) Diameter} & \multicolumn{2}{c}{(c) Distance} \\
\includegraphics[width=0.15\textwidth]{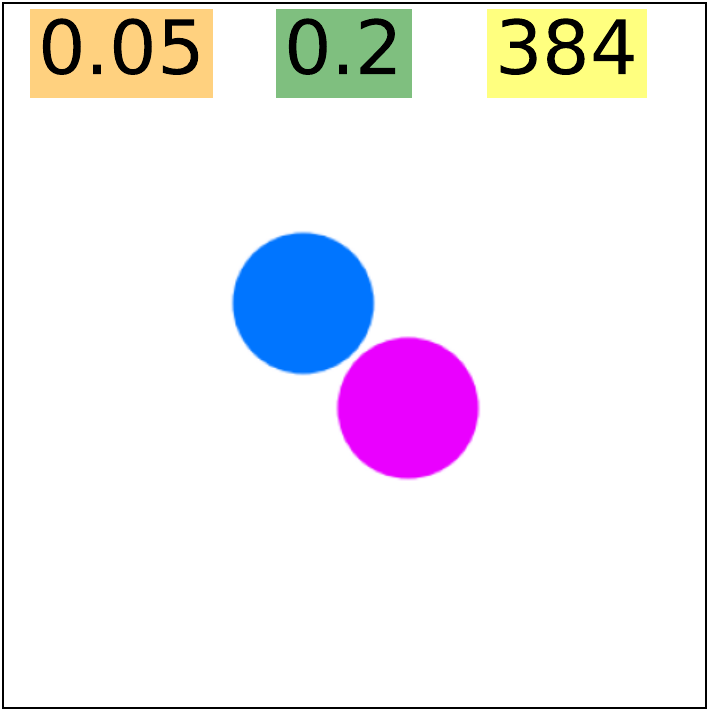} & \includegraphics[width=0.15\textwidth]{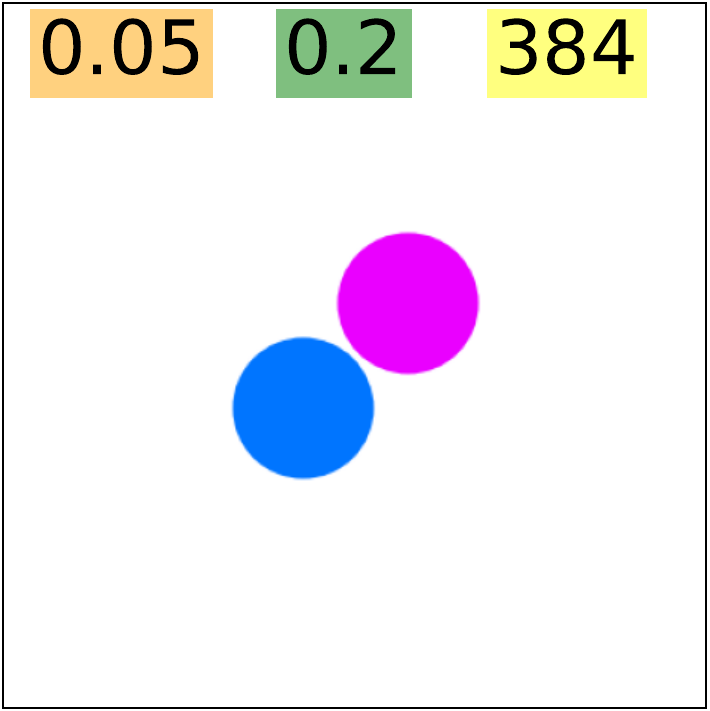} & \includegraphics[width=0.15\textwidth]{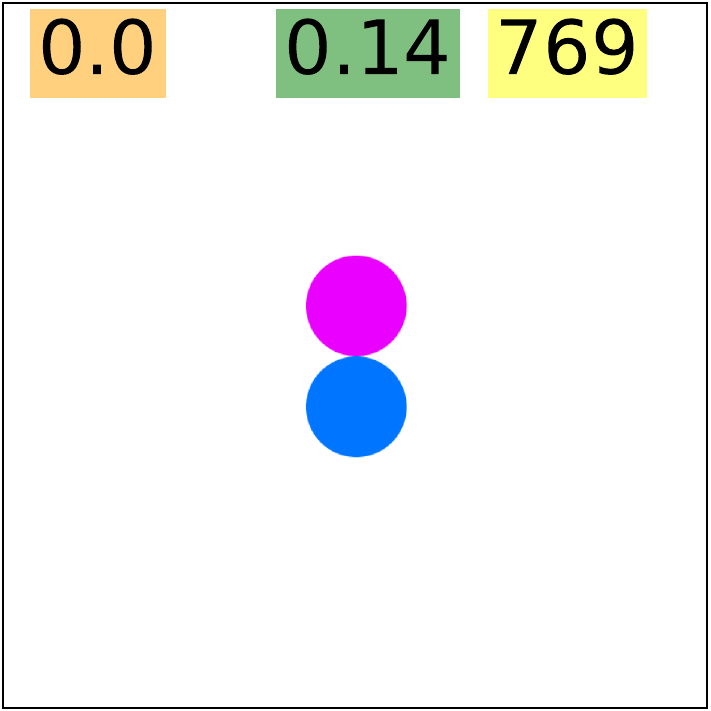} & \includegraphics[width=0.15\textwidth]{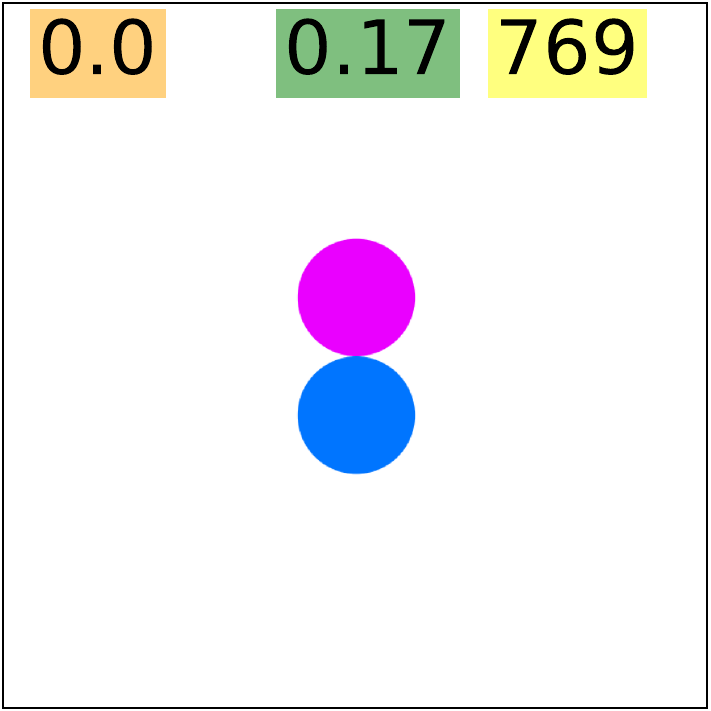} & \includegraphics[width=0.15\textwidth]{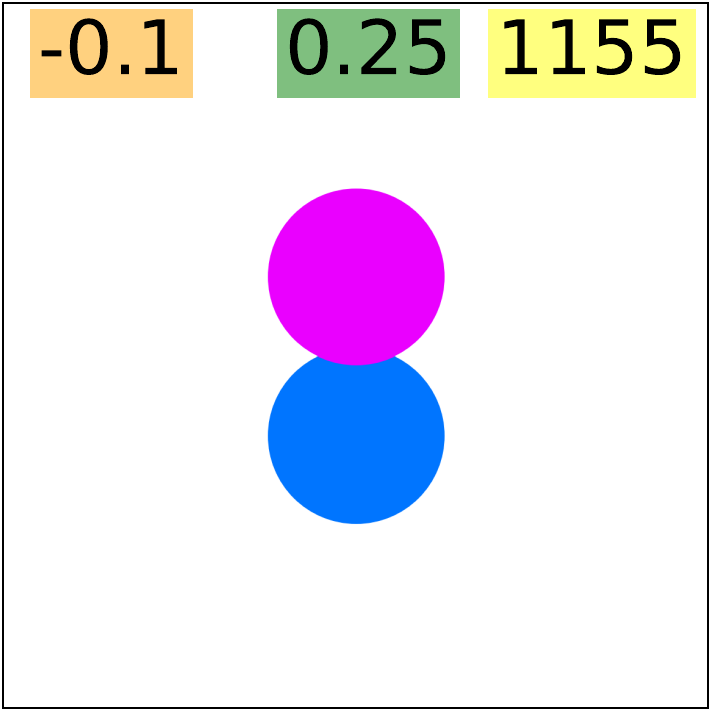} & \includegraphics[width=0.15\textwidth]{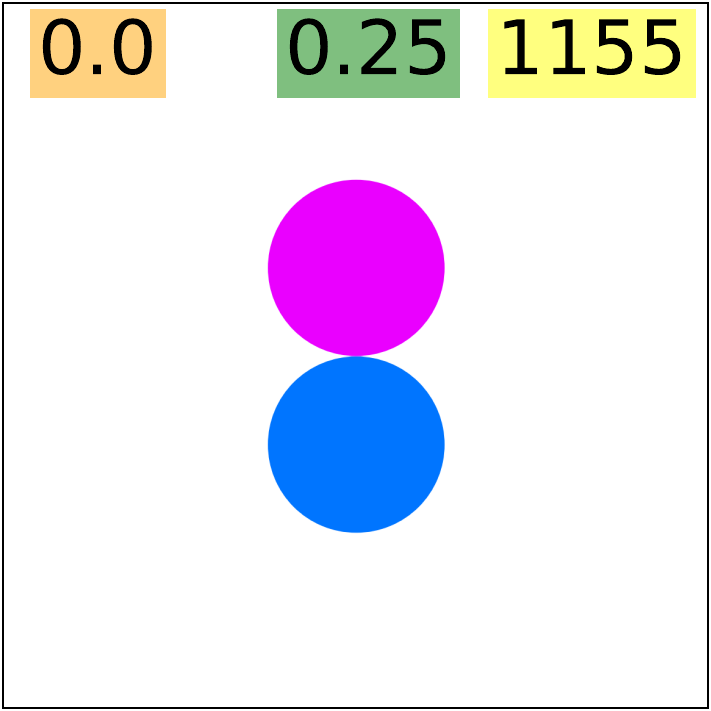} \\
\includegraphics[width=0.15\textwidth]{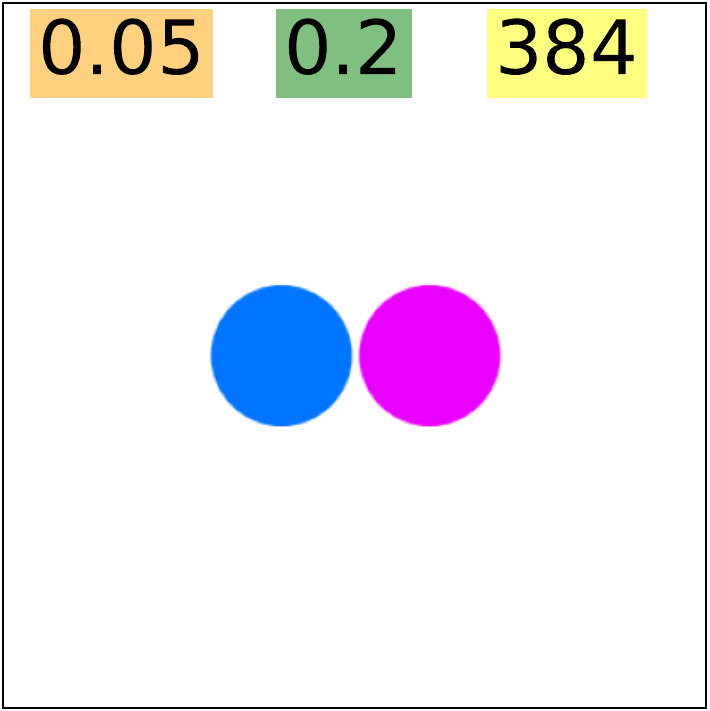} & \includegraphics[width=0.15\textwidth]{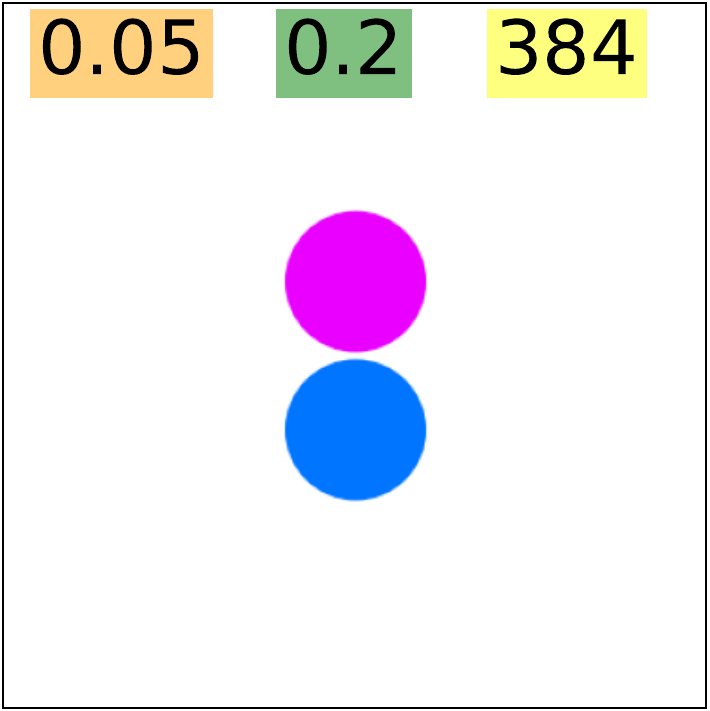} & \includegraphics[width=0.15\textwidth]{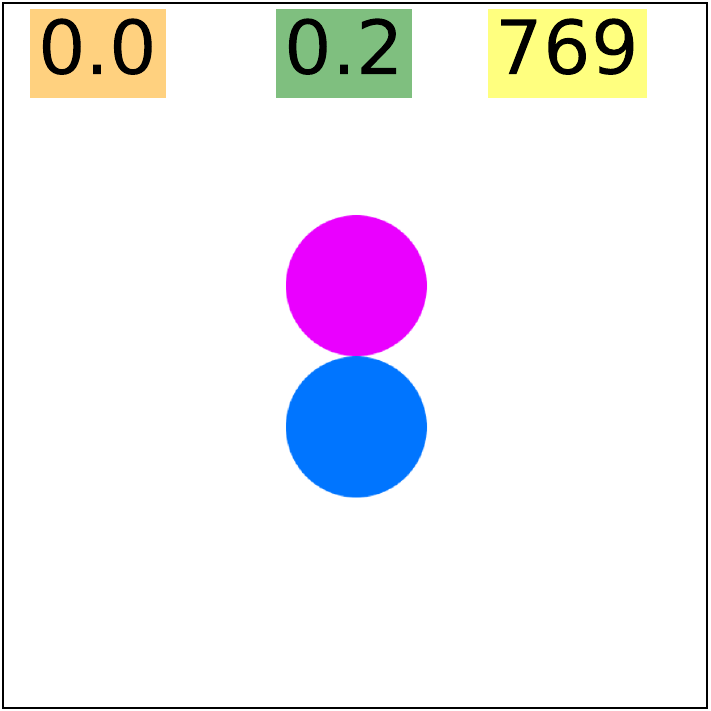} & \includegraphics[width=0.15\textwidth]{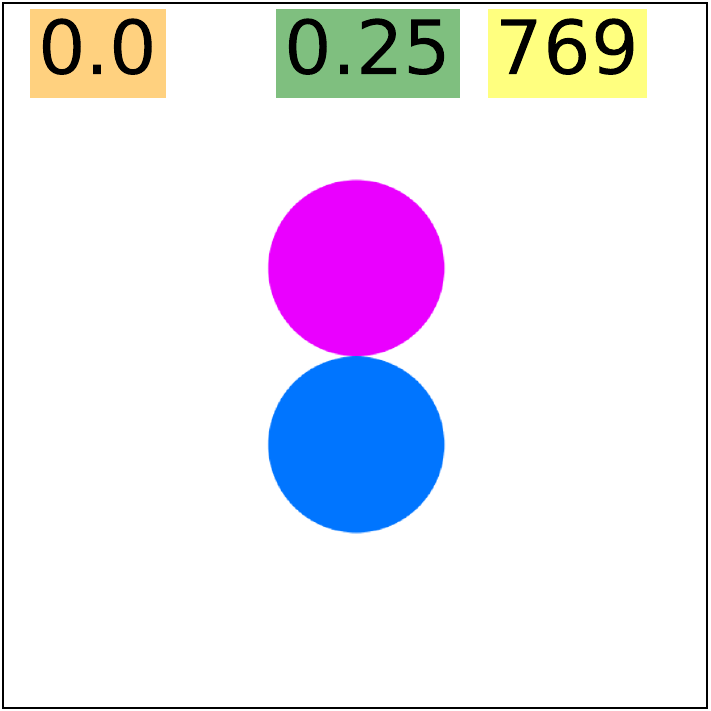} & \includegraphics[width=0.15\textwidth]{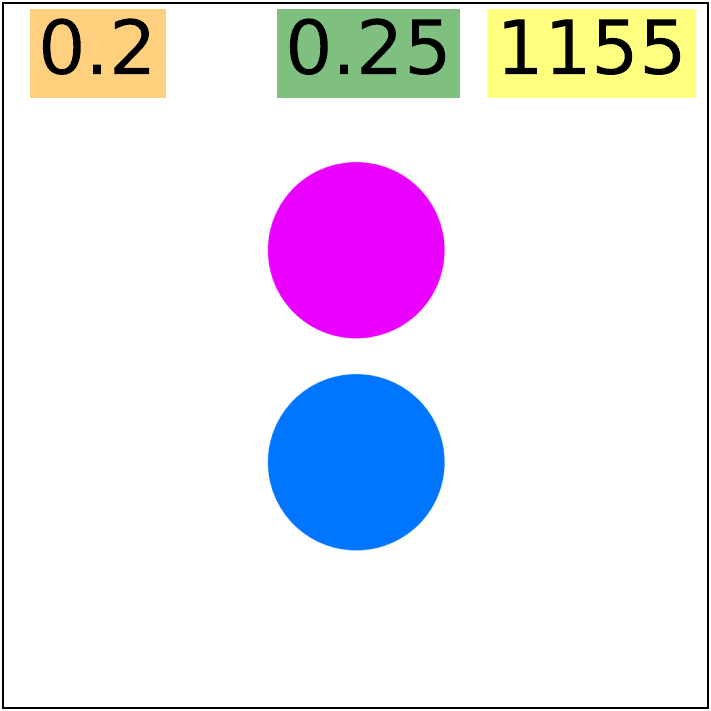} & \includegraphics[width=0.15\textwidth]{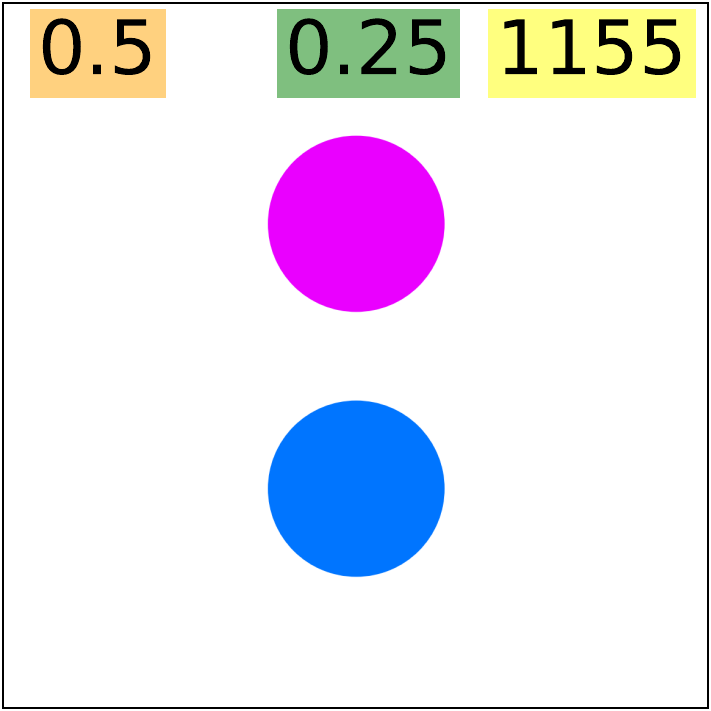}\\
\end{tabular}
\caption{Samples in the benchmark include various settings for drawing two circles. We start with choosing a rotation (a) and change other parameters of each plot, \eg, \sethlcolor{forestgreen(web)} \hl{the diameter} (b), \sethlcolor{amber} \hl{the distance} between perimeters (c), \sethlcolor{aureolin} and the \hl{image size} (in pixels). We show the parameters that can be changed to generate different samples inside the legend.}
\label{fig:2circles_examples}
\end{figure}

We use the center of the image as the origin so that it always aligns with the midpoint of distances between two circles. This systematic process results in a benchmark comprising 768 images (see \cref{table:2cir-b,fig:2circles_examples}).

\textbf{Code} The code is available at \url{https://github.com/anguyen8/vision-llms-are-blind/blob/main/src/TouchingCircle/TwoTouchingCircles.ipynb}.

\subsec{Prompts}
\begin{enumerate}
    \item \emph{Are the two circles touching each other? Answer with Yes/No.}
    
    \item \emph{Are the two circles overlapping? Answer with Yes/No.}
\end{enumerate}

\begin{figure}
\centering
\resizebox{\columnwidth}{!}{
\begin{tabular}{rc@{}c@{}c@{}ccc@{}c@{}c@{}c}
    &\multicolumn{4}{c}{\footnotesize (a) 4 diameter $\phi$}& &\multicolumn{4}{c}{\footnotesize (b) 4 orientations}\\
     & \includegraphics[trim={0 80 0 0},clip,width=0.12\textwidth]{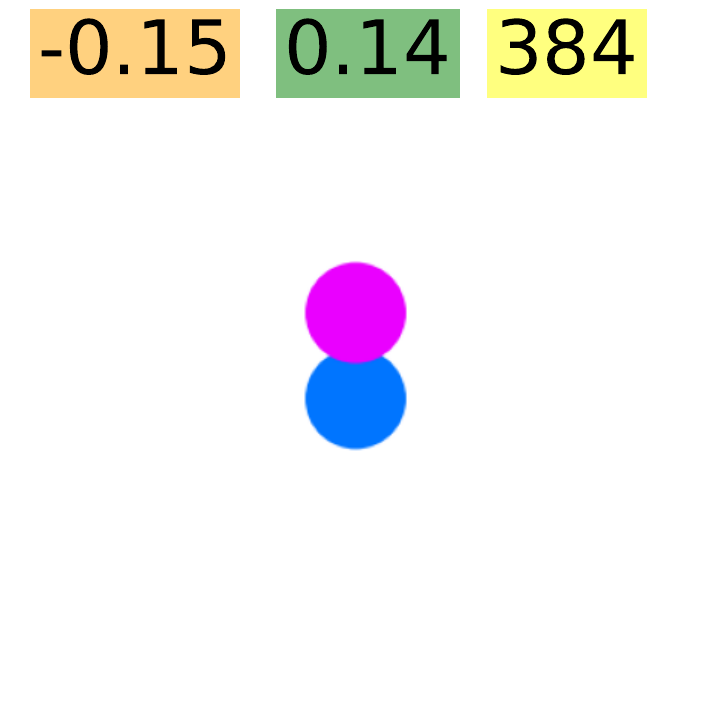}&\includegraphics[trim={0 80 0 0},clip,width=0.12\textwidth]{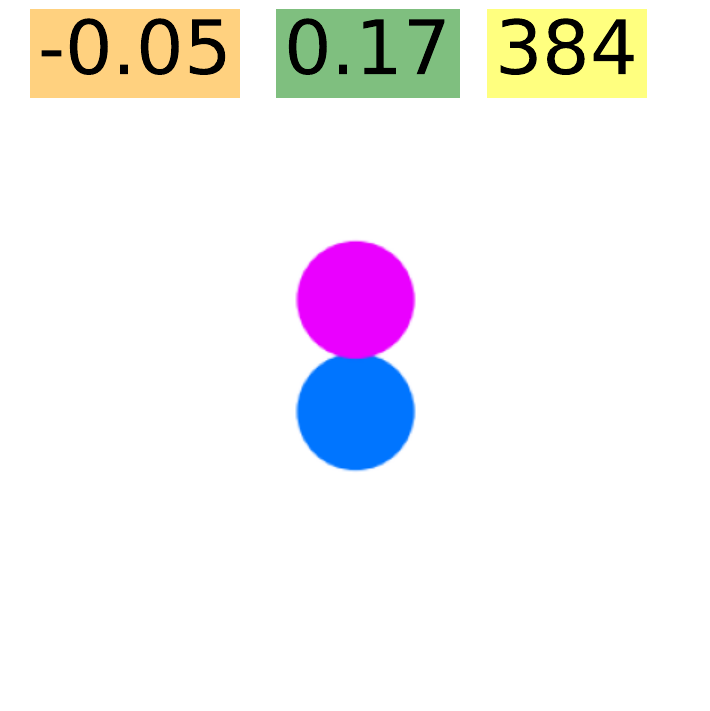}&\includegraphics[trim={0 80 0 0},clip,width=0.12\textwidth]{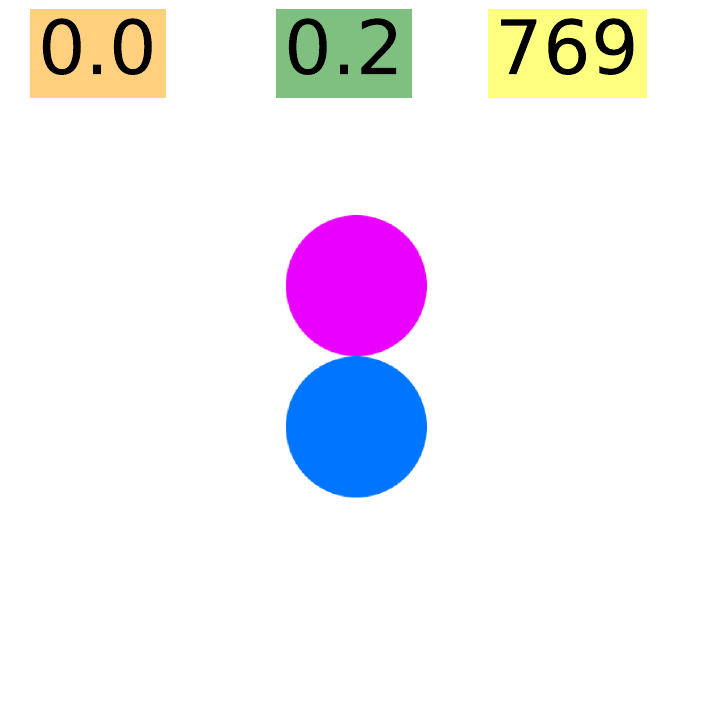}&\includegraphics[trim={0 80 0 0},clip,width=0.12\textwidth]{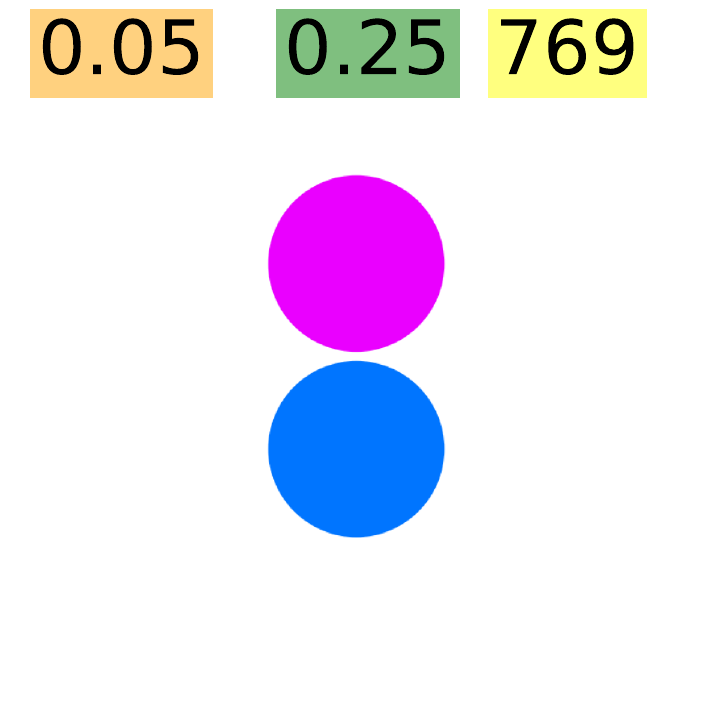}& & \includegraphics[trim={0 80 0 0},clip,width=0.12\textwidth]{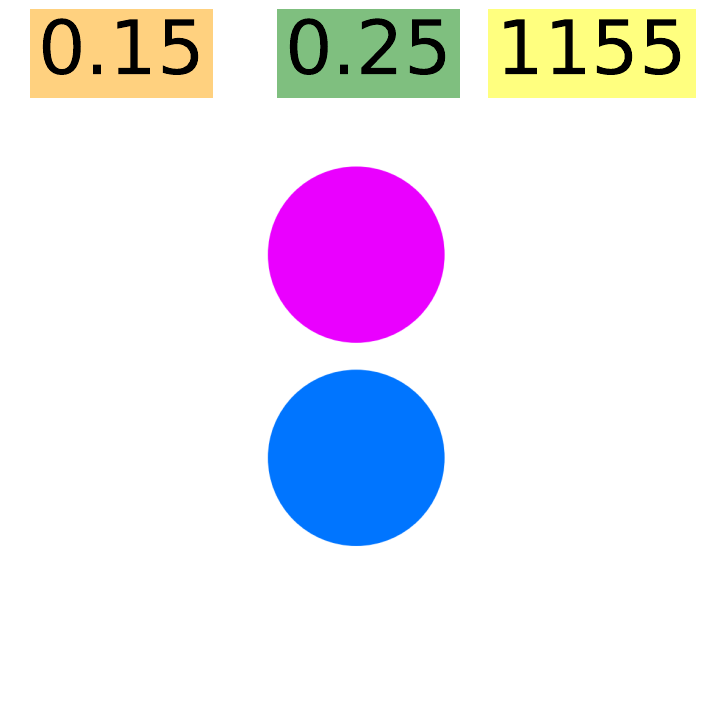}&\includegraphics[trim={0 80 0 0},clip,width=0.12\textwidth]{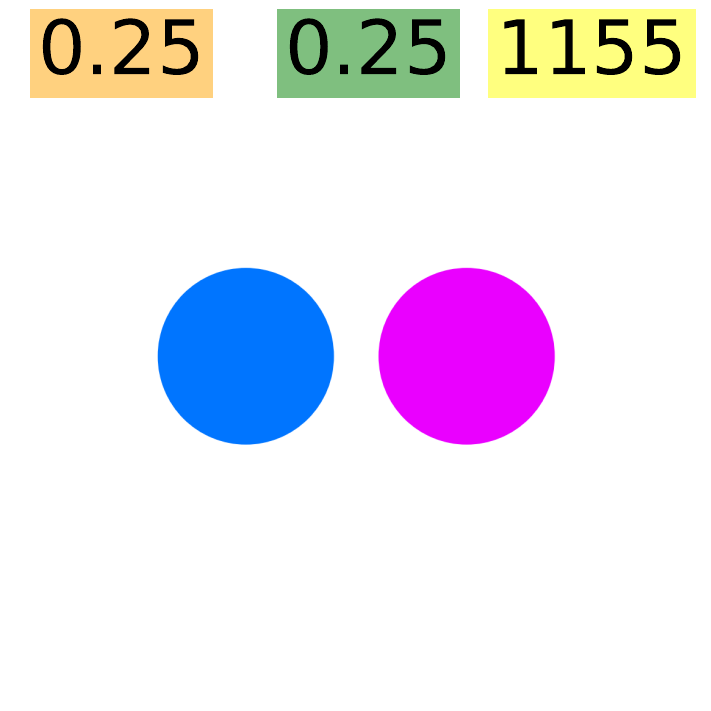}  &\includegraphics[trim={0 80 0 0},clip,width=0.12\textwidth]{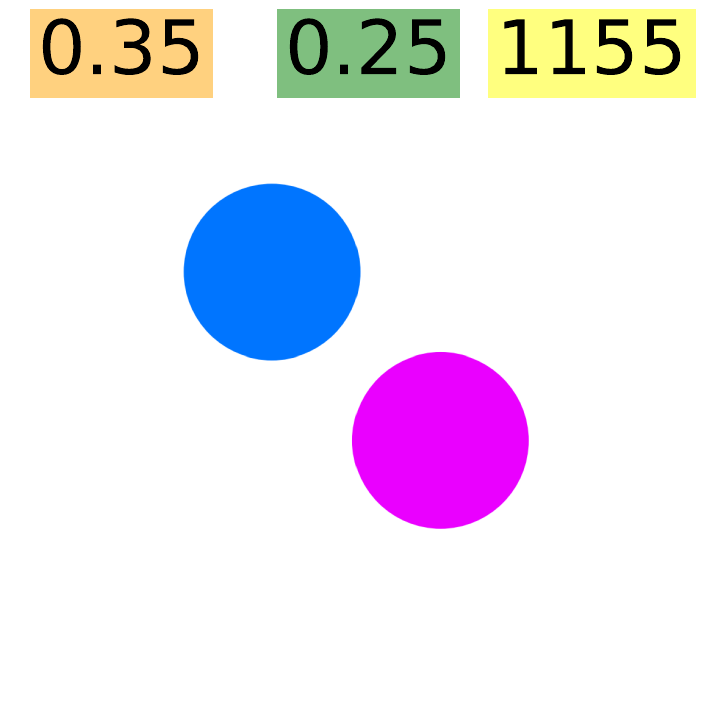}  &\includegraphics[trim={0 80 0 0},clip,width=0.12\textwidth]{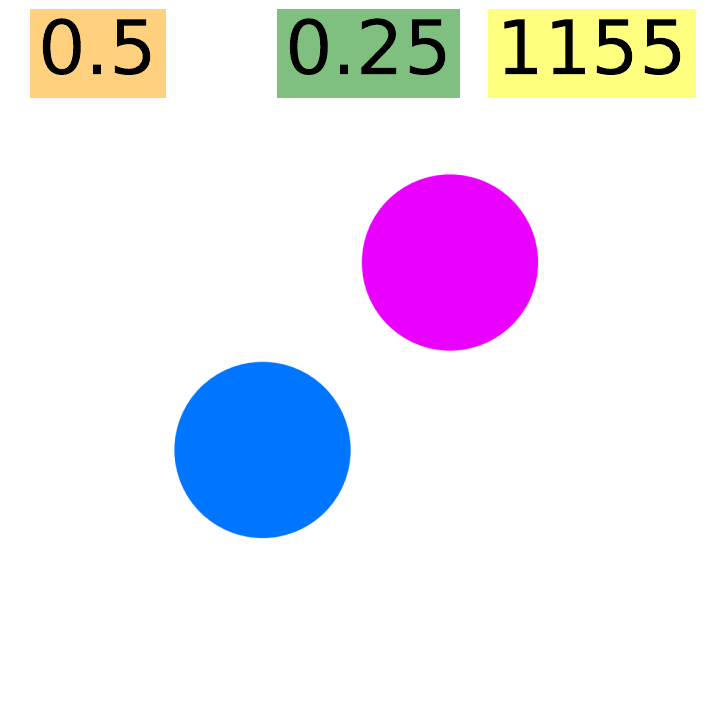} \\
     $d$ & -0.15 & -0.05 & 0.0 &  +0.05 & & +0.15 & +0.25  & +0.35 & +0.50 \\
     GT labels & \overlapping,\touching & \overlapping,\touching & \notoverlapping,\touching &  \notoverlapping,\nottouching &~~& \notoverlapping,\nottouching & \notoverlapping,\nottouching  & \notoverlapping,\nottouching & \notoverlapping,\nottouching \\
\end{tabular}}
\caption{For each image {\sethlcolor{yellow_sizes}\hl{size}} and {\sethlcolor{amber} \hl{distance}} $d$, we vary \sethlcolor{forestgreen(web)} \hl{diameter} (a) and orientation (b).
Groundtruth: 
\overlapping: overlapping. \touching: touching.
\notoverlapping: non-overlapping. \nottouching: non-touching.
}
\label{fig:2circles_examples_main}
\end{figure}

\subsec{Groundtruth} We consider two circles overlapping and touching (\overlapping, \touching) if $d < 0.0$; non-overlapping but touching (\notoverlapping, \touching) if $d = 0.0$; and non-overlapping \& non-touching (\notoverlapping, \nottouching) when $d > 0.0$ (\cref{fig:2circles_examples_main}).
Random-baseline accuracy: 50\%.

\begin{table}
\centering
\caption{Number of samples for each category is the same in our benchmark, where they sum to 768.}
\label{table:2cir-b}
\begin{tabular}{lccc}
\toprule
\textbf{Parameter} & \textbf{Values} & \textbf{Samples} & ~~\textbf{Total Samples}\\ 
\midrule
\multirow{3}{*}{Image size ($C$)} & 384px & 256 &~~\multirow{3}{*}{768}\\ 
                            & 769px & 256 \\ 
                            & 1155px & 256 \\ 
\midrule
\multirow{7}{*}{Diameter ($\phi$)} & $\frac{C}{4}$ & 192 & ~~\multirow{7}{*}{768}\\ 
\\
                                                                    & $\frac{C}{5}$ & 192 \\ 
                                                           \\         & $\frac{C}{6}$ & 192 \\ 
                                                            \\        & $\frac{C}{7}$ & 192 \\ 
\midrule
\multirow{17}{*}{Distance} & -0.25$\times\phi$ & 48 & ~~\multirow{17}{*}{768}\\ 
                                                                     & -0.2$\times\phi$ & 48 \\
                                                                     & -0.15$\times\phi$ & 48 \\
                                                                     & -0.1$\times\phi$ & 48 \\ 
                                                                     & -0.05$\times\phi$ & 48 \\ 
                                                                     & 0.0$\times\phi$ & 48 \\ 
                                                                     & 0.05$\times\phi$ & 48 \\ 
                                                                     & 0.1$\times\phi$ & 48 \\ 
                                                                     & 0.15$\times\phi$ & 48 \\ 
                                                                     & 0.2$\times\phi$ & 48 \\ 
                                                                     & 0.25$\times\phi$ & 48 \\ 
                                                                     & 0.3$\times\phi$ & 48 \\ 
                                                                     & 0.35$\times\phi$ & 48 \\ 
                                                                     & 0.4$\times\phi$ & 48 \\ 
                                                                     & 0.45$\times\phi$ & 48 \\ 
                                                                     & 0.5$\times\phi$ & 48 \\ 
\midrule
\multirow{4}{*}{Rotation} & Vertical & 192 & ~~\multirow{4}{*}{768}\\ 
                          & Horizontal & 192 \\ 
                          & Diagonal 1 & 192 \\ 
                          & Diagonal 2 & 192 \\ 
\bottomrule
\end{tabular}
\end{table}

\clearpage
\subsection{Finding: image resolution does not affect VLMs performance}
\label{app:2circle-resolution}

\cref{fig:all-2cricles}-left shows that VLMs are almost invariant to the image resolution. For example, \gpt and \sonnet's performance saturates at 769px, and \newsonnet slightly performs worse at 769 and 1155px compared to 384px. \gemini, however, is fairly consistent across different resolutions. Based off these results, we conclude that VLMs' ability to see the intersection of two circles does not depend on the quality of the image.

\subsection{Finding: the vertical rotation closes the gap between models' performance}
\label{app:2circle-vertical}

As shown in \cref{fig:all-2cricles}-middle, arranging the circles in vertical rotation causes the models to perform similarly on the benchmark. Although \gemini slightly performs better at diagonal and \newsonnet at horizontal rotation, VLMs perform relatively better at vertical rotation. This suggests that the task complexity due to various rotations is not the main source of low performance in VLMs.

\subsection{Finding: Increasing the distance improves the VLMs' accuracy}
\label{app:2circle-simdist}

\begin{figure}
    \centering
        \includegraphics[width=\textwidth]{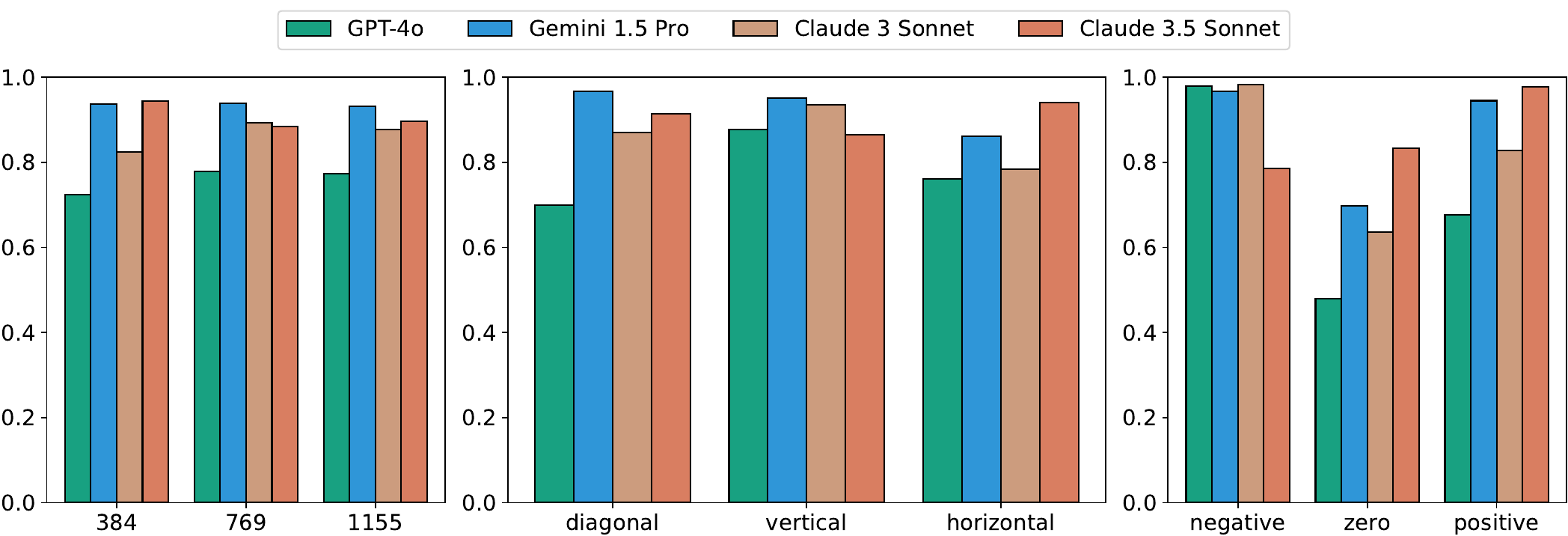}
    \caption{There is no correlation between the resolution of the image (left) and VLMs' performance. Across various rotations (middle), VLMs perform almost the same at vertical. Most failure cases are at boundary distances (right).}
    \label{fig:all-2cricles}
\end{figure}

VLMs perform better when the distance increases from zero to positive values (see \cref{fig:all-2cricles}-right). However, \newsonnet is more conservative than other VLMs that mostly answer ``Yes'', which results in its lowest performance at negative distances.

\clearpage
\subsection{Finding: VLMs prefer a specific rotation}
\label{app:2circle-rotation}
\cref{tab:2cricles} shows that VLMs prefer different rotations. For example, \gpt performs best at vertical, \gemini at diagonal, \newsonnet at horizontal, and \sonnet at vertical.

\begin{table}
\centering
\caption{
VLM accuracy is often best at a specific two-circle orientation.
Across three different resolutions, \gpt and \sonnet perform much better when two circles are arranged vertically.
In contrast, \gemini and \newsonnet prefer the diagonal and horizontal orientations, respectively.
}
\label{tab:twoCircles}
\begin{tabular}{lcccccccccc}
\toprule
&{Resolution} &{Rotation} && \raisebox{-.1\height}\gptlogo  && \raisebox{-.1\height}\geminilogo  && \raisebox{-.1\height}\sonnetlogo  && \raisebox{-.1\height}\newsonnetlogo \\
\midrule
\multirow{3}{*}{\emph{a}} & \multirow{3}{*}{384px} & \textsc{vertical}          && \textcolor{gpt_green}{\textbf{82.03}}  && 93.75          && \textcolor{sonnet3_brown}{\textbf{96.09}} && 90.62\\
                          &                        & \textsc{horizontal}        && 71.87  && 89.84          && 75.78 && \newsonnetnumber{96.87}\\
                          &                        & \textsc{diagonal}          && 69.97  && \gemininumber{95.70} && 78.90 && 94.92\\
\midrule
\multirow{3}{*}{\emph{b}} & \multirow{3}{*}{769px} & \textsc{vertical}          && \textcolor{gpt_green}{\textbf{89.84}}  && 95.31 && \textcolor{sonnet3_brown}{\textbf{92.97}} && 84.37\\
                          &                        & \textsc{horizontal}        && 80.47  && 85.94          && 83.59 && \newsonnetnumber{91.41}\\
                          &                        & \textsc{diagonal}          && 70.70  && \gemininumber{97.26} && 90.23 && 89.06\\
\midrule
\multirow{3}{*}{\emph{c}} & \multirow{3}{*}{1155px}& \textsc{vertical}          && \gptnumber{91.41}  && 96.09 && 91.41 && 84.37\\
                          &                        & \textsc{horizontal}        && 75.78  && 82.81          && 75.78 && \newsonnetnumber{93.75}\\
                          &                        & \textsc{diagonal}          && 71.09  && \gemininumber{96.87} && \textcolor{sonnet3_brown}{\textbf{91.80}} && 90.23\\
\bottomrule
\end{tabular}
\label{tab:2cricles}
\end{table}

\subsection{Finding: Various coloring of the two circles has minimal effects on the overall accuracy}
\label{app:color-circles}
VLMs' performance in the \twocirclelogo task does not change substantially ($\pm 5.79\%$ for \newsonnet and $\pm$10.81\% for \gpt) when tested against different colors (\cref{tab:color-circ}).

\begin{table}
    \centering
    \caption{The small $\triangle$ in VLMs' performance by changing the colors of the circles suggests that coloring does not significantly impact the VLMs' vision capabilities.}
    \begin{tabular}{lcc}
        \toprule
         Color &~~\newsonnetlogo&~~\gptlogo\\
         \midrule
         Magenta - Blue (baseline) & ~~90.93 & ~~76.10\\
         Red - Black & ~~96.72 &~~86.91\\
         Yellow - Green & ~~94.39 &~~80.52\\
         Dark gray - Light gray & ~~93.62 & ~~82.25\\
         \bottomrule
    \end{tabular}
    \label{tab:color-circ}
\end{table}

\clearpage
\subsection{Results for fine-tuning Bunny on the two touching circle}
\label{app:bunny}

In order to determine if fine-tuning could improve the model's performance on this task we attempted to fine-tune Bunny~\cite{he2024efficientApp} (Bunny-v1.1-Llama-3-8B-V with the \href{https://huggingface.co/BAAI/Bunny-v1_1-Llama-3-8B-V}{original weights}) on the two touching circles task. We fine-tuned Bunny using datasets of sizes: 10K, 20K, 50K, and 100K samples, each containing a balanced number of instances where the circles are either overlapping or separate (equal number of YES/NO answers in the training set).

The baseline model, without any fine-tuning, achieved 17.1\% accuracy for task overlap and 11.7\% for touching circles. After fine-tuning, we observed improvements with smaller datasets, such as 10K and 20K cases, where accuracy reached up to 36.8\%. However, increasing the number of samples did not lead to better performance. In some instances, such as with the 50K dataset, the model failed to predict anything and only generated the \textit{end-of-text} token.

The loss values for all these experiments were very close to zero, indicating that the model overfits the training set but fails to generalize. This suggests that training on this task is not straightforward and may require a combination of multiple tasks or that this problem does not have a simple solution.

\begin{figure}[ht]
    \centering
    \begin{subfigure}[b]{0.9\textwidth}
        \centering
        \includegraphics[width=\textwidth]{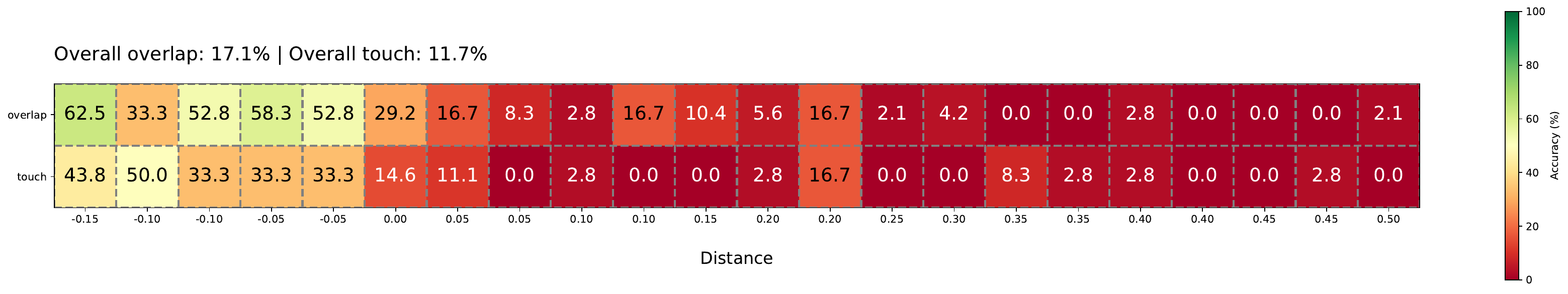}
        \caption{Accuracy by distance without fine-tuning}
        \label{appfig:bunny_no_fine_tuning}
    \end{subfigure}
    \hfill
    \begin{subfigure}[b]{0.9\textwidth}
        \centering
        \includegraphics[width=\textwidth]{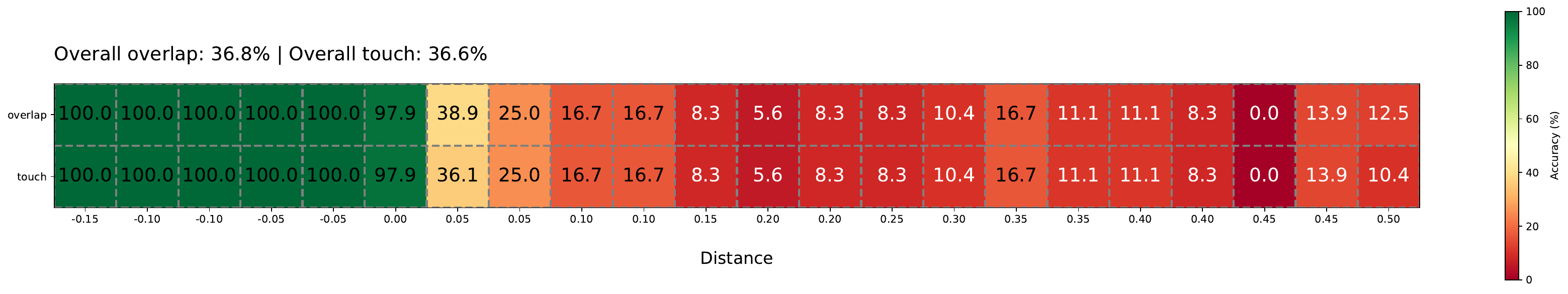}
        \caption{Accuracy by distance with 10K samples}
        \label{appfig:bunny_finetuned_10k}
    \end{subfigure}
    
    \begin{subfigure}[b]{0.9\textwidth}
        \centering
        \includegraphics[width=\textwidth]{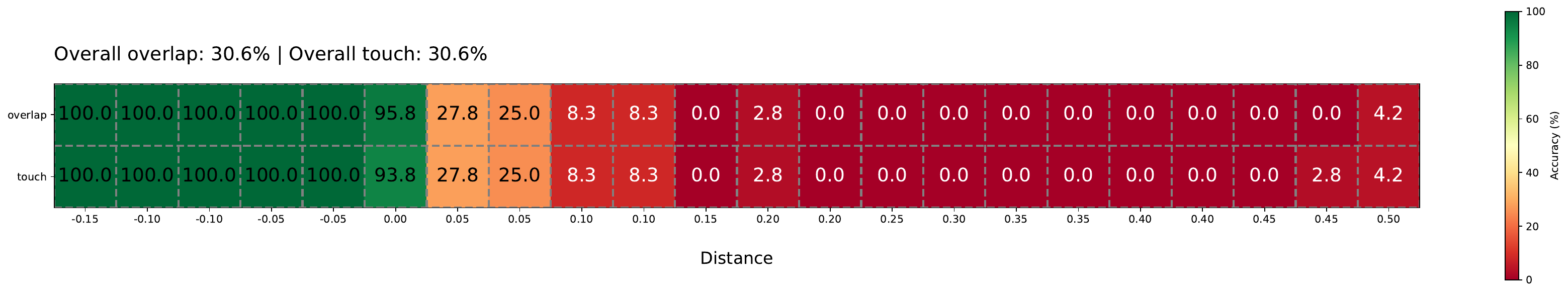}
        \caption{Accuracy by distance with 20K samples}
        \label{appfig:bunny_finetuned_20k}
    \end{subfigure}
    \hfill
    \begin{subfigure}[b]{0.9\textwidth}
        \centering
        \includegraphics[width=\textwidth]{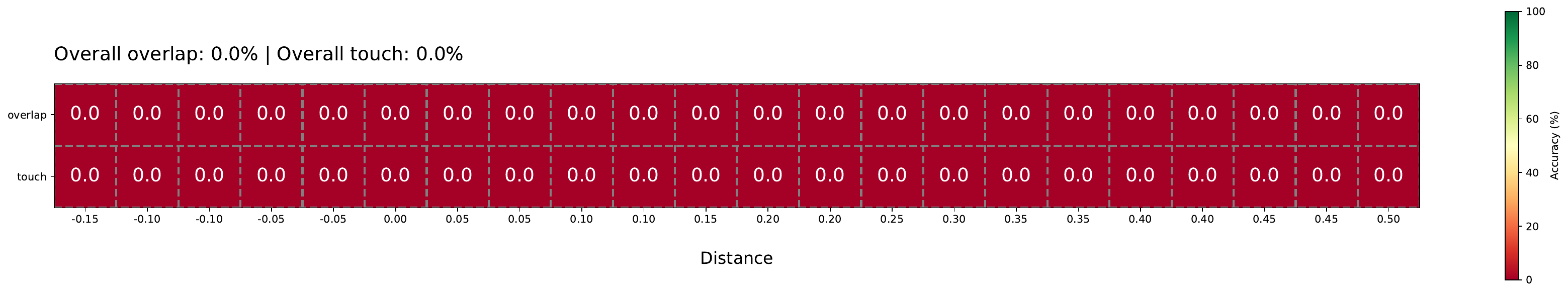}
        \caption{Accuracy by distance with 50K samples}
        \label{appfig:bunny_finetuned_50k}
    \end{subfigure}
    
    \begin{subfigure}[b]{0.9\textwidth}
        \centering
        \includegraphics[width=\textwidth]{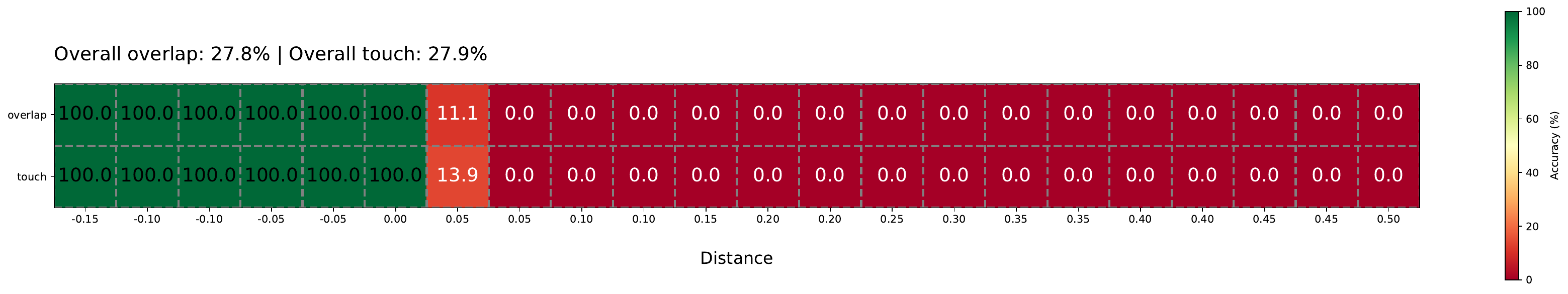}
        \caption{Accuracy by distance with 100K samples}
        \label{appfig:bunny_finetuned_100k}
    \end{subfigure}

    \caption{Comparison of Bunny's accuracy by distance with and without fine-tuning}
    \label{fig:bunny_comparison}
\end{figure}

\cref{fig:bunny_comparison} shows the accuracy breakdown by distance of the two circles. The model's performance improves when the circles are overlapping, but when there is a long distance between them, the model does not generalize well and cannot provide accurate answers.

\clearpage
\subsection{Additional examples}
\label{app:2circle-qual}
\begin{figure}
\centering
\begin{AIbox}{Are the two circles touching each other? Answer with Yes/No.}
\centering

    \begin{tabular}{lp{0.8cm}c|p{0.8cm}c|p{0.8cm}c|p{0.8cm}c|p{0.8cm}c|p{0.8cm}c}
     &\multicolumn{2}{c}{\includegraphics[width=0.15\textwidth]{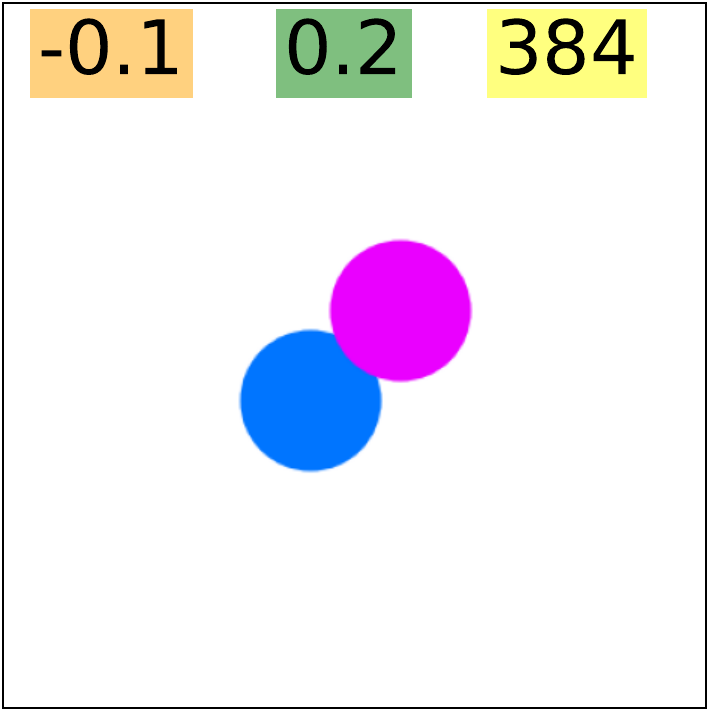}} & \multicolumn{2}{c}{\includegraphics[width=0.15\textwidth]{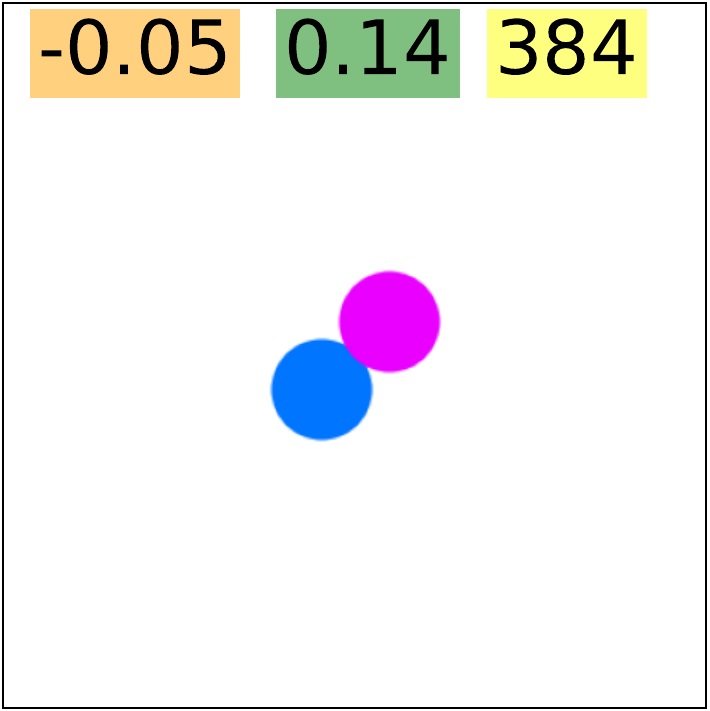}} &\multicolumn{2}{c}{\includegraphics[width=0.15\textwidth]{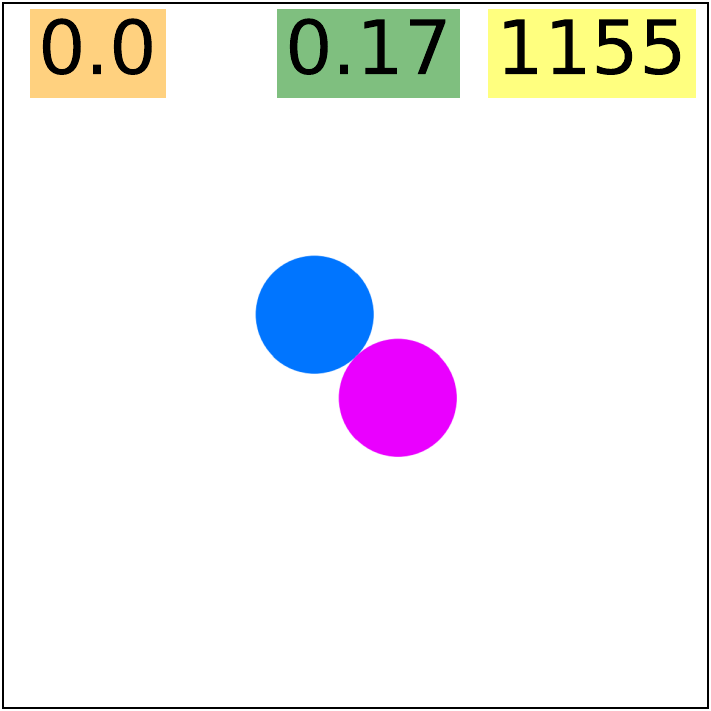}} &\multicolumn{2}{c}{\includegraphics[width=0.15\textwidth]{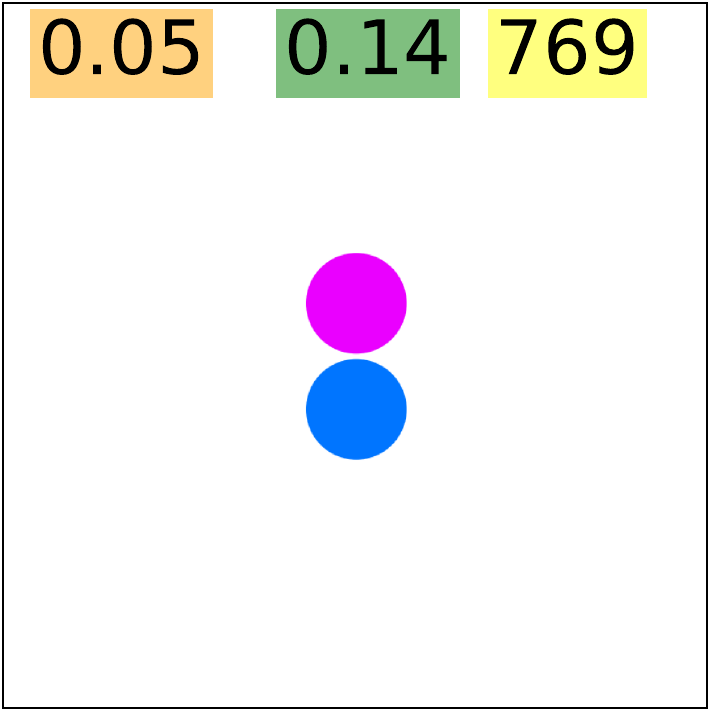}} &\multicolumn{2}{c}{\includegraphics[width=0.15\textwidth]{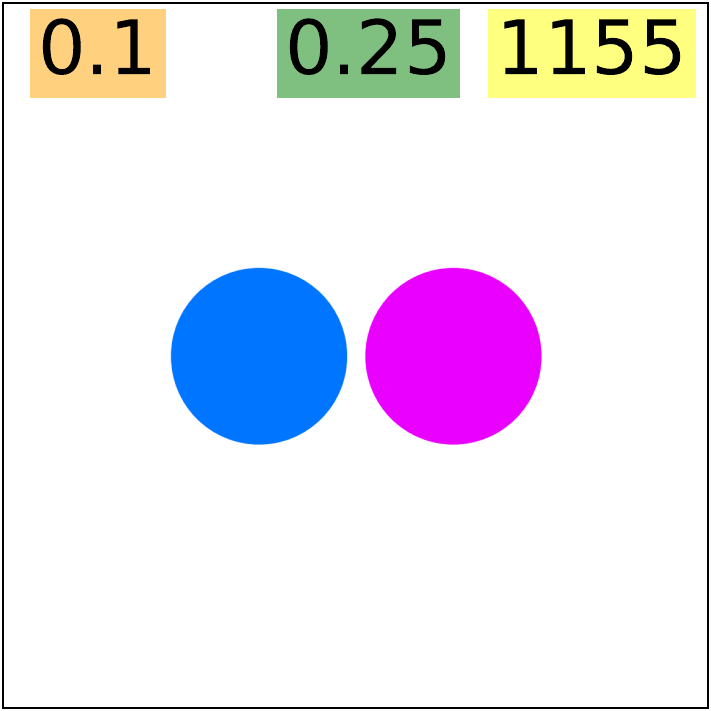}}&\multicolumn{2}{c}{\includegraphics[width=0.15\textwidth]{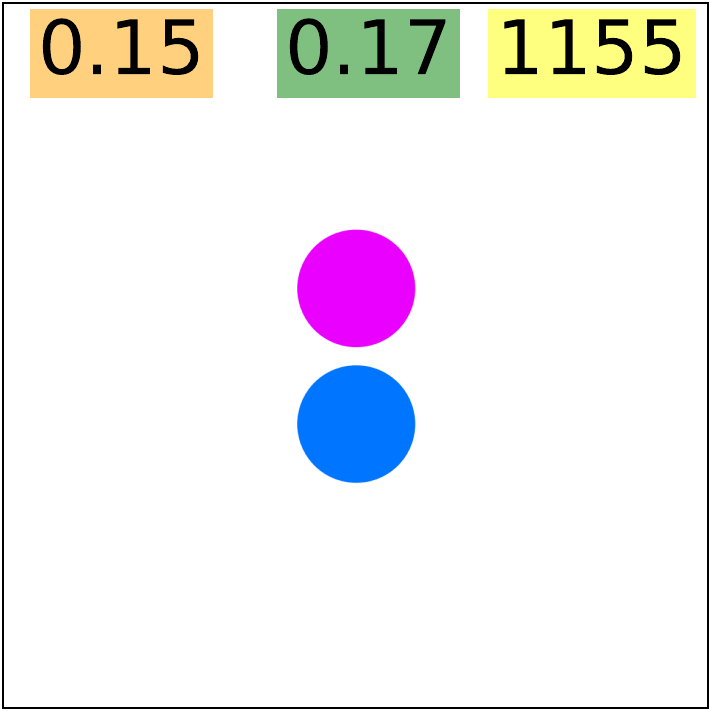}}\\
     \rowcolor{lightgray}
     \raisebox{-0.2\height}\gptlogo & \centering Yes & \textcolor{ForestGreen}{\cmark} &  \centering Yes & \textcolor{ForestGreen}{\cmark}&  \centering Yes & \textcolor{ForestGreen}{\cmark} &  \centering No & \textcolor{ForestGreen}{\cmark} &  \centering Yes & \textcolor{red}{\xmark}&  \centering Yes & \textcolor{red}{\xmark}\\
     \raisebox{-0.2\height}\geminilogo & \centering Yes & \textcolor{ForestGreen}{\cmark} & \centering Yes & \textcolor{ForestGreen}{\cmark}&  \centering Yes & \textcolor{ForestGreen}{\cmark} &  \centering Yes & \textcolor{red}{\xmark} &  \centering No & \textcolor{ForestGreen}{\cmark}&  \centering No & \textcolor{ForestGreen}{\cmark}\\
     \rowcolor{lightgray}
     \raisebox{-0.2\height}\sonnetlogo & \centering Yes & \textcolor{ForestGreen}{\cmark} & \centering Yes & \textcolor{ForestGreen}{\cmark}&  \centering Yes & \textcolor{ForestGreen}{\cmark} &  \centering Yes & \textcolor{red}{\xmark} &  \centering Yes & \textcolor{red}{\xmark}&  \centering Yes & \textcolor{red}{\xmark}\\
     \raisebox{-0.2\height}\newsonnetlogo & \centering Yes & \textcolor{ForestGreen}{\cmark} & \centering No & \textcolor{red}{\xmark}&  \centering Yes & \textcolor{ForestGreen}{\cmark} &  \centering No & \textcolor{ForestGreen}{\cmark} &  \centering No & \textcolor{ForestGreen}{\cmark}&  \centering No & \textcolor{ForestGreen}{\cmark}\\
     \end{tabular}

    \vspace{4pt}
    \centering
    \begin{tabular}{cccccccccccccc}
    \raisebox{-0.1\height}\gptlogo & \gpt &&&
      \raisebox{-0.1\height}\geminilogo & \gemini & &&\raisebox{-0.1\height}\sonnetlogo & \sonnet & && \raisebox{-0.12\height}\newsonnetlogo & \newsonnet \\
      \end{tabular}
     
\end{AIbox}
\caption{Some VLMs consistently fail by not seeing the distance between the two circles at large \orangedistance{distances} and \yellowsize{resolutions} (\gpt and \sonnet--rightmost). \gemini and \sonnet also struggle to see the small gaps between circles (\emph{\orangedistance{distance=0.05}})}
\label{fig:qual_touch}
\end{figure}

We show examples of models' responses to the prompts on the two touching circles task in \cref{fig:qual_touch}.

\begin{figure}[H]
\centering
\begin{AIbox}{Are the two circles overlapping? Answer with Yes/No.}
\centering
    \begin{tabular}{lp{0.8cm}c|p{0.8cm}c|p{0.8cm}c|p{0.8cm}c|p{0.8cm}c|p{0.8cm}c}
     &\multicolumn{2}{c}{\includegraphics[width=0.15\textwidth]{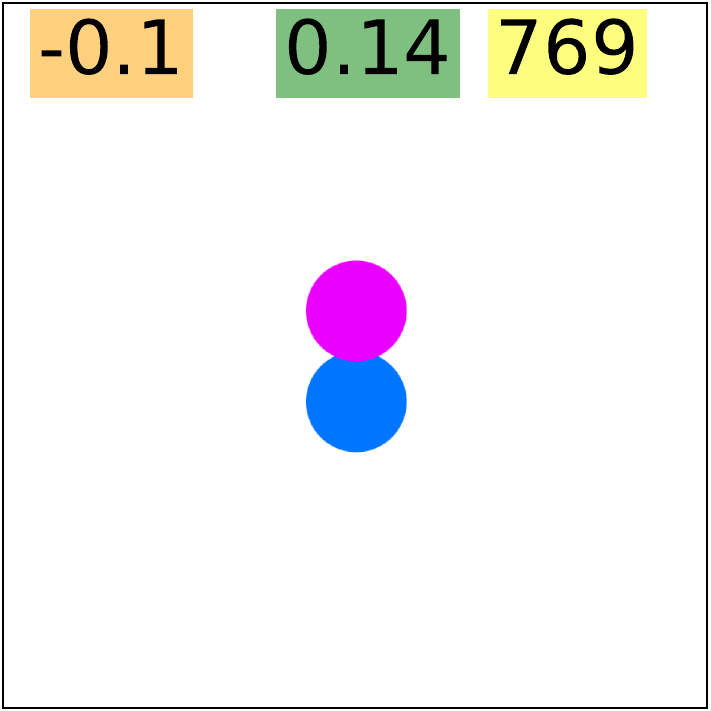}} & \multicolumn{2}{c}{\includegraphics[width=0.15\textwidth]{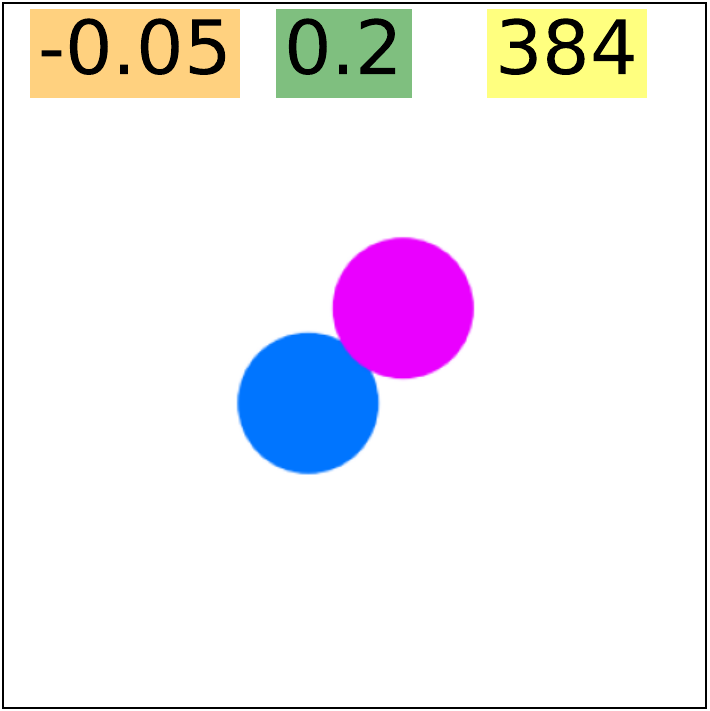}} &\multicolumn{2}{c}{\includegraphics[width=0.15\textwidth]{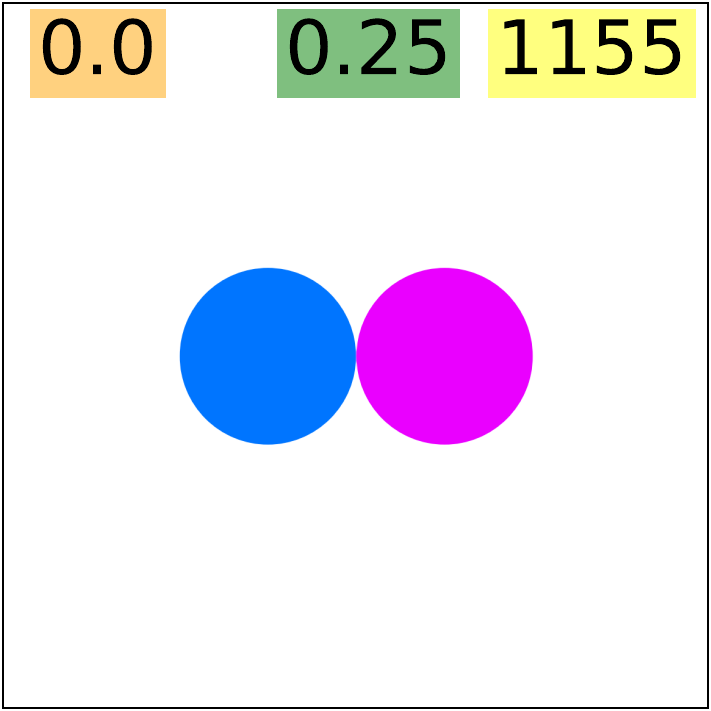}} &\multicolumn{2}{c}{\includegraphics[width=0.15\textwidth]{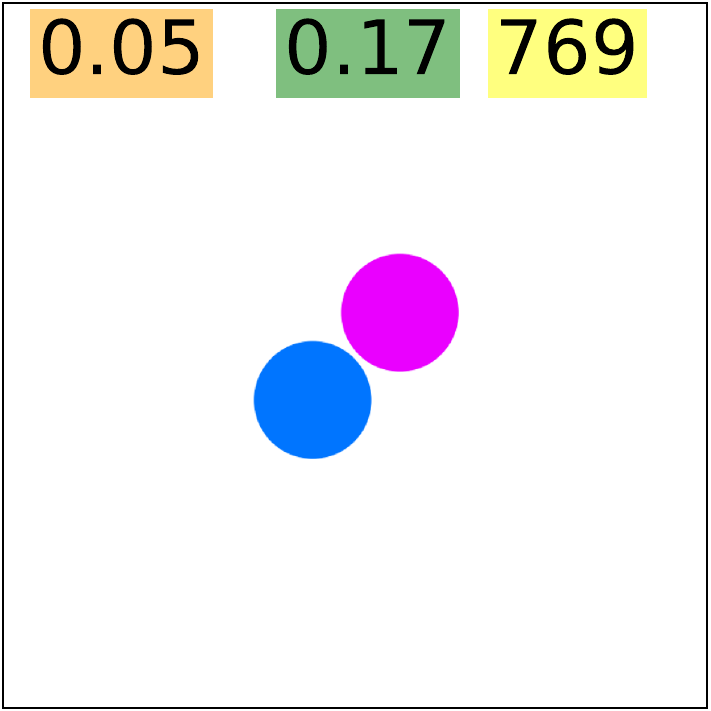}} &\multicolumn{2}{c}{\includegraphics[width=0.15\textwidth]{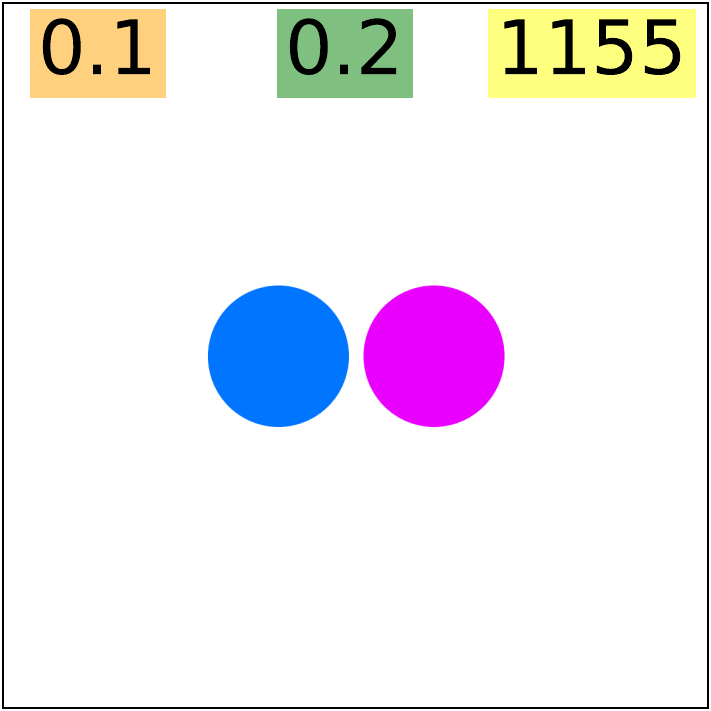}}&\multicolumn{2}{c}{\includegraphics[width=0.15\textwidth]{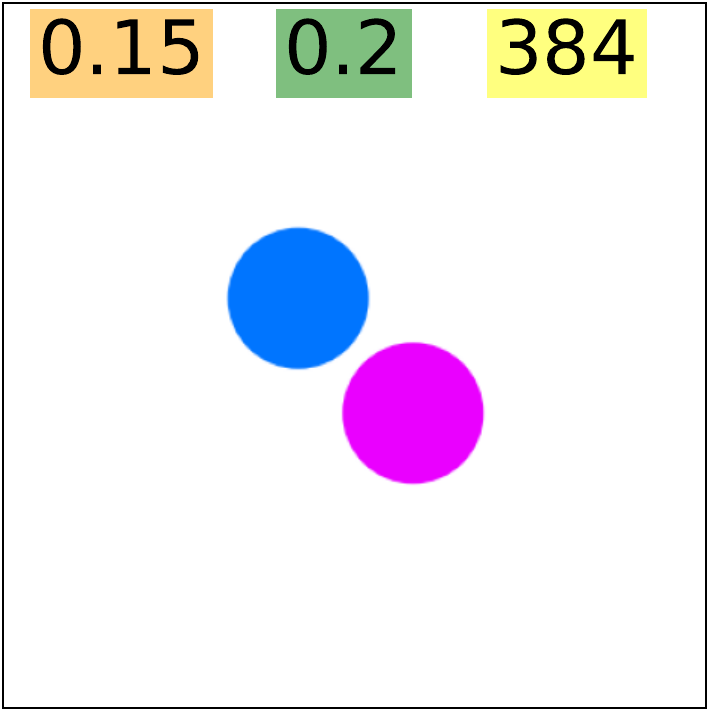}}\\
     \rowcolor{lightgray}
     \raisebox{-0.2\height}\gptlogo & \centering Yes & \textcolor{ForestGreen}{\cmark} &  \centering Yes & \textcolor{ForestGreen}{\cmark}&  \centering Yes & \textcolor{red}{\xmark} &  \centering Yes & \textcolor{red}{\xmark} &  \centering No & \textcolor{ForestGreen}{\cmark}&  \centering Yes & \textcolor{red}{\xmark}\\
     \raisebox{-0.2\height}\geminilogo & \centering No & \textcolor{red}{\xmark} & \centering Yes & \textcolor{ForestGreen}{\cmark}&  \centering Yes & \textcolor{red}{\xmark} &  \centering No & \textcolor{ForestGreen}{\cmark} &  \centering No & \textcolor{ForestGreen}{\cmark}&  \centering No & \textcolor{ForestGreen}{\cmark}\\
     \rowcolor{lightgray}
     \raisebox{-0.2\height}\sonnetlogo & \centering Yes & \textcolor{ForestGreen}{\cmark} & \centering Yes & \textcolor{ForestGreen}{\cmark}&  \centering Yes & \textcolor{red}{\xmark} &  \centering Yes & \textcolor{red}{\xmark} &  \centering Yes & \textcolor{red}{\xmark}&  \centering No & \textcolor{ForestGreen}{\cmark}\\
     \raisebox{-0.2\height}\newsonnetlogo & \centering No & \textcolor{red}{\xmark} & \centering No & \textcolor{red}{\xmark}&  \centering No & \textcolor{ForestGreen}{\cmark} &  \centering No & \textcolor{ForestGreen}{\cmark} &  \centering No & \textcolor{ForestGreen}{\cmark}&  \centering No & \textcolor{ForestGreen}{\cmark}\\
     \end{tabular}

    \vspace{4pt}
    \centering
    \begin{tabular}{cccccccccccccc}
    \raisebox{-0.1\height}\gptlogo & \gpt &&&
      \raisebox{-0.1\height}\geminilogo & \gemini & &&\raisebox{-0.1\height}\sonnetlogo & \sonnet & && \raisebox{-0.12\height}\newsonnetlogo & \newsonnet \\
      \end{tabular}
     
\end{AIbox}
\caption{VLMs consistently fail at smaller \distances.
However, when the \sethlcolor{forestgreen(web)} \hl{gap} is large and clearly visible, \gpt remains unreliable. 
\newsonnet tends to conservatively answer ``{No}'' regardless of the actual distance between the two circles.
}
\label{fig:qual_overlap}
\end{figure}

\clearpage
\section{Task 3: Identifying the circled letter in a string}
\label{app:circled_word}

\subsection{Task construction}
\label{app:redoval-bench}

Each image is created using a combination of the below variables.

\begin{itemize}
    \item \textbf{String}: We test three strings \{\Acknowledgement, \Subdermatoglyphic,\\ \tHyUiKaRbNqWeOpXcZvM\}. 
    While they all have letters of varying widths and heights, the first two are English words while the latter is a string of random characters.
    \Subdermatoglyphic is the longest English word that has no letter repetitions.
    \item \textbf{Letter}: We draw the oval (\redcircle{ }) over every letter, one at a time, in each string.
    \item \textbf{Font}: We use 2 different font families for each word, \opensansfont{OpenSans} and \helvetica{Helvetica}.
    \item \textbf{Oval (\redcircle{ }) thickness}: We generate the \redcircle{ } with 3 various line thicknesses.
    \item \textbf{Scaling factor}: Since each letter has a unique size, we use a scaling factor to control the size of the \redcircle{ }.
\end{itemize}

\begin{figure}
    \centering
    \begin{tabular}{ccc}
         \includegraphics[width=0.2\textwidth]{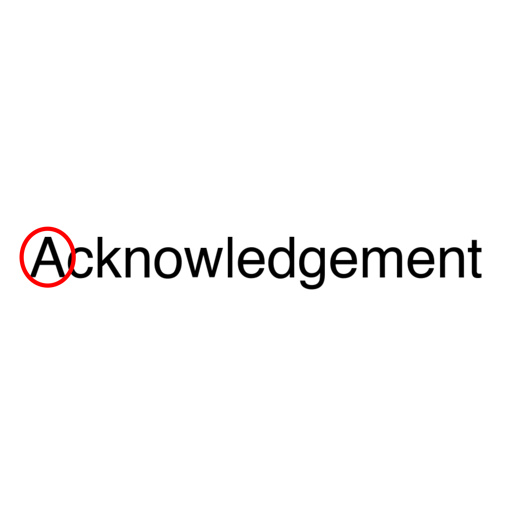}&\includegraphics[width=0.2\textwidth]{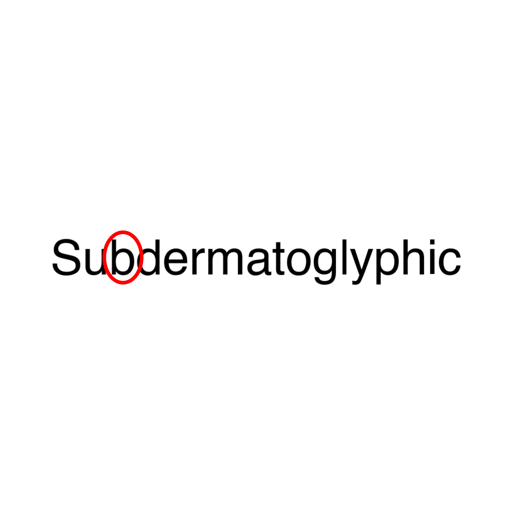}& \includegraphics[width=0.2\textwidth]{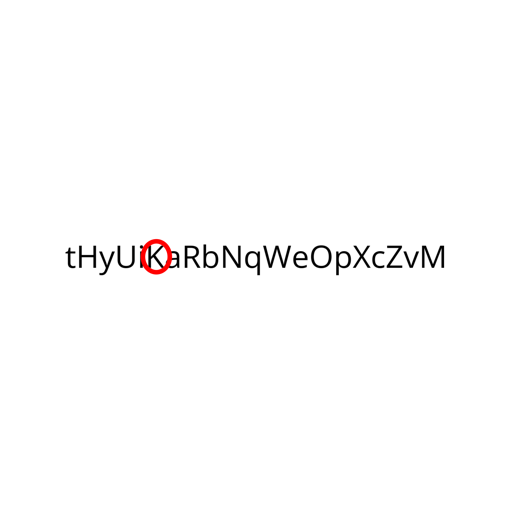} \\
         \includegraphics[width=0.2\textwidth]{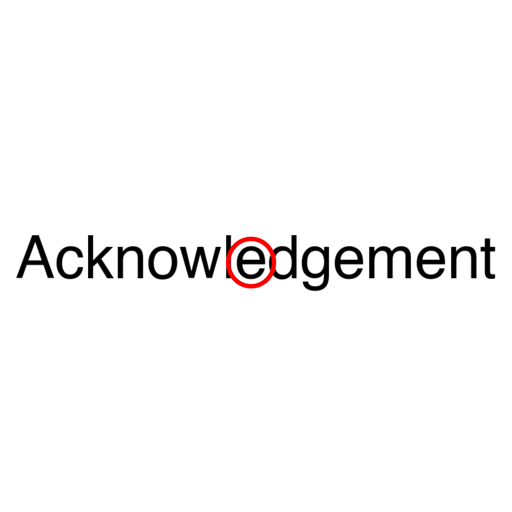}&\includegraphics[width=0.2\textwidth]{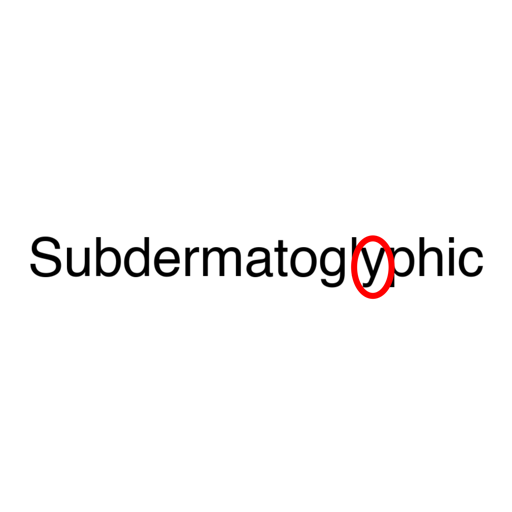}&\includegraphics[width=0.2\textwidth]{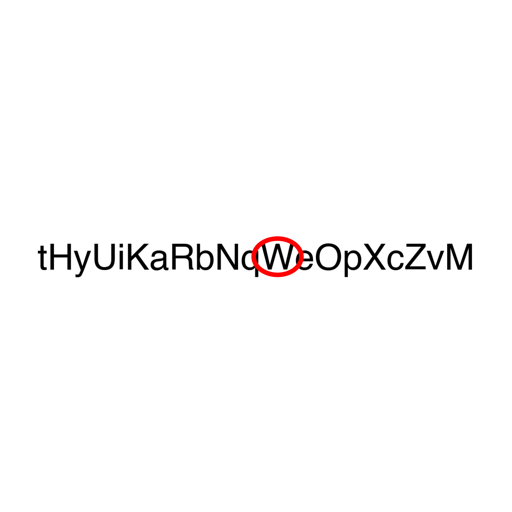}  \\
         \includegraphics[width=0.2\textwidth]{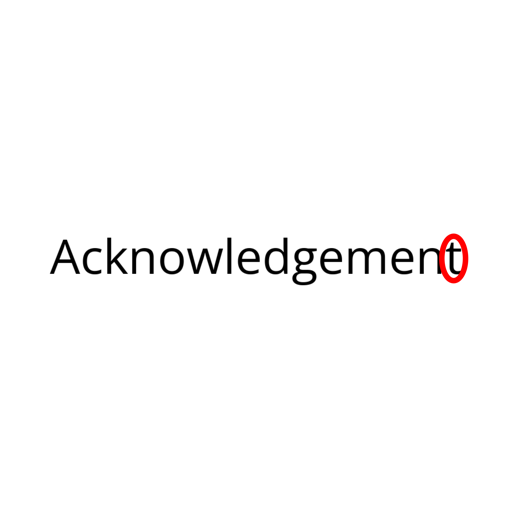}&\includegraphics[width=0.2\textwidth]{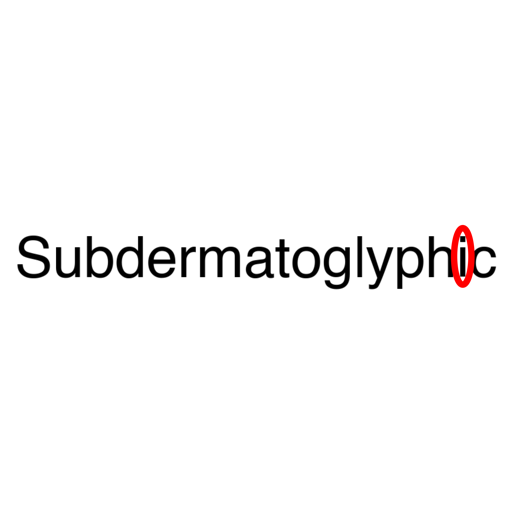}& \includegraphics[width=0.2\textwidth]{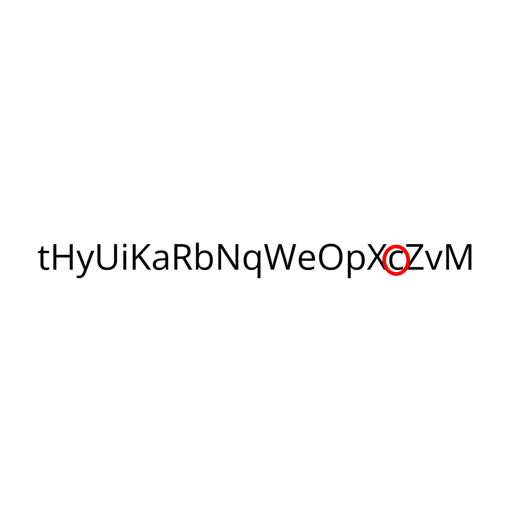} \\
         (a) \Acknowledgement & (b) \Subdermatoglyphic & (c) \tHyUiKaRbNqWeOpXcZvM \\
    \end{tabular}
    \caption{Our benchmark comprises three different words, of which one letter is circled by the red oval in each image.}
    \label{app:fig-redoval-examples}
\end{figure}

Finally, we render the text on a white canvas with a size of 1250$\times$1250 pixels, and we produce 90 images for \Acknowledgement, 102 for \Subdermatoglyphic, and 120 samples for  \tHyUiKaRbNqWeOpXcZvM (see \cref{table:word-oval-parameters,app:fig-redoval-examples}). 

\subsubsection{Evaluation}
To determine the models' prediction, we extract the character enclosed in \{curly braces\} in the models' response.

\subsubsection{Code} The code is available at \url{https://github.com/anguyen8/vision-llms-are-blind/blob/main/src/CircledWord/GenerateSamples.ipynb}.

\subsec{Prompts}
\begin{enumerate}
    \item \emph{Which letter is being circled? Please provide your answer in curly brackets, e.g. \{a\}}
    
    \item \emph{Which character is being highlighted with a red oval? Please provide your answer in curly brackets, e.g. \{a\}} 
\end{enumerate}

\subsec{Groundtruth} letters must match predicted letters exactly (case-insensitive).

\begin{table}[ht]
\centering
\caption{The breakdown of our benchmark based on the number of different parameters shows our data is balanced for each word.}
\label{table:word-oval-parameters}
\begin{tabular}{lccc}
\toprule
\textbf{Parameter} & \textbf{Values} & \textbf{Samples} & ~~\textbf{Total Samples}\\ 
\midrule
\multirow{3}{*}{Word} & \Acknowledgement & 90 & ~~\multirow{3}{*}{312}\\ 
                      & \Subdermatoglyphic & 102 \\ 
                      & \tHyUiKaRbNqWeOpXcZvM & 120 \\ 
\midrule
\multirow{2}{*}{Font} & \opensansfont{OpenSans} & 156 & ~~\multirow{2}{*}{312}\\ 
                      & \helvetica{Helvetica} & 156 \\ 
\midrule
Image size ($C$) & 1250x1250 pixels & 312 & ~~312 \\ 
\midrule
\multirow{5}{*}{Oval thickness} & $\frac{1}{200}\times C$ & 104 & ~~\multirow{5}{*}{312}\\ 
\\
                                & $\frac{1}{250}\times C$ & 104 \\ 
                                \\
                                & $\frac{1}{300}\times C$ & 104 \\ 
\midrule
\end{tabular}
\end{table}

\begin{figure}
        \centering
        \begin{tabular}{cc}
         \includegraphics[trim={20 150 20 200},clip,width=0.4\textwidth]{images/CircledLetter/benchmark_samples/gt_a_pred_a_text_image_3bac479b-38ae-467f-82b3-36ec1c1dd97b.png}& \includegraphics[trim={0 150 0 200},clip,width=0.5\textwidth]{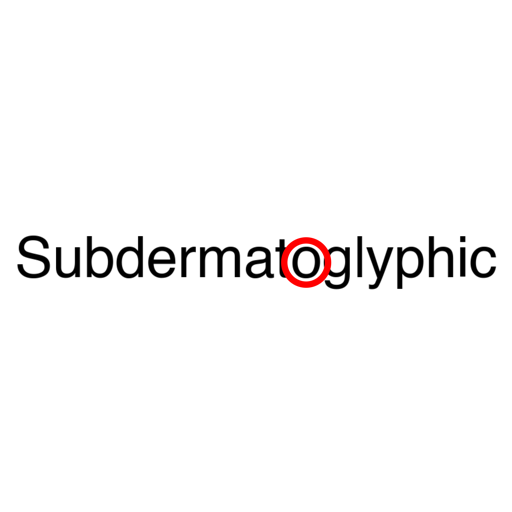}\\
         \includegraphics[trim={10 150 10 200},clip,width=0.4\textwidth]{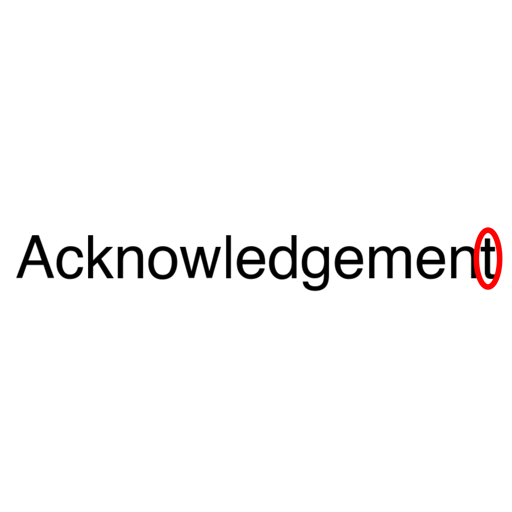}& \includegraphics[trim={20 150 20 200},clip,width=0.6\textwidth]{images/CircledLetter/benchmark_samples/gt_w_pred_w_text_image_b52e3649-ef9c-4dcb-a751-28d7444e5126.png} \\
    \end{tabular}
    \caption{Which letter is being red circled?}
    \label{fig:red_oval}
\end{figure}

\clearpage
\subsection{Finding: VLMs mostly confuse the adjacent character for the circled letter}
\label{app:redoval_adjacent}

\begin{figure}
    \centering
    \includegraphics[width=\textwidth]{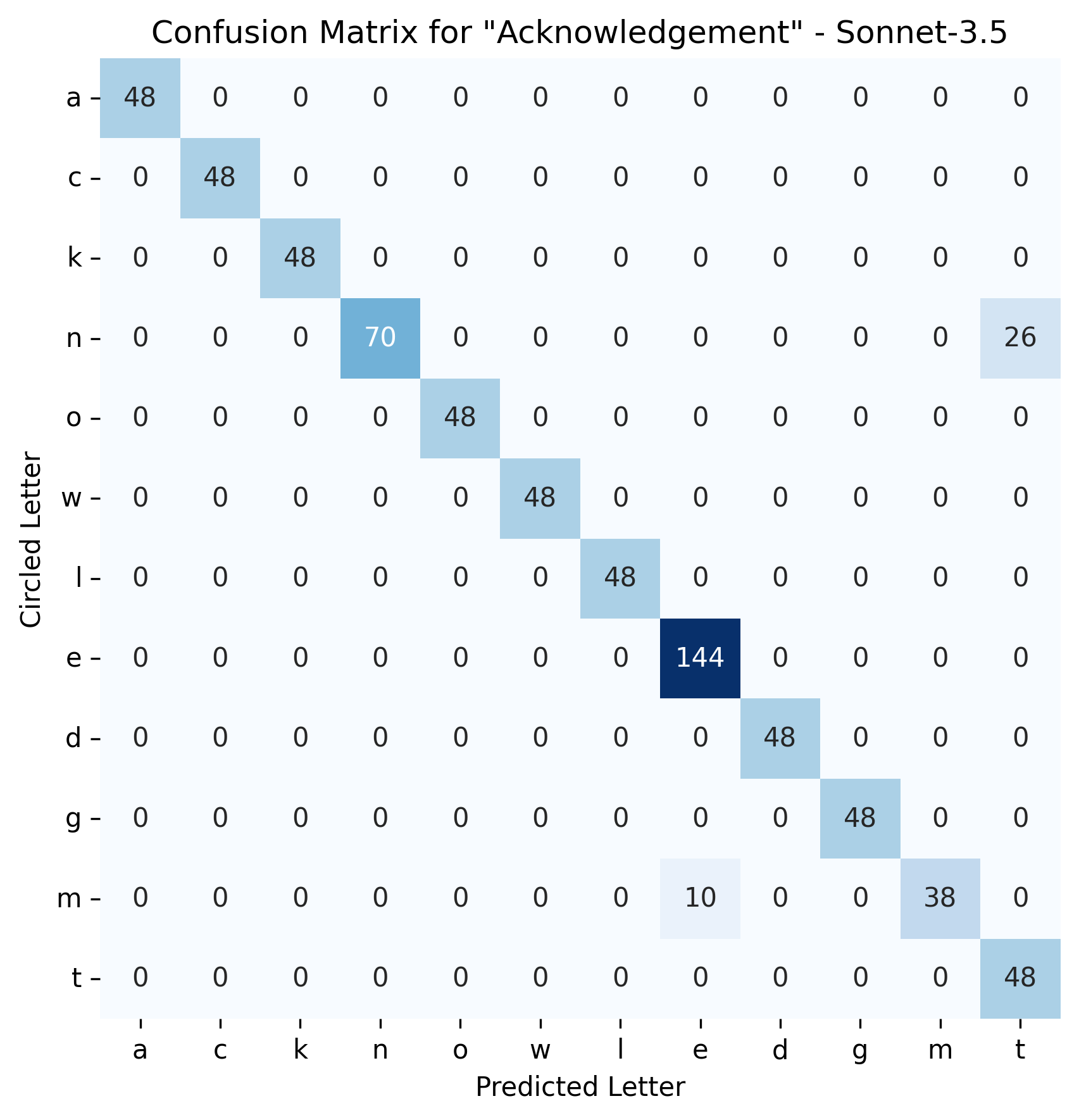}
    \caption{Adjacent letters are the most common wrong prediction for \newsonnet in \Acknowledgement, \eg, letter ``e'' is predicted instead of ``m'' 20.83\% of the time, or letter ``t'' is predicted instead of ``n'' 27.08\% of the time.}    
    \label{confusion:ack_newsonnet}
\end{figure}

\begin{figure}
    \centering
    \includegraphics[width=\textwidth,clip]{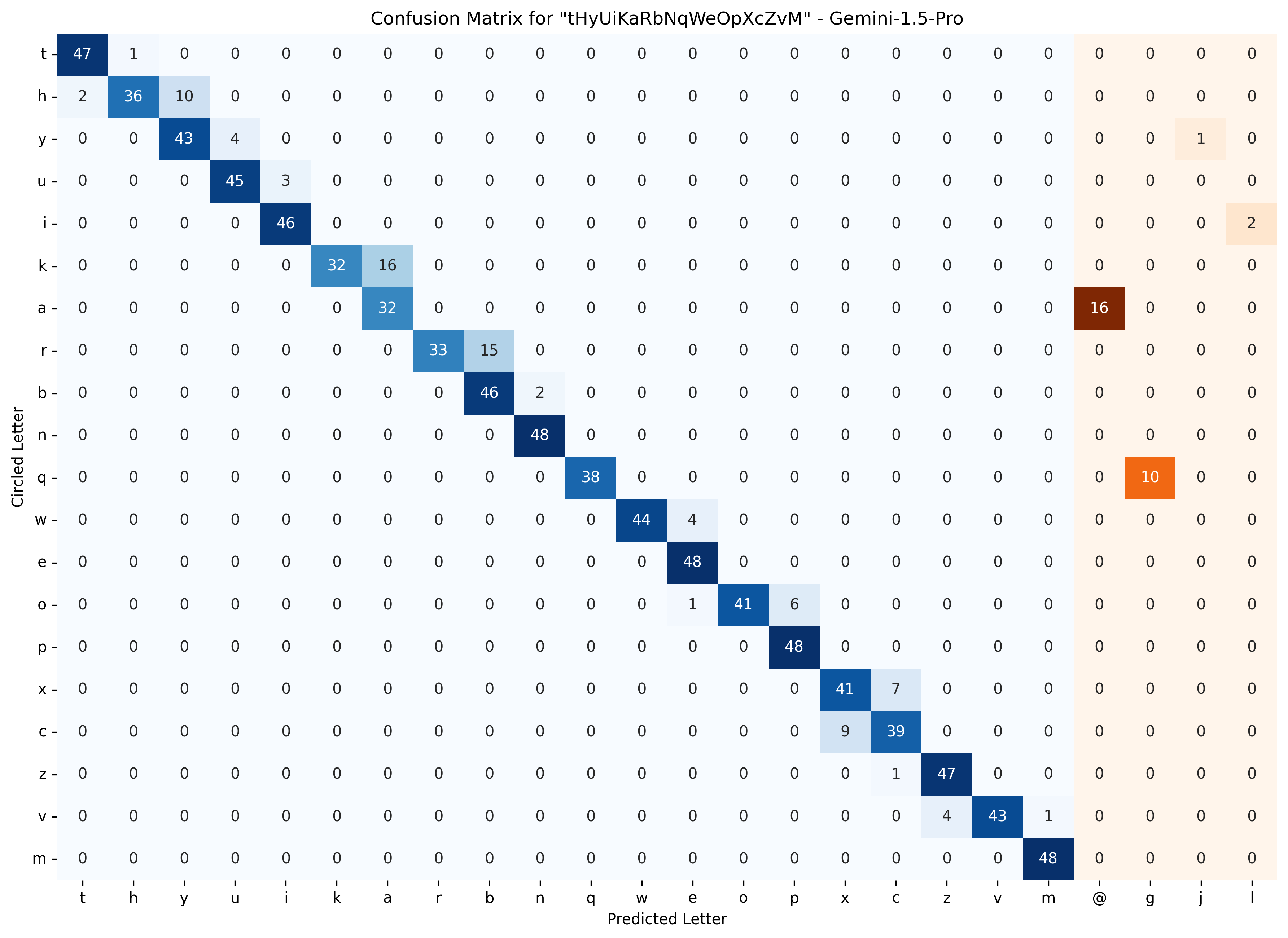}
    \caption{\gemini tends to confuse adjacent letters like ``a'' and ``K''. \gemini also hallucinates the circled letter ``a'' as being the character ``@'' and it confuses the letter ``q'' for the letter ``g''.}\label{confusion:rand_gemini}
\end{figure}

Models often mistake the neighboring characters as actual circled letters. For example, \cref{confusion:ack_newsonnet} shows that for \newsonnet in the word \Acknowledgement, all of the incorrect predictions for ``n'' are the neighboring letter ``t''. For \gemini, in the word \tHyUiKaRbNqWeOpXcZvM, where letters ``K'' and ``a'' are adjacent, 100\% of the mispredictions for ``K'' are the letter ``a'' (see \cref{confusion:rand_gemini}).

\begin{figure}
\centering
\includegraphics[width=\textwidth]{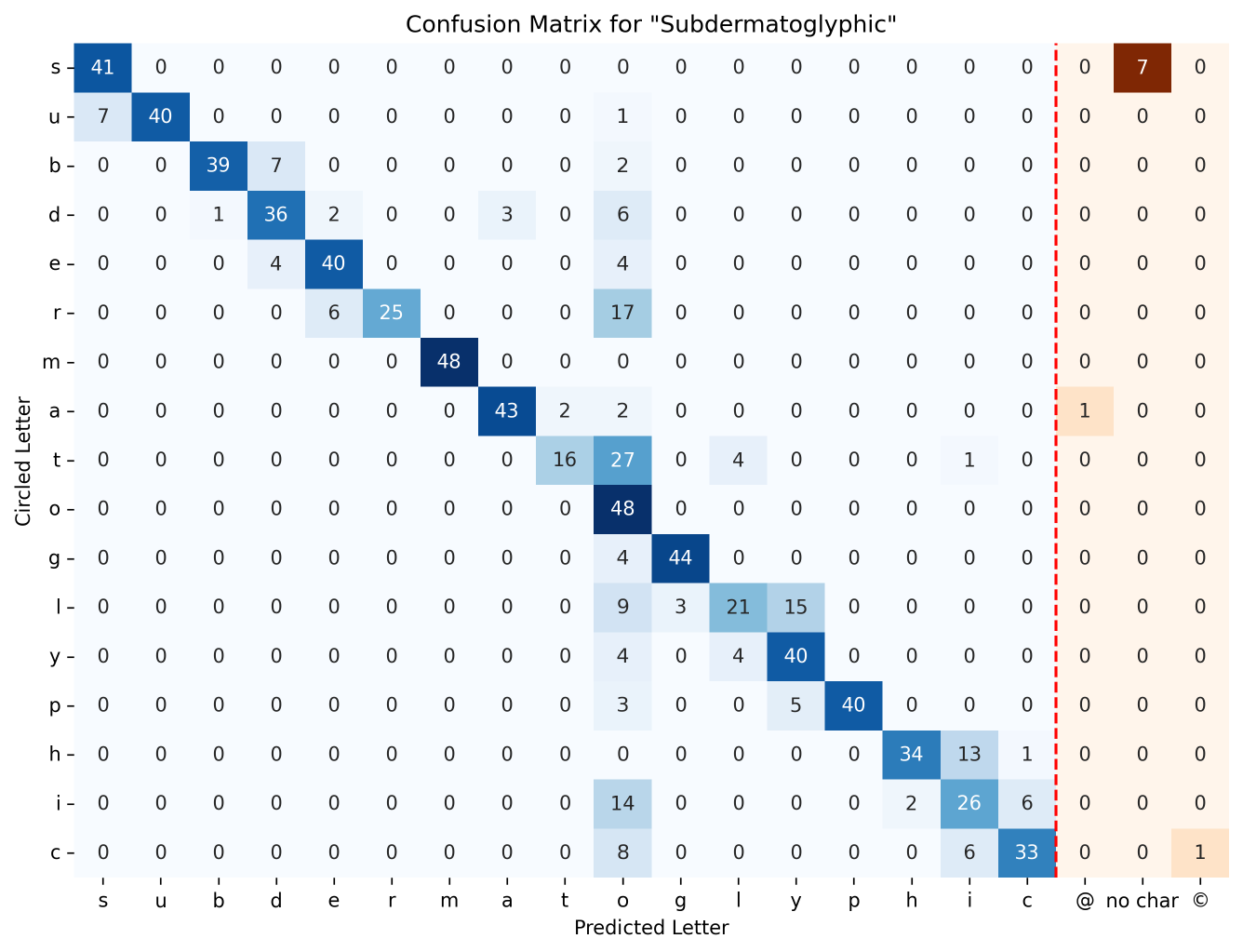}
\caption{Aggregate confusion matrix summed over all 4 VLMs and 12 responses for each letter in the word \SubdermatoglyphicCaption. Models mostly mispredict characters near the circled letter.
VLMs sometimes hallucinate characters that do not even exist in the word, \eg, ``{@}'' (\hallucinatepanel{right panel}).
}
\label{fig:confusion-main}
\end{figure}


\clearpage
\subsection{Finding: \gpt and \gemini confuse the red oval as part of the letter}
\label{app:redoval-confuses_oval}

\begin{figure}
    \includegraphics[width=\textwidth]{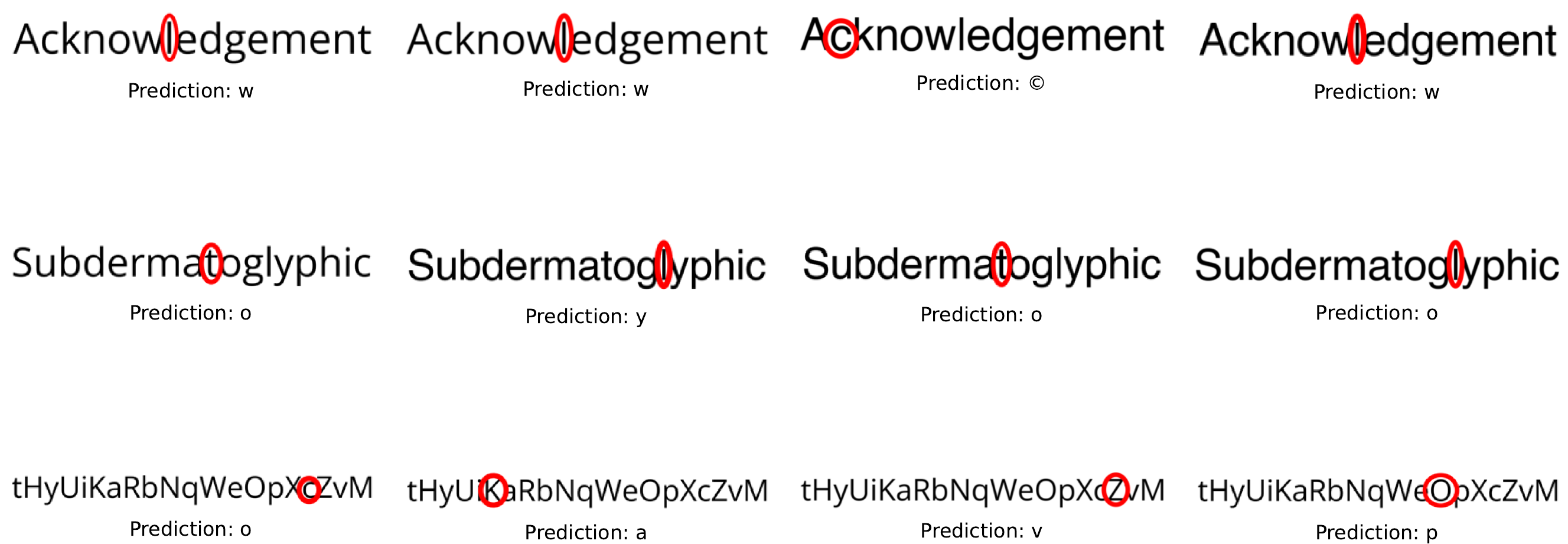}
    \caption{Random samples for different words, and \gemini's predictions, where \gemini mostly predicts the adjacent letters or confuses the red oval as part of the circled letter.}    
    \label{appfig:gemini_failure_cases}
\end{figure}

\begin{figure}
    \centering
    \includegraphics[width=\textwidth]{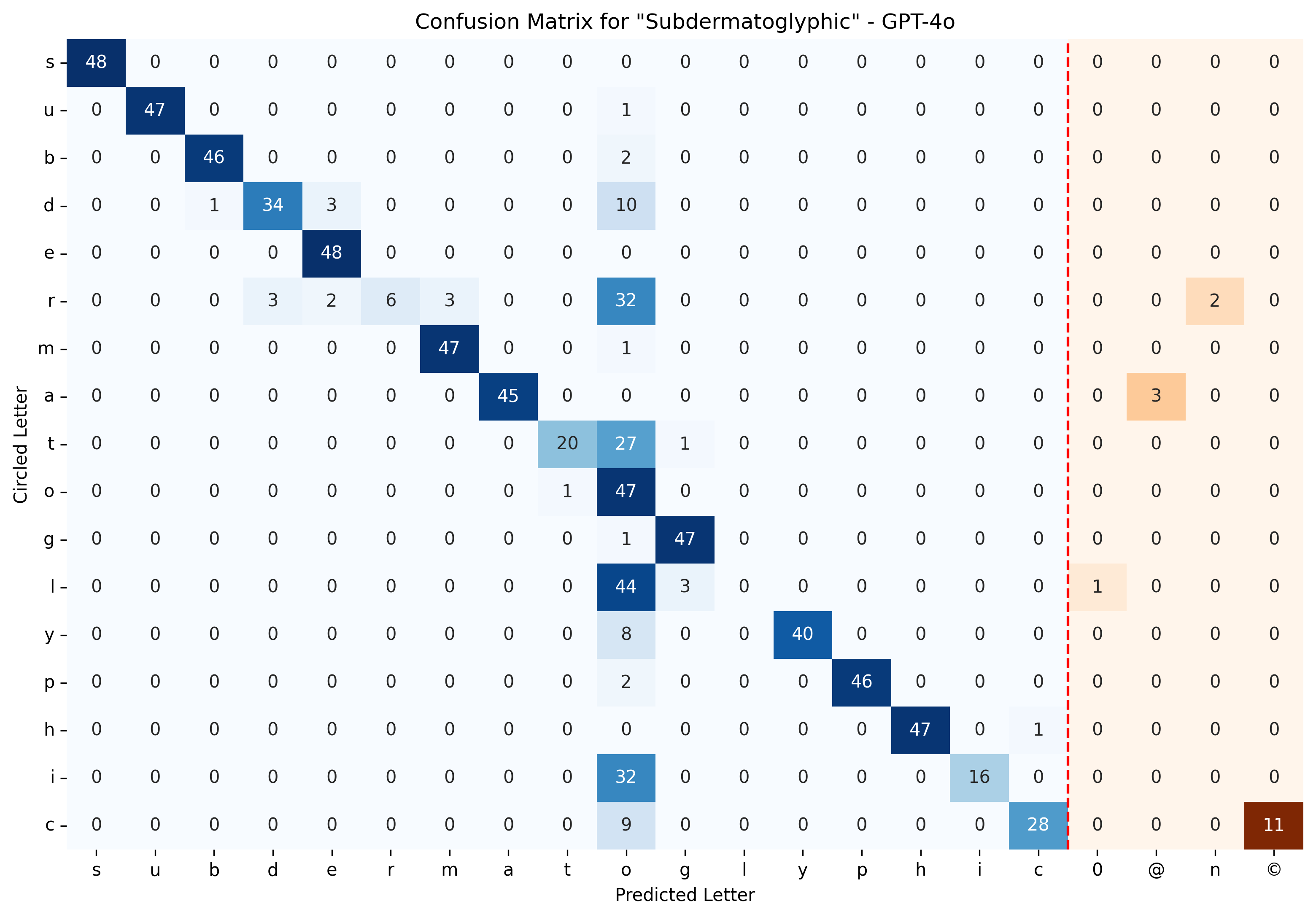}
    \caption{\gpt is the most sensitive to confuse the red oval as part of the letter, where it often predicts ``o'' instead of ``l'', ``y'', ``p'', and ``c'' in the word \Subdermatoglyphic.}    
    \label{confusion:sub_gpt}
\end{figure}

\cref{appfig:gemini_failure_cases,confusion:sub_gpt} show that \gemini and \gpt sometimes fail to recognize that the red oval is not part of the letter. \gemini tends to predict that the circled letter ``a'' is actually the ``@'' sign (see \cref{appfig:gemini_failure_cases}). \gpt on the other hand tends to predict ``o'' regardless of which letter is circled (\cref{confusion:sub_gpt}).

\clearpage

\subsection{Finding: Models perform similarly across two common font families}
\label{app:redoval_font}

\begin{figure}[ht]
    \centering
    \includegraphics[width=\textwidth]{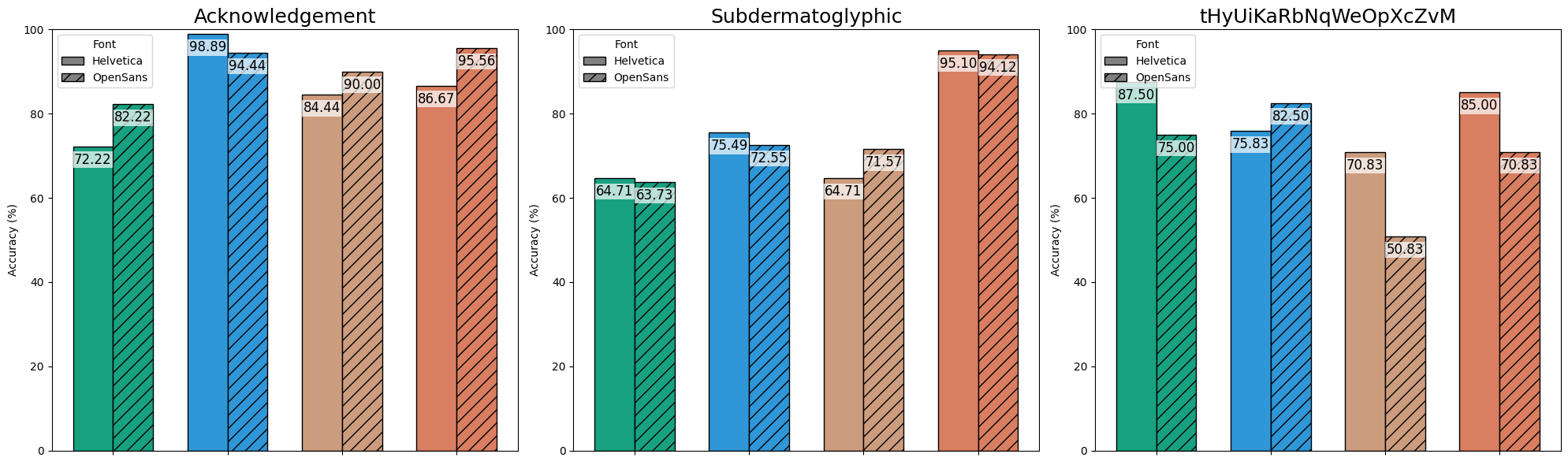}
    \vspace{0.25cm} 
    \hspace{0.04\textwidth}
    \raisebox{-0.5\height}{\gptlogo}\hspace{0.04\textwidth}
    \raisebox{-0.5\height}{\geminilogo}\hspace{0.04\textwidth}
    \raisebox{-0.5\height}{\sonnetlogo}\hspace{0.04\textwidth}
    \raisebox{-0.5\height}{\newsonnetlogo}\hspace{0.1\textwidth}
    \raisebox{-0.5\height}{\gptlogo}\hspace{0.04\textwidth}
    \raisebox{-0.5\height}{\geminilogo}\hspace{0.04\textwidth}
    \raisebox{-0.5\height}{\sonnetlogo}\hspace{0.04\textwidth}
    \raisebox{-0.5\height}{\newsonnetlogo}\hspace{0.1\textwidth}
    \raisebox{-0.5\height}{\gptlogo}\hspace{0.04\textwidth}
    \raisebox{-0.5\height}{\geminilogo}\hspace{0.04\textwidth}
    \raisebox{-0.5\height}{\sonnetlogo}\hspace{0.04\textwidth}
    \raisebox{-0.5\height}{\newsonnetlogo} \hfill
    \caption{VLMs do not show a consistent trend between fonts. This suggests that using different spacing between letters, letter styles, and letter size has minimal effects on the VLMs' ability to see the content of the red oval.}
    \label{suppfig:circledword_font_combined_average}
\end{figure}

As shown in \cref{suppfig:circledword_font_combined_average}, models do not show a significant variance over different fonts, suggesting our choice of font is not a reason for their unreliable performance.

\clearpage
\subsection{Finding: models are invariant to our choice of prompts}
\label{app:redoval_prompt}

\begin{figure}[ht]
    \centering
    \includegraphics[width=\textwidth]{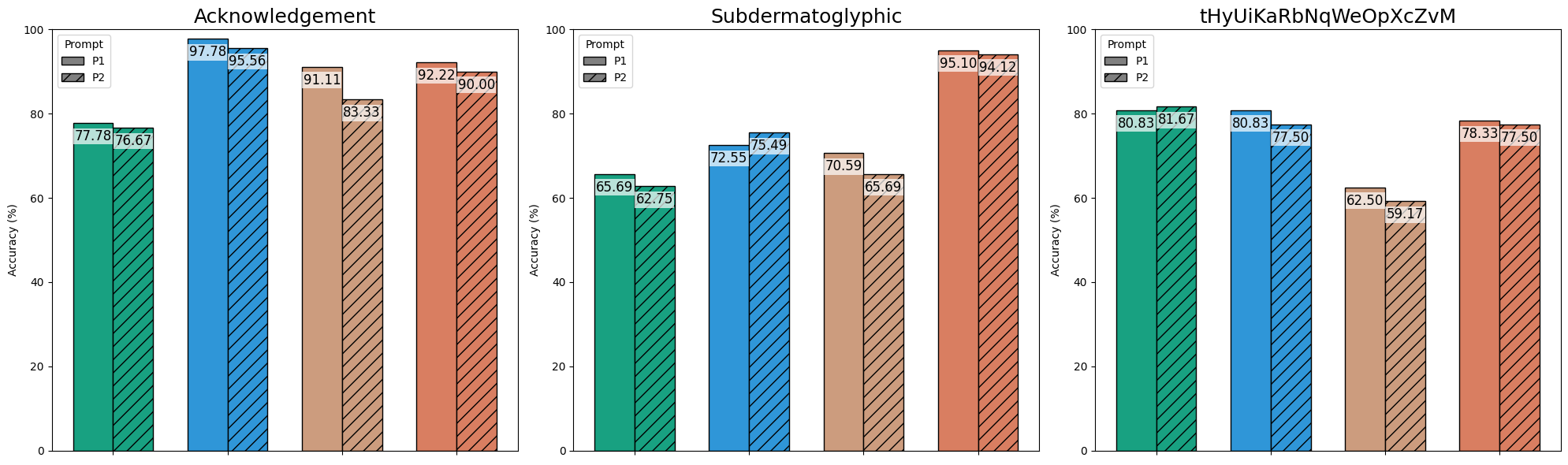}
    \vspace{0.25cm} 
    \hspace{0.007\textwidth}
    \raisebox{-0.5\height}{\gptlogo}\hspace{0.045\textwidth}
    \raisebox{-0.5\height}{\geminilogo}\hspace{0.045\textwidth}
    \raisebox{-0.5\height}{\sonnetlogo}\hspace{0.045\textwidth}
    \raisebox{-0.5\height}{\newsonnetlogo}\hspace{0.1\textwidth}
    \raisebox{-0.5\height}{\gptlogo}\hspace{0.045\textwidth}
    \raisebox{-0.5\height}{\geminilogo}\hspace{0.045\textwidth}
    \raisebox{-0.5\height}{\sonnetlogo}\hspace{0.045\textwidth}
    \raisebox{-0.5\height}{\newsonnetlogo}\hspace{0.085\textwidth}
    \raisebox{-0.5\height}{\gptlogo}\hspace{0.04\textwidth}
    \raisebox{-0.5\height}{\geminilogo}\hspace{0.04\textwidth}
    \raisebox{-0.5\height}{\sonnetlogo}\hspace{0.04\textwidth}
    \raisebox{-0.5\height}{\newsonnetlogo} \hfill
    \caption{Model performance breakdown for different prompts P1: \textit{"Which letter is being circled? Please provide your answer in curly brackets, e.g. \{a\}"} and P2: \textit{"Which character is being highlighted with a red oval? Please provide your answer in curly brackets, e.g. \{a\}"}. Regardless of the prompts, VLMs perform similarly in seeing the contents of the red oval.}
    \label{supptab:circledword_performance_different_prompts}
\end{figure}

Our choice of prompts has little impact on the performance of the models as depicted in \cref{supptab:circledword_performance_different_prompts}.

\clearpage
\subsection{Finding: VLMs fail to see the circled letter on various English words}
\label{app:english_words}
On average, SOTA VLMs show the same limitations across various English words (see \cref{tab:english_combined_average}). However, the mean accuracy over words with less and non-repetitive letters is higher than others (the mean accuracy over the word \questionWord is 90.36\% while it is 79.86\% on the word \civilization).

\begin{table}
\centering
\caption{Overall, VLMs perform better on short words without repetitive letters, \eg, \questionWord. Yet, the mean accuracy of four SOTA VLMs (86.43\%) is far from being 100\% accurate.
}
\label{tab:english_combined_average}
\resizebox{0.55\textwidth}{!}{%
\begin{tabular}{lrrrrr}
\toprule
\multicolumn{1}{l}{String} & \multicolumn{1}{c}{\gptlogo} & \multicolumn{1}{c}{~~\geminilogo} & \multicolumn{1}{c}{~~\sonnetlogo} & \multicolumn{1}{c}{~~\newsonnetlogo} & ~~Mean \\
\midrule
\questionWord     & ~~98.96 & ~~100.00 & ~~62.50 & ~~100.00 & ~~90.36 \\
\syndrome    & ~~91.67 & ~~91.67 & ~~70.83 & ~~91.58 & ~~86.44\\
\environmental & ~~82.69 & ~~89.10 & ~~73.72 & ~~89.74 & ~~83.81 \\
\civilization & ~~79.17 & ~~87.50 & ~~61.80 & ~~90.97 & ~~79.86 \\
\tournament & ~~89.17 & ~~92.50 & ~~85.00 & ~~100.00 & ~~91.67 \\
\midrule
Mean accuracy & ~~88.33 & ~~92.15 & ~~70.77 &  ~~94.46 & ~~86.43 \\
\bottomrule
\end{tabular}}
\end{table}

\begin{table}
\centering
\caption{Except for \gpt, all other models have a higher accuracy (\%) on the two English words than on the random string, suggesting that VLMs might leverage their familiarity with a known word to make educated guesses.
}
\label{tab:circledwords_combined_average}
\resizebox{0.95\textwidth}{!}{%
\begin{tabular}{lrrrrr}
\toprule
\multicolumn{1}{l}{String} & \multicolumn{1}{c}{\gptlogo} & \multicolumn{1}{c}{~~\geminilogo} & \multicolumn{1}{c}{~~\sonnetlogo} & \multicolumn{1}{c}{~~\newsonnetlogo} & ~~Mean \\
\midrule
\Acknowledgement     & ~~77.22 & ~~96.67 & ~~87.22 & ~~91.11 & ~~88.06 \\
\Subdermatoglyphic    & ~~64.42 & ~~74.02 & ~~68.14 & ~~94.10 & ~~75.25\\
\footnotesize{\tHyUiKaRbNqWeOpXcZvM} & 81.25 & ~~79.17 & ~~60.83 & ~~77.92 & ~~74.79 \\
\midrule
Mean accuracy & ~~74.23 & ~~83.29 & ~~72.06 &  ~~87.88 & ~~79.37 \\
\bottomrule
\end{tabular}}
\end{table}


\clearpage
\subsection{Adding white space between letters consistently improves the accuracy across all words}
\label{app:space-words}

\begin{figure}
    \centering
    \begin{tabular}{ccccc}
    \raisebox{-0.8\height}{\gptlogo} & \multirow{4}{*}{\includegraphics[width=0.2\textwidth]{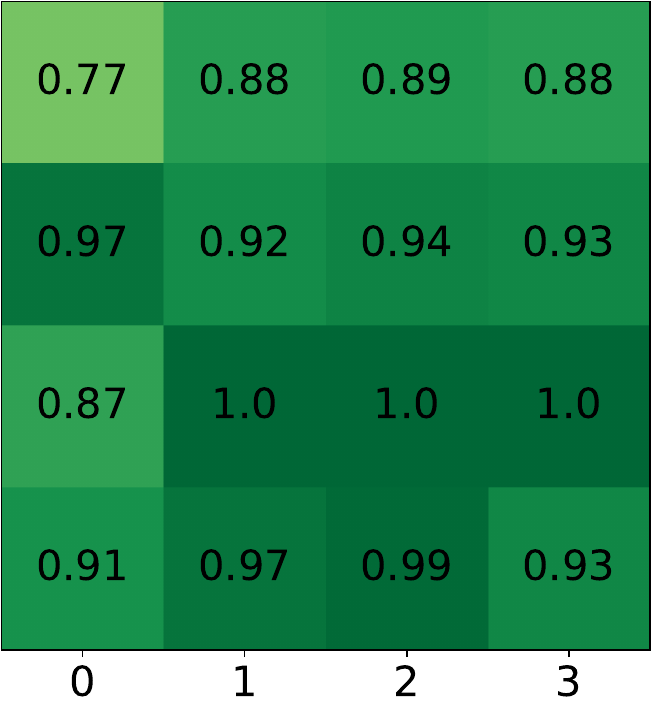}} & \multirow{4}{*}{\includegraphics[width=0.2\textwidth]{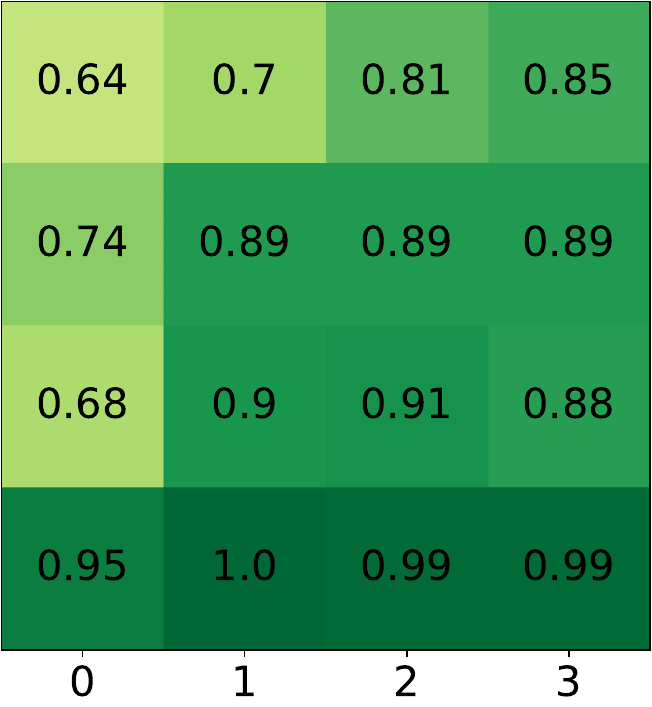}} & \multirow{4}{*}{\includegraphics[width=0.2\textwidth]{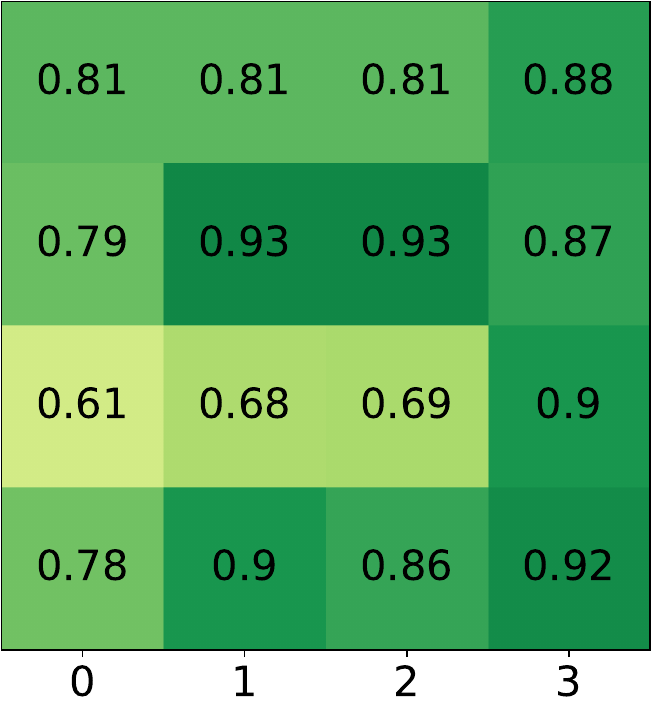}} & \multirow{4}{*}{\includegraphics[width=0.2\textwidth]{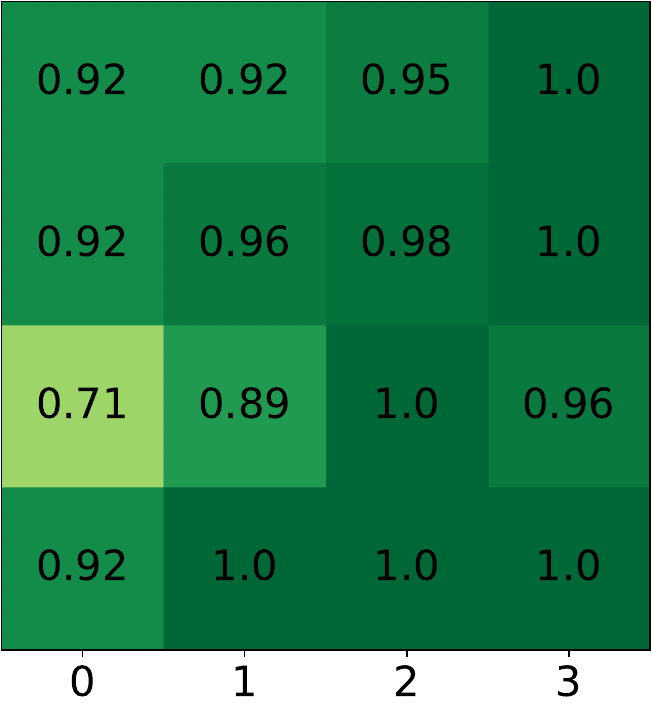}} \\
    \raisebox{-1.2\height}{\geminilogo} &&&&\\
    \raisebox{-1.0\height}{\sonnetlogo} &&&&\\
    \raisebox{-1.2\height}{\newsonnetlogo} &&&&\\ 
    \\
    &\scalebox{0.6}{(a) \Acknowledgement} & \scalebox{0.6}{(b) \Subdermatoglyphic} & \scalebox{0.6}{(c) \tHyUiKaRbNqWeOpXcZvM} & \scalebox{0.6}{(d) \syndrome}\\
    \end{tabular}
    \begin{tabular}{ccccc}
    \raisebox{-0.8\height}{\gptlogo} & \multirow{4}{*}{\includegraphics[width=0.2\textwidth]{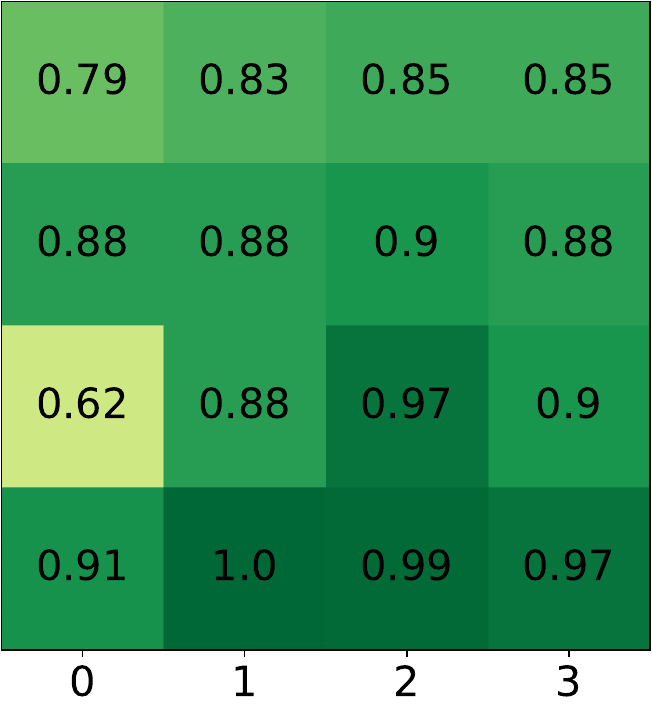}} & \multirow{4}{*}{\includegraphics[width=0.2\textwidth]{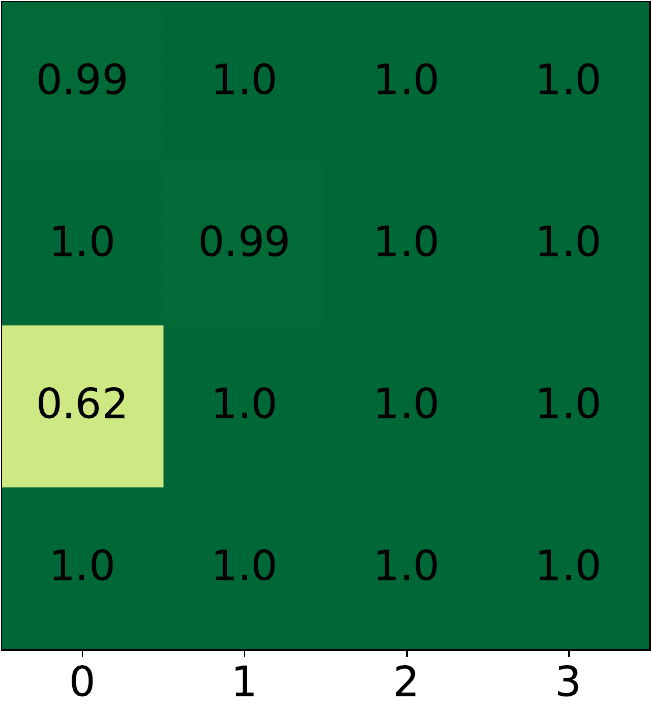}} & \multirow{4}{*}{\includegraphics[width=0.2\textwidth]{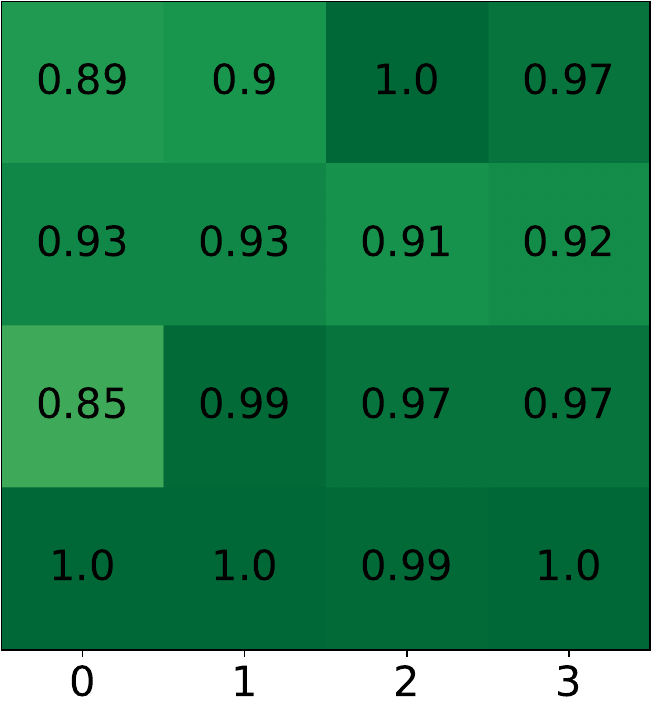}} & \multirow{4}{*}{\includegraphics[width=0.2\textwidth]{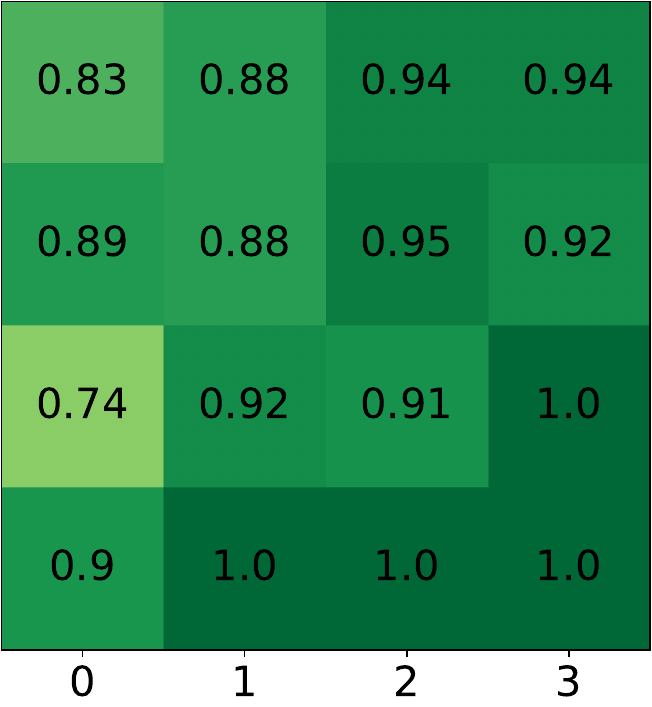}} \\
    \raisebox{-1.2\height}{\geminilogo} &&&&\\
    \raisebox{-1.0\height}{\sonnetlogo} &&&&\\
    \raisebox{-1.2\height}{\newsonnetlogo} &&&&\\ 
    \\
    &\scalebox{0.6}{(e) \civilization} & \scalebox{0.6}{(f) \questionWord} & \scalebox{0.6}{(g) \tournament} & \scalebox{0.6}{(h) \environmental}\\
    \end{tabular}
    \caption{VLMs tend to be significantly more accurate when there are white spaces between letters.}
    \label{fig:mult-word-space}
\end{figure}

As shown in \cref{fig:mult-word-space}, VLMs perform better when we add more white space between letters of a word, \eg, $\triangle$ for \gpt on the words \Acknowledgement and \Subdermatoglyphic from 0 to 3 spaces are 11\% and 21\%, respectively.

\clearpage
\subsection{Additional Examples}
\label{app:redoval_qual}
\begin{figure}
\centering
\begin{AIbox}{Which character is being highlighted with a red oval?  Please provide your answer in curly brackets{,} e.g. \{a\}}
\centering

    \begin{tabular}{lp{0.8cm}c|p{0.8cm}c|p{0.8cm}c|p{0.8cm}c|p{0.8cm}c|p{0.8cm}c}
     & \multicolumn{2}{c}{\includegraphics[width=0.15\textwidth]{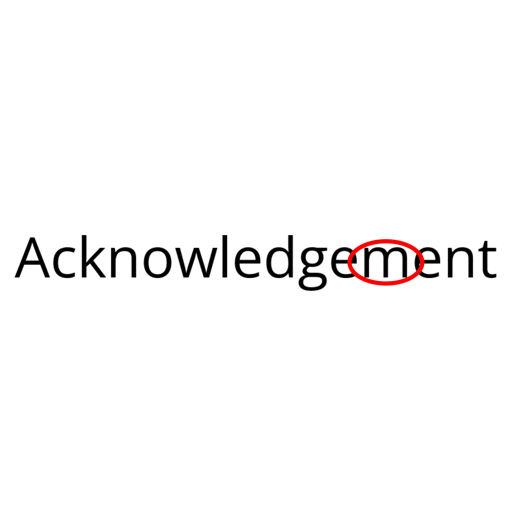}} &
     \multicolumn{2}{c}{\includegraphics[width=0.15\textwidth]{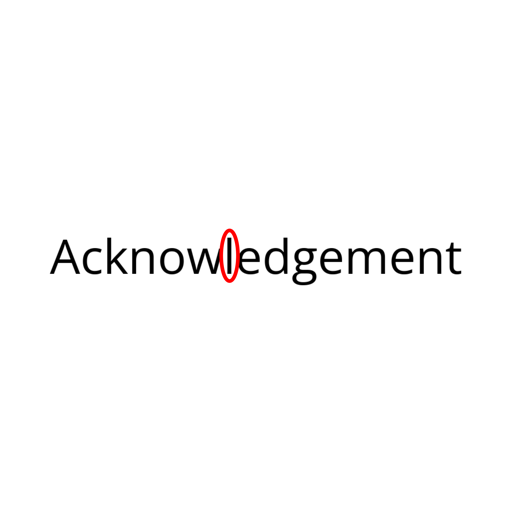}} &
     \multicolumn{2}{c}{\includegraphics[width=0.15\textwidth]{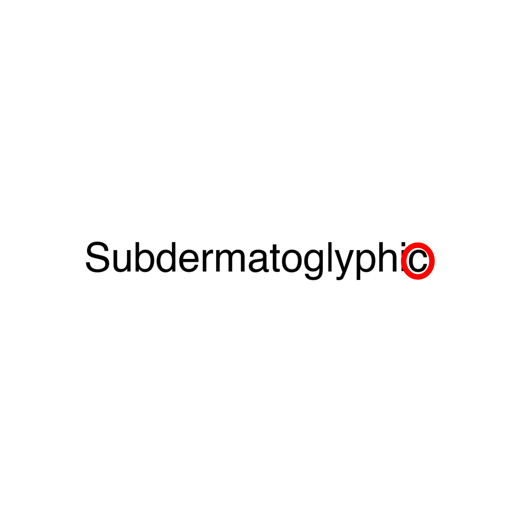}} & 
     \multicolumn{2}{c}{\includegraphics[width=0.15\textwidth]{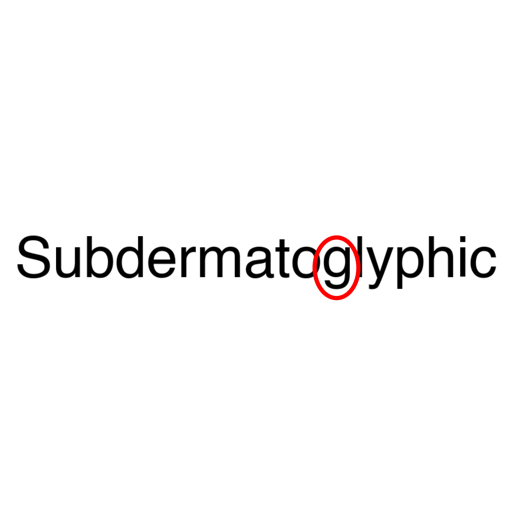}} &
     \multicolumn{2}{c}{\includegraphics[width=0.15\textwidth]{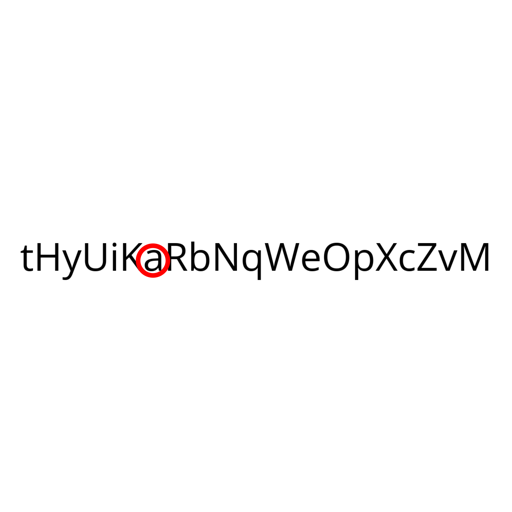}} &
     \multicolumn{2}{c}{\includegraphics[width=0.15\textwidth]{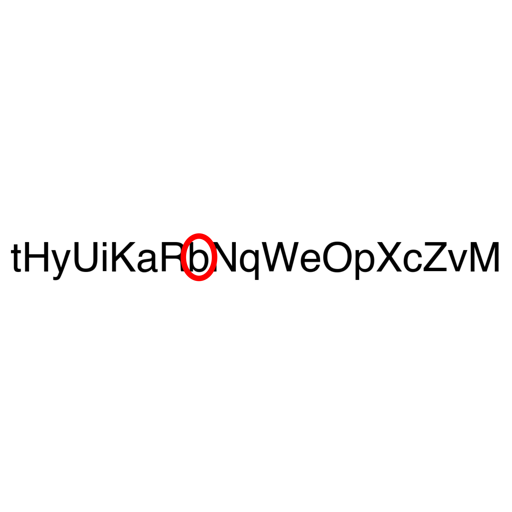}} \\
     \rowcolor{lightgray}
     \raisebox{-0.2\height}\gptlogo & \centering e & \textcolor{red}{\xmark} &  \centering e & \textcolor{red}{\xmark}&  \centering c & \textcolor{ForestGreen}{\cmark} &  \centering g & \textcolor{ForestGreen}{\cmark} &  \centering @  & \textcolor{red}{\xmark}&  \centering  r & \textcolor{red}{\xmark}\\
     \raisebox{-0.2\height}\geminilogo & \centering m & \textcolor{ForestGreen}{\cmark} & \centering l & \textcolor{ForestGreen}{\cmark}&  \centering c & \textcolor{ForestGreen}{\cmark} &  \centering o & \textcolor{red}{\xmark} &  \centering a & \textcolor{ForestGreen}{\cmark}&  \centering b & \textcolor{ForestGreen}{\cmark}\\
     \rowcolor{lightgray}
     \raisebox{-0.2\height}\sonnetlogo & \centering e & \textcolor{red}{\xmark} & \centering n & \textcolor{red}{\xmark}&  \centering i & \textcolor{red}{\xmark} &  \centering o & \textcolor{red}{\xmark} &  \centering i & \textcolor{red}{\xmark}&  \centering n & \textcolor{red}{\xmark}\\
       \raisebox{-0.2\height}\newsonnetlogo & \centering e & \textcolor{red}{\xmark} & \centering l & \textcolor{ForestGreen}{\cmark}&  \centering © & \textcolor{red}{\xmark} &  \centering g & \textcolor{ForestGreen}{\cmark} &  \centering a & \textcolor{ForestGreen}{\cmark}&  \centering b & \textcolor{ForestGreen}{\cmark}\\

     \end{tabular}

    \vspace{4pt}
    \centering
    \begin{tabular}{cccccccccccccc}
    \raisebox{-0.1\height}\gptlogo & \gpt &&&
      \raisebox{-0.1\height}\geminilogo & \gemini & &&\raisebox{-0.1\height}\sonnetlogo & \sonnet & && \raisebox{-0.12\height}\newsonnetlogo & \newsonnet \\
      \end{tabular}
     
\end{AIbox}
\caption{Most failure cases consist of predicting the adjacent letters (predicting ``e'' instead of ``m'' in \gpt, \sonnet, and \newsonnet leftmost), and confusing the red oval as part of the circled letter (predicting ``\copyright'' instead of ``c'' in \newsonnet).}
\label{appfig:circled_word_prompt2}
\end{figure}

Examples from our evaluation of VLMs on the circled letter task are shown in \cref{appfig:circled_word_prompt2}.

\clearpage
\section{Counting the number of line intersections task}
\label{app:lineplots_intersection}

\subsection{Benchmark Construction Details}
\label{app:2dline-bench}
To create our benchmark, we use 5 parameters to control the diversity of the samples.

\begin{itemize}
    \item \textbf{Color}: We fix the colors for each line to use \{\emph{blue}, \emph{red}\}
    \item \textbf{Image size}: We generate 3 different image resolutions, \{384, 768, 1152\}px, to include high and low-resolution samples in our tests.
    \item \textbf{Grid size}: We divide the whole image into a 12$\times$12 grid with 144 equal sized cells. 
    Then, we choose 3 points for each (blue and red) line on the grid to draw the line plots. 
    To make sure the line plots do not overlap and the intersections are well distanced from the edges, we fix the x-coordinates, and randomly sample y-coordinates from the invisible 12$\times$12 grid. 
    The grid sampling also avoids picking the same coordinates for both blue and red line.
    \item \textbf{X-coordinate}: For each point, we choose x$ = 0, \frac{C}{2}, C$ px.
    \item \textbf{Y-coordinate}: 
    For each x-coordinate above, we randomly sample a y-coordinate from 12 pre-defined rows in the 12$\times$12 grid.
    We also ensure that no blue and red points share the exact same (x,y) coordinates.
    \item \textbf{Line thickness}: We vary the line widths with standard \emph{matplotlib} values (2 and 4), which renders into a width of $0.005\times C$ and $0.01\times C$ pixels, respectively.
    
    \item \textbf{Number of intersections}: We count the intersections based on the three points defined for each line (\((x, y_1)\), \((x, y_2)\), and \((x, y_3)\)).
\end{itemize}

\begin{figure}
    \centering
    \begin{tabular}{ccc}
         \includegraphics[width=0.2\textwidth]{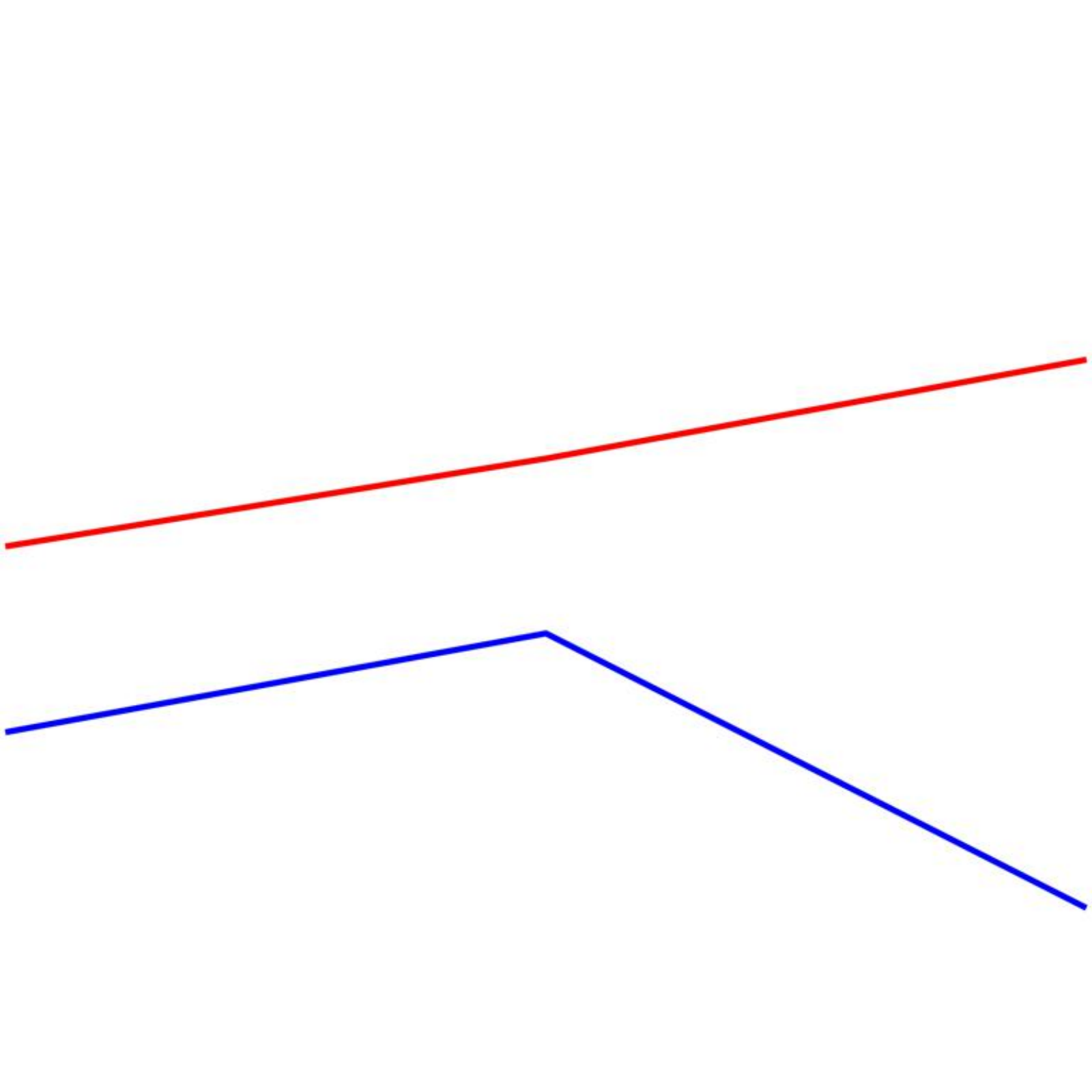}& \includegraphics[width=0.2\textwidth]{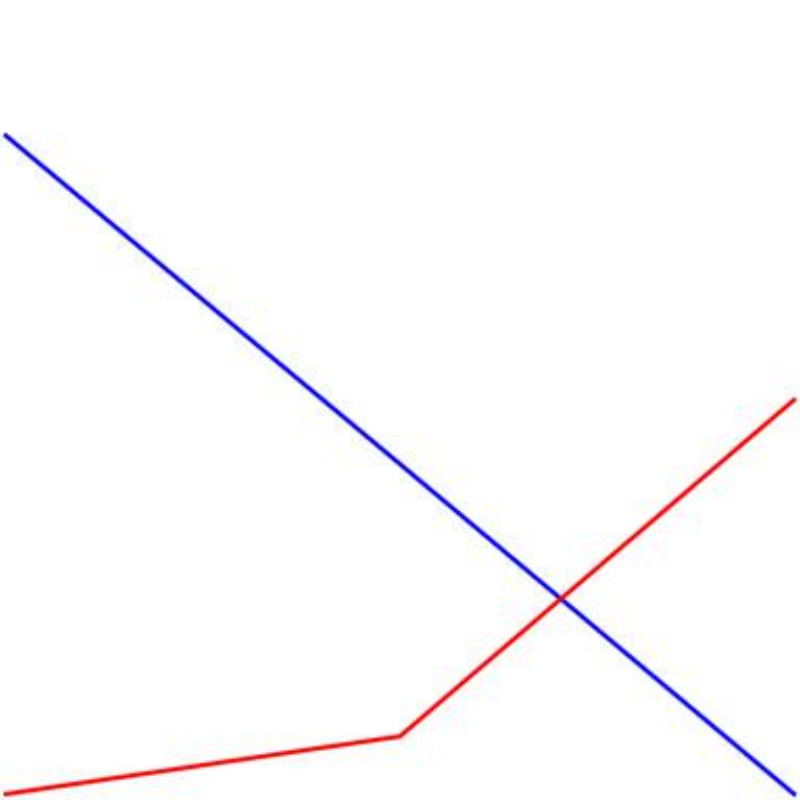} & \includegraphics[width=0.2\textwidth]{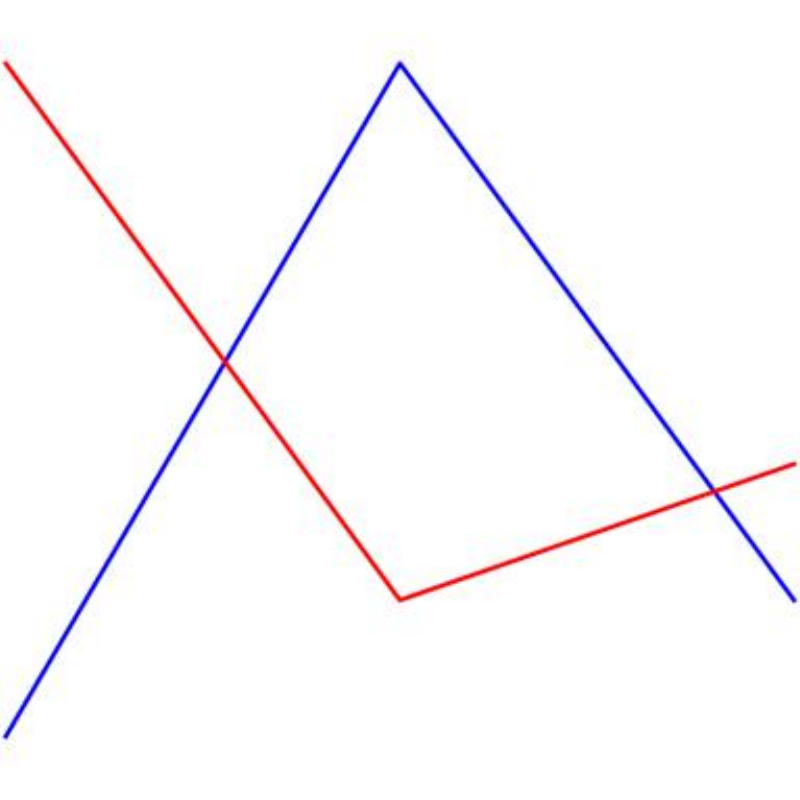}\\
         \includegraphics[width=0.2\textwidth]{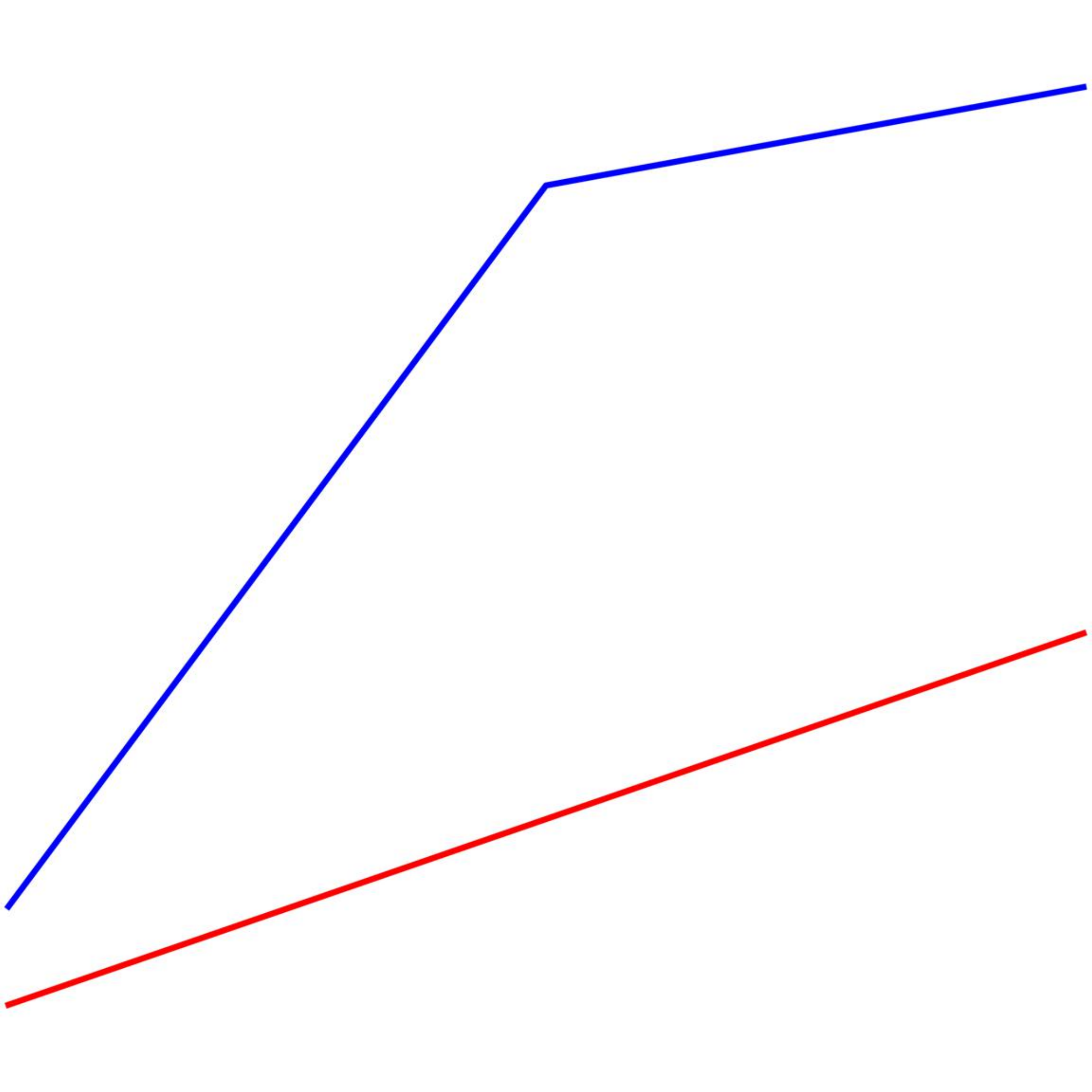}& \includegraphics[width=0.2\textwidth]{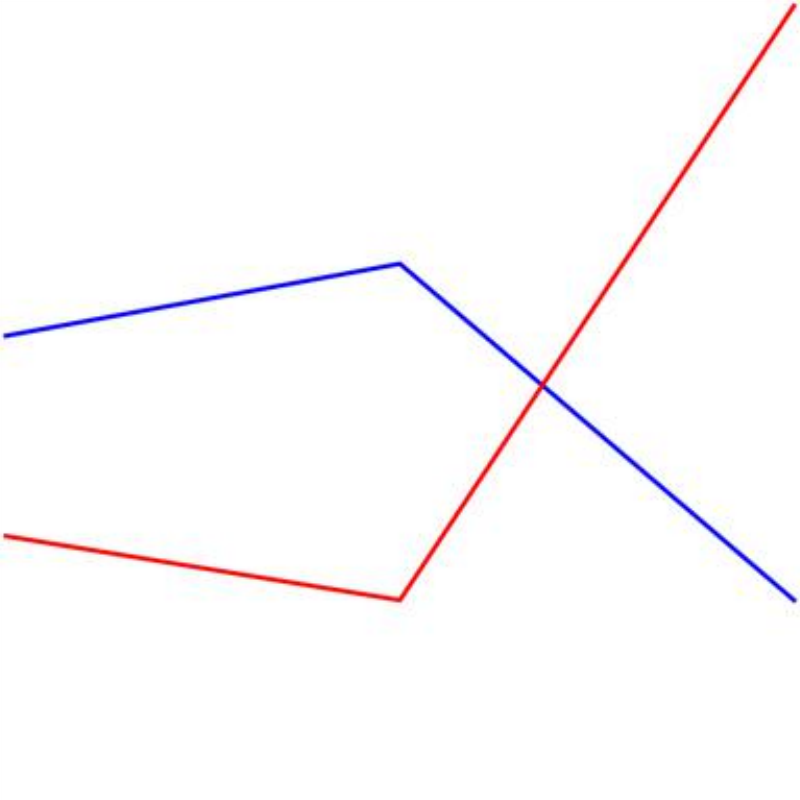} & \includegraphics[width=0.2\textwidth]{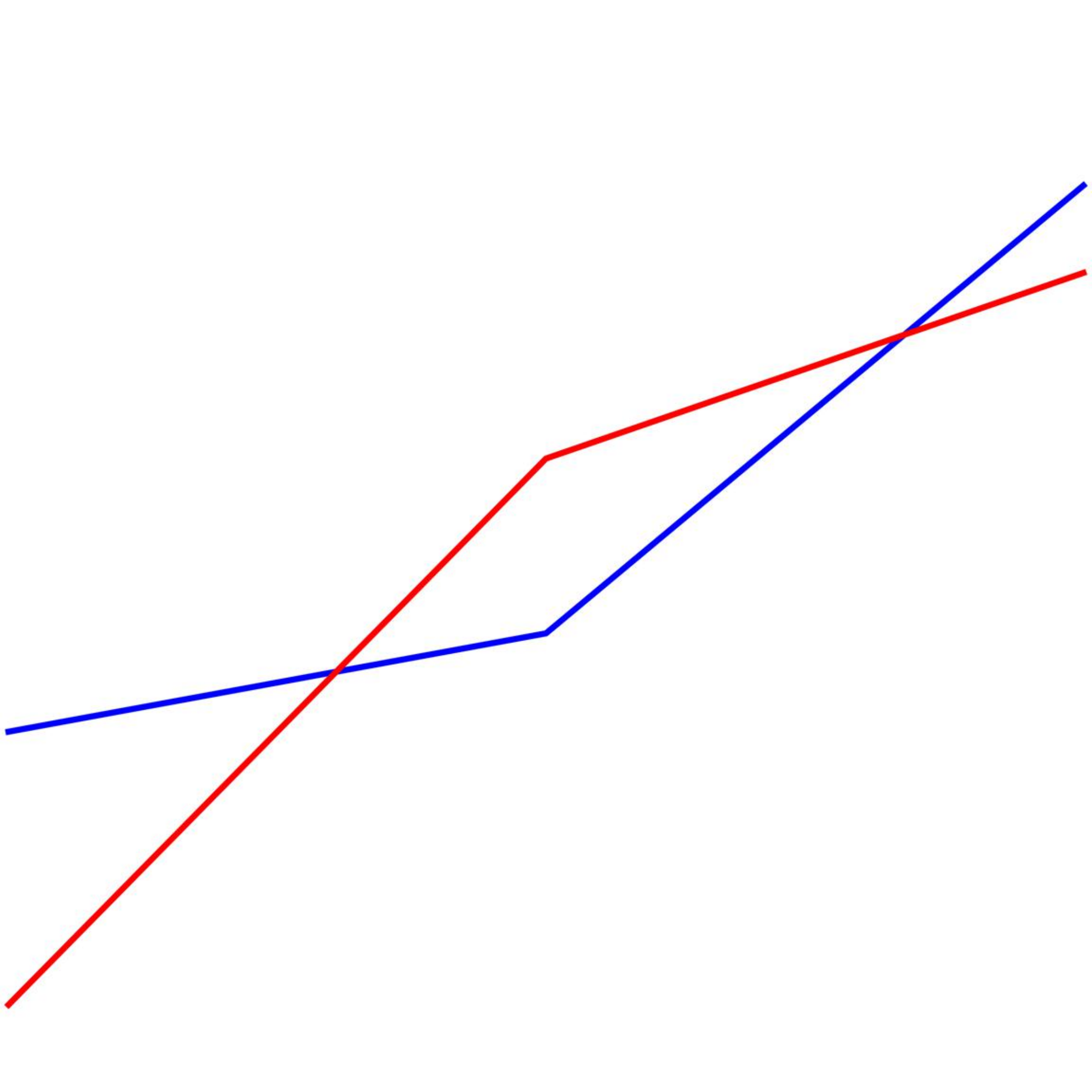}\\
         \includegraphics[width=0.2\textwidth]{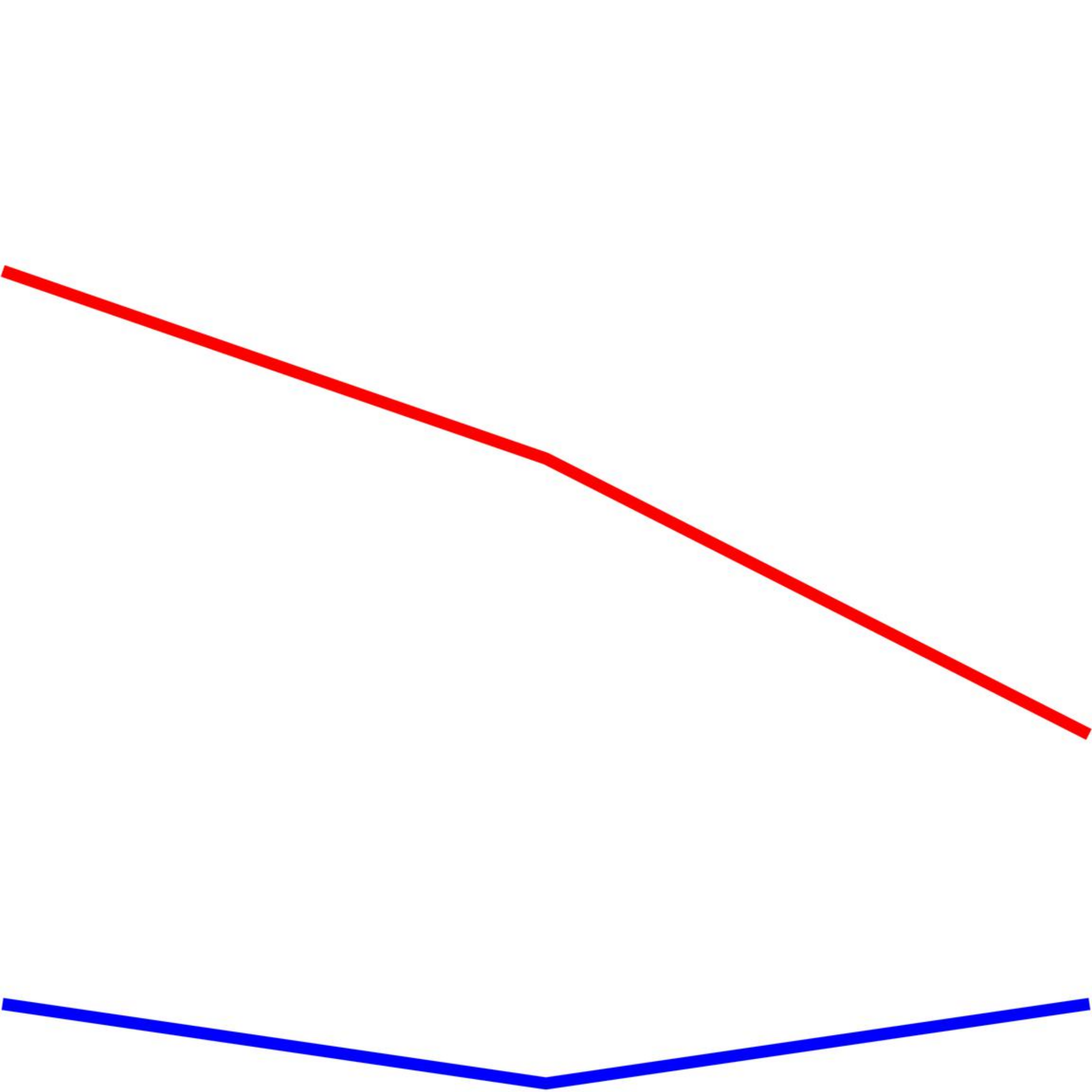}& \includegraphics[width=0.2\textwidth]{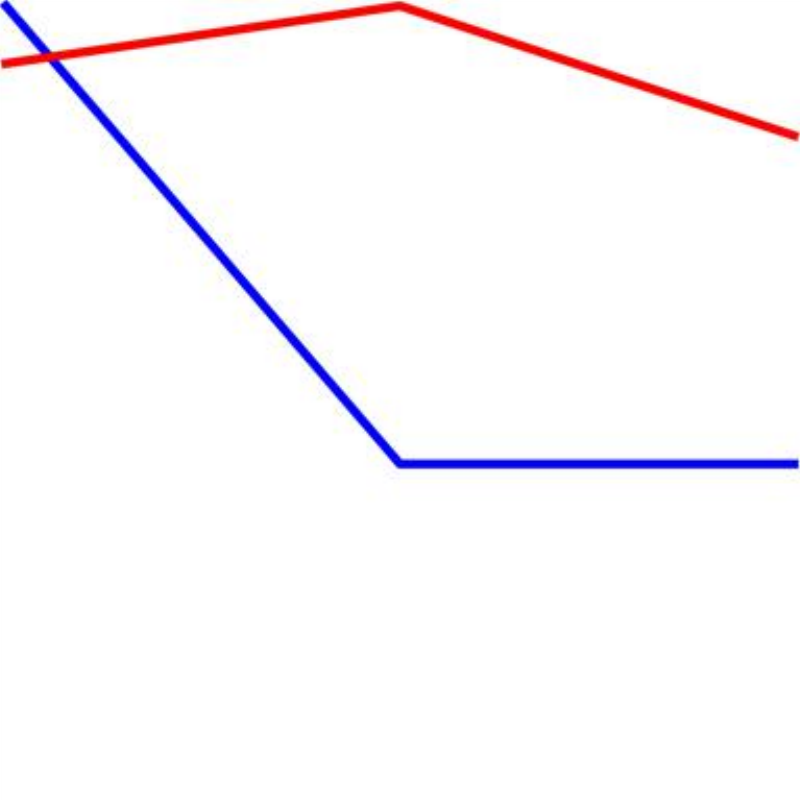} & \includegraphics[width=0.2\textwidth]{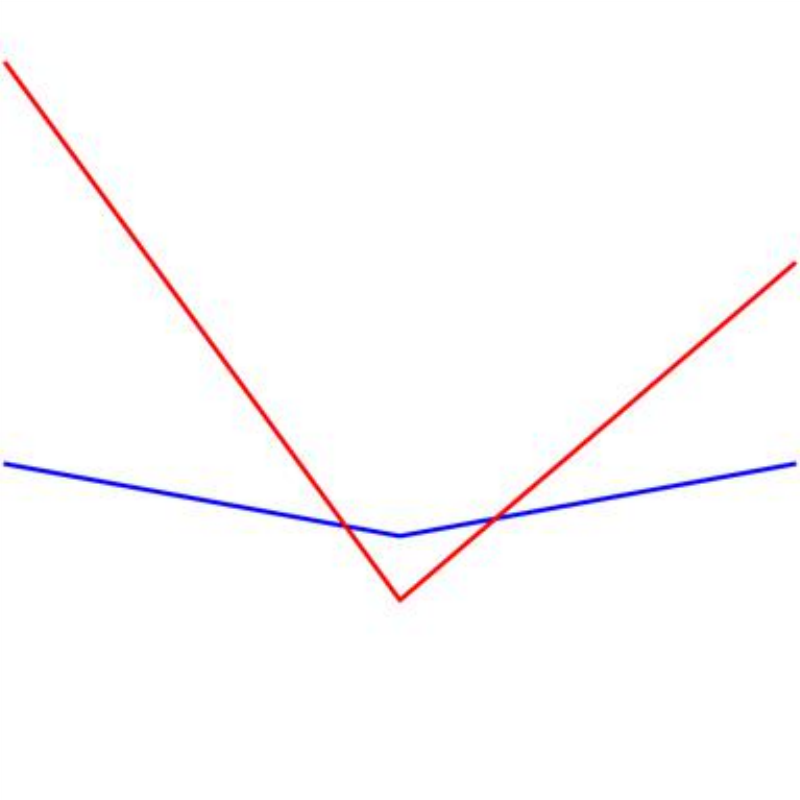}\\
         (a) zero intersection & (b) one intersection & (c) two intersections \\
    \end{tabular}
    \caption{Samples from the two intersecting lines benchmark that contain 0, 1, or 2 line intersections.}
    \label{app:fig-2dline-examples}
\end{figure}

We repeat the process until we have 2 line widths$\times$3 image sizes$\times$100 samples of 0, 1, and 2 intersections, resulting in 1800 images (see \cref{table:line-intersection-parameters,app:fig-2dline-examples}). 

\subsec{Code} The code to generate the images is at \url{https://github.com/anguyen8/vision-llms-are-blind/blob/main/src/LineIntersection/2dline.ipynb}.

\subsec{Prompts}    
We ask each question using two different wordings:
\begin{enumerate}
    \item \emph{How many times do the blue and red lines touch each other? Answer with a number in curly brackets, e.g., \{5\}.}
    \item \emph{Count the intersection points where the blue and red lines meet. Put your answer in curly brackets, e.g., \{2\}.}
\end{enumerate}

\noindent\textbf{Groundtruth} answers are $\in \{0, 1, 2\}$ (random-baseline accuracy: 33.33\%).

\begin{figure}
\centering
        \begin{tabular}{ll}
        {GT = 0}\includegraphics[trim={0 0 0 300},clip,width=0.4\textwidth]{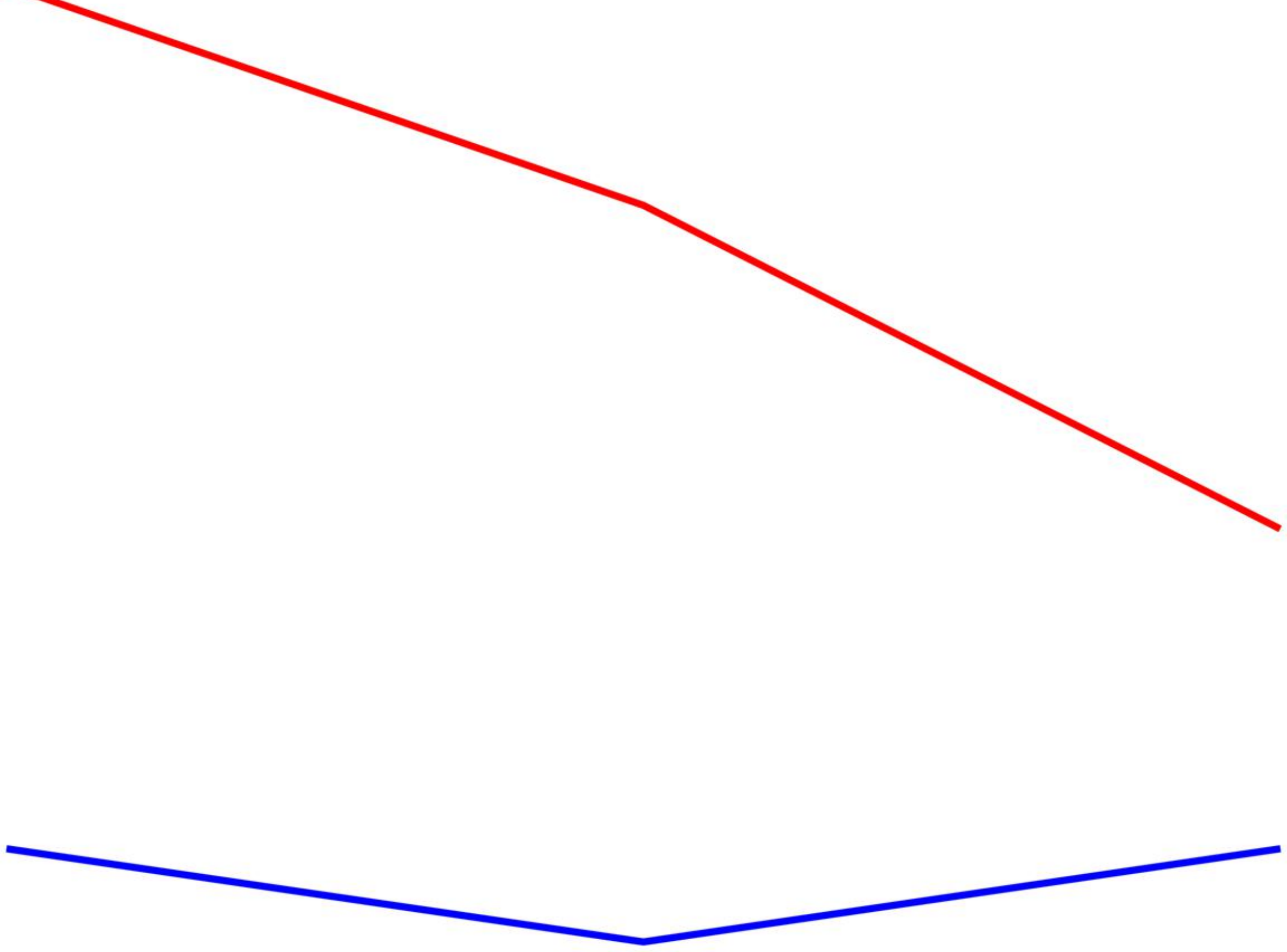}& 
         {GT = 1}\includegraphics[trim={0 400 0 0},clip,width=0.4\textwidth]{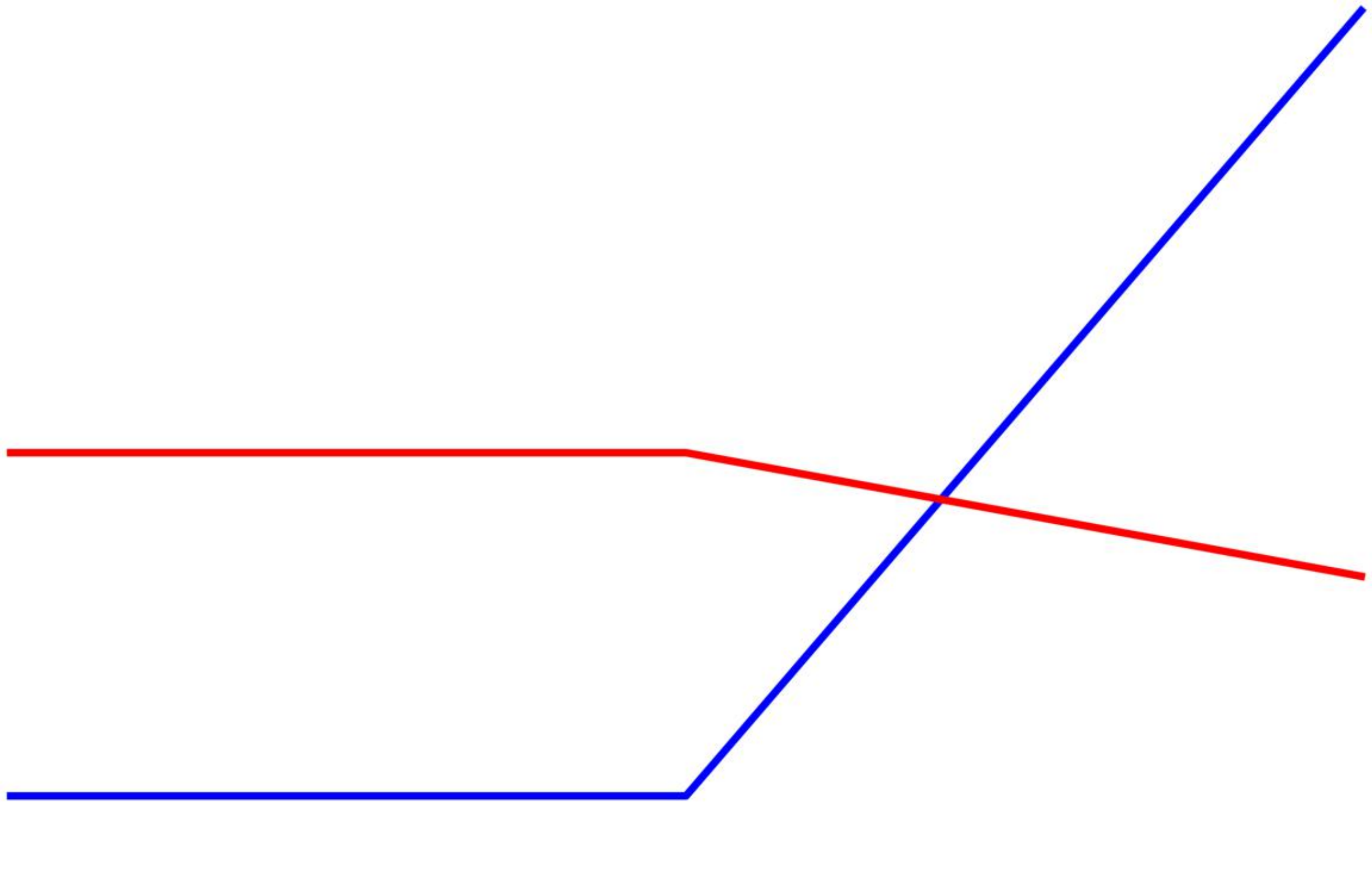}\\
         {GT = 1}\includegraphics[trim={0 0 0 0},clip,width=0.4\textwidth]{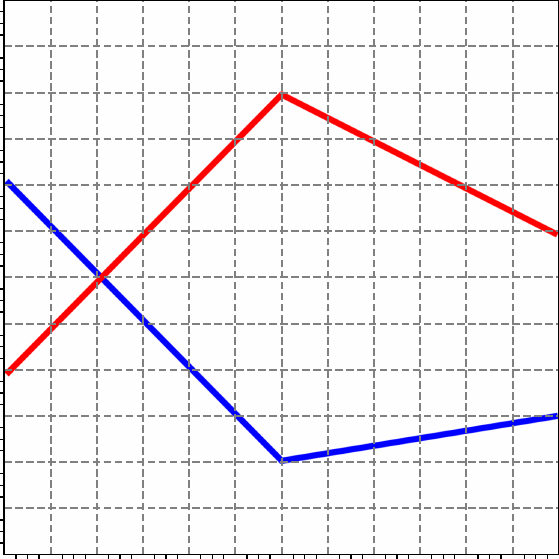}& 
         {GT = 2}\includegraphics[trim={0 0 0 0},clip,width=0.4\textwidth]{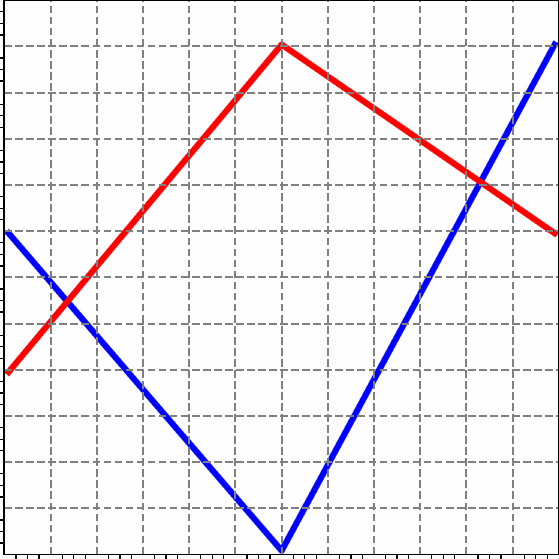} \\
    \end{tabular}
    \caption{Count intersections}
    \label{fig:2dline}
\end{figure}

\begin{table}
\centering
\caption{We generate the same number of images based on various parameters to have a balanced benchmark.}
\label{table:line-intersection-parameters}
\begin{tabular}{lccc}
\toprule
\textbf{Parameter} & \textbf{Values} & \textbf{Samples} & ~~\textbf{Total Samples}\\
\midrule
\multirow{3}{*}{Image size ($C$)} & 384px & 600 & ~~\multirow{2}{*}{1800}\\
& 768px & 600 \\
& 1152px & 600 \\
\midrule
\multirow{2}{*}{Line thickness} & $0.005\times C$ & 900 & ~~\multirow{2}{*}{1800}\\
 & $0.01\times C$ & 900 \\
\midrule
\multirow{3}{*}{Number of intersections} & 0 & 600 & ~~\multirow{3}{*}{1800}\\
 & 1 & 600 \\
 & 2 & 600 \\
\bottomrule
\end{tabular}
\end{table}

\clearpage
\subsection{Finding: image resolution does not influence VLM's ability to count the line intersections}
\label{app:2dline-thickness}

\cref{suppplot:linewidth-line-intersection} depicts that increasing the resolution in our plots does not help the VLMs see and count the intersections of 2D lines.

\subsection{Finding: VLMs perform similarly when the lines intersect twice}
\label{app:2dline-gt}
\begin{figure}
    \centering
    \includegraphics[width=\textwidth]{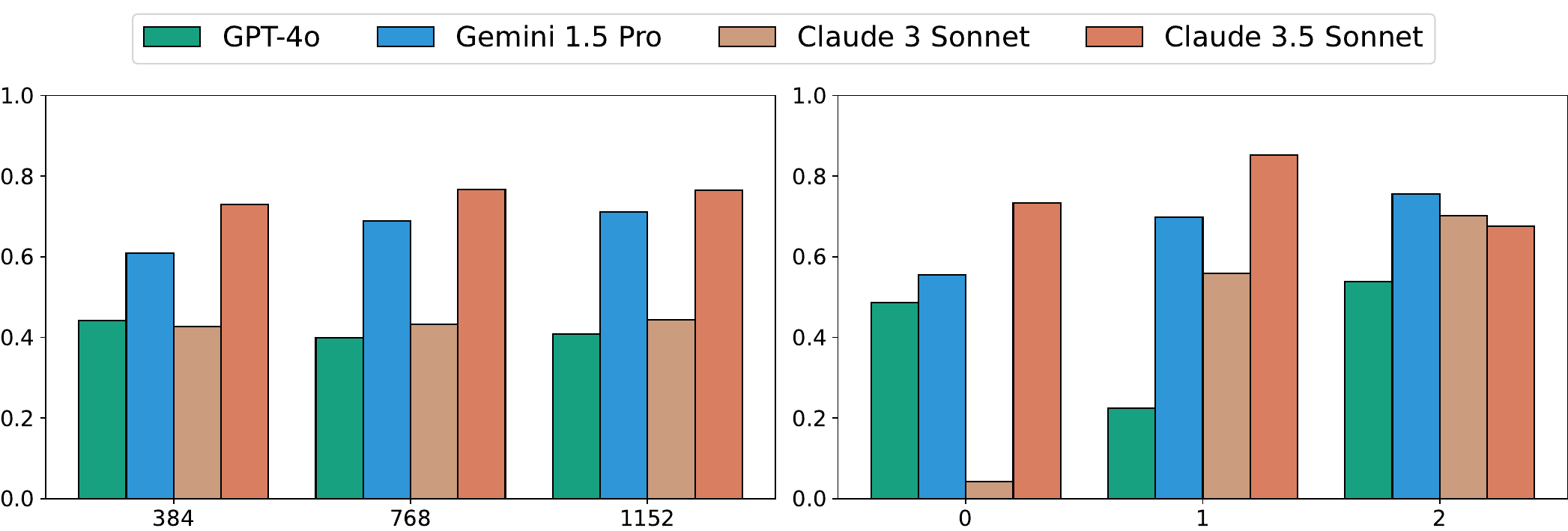}
    \caption{VLMs are not sensitive to the image resolution to see and count the number of intersections (left). When the lines intersect twice, VLMs perform similarly, while \newsonnet has the least variance over different numbers of intersections (right).}
    \label{suppplot:linewidth-line-intersection}
\end{figure}

Our benchmark consists of images with 0 to 1 line intersections with an identical number of samples in each category. We break down the performance of each model based on the groundtruth in \cref{suppplot:linewidth-line-intersection}-right to analyze where VLMs mostly fail in their overall performance. As depicted in \cref{suppplot:linewidth-line-intersection}-right, VLMs perform relatively better on 2 intersections (except \newsonnet that is best on 1 intersection) and perform worse on 0 intersections (\sonnet is the worst).

\clearpage
\subsection{Additional Examples}
\label{app:2dline-qual}

\begin{figure}
\centering
\begin{AIbox}{Count the intersection points where the blue and red lines meet. Put your answer in curly brackets{,} e.g.{,} \{2\}.}
\centering

    \begin{tabular}{lp{0.8cm}c|p{0.8cm}c|p{0.8cm}c|p{0.8cm}c|p{0.8cm}c|p{0.8cm}c}
     &\multicolumn{2}{c}{\includegraphics[width=0.15\textwidth]{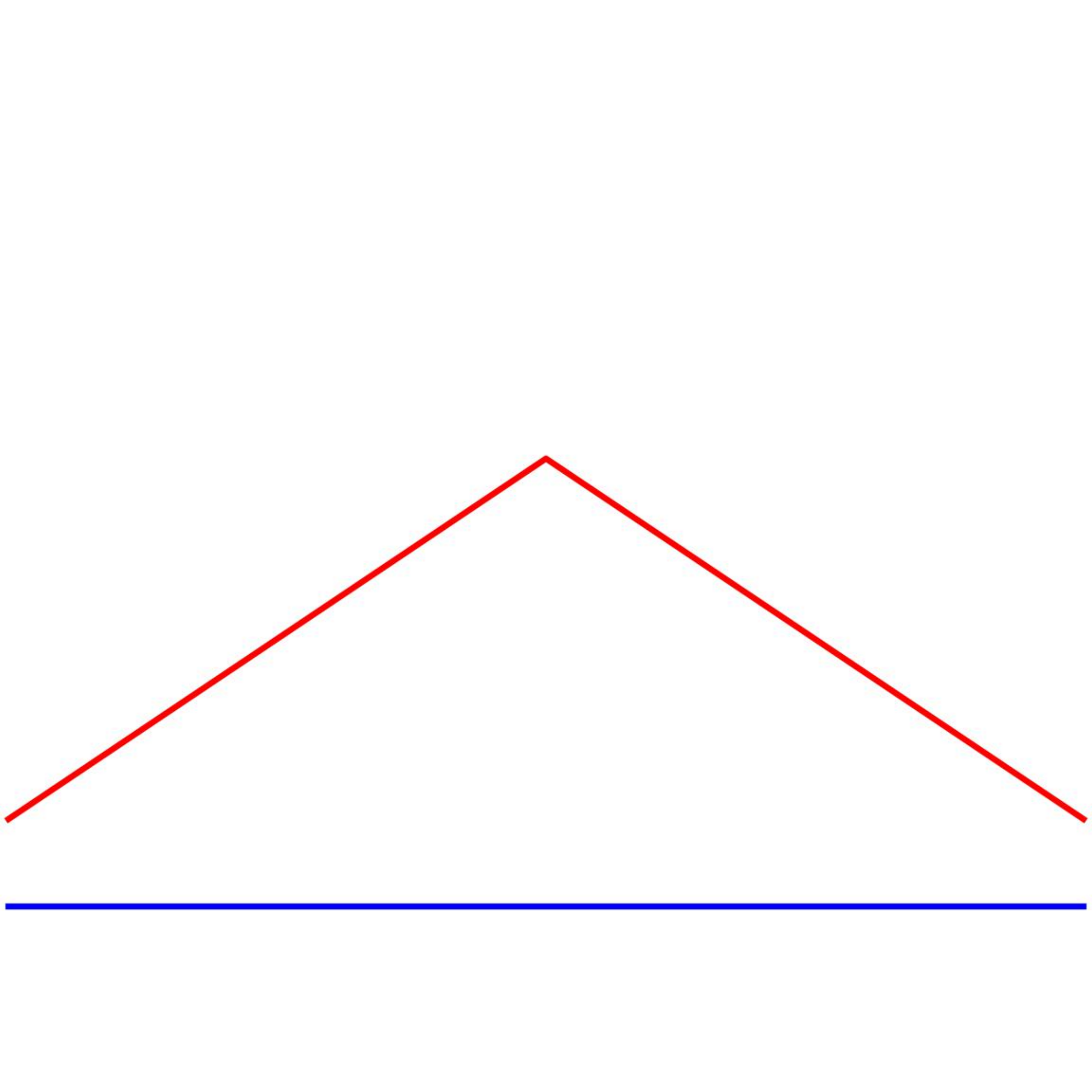}} & 
     \multicolumn{2}{c}{\includegraphics[width=0.15\textwidth]{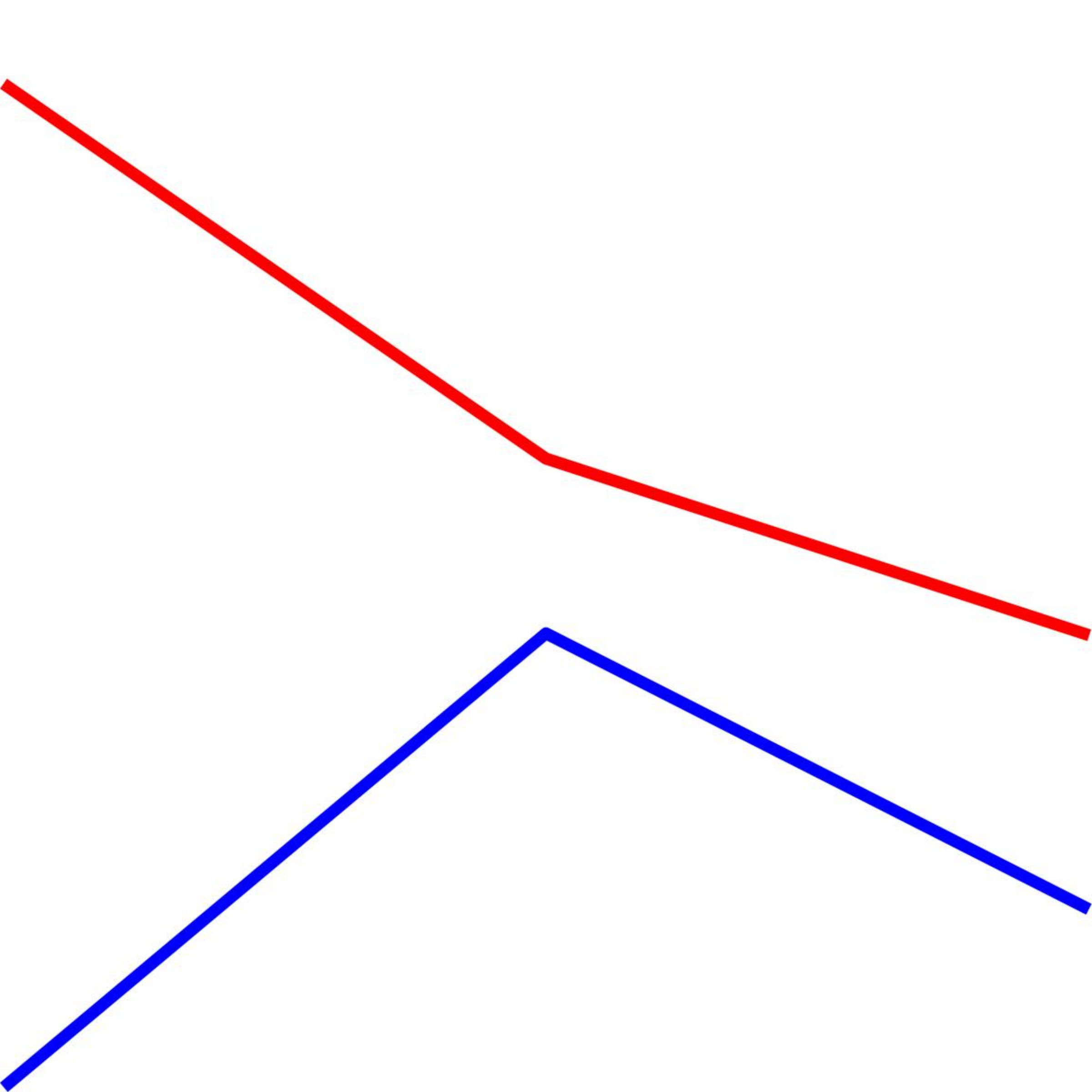}} &
     \multicolumn{2}{c}{\includegraphics[width=0.15\textwidth]{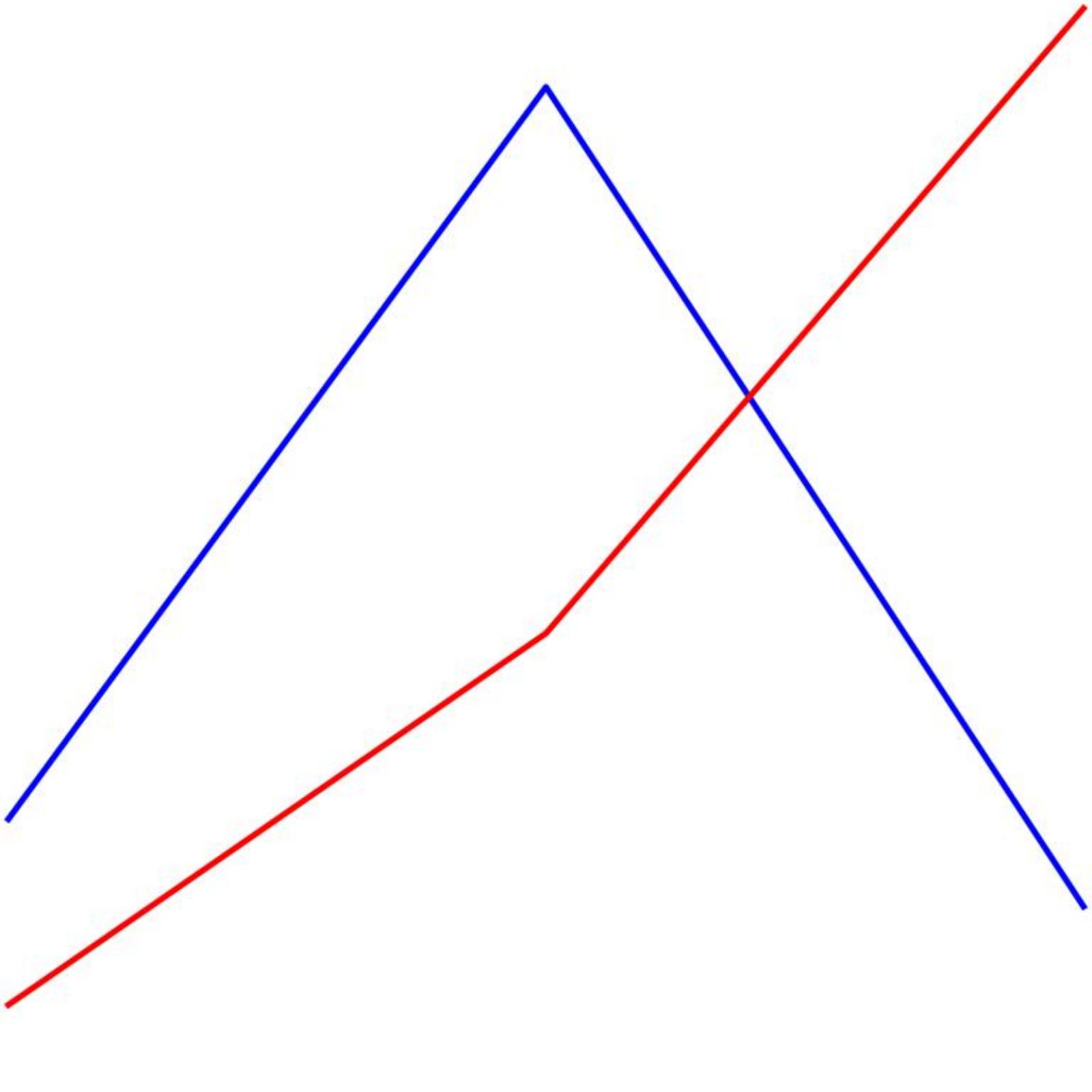}} &
     \multicolumn{2}{c}{\includegraphics[width=0.15\textwidth]{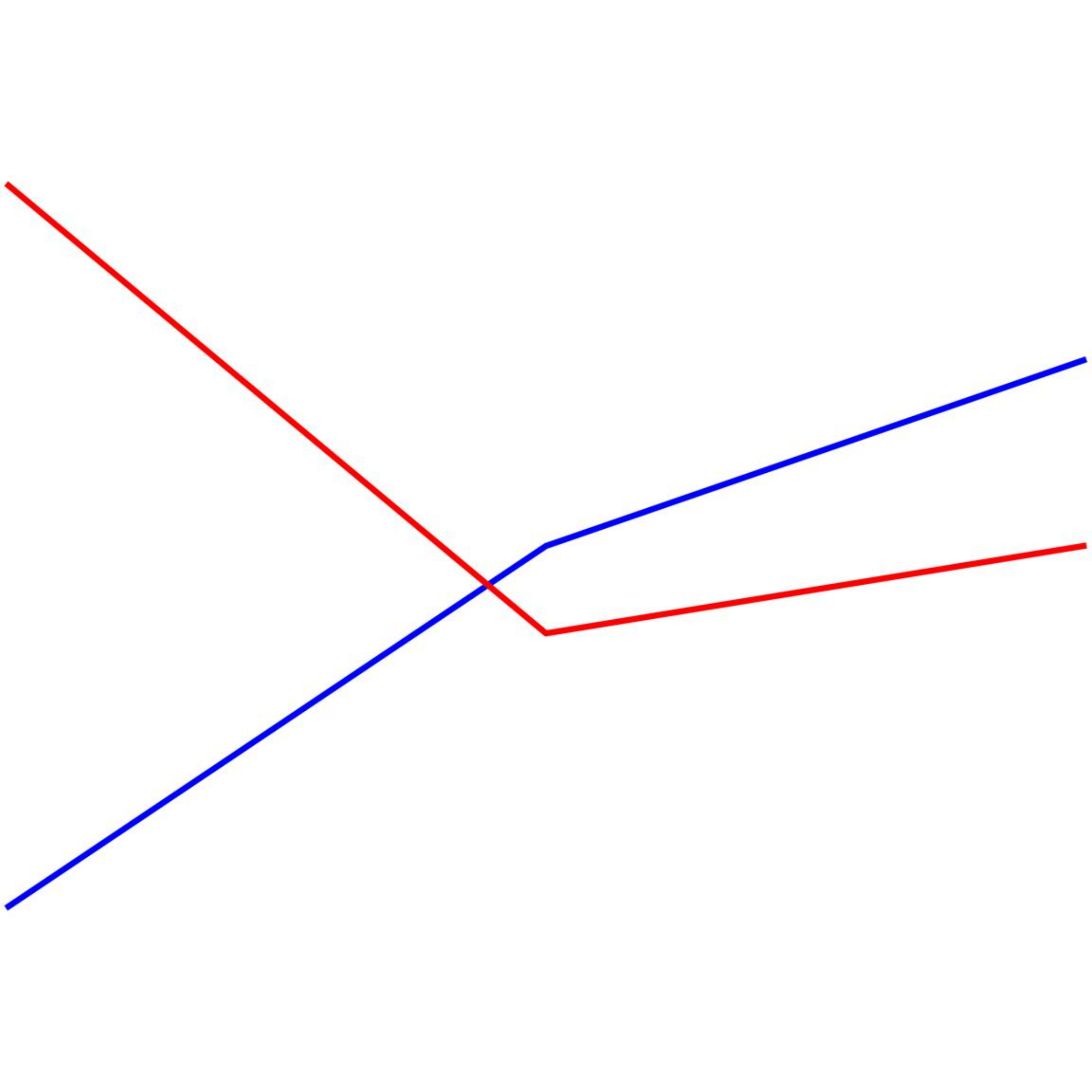}} &
     \multicolumn{2}{c}{\includegraphics[width=0.15\textwidth]{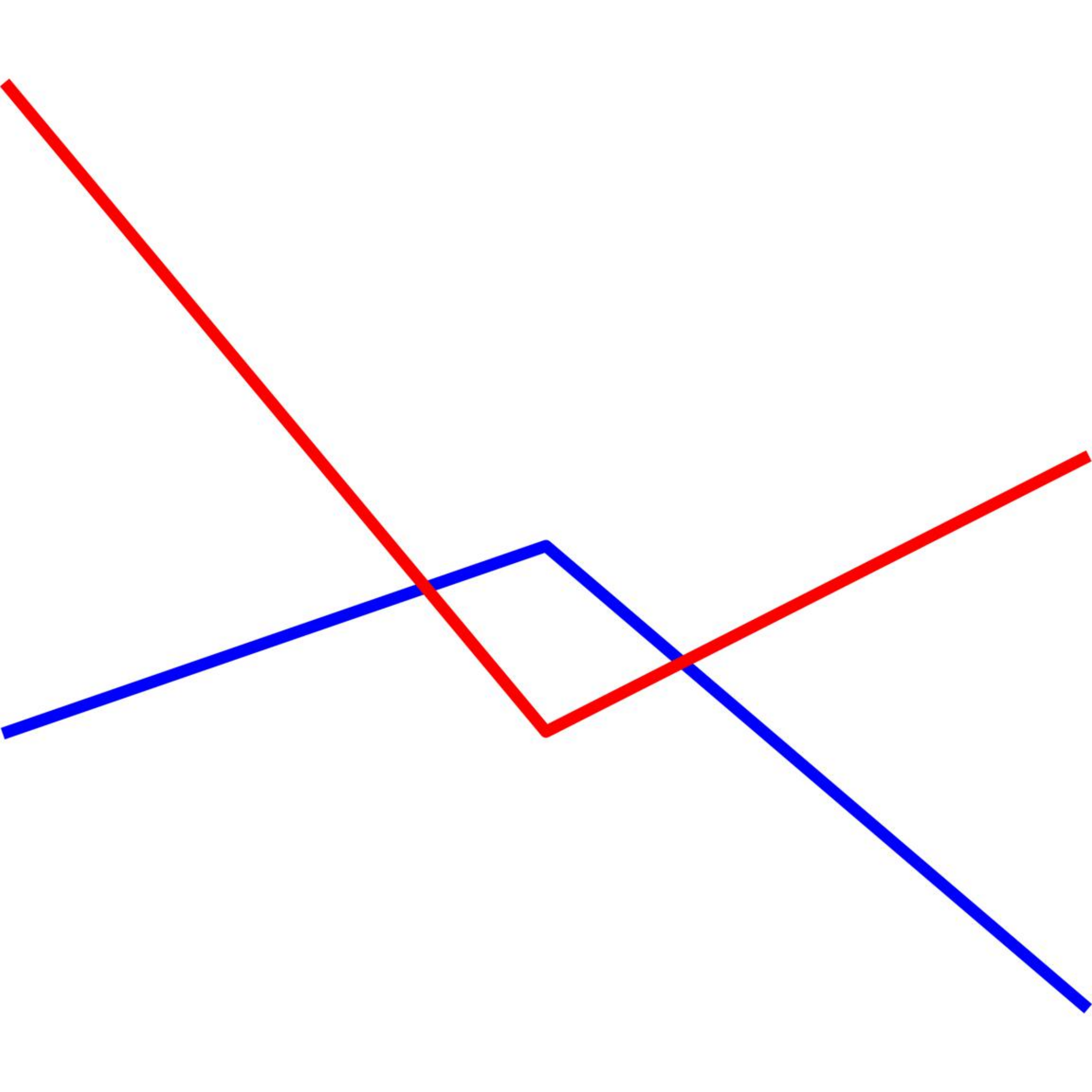}}&
     \multicolumn{2}{c}{\includegraphics[width=0.15\textwidth]{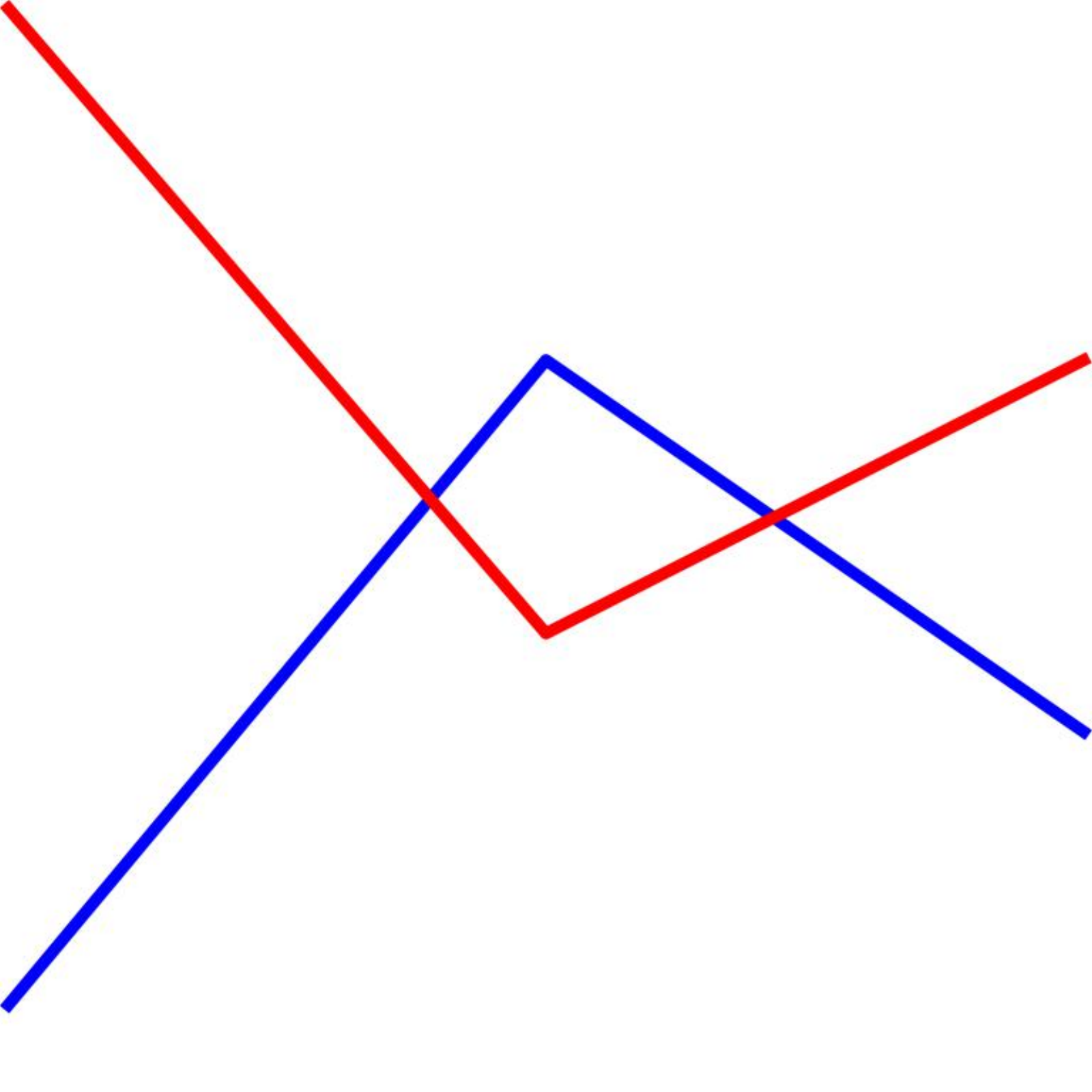}}\\
     \rowcolor{lightgray}
     \raisebox{-0.2\height}\gptlogo & \centering 2 & \textcolor{red}{\xmark} &  \centering 0 & \textcolor{ForestGreen}{\cmark}&  \centering 2 & \textcolor{red}{\xmark} &  \centering 2 & \textcolor{red}{\xmark} &  \centering 3 & \textcolor{red}{\xmark}&  \centering 4 & \textcolor{red}{\xmark}\\
     \raisebox{-0.2\height}\geminilogo & \centering 2 & \textcolor{red}{\xmark} & \centering 1 & \textcolor{red}{\xmark}&  \centering 1 & \textcolor{ForestGreen}{\cmark} &  \centering 1 & \textcolor{ForestGreen}{\cmark} &  \centering 2 & \textcolor{ForestGreen}{\cmark}&  \centering 2 & \textcolor{ForestGreen}{\cmark}\\
     \rowcolor{lightgray}
     \raisebox{-0.2\height}\sonnetlogo & \centering 2 & \textcolor{red}{\xmark} & \centering 2 & \textcolor{red}{\xmark}&  \centering 2 & \textcolor{red}{\xmark} &  \centering 1 & \textcolor{ForestGreen}{\cmark} &  \centering 4 & \textcolor{red}{\xmark}&  \centering 1 & \textcolor{red}{\xmark}\\
       \raisebox{-0.2\height}\newsonnetlogo & \centering 2 & \textcolor{red}{\xmark} & \centering 1 & \textcolor{red}{\xmark}&  \centering 1 & \textcolor{ForestGreen}{\cmark} &  \centering 1 & \textcolor{ForestGreen}{\cmark} &  \centering 2 & \textcolor{ForestGreen}{\cmark}&  \centering 2 & \textcolor{ForestGreen}{\cmark}\\

     \end{tabular}

    \vspace{4pt}
    \centering
    \begin{tabular}{ccccccccccc}
    \raisebox{-0.1\height}\gptlogo & \gpt &&
      \raisebox{-0.1\height}\geminilogo & \gemini && \raisebox{-0.1\height}\sonnetlogo & \sonnet && \raisebox{-0.12\height}\newsonnetlogo & \newsonnet \\
      \end{tabular}
     
\end{AIbox}
\caption{All VLMs, except \newsonnet, fail at counting when there is no intersection, even when the gap between 2 lines is large (second image on the left).}
\label{appfig:lineintersection_prompt1}
\end{figure}

We show examples of models' responses to the prompts on the counting the number of line intersections task in \cref{appfig:lineintersection_prompt1}.

\begin{figure}[H]
\centering
\begin{AIbox}{How many times do the blue and red lines touch each other? Answer with a number in curly brackets{,} e.g.{,} \{5\}.}
\centering

    \begin{tabular}{lp{0.8cm}c|p{0.8cm}c|p{0.8cm}c|p{0.8cm}c|p{0.8cm}c|p{0.8cm}c}
     &\multicolumn{2}{c}{\includegraphics[width=0.15\textwidth]{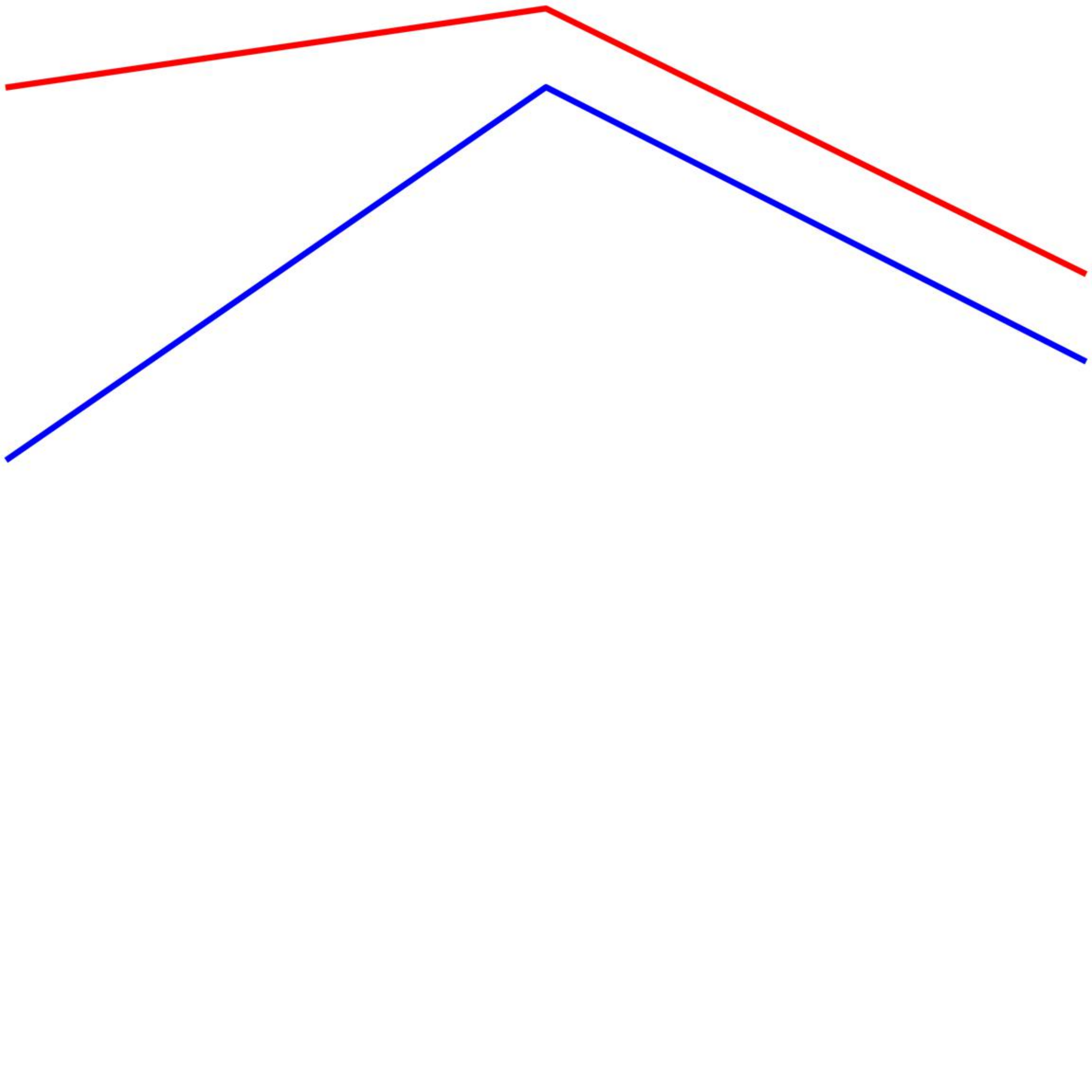}} & 
     \multicolumn{2}{c}{\includegraphics[width=0.15\textwidth]{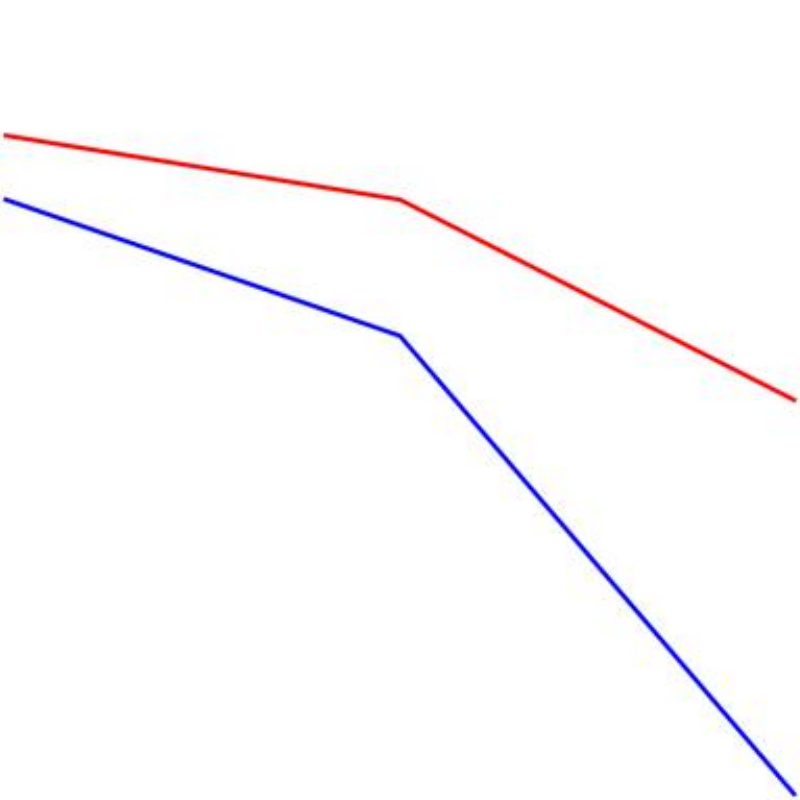}} &
     \multicolumn{2}{c}{\includegraphics[width=0.15\textwidth]{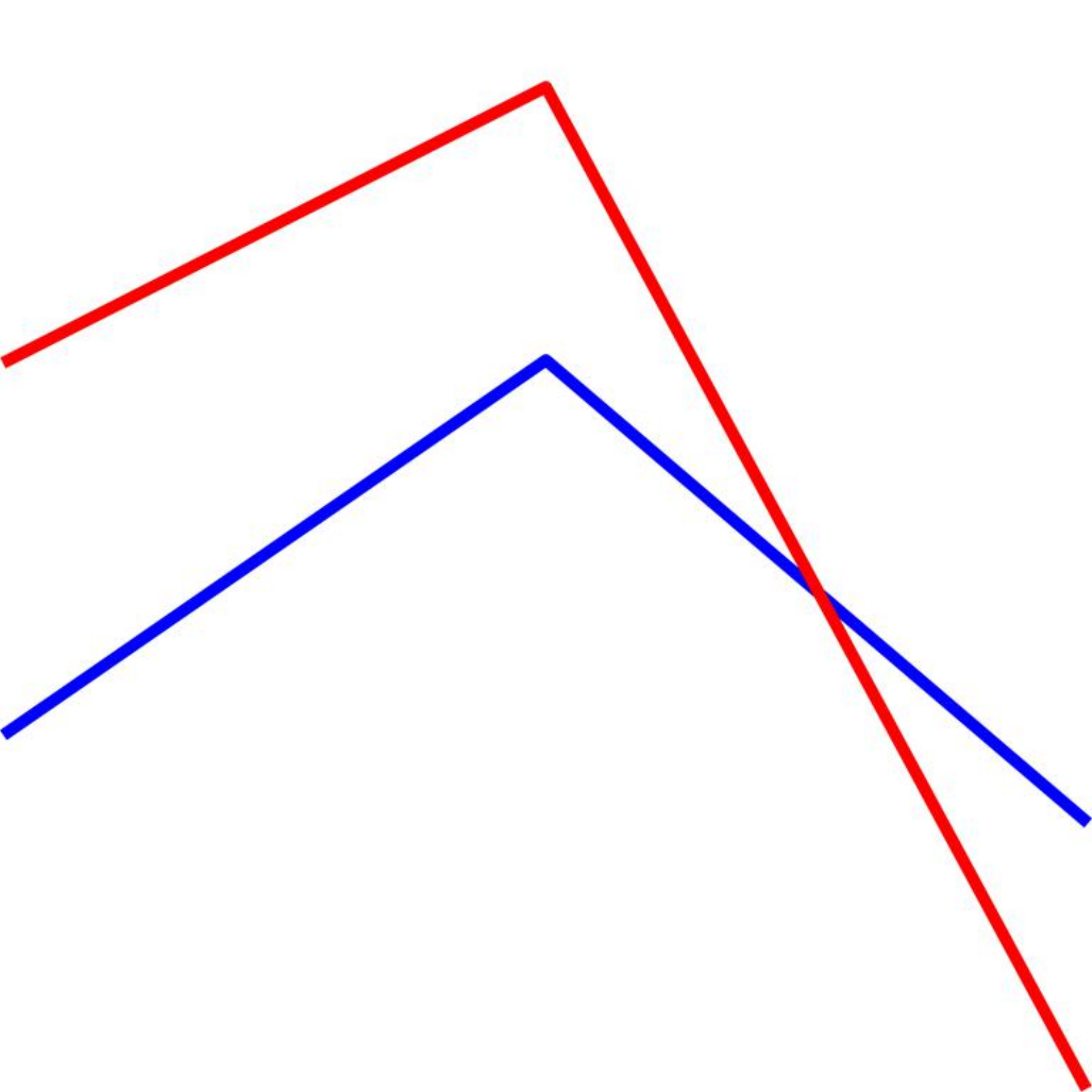}} &
     \multicolumn{2}{c}{\includegraphics[width=0.15\textwidth]{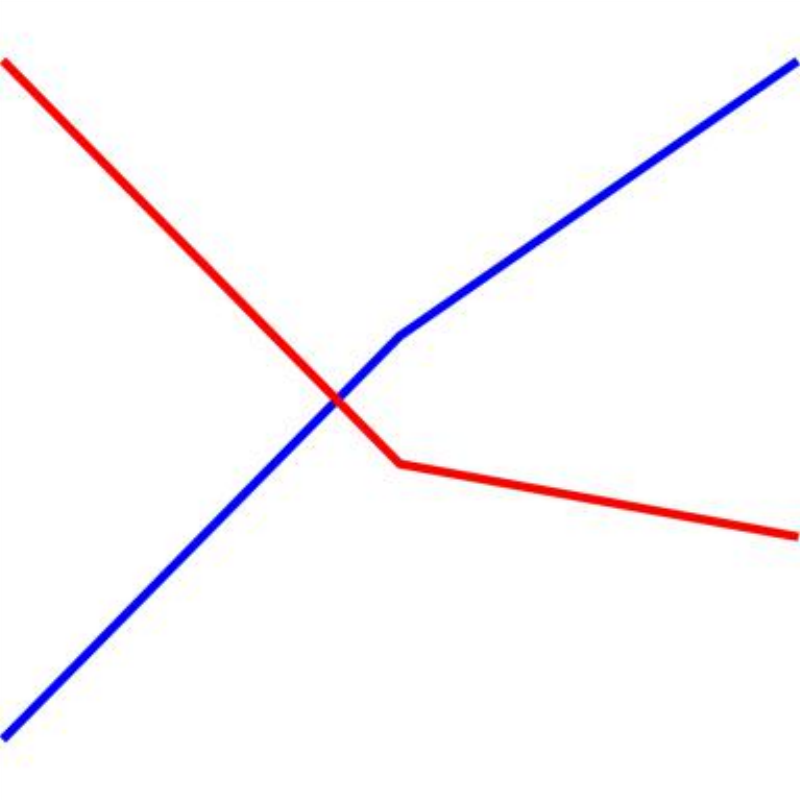}} &
     \multicolumn{2}{c}{\includegraphics[width=0.15\textwidth]{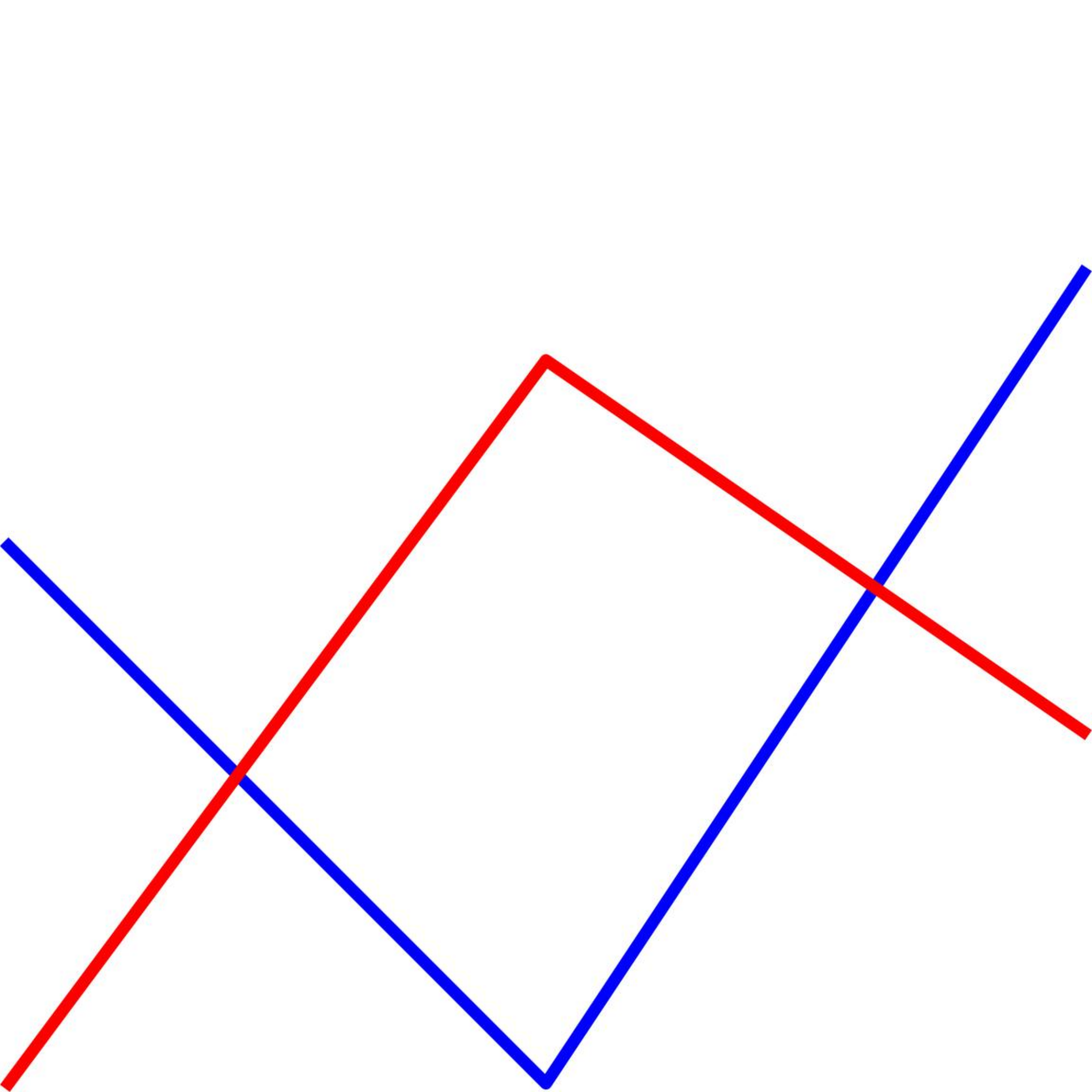}}&
     \multicolumn{2}{c}{\includegraphics[width=0.15\textwidth]{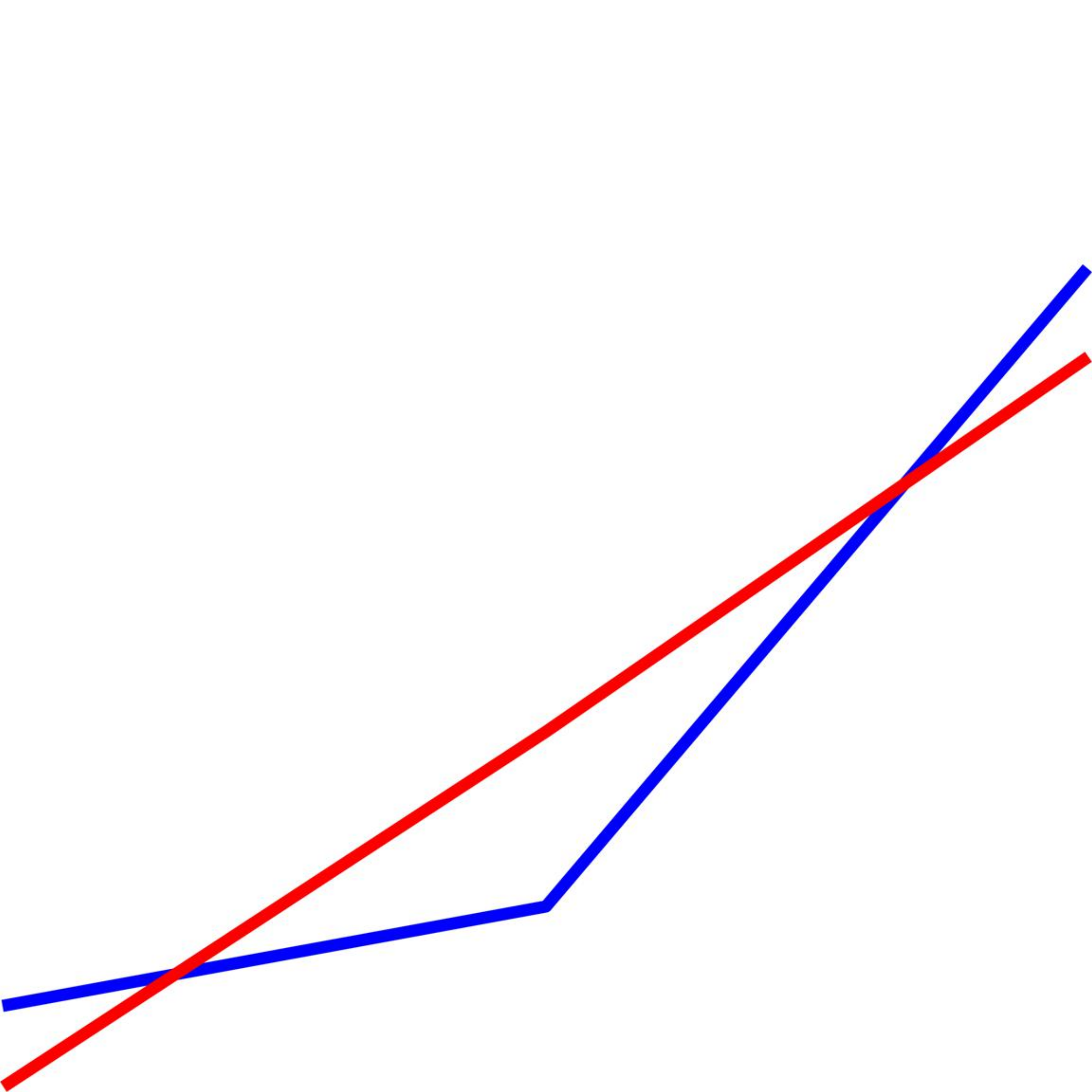}}\\
     \rowcolor{lightgray}
     \raisebox{-0.2\height}\gptlogo & \centering 1 & \textcolor{red}{\xmark} &  \centering 0 & \textcolor{ForestGreen}{\cmark}&  \centering 2 & \textcolor{red}{\xmark} &  \centering 2 & \textcolor{red}{\xmark} &  \centering 4 & \textcolor{red}{\xmark}&  \centering 1 & \textcolor{red}{\xmark}\\
     \raisebox{-0.2\height}\geminilogo & \centering 1 & \textcolor{red}{\xmark} & \centering 1 & \textcolor{red}{\xmark}&  \centering 1 & \textcolor{ForestGreen}{\cmark} &  \centering 2 & \textcolor{red}{\xmark} &  \centering 1 & \textcolor{red}{\xmark}&  \centering 1 & \textcolor{red}{\xmark}\\
     \rowcolor{lightgray}
     \raisebox{-0.2\height}\sonnetlogo & \centering 4 & \textcolor{red}{\xmark} & \centering 1 & \textcolor{red}{\xmark}&  \centering 2 & \textcolor{red}{\xmark} &  \centering 1 & \textcolor{ForestGreen}{\cmark} &  \centering 4 & \textcolor{Red}{\xmark}&  \centering 1 & \textcolor{red}{\xmark}\\
       \raisebox{-0.2\height}\newsonnetlogo & \centering 1 & \textcolor{red}{\xmark} & \centering 0 & \textcolor{ForestGreen}{\cmark}&  \centering 2 & \textcolor{red}{\xmark} &  \centering 1 & \textcolor{ForestGreen}{\cmark} &  \centering 3 & \textcolor{red}{\xmark}&  \centering 2 & \textcolor{ForestGreen}{\cmark}\\

     \end{tabular}

    \vspace{4pt}
    \centering
    \begin{tabular}{cccccccccccccc}
    \raisebox{-0.1\height}\gptlogo & \gpt &&&
      \raisebox{-0.1\height}\geminilogo & \gemini & &&\raisebox{-0.1\height}\sonnetlogo & \sonnet & && \raisebox{-0.12\height}\newsonnetlogo & \newsonnet \\
      \end{tabular}
     
\end{AIbox}
\caption{VLMs cannot reliably count the intersections between the \textcolor{blue}{blue} and \textcolor{red}{red} plots.}
\label{appfig:lineintersection_prompt2}
\end{figure}

\clearpage
\section{Counting the number of nested squares \nestedsquareslogo}
\label{app:nested_squares}

\subsection{Benchmark Construction Details}
\label{app:squares-bench}

We use 5 parameters to create the images of nested squares.

\begin{itemize}
    \item \textbf{Depth}: For each image, we draw $N\in\{2, 3, 4, 5\}$ nested squares on the image. We refer to each square in this collection as a depth.
    \item \textbf{Initial size}: We choose a random size for the first square in the bounds of the image size.
    \item \textbf{Reduction factor}: We draw squares such that each depth is entirely contained by its previous depth. We use a reduction factor to scale the square sizes.
    \item \textbf{Center}: The first square's center is chosen to ensure it is entirely visible in the image. For the remaining squares, we choose the center based on the space between the previous square and the new reduced size.  
    \item \textbf{Line thickness}: We use standard \emph{matplotlib} line width parameter of (2=3px, 3=4px, 4=6px).
\end{itemize}

\begin{figure}
    \centering
    \begin{tabular}{cccc}
         \includegraphics[width=0.18\textwidth]{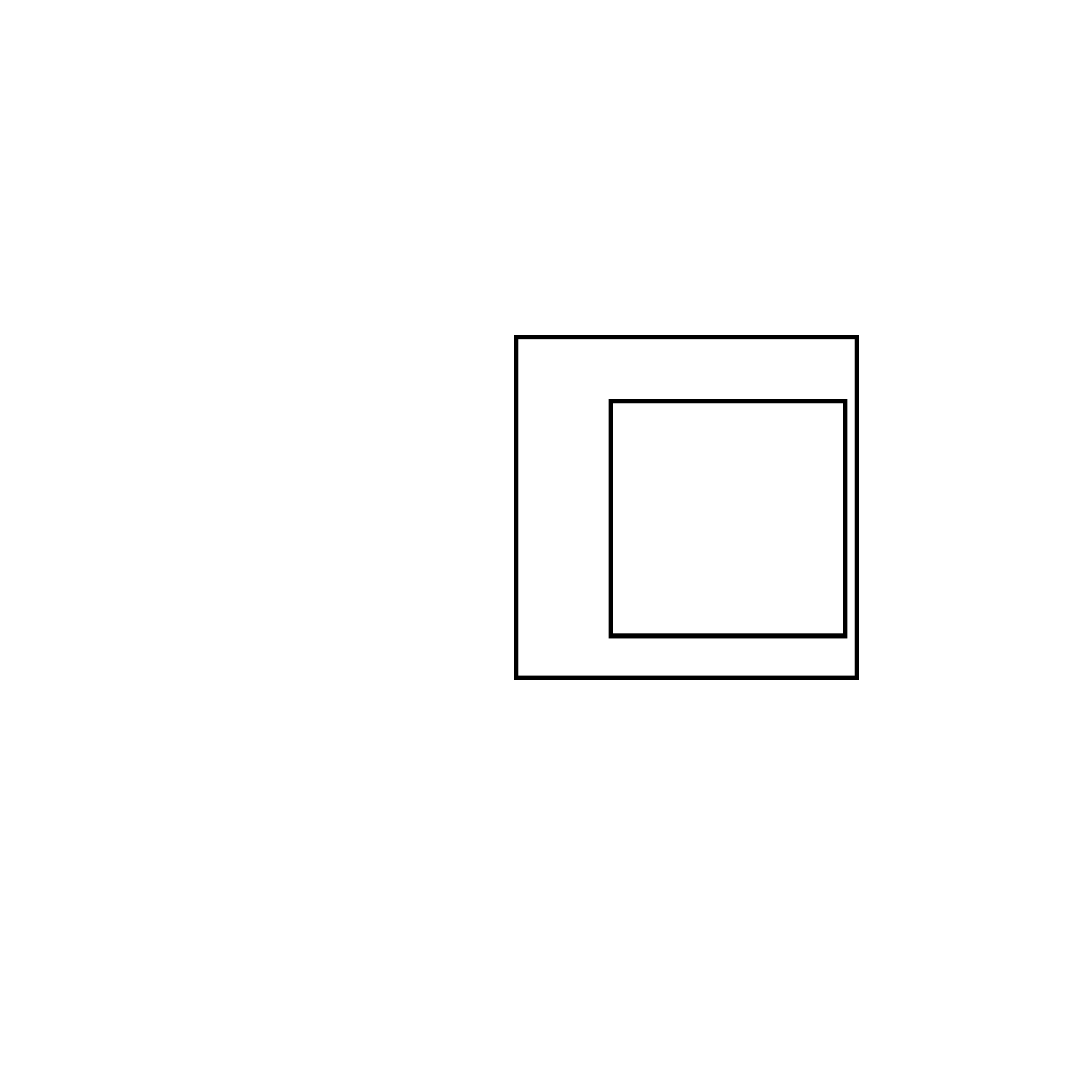}&\includegraphics[width=0.18\textwidth]{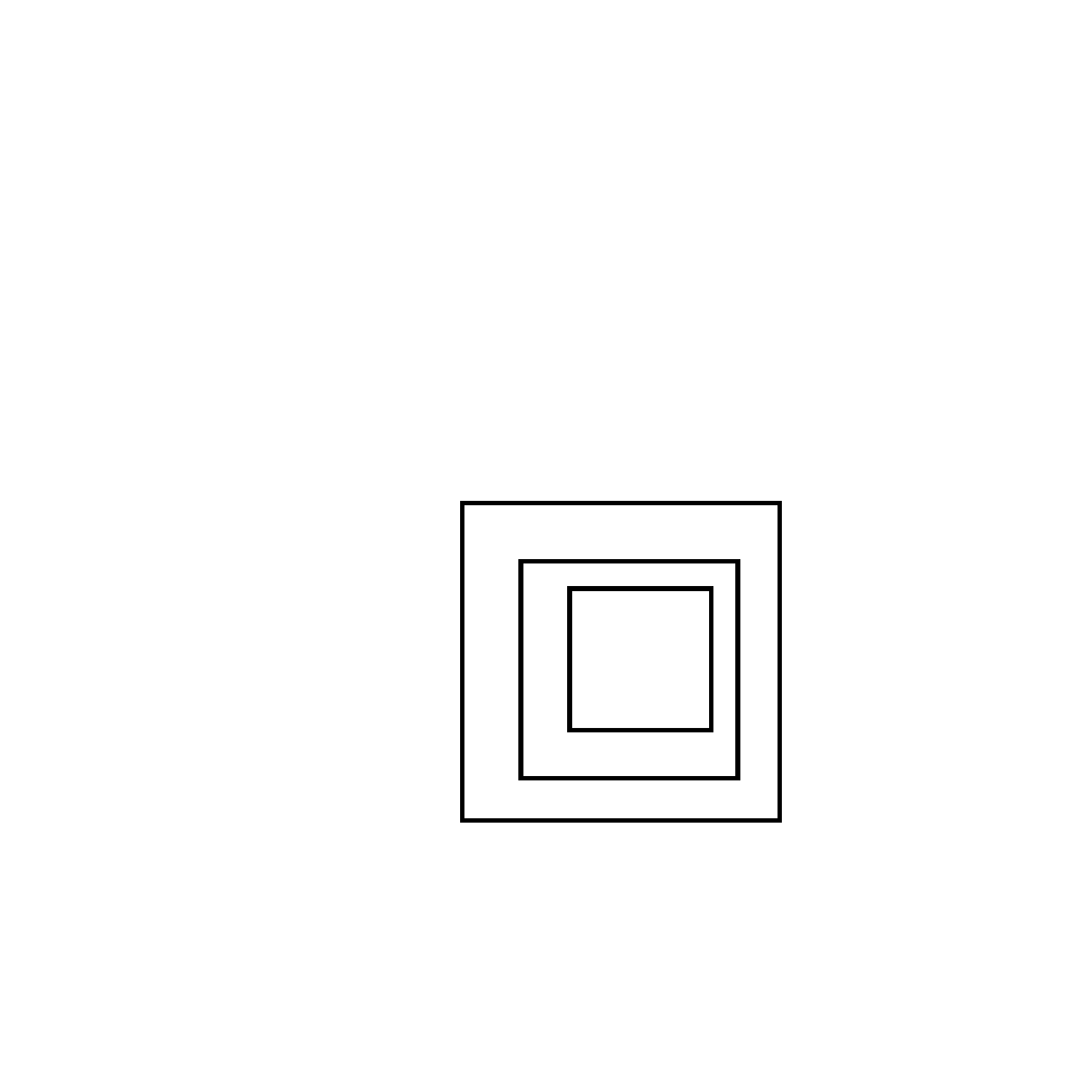}  &\includegraphics[width=0.18\textwidth]{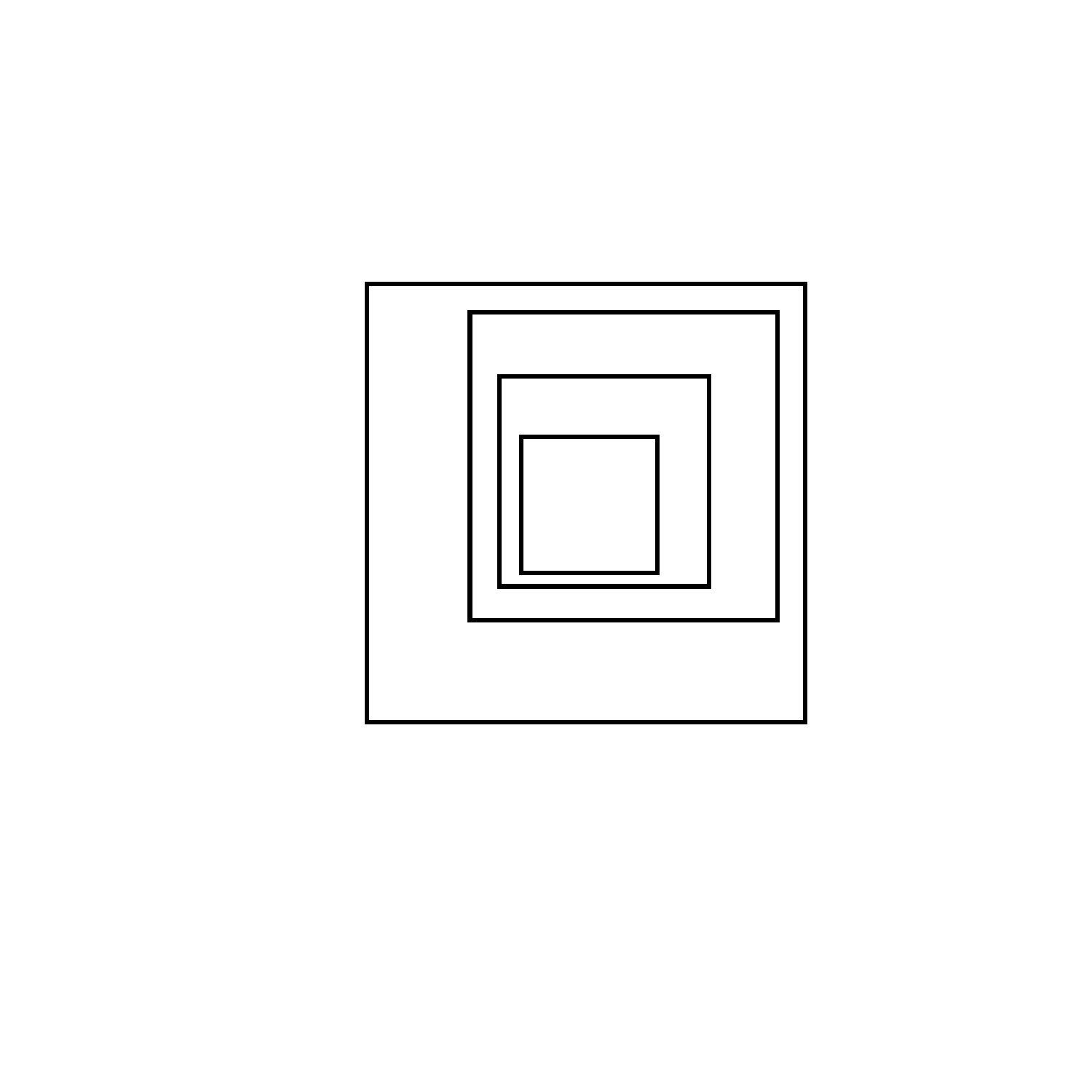}&\includegraphics[width=0.18\textwidth]{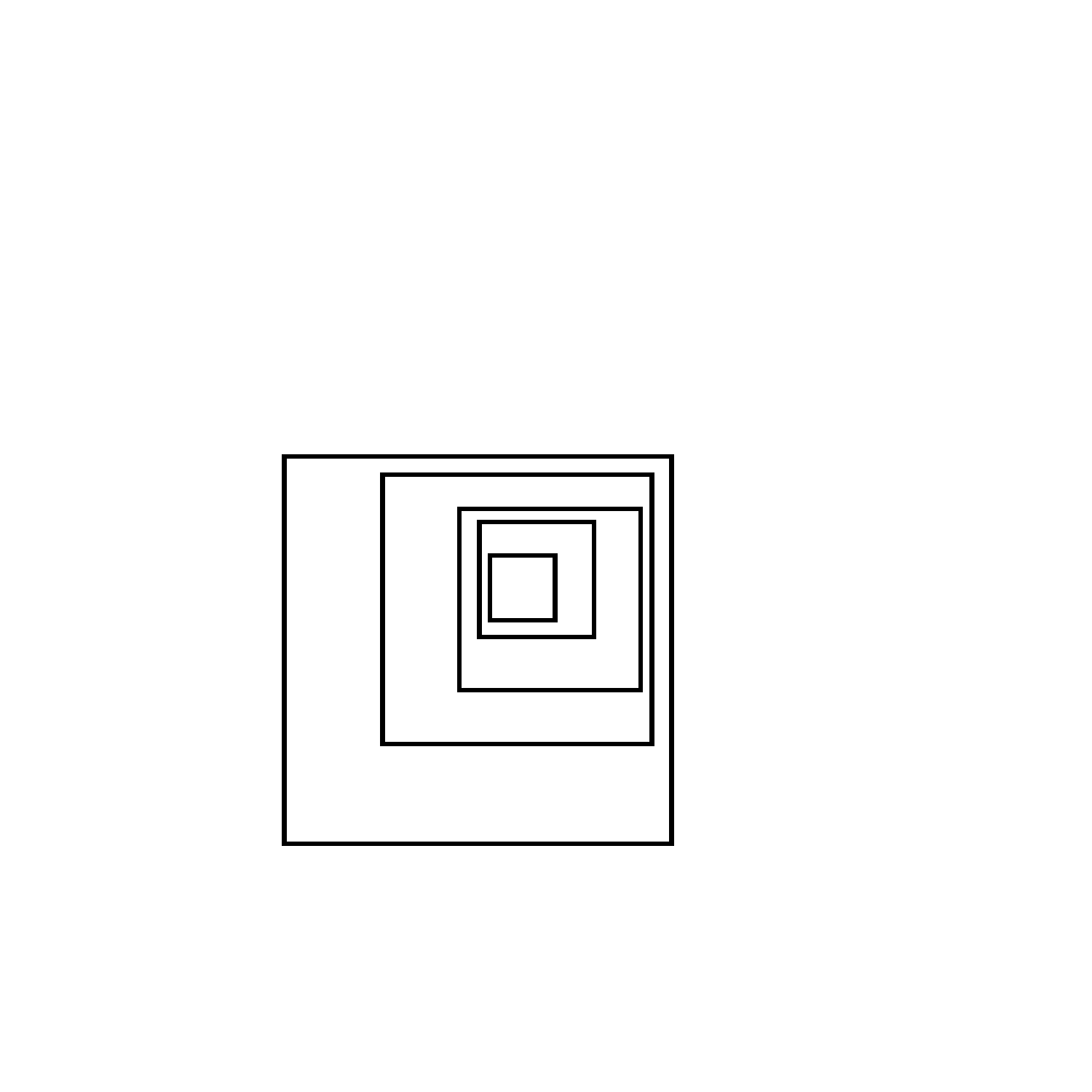}\\
         \includegraphics[width=0.18\textwidth]{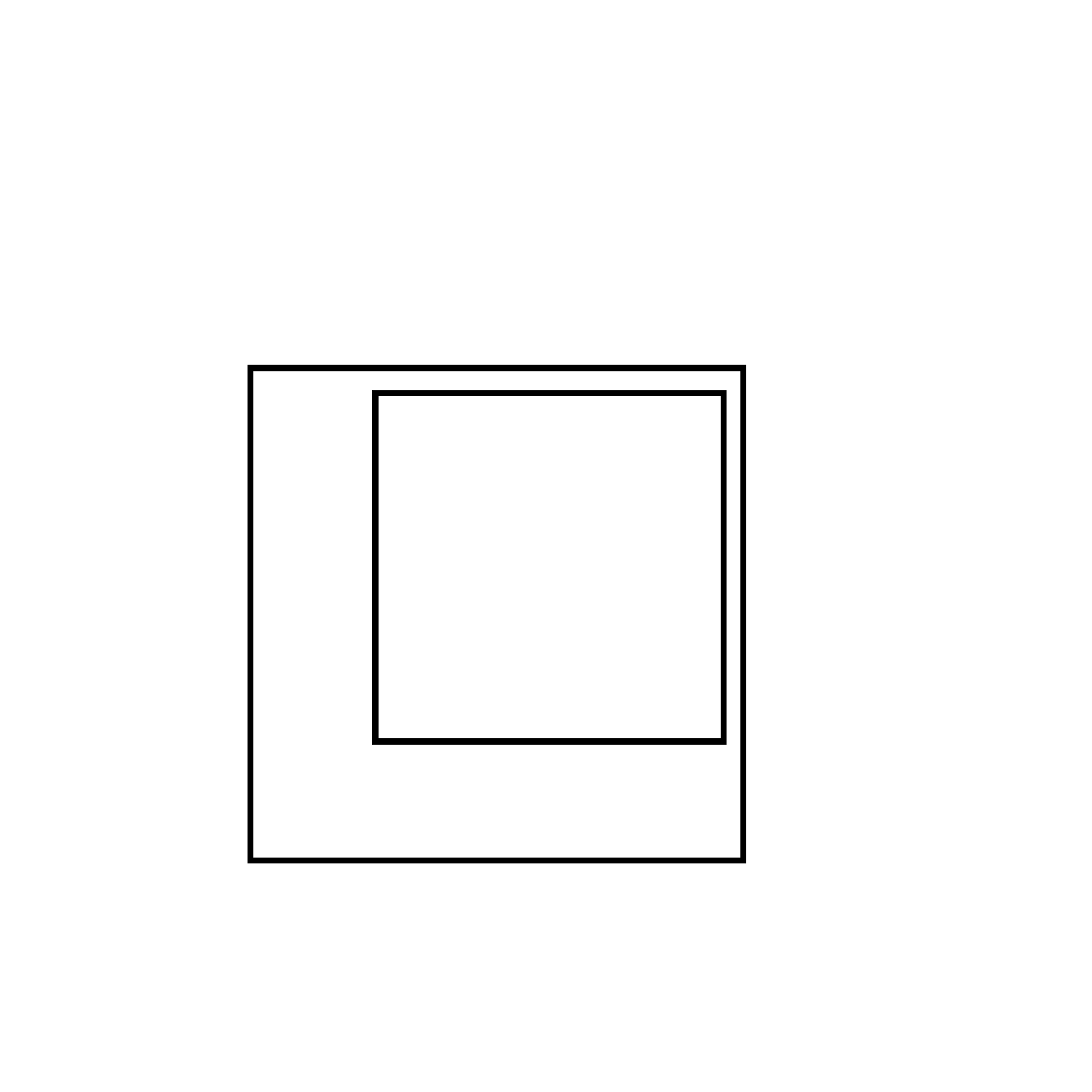}&\includegraphics[width=0.18\textwidth]{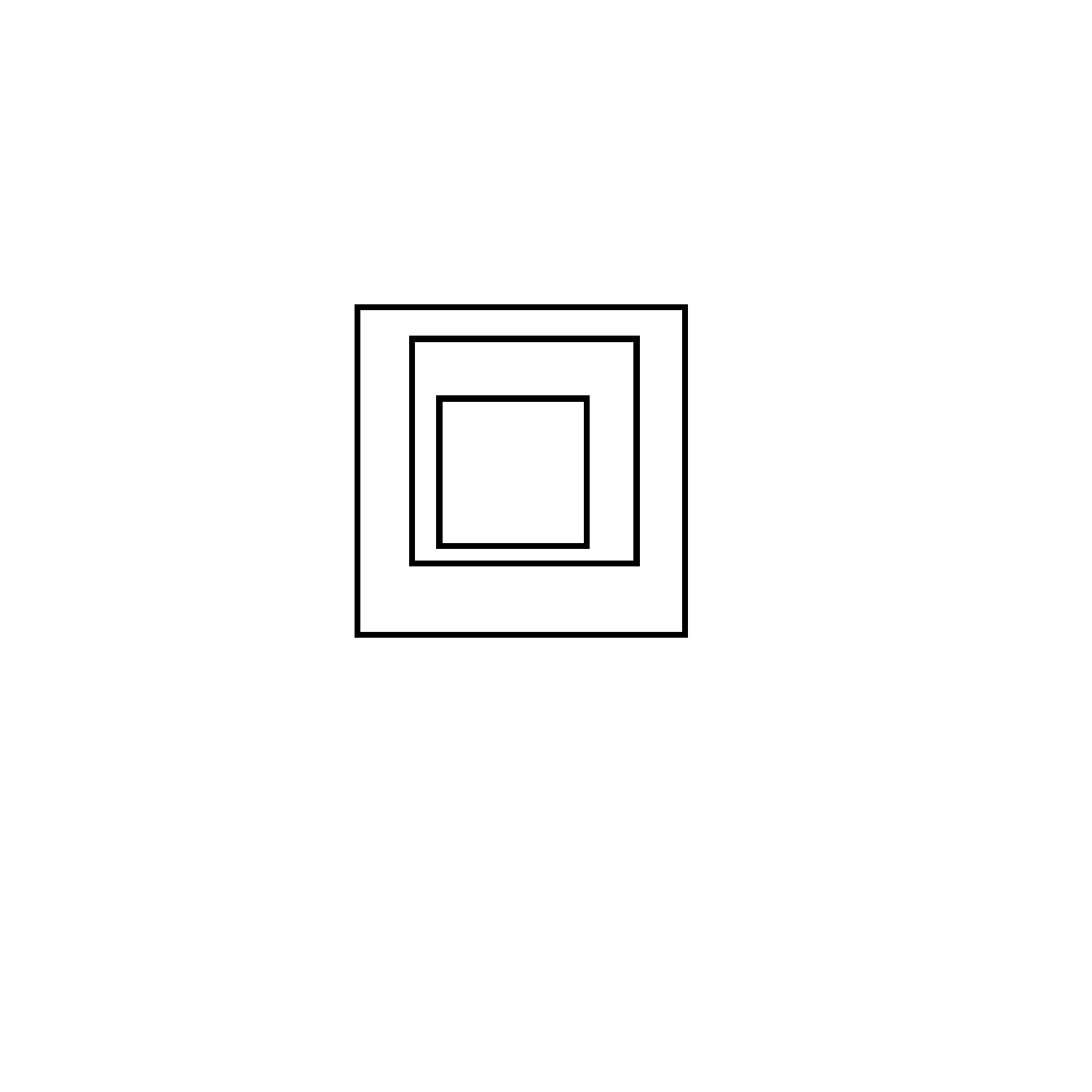}  &\includegraphics[width=0.18\textwidth]{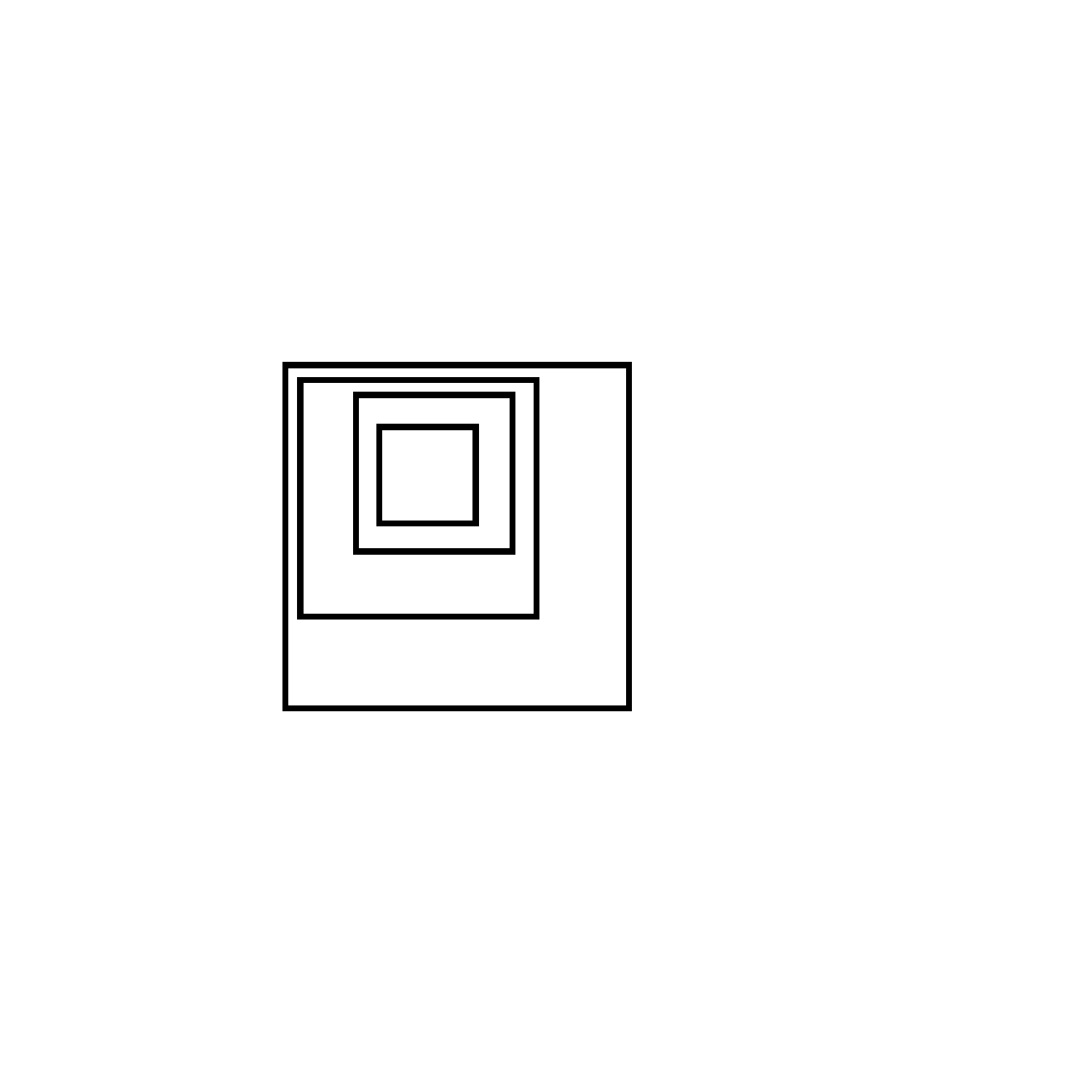}&\includegraphics[width=0.18\textwidth]{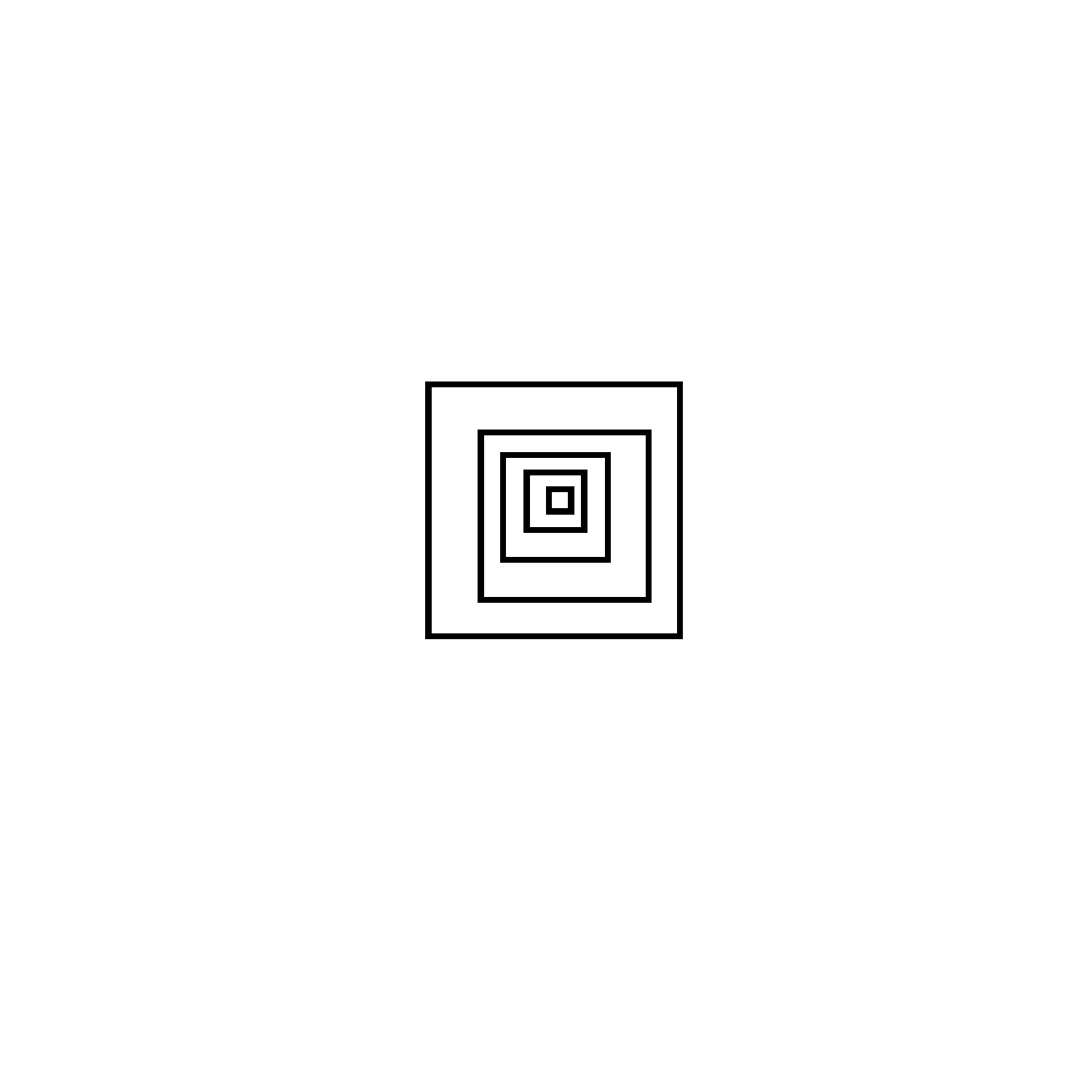}\\
         (a) 2 squares & (b) 3 squares  & (c) 4 squares & (d) 5 squares \\
    \end{tabular}
    \caption{Random examples from the nested square task that have 2, 3, 4 or 5 squares in the image.}
    \label{app:fig-nested-examples}
\end{figure}

We continue to generate images until we have 30 samples for each depth, resulting in 120 images overall (see \cref{table:squares-parameters,app:fig-nested-examples}).

\textbf{Code} The code is available at \url{https://github.com/anguyen8/vision-llms-are-blind/blob/main/src/NestedSquares/GenerateSamples.ipynb}.

\subsec{Post-processing:} To determine the model's answer, we use Python to extract the number from curly braces in the response e.g. \{3\}.

\subsec{Prompts}

\begin{enumerate}
    \item \emph{How many squares are in the image? Please answer with a number in curly brackets e.g., \{10\}.}
    
    \item \emph{Count total number of squares in the image. Answer with only the number in numerical format in curly brackets e.g. \{3\}.}

\end{enumerate}

\noindent\textbf{Groundtruth} answers are $\in \{2, 3, 4, 5\}$ (random-baseline accuracy: 25\%).

\begin{figure}
    \centering
        \begin{tabular}{c@{}c}
         GT = 2
         \includegraphics[trim={300 270 100 200},clip,width=0.4\textwidth]{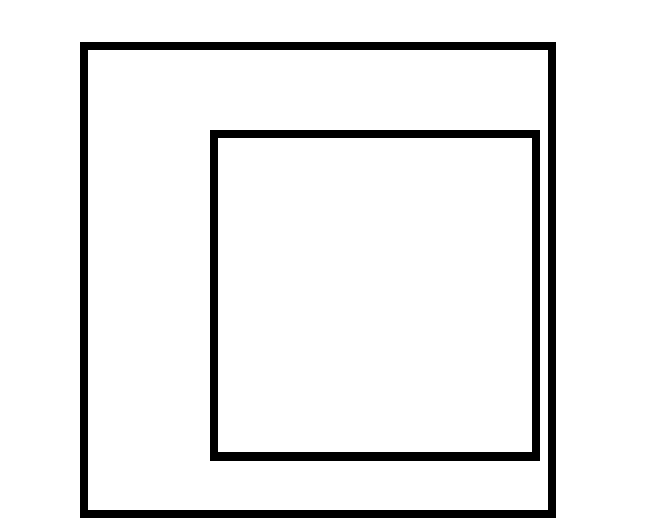}& GT = 3 \includegraphics[trim={220 300 200 200},clip,width=0.4\textwidth]{images/NestedSquares/benchmark_samples/gt_3_predicted_4_nested_squares_depth_3_image_10_thickness_4.pdf}\\
         GT = 4 \includegraphics[trim={200 240 120 170},clip,width=0.4\textwidth]{images/NestedSquares/benchmark_samples/gt_4_predicted_5_nested_squares_depth_4_image_7_thickness_3.pdf}& 
         GT = 5 \includegraphics[trim={170 160 200 300},clip,width=0.4\textwidth]{images/NestedSquares/benchmark_samples/gt_5_predicted_1_nested_squares_depth_5_image_4_thickness_3.pdf} \\
    \end{tabular}
    \caption{Count squares}
    \label{fig:nested_sq}
\end{figure}


\begin{table}
\centering
\caption{Nested squares include the same number of samples for each parameter value in our benchmark.}
\label{table:squares-parameters}
\begin{tabular}{lccc}
\toprule
\textbf{Parameter} & \textbf{Values} & ~~\textbf{Samples} & ~~\textbf{Total Samples}\\
\midrule
\multirow{1}{*}{Image size} & 1000px & ~~120 & ~~\multirow{1}{*}{120}\\
\midrule
\multirow{3}{*}{Line thickness} & 3px & ~~40 & ~~\multirow{3}{*}{120}\\
 & 4px & ~~40 \\
 & 6px & ~~40 \\
\midrule
\multirow{4}{*}{Number of Squares} & 2 & ~~30 & ~~\multirow{4}{*}{120}\\
 & 3 & ~~30 \\
 & 4 & ~~30 \\
 & 5 & ~~30 \\
\bottomrule
\end{tabular}
\end{table}

\clearpage
\subsection{Finding: the best-performing model is affected by line width}
\label{app:square-thickness}
\cref{tab:nested_squares_depth} shows that the best VLM on the nested square task (\newsonnet) is more accurate as line width increases. In contrast, the worst model (\gpt) shows the opposite trend as the line thickness changes, suggesting that \gpt confuses squares when the borderline is thick.

\begin{table}
\centering
\caption{Line thickness has minimal effect on VLMs' performance, suggesting that visual attributes of shapes are not critical to VLMs when asked to count the shapes.}
\label{tab:nested_squares_depth}
\begin{tabular}{lcccc}
\toprule
Line width & ~~\gptlogo  & ~~\geminilogo  & ~~\sonnetlogo  & ~~\newsonnetlogo\\
\midrule
2 & ~~58.75 & ~~86.25 & ~~61.25 & ~~90.00\\
3 & ~~56.25 & ~~85.00 & ~~67.50 & ~~91.25 \\
4 & ~~52.50 & ~~90.00 & ~~66.25 & ~~95.00 \\ \midrule
\textbf{Average} & ~~55.83 & ~~87.08 & ~~65.00 & ~~92.08 \\
\bottomrule
\end{tabular}
\end{table}

\clearpage
\subsection{Additional Examples}
\begin{figure}
\centering
\begin{AIbox}{Count total number of squares in the image. Answer with only the number in numerical format in curly brackets e.g. \{3\}.}
\centering

    \begin{tabular}{lp{0.8cm}c|p{0.8cm}c|p{0.8cm}c|p{0.8cm}c|p{0.8cm}c|p{0.8cm}c}
     &\multicolumn{2}{c}{\includegraphics[trim={130 100 200 230},clip,width=0.15\textwidth]{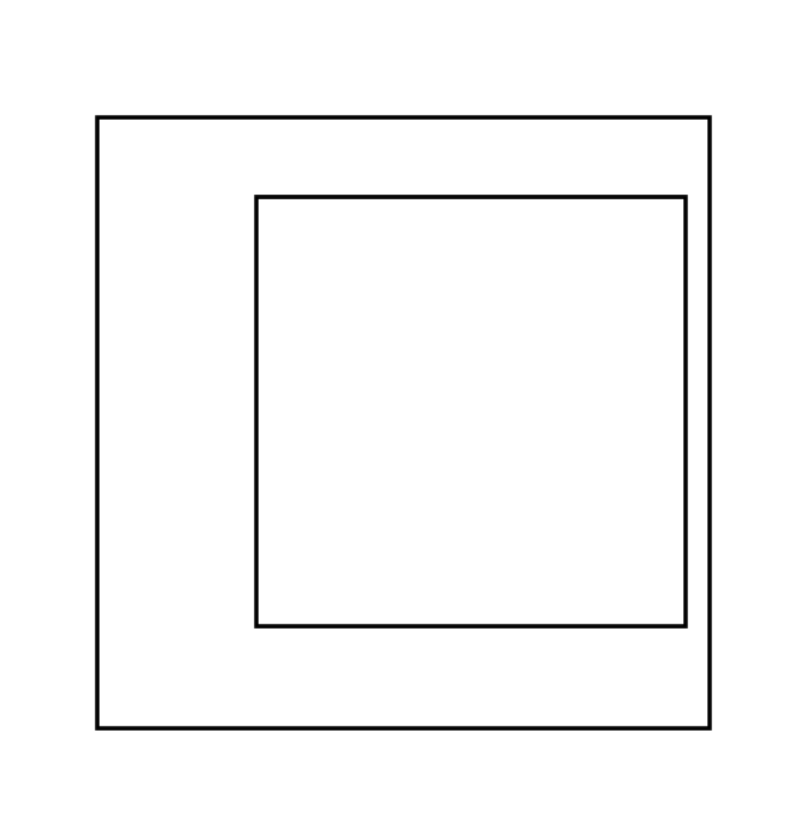}} &
     \multicolumn{2}{c}{\includegraphics[trim={150 120 150 150},clip,width=0.15\textwidth]{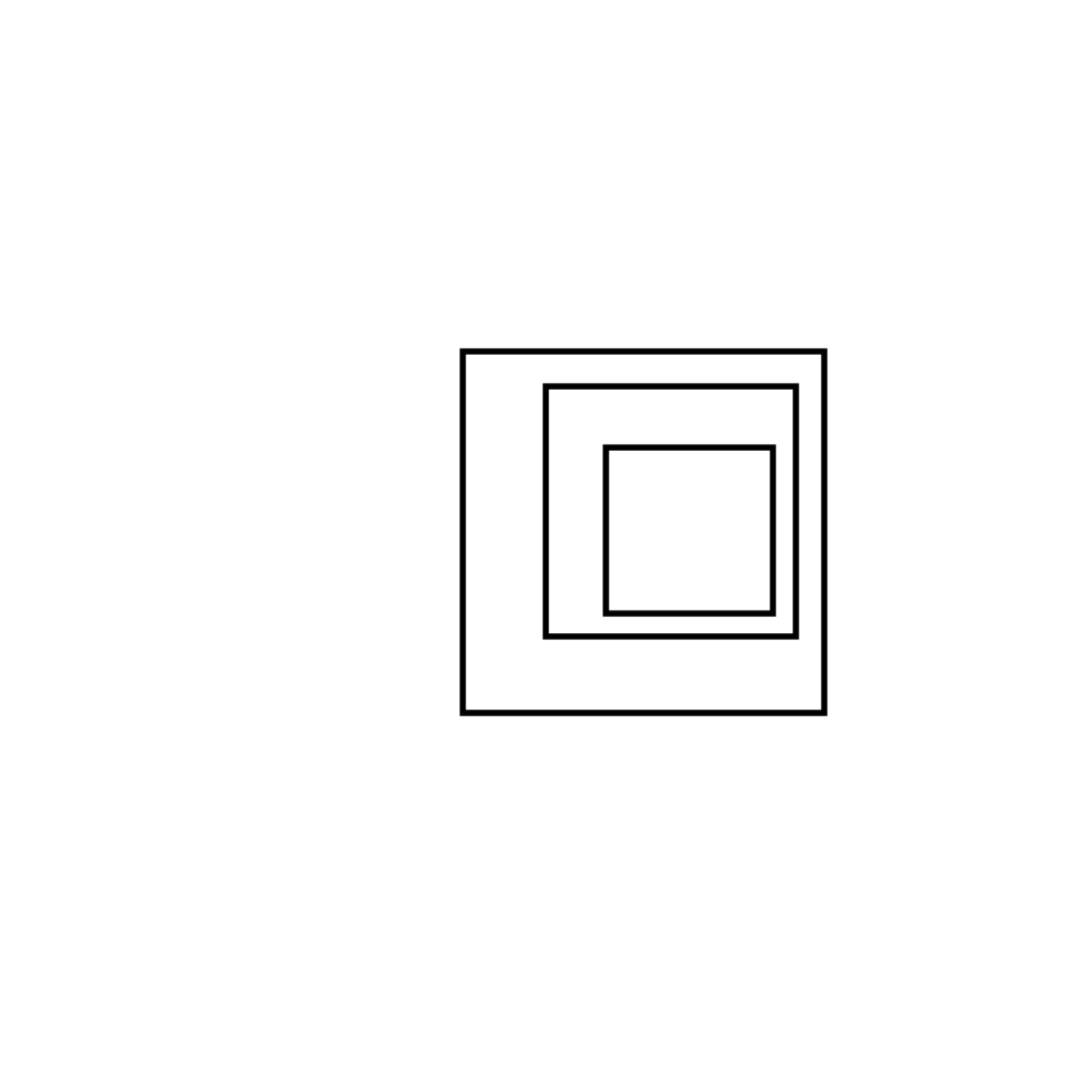}} &
     \multicolumn{2}{c}{\includegraphics[trim={130 100 200 230},clip,width=0.15\textwidth]{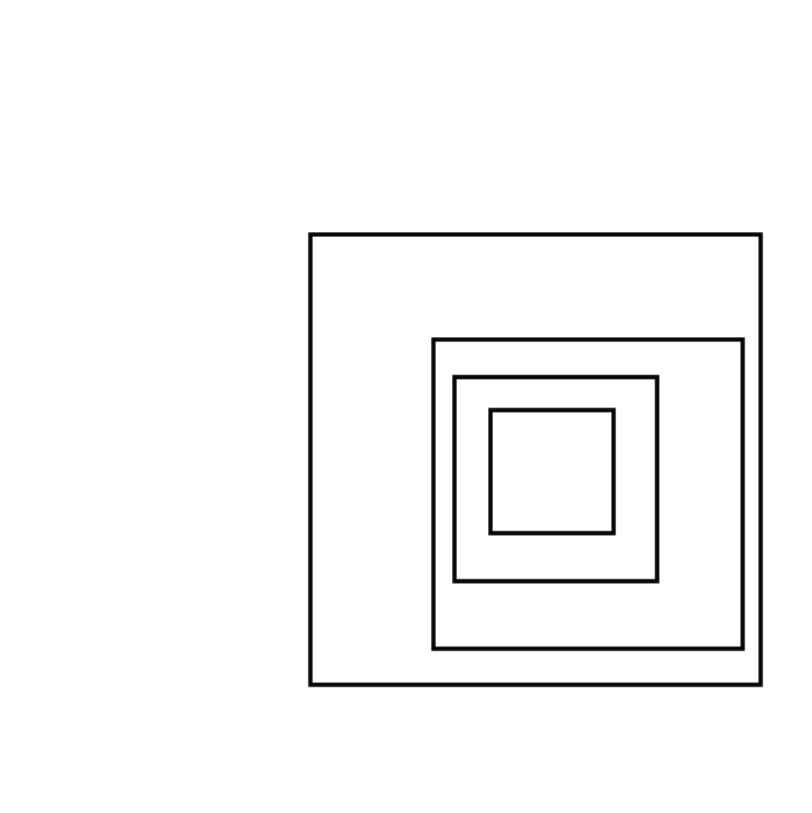}} &
     \multicolumn{2}{c}{\includegraphics[trim={130 200 200 130},clip,width=0.15\textwidth]{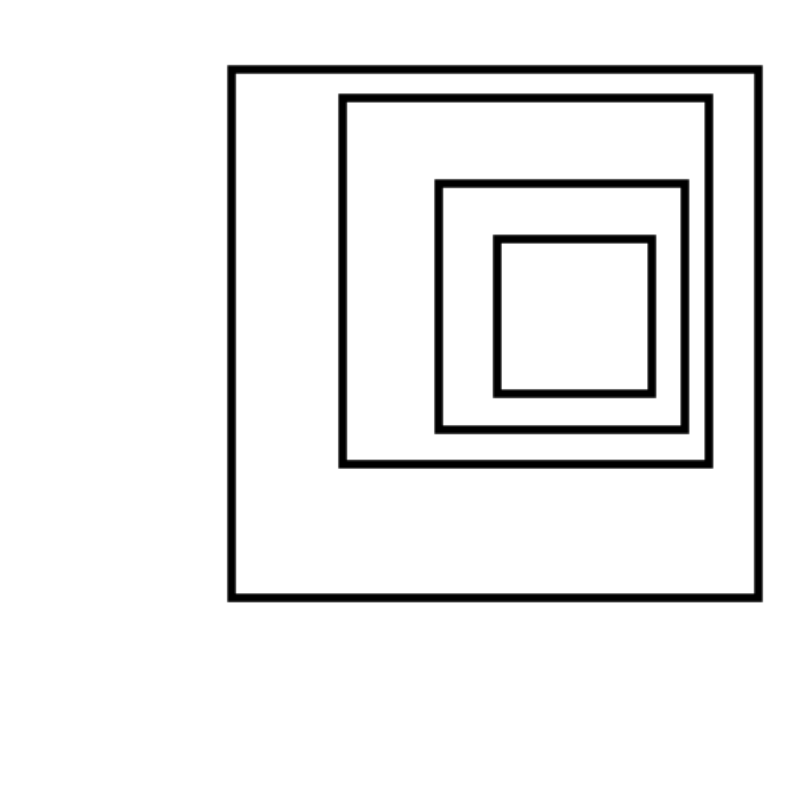}} &
     \multicolumn{2}{c}{\includegraphics[trim={230 200 200 230},clip,width=0.15\textwidth]{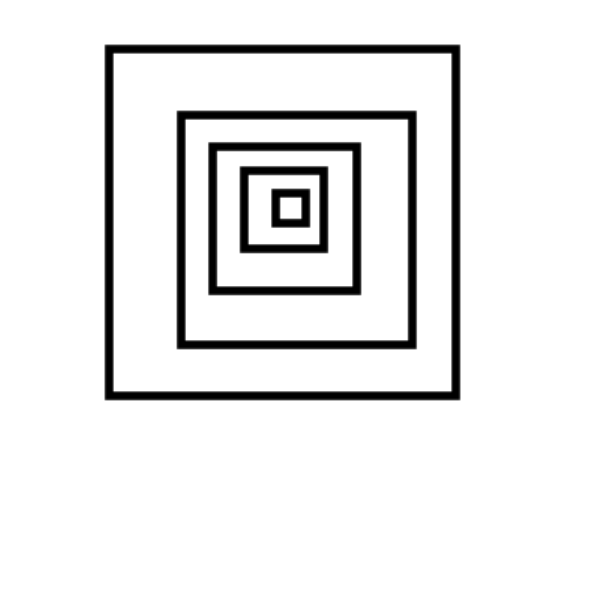}}&
     \multicolumn{2}{c}{\includegraphics[trim={200 150 200 230},clip,width=0.15\textwidth]{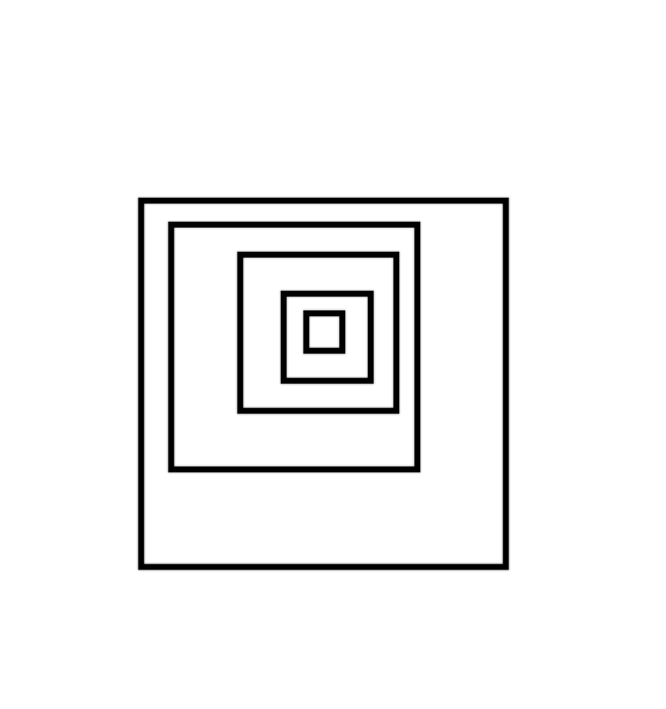}}\\
     \rowcolor{lightgray}
     \raisebox{-0.2\height}\gptlogo & 
     \centering 2 & \textcolor{ForestGreen}{\cmark} &  
     \centering 5 & \textcolor{red}{\xmark}&  
     \centering 5 & \textcolor{red}{\xmark} &  
     \centering 5 & \textcolor{red}{\xmark} &  
     \centering 5 & \textcolor{ForestGreen}{\cmark}&  
     \centering 5 & \textcolor{ForestGreen}{\cmark}\\
     
     \raisebox{-0.2\height}\geminilogo & 
     \centering 2 & \textcolor{ForestGreen}{\cmark} & 
     \centering 3 & \textcolor{ForestGreen}{\cmark}&  
     \centering 5& \textcolor{red}{\xmark} &  
     \centering 5 & \textcolor{red}{\xmark} &  
     \centering 5 & \textcolor{ForestGreen}{\cmark}&  
     \centering 5 & \textcolor{ForestGreen}{\cmark}\\
     
     \rowcolor{lightgray}
     \raisebox{-0.2\height}\sonnetlogo & 
     \centering 5 & \textcolor{red}{\xmark} & 
     \centering 4 & \textcolor{red}{\xmark}&  
     \centering 5 & \textcolor{red}{\xmark} &  
     \centering 5 & \textcolor{red}{\xmark} &  
     \centering 5 & \textcolor{ForestGreen}{\cmark}&  
     \centering 4 & \textcolor{red}{\xmark}\\
     
      \raisebox{-0.2\height}\newsonnetlogo & 
      \centering 2 & \textcolor{ForestGreen}{\cmark} & 
      \centering 3 & \textcolor{ForestGreen}{\cmark}&  
      \centering 4 & \textcolor{ForestGreen}{\cmark} &  
      \centering 4 & \textcolor{ForestGreen}{\cmark} &  
      \centering 4 & \textcolor{red}{\xmark}&  
      \centering 4 & \textcolor{red}{\xmark}\\
     \end{tabular}

    \vspace{4pt}
    \centering
    \begin{tabular}{cccccccccccccc}
    \raisebox{-0.1\height}\gptlogo & \gpt &&&
      \raisebox{-0.1\height}\geminilogo & \gemini & &&\raisebox{-0.1\height}\sonnetlogo & \sonnet & && \raisebox{-0.12\height}\newsonnetlogo & \newsonnet \\
      \end{tabular}
     
\end{AIbox}
\caption{VLMs fail to reliably count the number of nested squares in an image.}
\label{appfig:nested_squares}
\end{figure}

We show examples of models' responses to the counting the number of nested squares task in \cref{appfig:nested_squares}.
\label{app:square-qual}

\begin{figure}[H]
\centering
\begin{AIbox}{How many squares are in the image? Please answer with a number in curly brackets e.g.{,} \{10\}.}
\vspace{-.3cm}
\hspace{-0.039\textwidth}
    \begin{tabular}{lp{0.5cm}c|p{0.5cm}c|p{0.5cm}c|p{0.5cm}c|p{0.5cm}c|p{0.5cm}c|p{0.5cm}c|p{0.5cm}c}
     &\multicolumn{2}{c}{\includegraphics[trim={150 120 220 230},clip,width=0.12\textwidth]{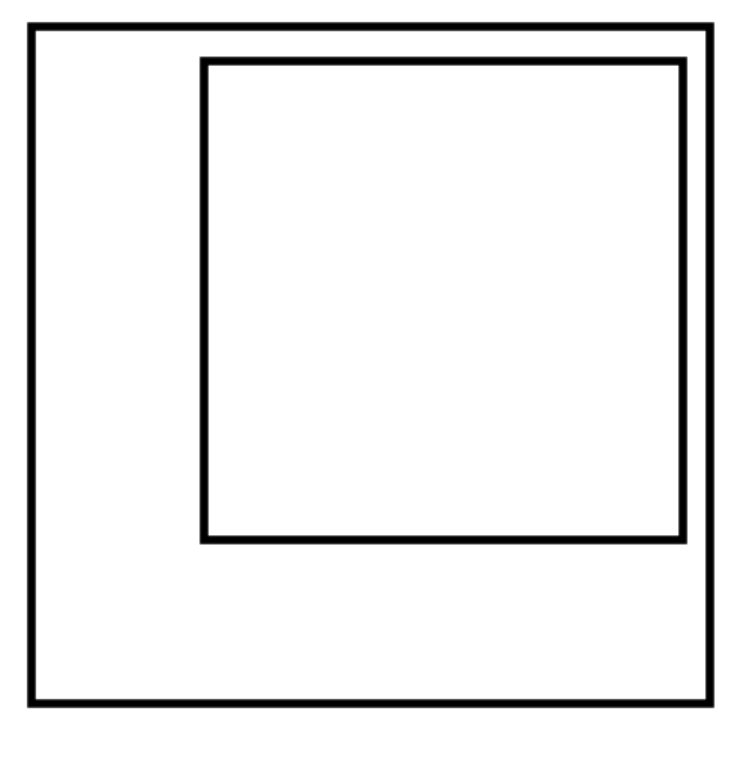}} &
     \multicolumn{2}{c}{\includegraphics[trim={250 210 120 180},clip,width=0.12\textwidth]{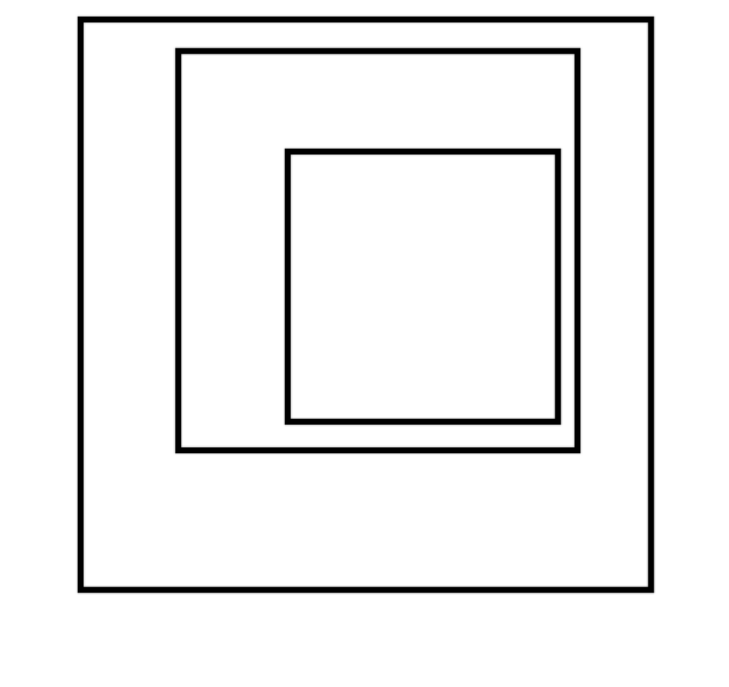}} &
     \multicolumn{2}{c}{\includegraphics[trim={170 290 240 140},clip,width=0.12\textwidth]{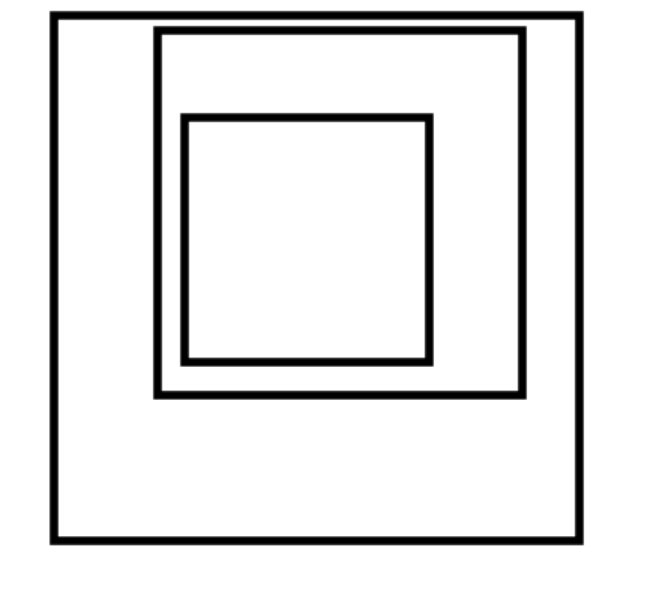}} &
     \multicolumn{2}{c}{\includegraphics[trim={200 220 240 220},clip,width=0.12\textwidth]{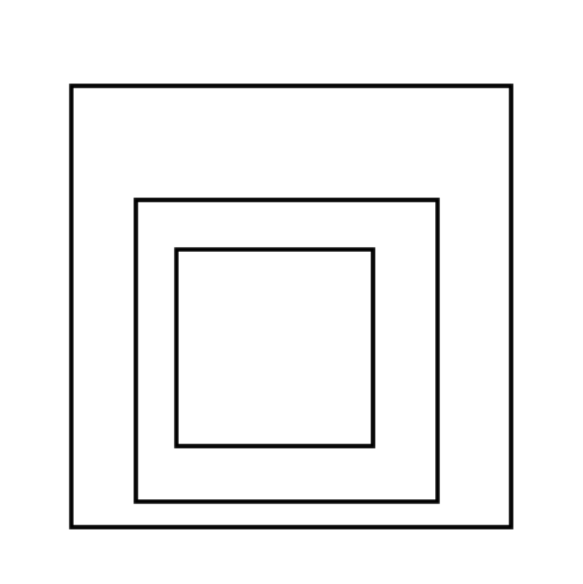}} &
     \multicolumn{2}{c}{\includegraphics[trim={220 150 200 300},clip,width=0.12\textwidth]{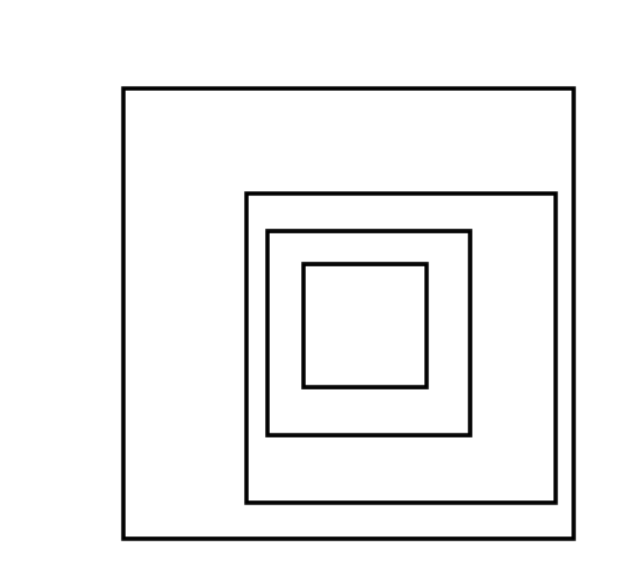}}&
     \multicolumn{2}{c}{\includegraphics[trim={200 250 200 150},clip,width=0.12\textwidth]{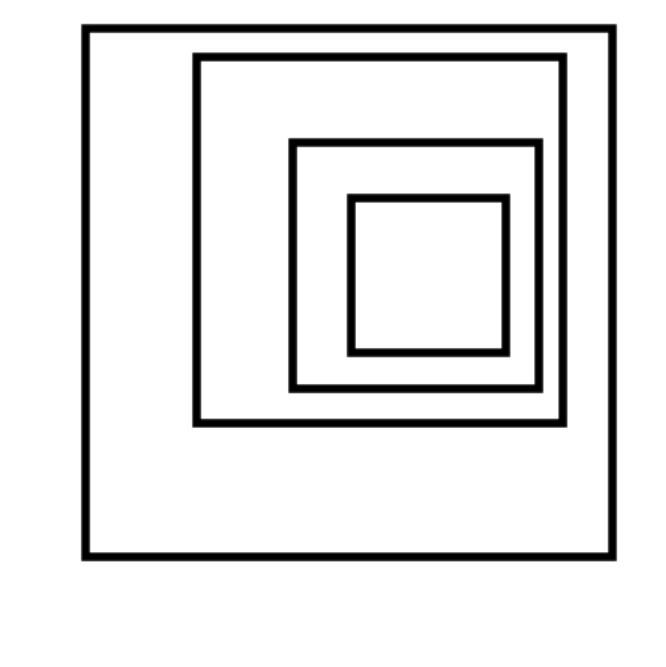}}&
     \multicolumn{2}{c}{\includegraphics[trim={250 200 250 290},clip,width=0.12\textwidth]{images/NestedSquares/updated_samples/howmany/nested_squares_depth_5_image_7_thickness_3.pdf}}&
     \multicolumn{2}{c}{\includegraphics[trim={300 280 220 250},clip,width=0.12\textwidth]{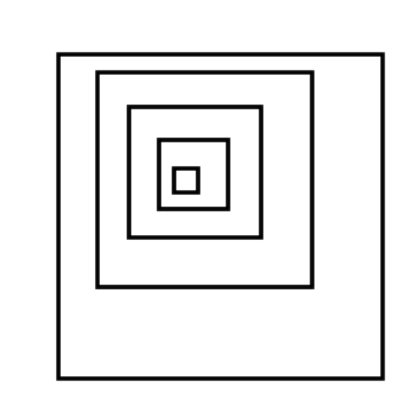}}\\
     \rowcolor{lightgray}
     \raisebox{-0.2\height}\gptlogo & 
     \centering 3 & \textcolor{red}{\xmark} &  
     \centering 5 & \textcolor{red}{\xmark}&  
     \centering 3 & \textcolor{ForestGreen}{\cmark} &  
     \centering 5 & \textcolor{red}{\xmark} &  
     \centering 5 & \textcolor{red}{\xmark}&  
     \centering 5 & \textcolor{red}{\xmark}&  
     \centering 6 & \textcolor{red}{\xmark}&  
     \centering 6 & \textcolor{red}{\xmark}\\

     \raisebox{-0.2\height}\geminilogo 
     & \centering 2 & \textcolor{ForestGreen}{\cmark} 
     & \centering 3 & \textcolor{ForestGreen}{\cmark}
     &  \centering 2 & \textcolor{red}{\xmark} 
     &  \centering 3 & \textcolor{ForestGreen}{\cmark} 
     &  \centering 5 & \textcolor{red}{\xmark}
     &  \centering 5 & \textcolor{red}{\xmark}
     &  \centering 5 & \textcolor{ForestGreen}{\cmark}
     &  \centering 4 & \textcolor{red}{\xmark}\\
     
     \rowcolor{lightgray}
     \raisebox{-0.2\height}\sonnetlogo & 
     \centering 2 & \textcolor{ForestGreen}{\cmark} & 
     \centering 4 & \textcolor{red}{\xmark}&  
     \centering 2 & \textcolor{red}{\xmark} &  
     \centering 4 & \textcolor{red}{\xmark} &  
     \centering 5 & \textcolor{red}{\xmark}&  
     \centering 4 & \textcolor{ForestGreen}{\cmark}&  
     \centering 4 & \textcolor{red}{\xmark}&  
     \centering 5 & \textcolor{ForestGreen}{\cmark}\\

      \raisebox{-0.2\height}\newsonnetlogo & 
      \centering 2 & \textcolor{ForestGreen}{\cmark} & 
      \centering 3 & \textcolor{ForestGreen}{\cmark}&  
      \centering 3 & \textcolor{ForestGreen}{\cmark} &  
      \centering 3 & \textcolor{ForestGreen}{\cmark} &  
      \centering 4 & \textcolor{ForestGreen}{\cmark}&  
      \centering 4 & \textcolor{ForestGreen}{\cmark}&  
      \centering 4 & \textcolor{red}{\xmark}&  
      \centering 5 & \textcolor{ForestGreen}{\cmark}\\
     \end{tabular}

    \vspace{4pt}
    \centering
    \begin{tabular}{cccccccccccccc}
    \raisebox{-0.1\height}\gptlogo & \gpt &&&
      \raisebox{-0.1\height}\geminilogo & \gemini & &&\raisebox{-0.1\height}\sonnetlogo & \sonnet & && \raisebox{-0.12\height}\newsonnetlogo & \newsonnet \\
      \end{tabular}
     
\end{AIbox}
\caption{
Counting nested squares is not easy for VLMs even when there are only two squares (leftmost).
The task becomes harder as the count increases from 2 to 5.
\newsonnet performs the best (92.08\%) but still not at the 100\% by humans.
}
\label{fig:nested_squares}
\end{figure}

\clearpage
\section{Counting the shapes in an Olympic-like logo}
\label{app:olympic}

\subsection{Benchmark Construction Details}
\label{app:olympic-bench}

We create the benchmark by generating images containing shapes resembling the Olympic logo by choosing a combination of settings.

\begin{itemize}
    \item \textbf{Image size}: We fix the physical size of the image in \emph{matplotlib} to 5$\times$5, and change the resolution by changing the DPI value, which is $\in\{100, 200, 300\}$ to get images with sizes $\{384, 769, 1155\}$px.
    \item \textbf{Number of shapes}: We choose a number from \{5, 6, 7, 8, 9\}.
    \item \textbf{Color}: Each image is generated using two different coloring schemes. We generate an all-black version and a second version by randomly sampling colors from a colormap in \emph{matplotlib}.
    \item \textbf{Distance}: To generate the interlaced shapes, we use a small boundary-to-boundary distance factor for each row of the shapes. We fix this value to 0.1 proportional to the diameter of circles or side length of pentagons. 
    \item \textbf{Diameter}: We choose a uniform diameter for all the circles in each image from \(\{\frac{1}{7}, \frac{1}{10}\}\) proportional to the image size.
    \item \textbf{Side length}: We follow the same policy for the diameter to choose the side length of the pentagons.
    \item \textbf{Line thickness}: We generate each image with \{0.5, 1.0\} line width of \emph{matplotlib} standard. This results in \{1px, 2px\} lines in 384px images, \{2px, 3px\} lines in 769px images and \{3px, 5px\} lines in 1155px images.
     
\end{itemize}

\begin{figure}
\centering
\begin{tabular}{c@{}cccc}
\multicolumn{2}{c}{(a) Number of circles} & (b) Linewidth & (c) Color & (d) Pentagons\\
\includegraphics[width=0.15\textwidth]{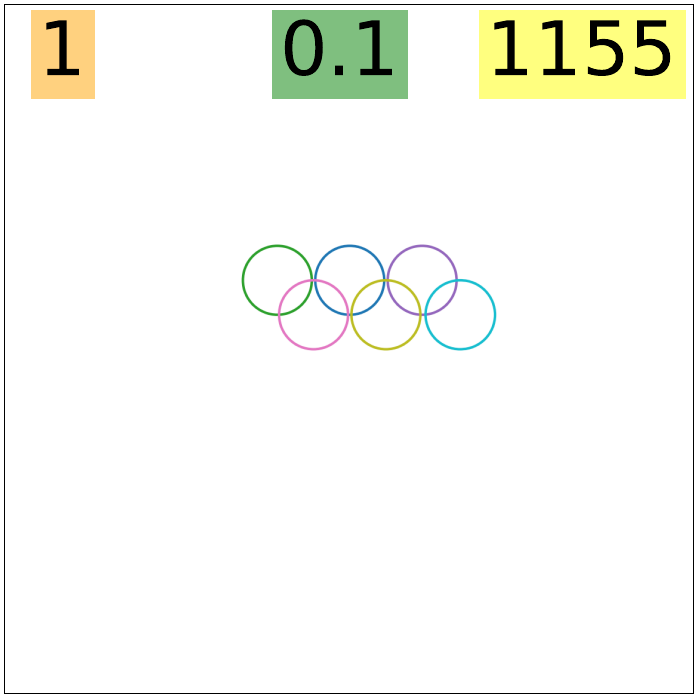} & \includegraphics[width=0.15\textwidth]{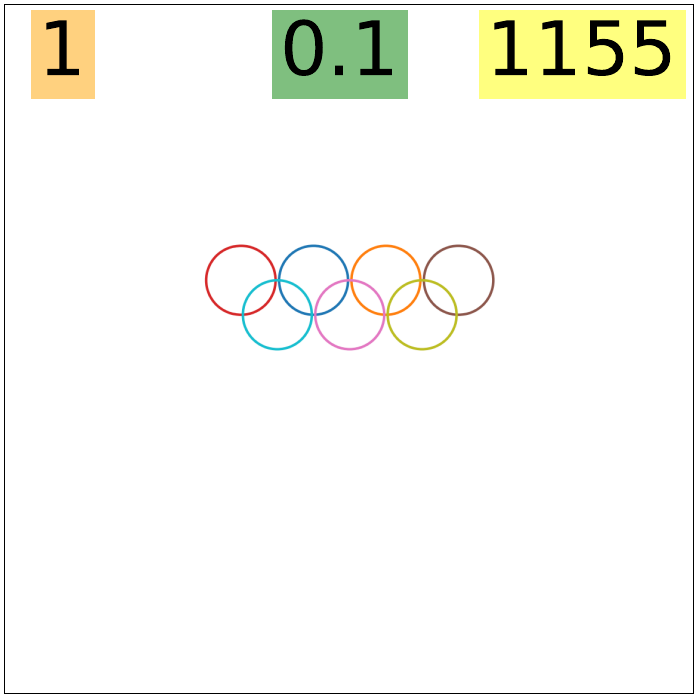} & \includegraphics[width=0.15\textwidth]{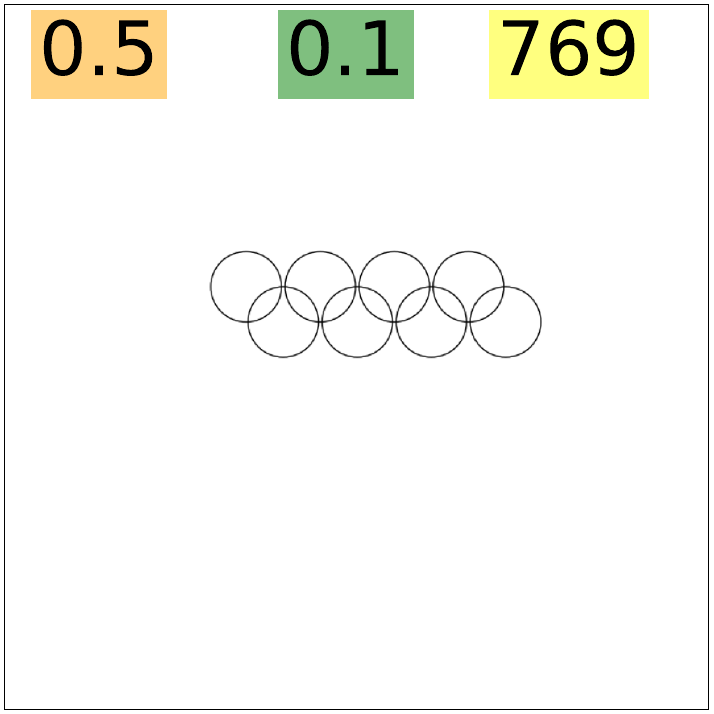} & \includegraphics[width=0.15\textwidth]{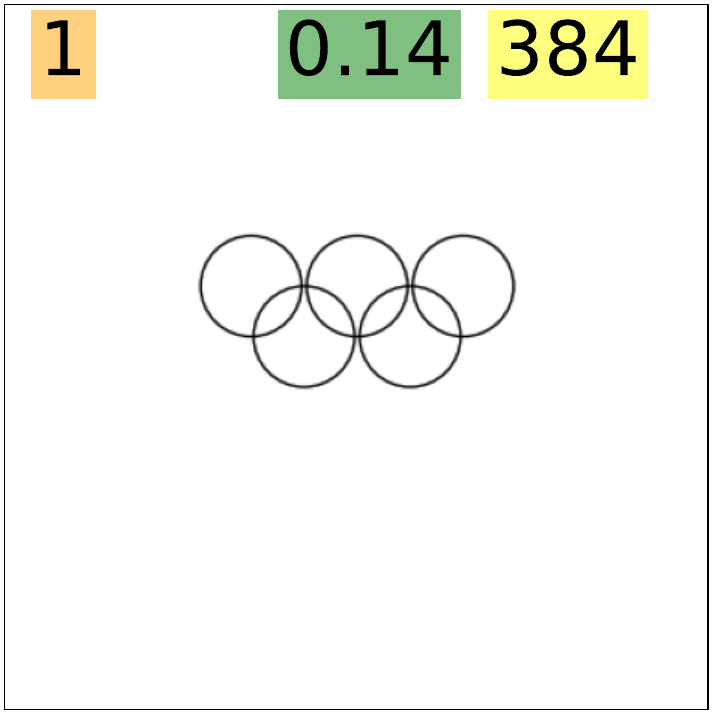} & \includegraphics[width=0.15\textwidth]{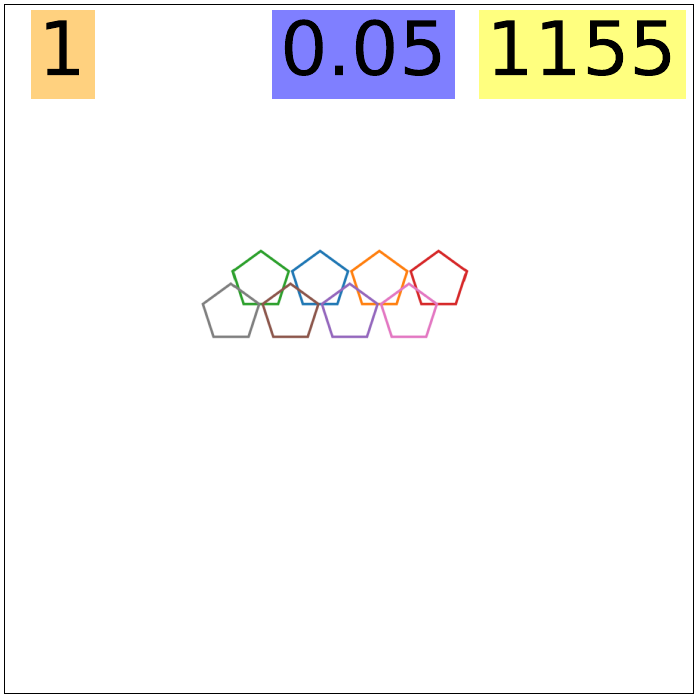}\\
\includegraphics[width=0.15\textwidth]{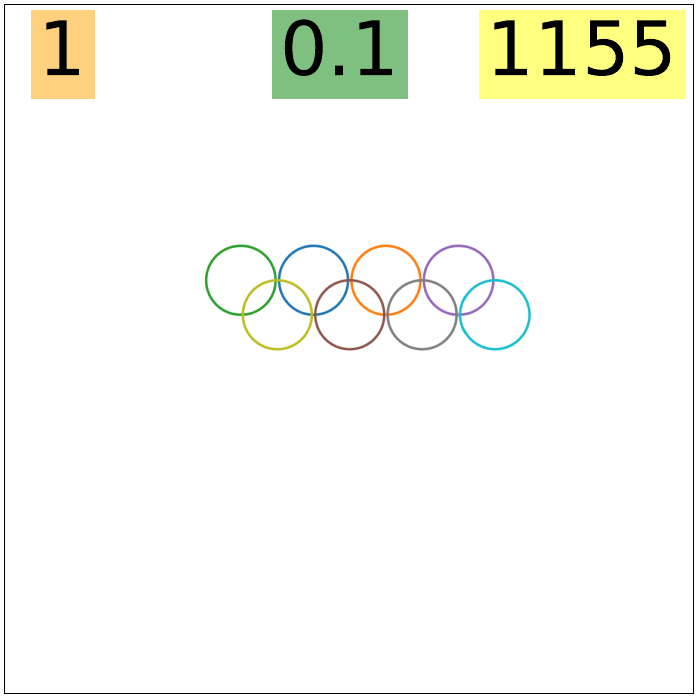} & \includegraphics[width=0.15\textwidth]{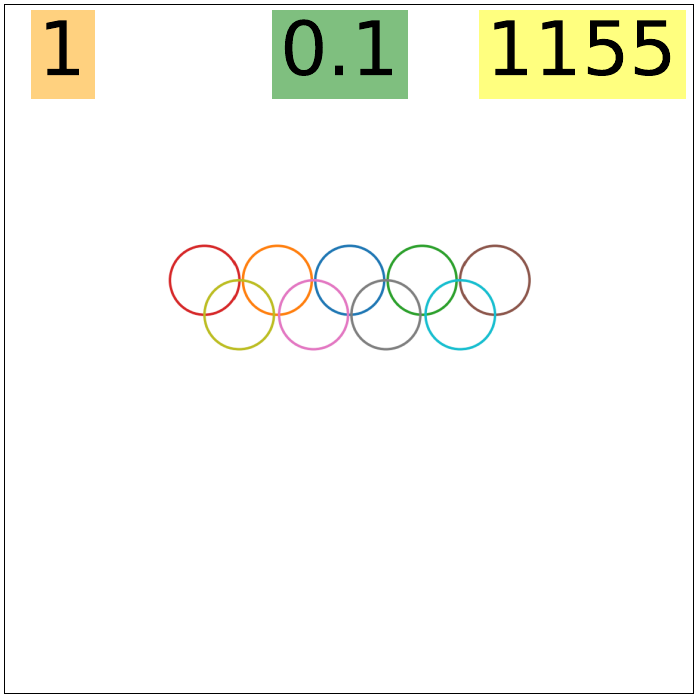} & \includegraphics[width=0.15\textwidth]{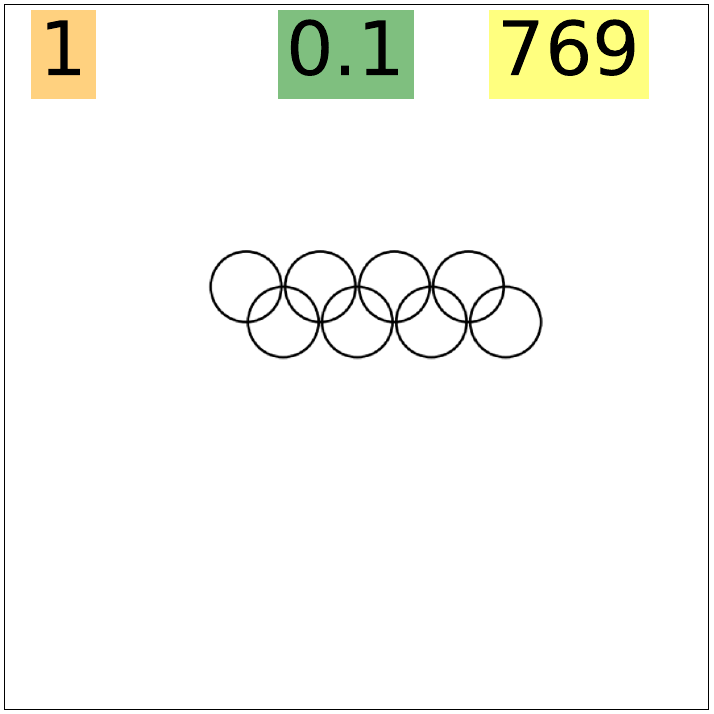} & \includegraphics[width=0.15\textwidth]{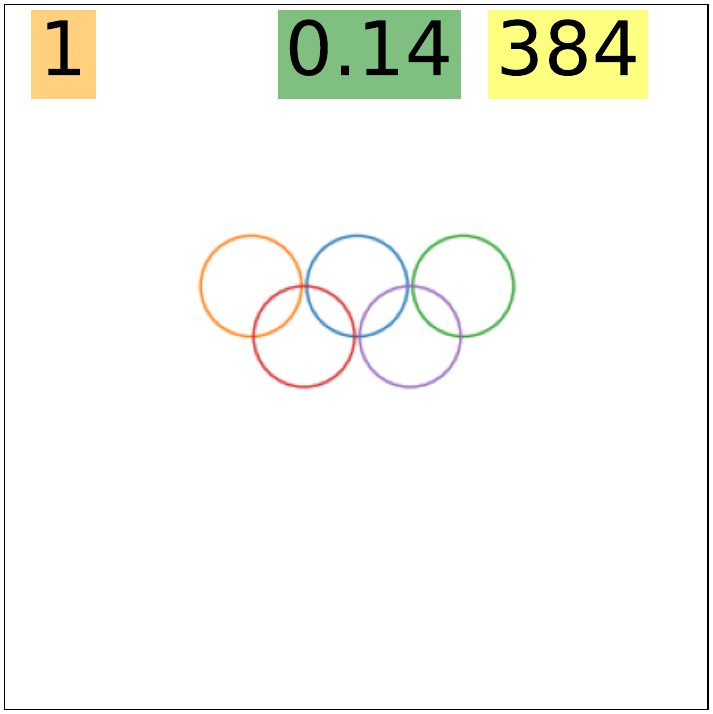}&\includegraphics[width=0.15\textwidth]{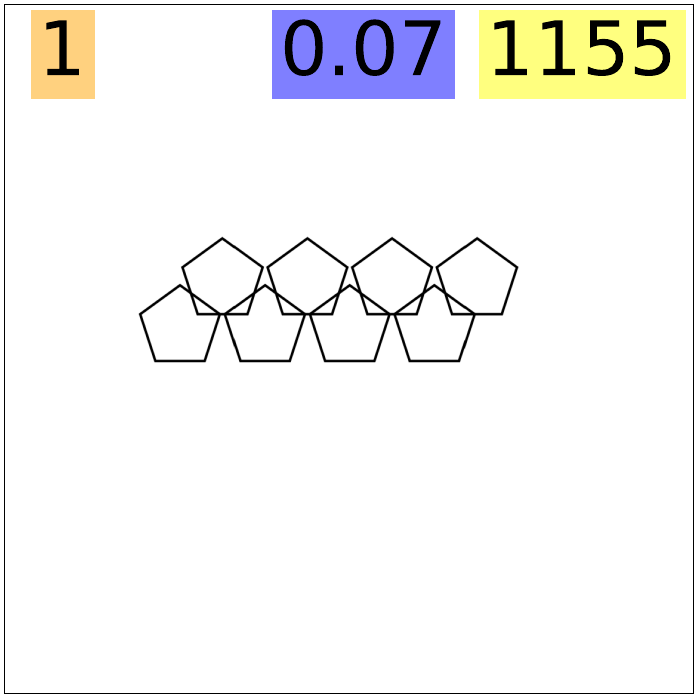}\\
\end{tabular}
\caption{We generate images of (a) different numbers of circles with various parameter changes, \eg, \sethlcolor{forestgreen(web)} \hl{the diameter}, \sethlcolor{amber} (b) \hl{the linewidth} (in points) (c) colorings, \sethlcolor{aureolin} and the \hl{image size} (in pixels). For the pentagons, we vary the \sethlcolor{mediumslateblue} \hl{side length} instead of the diameter.}
\label{fig:olympic_examples}
\end{figure}

We center the shape collection on the center of the image in two rows, and generate 480 images (see \cref{fig:olympic_examples,table:circles-parameters}), 240 images of circles, and 240 images of pentagons. 

\textbf{Code} The code is available at \url{https://github.com/anguyen8/vision-llms-are-blind/blob/main/src/CountingCircles/OlympicCircles.ipynb} for circles and at \url{https://github.com/anguyen8/vision-llms-are-blind/blob/main/src/CountingCircles/OlympicPentagons.ipynb} for pentagons.

\subsec{Prompts} 

\begin{enumerate}
    \item \emph{How many \texttt{\{shapes\}} are in the image? Answer with only the number in numerical format.}

    \item \emph{Count the \texttt{\{shapes\}} in the image. Answer with a number in curly brackets e.g. \{3\}.}
\end{enumerate}

where \texttt{\emph{\{shapes\}}} =  \emph{circles} or \emph{pentagons}.

\noindent\textbf{Groundtruth} answers are $\in \{5, 6, 7, 8, 9\}$ (random-baseline accuracy: 20\%).

\begin{figure}
\centering
\resizebox{\columnwidth}{!}{
\begin{tabular}{cc@{}cc@{}cc@{}c}
    &\multicolumn{2}{c}{(a) diameter $\phi$}&\multicolumn{2}{c}{(b) color: black vs. colorful}&\multicolumn{2}{c}{(c) shape: \BigCircle~ vs. \BigPentagon}\\
    &\includegraphics[trim={0 130 0 0},clip,width=0.16\textwidth]{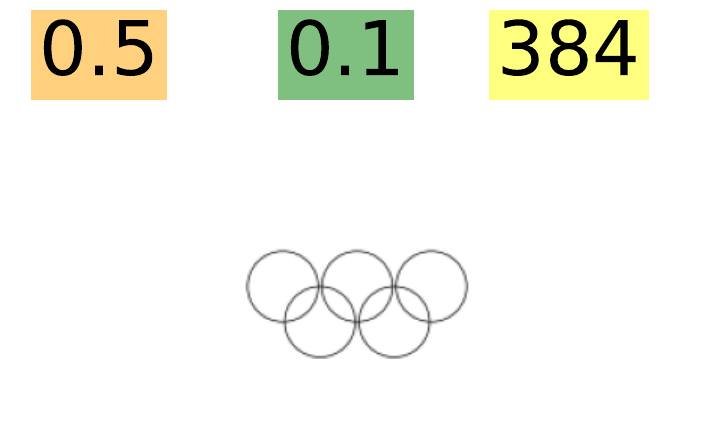}&\includegraphics[trim={0 130 0 0},clip,width=0.16\textwidth]{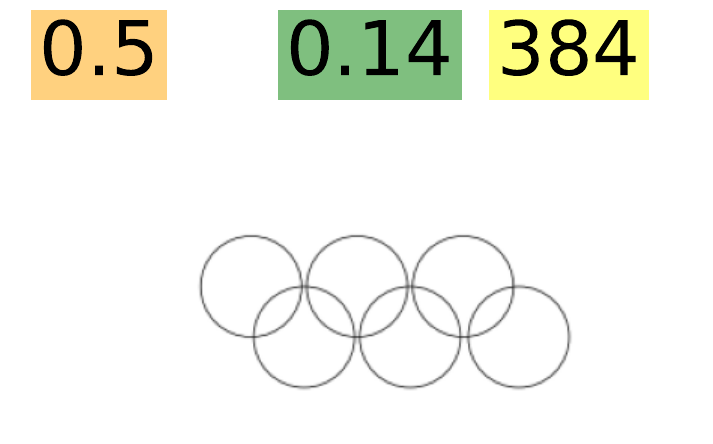}&\includegraphics[trim={0 130 0 0},clip,width=0.16\textwidth]{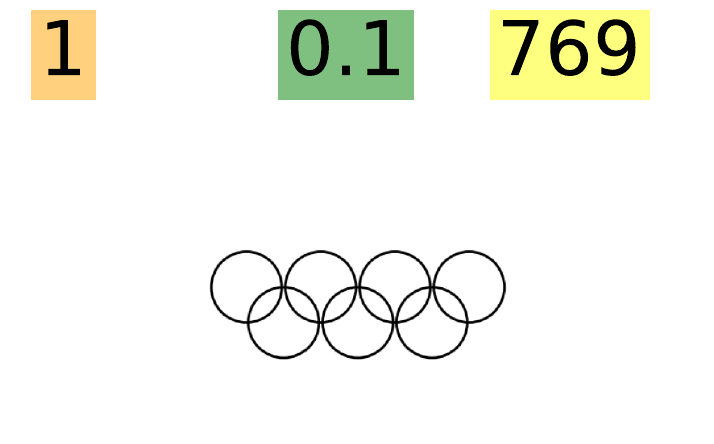}&\includegraphics[trim={0 130 0 0},clip,width=0.16\textwidth]{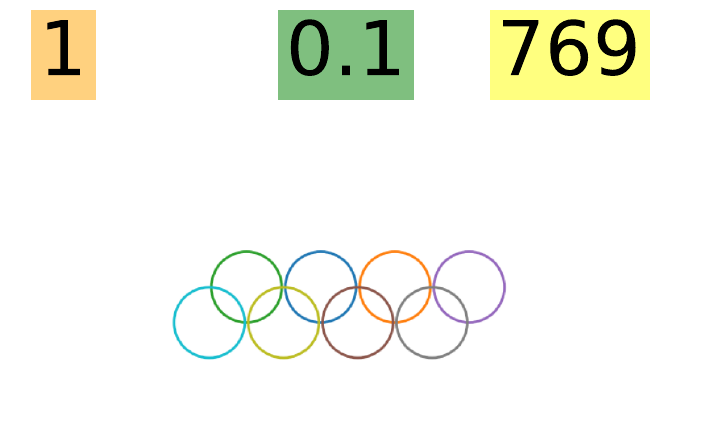}&\includegraphics[trim={0 130 0 0},clip,width=0.16\textwidth]{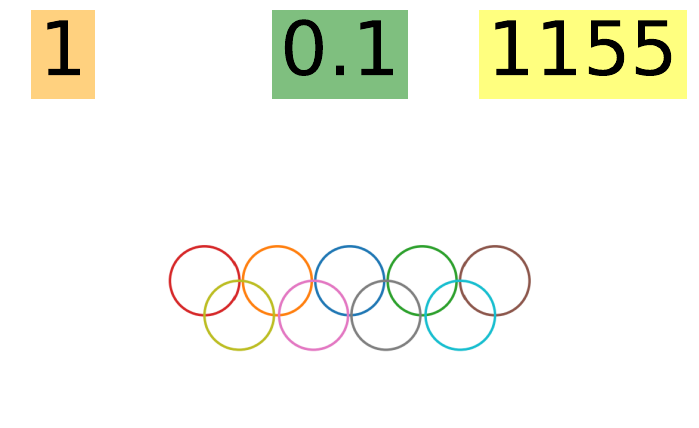}&\includegraphics[trim={0 130 0 0},clip,width=0.16\textwidth]{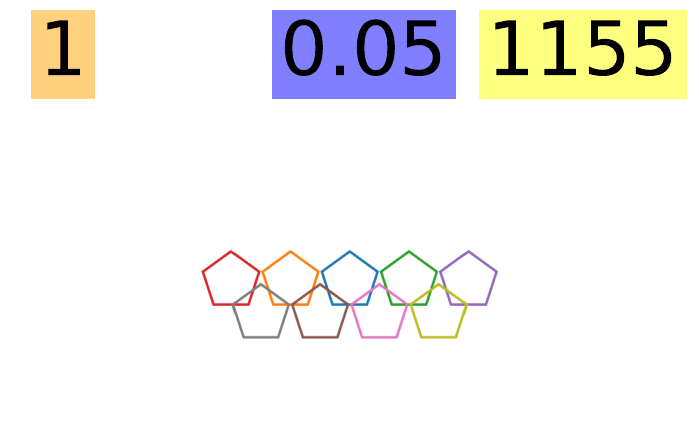}  \\
    \# shapes &5&6&7&8&9&9\\
\end{tabular}}
\caption{Images span across three \yellowsize{sizes} and shapes span across two \sethlcolor{forestgreen(web)} \hl{diameters} (and two \sethlcolor{mediumslateblue} \hl{side lengths} for \BigPentagon), two color options (black vs. colored), and two line \sethlcolor{amber}\hl{widths}.
}
\label{fig:olympic_examples_main}
\end{figure}


\begin{table}
\centering
\caption{We create 480 images containing circles and pentagons that are uniformly distributed over various parameters.}
\label{table:circles-parameters}
\begin{tabular}{lccc}
\toprule
\textbf{Parameter} & \textbf{Values} & \textbf{Samples} & ~~\textbf{Total Samples}\\
\midrule
\multirow{2}{*}{Shape} & Circles & 240 & ~~\multirow{2}{*}{480}\\
& Pentagons & 240 \\
\midrule
\multirow{3}{*}{Image size ($C$)} & 384px & 80 & ~~\multirow{3}{*}{240}\\
& 769px & 80 \\
& 1155px & 80 \\
\midrule
\multirow{2}{*}{Line thickness} & $0.0025\times C$ pixels & 120 & ~~\multirow{2}{*}{240}\\
 & $0.005\times C$ pixels & 120 \\
\midrule
\multirow{5}{*}{Number of shapes} & 5 & 48 & ~~\multirow{5}{*}{240}\\
 & 6 & 48 \\
 & 7 & 48 \\
& 8 & 48 \\
& 9 & 48 \\
\midrule
\multirow{3}{*}{Diameter/Side} & $\frac{C}{7}$ & 120 & ~~\multirow{3}{*}{240}\\
\\
 & $\frac{C}{10}$ & 120 \\
\bottomrule
\end{tabular}
\end{table}

\clearpage
\subsection{Finding: different resolutions have no impact on most VLMs' performance}
\label{app:resolu-olympic}

\begin{figure}
    \centering
    \begin{tabular}{ccc}
        \multicolumn{3}{c}
        {\includegraphics[width=\textwidth]{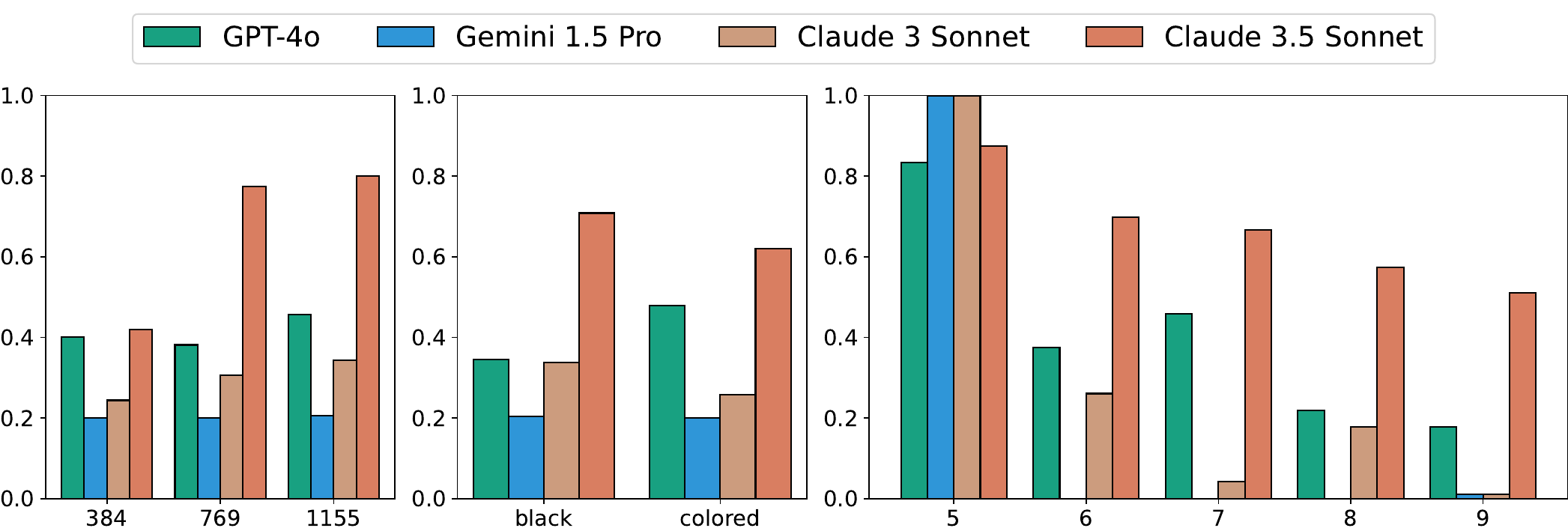}} \\
        \multicolumn{3}{c}{Counting overlapping {circles} \olympiclogo} \\
        \\
        \\
        \hspace{0.8cm}(a) Resolution & \hspace{1.6cm}(b) Color & \hspace{1.3cm}(c) Number of shapes \\
        \multicolumn{3}{c}{\includegraphics[width=\textwidth]{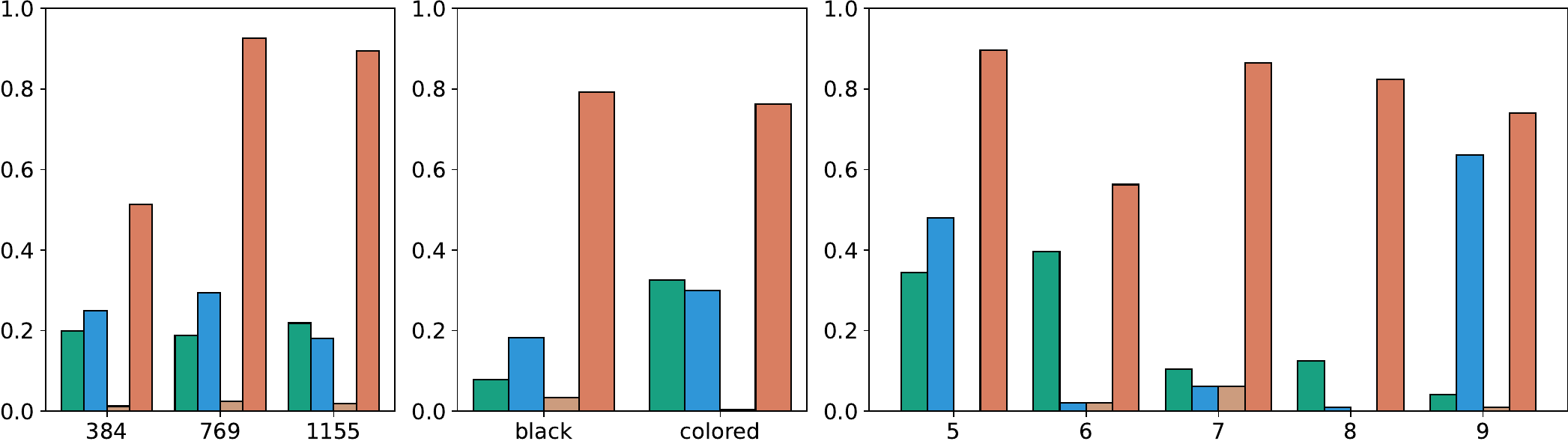}} \\
        \multicolumn{3}{c}{Counting overlapping {pentagons} \pentagonlogo} \\        
    \end{tabular}
    \caption{
    VLMs perform better on counting overlapping circles \olympiclogo (top) than overlapping pentagons \pentagonlogo (bottom).
    For most models, resolution (a) and colors (b) have minimal impact on performance. 
    \newsonnet performs better as the image size increases (a). 
    \gpt performs better on colored shapes than on black shapes.
    }
    \label{fig:all-olympic}
\end{figure}

\cref{fig:all-olympic}-a shows that VLMs are invariant to the resolution when asked to count the overlapping shapes. This suggests that the image quality has almost no effect on the performance, and VLMs cannot see the shapes.

\subsection{Finding: color-coding does not generally help the VLMs}
\label{app:color-olympic}
While we expect the color-coding to make the shapes more distinct for the models, \cref{fig:all-olympic}-b suggests that, except for \gpt, coloring the shapes has an opposite effect on the performance of the models.

\clearpage
\subsection{Finding: \gemini has the most biased predictions to the Olympic logo}
\label{app:bias-gemini-fig}

\begin{figure}
\centering
\begin{subfigure}[]{0.475\textwidth}
\centering
    \includegraphics[width=\textwidth]{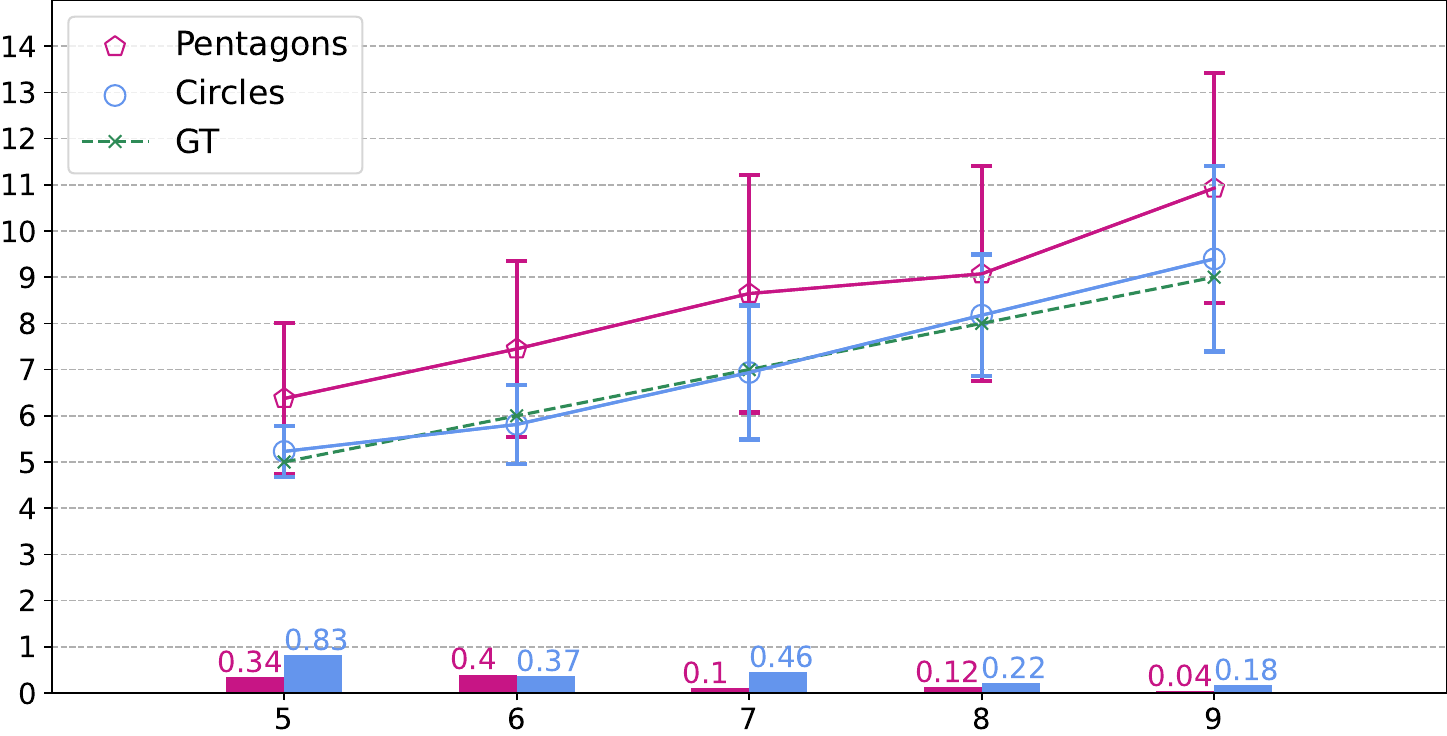}
    \caption{\gpt}
\label{app:gpt-trend}
\end{subfigure}
\begin{subfigure}[]{0.475\textwidth}
\centering
    \includegraphics[width=\textwidth]{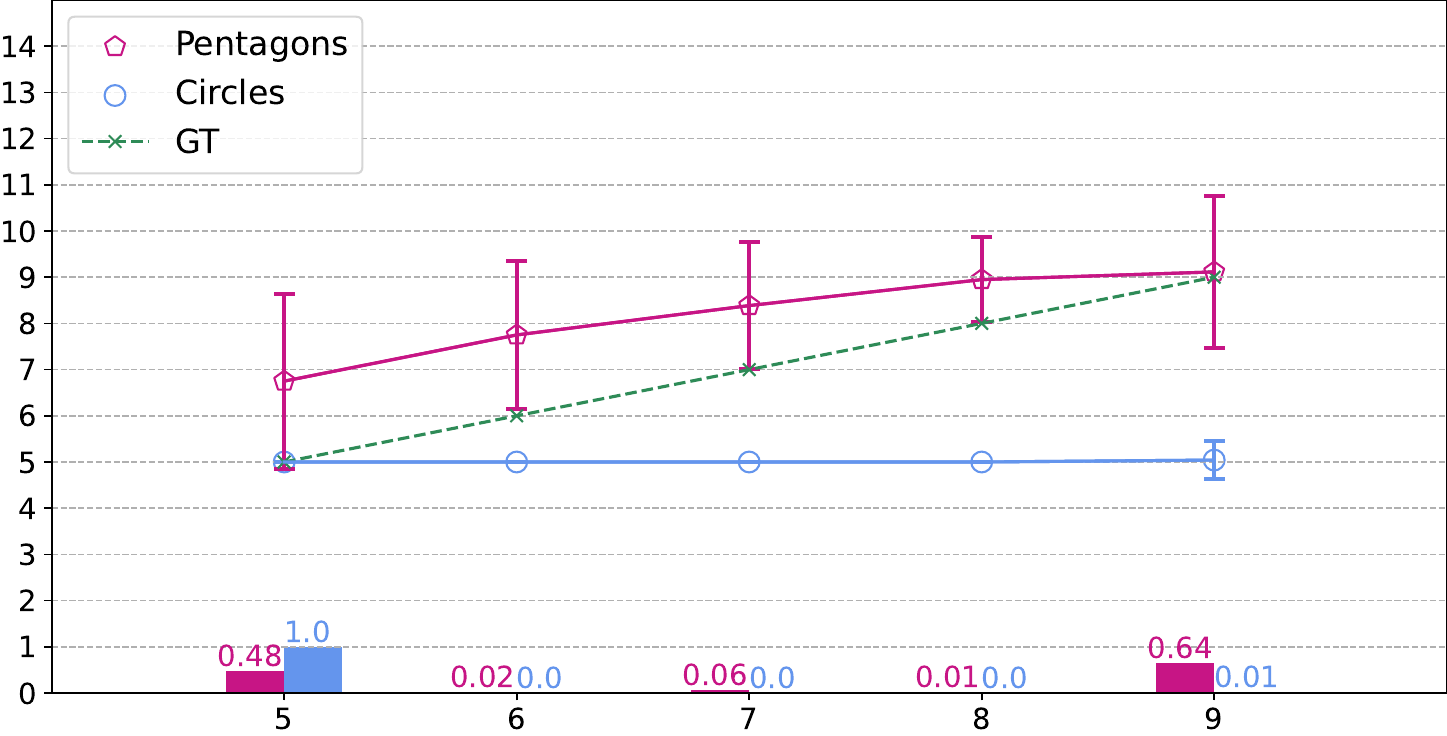}
    \caption{\gemini}
\label{app:gemini-trend}
\end{subfigure}
\begin{subfigure}[]{0.475\textwidth}
\centering
    \includegraphics[width=\textwidth]{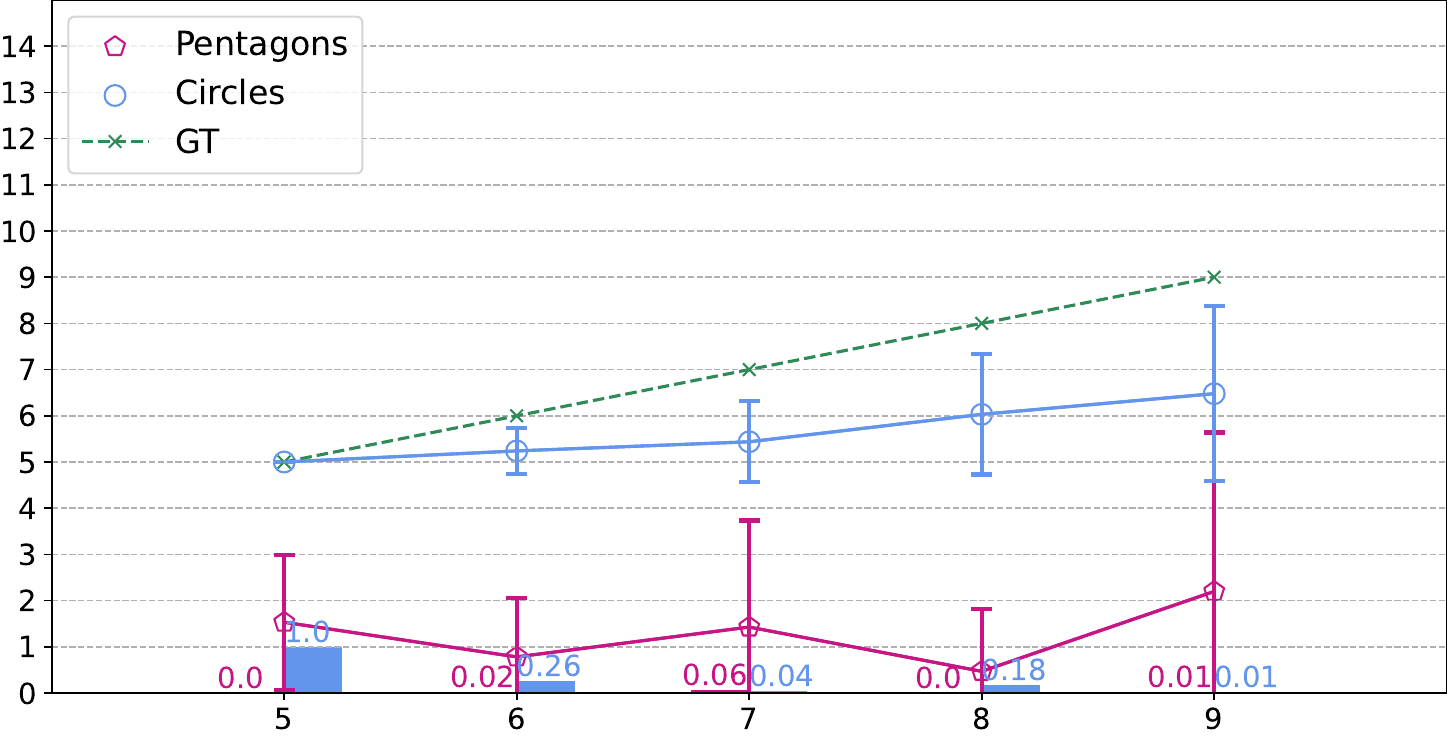} 
\caption{\sonnet}

\label{app:claude-trend}
\end{subfigure}
\begin{subfigure}[]{0.475\textwidth}
\centering
    \includegraphics[width=\textwidth]{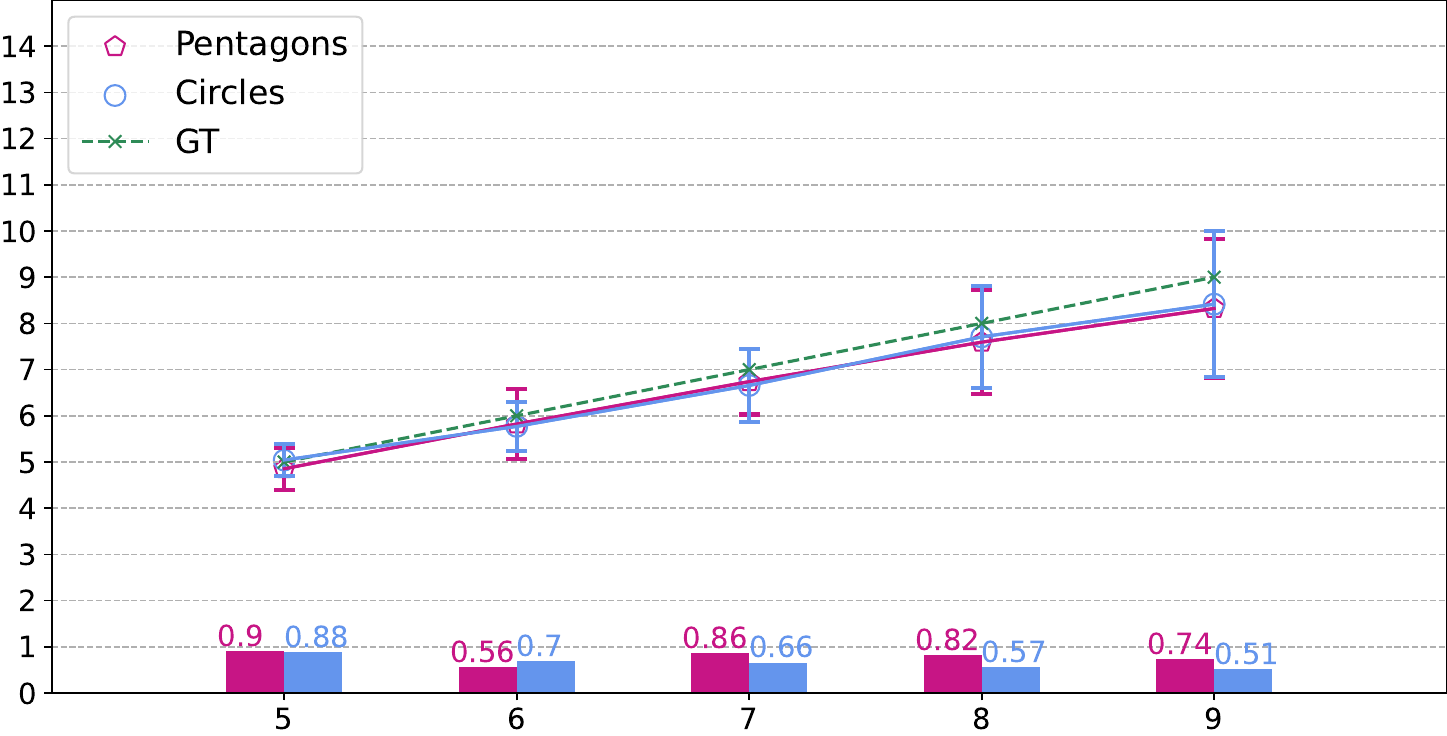}
    \caption{\newsonnet}
    \label{app:cl35-trend}
\end{subfigure}
\caption{Prediction trend for each VLM shows (a) \gpt has less variance in counting circles versus pentagons, (b) \gemini is biased to predicting the number of circles to be 5, (c) \sonnet tends to under count the number of shapes, and (d) \newsonnet has the least relative variance over both shapes.}
\label{app:fig_trends_olympic}
\end{figure}

\cref{app:fig_trends_olympic} shows the overall trend of the predictions among SOTA VLMs. \gemini (see \cref{app:gemini-trend}) tends to predict ``5'' significantly more often when asked to count the circles, while its predictions are more random for pentagons. This suggests the model's bias toward the Olympic logo.

\begin{table}[htb]
    \centering
    \caption{
    Frequency (\%) of predicting ``5'' when there are more than 5 circles (\BigCircle) or pentagons (\BigPentagon), \ie, $N = 6, 7, 8, 9$ shapes in the image.
    For example, \gemini predicts ``5'' circles 99.74\% of the time but this tendency disappears in the case of \BigPentagon (10.94\%), showing a strong bias towards the 5-circle Olympic logo (among four models).}
    \label{tab:biased-pred}
    \begin{tabular}{lcccccccccccc}
    \toprule
    &&\multicolumn{2}{c}{\gptlogo} &&\multicolumn{2}{c}{\geminilogo} &&\multicolumn{2}{c}{\sonnetlogo} &&\multicolumn{2}{c}{\newsonnetlogo}\\
    \cmidrule(rl){3-4}
    \cmidrule(rl){6-7}
    \cmidrule(rl){9-10}
    \cmidrule(rl){12-13}
    &&  \BigCircle & \BigPentagon &&  \BigCircle & \BigPentagon &&  \BigCircle & \BigPentagon &&  \BigCircle & \BigPentagon\\
    \midrule
    &&13.54&5.47&&99.74&10.94&&64.32&5.73&&15.89&10.16 \\
    \bottomrule
    \end{tabular}
\end{table}

\clearpage
\subsection{Additional Examples}
\label{app:circles-qual}

\begin{figure}
\centering
\begin{AIbox}{Count the circles in the image. Answer with a number in curly brackets e.g. \{3\}}
\centering
\begin{tabular}{lp{0.8cm}c|p{0.8cm}c|p{0.8cm}c|p{0.8cm}c|p{0.8cm}c|p{0.8cm}c}
     & \multicolumn{2}{c}{\includegraphics[width=0.15\textwidth]{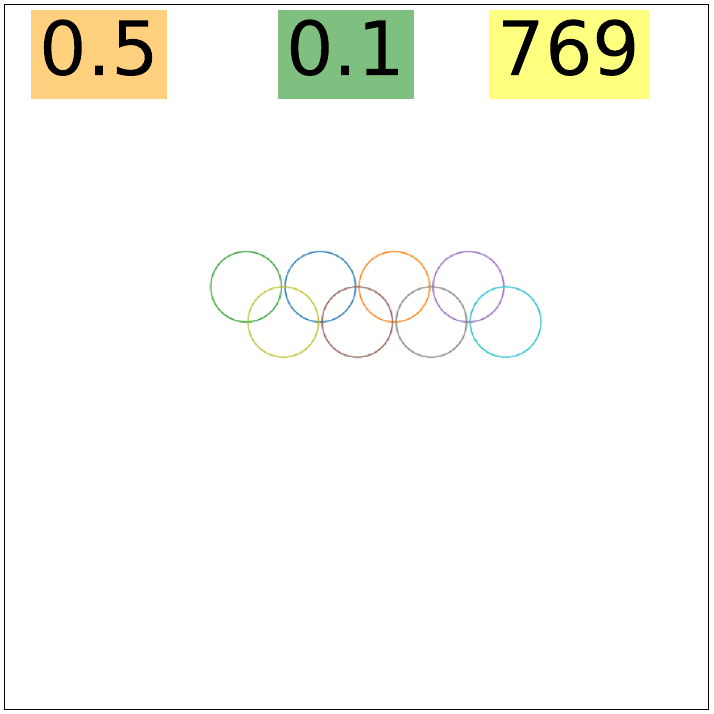}} &
     \multicolumn{2}{c}{\includegraphics[width=0.15\textwidth]{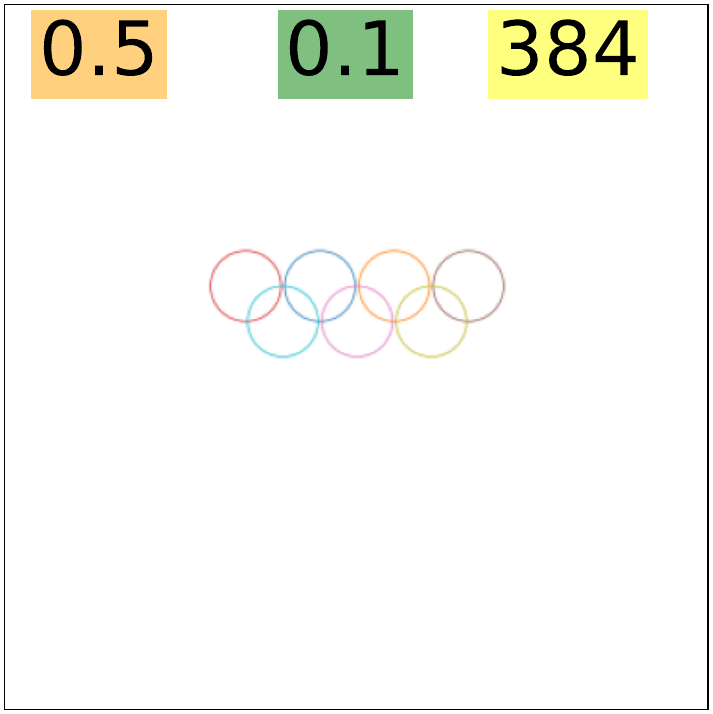}} &
     \multicolumn{2}{c}{\includegraphics[width=0.15\textwidth]{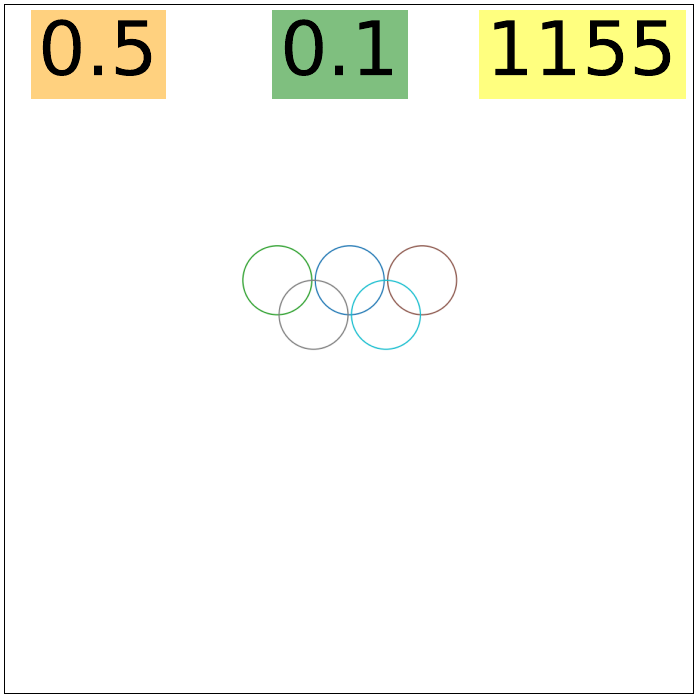}} & 
     \multicolumn{2}{c}{\includegraphics[width=0.15\textwidth]{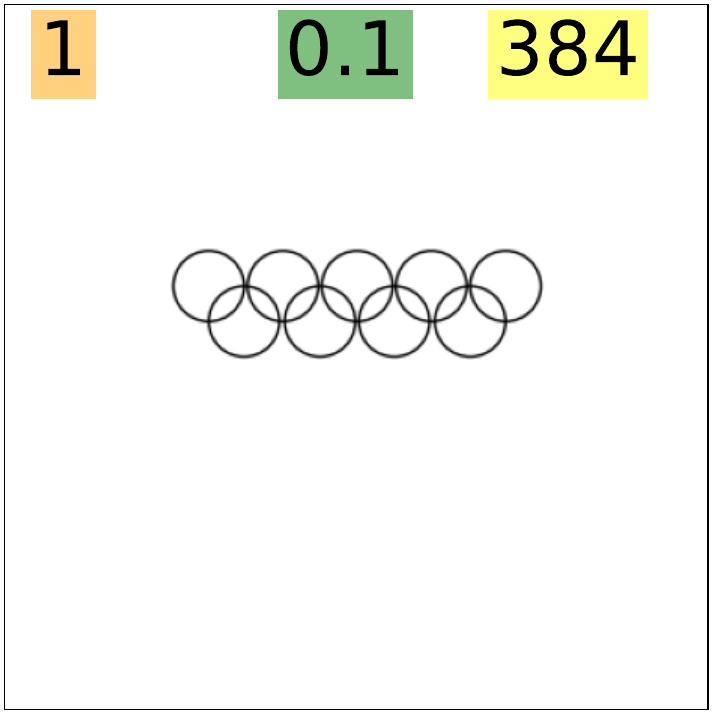}} &
     \multicolumn{2}{c}{\includegraphics[width=0.15\textwidth]{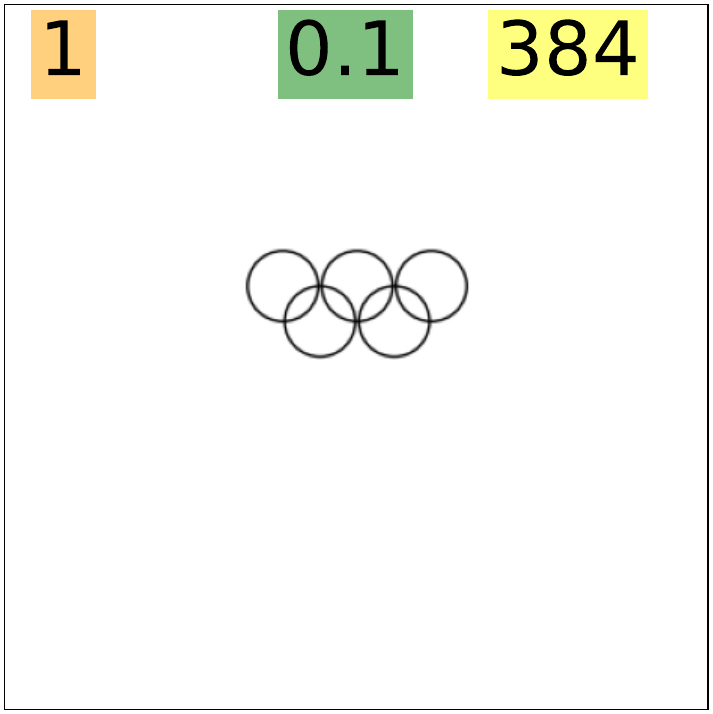}} &
     \multicolumn{2}{c}{\includegraphics[width=0.15\textwidth]{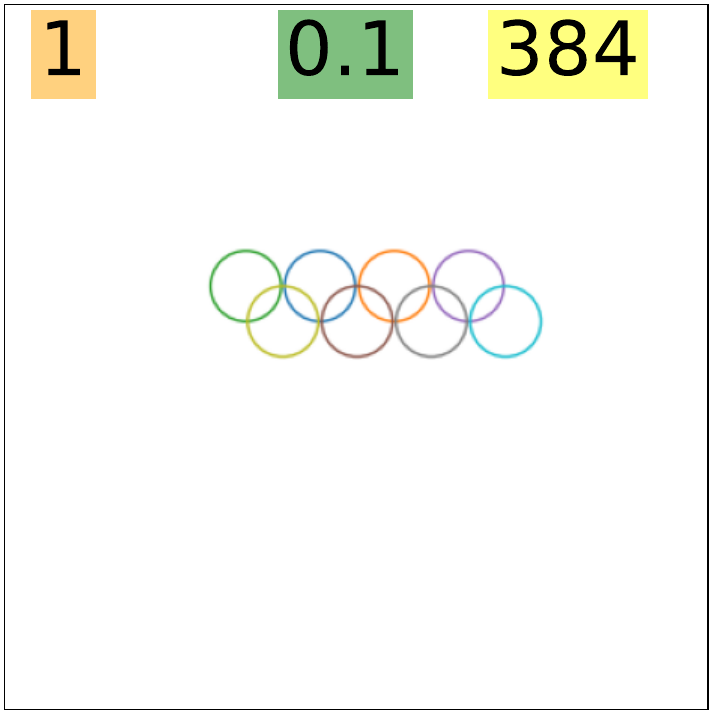}} \\
     \rowcolor{lightgray}
     \raisebox{-0.2\height}\gptlogo & \centering 9 & \textcolor{red}{\xmark} &  \centering 6 & \textcolor{red}{\xmark}&  \centering 5 & \textcolor{ForestGreen}{\cmark} &  \centering 8 & \textcolor{red}{\xmark} &  \centering 5  & \textcolor{ForestGreen}{\cmark}&  \centering  8 & \textcolor{ForestGreen}{\cmark}\\
     \raisebox{-0.2\height}\geminilogo & \centering 5 & \textcolor{red}{\xmark} & \centering 5 & \textcolor{red}{\xmark}&  \centering 5 & \textcolor{ForestGreen}{\cmark} &  \centering 5 & \textcolor{red}{\xmark} &  \centering 5 & \textcolor{red}{\xmark}&  \centering 5 & \textcolor{red}{\xmark}\\
     \rowcolor{lightgray}
     \raisebox{-0.2\height}\sonnetlogo & \centering 5 & \textcolor{red}{\xmark} & \centering 5 & \textcolor{red}{\xmark}&  \centering 5 & \textcolor{ForestGreen}{\cmark} &  \centering 8 & \textcolor{red}{\xmark} &  \centering 5 & \textcolor{ForestGreen}{\cmark}&  \centering 8 & \textcolor{ForestGreen}{\cmark}\\
       \raisebox{-0.2\height}\newsonnetlogo & \centering 8 & \textcolor{ForestGreen}{\cmark} & \centering 5 & \textcolor{red}{\xmark}&  \centering 5 & \textcolor{ForestGreen}{\cmark} &  \centering 5& \textcolor{red}{\xmark} &  \centering 5 & \textcolor{ForestGreen}{\cmark}&  \centering 5 & \textcolor{red}{\xmark}\\

     \end{tabular}

    \vspace{4pt}
    \centering
    \begin{tabular}{cccccccccccccc}
    \raisebox{-0.1\height}\gptlogo & \gpt &&&
      \raisebox{-0.1\height}\geminilogo & \gemini & &&\raisebox{-0.1\height}\sonnetlogo & \sonnet & && \raisebox{-0.12\height}\newsonnetlogo & \newsonnet \\
      \end{tabular}
     
\end{AIbox}
\caption{VLMs tend only count the correct number of shapes when the image includes 5 circles. This shows that their predictions are biased toward the actual Olympic logo.}
\label{appfig:circles_p2}
\end{figure}

We provide examples of VLMs' responses in \cref{appfig:circles_p2}.

\begin{figure}[H]
\centering
\begin{AIbox}{How many circles are in the image? Answer with only the number in numerical format.}
\centering
\begin{tabular}{lp{0.8cm}c|p{0.8cm}c|p{0.8cm}c|p{0.8cm}c|p{0.8cm}c|p{0.8cm}c}
     & \multicolumn{2}{c}{\includegraphics[width=0.15\textwidth]{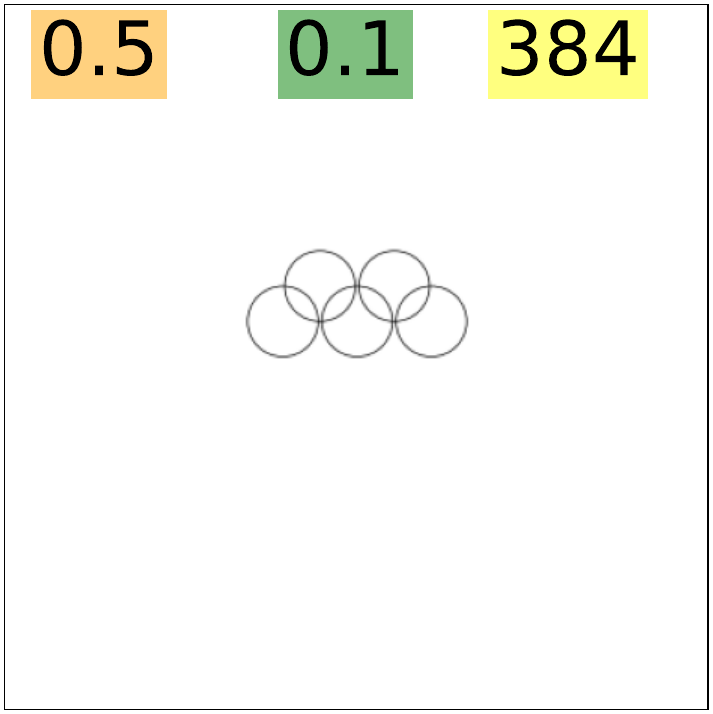}} &
     \multicolumn{2}{c}{\includegraphics[width=0.15\textwidth]{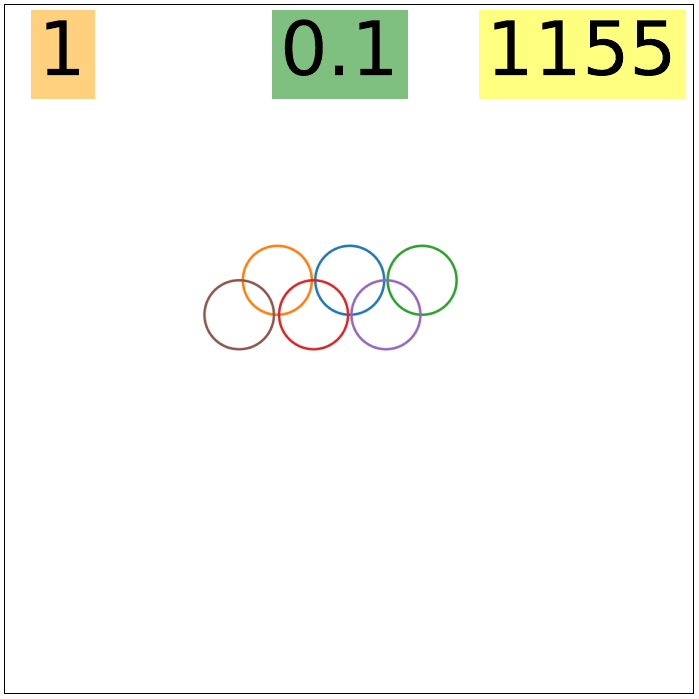}} &
     \multicolumn{2}{c}{\includegraphics[width=0.15\textwidth]{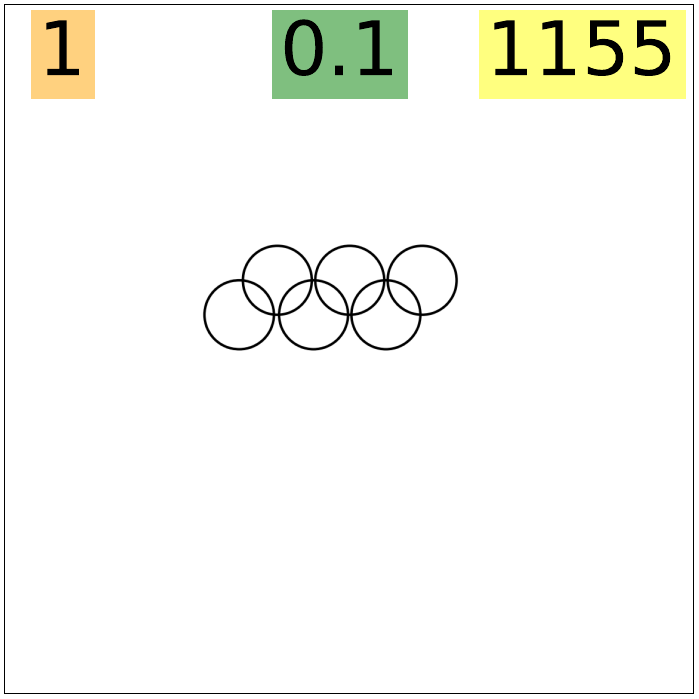}} & 
     \multicolumn{2}{c}{\includegraphics[width=0.15\textwidth]{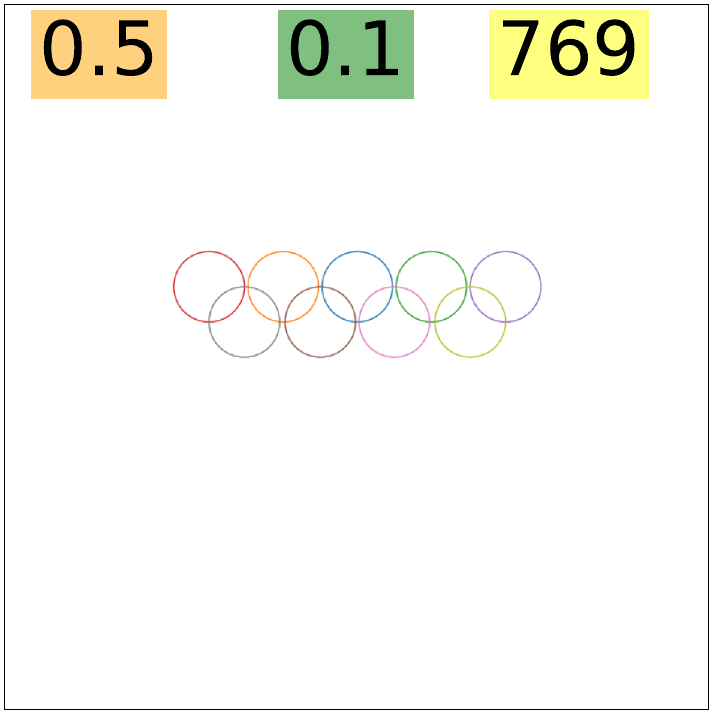}} &
     \multicolumn{2}{c}{\includegraphics[width=0.15\textwidth]{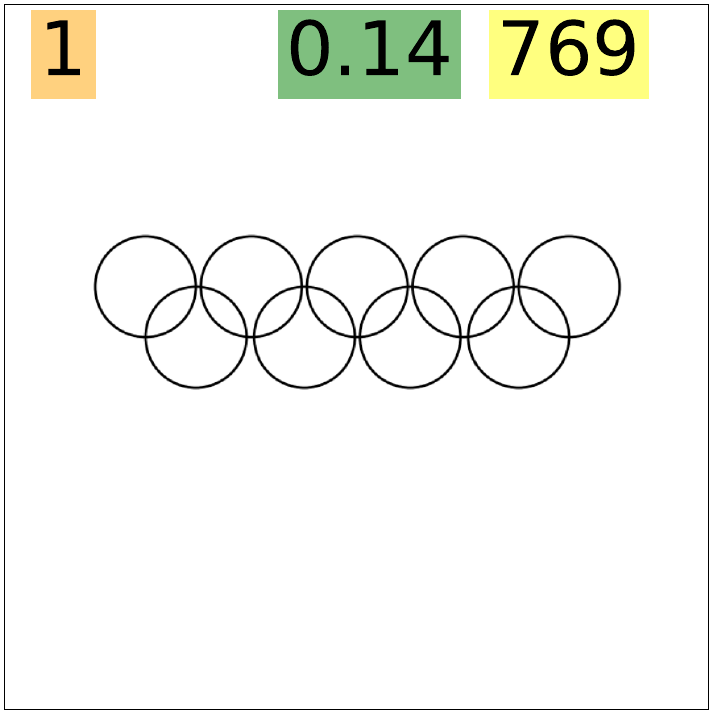}} &
     \multicolumn{2}{c}{\includegraphics[width=0.15\textwidth]{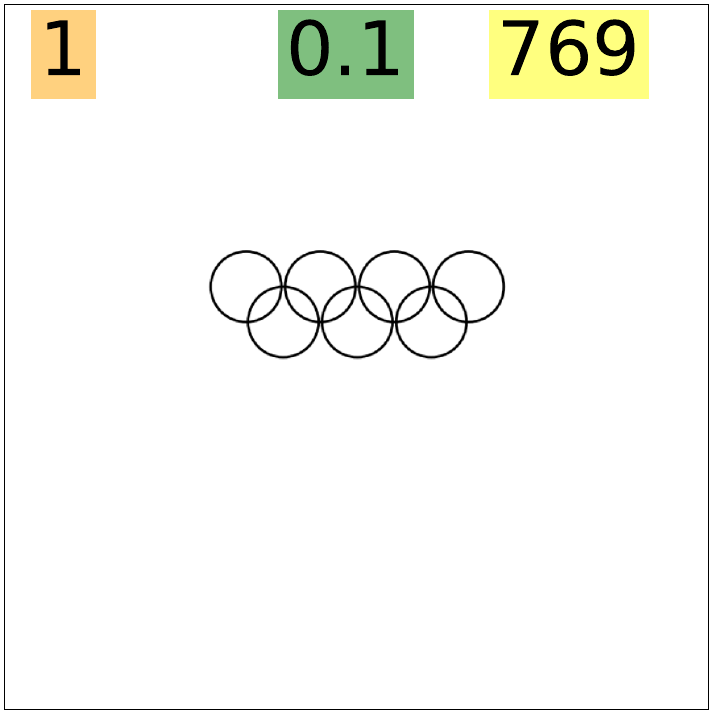}} \\
     \rowcolor{lightgray}
     \raisebox{-0.2\height}\gptlogo & \centering 5 & \textcolor{ForestGreen}{\cmark} &  \centering 5 & \textcolor{red}{\xmark}&  \centering 7 & \textcolor{red}{\xmark} &  \centering 12 & \textcolor{red}{\xmark} &  \centering 11  & \textcolor{red}{\xmark}&  \centering  5 & \textcolor{red}{\xmark}\\
     \raisebox{-0.2\height}\geminilogo & \centering 5 & \textcolor{ForestGreen}{\cmark} & \centering 5 & \textcolor{red}{\xmark}&  \centering 5 & \textcolor{red}{\xmark} &  \centering 5 & \textcolor{red}{\xmark} &  \centering 5 & \textcolor{red}{\xmark}&  \centering 5 & \textcolor{red}{\xmark}\\
     \rowcolor{lightgray}
     \raisebox{-0.2\height}\sonnetlogo & \centering 3 & \textcolor{red}{\xmark} & \centering 5 & \textcolor{red}{\xmark}&  \centering 5 & \textcolor{red}{\xmark} &  \centering 10 & \textcolor{red}{\xmark} &  \centering 10 & \textcolor{red}{\xmark}&  \centering 5 & \textcolor{red}{\xmark}\\
       \raisebox{-0.2\height}\newsonnetlogo & \centering 4 & \textcolor{red}{\xmark} & \centering 6 & \textcolor{ForestGreen}{\cmark}&  \centering 6 & \textcolor{ForestGreen}{\cmark} &  \centering 10 & \textcolor{red}{\xmark} &  \centering 9 & \textcolor{ForestGreen}{\cmark}&  \centering 7 & \textcolor{ForestGreen}{\cmark}\\

     \end{tabular}

    \vspace{4pt}
    \centering
    \begin{tabular}{cccccccccccccc}
    \raisebox{-0.1\height}\gptlogo & \gpt &&&
      \raisebox{-0.1\height}\geminilogo & \gemini & &&\raisebox{-0.1\height}\sonnetlogo & \sonnet & && \raisebox{-0.12\height}\newsonnetlogo & \newsonnet \\
      \end{tabular}
     
\end{AIbox}
\caption{Counting overlapped circles is not easy to VLMs regardless of circle colors, \orangedistance{line widths}, and \yellowsize{resolutions}.
\gemini often predicts ``5'' regardless of the actual circle count, suggesting a strong bias towards the well-known Olympic logo.}
\label{appfig:circles_p1}
\end{figure}


\clearpage
\section{Counting the rows and columns of a grid task}
\label{app:grids}

\subsection{Benchmark Construction Details}
\label{app:grid-bench}

Our benchmark specifications consist of various parameters for the grid generation process.

\begin{itemize}
    \item \textbf{Image size}: We include three different sizes \{500, 1250, 2000\} to create the grids on the image.
    \item \textbf{Number of rows/columns}: We choose a base size $N\in\{3, 4, 5, 6, 7, 8, 9\}$, and initialize the sizes to \(N\times N\), \(N\times N'\), and \(N'\times N\) where $N' = N + 1$. We also create 10$\times$10 grids to balance the square grids.
    \item \textbf{Line thickness}: We use a line thickness of 1\% of the image size and 0.5\% of the image size.
    \item \textbf{Entry}: Each table is generated in two versions, one that includes blank entries, and the second with random text entries.
     
\end{itemize}

\begin{figure}
    \centering
    \begin{tabular}{cc}
         \includegraphics[width=0.33\textwidth]{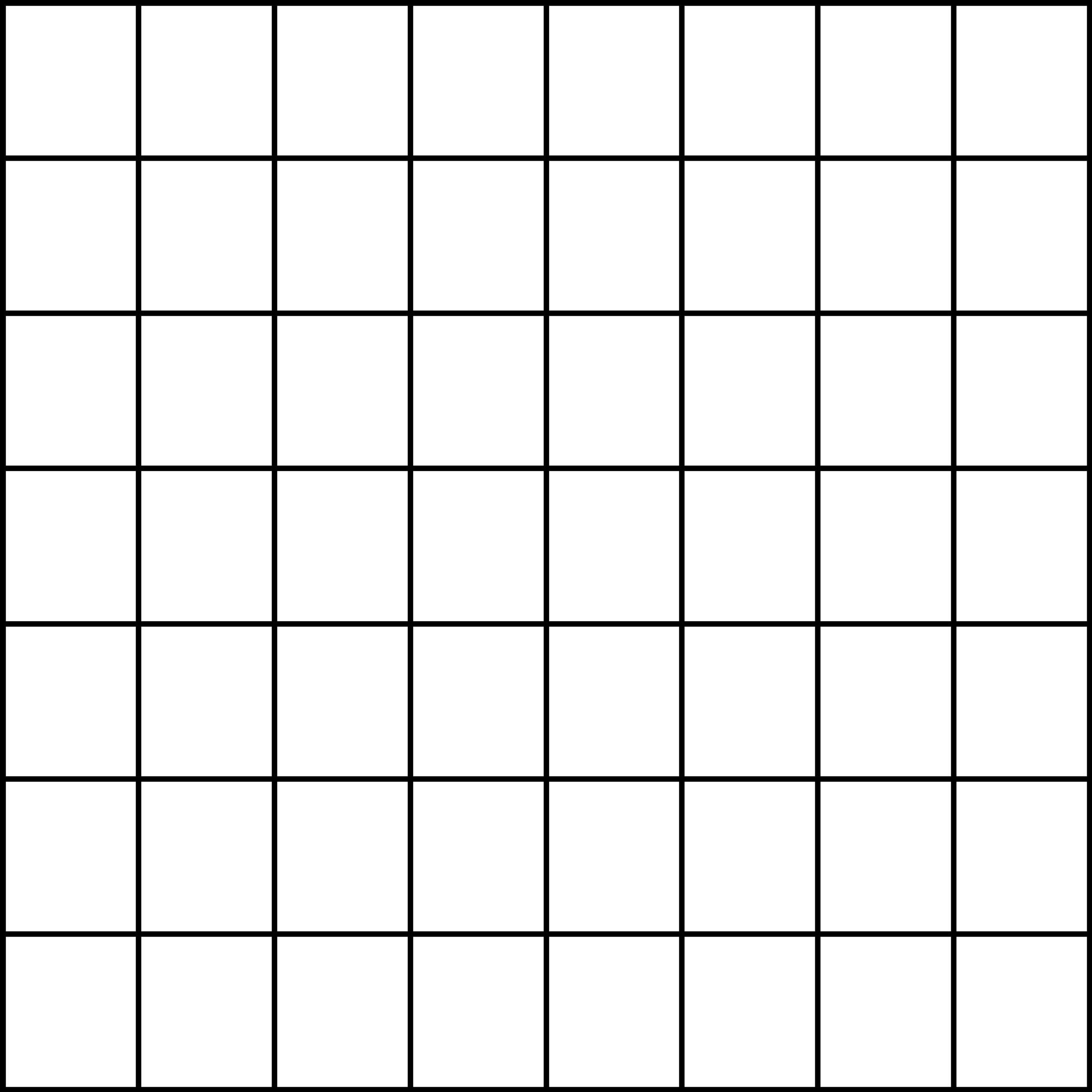}& \includegraphics[width=0.33\textwidth]{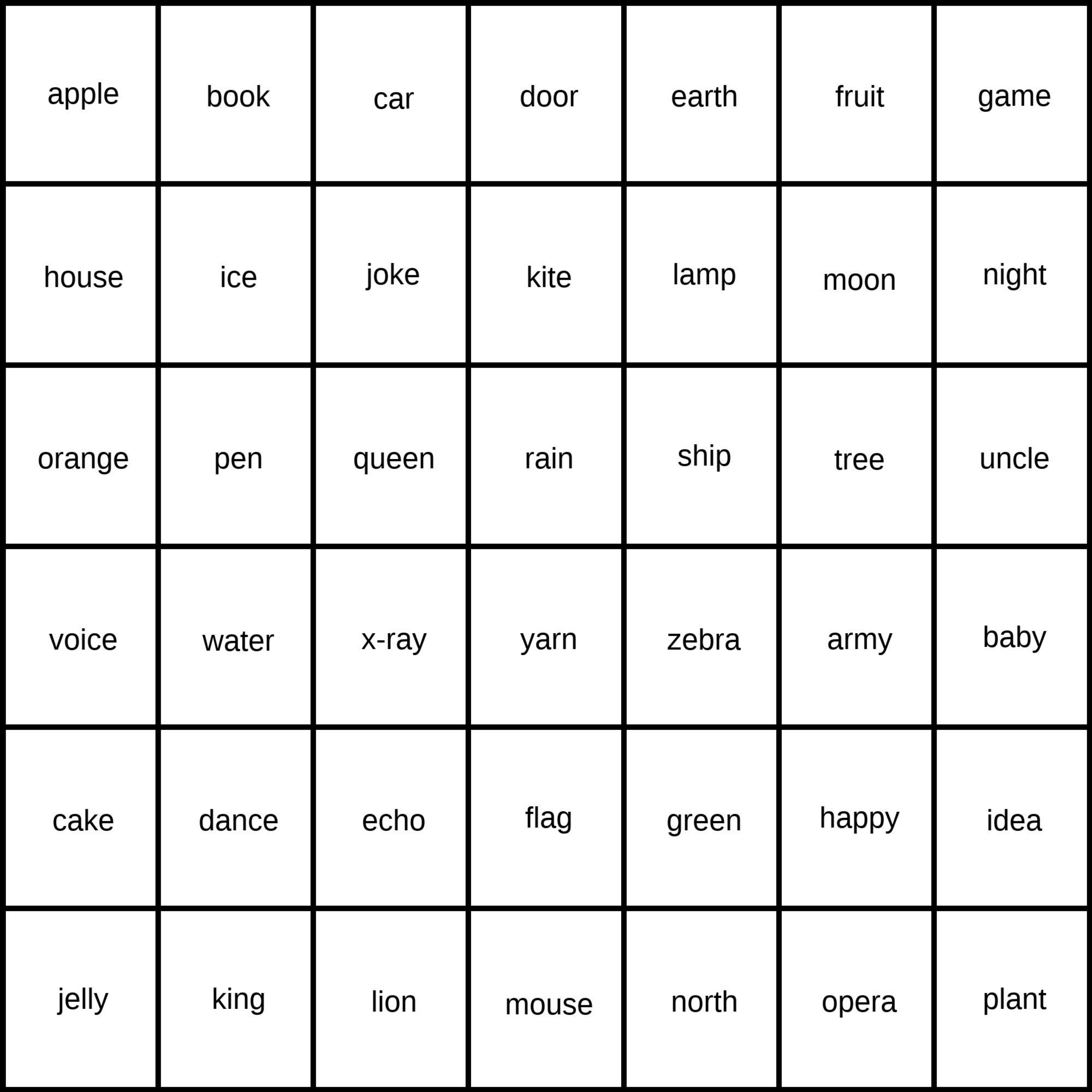} \\
         \includegraphics[width=0.33\textwidth]{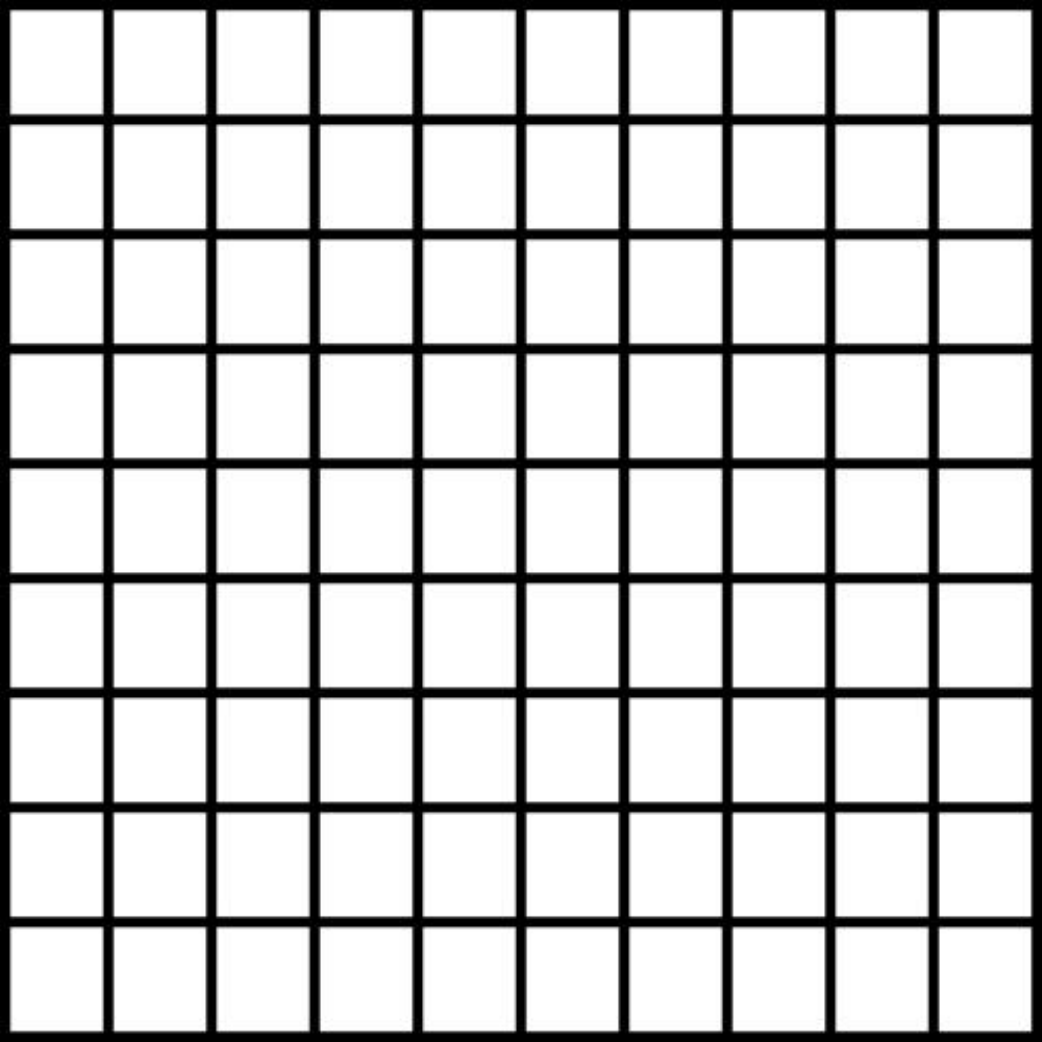}& \includegraphics[width=0.33\textwidth]{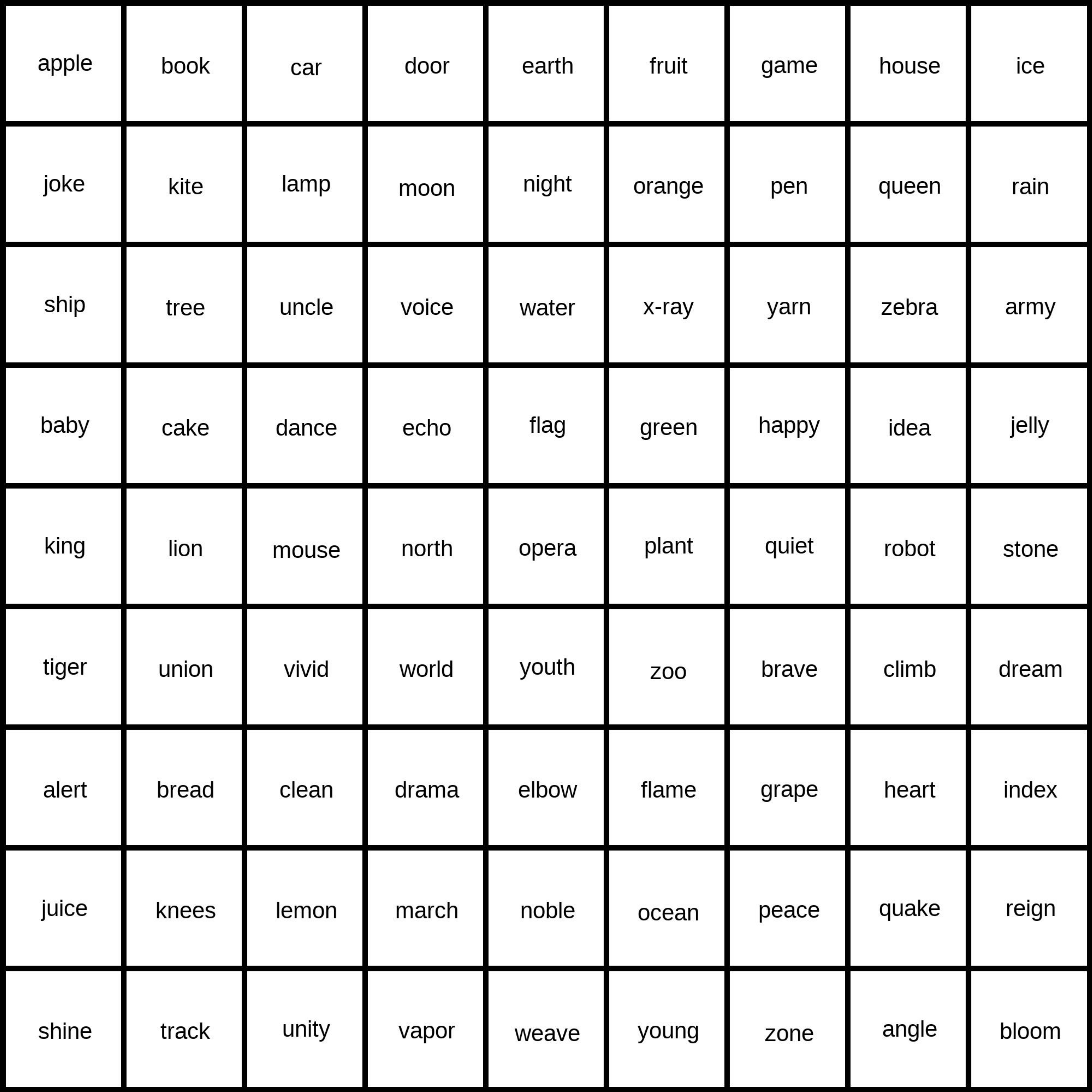} \\
         \includegraphics[width=0.33\textwidth]{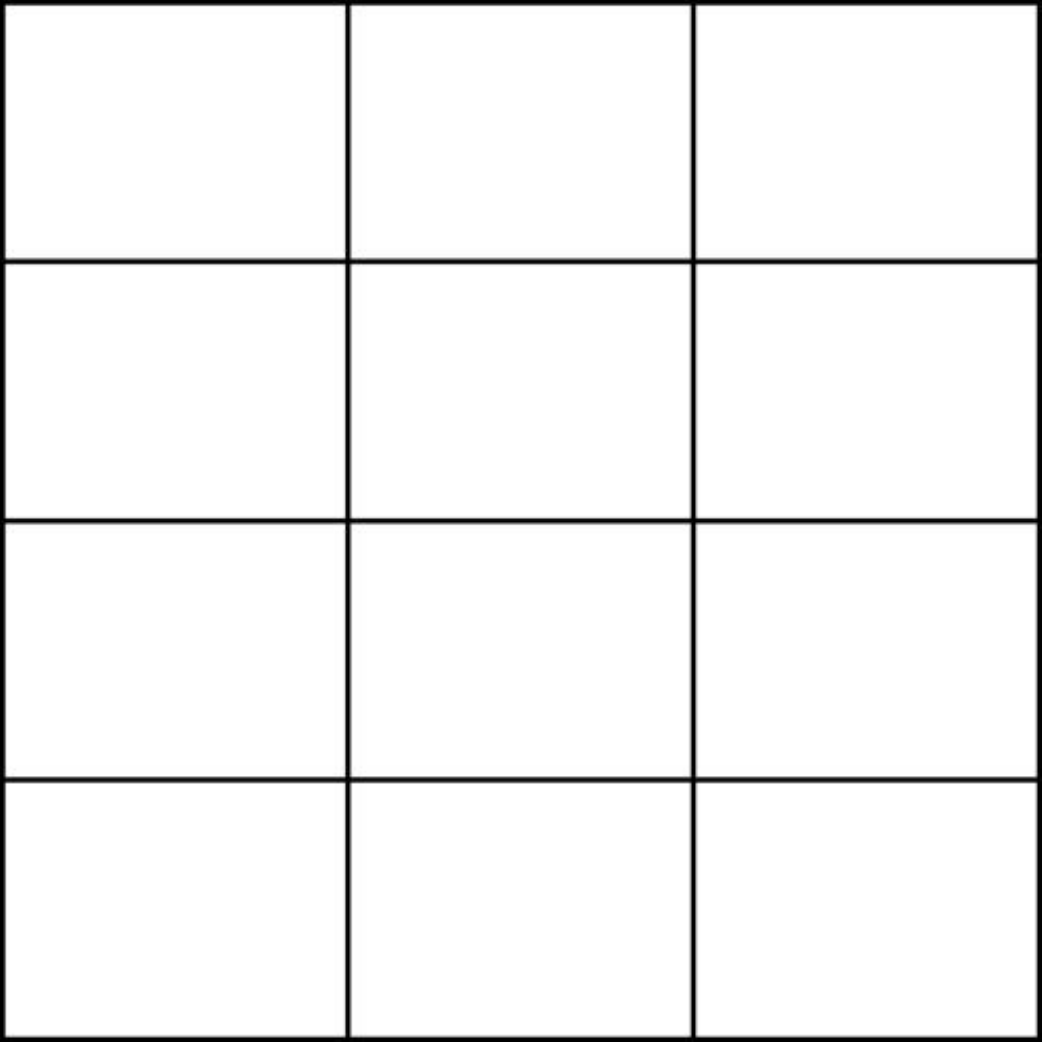}& \includegraphics[width=0.33\textwidth]{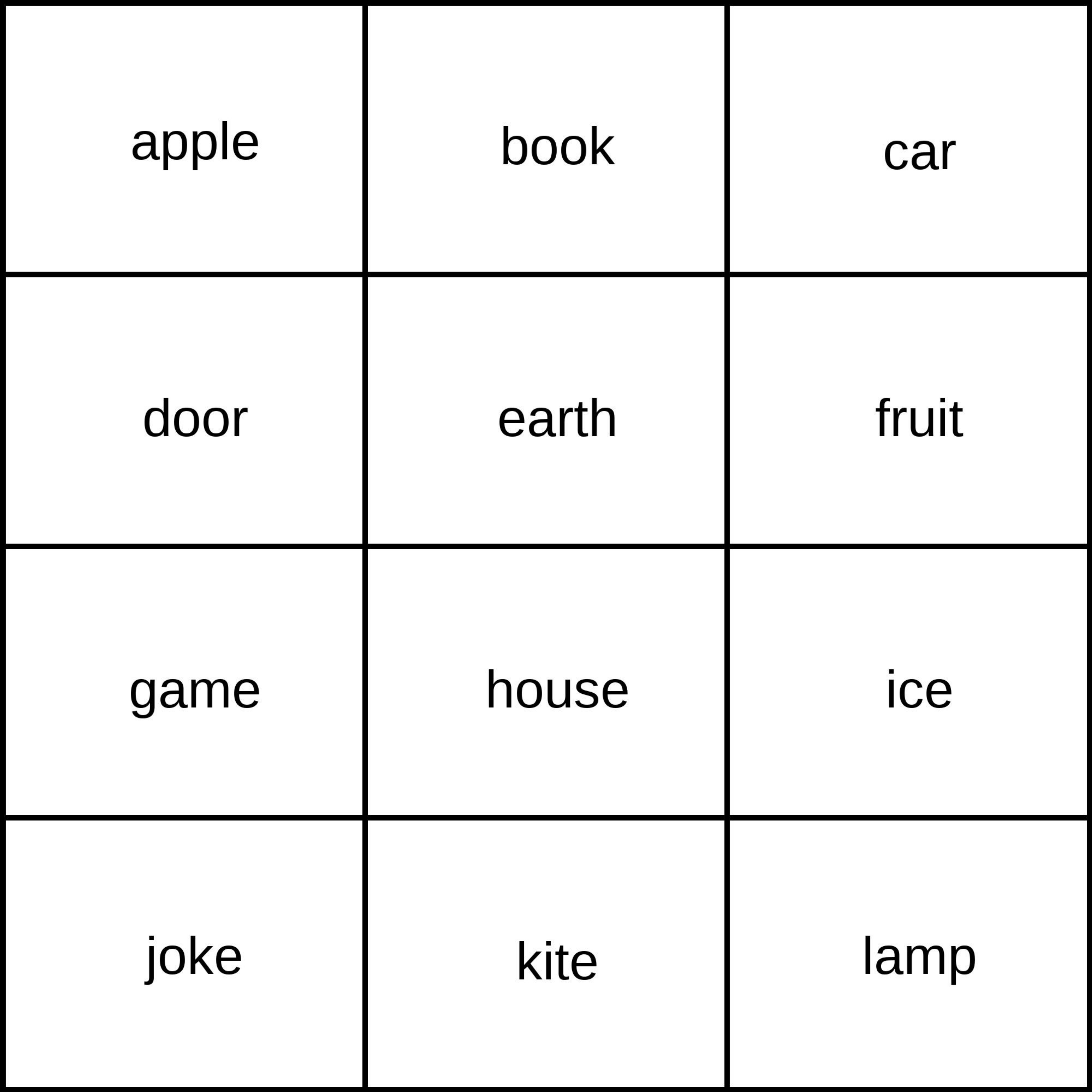} \\
         \includegraphics[width=0.33\textwidth]{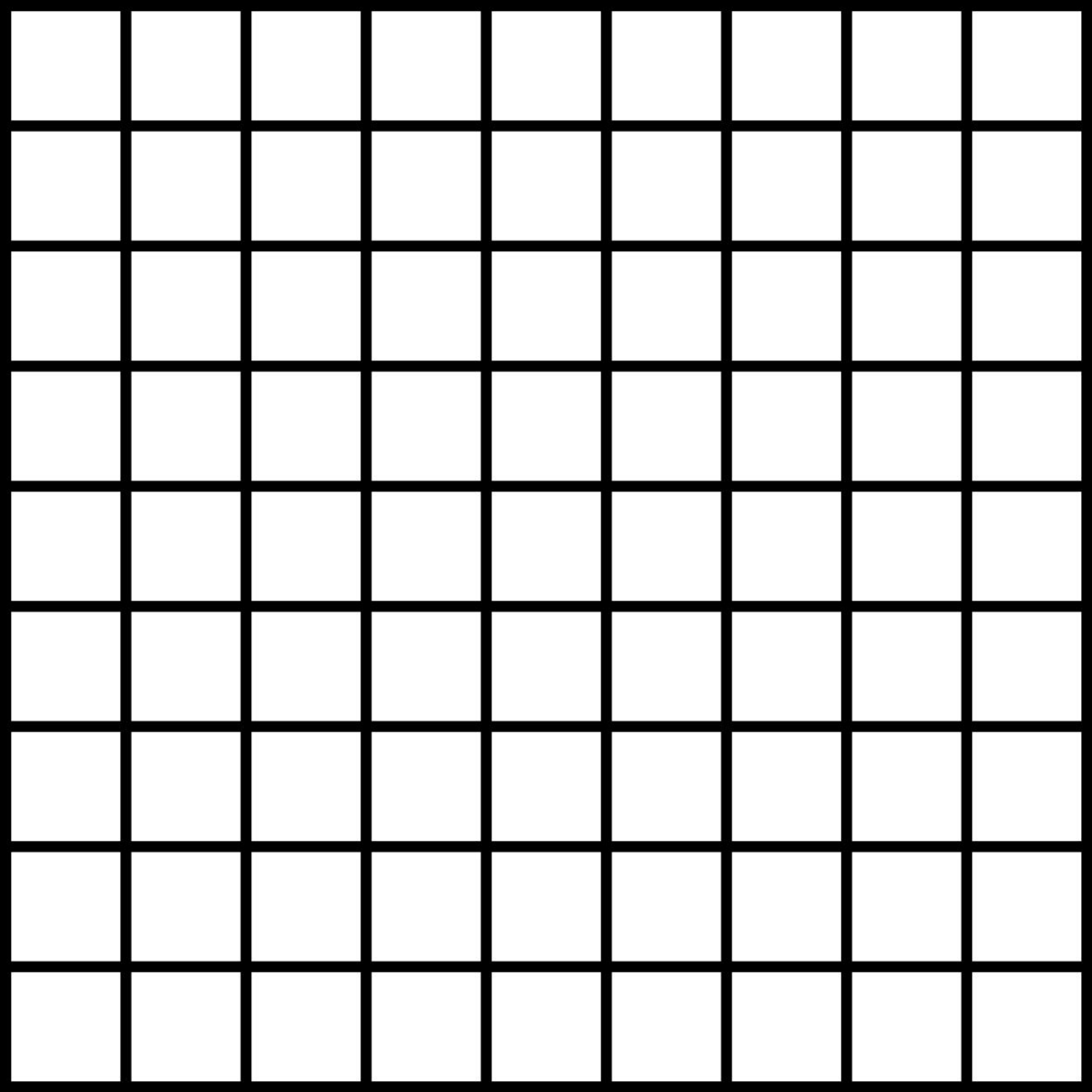}& \includegraphics[width=0.33\textwidth]{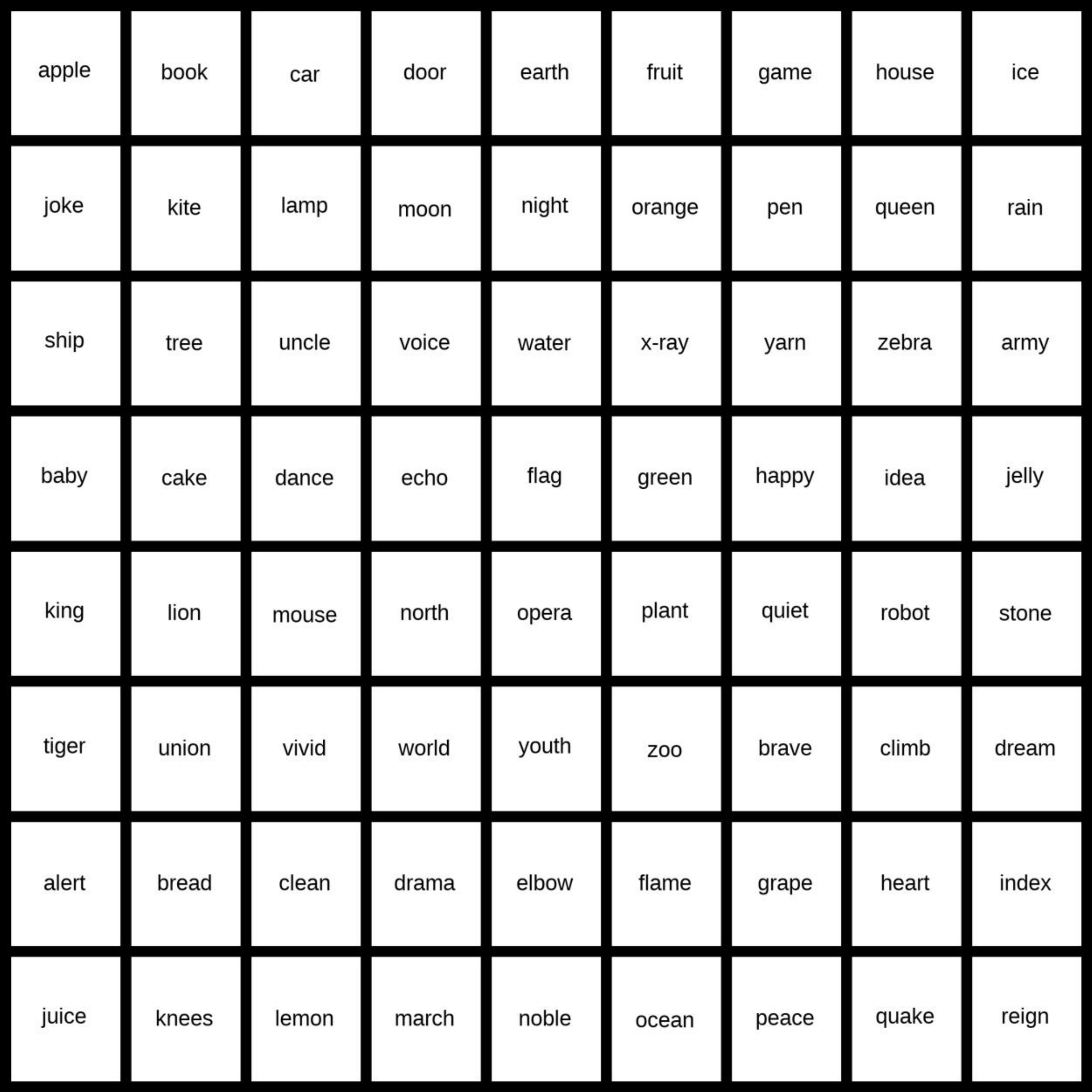} \\
         (a) blank & (b) text \\
    \end{tabular}
    \caption{We create grids with various sizes and line thicknesses. Each grid has a blank (a) and text (b) version.}
    \label{app:fig-grids-examples}
\end{figure}

We divide the image size by the number of rows and columns to find the coordinates for drawing the borderlines. Then, we draw the lines on the four edges of the image and draw the remaining lines in between. Our benchmark comprises 264 images of blank and text-containing grids (see \cref{table:grid-parameters,app:fig-grids-examples}). 

\textbf{Code} The code is available at \url{https://github.com/anguyen8/vision-llms-are-blind/blob/main/src/CountingRowsAndColumns/Grids.ipynb}.

\subsec{Prompts}
\begin{enumerate}
    \item \emph{Count the number of rows and columns and answer with numbers in curly brackets. For example, rows=\{5\} columns=\{6\}}

    \item \emph{How many rows and columns are in the table? Answer with only the numbers in a pair (row, column), e.g., (5,6)}.
\end{enumerate}

\noindent\textbf{Groundtruth} answers include both the number of rows and columns. 
An answer is correct when both column and row counts are correctly predicted (random-chance accuracy is 1/22, \ie, 4.55\%).

\begin{table}
\centering
\caption{\blindtest consists of 264 empty and text-containing grids. We create equal number of images for each subcategory.}
\label{table:grid-parameters}
\begin{tabular}{lccc}
\toprule
\textbf{Parameter} & \textbf{Values} & \textbf{Samples} & ~~\textbf{Total Samples}\\
\midrule
\multirow{2}{*}{Cell} & Empty & 132 & ~~\multirow{2}{*}{264}\\
& Text & 132 \\
\midrule
\multirow{3}{*}{Image size ($C$)} & 500px & 44 & ~~\multirow{3}{*}{132}\\
& 1250px & 44 \\
& 2000px & 44 \\
\midrule
\multirow{2}{*}{Line thickness} & $0.005\times C$ & 66 & ~~\multirow{2}{*}{132}\\
 & $0.01\times C$ & 66 \\
\midrule
\multirow{22}{*}{Dimensions} & $3\times 3$ & 6 & ~~\multirow{22}{*}{132}\\
 &  $3\times 4$ & 6 \\
 &  $4\times 3$ & 6 \\
&  $4\times 4$ & 6 \\
&  $4\times 5$ & 6 \\
&  $5\times 4$ & 6 \\
&  $5\times 5$ & 6 \\
&  $5\times 6$ & 6 \\
&  $6\times 5$ & 6 \\
&  $6\times 6$ & 6 \\
&  $6\times 7$ & 6 \\
&  $7\times 6$ & 6 \\
&  $7\times 7$ & 6 \\
&  $7\times 8$ & 6 \\
&  $8\times 7$ & 6 \\
&  $8\times 8$ & 6 \\
&  $8\times 9$ & 6 \\
&  $9\times 8$ & 6 \\
&  $9\times 9$ & 6 \\
&  $9\times 10$ & 6 \\
&  $10\times 9$ & 6 \\
&  $10\times 10$ & 6 \\
\bottomrule
\end{tabular}
\end{table}

\begin{figure}
    \centering
    \resizebox{\textwidth}{!}{
    \begin{tabular}{cccc}
\rotatebox{90}{\hspace{0.3cm}with text}&\includegraphics[width=0.4\textwidth]{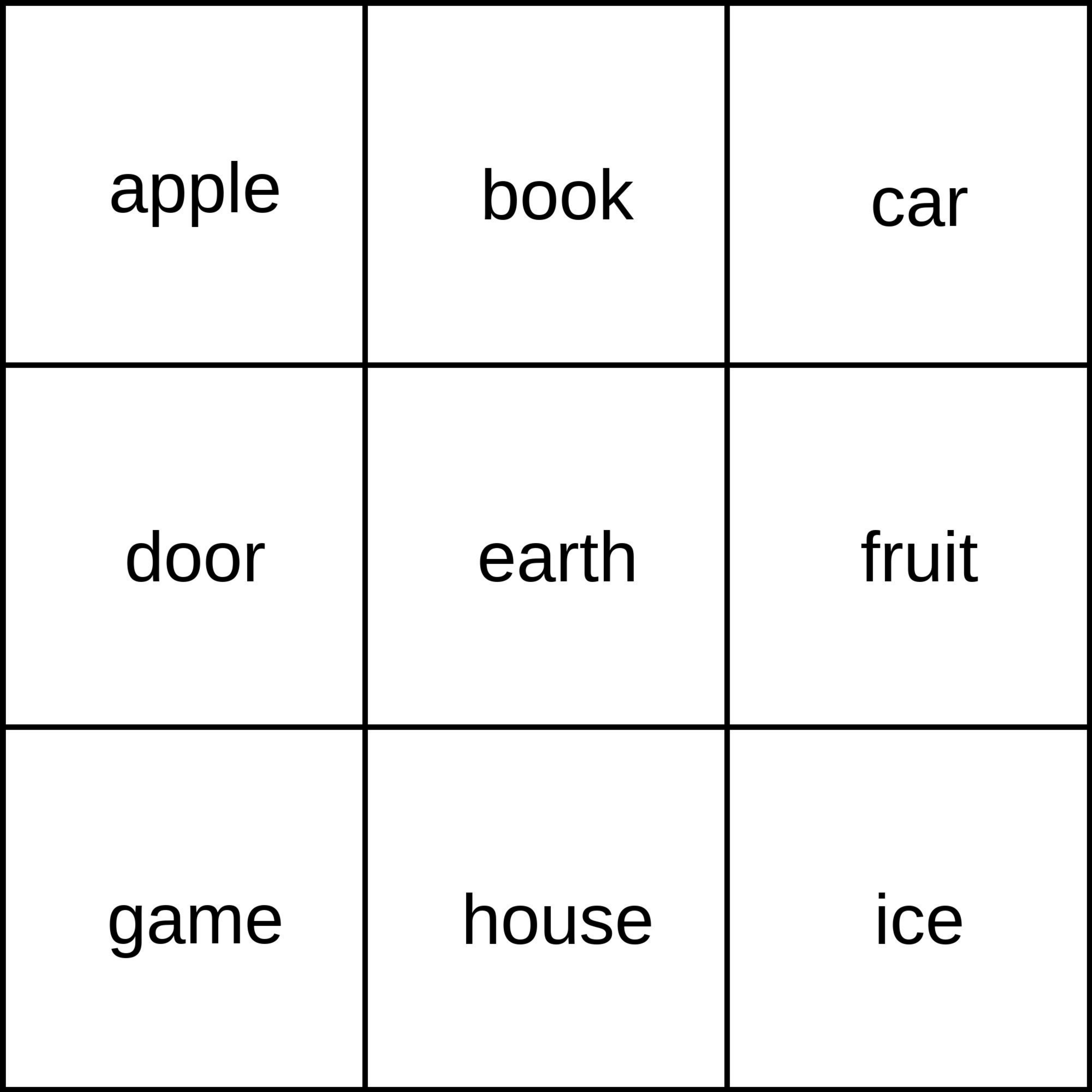}&\includegraphics[width=0.4\textwidth]{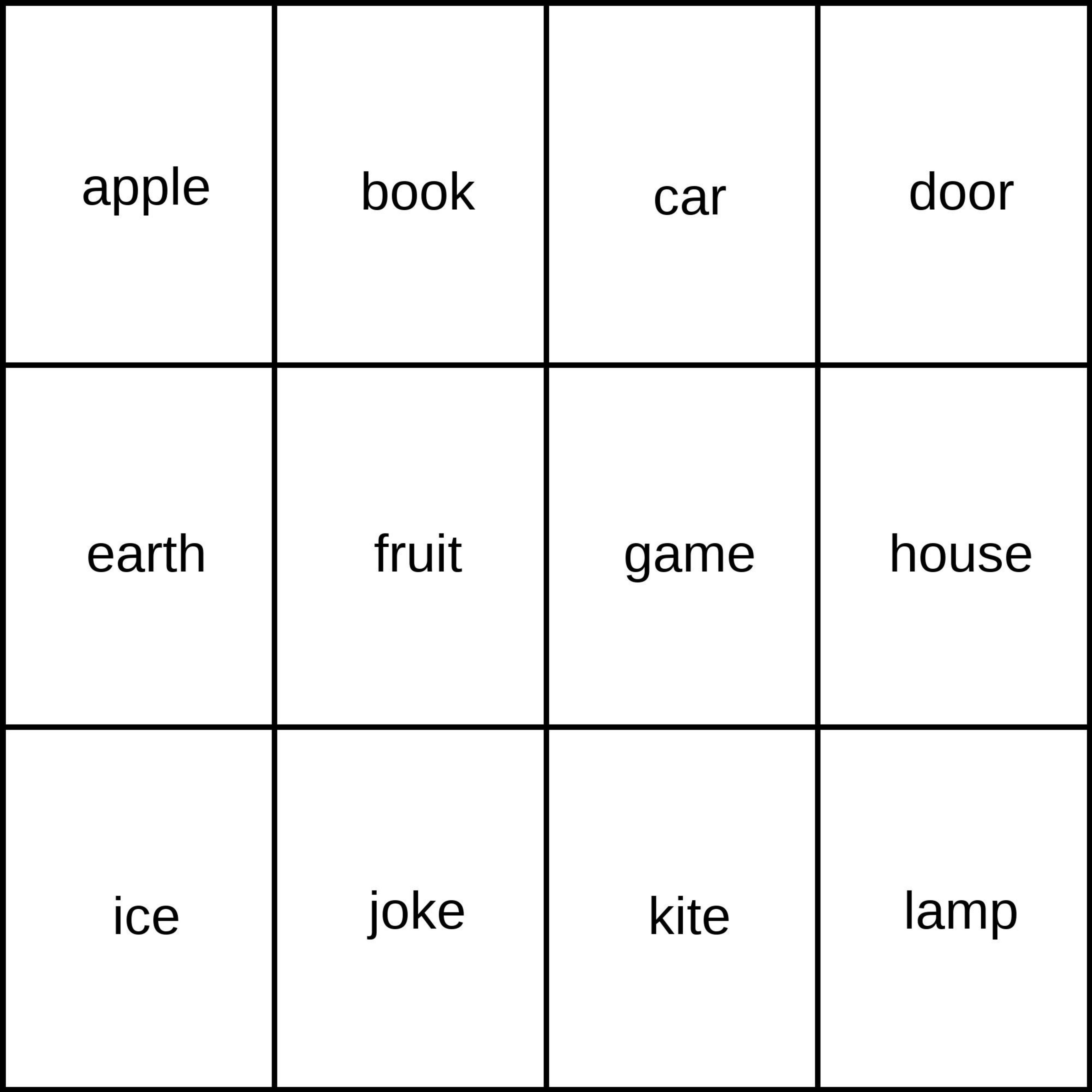}&\includegraphics[width=0.4\textwidth]{images/CountGridRowColumns/GridExamples/text_grid_4x3_2000_10.pdf}  \\
\rotatebox{90}{\hspace{0.55cm}empty}&\includegraphics[width=0.39\textwidth]{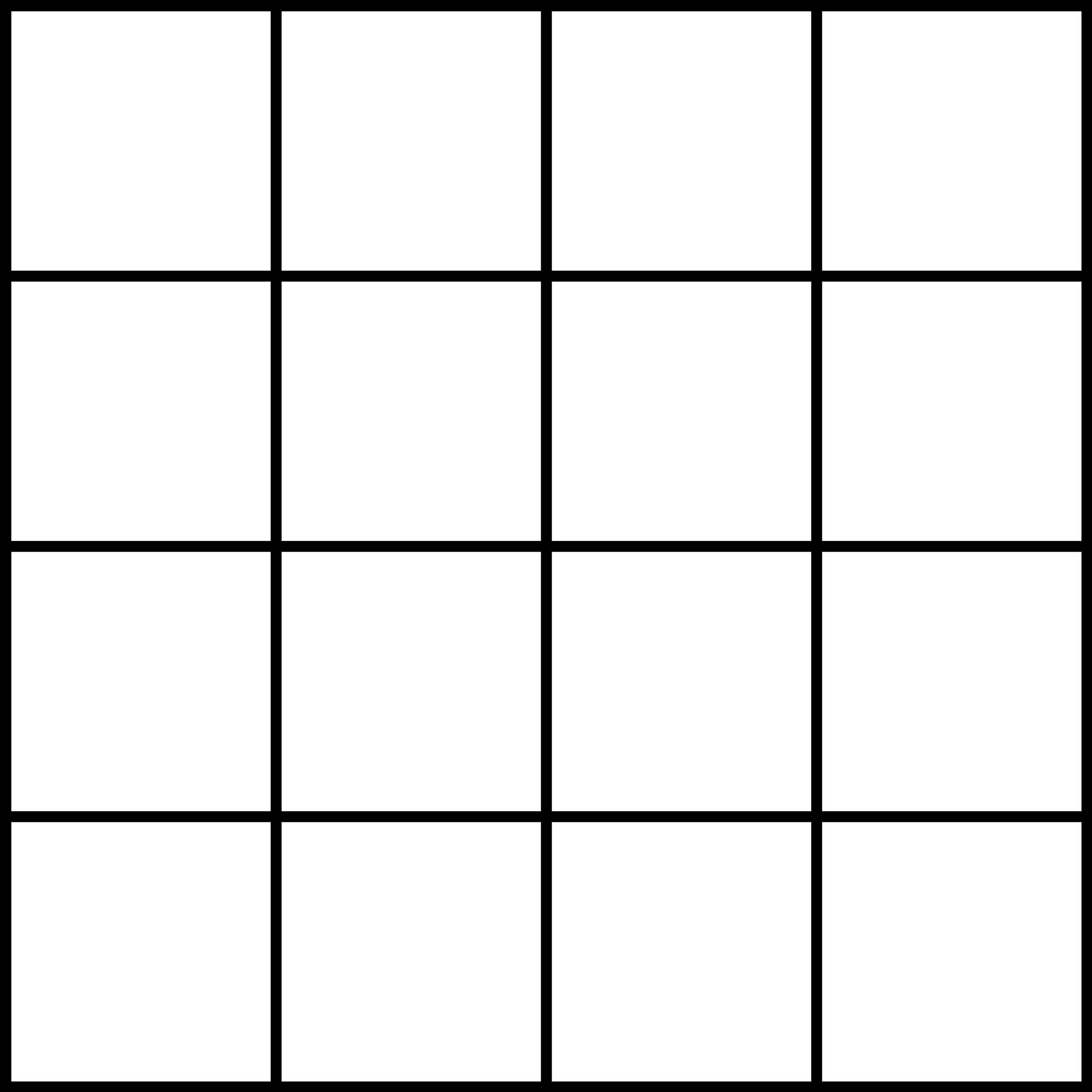} {(4,4)} &\includegraphics[width=0.39\textwidth]{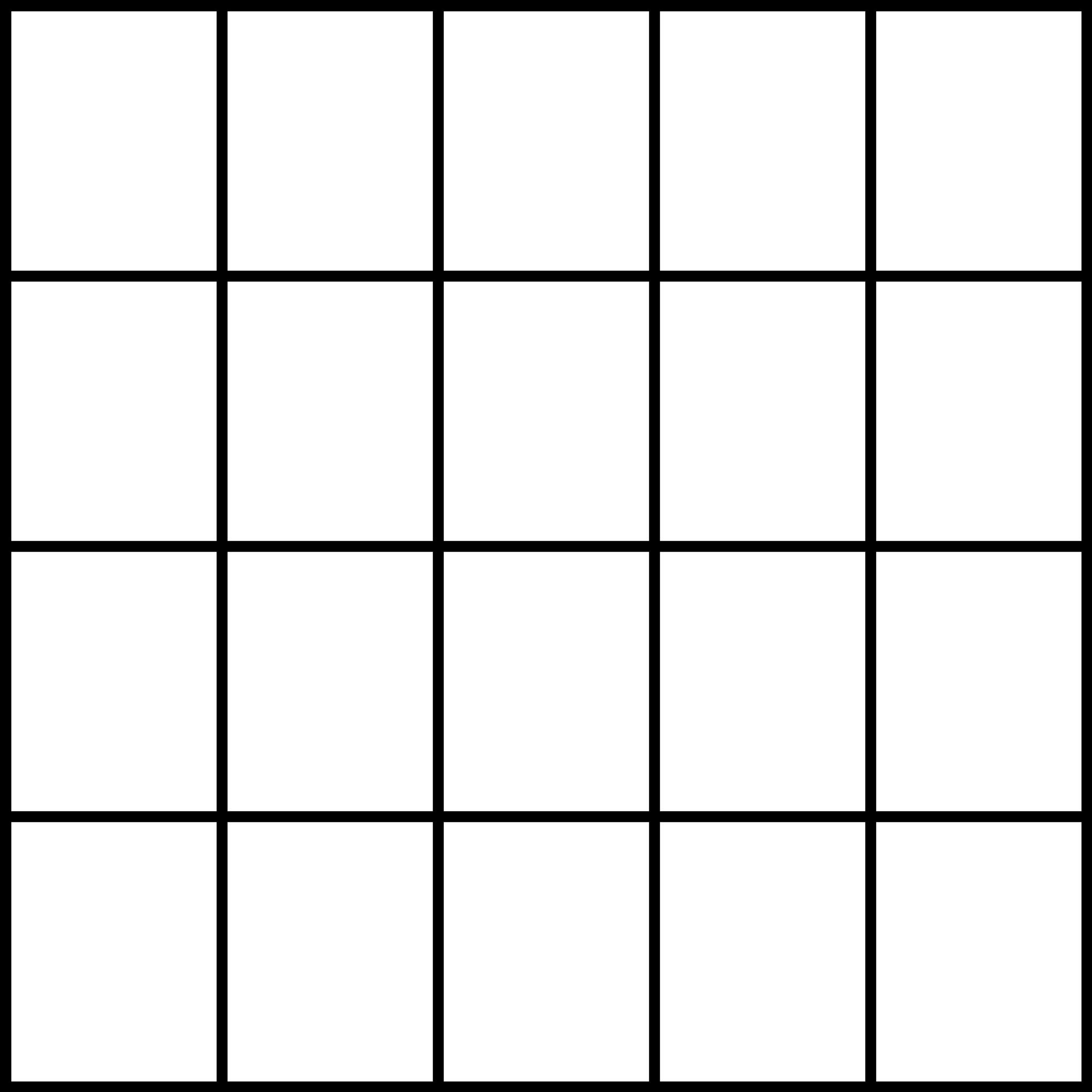} {(4,5)} &\includegraphics[width=0.39\textwidth]{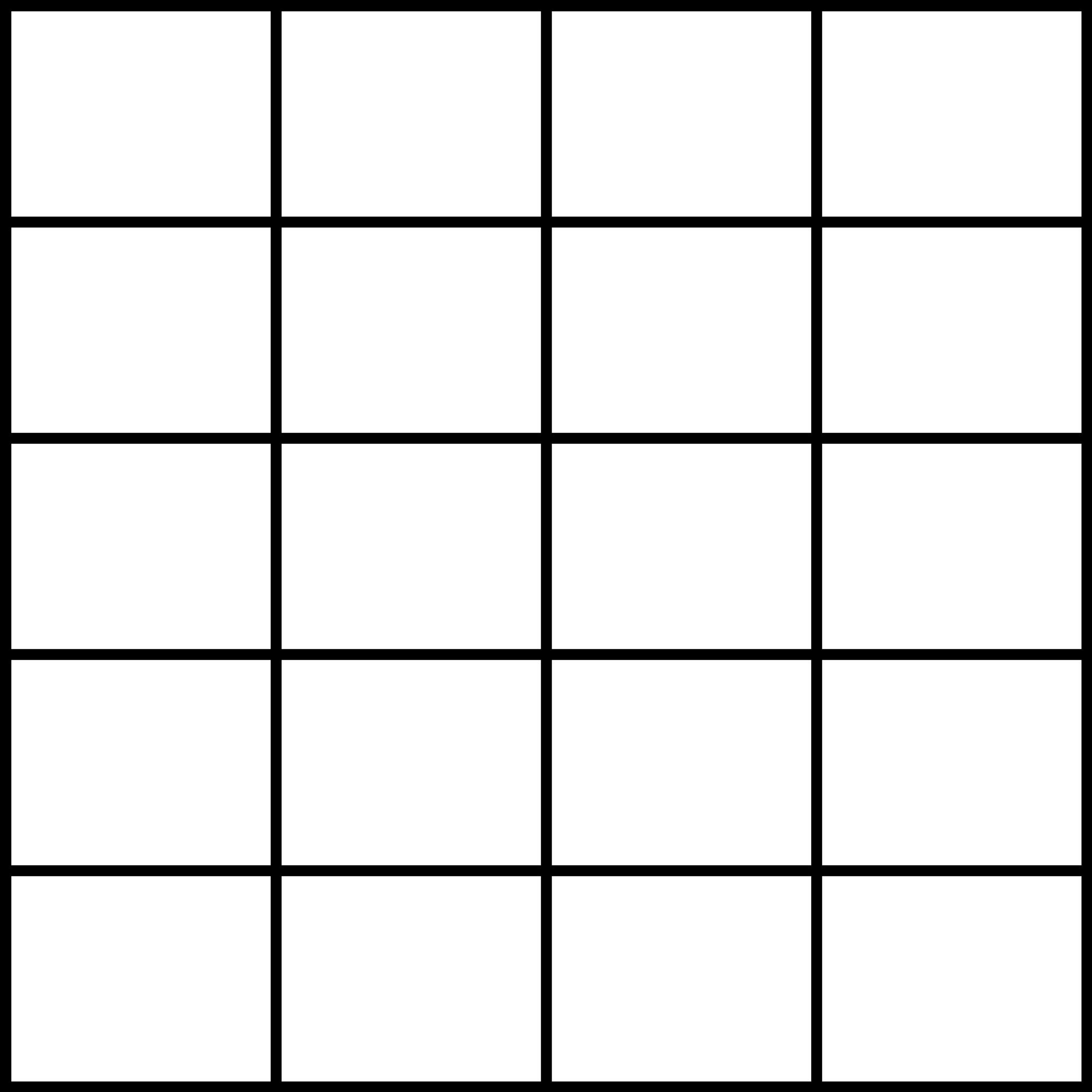} {(5,4)}\\
\end{tabular}}
\caption{Empty and text-containing grids are generated with various image sizes. 
}
\label{fig:grid-sub-examples}
\end{figure}

\clearpage

\subsection{Finding: VLMs cannot reliably count either rows or columns}
\label{app:grids-row}

We expect counting rows and columns separately to be hard for the VLMs, thus we analyze counting the rows and grids individually to see how VLMs perform. As shown in \cref{tab:app-grids}, VLMs cannot count either rows or columns alone. 

\begin{table}
    \centering
    \caption{Including text inside grids improves all model accuracies. 
\newsonnet, yet, outperforms other models on both empty and text-containing grids.}
    \resizebox{0.5\columnwidth}{!}{
    \begin{tabular}{lccccr}
    \toprule
    Grid & 
    ~~\raisebox{-.1\height}\gptlogo & ~~\raisebox{-.1\height}\geminilogo & ~~\raisebox{-.1\height}\sonnetlogo & ~~\raisebox{-.2\height}\newsonnetlogo & Mean \\
    \midrule
    Empty & ~~26.13 & ~~26.51 & ~~25.00 & ~~59.84 & 34.37\\
    Text & ~~\textbf{53.03} & ~~\textbf{52.27}& ~~\textbf{47.34} & ~~\textbf{88.68} & 60.33\\
    \midrule
    Mean & ~~39.58 & ~~39.39 & ~~36.17 & ~~74.26 &47.35\\
    \bottomrule
    \end{tabular}}
    \label{tab:overall_acc_grid}
\end{table}

\begin{table}
\centering
\caption{
Average row and column counting accuracy (\%) of VLMs. 
VLMs perform better at counting columns (70.53\% accuracy) than counting rows (60.83\%)---both of which are far from the expected 100\% accuracy.
}
\begin{tabular}{l@{\hskip 20pt}c@{\hskip 20pt}c@{\hskip 20pt}c@{\hskip 20pt}cr}
\toprule
Axis & \raisebox{-.1\height}\gptlogo & \raisebox{-.1\height}\geminilogo & \raisebox{-.1\height}\sonnetlogo & \raisebox{-.1\height}\newsonnetlogo & ~~Mean \\
\midrule
\text{Rows} & 65.54 & 52.95 & 42.19 & 82.64 & ~~60.83 \\
\text{Columns} & 58.42 & 58.50 & 74.65 & 90.54 & ~~70.53 \\
\bottomrule
\end{tabular}
\label{tab:app-grids}
\end{table}

\clearpage
\subsection{Additional Examples}
\begin{figure}[H]
\centering
\begin{AIbox}{Count the number of rows and columns and answer with numbers in curly brackets. For example{,} rows{=\{5\}} columns{=\{6\}}}
\centering

    \begin{tabular}{lp{0.8cm}c|p{0.8cm}c|p{0.8cm}c|p{0.8cm}c|p{0.8cm}c|p{0.8cm}c}
     &\multicolumn{2}{c}{\includegraphics[width=0.15\textwidth]{images/CountGridRowColumns/GridExamples/blank_grid_4x5_2000_20.pdf}} & \multicolumn{2}{c}{\includegraphics[width=0.15\textwidth]{images/CountGridRowColumns/GridExamples/blank_grid_6x7_2000_20.pdf}} &\multicolumn{2}{c}{\includegraphics[width=0.15\textwidth]{images/CountGridRowColumns/GridExamples/blank_grid_7x6_2000_20.pdf}} &\multicolumn{2}{c}{\includegraphics[width=0.15\textwidth]{images/CountGridRowColumns/GridExamples/blank_grid_8x7_1250_20.pdf}} &\multicolumn{2}{c}{\includegraphics[width=0.15\textwidth]{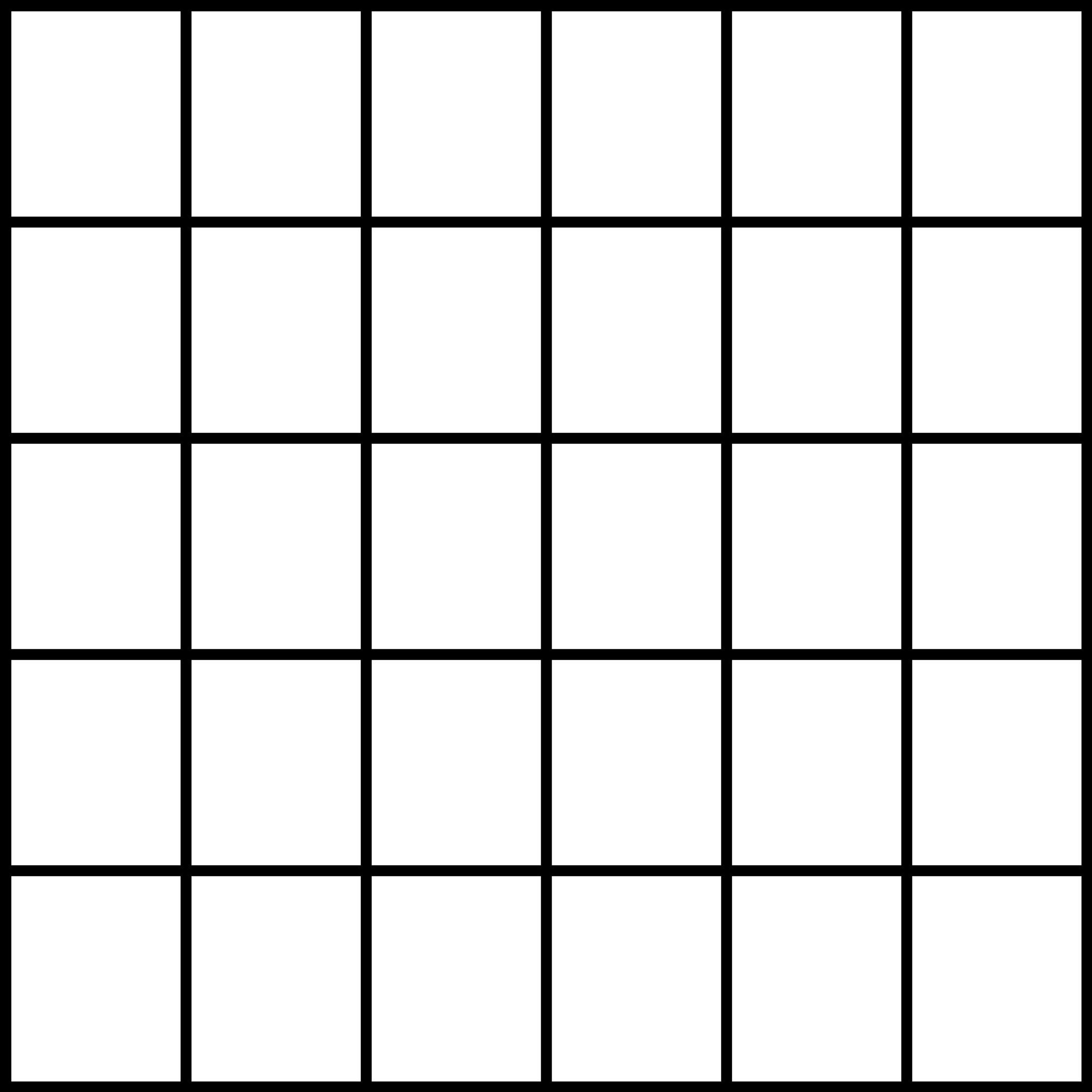}}&\multicolumn{2}{c}{\includegraphics[width=0.15\textwidth]{images/CountGridRowColumns/GridExamples/blank_grid_7x8_2000_10.pdf}}\\
     {GT} & \multicolumn{2}{c}{\centering 4$\times$5} &  \multicolumn{2}{c}{\centering 6$\times$7}&  \multicolumn{2}{c}{\centering 7$\times$6} &  \multicolumn{2}{c}{\centering 8$\times$7} & \multicolumn{2}{c}{\centering 5$\times$6}  &  \multicolumn{2}{c}{\centering 7$\times$8} \\
     \rowcolor{lightgray}
     \raisebox{-0.2\height}\gptlogo & \centering 4$\times$4 & \textcolor{red}{\xmark} &  \centering 6$\times$6 & \textcolor{red}{\xmark} &  \centering 7$\times$7 & \textcolor{red}{\xmark} &  \centering 6$\times$6 & \textcolor{red}{\xmark} &  \centering 6$\times$6 & \textcolor{red}{\xmark}&  \centering 6$\times$6 & \textcolor{red}{\xmark}\\
     \raisebox{-0.2\height}\geminilogo & \centering 5$\times$5 & \textcolor{red}{\xmark} & \centering 6$\times$6 & \textcolor{red}{\xmark}&  \centering 7$\times$7 & \textcolor{red}{\xmark} &  \centering 10$\times$10 & \textcolor{red}{\xmark} &  \centering 5$\times$6 & \textcolor{ForestGreen}{\cmark}&  \centering 10$\times$10 & \textcolor{red}{\xmark}\\
     \rowcolor{lightgray}
     \raisebox{-0.2\height}\sonnetlogo & \centering 5$\times$5 & \textcolor{red}{\xmark} & \centering 7$\times$8 & \textcolor{red}{\xmark}&  \centering 6$\times$6 & \textcolor{red}{\xmark} &  \centering 9$\times$9 & \textcolor{red}{\xmark} &  \centering 6$\times$6 & \textcolor{red}{\xmark}&  \centering 9$\times$12 & \textcolor{red}{\xmark}\\
     \raisebox{-0.2\height}\newsonnetlogo & \centering 4$\times$5 & \textcolor{ForestGreen}{\cmark} & \centering 6$\times$7 & \textcolor{ForestGreen}{\cmark}&  \centering 7$\times$7 & \textcolor{red}{\xmark} &  \centering 8$\times$7 & \textcolor{ForestGreen}{\cmark} &  \centering 5$\times$6 & \textcolor{ForestGreen}{\cmark}&  \centering 8$\times$8 & \textcolor{red}{\xmark}\\
     \end{tabular}

    \vspace{4pt}
    \centering
    \begin{tabular}{cccccccccccccc}
    \raisebox{-0.1\height}\gptlogo & \gpt &&&
      \raisebox{-0.1\height}\geminilogo & \gemini & &&\raisebox{-0.1\height}\sonnetlogo & \sonnet & && \raisebox{-0.12\height}\newsonnetlogo & \newsonnet \\
      \end{tabular}
     
\end{AIbox}
\caption{Examples from the benchmark show that models consistently fail at counting rows and columns of empty grids. \newsonnet, however, is more consistent in the lower number of rows and columns.}
\label{fig:qual_grid1}
\end{figure}

\begin{figure}[ht]
\centering
\begin{AIbox}{How many rows and columns are in the table? Answer with only the numbers in a pair (row{,} column){,} e.g.{,} (5{,}6).}
\centering

    \begin{tabular}{lp{0.8cm}c|p{0.8cm}c|p{0.8cm}c|p{0.8cm}c|p{0.8cm}c|p{0.8cm}c}
     &\multicolumn{2}{c}{\includegraphics[width=0.15\textwidth]{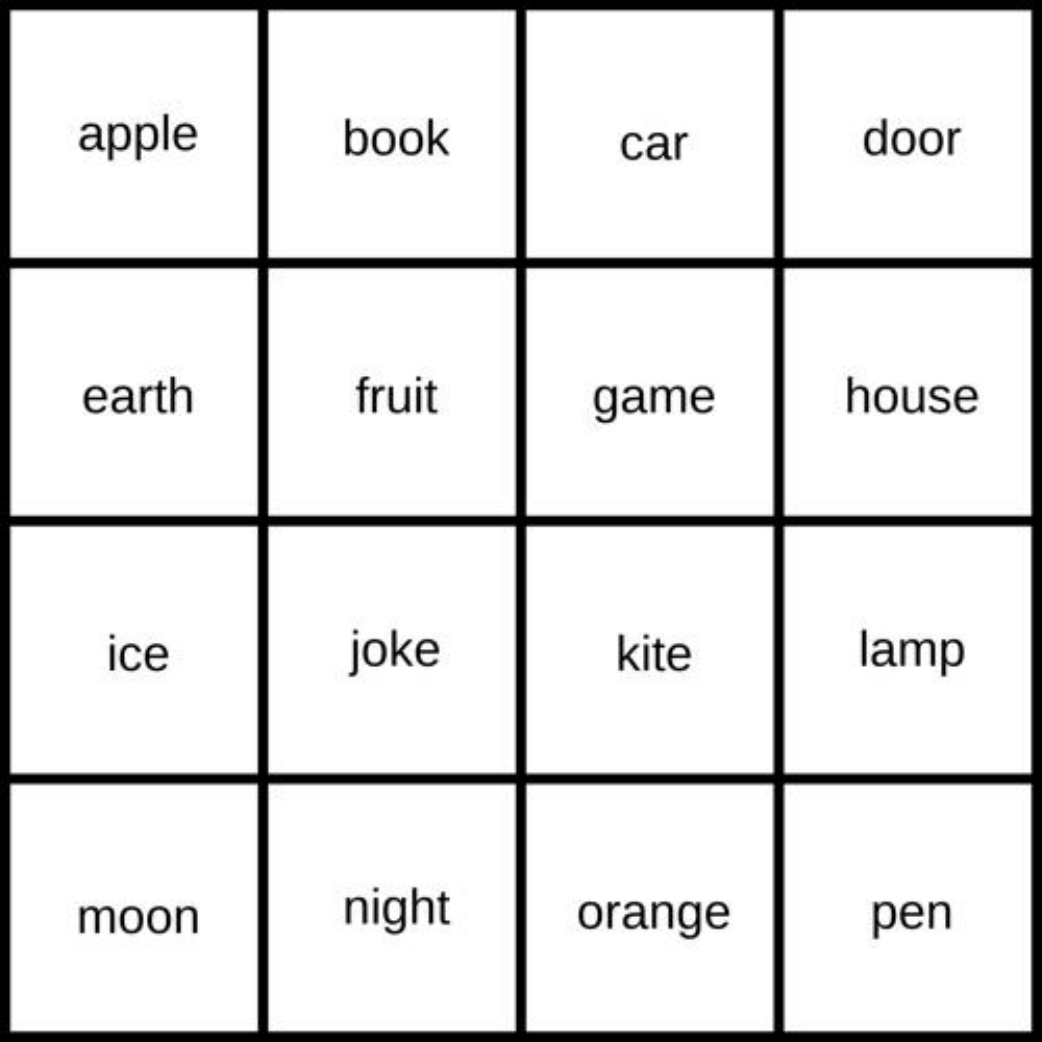}} & \multicolumn{2}{c}{\includegraphics[width=0.15\textwidth]{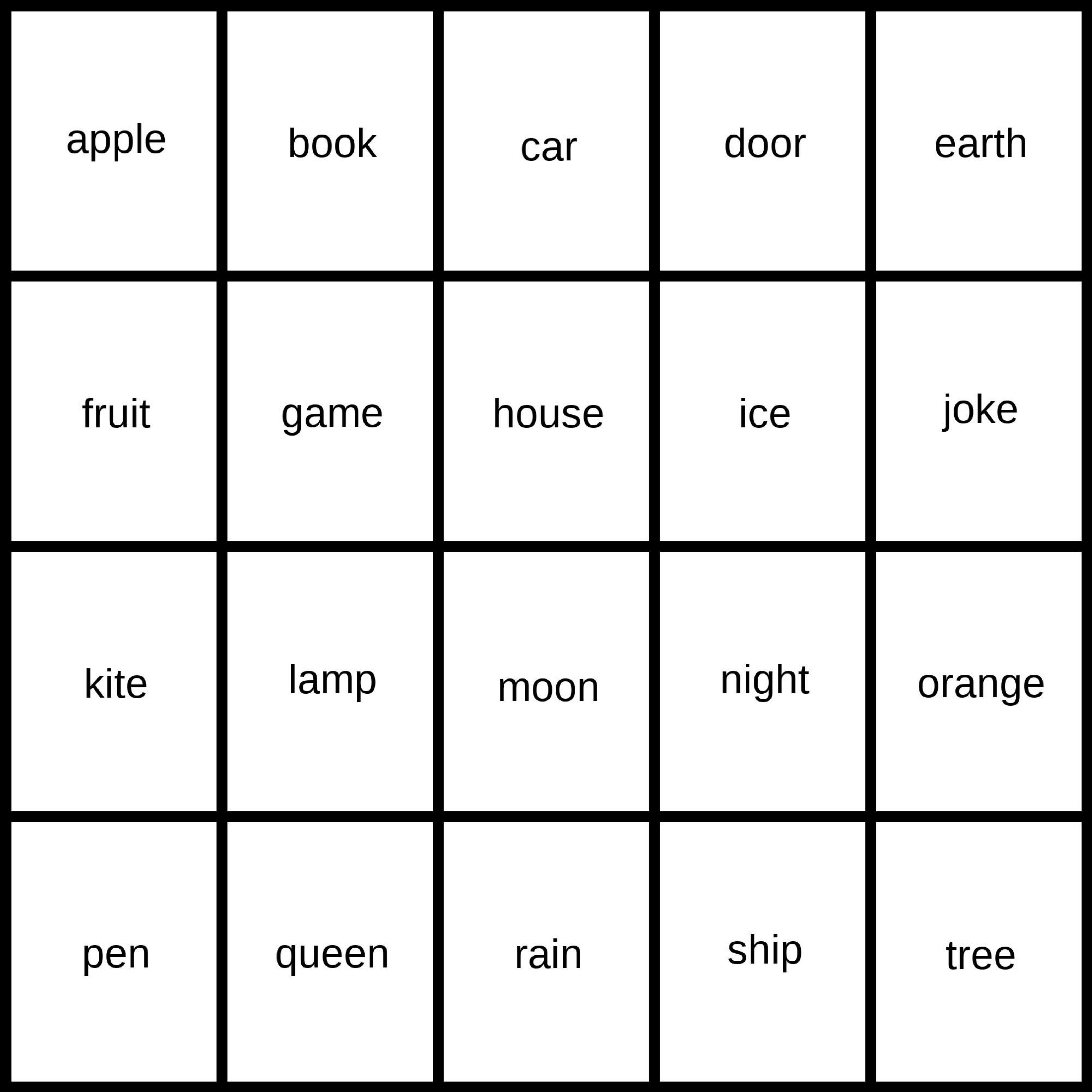}} &\multicolumn{2}{c}{\includegraphics[width=0.15\textwidth]{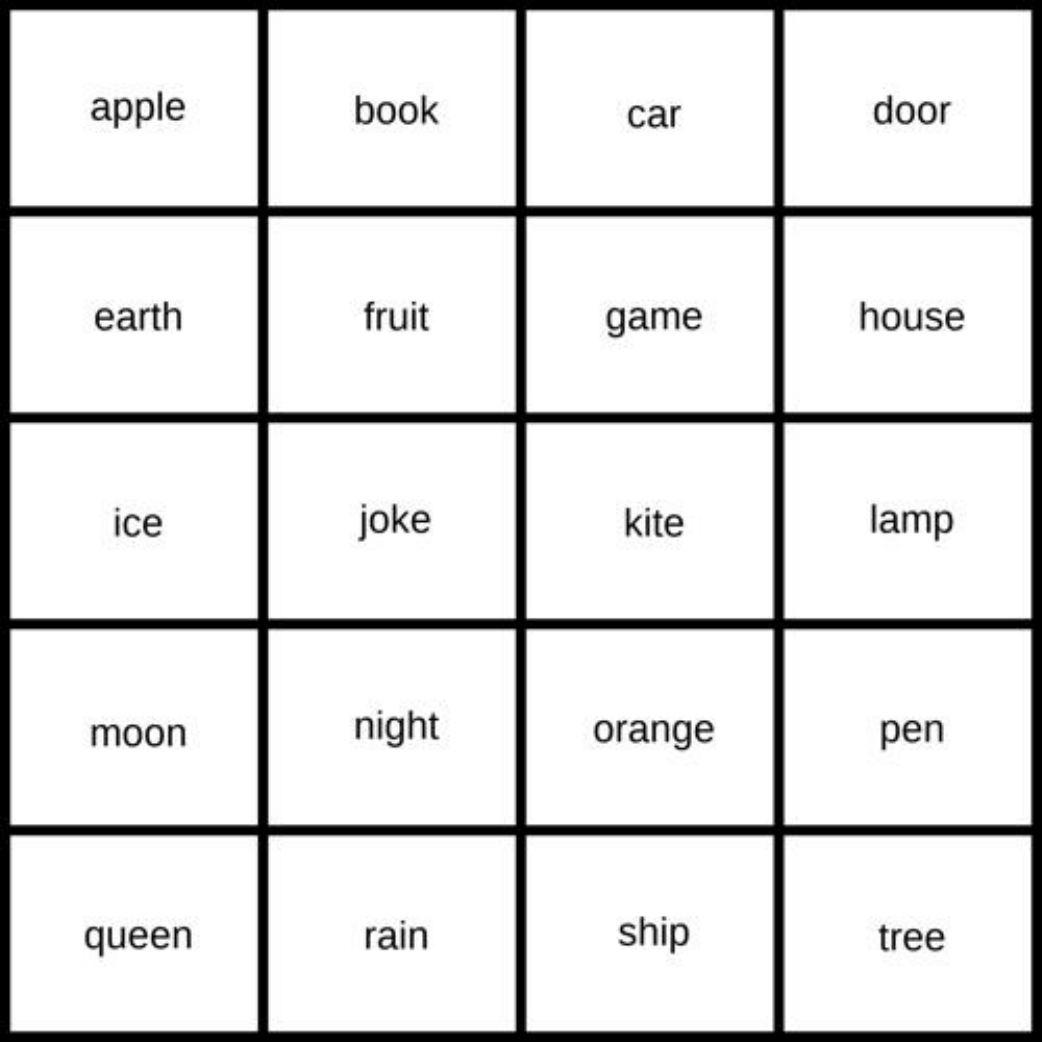}} &\multicolumn{2}{c}{\includegraphics[width=0.15\textwidth]{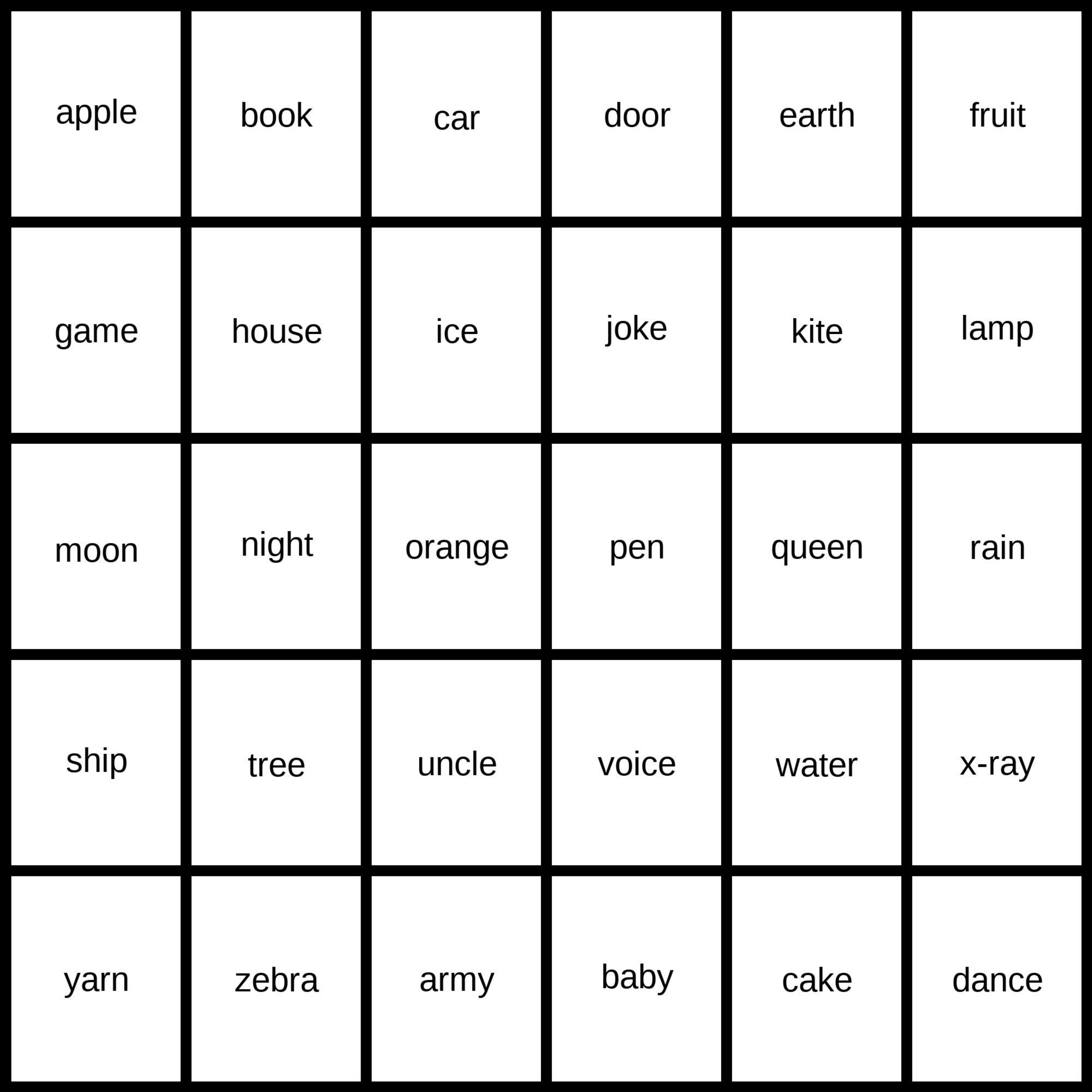}} &\multicolumn{2}{c}{\includegraphics[width=0.15\textwidth]{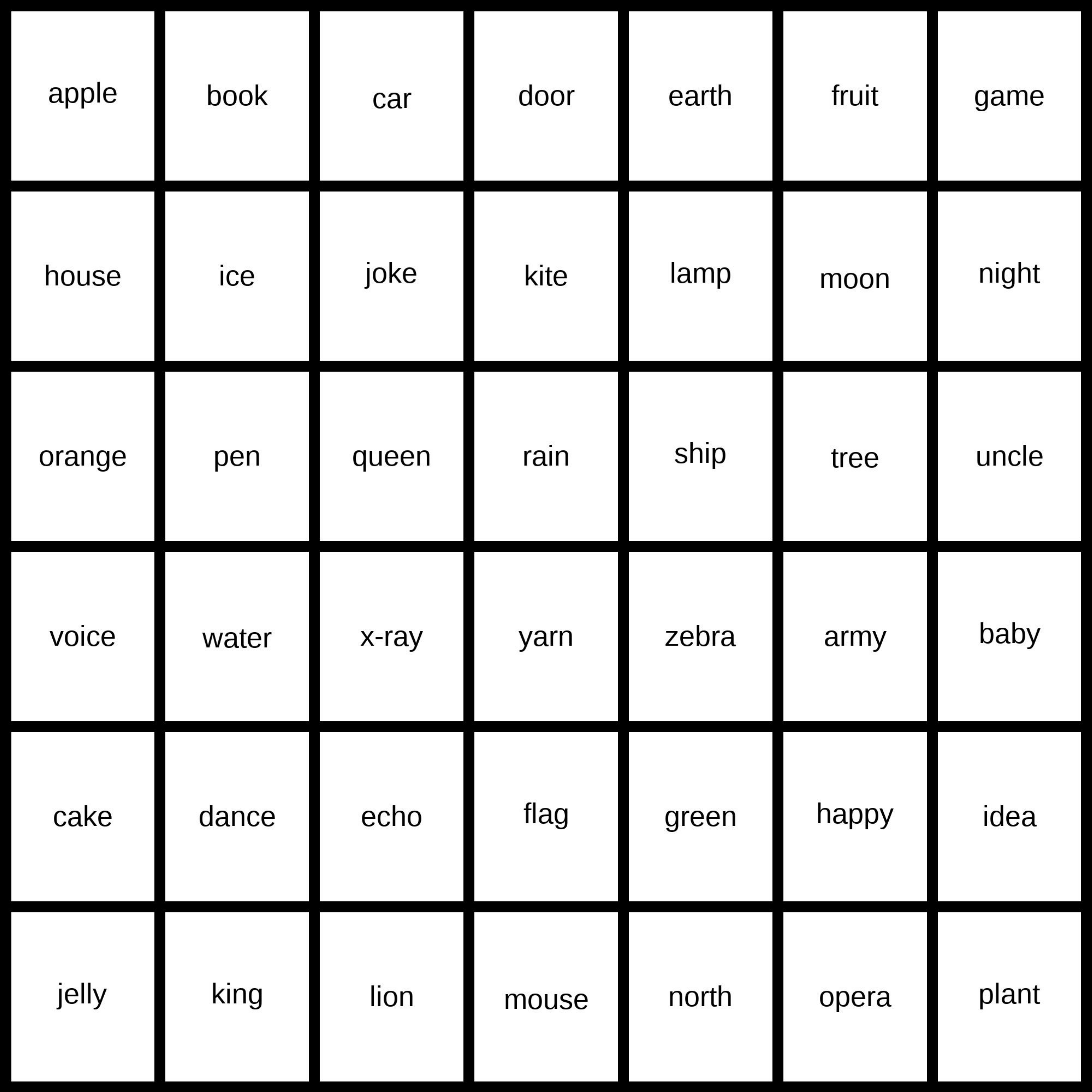}}&\multicolumn{2}{c}{\includegraphics[width=0.15\textwidth]{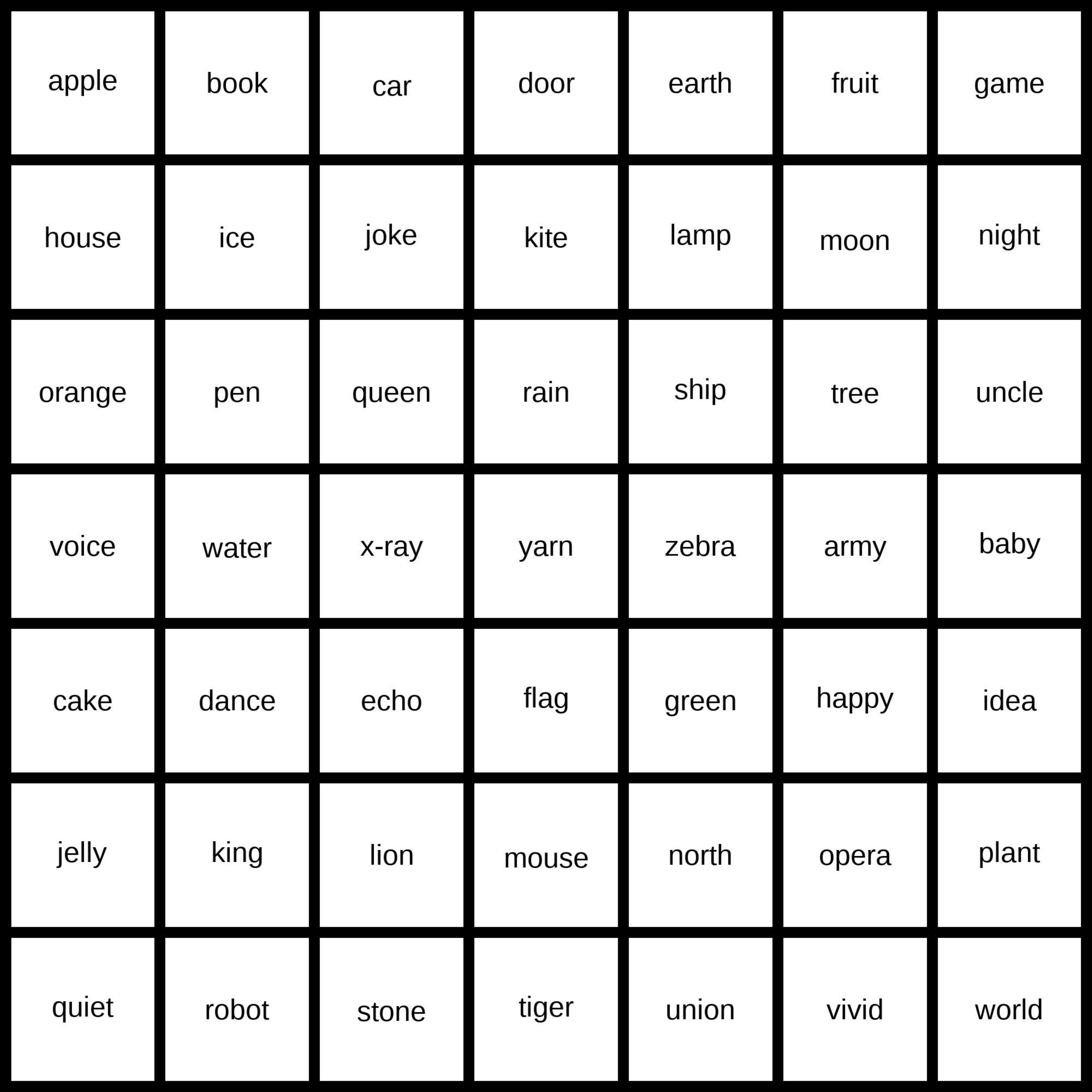}}\\
     {GT} & \multicolumn{2}{c}{\centering 4$\times$4} &  \multicolumn{2}{c}{\centering 4$\times$5}&  \multicolumn{2}{c}{\centering 5$\times$4} &  \multicolumn{2}{c}{\centering 5$\times$6} & \multicolumn{2}{c}{\centering 6$\times$7}  &  \multicolumn{2}{c}{\centering 7$\times$7} \\
     \rowcolor{lightgray}
     \raisebox{-0.2\height}\gptlogo & \centering 4$\times$4 & \textcolor{ForestGreen}{\cmark} &  \centering 4$\times$5 & \textcolor{ForestGreen}{\cmark}&  \centering 5$\times$4 & \textcolor{ForestGreen}{\cmark} &  \centering 5$\times$6 & \textcolor{ForestGreen}{\cmark} &  \centering 6$\times$8 & \textcolor{red}{\xmark} & \centering 7$\times$8 & \textcolor{red}{\xmark}\\
     \raisebox{-0.2\height}\geminilogo & \centering 4$\times$4 & \textcolor{ForestGreen}{\cmark} & \centering 4$\times$5 & \textcolor{ForestGreen}{\cmark}&  \centering 5$\times$4 & \textcolor{ForestGreen}{\cmark} &  \centering 5$\times$6 & \textcolor{ForestGreen}{\cmark} &  \centering 6$\times$8 & \textcolor{red}{\xmark} & \centering 7$\times$8 & \textcolor{red}{\xmark}\\
     \rowcolor{lightgray}
     \raisebox{-0.2\height}\sonnetlogo & \centering 4$\times$4 & \textcolor{ForestGreen}{\cmark} & \centering 5$\times$5 & \textcolor{red}{\xmark}&  \centering 5$\times$4 & \textcolor{ForestGreen}{\cmark} &  \centering 6$\times$6 & \textcolor{red}{\xmark} &  \centering 7$\times$7 & \textcolor{red}{\xmark}&  \centering 8$\times$7 & \textcolor{red}{\xmark}\\
     \raisebox{-0.2\height}\newsonnetlogo & \centering 4$\times$4 & \textcolor{ForestGreen}{\cmark} & \centering 4$\times$5 & \textcolor{ForestGreen}{\cmark}&  \centering 5$\times$4 & \textcolor{ForestGreen}{\cmark} &  \centering 5$\times$6 & \textcolor{ForestGreen}{\cmark} &  \centering 6$\times$7 & \textcolor{ForestGreen}{\cmark}&  \centering 7$\times$7 & \textcolor{ForestGreen}{\cmark}\\
     \end{tabular}

    \vspace{4pt}
    \centering
    \begin{tabular}{cccccccccccccc}
    \raisebox{-0.1\height}\gptlogo & \gpt &&&
      \raisebox{-0.1\height}\geminilogo & \gemini & &&\raisebox{-0.1\height}\sonnetlogo & \sonnet & && \raisebox{-0.12\height}\newsonnetlogo & \newsonnet \\
      \end{tabular}
     
\end{AIbox}
\caption{When text is included in the cells of the grid, the performance of all VLMs improves, especially \newsonnet.}
\label{fig:qual_grid2}
\end{figure}

We show examples of models' responses to the counting the number of rows and columns task in \cref{fig:qual_grid1,fig:qual_grid2}.
\label{app:grids-qual}



\begin{figure}
\centering
\begin{AIbox}{Count the number of rows and columns and answer with numbers in curly brackets. For example{,} rows{=\{5\}} columns{=\{6\}.}}
\centering
    \begin{tabular}{lp{0.8cm}c|p{0.8cm}c|p{0.8cm}c|p{0.8cm}c|p{0.8cm}c|p{0.8cm}c}
     &\multicolumn{2}{c}{\includegraphics[width=0.15\textwidth]{images/CountGridRowColumns/GridExamples/blank_grid_4x5_2000_20.pdf}} & \multicolumn{2}{c}{\includegraphics[width=0.15\textwidth]{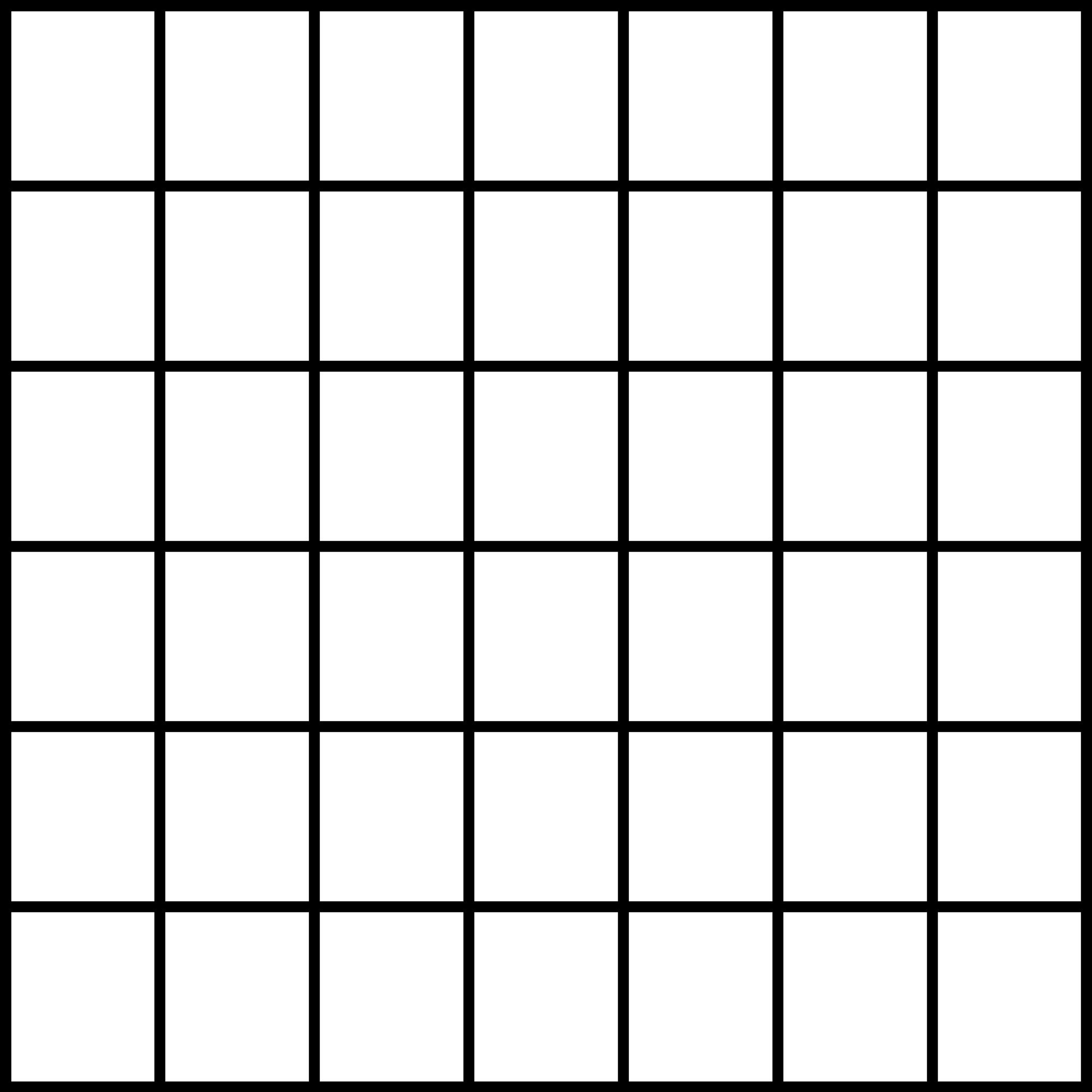}} &\multicolumn{2}{c}{\includegraphics[width=0.15\textwidth]{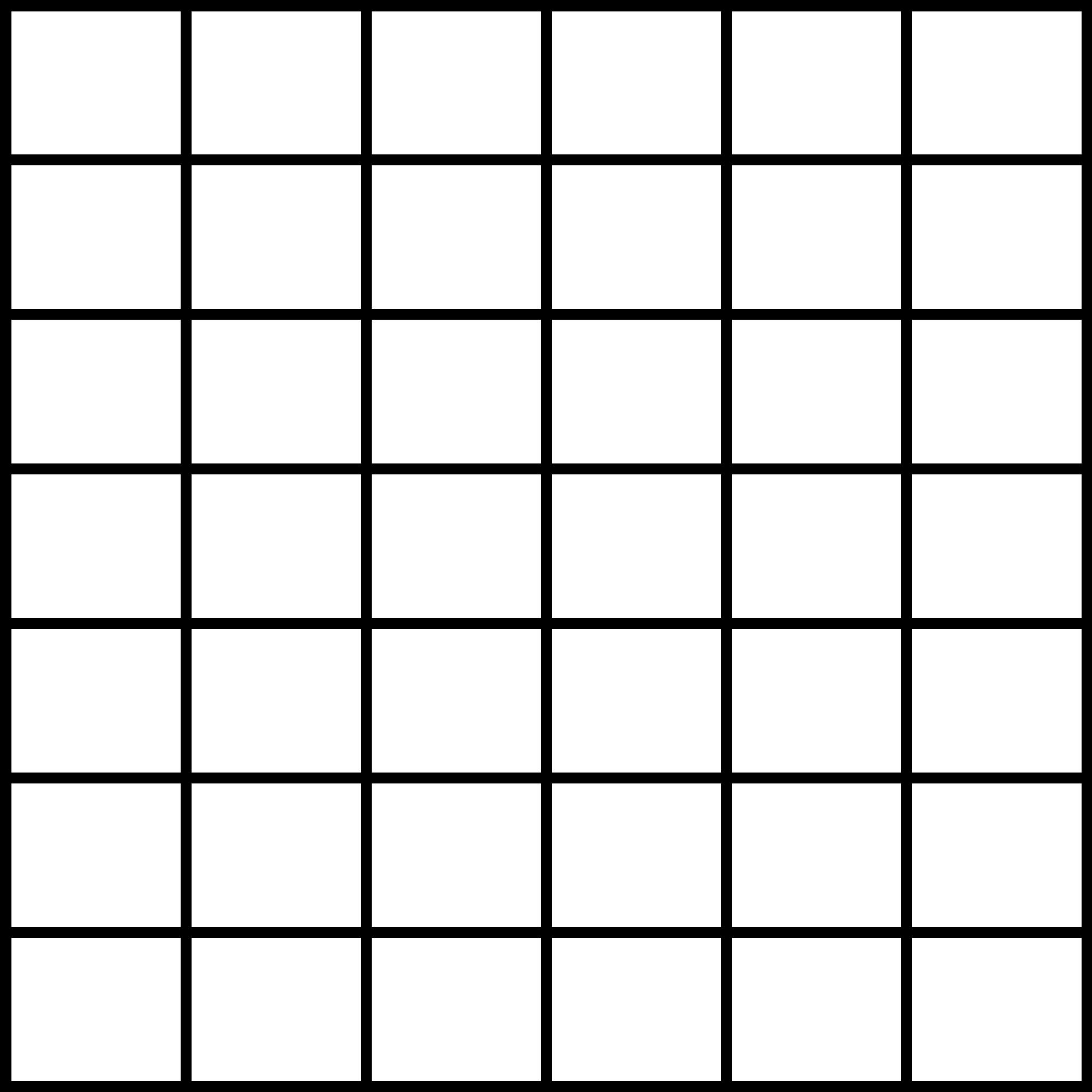}} &\multicolumn{2}{c}{\includegraphics[width=0.15\textwidth]{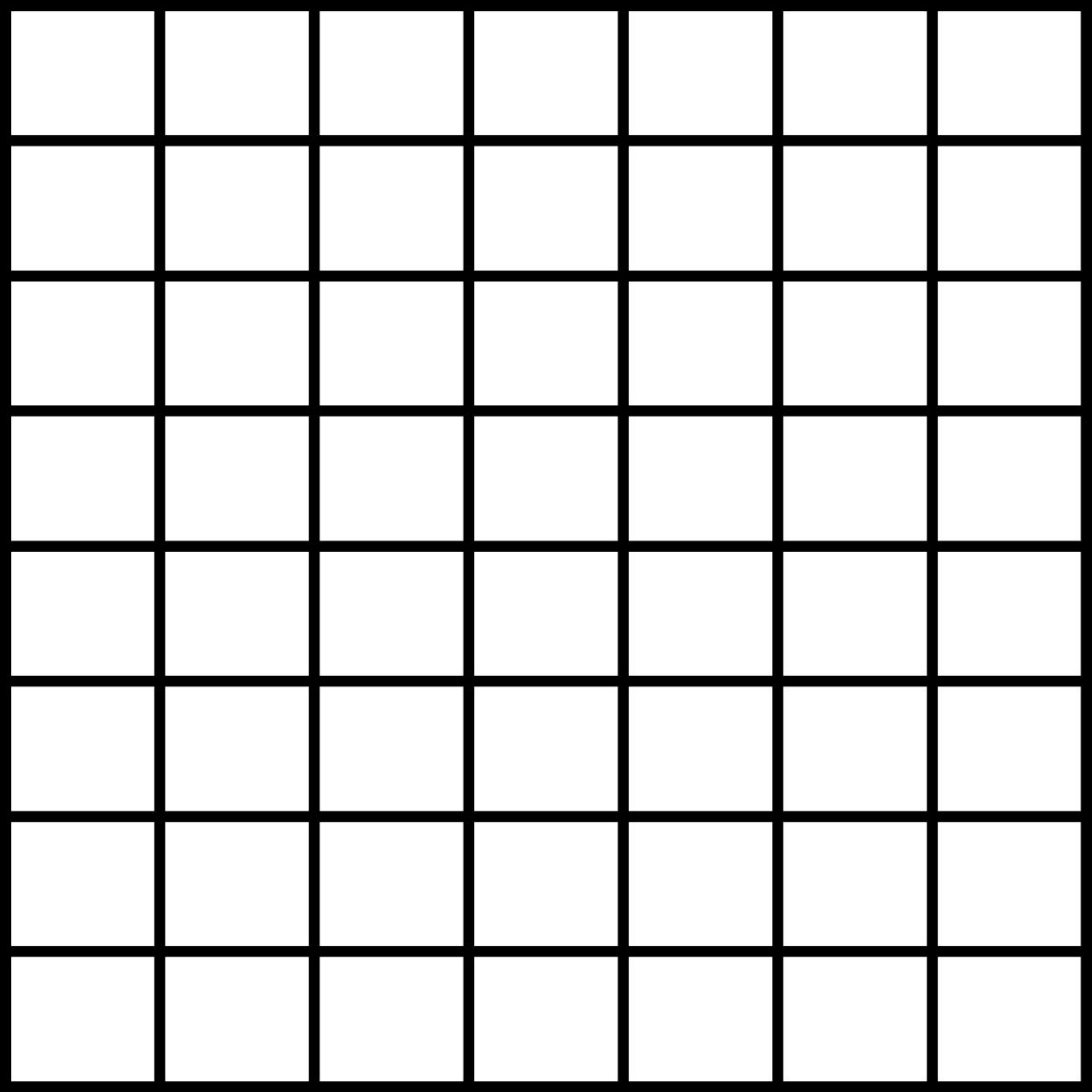}} &\multicolumn{2}{c}{\includegraphics[width=0.15\textwidth]{images/CountGridRowColumns/GridExamples/text_grid_3x4_500_20.pdf}}&\multicolumn{2}{c}{\includegraphics[width=0.15\textwidth]{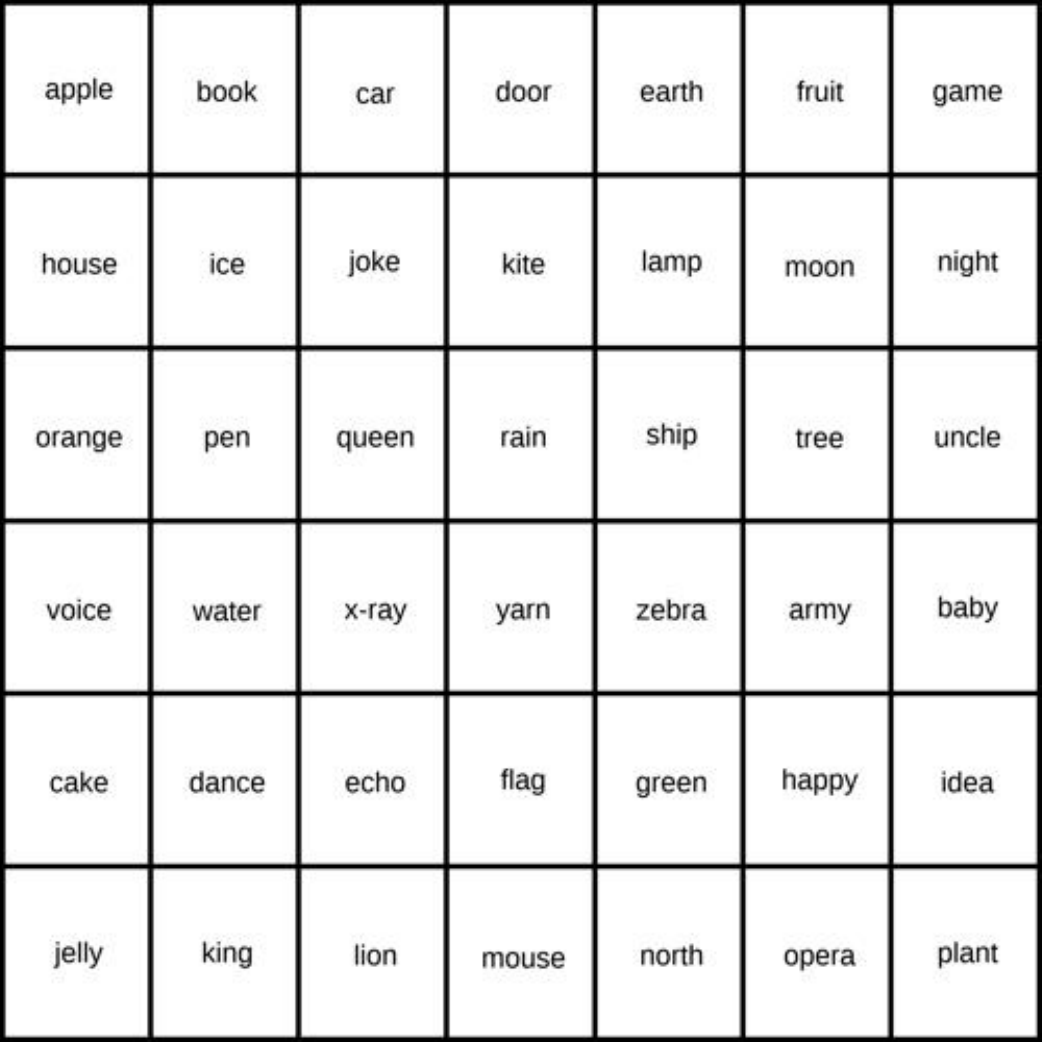}}\\
     {GT} & \multicolumn{2}{c}{\centering 4$\times$5} &  \multicolumn{2}{c}{\centering 6$\times$7}&  \multicolumn{2}{c}{\centering 7$\times$6} &  \multicolumn{2}{c}{\centering 8$\times$7} & \multicolumn{2}{c}{\centering 3$\times$4}  &  \multicolumn{2}{c}{\centering 6$\times$7} \\
     \rowcolor{lightgray}
     \raisebox{-0.2\height}\gptlogo & \centering 4$\times$4 & \textcolor{red}{\xmark} &  \centering 6$\times$6 & \textcolor{red}{\xmark} &  \centering 7$\times$7 & \textcolor{red}{\xmark} &  \centering 6$\times$6 & \textcolor{red}{\xmark} &  \centering 3$\times$4 & \textcolor{ForestGreen}{\cmark}&  \centering 7$\times$7 & \textcolor{red}{\xmark}\\
     \raisebox{-0.2\height}\geminilogo & \centering 5$\times$5 & \textcolor{red}{\xmark} & \centering 6$\times$6 & \textcolor{red}{\xmark}&  \centering 7$\times$7 & \textcolor{red}{\xmark} &  \centering 10$\times$10 & \textcolor{red}{\xmark} &  \centering 3$\times$4 & \textcolor{ForestGreen}{\cmark}&  \centering 7$\times$8 & \textcolor{red}{\xmark}\\
     \rowcolor{lightgray}
     \raisebox{-0.2\height}\sonnetlogo & \centering 5$\times$5 & \textcolor{red}{\xmark} & \centering 7$\times$8 & \textcolor{red}{\xmark}&  \centering 6$\times$6 & \textcolor{red}{\xmark} &  \centering 9$\times$9 & \textcolor{red}{\xmark} &  \centering 4$\times$4 & \textcolor{red}{\xmark}&  \centering 7$\times$7 & \textcolor{red}{\xmark}\\
     \raisebox{-0.2\height}\newsonnetlogo & \centering 4$\times$5 & \textcolor{ForestGreen}{\cmark} & \centering 6$\times$7 & \textcolor{ForestGreen}{\cmark}&  \centering 7$\times$7 & \textcolor{red}{\xmark} &  \centering 8$\times$7 & \textcolor{ForestGreen}{\cmark} &  \centering 3$\times$4 & \textcolor{ForestGreen}{\cmark}&  \centering 7$\times$7 & \textcolor{red}{\xmark}\\
     \end{tabular}

    \vspace{4pt}
    \centering
    \begin{tabular}{cccccccccccccc}
    \raisebox{-0.1\height}\gptlogo & \gpt &&&
      \raisebox{-0.1\height}\geminilogo & \gemini & &&\raisebox{-0.1\height}\sonnetlogo & \sonnet & && \raisebox{-0.12\height}\newsonnetlogo & \newsonnet \\
      \end{tabular}
     
\end{AIbox}
\caption{VLMs are often off by one or two in counting rows and columns in an empty grid.
The same is true when a grid is small (\eg, 3$\times$4) and contains a word in each cell.}
\label{fig:qual_grid}
\end{figure}

\clearpage
\section{Following single-colored paths task}
\label{app:subway}

\subsection{Benchmark Construction Details}
\label{app:subway-bench}

Our subway-like graphs are generated using a set of parameters defining the characteristics of the plot.

\begin{itemize}
    \item \textbf{Image size}: We use two different sizes \{512, 1024\}px for the images to include various resolutions.
    \item \textbf{Grid size}: We assume a hypothetical grid on the image that determines the position of the paths. We used an 18$\times$18 grid, which means each path segment is \(\frac{1}{18}\) of the image size.
    \item \textbf{Number of stations}: We use four station labels, \{A, B, C, D\}.
    \item \textbf{Starting points}: Each station in our maps has three different starting points which are exactly $\frac{1}{18}$ of the image size to one side of the stations.
    \item \textbf{Path thickness}: We use two line thicknesses, 10 and 20 pixels to have bold and light visualizations of the same path.
    \item \textbf{Number of paths}: Considering the number of starting points in our setup, each image can include stations from which exactly 1, 2, or 3 paths exit.
     
\end{itemize}

\begin{figure}
    \centering
    \begin{tabular}{cc}
         \includegraphics[width=0.33\textwidth]{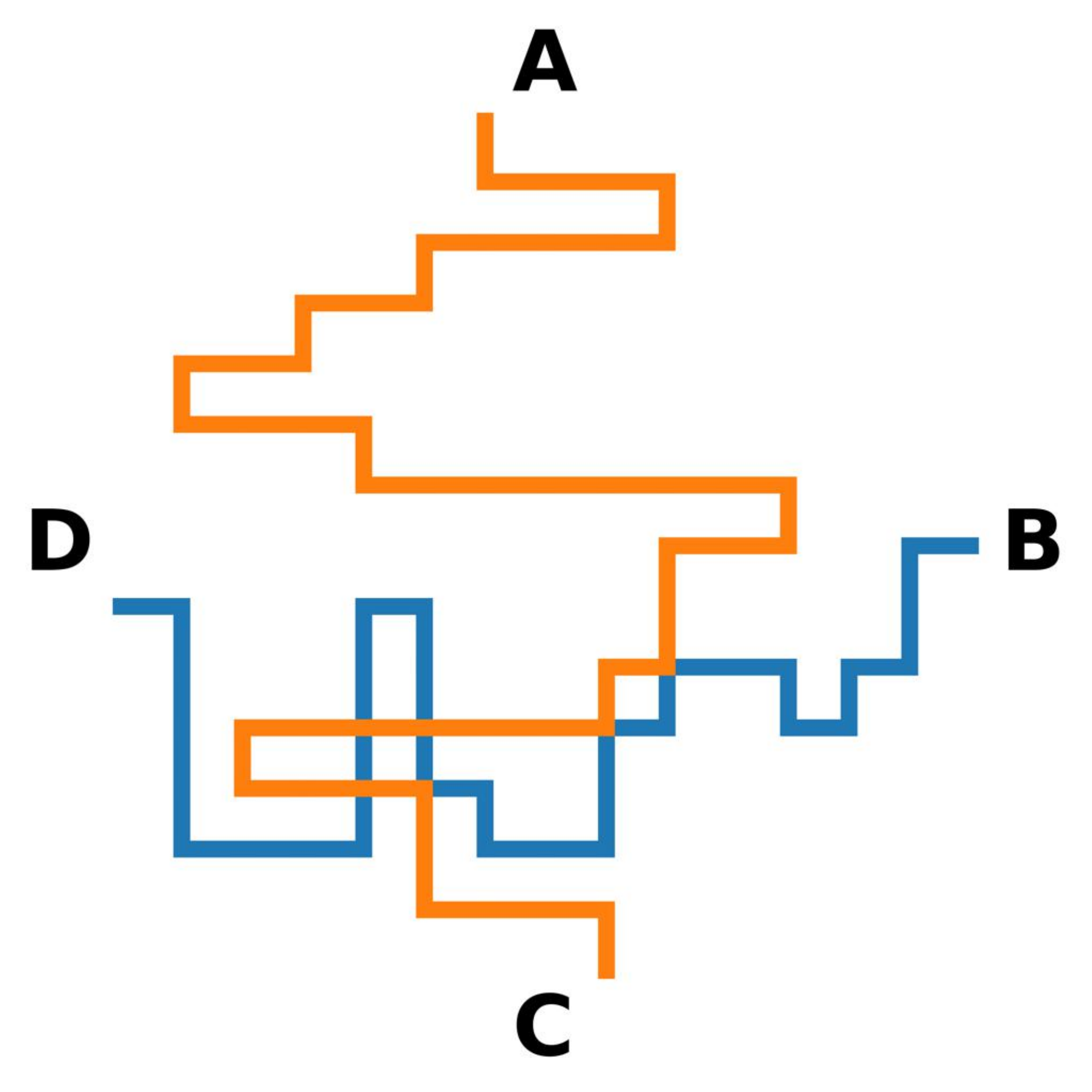}&\includegraphics[width=0.33\textwidth]{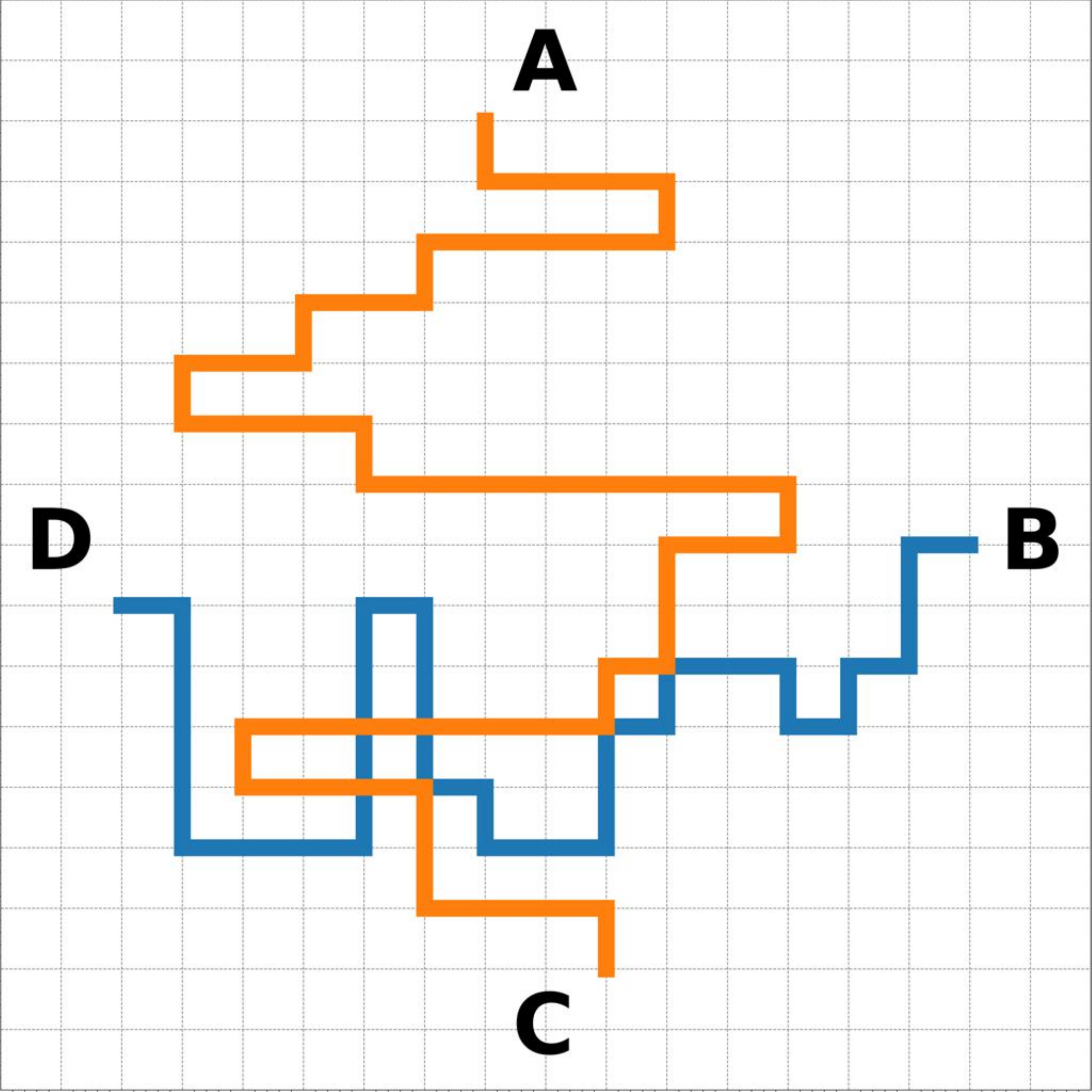} \\
         \includegraphics[width=0.33\textwidth]{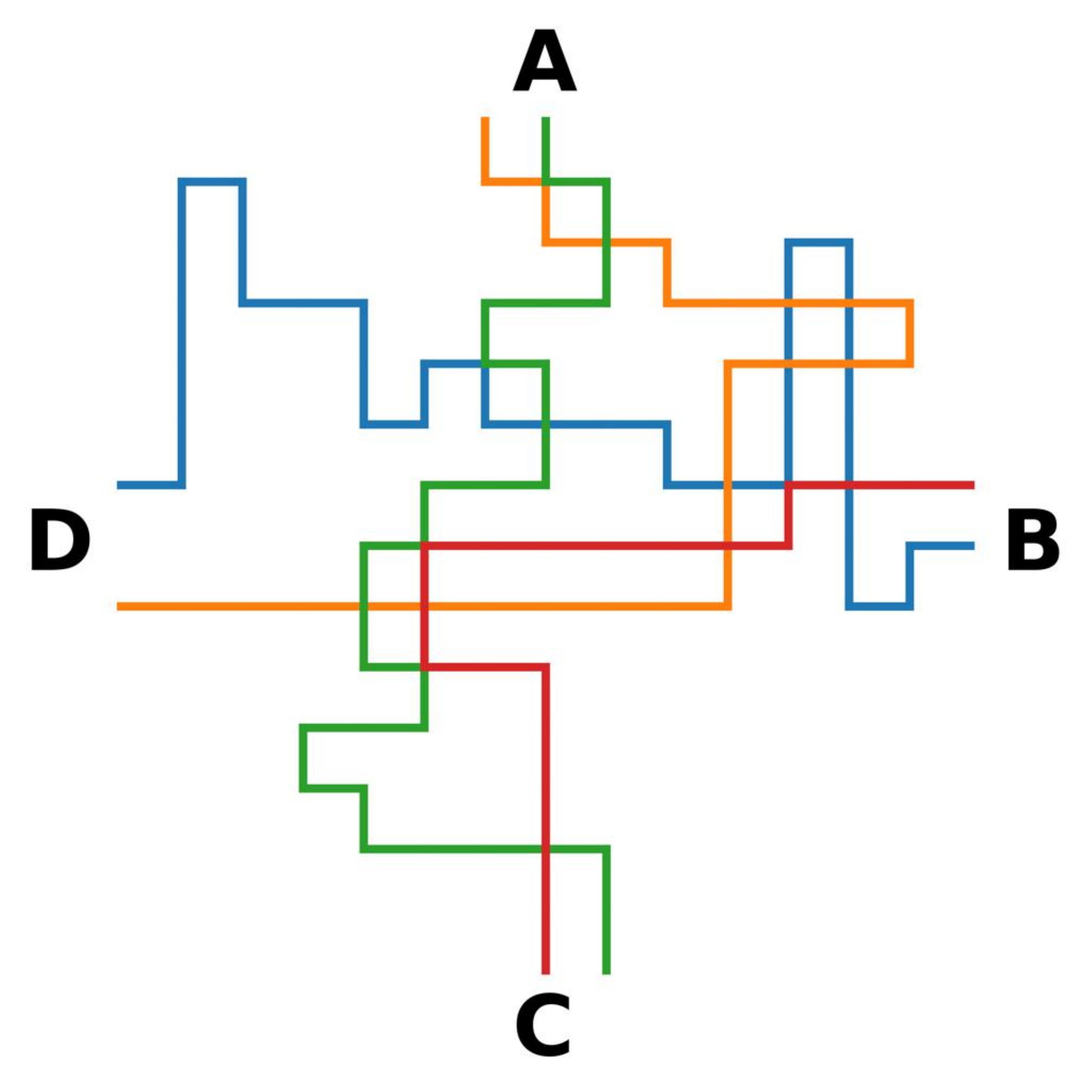}&\includegraphics[width=0.33\textwidth]{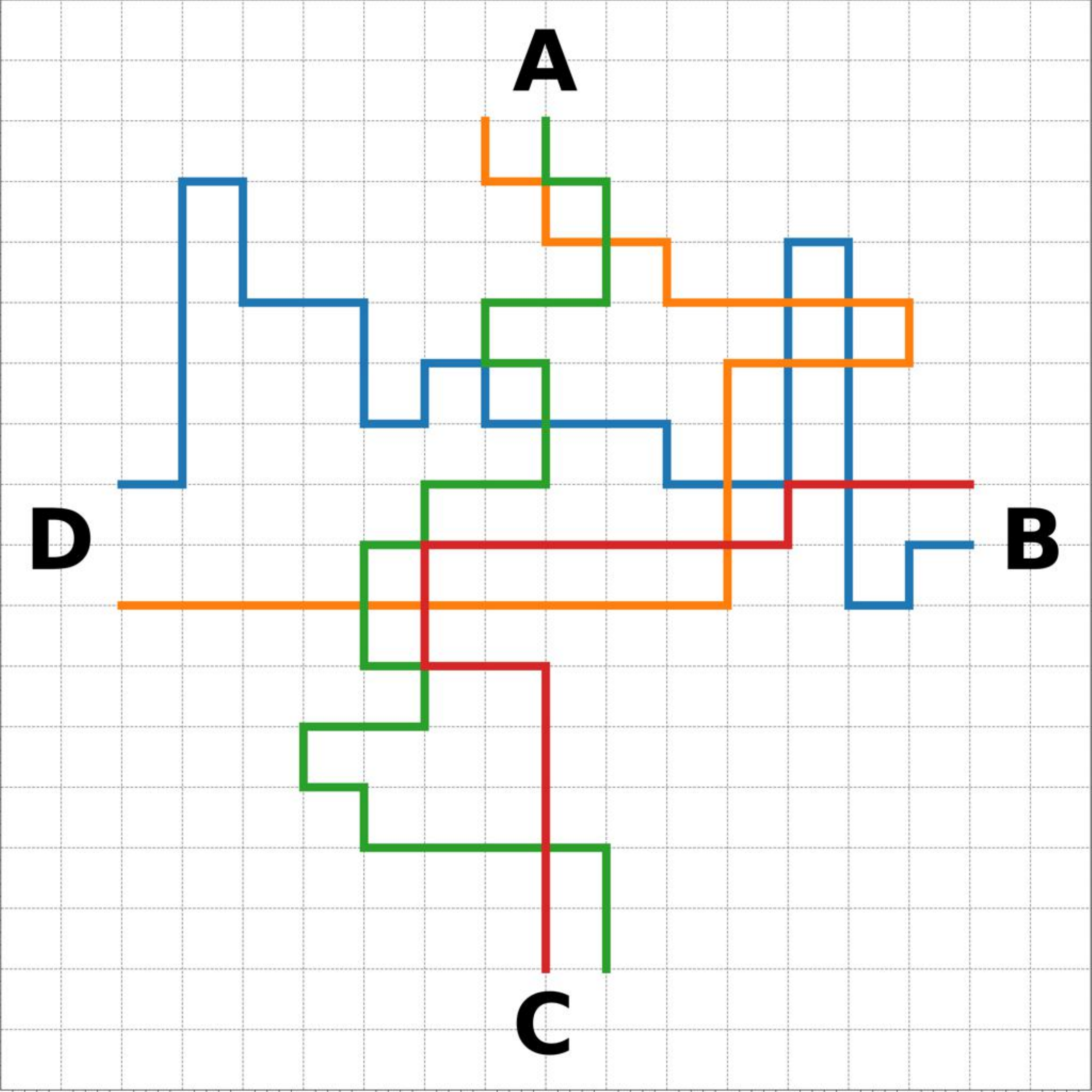} \\
         \includegraphics[width=0.33\textwidth]{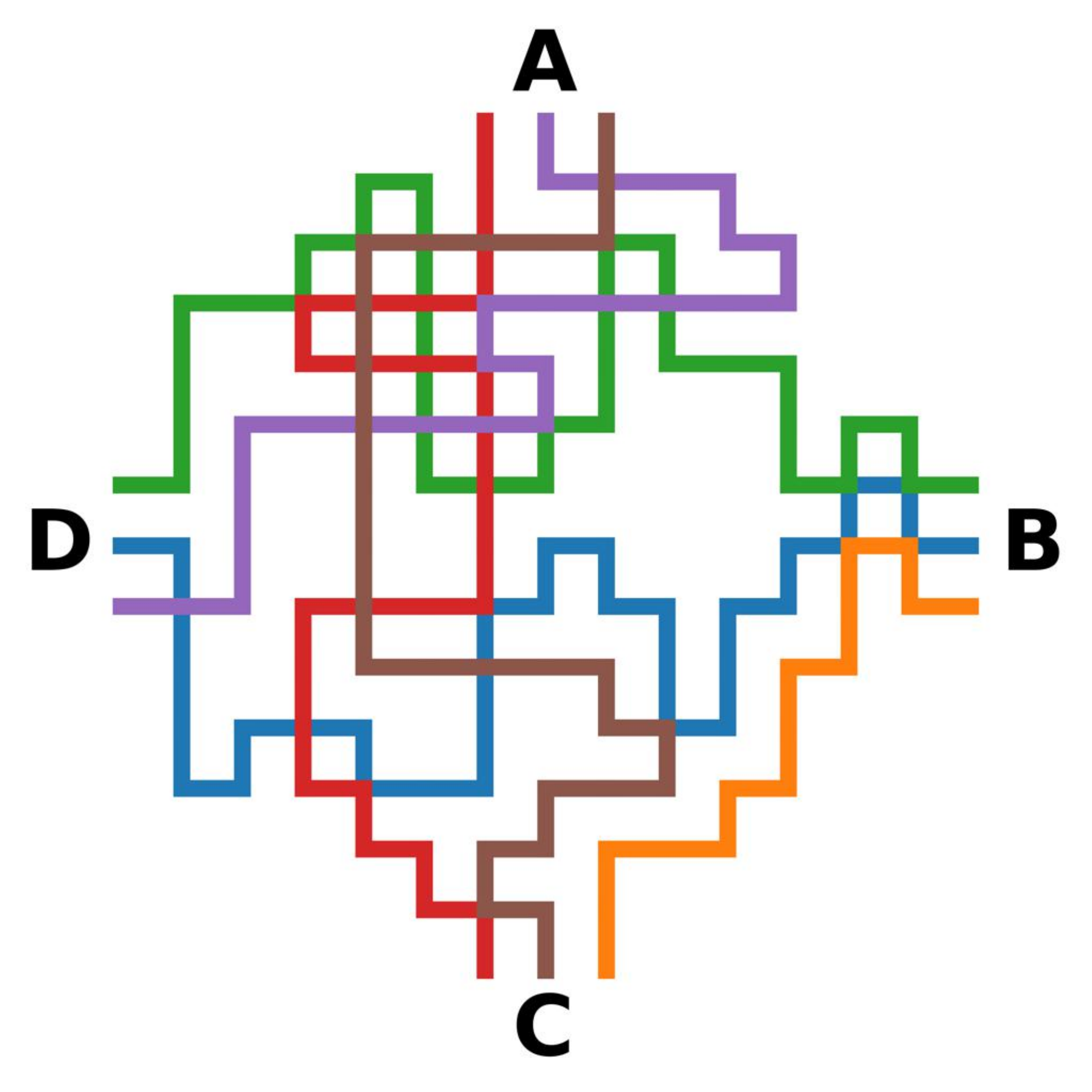} &\includegraphics[width=0.33\textwidth]{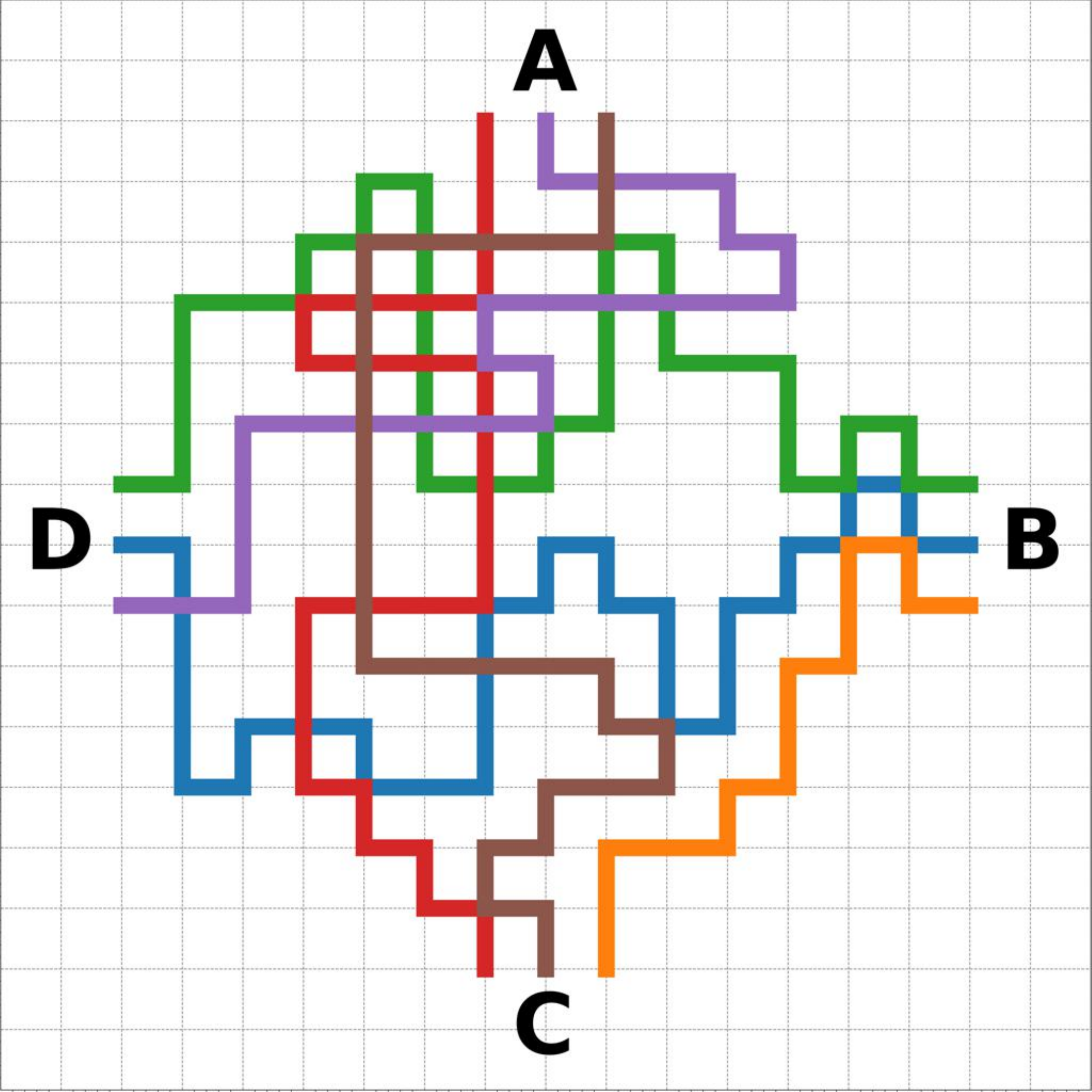} \\
    \end{tabular}
    \caption{Images in our benchmark (left) have exactly 1, 2, or 3 paths exiting each station. The hypothetical grids (right) are used when generating the paths.}
    \label{app:fig-subwway-ex}
\end{figure}

We keep generating the images until we have 15 samples for each number of paths which results in 180 images (see \cref{table:sub-parameters,app:fig-subwway-ex}).

\textbf{Code} The code is available at \url{https://github.com/anguyen8/vision-llms-are-blind/blob/main/src/SubwayMap/SubwayMap.ipynb}

\subsec{Prompts}
\begin{enumerate}
    \item \emph{How many single-colored paths go from \texttt{\{station 1\}} to \texttt{\{station 2\}}? Answer with a number in curly brackets, e.g., \{3\}.}

    \item \emph{Count the one-colored routes that go from \texttt{\{station 1\}} to \texttt{\{station 2\}}. Answer with a number in curly brackets, e.g., \{3\}.}
\end{enumerate}
where the two stations are different and sampled from $\{ \text{ {\textsf{\footnotesize A}}, {\textsf{\footnotesize B}}, {\textsf{\footnotesize C}}, {\textsf{\footnotesize D}} }\}$.

\noindent\textbf{Groundtruth} answers are $\in \{1, 2, 3\}$ (random-baseline accuracy: 33.33\%).\\
Note that, to make the task easier, 0 is excluded from the groundtruth set (\ie, we never ask VLMs to count when no path exists between two given stations).

\begin{table}
\centering
\caption{We create 2 different image resolutions, and 3 various line widths to have balanced number of colored paths.}
\label{table:sub-parameters}
\begin{tabular}{lccc}
\toprule
\textbf{Parameter} & \textbf{Values} & ~~\textbf{Samples} & ~~\textbf{Total Samples}\\
\midrule
\multirow{2}{*}{Image size} & 512px & ~~90 & ~~\multirow{2}{*}{180}\\
& 1024px & ~~90 \\
\midrule
\multirow{2}{*}{Line thickness} & 10px & ~~90 & ~~\multirow{2}{*}{180}\\
 & 20px & ~~90 \\
\midrule
\multirow{3}{*}{Paths} & 1 & ~~60 & ~~\multirow{3}{*}{180}\\
 & 2 & ~~60 \\
  & 3 & ~~60 \\
\bottomrule
\end{tabular}
\end{table}

\begin{figure}
    \centering
    \begin{tabular}{cc}
    \includegraphics[width=0.4\textwidth]{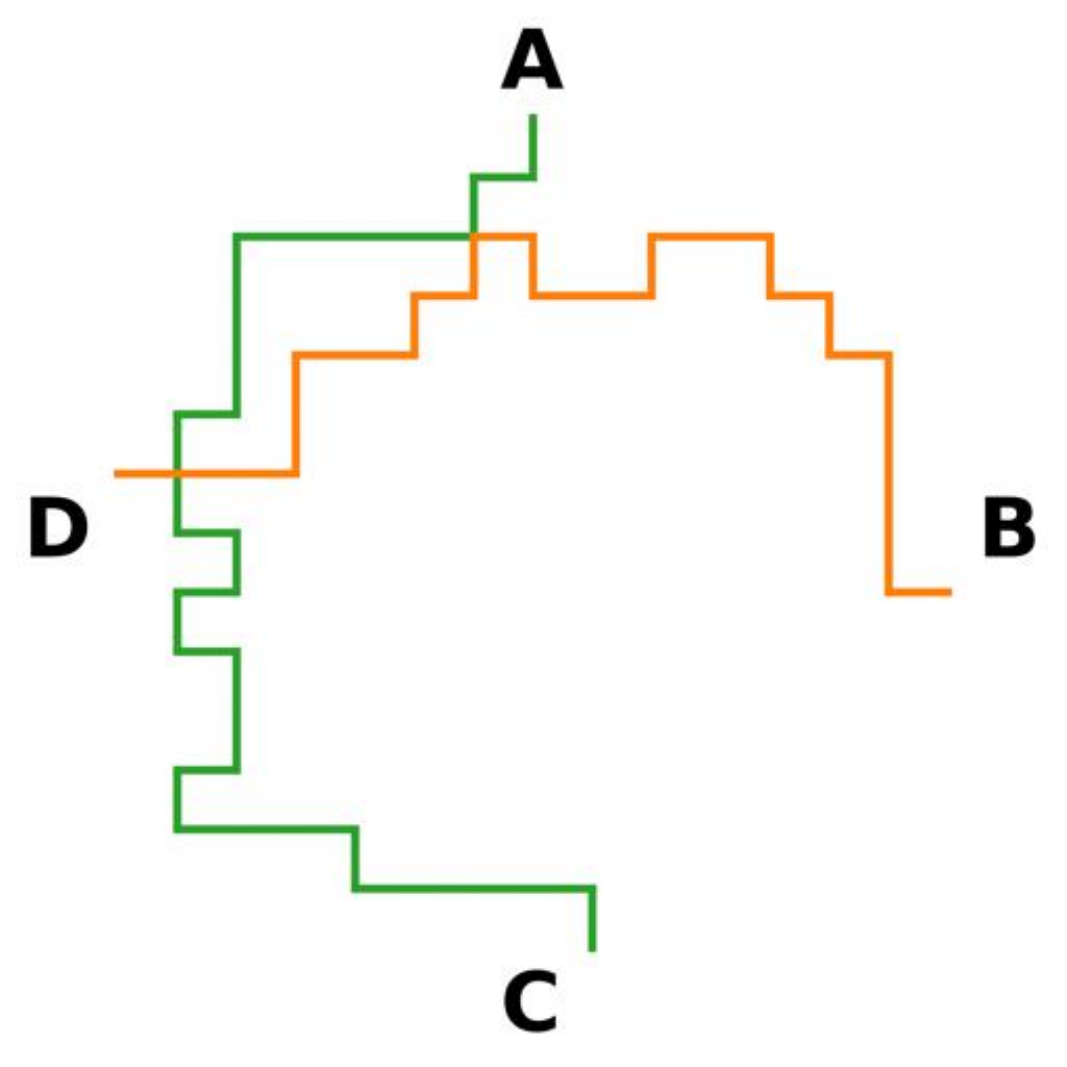} & \includegraphics[width=0.4\textwidth]{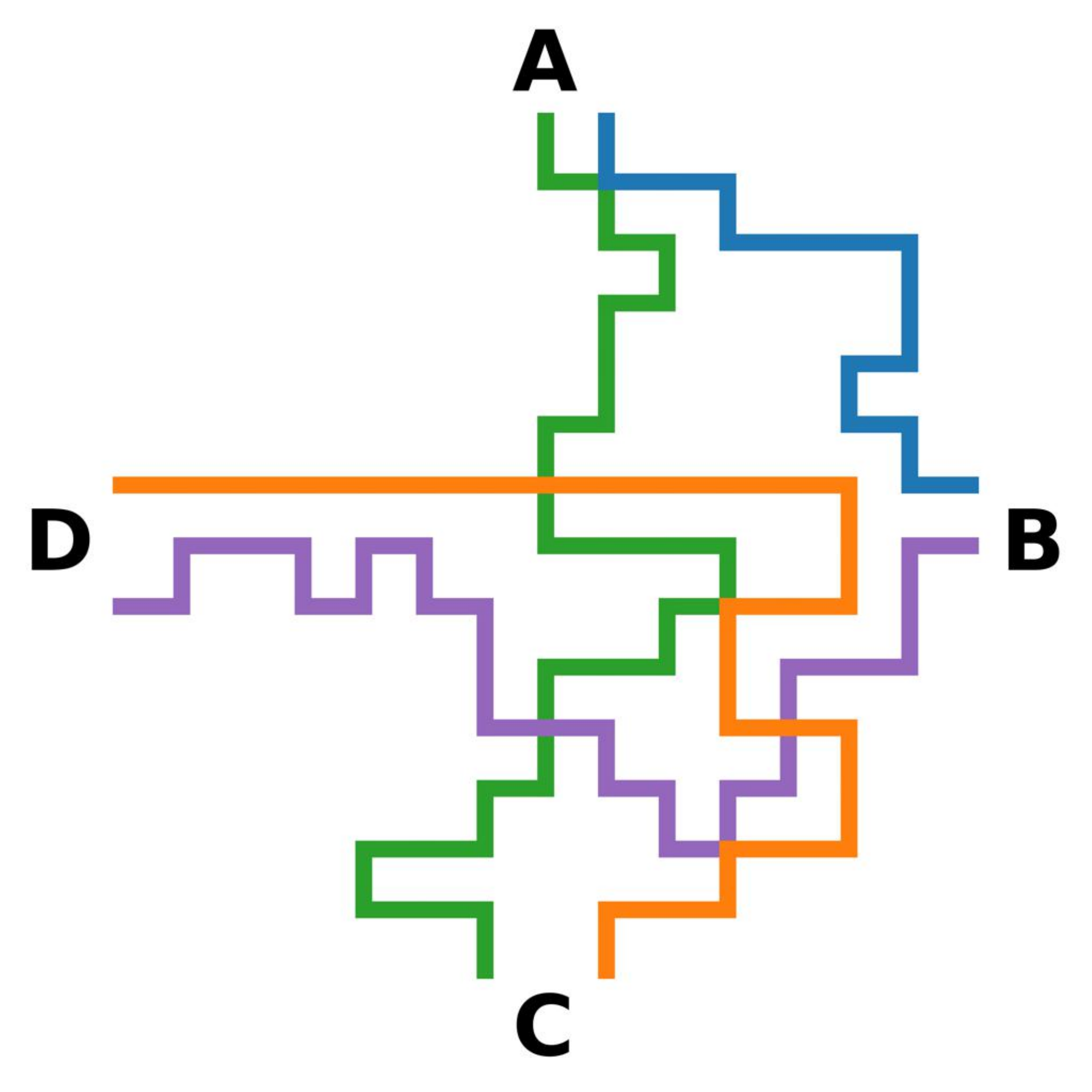} \\
     \includegraphics[width=0.4\textwidth]{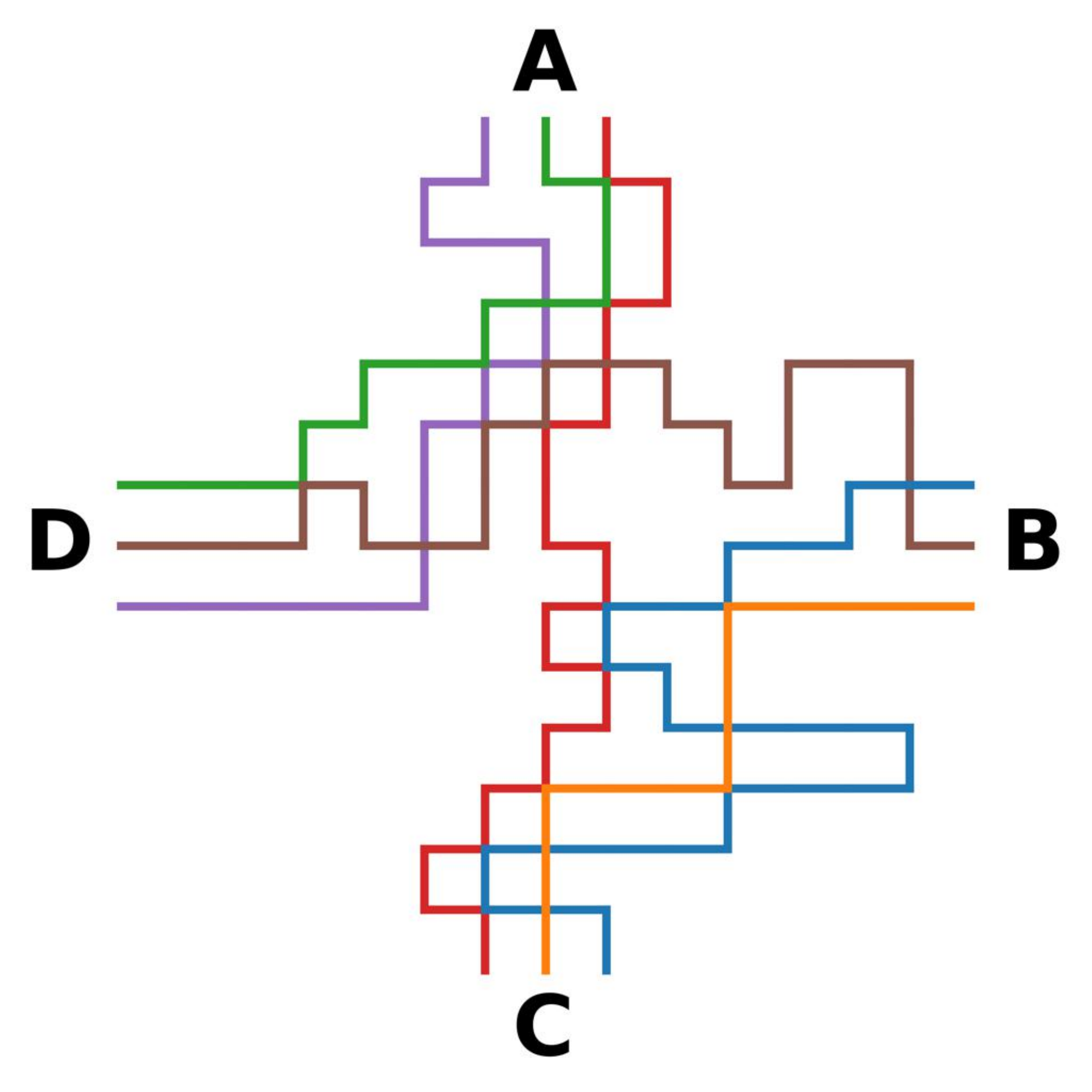}& \includegraphics[width=0.4\textwidth]{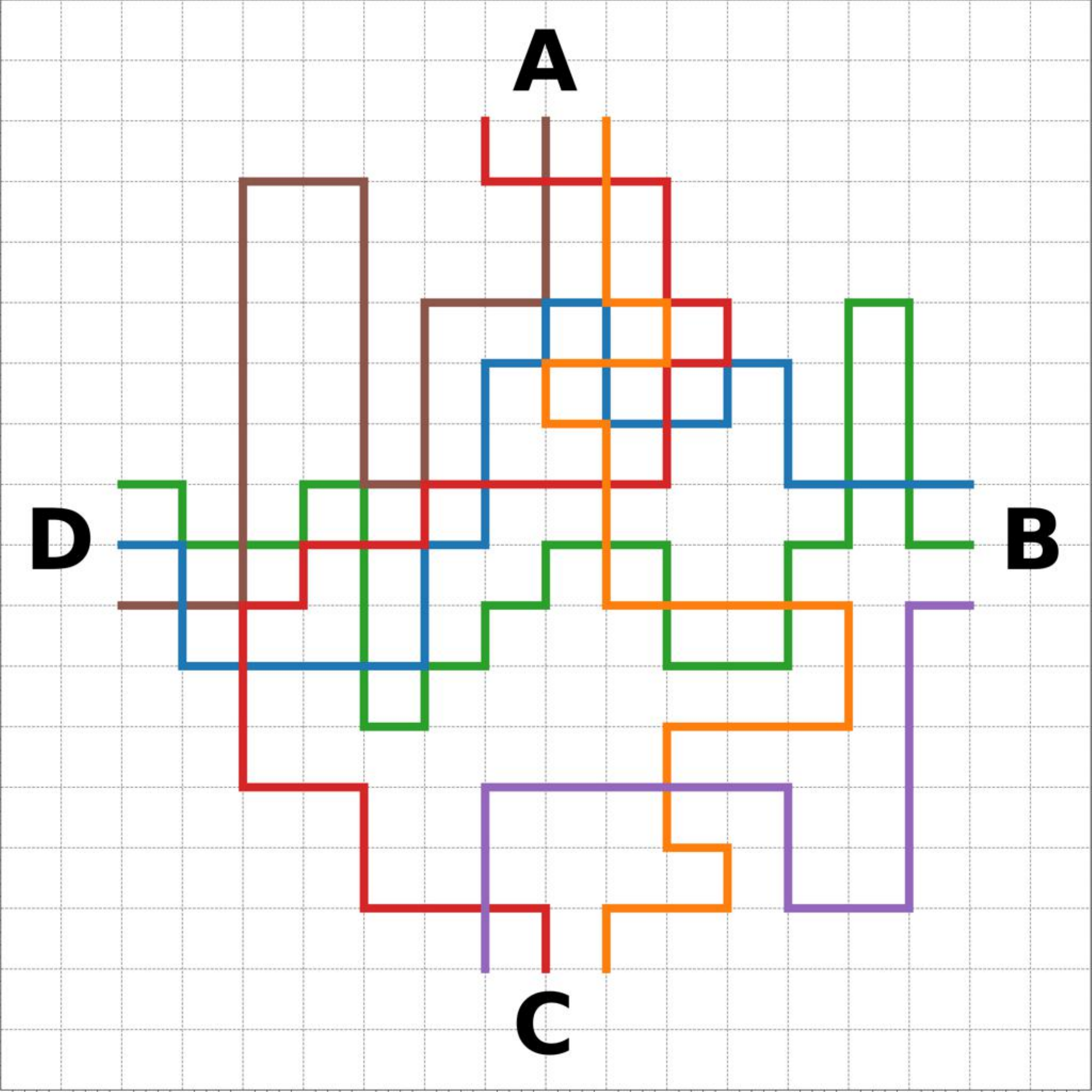}  \\
\end{tabular}
\caption{
On an invisible 18$\times$18 grid (bottom right), we randomly generate random paths from one station to another. All stations have an equal $N$ = 1, 2 or 3 outgoing paths.
}
\label{fig:sub-examples-f}
\end{figure}

\clearpage
\subsection{Finding: VLMs fail to see the colored paths across various map complexities}
\begin{figure}[h] 
        \centering
        \resizebox{0.5\textwidth}{!}{
        \begin{tabular}{cc}
            \raisebox{-2\height}{\scalebox{2.0}{\gptlogo}} & \multirow{4}{*}{\includegraphics[width=0.6\textwidth]{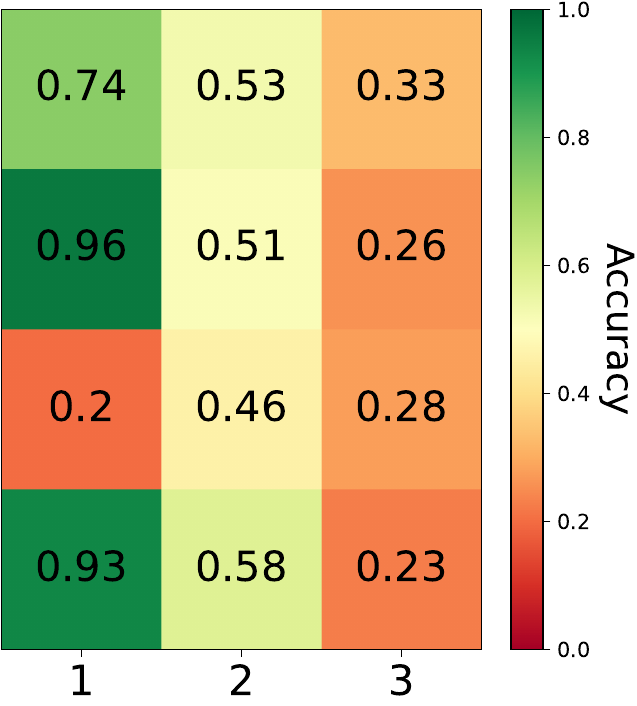}}\\
            \raisebox{-2.8\height}{\scalebox{2.0}{\geminilogo}} &\\
            \raisebox{-2.7\height}{\scalebox{2.0}{\sonnetlogo}} &\\
            \raisebox{-2.5\height}{\scalebox{2.0}{\newsonnetlogo}} &\\
        \end{tabular}}
        \vspace{5em}
        \caption{
        VLM accuracy decreases as the number of paths increases from 1 to 3.
    Overall, counting colored paths poses a challenge to VLMs.
        }\label{fig:overall_acc_subway} 
\end{figure}

All VLMs, except \sonnet, start to perform worse as the number of paths per station increases in the images of subway-like maps (\cref{fig:overall_acc_subway}).

\clearpage

\subsection{\newsonnet can more accurately count subway paths on simplified images compared to other VLMs}
\label{app:simp_sub}

\begin{figure}
    \centering
    \resizebox{\textwidth}{!}{
    \begin{tabular}{c}
         \begin{tabular}{p{0.5cm}p{1.5cm}p{1.5cm}p{1.5cm}p{1.5cm}p{1.5cm}p{1.5cm}p{1.5cm}}
            &\hspace{1.1cm}\gptlogo & 
            \hspace{0.8cm}\geminilogo &
            \hspace{0.55cm}\sonnetlogo &
            \hspace{0.2cm}\newsonnetlogo &
            \hspace{-0.15cm}\llavalogolarge&
            \hspace{-0.45cm}\llavalogosmall&
            \hspace{-0.7cm}\philogo\\
        \end{tabular} \\
         \includegraphics[width=0.85\textwidth]{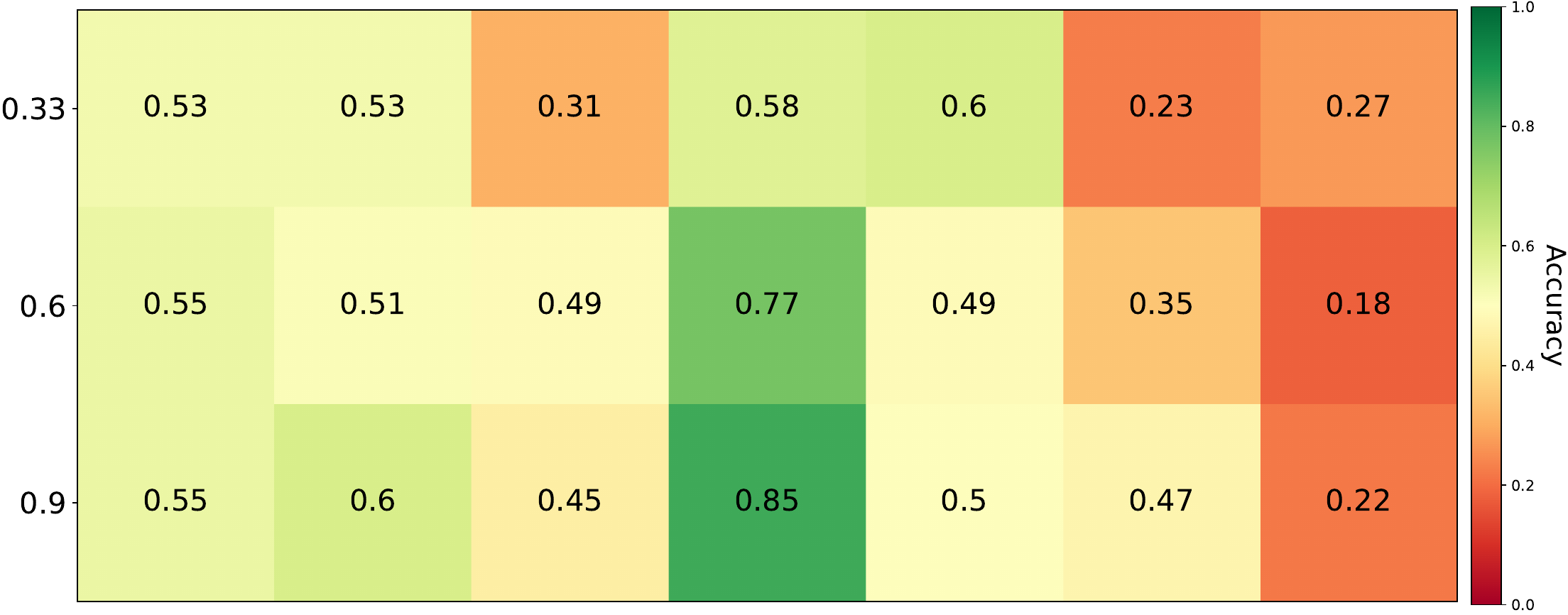}
         \\
    \end{tabular}
    }
    \caption{\newsonnet has the largest mean performance gain \increase{27} from $P=$ 0.33 to 0.9 when we increase the probability of choosing a straight path.}
    \label{fig:simp_subway_res}
\end{figure}

\newsonnet consistently benefits from increasing the probability of choosing the straight path across all image settings (\cref{fig:simp_subway_res}), suggesting its counting ability is generally better than other VLMs, and perhaps the complexity of images is the reason for its initial low performance.

\clearpage
\subsection{Additional Examples}
\begin{figure}[ht]
\centering
\begin{AIbox}{How many single-color paths go from A to B? Answer with a number in curly brackets e.g. \{3\}}
\centering

    \begin{tabular}{lp{0.8cm}c|p{0.8cm}c|p{0.8cm}c|p{0.8cm}c|p{0.8cm}c|p{0.8cm}c}
     &\multicolumn{2}{c}{\includegraphics[width=0.15\textwidth]{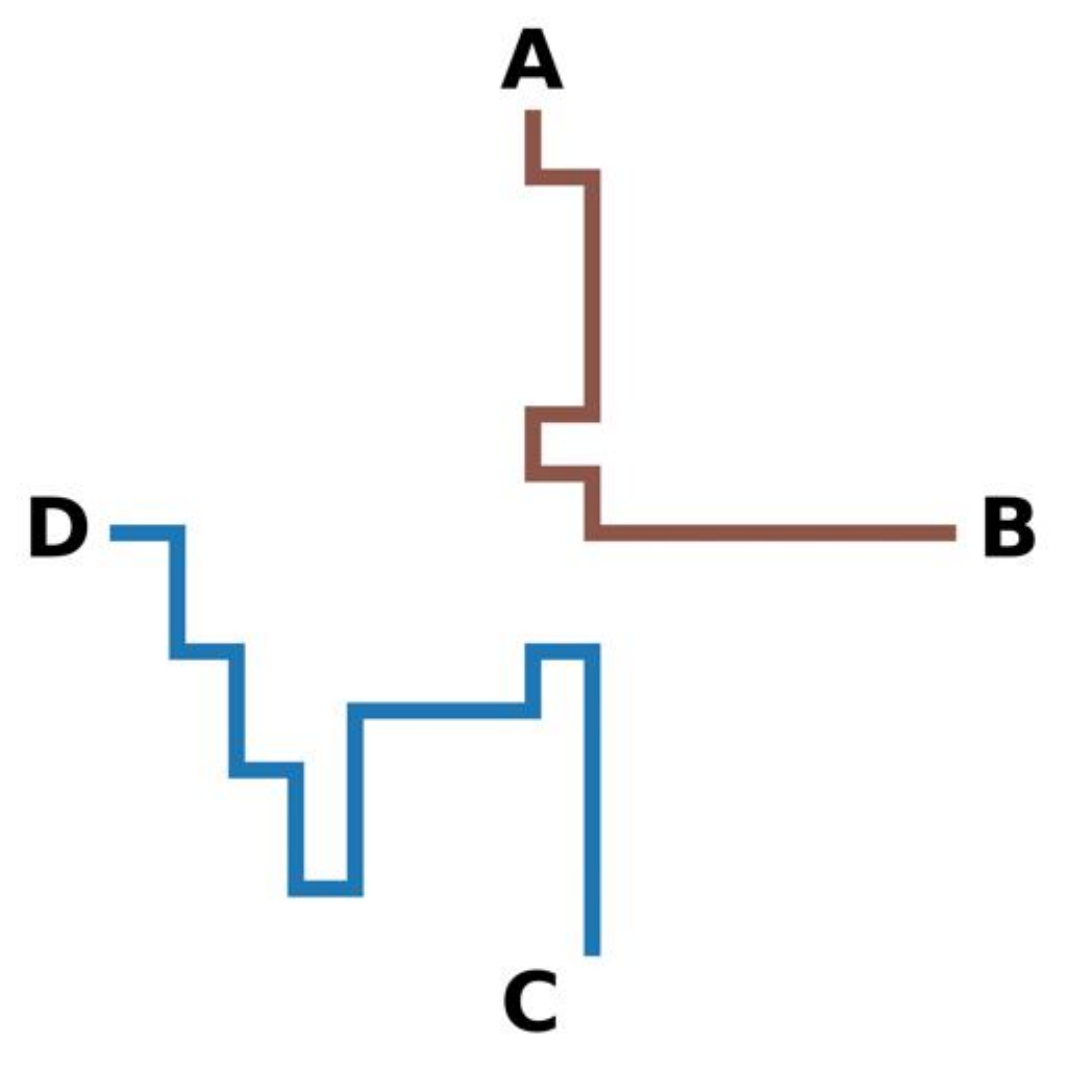}} & \multicolumn{2}{c}{\includegraphics[width=0.15\textwidth]{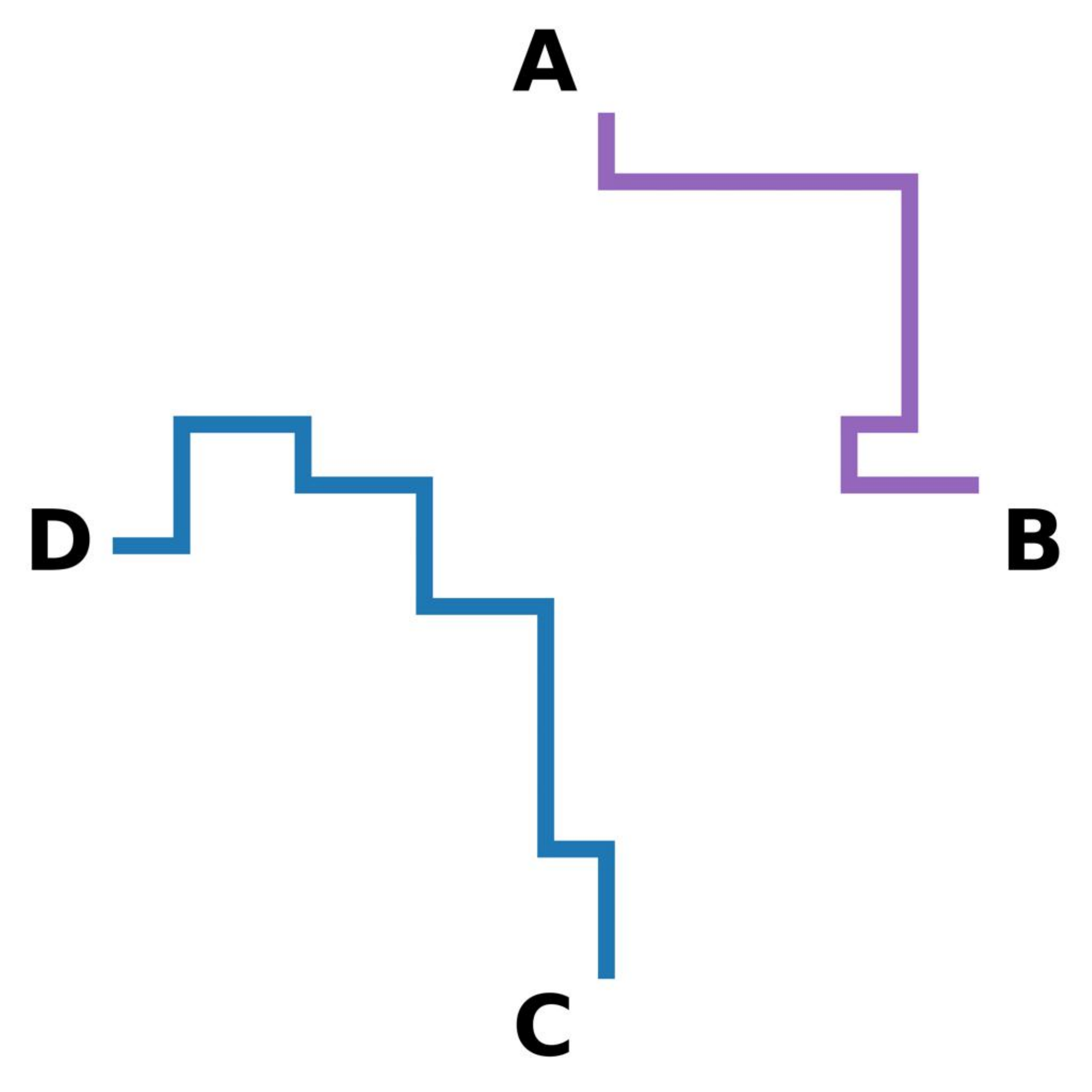}} &\multicolumn{2}{c}{\includegraphics[width=0.15\textwidth]{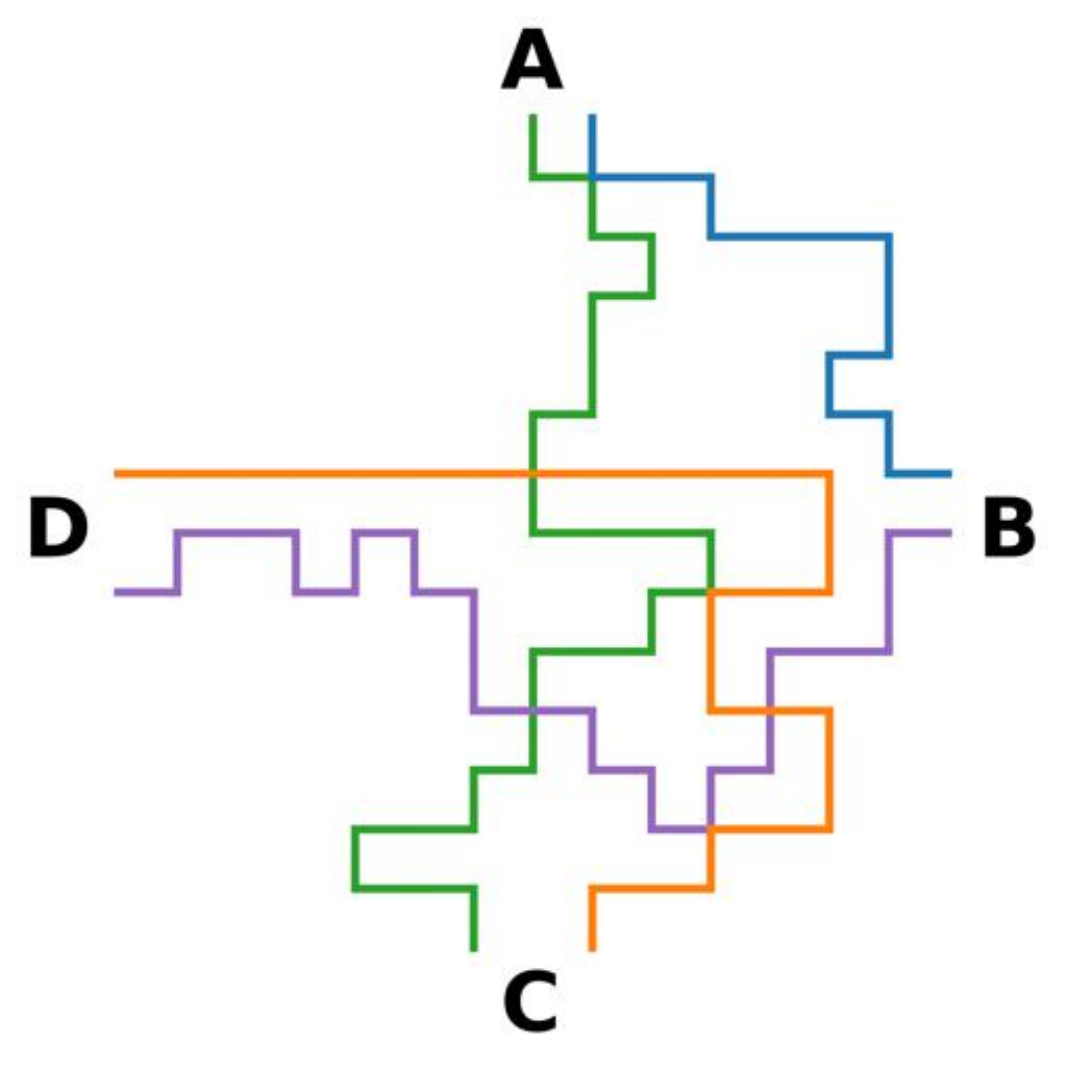}} &\multicolumn{2}{c}{\includegraphics[width=0.15\textwidth]{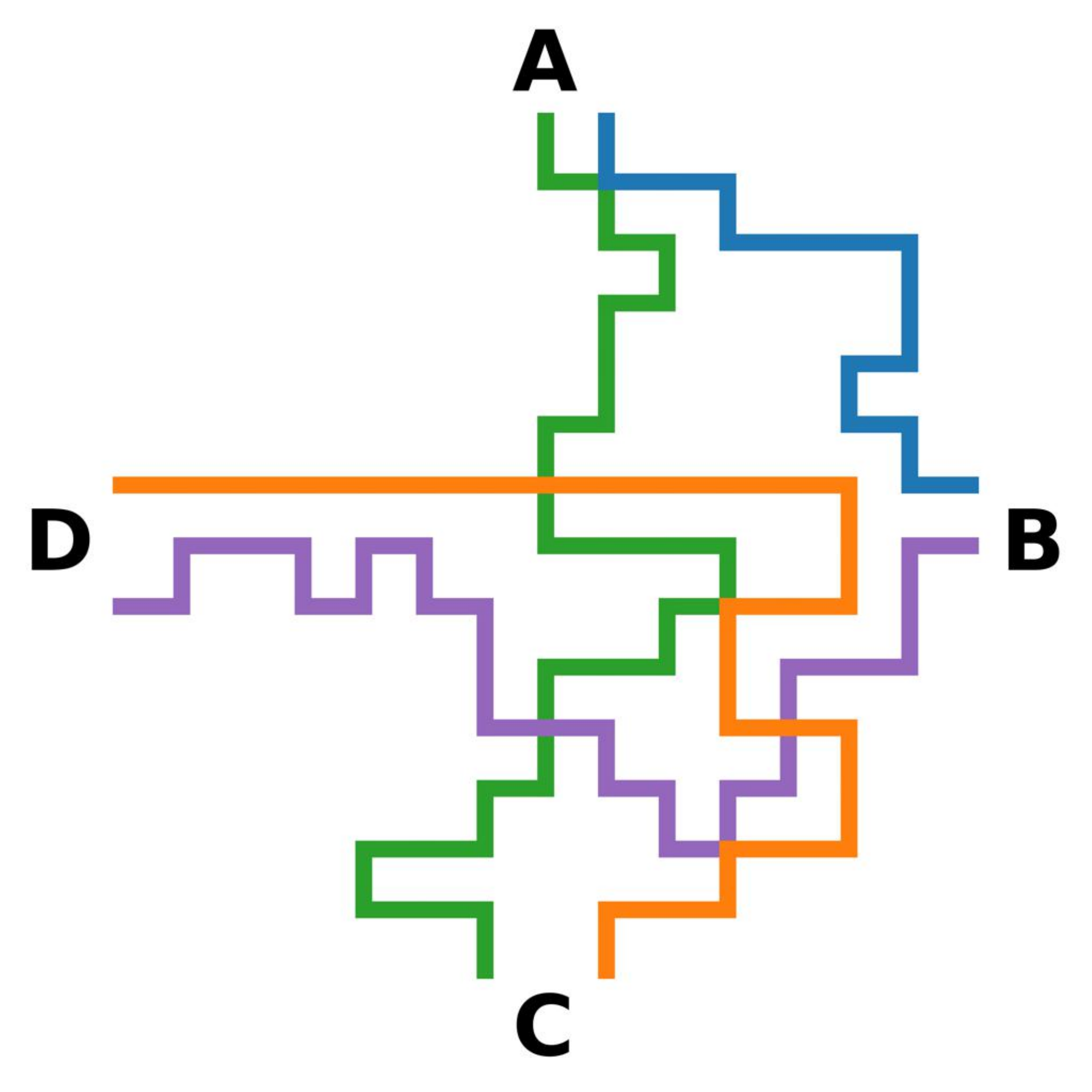}} &\multicolumn{2}{c}{\includegraphics[width=0.15\textwidth]{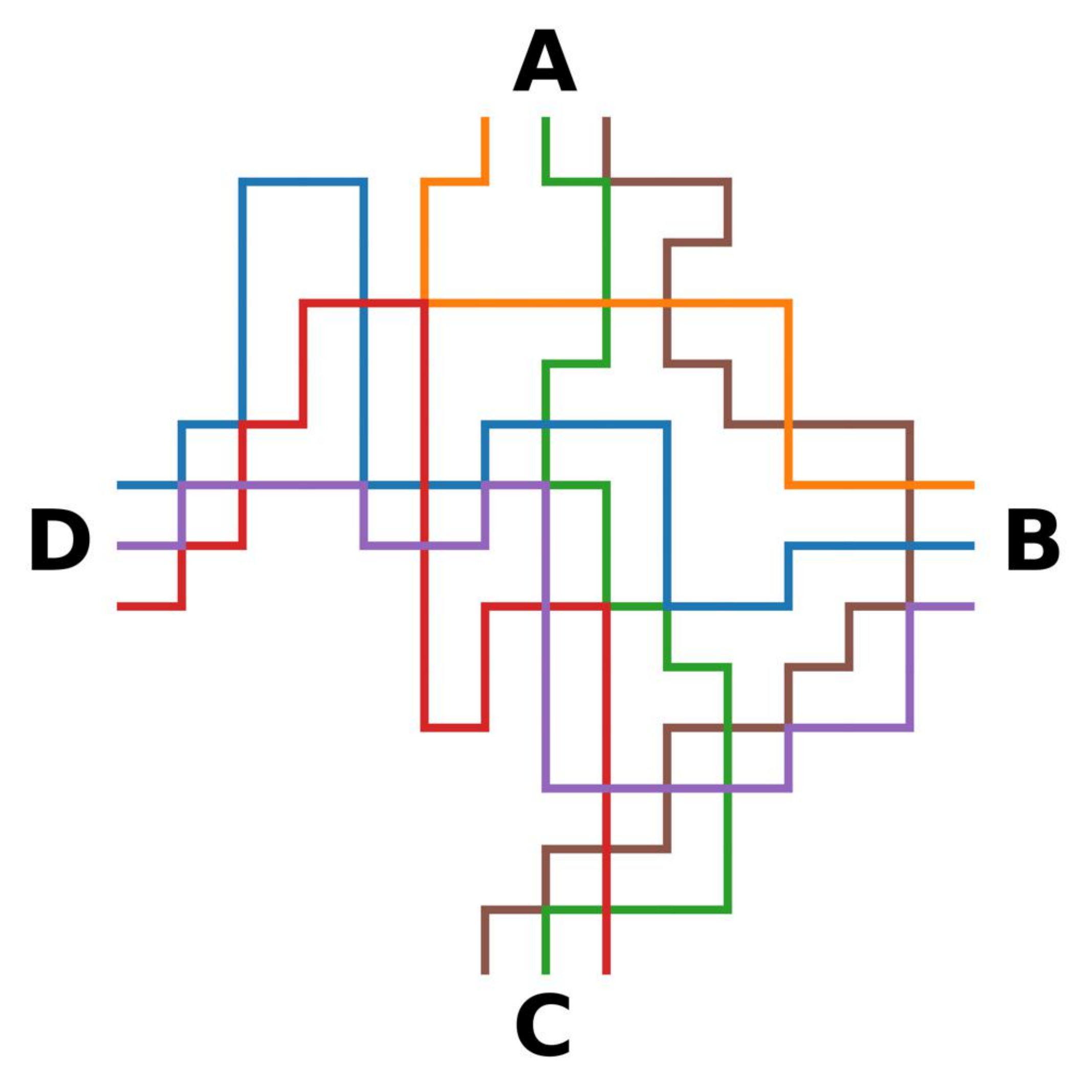}}&\multicolumn{2}{c}{\includegraphics[width=0.15\textwidth]{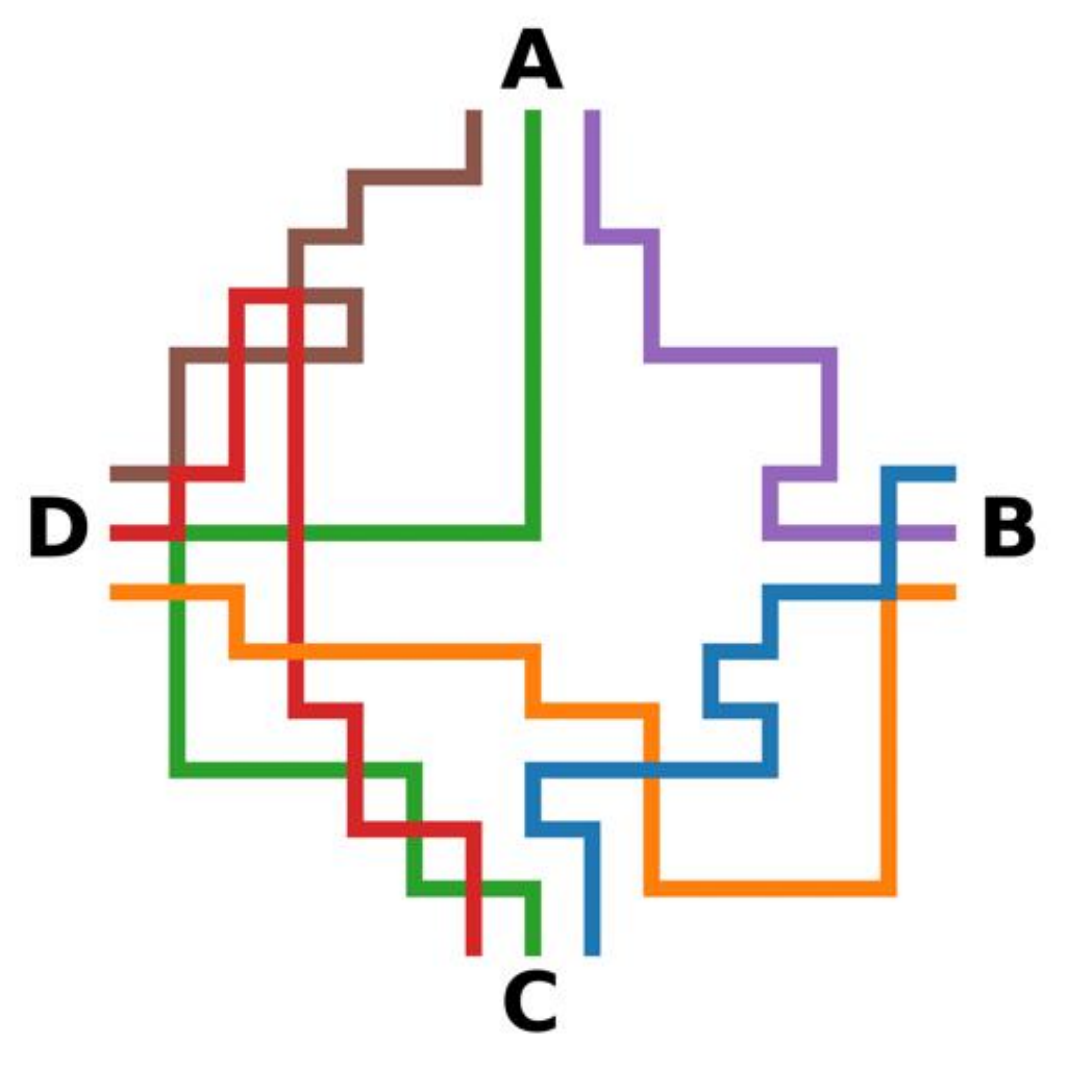}}\\
     \rowcolor{lightgray}
     \raisebox{-0.2\height}\gptlogo & \centering 1 & \textcolor{ForestGreen}{\cmark} &  \centering 1 & \textcolor{ForestGreen}{\cmark}&  \centering 1 & \textcolor{ForestGreen}{\cmark} &  \centering 2 & \textcolor{red}{\xmark} &  \centering 2 & \textcolor{red}{\xmark}&  \centering 0 & \textcolor{red}{\xmark}\\
     \raisebox{-0.2\height}\geminilogo & \centering 2 & \textcolor{red}{\xmark} & \centering 2 & \textcolor{red}{\xmark}&  \centering 3 & \textcolor{red}{\xmark} &  \centering 3 & \textcolor{red}{\xmark} &  \centering 1 & \textcolor{ForestGreen}{\cmark}&  \centering 6 & \textcolor{red}{\xmark}\\
     \rowcolor{lightgray}
     \raisebox{-0.2\height}\sonnetlogo & \centering 2 & \textcolor{red}{\xmark} & \centering 2 & \textcolor{red}{\xmark}&  \centering 2 & \textcolor{red}{\xmark} &  \centering 2 & \textcolor{red}{\xmark} &  \centering 4 & \textcolor{red}{\xmark}&  \centering 2 & \textcolor{red}{\xmark}\\
     \raisebox{-0.2\height}\newsonnetlogo & \centering 1 & \textcolor{ForestGreen}{\cmark} & \centering 2 & \textcolor{red}{\xmark}&  \centering 1 & \textcolor{ForestGreen}{\cmark} &  \centering 3 & \textcolor{red}{\xmark} &  \centering 1 & \textcolor{ForestGreen}{\cmark}&  \centering 3 & \textcolor{red}{\xmark}\\
     \end{tabular}

    \vspace{4pt}
    \centering
    \begin{tabular}{cccccccccccccc}
    \raisebox{-0.1\height}\gptlogo & \gpt &&&
      \raisebox{-0.1\height}\geminilogo & \gemini & &&\raisebox{-0.1\height}\sonnetlogo & \sonnet & && \raisebox{-0.12\height}\newsonnetlogo & \newsonnet \\
      \end{tabular}
     
\end{AIbox}
\caption{VLMs generally fail at images with more intersecting paths. As the number of paths increases, models tend to make an educated guess, especially \gemini and \newsonnet.}
\label{fig:qual_subway1}
\end{figure}

We show examples of models' responses to the counting the number of single-colored connecting paths in \cref{fig:qual_subway1}.
\label{app:subway-qual}

\begin{figure}[H]
\centering
\begin{AIbox}{How many single-color paths go from A to C? Answer with a number in curly brackets e.g. \{3\}.}
\centering
\vspace{-0.4cm}
    \begin{tabular}{lp{0.8cm}c|p{0.8cm}c|p{0.8cm}c|p{0.8cm}c|p{0.8cm}c|p{0.8cm}c}
     &\multicolumn{2}{c}

     {\includegraphics[width=0.15\textwidth]{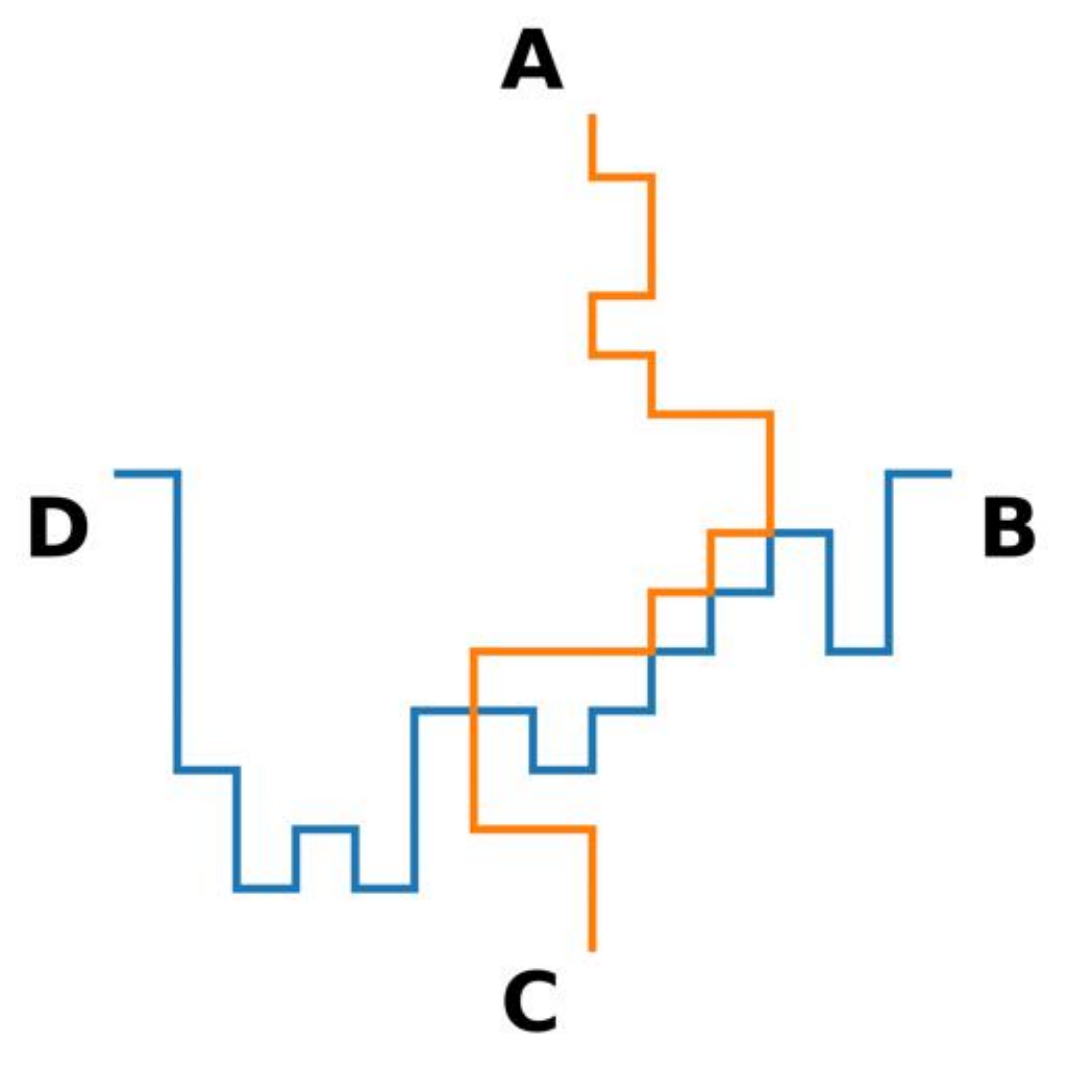}} & \multicolumn{2}{c}{\includegraphics[width=0.15\textwidth]{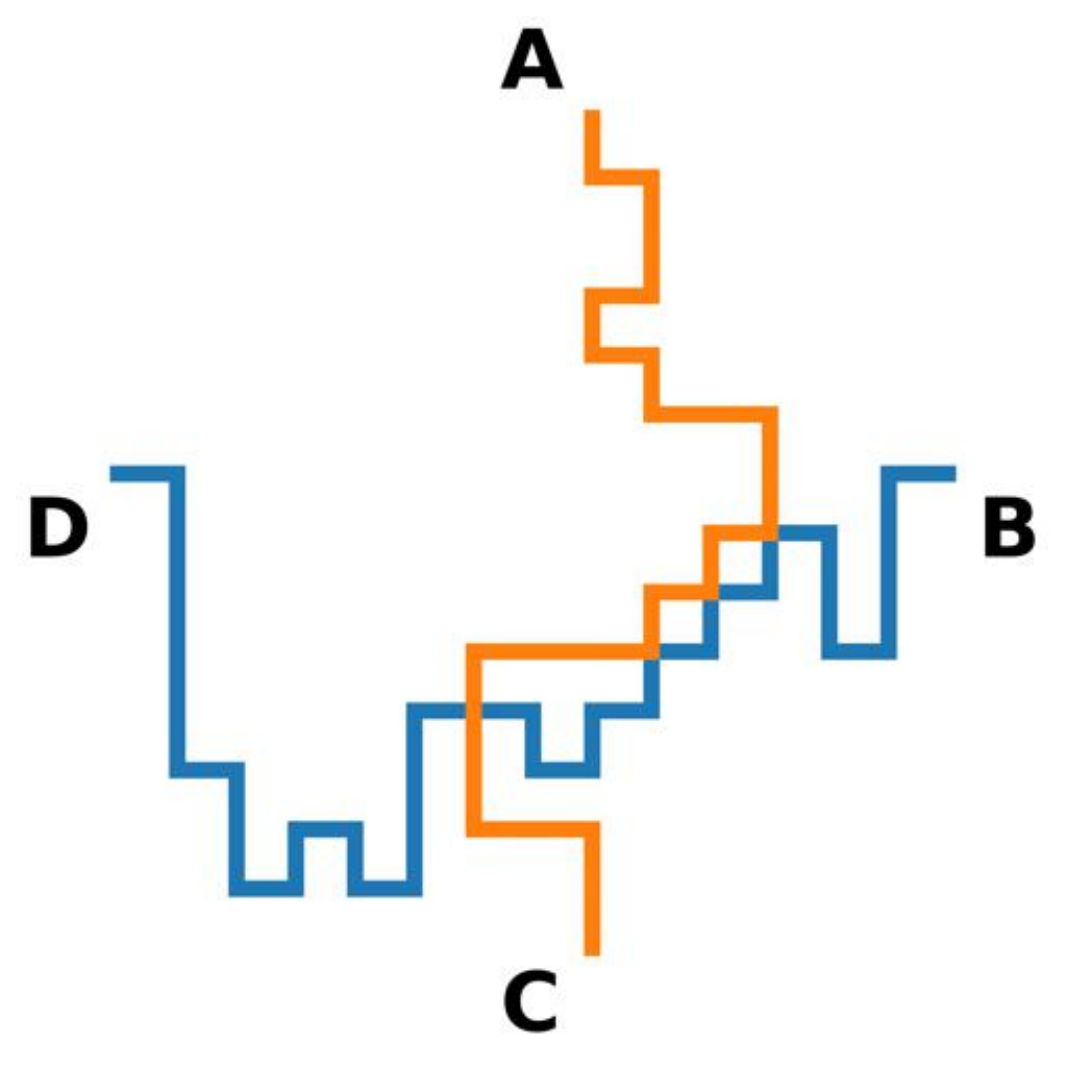}} &\multicolumn{2}{c}{\includegraphics[width=0.15\textwidth]{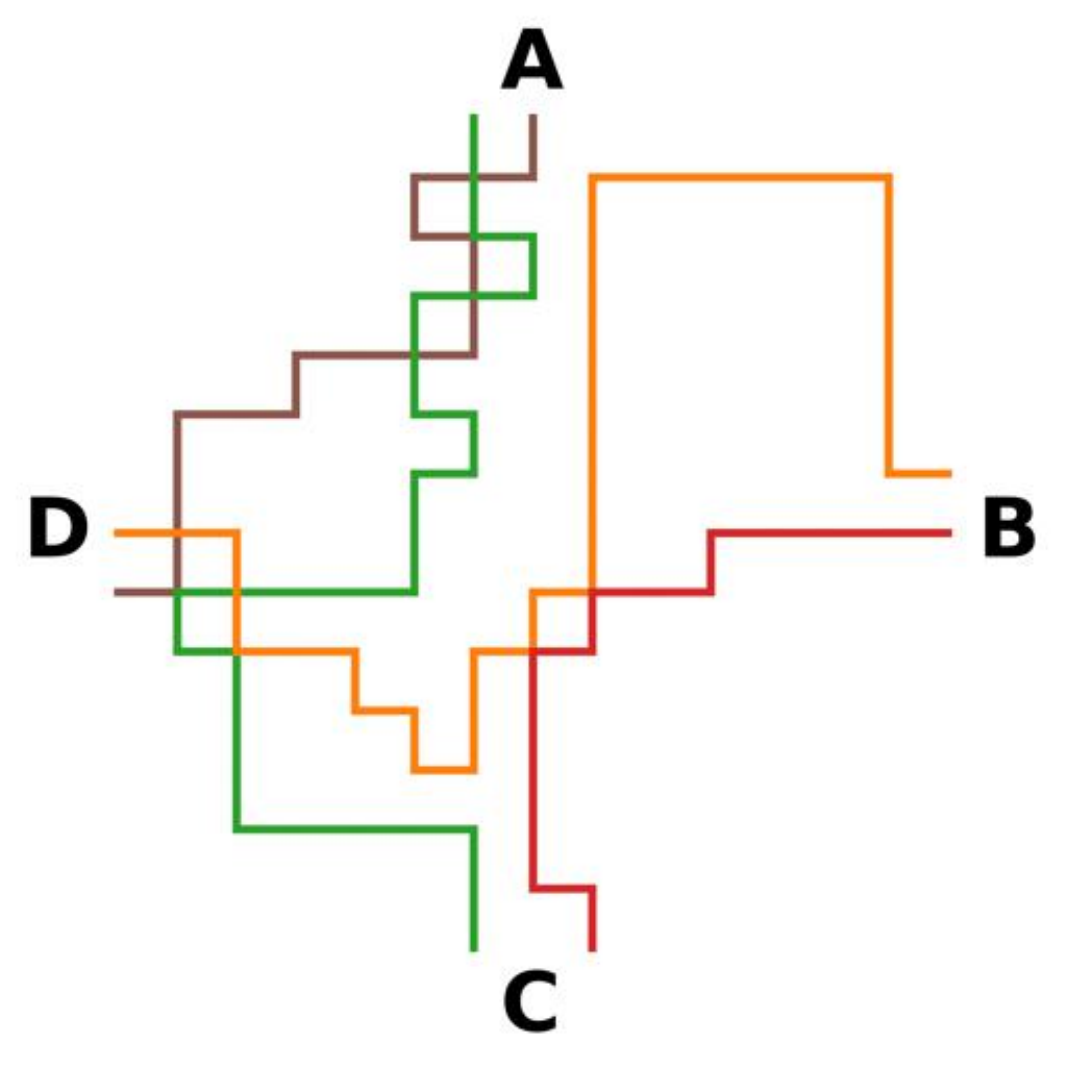}} &\multicolumn{2}{c}{\includegraphics[width=0.15\textwidth]{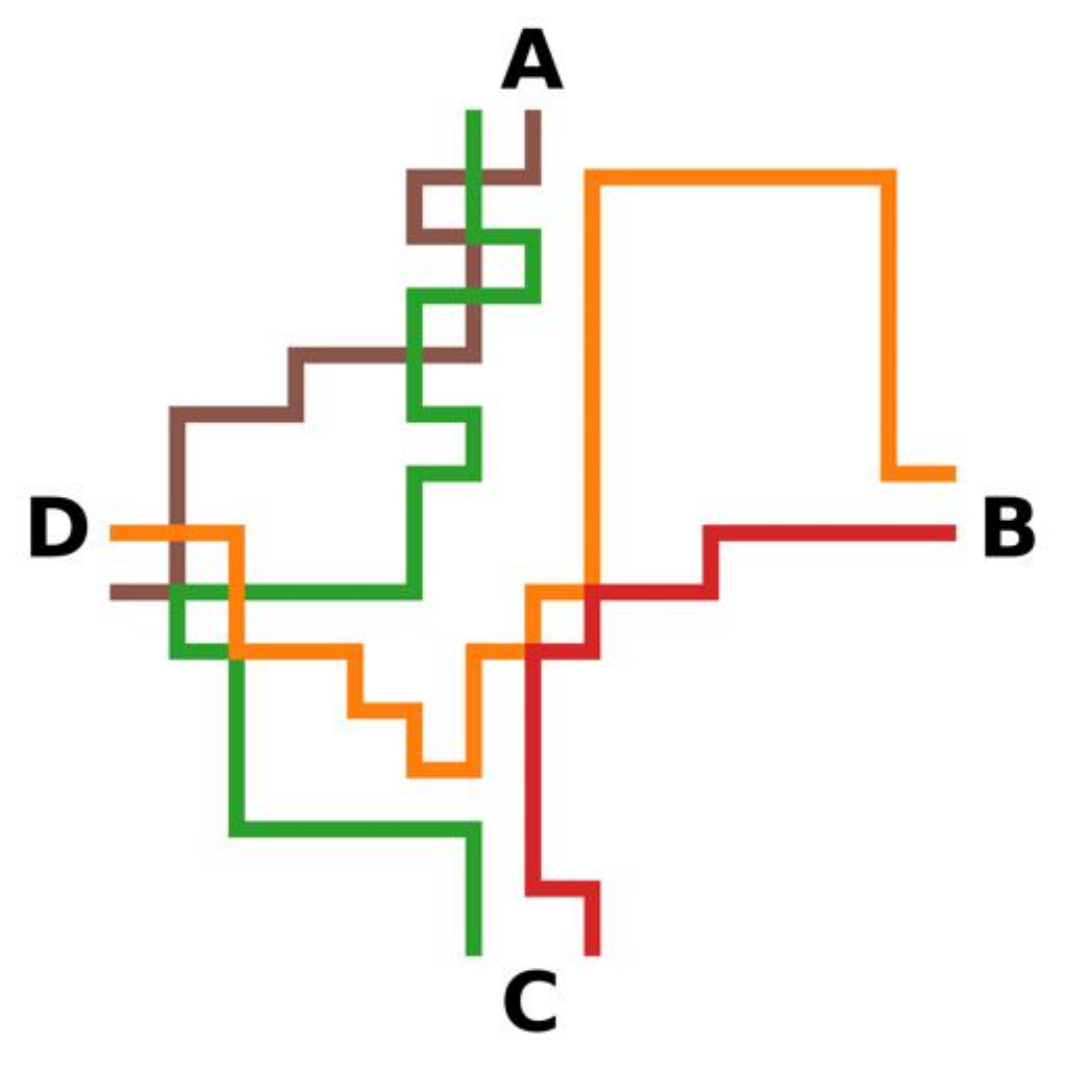}} &\multicolumn{2}{c}{\includegraphics[width=0.15\textwidth]{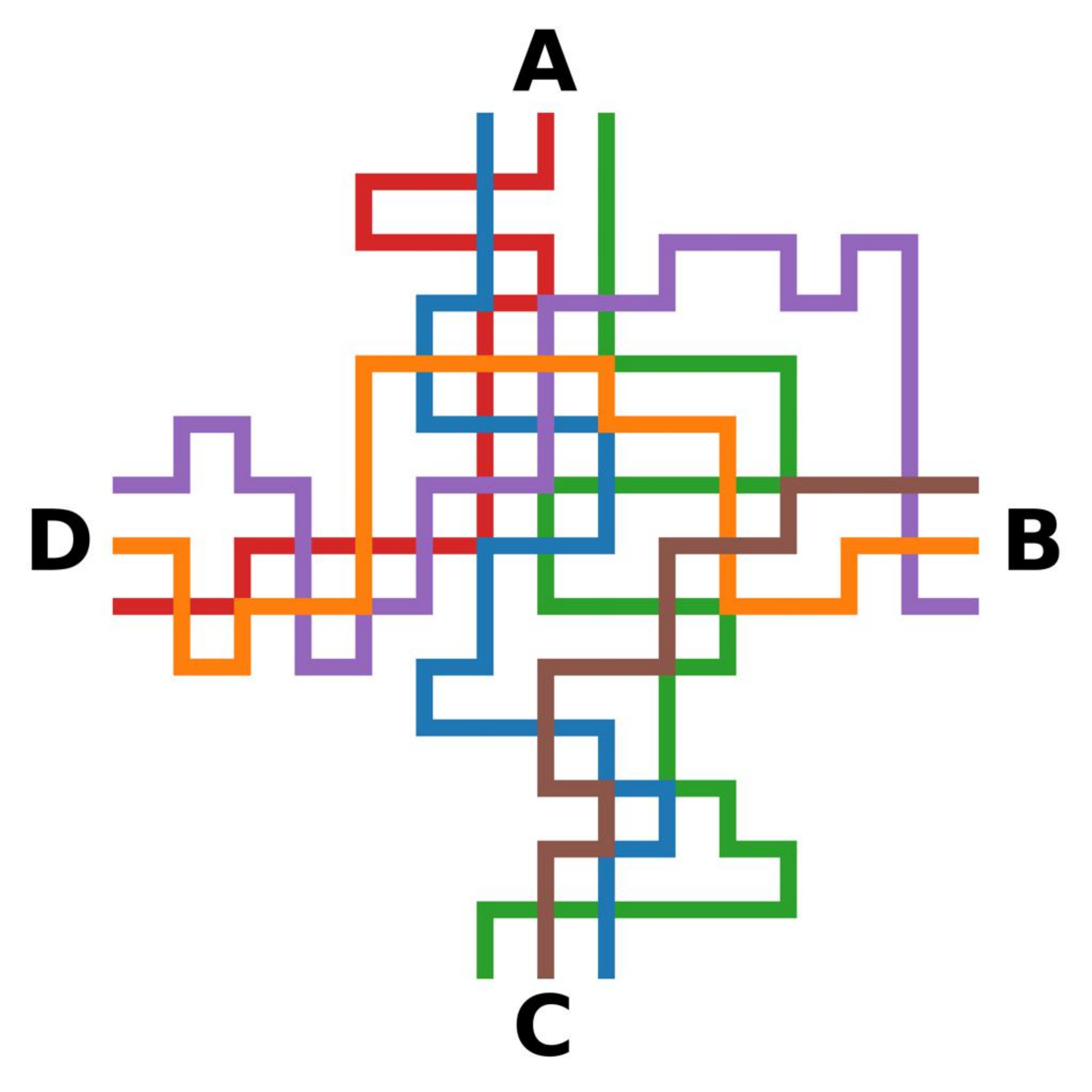}}&\multicolumn{2}{c}{\includegraphics[width=0.15\textwidth]{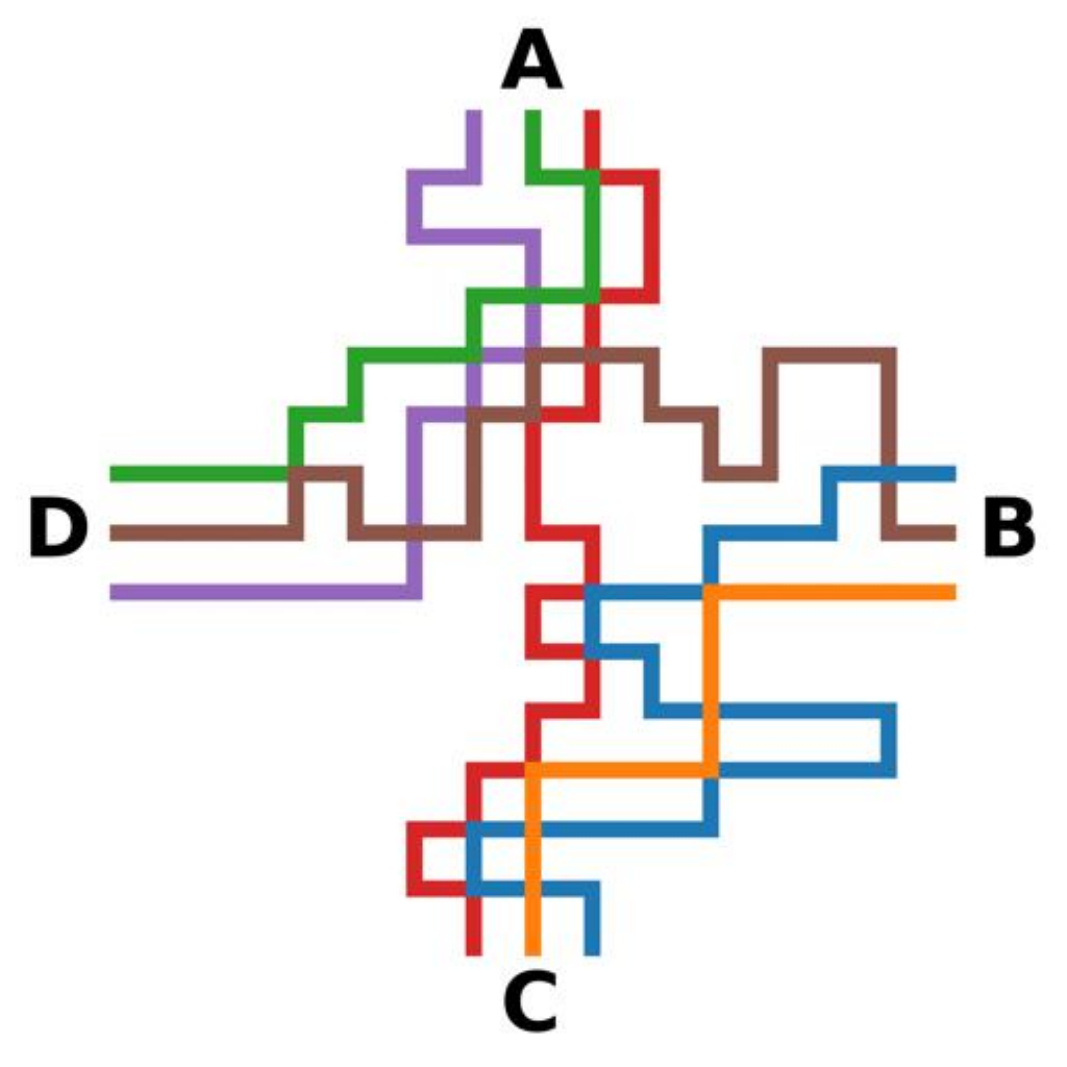}}\\
     GT & \centering 1 &  &  \centering 1 & &  \centering 1 &  &  \centering 1 &  &  \centering 2 & &  \centering 1 & \\
     \rowcolor{lightgray}
     \raisebox{-0.2\height}\gptlogo & \centering 2 & \textcolor{red}{\xmark} &  \centering 0 & \textcolor{red}{\xmark}&  \centering 2 & \textcolor{red}{\xmark} &  \centering 3 & \textcolor{red}{\xmark} &  \centering 3 & \textcolor{red}{\xmark}&  \centering 1 & \textcolor{ForestGreen}{\cmark}\\
     \raisebox{-0.2\height}\geminilogo & \centering 2 & \textcolor{red}{\xmark} & \centering 2 & \textcolor{red}{\xmark}&  \centering 4 & \textcolor{red}{\xmark} &  \centering 1 & \textcolor{ForestGreen}{\cmark} &  \centering 2 & \textcolor{ForestGreen}{\cmark}&  \centering 5 & \textcolor{red}{\xmark}\\
     \rowcolor{lightgray}
     \raisebox{-0.2\height}\sonnetlogo & \centering 2 & \textcolor{red}{\xmark} & \centering 2 & \textcolor{red}{\xmark}&  \centering 3 & \textcolor{red}{\xmark} &  \centering 2 & \textcolor{red}{\xmark} &  \centering 3 & \textcolor{red}{\xmark}&  \centering 3 & \textcolor{red}{\xmark}\\
     \raisebox{-0.2\height}\newsonnetlogo & \centering 1 & \textcolor{ForestGreen}{\cmark} & \centering 1 & \textcolor{ForestGreen}{\cmark}&  \centering 3 & \textcolor{red}{\xmark} &  \centering 3 & \textcolor{red}{\xmark} &  \centering 2 & \textcolor{ForestGreen}{\cmark}&  \centering 1 & \textcolor{ForestGreen}{\cmark}\\
     \end{tabular}

    \vspace{4pt}
    \centering
    \begin{tabular}{cccccccccccccc}
    \raisebox{-0.1\height}\gptlogo & \gpt &&&
      \raisebox{-0.1\height}\geminilogo & \gemini & &&\raisebox{-0.1\height}\sonnetlogo & \sonnet & && \raisebox{-0.12\height}\newsonnetlogo & \newsonnet \\
      \end{tabular}
     
\end{AIbox}
\caption{Some VLMs (\gptlogo, \geminilogo, \sonnetlogo) surprisingly fail in even extremely easy cases across both line widths (leftmost).
VLMs tend to perform worse as the number of paths connecting stations increases.}
\label{fig:qual_subway3}
\end{figure}

\clearpage
\section{VLM failures on real-world data}
\label{sec:real_world_failures}

\begin{figure}
\centering

\begin{minipage}{0.9\textwidth}
    \begin{AIbox}{List out stations on a subway path.}
    \hspace{0.2\textwidth}\includegraphics[width=0.6\textwidth]{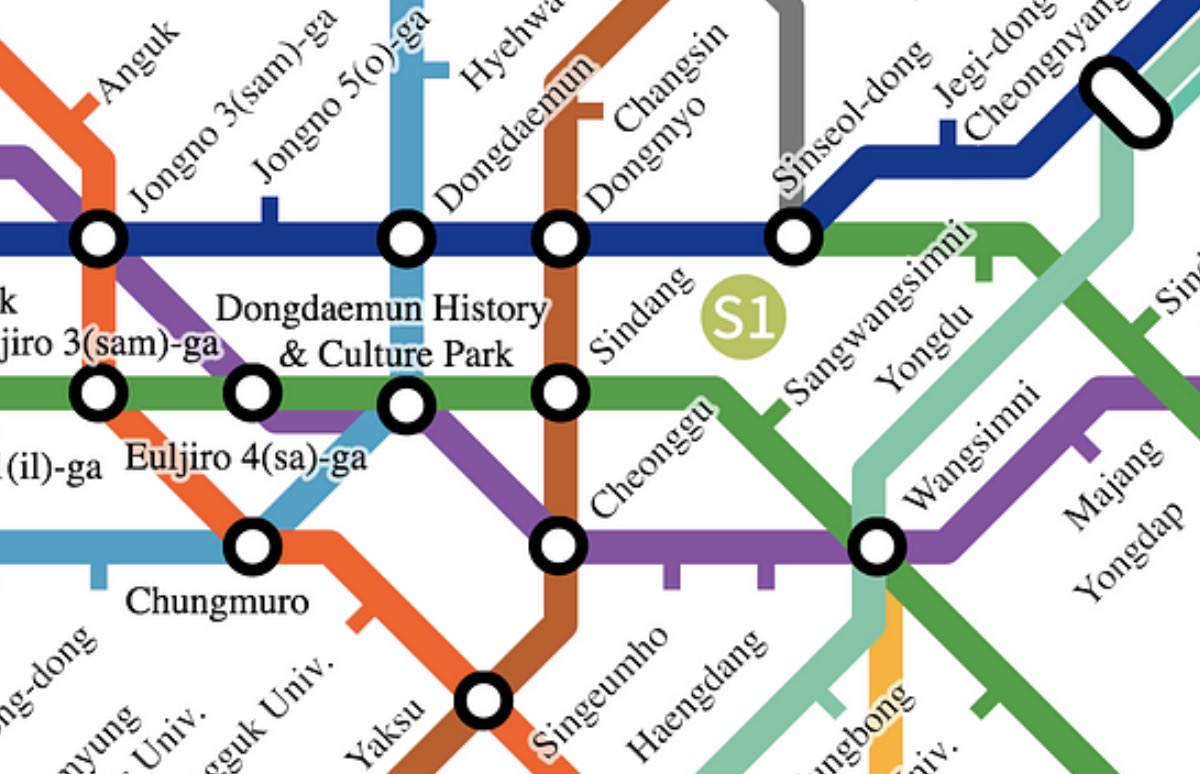}
    \\
    \footnotesize
    \textbf{\textcolor{black}{User}}:  This is a subway map. List out the stations between Chungmuro and Wangsimni.\\

    \gemini: Here are the stations between Chungmuro and Wangsimni on this map, assuming you're traveling along the dark blue line:
    \begin{itemize}
        \item Dongdaemun History \& Culture Park \redxmark
        \item Sinseol-dong \redxmark
        \item Jegi-dong \redxmark
    \end{itemize}

    \newsonnet: 
    Based on the subway map shown, the stations between Chungmuro and Wangsimni are:
    \begin{enumerate}
        \item Chungmuro
        \item Dongdaemun History \& Culture Park
        \item Sindang
        \item Cheonggu
        \item \textcolor{red}{Sangwangsimni}  \redxmark
        \item Wangsimni
    \end{enumerate}

    \end{AIbox}   
\end{minipage}

\caption{Examples of how VLMs fail on real-world data.}
\label{fig:real_task_failures_subway}
\end{figure}


\begin{figure}[t]
\centering
\begin{minipage}{0.48\textwidth}
    \begin{AIbox}{Reading music sheets}
    \includegraphics[width=1.0\textwidth]{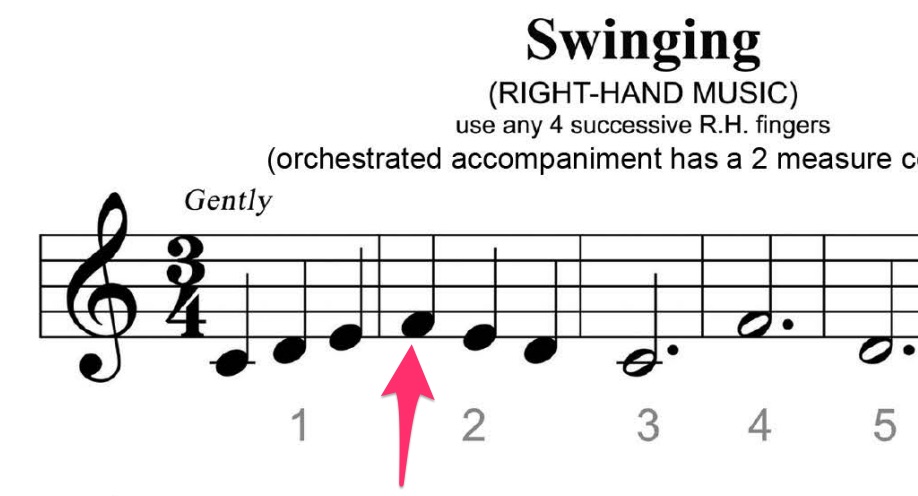}
    \\
    \footnotesize
    \textbf{\textcolor{black}{User}}: What is the name of this note?

    \newsonnet: The note indicated by the pink arrow is F. \greentick This can be determined by its position on the musical staff - it sits on the top line of the treble clef, which corresponds to the note F.\\

    \textbf{\textcolor{black}{User}}: Is the note on a line?

    \newsonnet: Yes, \textcolor{red}{this note is on a line}. \redxmark  Specifically, it's on the third line from the bottom of the treble clef staff. \redxmark

    \end{AIbox}   
\end{minipage}
\begin{minipage}{0.48\textwidth}
    \begin{AIbox}{Reading NYC street map}
    \includegraphics[width=1.0\textwidth]{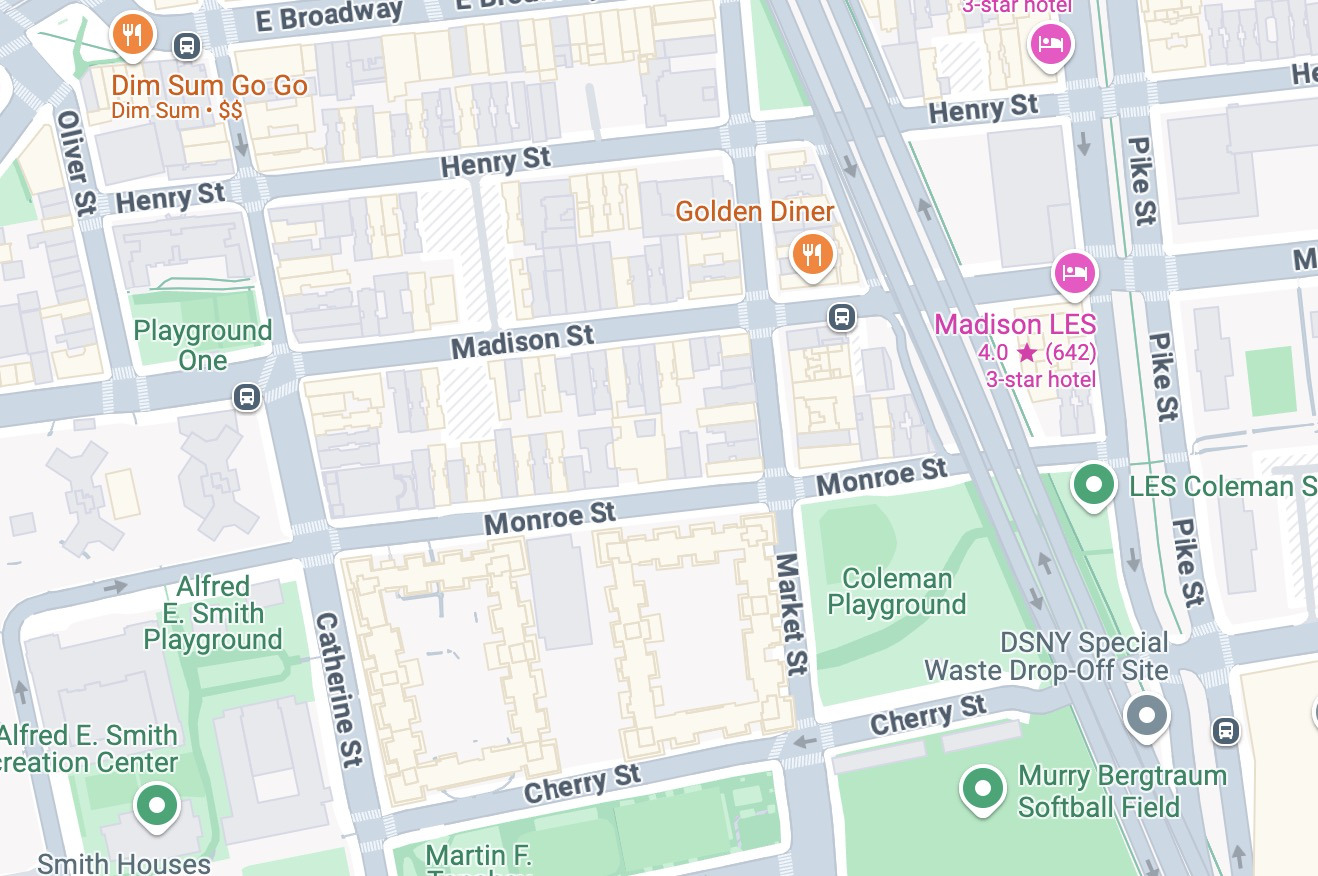}
    \\
    \footnotesize
    \textbf{\textcolor{black}{User}}:  Is Catherine St intersecting with Market St?\\

    \gpt: Yes, Catherine St \textcolor{red}{does intersect} with Market St as shown in the map \redxmark. They intersect near the Alfred E. Smith Playground and close to the Monroe St and Cherry St intersections. \redxmark
    \\\\
    \gemini: No, Catherine St and Market St do not intersect in this map. \greentick

    \end{AIbox}   
\end{minipage}

\caption{Examples of how VLMs fail on real-world data.}
\label{fig:real_task_failures}
\end{figure}

\clearpage
\section{Linear Probing Experiment for \twocirclelogo and \lineintersectlogo tasks}

\subsection{Implementation details}
\label{appsec:implementation_linear}
We use identical hyperparameters for all probing experiments and provide their values in \cref{apptab:ex_lin_prob_details}. We also provide the feature dimensions of each model in \cref{apptab:feat_dim}.

\begin{table}[]
    \centering
    \caption{The list of hyperparameters for training the linear classifier on frozen features of \philogo~and \llavalogosmall~in scikit-learn.}
    \label{apptab:ex_lin_prob_details}
    \begin{tabular}{c|c}
         \toprule
         Config &  Value\\
         \midrule
         Model & Logistic Regression\\
         Regularization & L2\\
         Regularization Weight & 1.0\\
         Random seed & 42\\
         Training epochs & 1000\\
         \bottomrule
    \end{tabular}
\end{table}

\begin{table}[]
    \centering
    \caption{\llavaSI generates features with various dimensions for each input size. However, after the average pooling the frozen features for different image sizes have identical dimensions.}
    \label{apptab:feat_dim}
    \begin{tabular}{ccccc}
    \toprule
         Model & Stage & input dim. & output dim. & after pooling\\
         \midrule
         \multirow{6}{*}{\llavalogosmall}& \multirow{3}{*}{vision encoder} & 384$\times$385 & [3, 729, 1152] & \multirow{3}{*}{[1, 1152]}\\
         & &769$\times$770 & [10, 729, 1152] & \\
         &  & 1155$\times$1155 & [17, 729, 1152] & \\
         \cmidrule(lr){2-5}
         & \multirow{3}{*}{projection layer} & 384$\times$385 & [3, 729, 896] & \multirow{3}{*}{[1, 896]}\\
         & &769$\times$770 & [10, 729, 896] &\\
         &  & 1155$\times$1155 & [17, 729, 896] & \\
         \midrule
         \multirow{6}{*}{\philogo} & \multirow{3}{*}{vision encoder} & 384$\times$384 & \multirow{3}{*}{[1, 5, 576, 1024]} & \multirow{3}{*}{[1, 1024]}\\
         & &768$\times$768 &\\
         &  & 1152$\times$1152 & \\
         \cmidrule(lr){2-5}
         & \multirow{3}{*}{projection layer} & 384$\times$384 & \multirow{3}{*}{[757, 3072]} & \multirow{3}{*}{[1, 3072]}\\
         & &768$\times$768 &\\
         &  & 1152$\times$1152 &\\
         \bottomrule
    \end{tabular}
\end{table}

\clearpage
\subsection{Additional results}
\label{app:probe-results_splits}

\begin{figure}
    \centering
    \begin{tabular}{ccc}
        &vision encoder& projection layer\\
         \raisebox{2cm}{\philogo}& \includegraphics[trim={0 120 0 0},clip,width=0.45\textwidth]{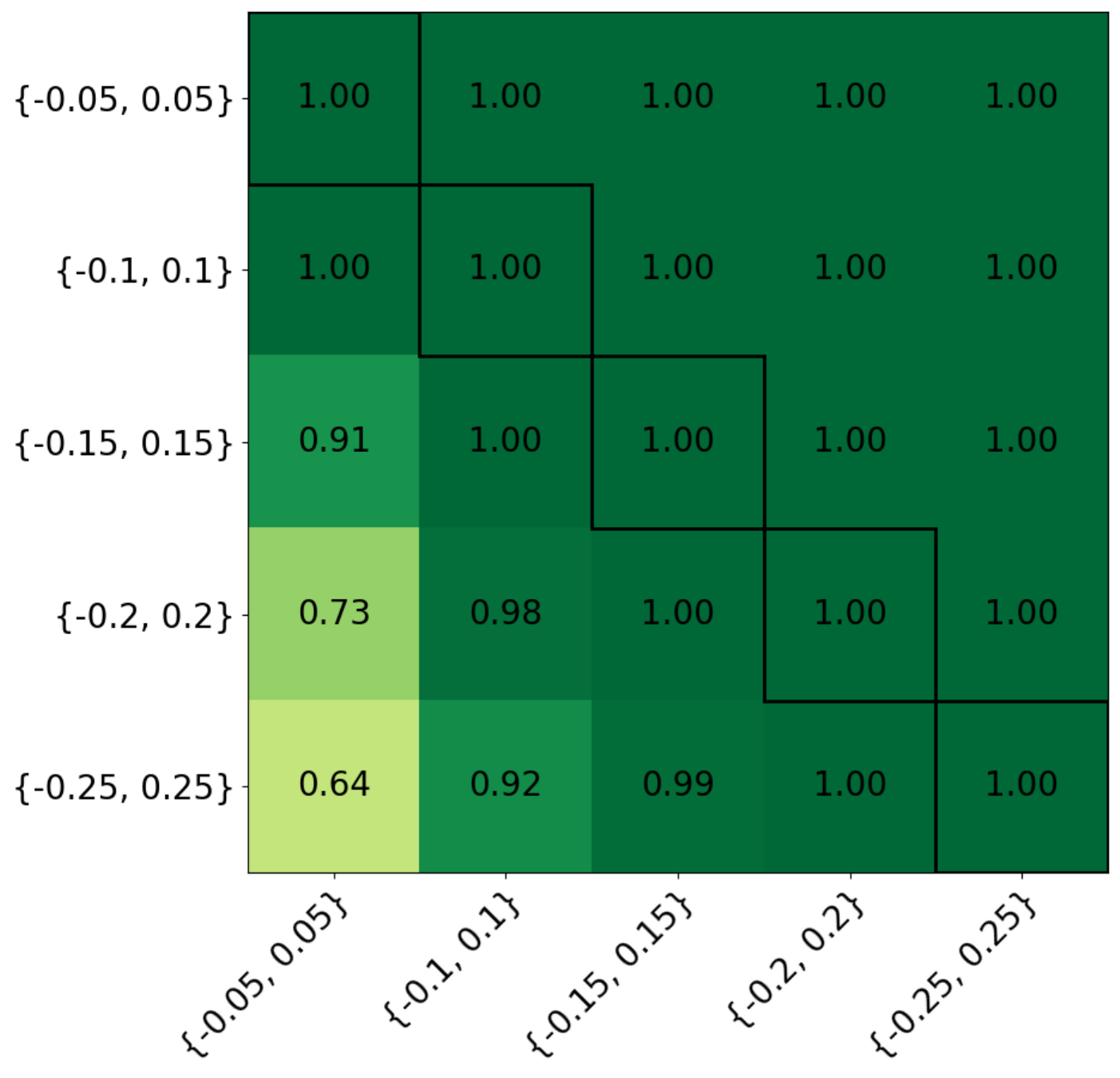}& \includegraphics[trim={0 120 0 0},clip,width=0.45\textwidth]{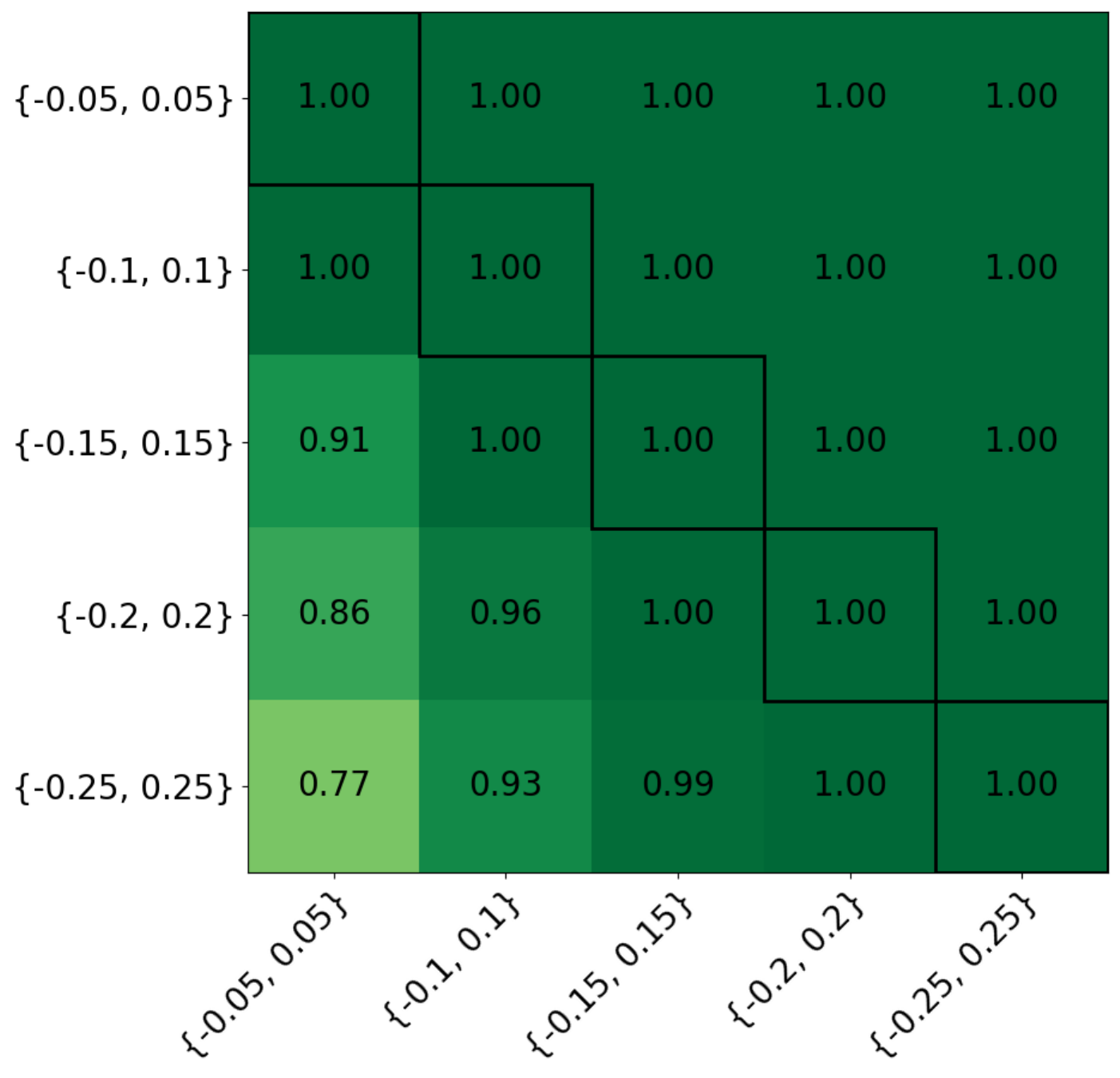} \\
         \midrule
         \raisebox{3cm}{\llavalogosmall}& \includegraphics[width=0.45\textwidth]{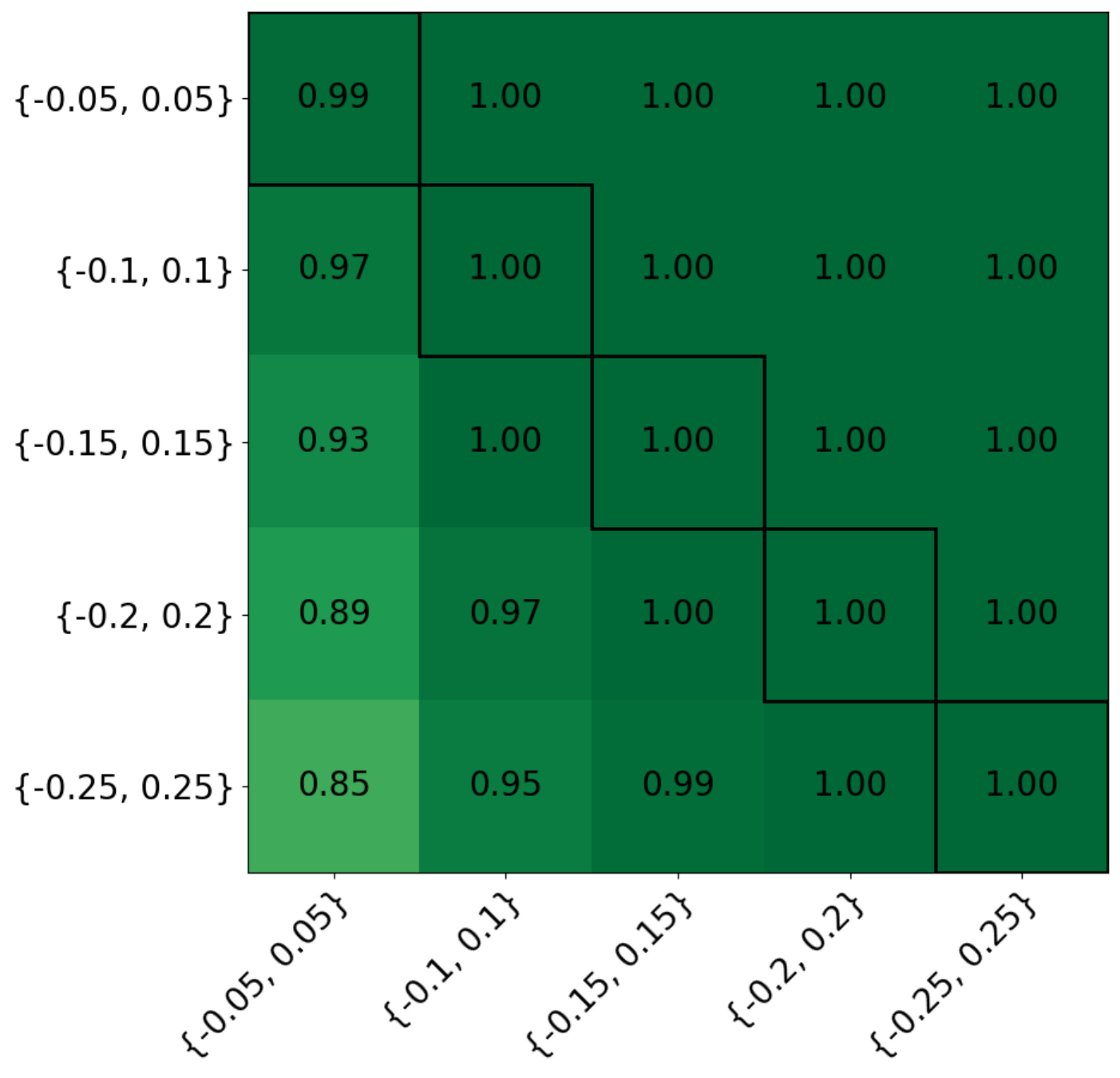}& \includegraphics[width=0.45\textwidth]{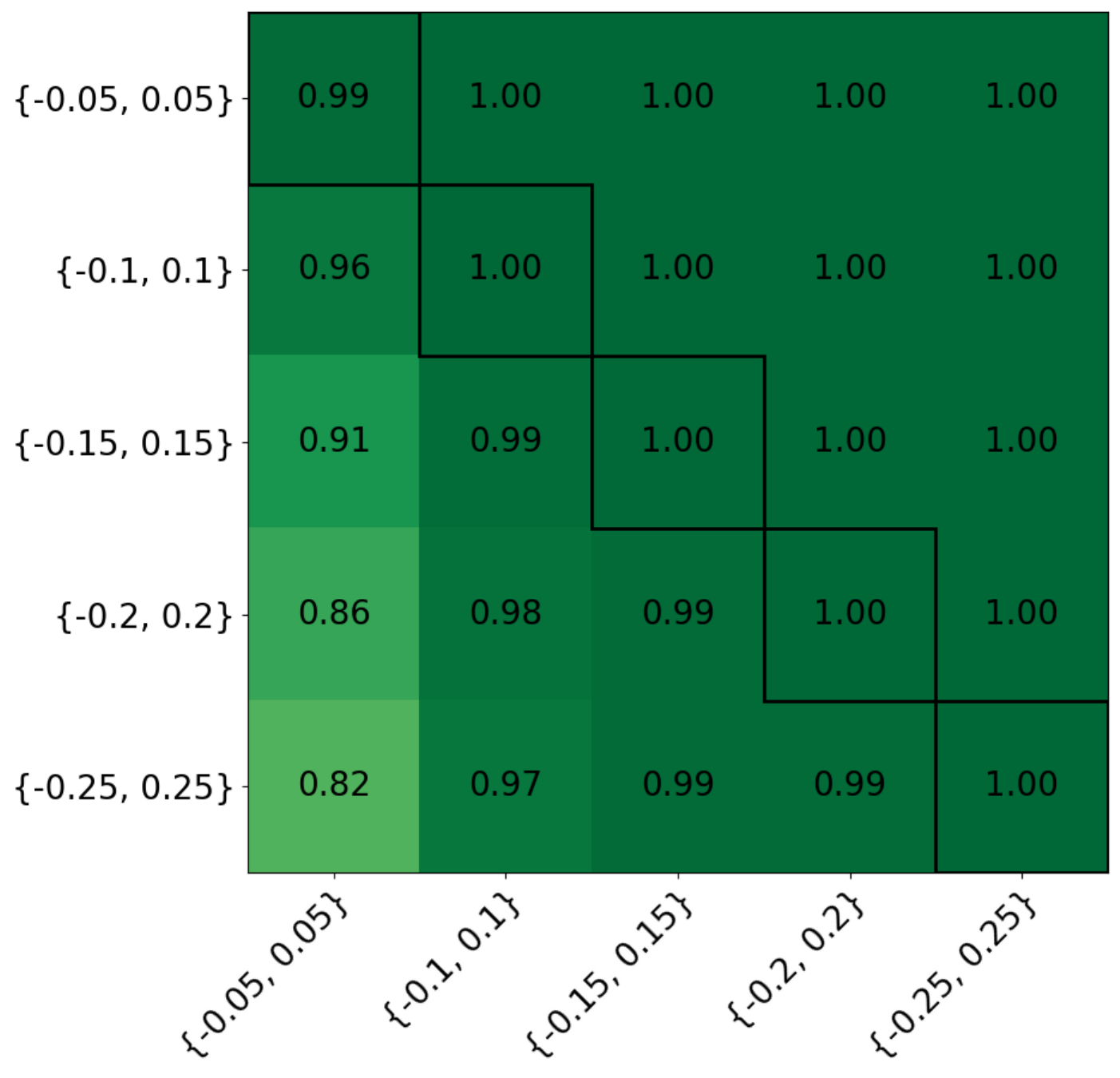} \\
    \end{tabular}
    \begin{flushleft}
    Distance between circles:\\
    $y$-axis \text{$\uparrow$}~~~training sets.\\
    $x$-axis \text{$\rightarrow$} test sets
    \end{flushleft}
    \caption{The linear classifier trained on frozen features of images, where objects (\twocirclelogo) are close ($distance\in\{-0.05, 0.05\}$), performs well (100\%) on images that contain objects that are far apart ($distance\in\{-0.25, 0.25\}$), for both models and stages. Conversely, the classifier performs worse (at least 18\%), when transferred from large to small distances.}
    \label{appfig:linprobe-circles}
\end{figure}

\begin{figure}
    \centering
    \begin{tabular}{ccc}
        &vision encoder& projection layer\\
         \raisebox{2cm}{\philogo}& \includegraphics[trim={0 68 0 0},clip,width=0.45\textwidth]{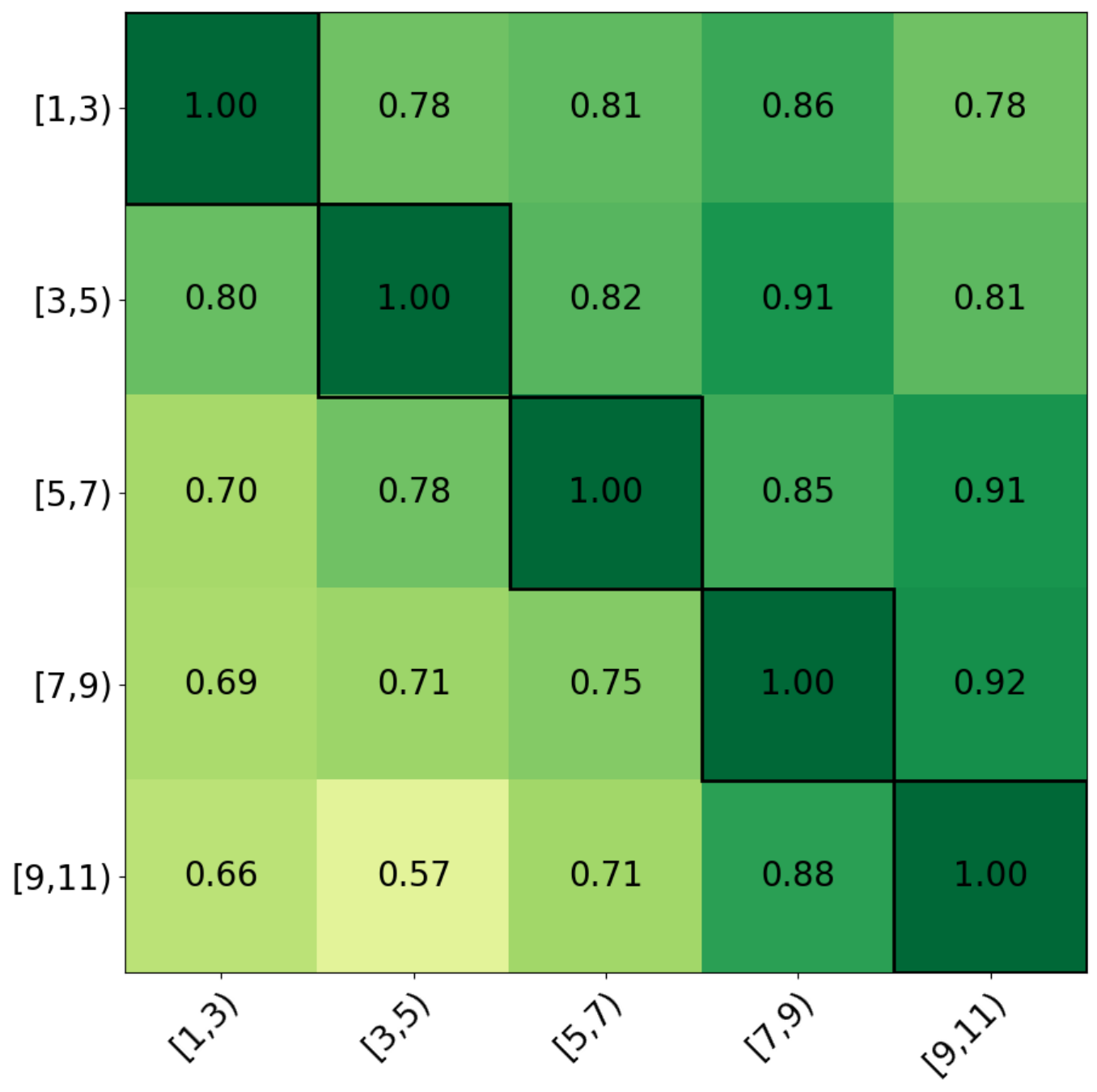}& \includegraphics[trim={0 68 0 0},clip,width=0.45\textwidth]{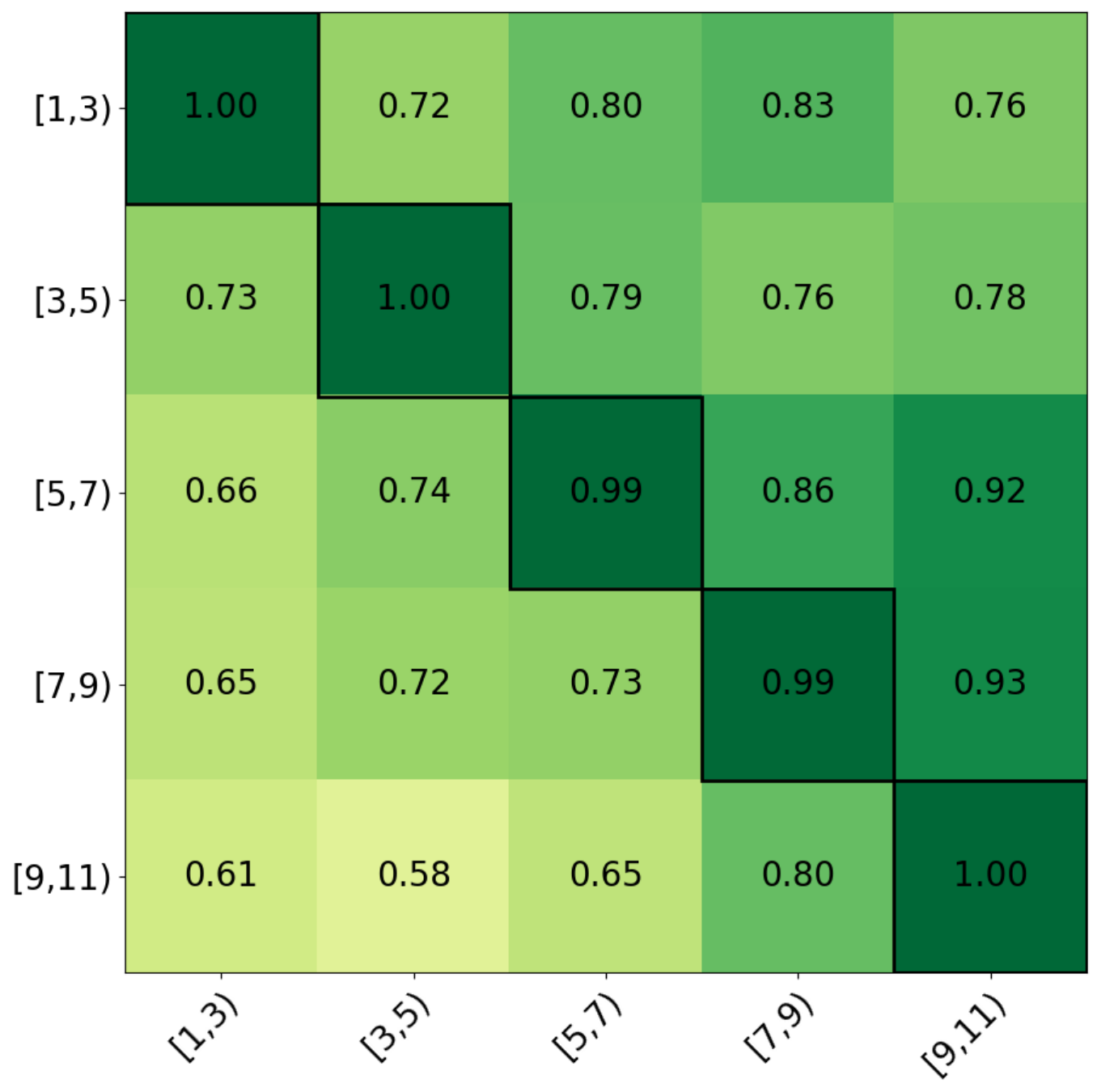} \\
         \midrule
         \raisebox{3cm}{\llavalogosmall}& \includegraphics[width=0.45\textwidth]{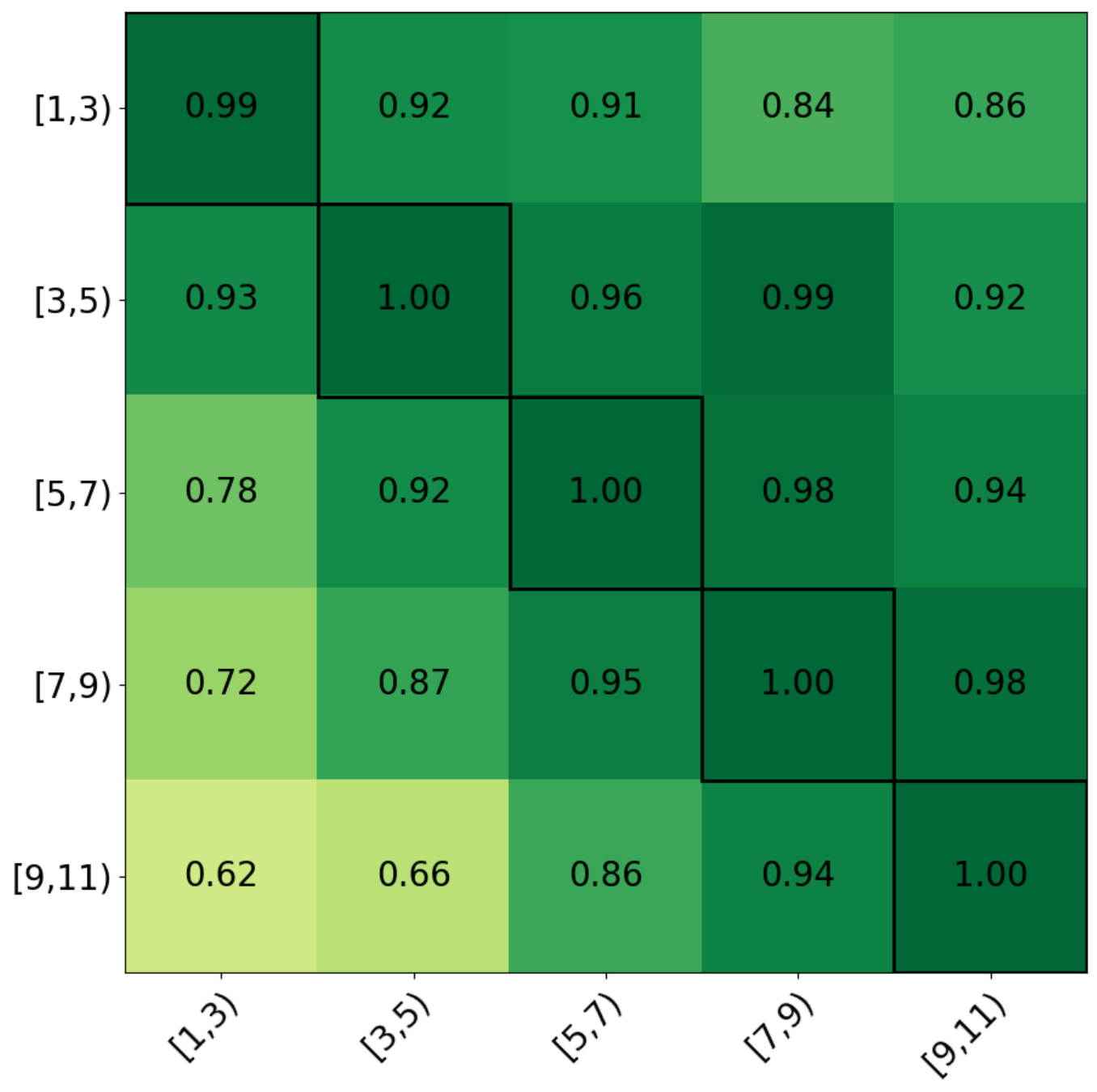}& \includegraphics[width=0.45\textwidth]{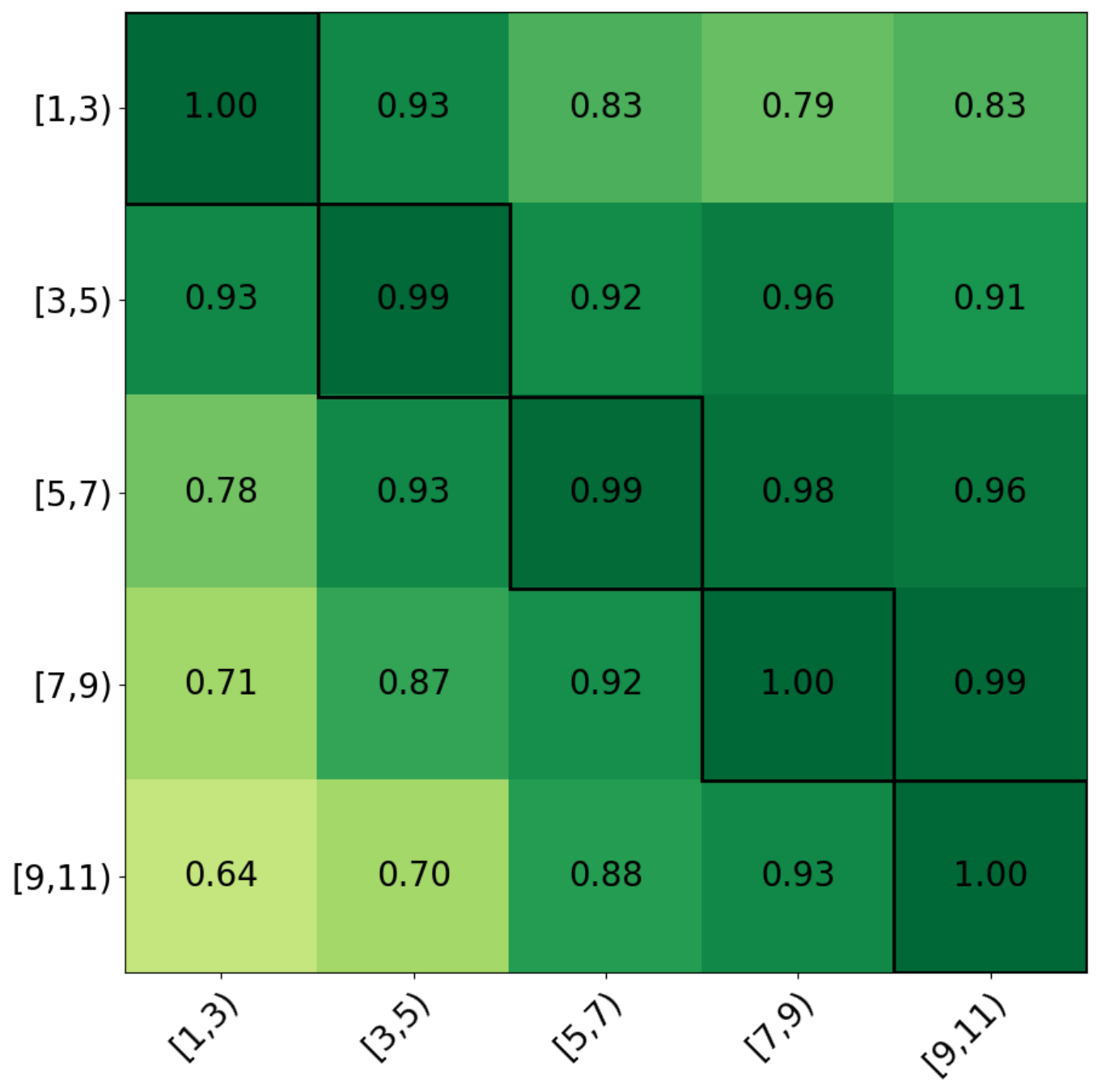} \\
    \end{tabular}
    \begin{flushleft}
    Mean distance brackets:
    \text{$\uparrow$ training sets}
    \text{$\rightarrow$ testing sets}
    \end{flushleft}
    \caption{The frozen features of both the projection layer and the VE output contain sufficient information to count the intersections between the line charts \lineintersectlogo. Moreover, the mean distance between the y coordinates of line charts controls the visual complexity of the images, where transferring the trained classifier from small ($[1, 3)$) to large ($[9, 11)$) distances, compared to transfer from large to small distances, has lower impacts on the classification performance (24\% vs. 39\%; the first and last row in the top-right heatmap).}
    \label{appfig:linprobe-lines}
\end{figure}

The linear classifier, \ie, logistic regression, performs with 99\% accuracy on the two circles (\cref{appfig:linprobe-circles}) and the line charts tasks (\cref{appfig:linprobe-lines}), when trained and tested on the images with the same visual complexities (\cref{appfig:linprobe-circles,appfig:linprobe-lines}; diagonal axis highlighted with black outline).

Moreover, transferring the classifier trained in small to large gaps shows an accuracy comparable to the baseline accuracy of these VLMs, suggesting the sufficiency of frozen features for solving the selected tasks from \blindtest.

\end{document}